\documentclass[12pt, a4paper, hidelinks]{book}

\usepackage[utf8]{inputenc}

\usepackage{etex}
\usepackage[a4paper,left=2.5cm,top=3.0cm,centering,bindingoffset=.8cm]{geometry}	
\usepackage{disspage}					
\usepackage{xspace}
\usepackage{url}
\usepackage{makeidx}
\usepackage{minitoc}
\usepackage{fancyhdr}
\usepackage[capitalise]{cleveref}
\usepackage{amsthm}
\usepackage{framed}
\usepackage{pdflscape}
\usepackage{afterpage}
\usepackage{setspace}

\usepackage[T1]{fontenc}
\usepackage{type1ec}
\usepackage[scaled=0.92]{helvet}
\usepackage{mathpazo}
\usepackage[rm,bf]{titlesec}
\usepackage{titletoc}		
\usepackage[english]{twolang}
\usepackage{textcomp}
\usepackage{pifont}
\newcommand{\cmark}{\ding{51}}%
\newcommand{\xmark}{\ding{55}}%

\usepackage{amsmath}
\usepackage{amssymb}
\usepackage{mathdots}
\usepackage{mathtools}
\usepackage{bm}

\usepackage{etoolbox}
\robustify
\robustify\addtocounter
\robustify
\robustify

\usepackage{graphicx}
\usepackage{adjustbox}
\usepackage{subfig}
\usepackage{pgfplots}
\usepackage{tikz,pgfplots}
\usepackage{tikz-3dplot}
\usetikzlibrary{decorations.pathreplacing}
\usetikzlibrary{calc,3d}
\usetikzlibrary{positioning,shapes,decorations,shadows,arrows,plotmarks,fit}
\usetikzlibrary{spy}

\usepgfplotslibrary{external}
\usepackage{ulem}
\usepackage[font=small,labelfont=bf]{caption}
\usepackage{pstricks}

\usepackage{array}
\usepackage{multirow}
\usepackage{booktabs}

\usepackage{algorithmicx}
\usepackage[chapter]{algorithm}
\usepackage[end]{algpseudocode}
\ifpdf
\usepackage{hyperref}	
\pdfpagewidth=\the\paperwidth
\else
\usepackage{hyperref}	
\fi

\usepackage[toc,section=chapter,indexonlyfirst]{glossaries}
\makeglossaries

\newglossarystyle{dottedlocations}{%
  \glossarystyle{list}%
  \renewcommand*{\glossaryentryfield}[5]{%
    \item[\glsentryitem{##1}\glstarget{##1}{##2}] \emph{##3}%
      \unskip\leaders\hbox to 2.9mm{\hss.}\hfill##5}%
}
\glossarystyle{dottedlocations}

\usepackage{marginnote}

\usepackage{thmtools}
\usepackage{thm-restate}

\newacronym{dct}{DCT}{discrete cosine transform}
\newacronym{ml}{ML}{maximum likelihood}
\newacronym{map}{MAP}{maximum a-posteriori}
\newacronym{pocs}{POCS}{projection onto convex sets}
\newacronym{ccd}{CCD}{charge-coupled device}
\newacronym{cmos}{CMOS}{complementary metaloxide semiconductor}
\newacronym{psf}{PSF}{point spread function}
\newacronym{snr}{SNR}{signal-to-noise ratio}
\newacronym{cft}{CFT}{continuous Fourier transform}
\newacronym{dft}{DFT}{discrete Fourier transform}
\newacronym{fft}{FFT}{Fast Fourier Transform}
\newacronym{lsi}{LSI}{linear shift invariant}
\newacronym{cfa}{CFA}{color-filter array}
\newacronym{pdf}{PDF}{probability density function}
\newacronym{mggd}{MGGD}{multivariate generalized Gaussian distribution}
\newacronym{cg}{CG}{conjugate gradient}
\newacronym{scg}{SCG}{scaled conjugate gradient}
\newacronym{tv}{TV}{total variation}
\newacronym{btv}{BTV}{bilateral total variation}
\newacronym{wbtv}{WBTV}{weighted bilateral total variation}
\newacronym{em}{EM}{expectation maximization}
\newacronym{mm}{MM}{majorization-minimization}
\newacronym{mad}{MAD}{median absolute deviation}
\newacronym{gcv}{GCV}{generalized cross validation}
\newacronym{psnr}{PSNR}{peak-signal-to-noise ratio}
\newacronym{ssim}{SSIM}{structural similarity}
\newacronym{db}{dB}{decibel}
\newacronym{ecc}{ECC}{enhanced correlation coefficient}
\newacronym{tof}{ToF}{Time-of-Flight}
\newacronym{pet}{PET}{positron emission tomography}
\newacronym{ct}{CT}{computed tomography}
\newacronym{mri}{MRI}{magnetic resonance imaging}
\newacronym{lmi}{LMI}{local mutual information}
\newacronym{ncc}{NCC}{normalized cross correlation}
\newacronym{gmm}{GMM}{Gaussian mixture model}
\newacronym{msac}{MSAC}{M-estimator sample consensus}
\newacronym{ritk}{RITK}{Range Imaging Toolkit}
\newacronym{svd}{SVD}{singular value decomposition}
\newacronym{cpbd}{CPBD}{cumulative probability of blur detection}
\newacronym{roc}{ROC}{receiver operator characteristic}
\newacronym{fov}{FOV}{field of view}
\newacronym{llr}{LLR}{locally linear regression}
\newacronym{nmad}{NMAD}{normalized mean absolute deviation}
\newacronym{pca}{PCA}{principal component analysis}
\newacronym[longplural={picture elements}]{pixel}{pixel}{picture element}
\newacronym{oct}{OCT}{optical coherence tomography}
\newacronym{irls}{IRLS}{iteratively re-weighted least squares}
\newacronym{idp}{IDP}{inter-color dependency penalty}

\newacronym{ssr}{SSR}{single-sensor super-resolution}
\newacronym{msr}{MSR}{multi-sensor super-resolution}
\newacronym{amsr}{AMSR}{adaptive multi-sensor super-resolution}
\newacronym{amsrod}{AMSR-OD}{adaptive multi-sensor super-resolution with outlier detection}

\def\titlefamily{\sffamily} 

\titlecontents{chapter}[0pc]
{\addvspace{1pc} \color{Blue} \normalsize \bfseries \titlefamily \filright}
{{\chaptername} \thecontentslabel \hspace{1pc}}
{}
{\hfill\contentspage}
[\addvspace{2pt}]

\titlecontents{section}[0pc]
{\addvspace{0pt} \normalsize \titlefamily \filright}
{\thecontentslabel\enskip}
{}
{\titlerule*[0.6em]{.}\contentspage}
[\addvspace{0pt}]

\titlecontents{subsection}[2pc]
{\addvspace{0pt} \normalsize \titlefamily \filright}
{\thecontentslabel\enskip}
{}
{\titlerule*[0.6em]{.}\contentspage}
[\addvspace{0pt}]

\titlecontents{subsubsection}[3pc]
{\addvspace{0pt} \normalsize \titlefamily \filright}
{\thecontentslabel\enskip}
{}
{\titlerule*[0.6em]{.}\contentspage}
[\addvspace{0pt}]

\titleformat{\chapter}[display]
{\color{Blue}\titlefamily\Large}
{\vspace{0pc}\filleft\MakeUppercase{c h a p t e r}\hspace{1pc}{\Huge\thechapter}}
{3ex}
{\fontsize{32}{32}\selectfont\bfseries\filleft}
[\vspace{-1ex}]

\titleformat{\section}
{\titlefamily\Large\bfseries}{\thesection}{1em}{}
\titleformat{\subsection}
{\titlefamily\large\bfseries}{\thesubsection}{1em}{}
\titleformat{\subsubsection}
{\titlefamily\normalsize\bfseries}{\thesubsubsection}{1em}{}

\titlespacing{\chapter}{-0.25pc}{0pc}{4pc}

\nomtcrule
\def\myminitoc{
\vspace*{-1.0pc}
\hfill
\begin{minipage}{0.95\textwidth}
\minitoc
\end{minipage}
\vspace{1.0pc}
}

\fancypagestyle{plain}{
    \fancyhead{}

    \cfoot{\titlefamily \thepage}
}

\def\titlefamily{\rmfamily}

\titlecontents{chapter}[0pc]
	{\addvspace{1pc} \color{Blue} \normalsize \bfseries \sffamily \filright}
	{{\chaptername} \thecontentslabel \hspace{1pc}}
	{}
	{\hfill\contentspage}
	[\addvspace{2pt}]
	
\titlecontents{section}[0pc]
	{\addvspace{0pt} \normalsize \titlefamily \filright}
	{\thecontentslabel\enskip}
	{}
	{\titlerule*[0.6em]{.}\contentspage}
	[\addvspace{0pt}]
	
\titlecontents{subsection}[2pc]
	{\addvspace{0pt} \normalsize \titlefamily \filright}
	{\thecontentslabel\enskip}
	{}
	{\titlerule*[0.6em]{.}\contentspage}
	[\addvspace{0pt}]

\titleformat{\chapter}[display]
{\color{Blue}\sffamily\Large}
{\vspace{0pc}\filleft\MakeUppercase{c h a p t e r}\hspace{1pc}{\Huge\thechapter}}
{3ex}
{\fontsize{32}{32}\selectfont\bfseries\filleft}
[\vspace{-1ex}]

\titlespacing{\chapter}{-0.25pc}{0pc}{4pc}

\usetikzlibrary{positioning,shapes,decorations,shadows,arrows,plotmarks}
\pgfplotsset{compat=newest}

\nomtcrule
\def\myminitoc{
\vspace*{-1.0pc}
\hfill
\begin{minipage}{0.93\textwidth}
\minitoc
\end{minipage}
}

\definecolor{faublue}{RGB}{0,51,102}

\hypersetup{
  colorlinks 					= true, 
  linkcolor 					= faublue,
  citecolor 					= cyan,
	plainpages 					= false,
  hypertexnames 			= true,
  linktocpage 				= true,
  bookmarksopen 			= true,
  bookmarksnumbered 	= true,
  bookmarksopenlevel	= 0,
  pdftitle 						= {Multi-Frame Super-Resolution Reconstruction with Applications to Medical Imaging},
  pdfauthor 					= {Thomas K\"ohler},
  pdfsubject 					= {Dissertation Friedrich-Alexander-Universit\"at Erlangen-N\"urnberg},
  pdfkeywords 				= {Super-resolution, multi-frame algorithms, medical imaging},
	pdfproducer         = {}
}

\input{hyphenation.own}

\input{commands.own}

\makeindex

\begin{document}

\ifx\compilechapter\undefined
  \frontmatter

  \pagestyle{empty}
    

%
\pdfbookmark[1]{Cover}{cov}
\begin{titlepage}
\centering
\huge
\vspace*{\fill}
\Huge
Multi-Frame Super-Resolution Reconstruction with Applications to Medical Imaging\\
\vspace{1cm}
\Huge
Bildfolgenbasierte Verfahren zur Aufl\"osungserh\"ohung mit Anwendungen in der Medizinischen Bildgebung 
\vspace{2cm}\\
\Large
Der Technischen Fakult\"at der\\
Friedrich-Alexander-Universit\"at Erlangen-N\"urnberg \vspace{1.0cm}\\
zur Erlangung des Grades \vspace{1.5cm}\\
\Huge 
Doktor-Ingenieur (Dr.-Ing.) \vspace{1.0cm}\\
\Large
vorgelegt von \vspace{1.8cm}\\
\Large
Thomas K\"ohler\vspace{0.25cm}

aus\vspace{0.25cm}

Bamberg, Deutschland
\vspace*{\fill}
\end{titlepage}

\newpage
{
\normalsize
\thispagestyle{empty}
\vspace*{\fill}
\begin{center}
Als Dissertation genehmigt von der \\
Technischen Fakult\"at\\
der Friedrich-Alexander-Universität Erlangen-Nürnberg\\
\begin{tabular}{p{8cm}p{7cm}}
\rule{4.8cm}{0pt} & \rule{6.72cm}{0pt} \\
Tag der mündlichen Prüfung: & 11.09.2017 \\
Vorsitzender des Promotionsorgans: & Prof. Dr.-Ing. Reinhard Lerch\\
Gutachter: 		& Prof. Dr.-Ing. Joachim Hornegger\\
							& Prof. Sina Farsiu, Ph.D.\\

\end{tabular}
\end{center}
\vspace*{1cm}
}
 \cleardoublepage

\begin{center}
  {\bf Abstract}
\end{center}
The optical resolution of a digital camera is one of its most crucial parameters with broad relevance for consumer electronics, surveillance systems, remote sensing, or medical imaging. However, resolution is physically limited by the optics and sensor characteristics. In addition, practical and economic reasons often stipulate the use of out-dated or low-cost hardware. Super-resolution is a class of retrospective techniques that aims at high-resolution imagery by means of software. Multi-frame algorithms approach this task by fusing multiple low-resolution frames to reconstruct high-resolution images. This work covers novel super-resolution methods along with new applications in medical imaging.

The first contribution of this thesis concerns computational methods to super-resolve image data of a single modality. The emphasis lies on motion-based algorithms that are derived from a Bayesian statistics perspective, where subpixel motion of low-resolution frames is exploited to reconstruct a high-resolution image. More specifically, we introduce a confidence-aware Bayesian observation model to account for outliers in the image formation, \eg invalid pixels. In addition, we propose an adaptive prior for sparse regularization to model natural images appropriately. We then develop a robust optimization algorithm for super-resolution using this model that features a fully automatic selection of latent hyperparameters. The proposed approach is capable of meeting the requirements regarding robustness of super-resolution in real-world systems including challenging conditions ranging from inaccurate motion estimation to space variant noise. For instance, in case of inaccurate motion estimation, the proposed method improves the \gls{psnr} by $0.7$\,\gls{db} over the state-of-the-art. 

The second contribution concerns super-resolution of multiple modalities in the area of hybrid imaging. We introduce novel multi-sensor super-resolution techniques and investigate two complementary problem statements. For super-resolution in the presence of a guidance modality, we introduce a reconstruction algorithm that exploits guidance data for motion estimation, feature driven adaptive regularization, and outlier detection to reliably super-resolve a second modality. For super-resolution in the absence of guidance data, we generalize this approach to a reconstruction algorithm that jointly super-resolves multiple modalities. These multi-sensor methodologies boost accuracy and robustness compared to their single-sensor counterparts. The proposed techniques are widely applicable for resolution enhancement in a variety of multi-sensor vision applications including color-, multispectral- and range imaging. For instance in color imaging as a classical application, joint super-resolution of color channels improves the \gls{psnr} by $1.5$\,\gls{db} compared to conventional channel-wise processing.  

The third contribution transfers super-resolution to workflows in healthcare. As one use case in ophthalmology, we address retinal video imaging to gain spatio-temporal measurements on the human eye background non-invasively. In order to enhance the diagnostic usability of current digital cameras, we introduce a framework to gain high-resolution retinal images from low-resolution video data by exploiting natural eye movements. This framework enhances the mean sensitivity of automatic blood vessel segmentation by $10$\,\% when using super-resolution for image preprocessing. As a second application in image-guided surgery, we investigate hybrid range imaging. To overcome resolution limitations of current range sensor technologies, we propose multi-sensor super-resolution based on domain-specific system calibrations and employ high-resolution color images to steer range super-resolution. In ex-vivo experiments for minimally invasive and open surgery procedures using \gls{tof} sensors, this technique improves the reliability of surface and depth discontinuity measurements compared to raw range data by more than $24$\,\% and $68$\,\%, respectively.  

\glsresetall

\newpage

\begin{center}
  {\bf Kurz\"{u}bersicht}
\end{center}
Die optische Auflösung einer Kamera ist eine ihrer wichtigsten Kenngrö{\ss}en mit hohem Stellenwert für Unterhaltungselektronik, Überwachungssysteme, Fern\-er\-kun\-dung oder medizinische Bildgebung. Jedoch ist die Auflösung durch Optik und Sensoren physikalisch beschränkt. Daneben bedingen praktische oder öko\-no\-mi\-sche Gründe den Einsatz veralteter oder preiswerter Hardware. Verfahren zur Auf\-lösungserhöhung sind eine Klasse retrospektiver Techniken mit dem Ziel hochauflösende Bild\-gebung softwarebasiert zu gewährleisten. Bild\-folgenbasierte Algorithmen ermöglichen dies durch Fusion mehrerer niedrigauf\-lös\-ender Bilder zur Rekonstruktion hoch\-auf\-lös\-ender Bilder. Diese Arbeit behandelt neuartige Me\-tho\-den zur Auflösungserhöhung, sowie neue Anwendungen für die medizinische Bildgebung.

Der erste Beitrag dieser Arbeit betrifft Verfahren zur Auflösungserhöhung von Bildern einer einzelnen Modalität. Den Schwerpunkt bilden bewegungsbasierte und mit Bayesscher Statistik hergeleitete Algorithmen, bei denen Subpixel-Ver\-schiebungen zwischen niedrig auflösenden Bildern zur Rekonstruktion eines hoch\-auflösenden Bildes ge\-nutzt werden. Konkret führen wir ein konfidenz\-ge\-wichtetes Be\-ob\-ach\-tungs\-modell zur Behandlung von Ausrei{\ss}ern, z.\,B. defekte Pixel, in der Bild\-auf\-nahme ein. Zusätzlich stellen wir eine neue adaptive Verteilungsfunktion für die Regularisierung zur adäquaten Modellierung natürlicher Bilder vor. Wir ent\-wi\-ckeln ferner einen robusten Optimierungsalgorithmus mit diesem Mo\-dell, der Hyperparameter vollautomatisch auswählt. Der vorgestellte Ansatz zur Auflösungserhöhung erfüllt in der Praxis Anforderungen hinsichtlich Robustheit, welche schwierige Rahmenbedingungen von ungenauer Be\-we\-gungs\-schät\-zung bis orts\-variantem Rauschen umfassen. Im beispielhaften Fall einer ungenauen Be\-we\-gungs\-schät\-zung verbessert die vorgeschlagene Methode das Spitzen-Signal-Rausch-Verhältnis (PSNR) um $0.7$\,\gls{db} gegenüber dem Stand der Technik.

Der zweite Beitrag betrifft Ansätze zur Auflösungserhöhung für mehrere Mo\-da\-litäten in der hybriden Bildgebung. Wir führen hierfür neue Mehrsensor-Ver\-fahren ein und untersuchen zwei gegensätzliche Problemstellungen. Für die Auf\-lösungserhöhung unter Verwendung einer Führungsmodalität stellen wir einen Algorithmus vor, der diese zur Bewegungsschätzung, merkmalsbasierten ad\-ap\-ti\-ven Regularisierung und Ausrei{\ss}erdetektion zur zuverlässigen Auflösungserhöhung einer zweiten Modalität einsetzt. Für den Fall, dass Führungsdaten fehlen, verallgemeinern wir diesen Ansatz zu einem Algorithmus, der mehrere Mo\-da\-li\-tä\-ten simultan verarbeitet. Diese Mehrsensor-Methodik steigert Ge\-nau\-ig\-keit und Robustheit gegenüber Einzelsensor-Ansätzen. Die neu eingeführten Techniken sind vielfältig für eine Auflösungserhöhung in zahlreichen Anwendungen von Mehr\-sensor-\-Bild\-geb\-ung einsetzbar, was Farb-, Multispektral- sowie Tiefen\-bild\-ge\-bung umfasst. Im Bereich der Farb\-bild\-gebung als Beispiel für ein klassisches Anwendungsfeld, verbessert die simultane Auf\-lösungs\-er\-höhung von Farbkanälen das PSNR um $1.5$\,\gls{db} gegen\-über einer konventionellen kanalweisen Verarbeitung.

Der dritte Beitrag überträgt Verfahren zur Auflösungserhöhung in die Medizin. Als Anwendung in der Ophthalmologie behandeln wir Videobildgebung zur nicht-invasiven, örtlich-zeitlichen Untersuchung der menschlichen Retina. Um den dia\-gnos\-ti\-schen Nutzen aktueller Digitalkameras zu ver\-bessern, stellen wir ein Verfahren zur Ge\-winnung hoch\-auf\-lösender Retinabilder aus niedrigauflösenden Videodaten unter Aus\-nutz\-ung natürlicher Augenbewegungen vor. Das Verfahren verbessert die mittlere Sensitivität einer automatischen Blutgefä{\ss}segmentierung um $10$\,\%, wenn eine Auflösungserhöhung zur Bildvorverarbeitung genutzt wird. Als eine weitere Anwendung in der bildgeführten Chirurgie untersuchen wir hybride Tiefenbildgebung. Um Auf\-lösungs\-be\-schränk\-ungen heutiger Tiefensensoren zu überwinden, führen wir an\-wen\-dungs\-spe\-zi\-fi\-sche Kalibrierverfahren ein und ver\-wenden hochauflösende Farbbilder für Mehrsensor-Auflösungserhöhung auf Tiefendaten. In ex-vivo Experimenten für mi\-ni\-mal-in\-va\-si\-ve und offene Chirurgie mit \gls{tof} Sensoren verbessert diese Technik die Zu\-ver\-lässig\-keit von Ober\-flächen- und Tiefenkantenmess\-ungen um mehr als $24$\,\% bzw. $68$\,\% gegen\-über Roh\-daten.

\glsresetall \cleardoublepage

\thispagestyle{empty}
\def\ackname{Acknowledgment}
\pdfbookmark[1]{\ackname}{ack}
\phantomsection
\noindent
{\large \bf \ackname}\\[4ex]

I would especially like to thank Prof.~Dr.~Joachim Hornegger for the opportunity of doing my Ph.D. in the highly fascinating area of super-resolution within the inspiring research environment at the Pattern Recognition Lab. I am certainly grateful for the freedom that he gave me in my research, the confidence that he put in me as well as his continuous encouragement within the last years. Let me also deeply thank Prof. Dr. Andreas Maier for his great support and outstanding scientific advice in the final phase of preparing this thesis. I am also particularly grateful to Prof. Dr. Claudius Schnörr, who awakened my enthusiasm for image processing during my Master's studies and pointed me towards a Ph.D study.   

I greatly appreciate the wonderful time at the lab over the past years. In particular, let me thank the former Range Imaging Group -- Dr. Sebastian Bauer, Dr. Sven Haase, and Jakob Wasza -- for discussing papers, the great advice in the field of range imaging, and the nice conference trips we had together. Special thanks go to Sven for the fruitful and friendly cooperation within his DFG project. This resulted in several exciting publications that built the foundation of this thesis. A big thanks also goes to Xiaolin~Huang~Ph.D. for his tremendous support on the convergence proof for the algorithm presented in Chapter 4. Many thanks also to my room mates André Aichert, Lennart Husvogt, and Martin Kraus for the relaxed working atmosphere, inspiring scientific discussions, and nice leisure activities. Let me also acknowledge André Aichert, Dr. Alexander Brost, Simone Gaffling, Wilhelm Haas, Dr. Sven Haase, Matthias Hoffmann, and the uncountable number of student tutors for their valuable assistance in teaching. Their awesome help ensured that I enjoyed an excellent balance between teaching and doing research.

I would like to express the immense gratitude for proofreading this thesis to my "`editor team"': André Aichert, Dr. Martin Berger, Martin Gabriel, Dr. Sven Haase, Matthias Hoffmann, and Dr. Ralf Peter Tornow. Their thorough review and invaluable feedback greatly improved this manuscript.

Many parts of my research would have been impossible without appropriate data and I deeply thank all project partners that supported my experimental studies: Peter F\"ursattel and Dr. Sven Haase for enabling the Time-of-Flight data acquisitions, Christiane K\"ohler, Prof.~Dr.~Georg Michelson, and Dr.~Ralf~Peter Tornow for providing clinical fundus images, and Dr. Johannes Jordan for providing the Gerbil framework to visualize multispectral images. Another big thank you goes to Michel B\"atz, Farzad Naderi, and Dr.~Christian~Riess for their great effort to capture super-resolution benchmark datasets within our ongoing collaboration.

I would like to acknowledge all students that I supervised during the last years, especially Cosmin Bercea, Florin C. Ghesu, Axel Heinrich, Katja Mogalle, and Anja Kürten. Their excellent works contributed to several scientific publications.

Last but not least, I would like to deeply thank my whole family and my wife Magdalena. Thank you very much for the patience and emotional support within the ups and downs of the last years.

\vspace{0.7em}
\begin{flushright}
	Thomas K\"ohler
\end{flushright}

 \cleardoublepage\relax	
\fi

  \pagestyle{plain}
  \pagenumbering{roman}

\pdfbookmark[1]{\contentsname}{toc}
\dominitoc
\begin{spacing}{1.06}
	\tableofcontents
\end{spacing}
\cleardoublepage
 \cleardoublepage

  \mainmatter
  \pagestyle{diss}
  \pagenumbering{arabic}

	\ifx\compilechapter\undefined

\chapter{Introduction}
\label{sec:Introduction}

\myminitoc

\noindent
The resolution of an imaging system characterizes the level of spatial detail at which it captures images and -- besides the contrast resolution -- it is considered as a major quality indicator. This is obvious in digital photography, where the camera resolution is directly related to the acquisition of fine textures in a scene. In remote sensing as another prominent example, one is interested in measuring information on a planet surface over long distances, which requires high-resolution cameras. Resolution is also crucial in the context of medical imaging to support interventional or diagnostic workflows. For instance, morphological imaging modalities need to provide precise information regarding human anatomy.

In most of these areas, a large effort has been made by researchers and system manufacturers to develop sensors and optical components that enable high-resolution imagery. However, resolution is inherently limited. Besides technological constraints, the use of improved hardware might lead to unacceptable high costs or sizes of commercial systems. In contrast to mass products with a limited life cycle that may allow the use of improved hardware in future product releases, a simple replacement of hardware components is often not feasible in existing long-lived systems, \eg in remote sensing or medical imaging. In these cases one has to optimally use existing hardware and needs to employ techniques to enhance the actual image resolution. This thesis investigates \textit{super-resolution} methods to reconstruct high-resolution images from low-resolution ones retrospectively. 

This chapter provides an introduction to the physics of digital imaging as well as different paradigms of super-resolution. Finally, the scientific contributions of this work are elaborated and an outline of the different chapters is presented. 

\section{Resolution of Digital Imaging Systems}
\label{sec:ResolutionOfDigitalImagingSystems}

Let us first discuss the meaning of \textit{optical resolution} in terms of imaging systems as it is the focus of any super-resolution method to improve this property. Optical resolution is an abstract term and depends on a variety of physical parameters. In this work, optical resolution is defined in accordance to optics literature and denotes the ability of an imaging system to capture spatial details. A classical approach to objectify this parameter are two-point criterions that measure the ability to resolve two point light sources without interference in the image \cite{DenDekker1997a}. Examples for these objective criterions are the Rayleigh \cite{Rayleigh1879} or the Sparrow criterion \cite{Sparrow1916}. As this thesis is focused on digital optical imaging, we consider an imaging system as the composition of optical components and a sensor array. There are two main aspects that influence the optical resolution\footnote{Notice that in this discussion we exclude conditions related to the scene, \eg motion blur or atmospheric turbulence, that may also affect the overall optical resolution.}.
\begin{figure}[!t]
	\centering
		\includegraphics[width=0.94\textwidth]{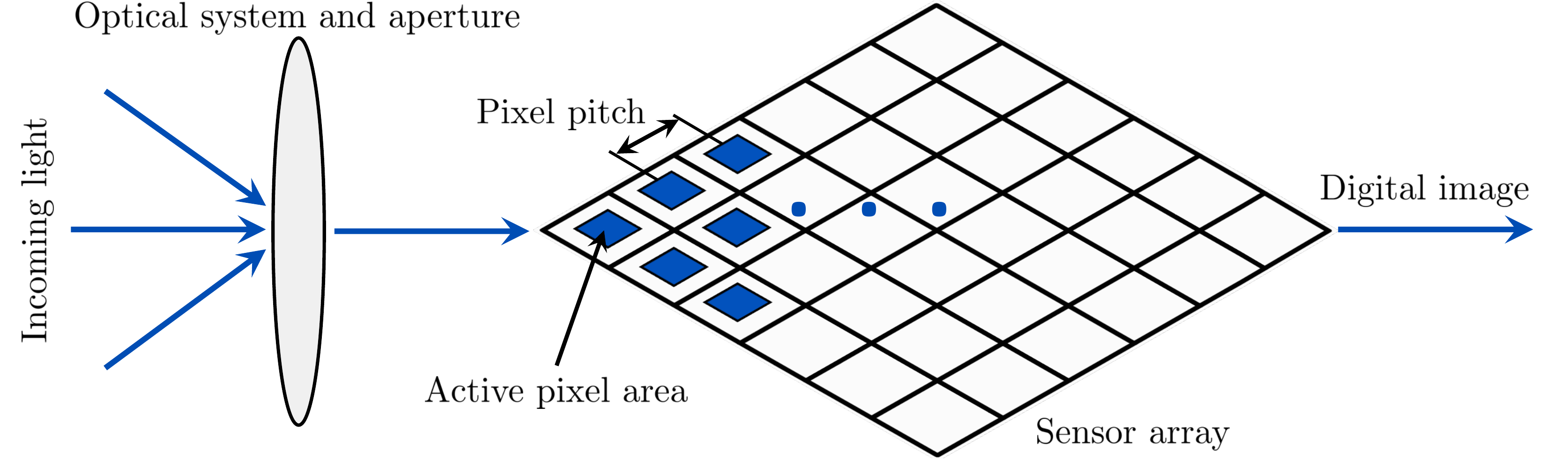}
	\caption[Illustration of the sensor array of a digital imaging system]{Illustration of the sensor array of a digital imaging system with square pixels. Incoming light passes an optical system and is integrated on the active pixel areas.}
	\label{fig:01_sensorArray}
\end{figure}

\paragraph{Limitations of the Optics.}
In terms of optics, the optical resolution is inherently limited by \textit{diffraction} that is related to the camera aperture size and the wavelength of light \cite{Ersoy2006}. The diffraction barrier results in a spread of incoming light waves when passing a small aperture and leads to distortions of the light signal. Moreover, unavoidable manufacturing uncertainties of lenses cause additional distortions. For these reasons, point light sources cannot be captured as ideal points and appear blurred in an acquired image. These distortions limit the optical resolution and are modeled by the optical \gls{psf}. This function denotes the impulse response of the optical system and causes a band-limitation in terms of spatial frequencies that can be actually resolved \cite{Lindberg2012}. 

\paragraph{Limitations of the Sensor.}
In addition to optical effects, the utilized sensor technology influences the optical resolution. In digital imaging, resolution is affected by the discretization of incoming light according to the sensor geometry. \Gls{ccd} or \gls{cmos} systems \cite{ElGamal2005} consist of \glspl{pixel} that represent the sampling positions for this discretization. Two major parameters of the sensor are the pixel pitch and the active pixel area, see \fref{fig:01_sensorArray}. These parameters define the \textit{pixel resolution} that denotes the number of pixels on the sensor array. The pixel resolution directly contributes to the optical resolution as long as the diffraction limit is not exceeded. According to the Nyquist-Shannon sampling theorem \cite{Shannon1948}, the sensor needs to provide a sufficiently high resolution to avoid \textit{aliasing} due to undersampling. 

However, simply putting a higher number of pixels to the sensor array is often impracticable as pixels have a non-infinitesimal size and the maximum sensor area is bounded. Notice that only a percentage of the pixel area is light sensitive, which is quantified by the \textit{fill factor} \cite{ElGamal2005}. The incoming light is integrated over this active area to gain a digital signal. This effectively applies a low-pass filter to the acquired images and leads to a loss of detail. Unfortunately, decreasing the fill factor would result in fewer photons being collected on each pixel. This attenuates the pixel sensitivity and causes an increase of shot noise \cite{Chen2000}. For this reason, simply reducing the fill factor does not necessarily contribute to a higher optical resolution as noise can be seen as another resolution limiting property \cite{DenDekker1997a}.

\section{Super-Resolution in this Work}

Over the past decades, a variety of super-resolution techniques emerged in different scientific disciplines. Common to all of these approaches is the goal to enhance the optical resolution of an imaging system by engineering the resolution restricting aspects discussed in \sref{sec:ResolutionOfDigitalImagingSystems}. To the best of our knowledge, there is no clear taxonomy regarding the meaning of super-resolution and the techniques may fundamentally differ \cite{Driggers2005}. This work distinguishes between \textit{instrumental} and \textit{computational} super-resolution according to Lindberg \cite{Lindberg2012}.

The instrumental approach, also known as optical super-resolution, is focused on an engineering of the optical \gls{psf} in order to increase the band-limitation of the system. Methods that are related to this class have been widely studied in physics and optical engineering and include, among others, the use of photoswitchable proteins in microscopy \cite{Hofmann2005a} or superlenses \cite{Zhang2008}. These techniques aim at breaking the diffraction limit as a resolution limiting property. However, it is in the very nature of the instrumental approach that it requires modifications on the underlying hardware, which is beyond the scope of this work.

Computational super-resolution \cite{Bertero2003} is a complementary approach and features resolution enhancement by means of software -- without considerable effort regarding hardware modifications. This methodology is well-suited for low-cost imaging or workflows that do not allow changes on the system hardware. This area can be further divided into two domains: \textit{diffractive} and \textit{geometrical} approaches \cite{Zalevsky2011}. On the one hand, diffractive super-resolution aims at overcoming the diffraction barrier related to the optical system retrospectively \cite{Garcia2006,Zalevsky2013,Ilovitsh2014}. Geometrical super-resolution on the other hand has the goal to circumvent the limitations related to the sensor. One approach is to address the active pixel area as a resolution limiting factor \cite{Borkowski2009,Borkowski2011}. This has the goal to alleviate the low-pass effect caused by spatial sampling with pixels of finite size. In contrast to these methods, the focus of this thesis lies in computational techniques that consider the pixel pitch as the limiting property. These methods have the objective to reconstruct images at finer pixel sampling from one or an entire sequence of undersampled images and have been widely investigated in image processing \cite{Milanfar2010}. Undersampling refers to the fact that raw images are sampled below the Nyquist-Shannon frequency \cite{Shannon1948} and are therefore affected by aliasing. This improvement of the pixel sampling is due to redundancies or complementary information encoded in low-resolution images. As the primary goal in this area lies in an enhancement of the pixel sampling, we use the pixel resolution as a synonym for the overall optical resolution. For the sake of brevity in the remainder of this thesis, \textit{resolution} simply refers to the pixel resolution of the imaging system. Accordingly, super-resolution denotes the process of reconstructing high-resolution images from low-resolution ones by enhancing the pixel sampling.

This thesis further distinguishes super-resolution according to the type of information that is exploited for resolution enhancement. \textit{Multi-frame} super-resolution obtains one or a set of high-resolution images from a sequence of low-resolution frames by using complementary information across the input frames. This can be achieved by utilizing relative motion \cite{Park2003} as depicted in \fref{fig:01_srExample} or more seldom by defocusing across the frames \cite{Rajagopalan2003a}. The algorithms studied in this work mostly fall into the first category. \textit{Single-image} super-resolution or upsampling recovers a high-resolution image from a single low-resolution one. This can be considered a special case of multi-frame super-resolution but the resulting reconstruction problem is highly underdetermined. State-of-the-art methods in this area are learning based \cite{Yang2010,Freeman2000a,Dong2014} or make use of redundancies within a single image \cite{Glasner2009}. These approaches need to be differentiated from the closely related \textit{deconvolution} methods \cite{Patrizio2016}. The goal of image deconvolution is to remove blur caused by diffraction, atmospheric turbulence, or motion but the pixel resolution of deblurred images remains the same.

\begin{figure}[!t]
	\centering
	\small
	\subfloat[Low-resolution frames]{
		\includegraphics[width=0.52\linewidth]{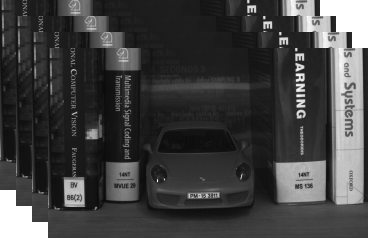}
		\label{fig:01_srExample:lr}
	}
	~
	\subfloat[Super-resolved image]{
		\includegraphics[width=0.4525\linewidth]{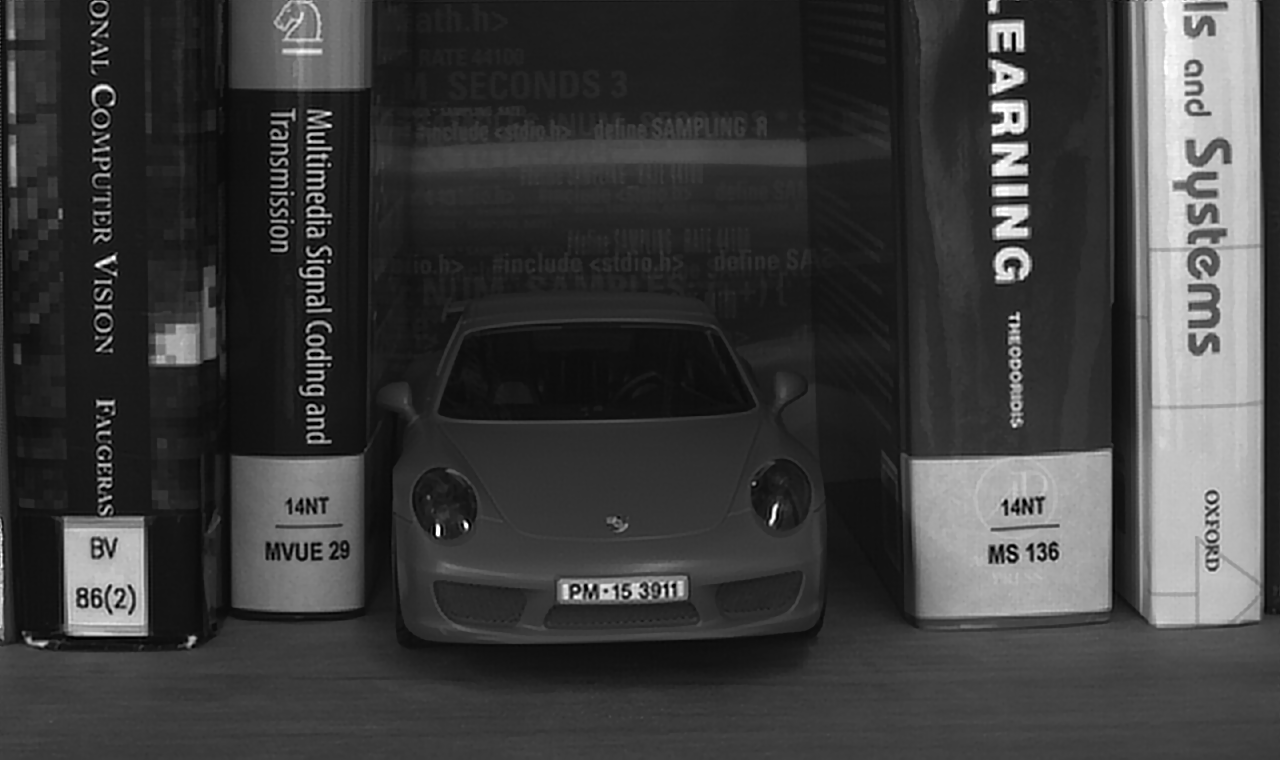}
		\label{fig:01_srExample:sr}
	}
	\caption[Example of multi-frame super-resolution using subpixel motion]{Example of multi-frame super-resolution by exploiting subpixel motion across a set of low-resolution frames. \protect\subref{fig:01_srExample:lr} Sequence of low-resolution frames. \protect\subref{fig:01_srExample:sr} Super-resolved image ($4\times$ magnification) gained from 17 frames using the method proposed in \cref{sec:RobustMultiFrameSuperResolutionWithSparseRegularization}.}
	\label{fig:01_srExample}
\end{figure}

\section{Scientific Contributions}
\label{sec:ScientificContributions}

The major contributions of this thesis concern the theory and the development of computational methods for multi-frame super-resolution. In addition, new applications in different domains of digital optical imaging including workflows in healthcare are studied. Let us outline these contributions that cover three parts. 
\\[0.8em]
\noindent
\textbf{Numerical Methods for Multi-Frame Super-Resolution.}
The first contribution concerns the development of general-purpose techniques for multi-frame super-resolution. This part puts the emphasis on the design of robust numerical algorithms that are well-suited under challenging conditions in real-world imaging systems, where super-resolution is prone to failure. 

In the field of robust numerical algorithms, we propose a novel optimization method based on \textit{space variant Bayesian modeling} of super-resolution. This formulation includes a \textit{confidence-aware observation model} that considers \textit{space variant noise and outliers} in the image formation process. Furthermore, we follow up on recent advances in the theory of compressed sensing \cite{Candes2008} and introduce a \textit{spatially adaptive prior distribution} to exploit \textit{sparsity} of natural images in the gradient domain as prior knowledge for super-resolution. The numerical optimization under this model leads to an \textit{iteratively re-weighted minimization} scheme, which facilitates the simultaneous reconstruction of high-resolution images along with the inference of latent model parameters. This approach can handle super-resolution for intensity images or 3-D range data under challenging conditions like inaccurate motion estimation, photometric variations, or space variant noise.

These methods have been originally published in a journal article \cite{Kohler2015c}.
\\[0.8em]
\noindent
\textbf{Multi-Sensor Super-Resolution for Hybrid Imaging.}
The second contribution concerns the development of super-resolution methods for \textit{hybrid imaging}. For this domain, we introduce novel \textit{multi-sensor} super-resolution algorithms that are applicable to various imaging setups. These algorithms exploit the existence of a set of imaging modalities in contrast to conventional methods that consider only a single one. Overall, we study two problem statements:
\begin{itemize}
	\item First, we investigate super-resolution of one imaging modality under the \textit{guidance of a complementary modality}. To this end, we introduce a novel framework that exploits high-resolution \textit{guidance} data to steer super-resolution on low-resolution \textit{input} data. This comprises guidance data driven motion estimation, spatially adaptive regularization, as well as outlier detection. The merit of this formulation over conventional super-resolution is demonstrated in hybrid 3-D range imaging, where high-resolution color images are utilized to reliably super-resolve low-resolution range data. 
	
	This methodology has been originally published in two conference proceedings \cite{Kohler2013a,Kohler2014a} and one journal article \cite{Kohler2015a}.
		
	\item Second, we examine multi-sensor super-resolution for an arbitrary number of modalities and in the absence of reliable guidance data. We introduce a novel Bayesian model based on \textit{linear regressions across the channels of multi-channel images} that represent the involved modalities. Furthermore, an energy minimization algorithm for the \textit{joint reconstruction} of the different channels and latent hyperparameters of this model is presented. This method is applicable to many target applications, including color-, multispectral-, and hybrid range imaging. The proposed multi-channel reconstruction algorithm exploits mutual dependencies between different channels as a strong prior to boost the accuracy of super-resolution.
		
	This methodology has been originally published in two conference proceedings \cite{Ghesu2014,Kohler2015b}.
\end{itemize}

\noindent
\textbf{Super-Resolution in Medical Imaging.}
The third contribution concerns the transfer of super-resolution algorithms to several fields in medical imaging with the goal to enhance medical workflows. The following applications are covered:
\begin{itemize}
	\item In terms of diagnostic imaging, we study super-resolution for a non-invasive examination of the human eye background by means of retinal fundus images. In contrast to single-shot photography, we propose video imaging and use a tailored approach to reconstruct \textit{high-resolution fundus images from low-resolution video frames}. This method utilizes natural eye motion during an examination as a cue for super-resolution. Furthermore, we introduce a \textit{fully automatic noise and sharpness measure for fundus images} to steer the selection of latent model hyperparameters. The proposed framework enables high-resolution fundus imaging using mobile and cost-effective video hardware. 
	
	These applications have been originally published in three conference proceedings \cite{Kohler2013,Kohler2014,Kohler2016}. 
	
	\item In the field of interventional workflows, we investigate super-resolution to facilitate image-guided surgery based on hybrid range imaging. We adopt multi-sensor super-resolution to \textit{hybrid range imaging in 3-D endoscopy and image guided open surgery} that are studied as example applications. For this purpose, we present two \textit{system calibration schemes} that are tailored for these applications in order to enable sensor data fusion of low-resolution range data and high-resolution color images. The proposed method enables high-resolution 3-D range measurements using current \gls{tof} sensors, which is studied in \textit{ex-vivo} experiments for minimally invasive and open surgery procedures. 
		
	These applications have been originally published in two conference proceedings \cite{Kohler2013a,Kohler2014a} and one journal article \cite{Kohler2015a}.
	
\end{itemize}
To foster reproducible research and future work of other groups, source code of the developed algorithms have been made publicly available in a \textit{multi-frame super-resolution toolbox} for MATLAB\footnote{\url{https://www5.cs.fau.de/research/software/multi-frame-super-resolution-toolbox/}}. This work also led to the publication of the \textit{Super-Resolution Erlangen} (SupER) benchmark \cite{Kohler2017} -- a comparative experimental validation of current super-resolution algorithms on a novel image database.

In addition to these research results, this work also contributed to related areas including multi-frame denoising \cite{Kohler2012,Schirrmacher2017}, blind deconvolution \cite{Kohler2015d}, joint image registration and super-resolution \cite{Bercea2016}, or hardware acceleration of super-resolution \cite{Wetzl2013}. This thesis also contributed to research in medical image analysis including image-based tracking \cite{Kurten2014}, super-resolved segmentation \cite{Haase2013c}, and computer-assisted diagnostics \cite{Kohler2015}.  

\section{Outline of this Thesis}
\label{sec:OutlineOfThisThesis}

This thesis is structured in an introductory part that covers the background of multi-frame super-resolution as well as three main parts as shown in \fref{fig:thesisOverview}. 

\cref{sec:MultiFrameSuperResolutionAndTheSamplingTheorem} presents an in-depth analysis of super-resolution in the frequency domain. This includes a study of \textit{single- and multi-channel sampling} along with a discussion of the \textit{sampling theorem} according to Nyquist and Shannon. The reconstruction of well sampled signals from undersampled ones as the core of super-resolution is modeled as an inverse problem based on the multi-channel theory. A mathematical discussion regarding the \textit{uniqueness} of this reconstruction and the \textit{effective magnification factor} obtainable by super-resolution is presented.

\begin{figure}[!p]
	\centering
		\includegraphics[width=1.00\textwidth]{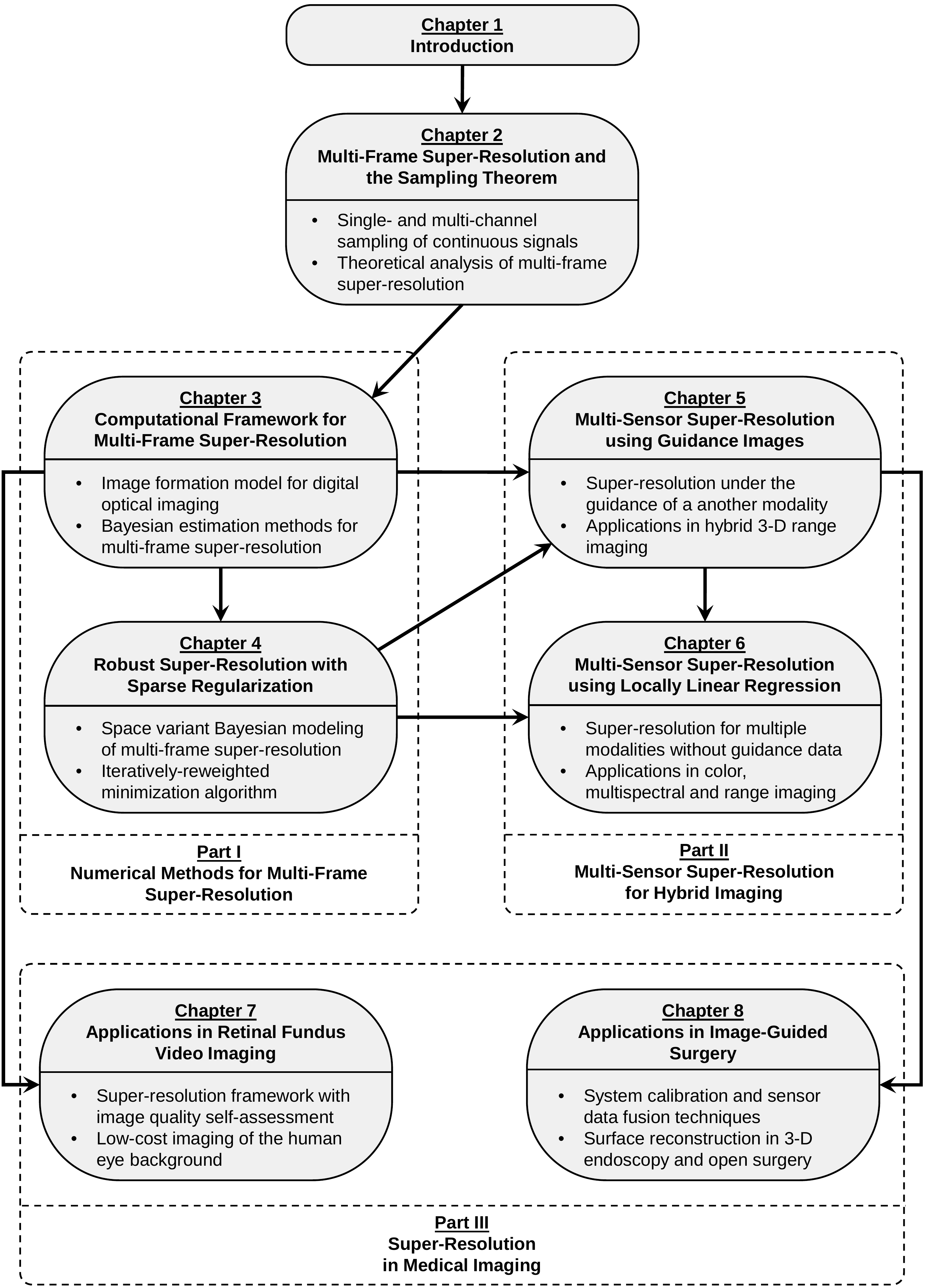}
	\caption[Structure of this thesis and relationship among the chapters]{Structure of this thesis and relationship among the individual chapters. The background part provides a theoretical discussion of the relationship between super-resolution and the Nyquist-Shannon sampling theorem. The main body covers numerical methods for multi-frame super-resolution algorithms (\pref{sec:NumericalMethodsForMultiFrameSuperResolution}), multi-sensor super-resolution for hybrid imaging (\pref{sec:MultiSensorSuperResolutionForHybridImaging}), as well as applications of super-resolutions in medical imaging (\pref{sec:SuperResolutionInMedicalImaging}).}
	\label{fig:thesisOverview}
\end{figure}

The main body is divided into three parts in accordance with the contributions outlined in \sref{sec:ScientificContributions}. \pref{sec:NumericalMethodsForMultiFrameSuperResolution} covers general-purpose methods for multi-frame super-resolution. In \cref{sec:ComputationalFrameworkForMultiFrameSuperResolution}, we introduce the \textit{computational framework for multi-frame super-resolution} that is widely employed in the algorithms presented in the remainder of this work. This chapter concerns the fundamentals of state-of-the-art algorithms ranging from the mathematical modeling of the image formation process to Bayesian methods that formulate super-resolution as a statistical parameter estimation problem. Subsequently, \cref{sec:RobustMultiFrameSuperResolutionWithSparseRegularization} introduces \textit{robust super-resolution with sparse regularization} that extends this framework by space variant observation and prior distributions. In this chapter, we present \textit{iteratively re-weighted minimization} for simultaneous super-resolution and model parameter estimation.

\pref{sec:MultiSensorSuperResolutionForHybridImaging} covers super-resolution for hybrid imaging and comprises the two multi-sensor techniques developed in this work. In \cref{sec:MultiSensorSuperResolutionUsingGuidanceImages}, we introduce \textit{multi-sensor super-resolution using guidance images}. This framework augments conventional super-resolution (\cref{sec:ComputationalFrameworkForMultiFrameSuperResolution}) by motion estimation, adaptive regularization, as well as outlier detection techniques that are steered by guidance data to leverage the reconstruction of a complementary modality. Subsequently, \cref{sec:SuperResolutionForMultiChannelImages} introduces \textit{multi-sensor super-resolution using locally linear regression} to jointly super-resolve a set of imaging modalities without explicitly using one of them as a guidance. This provides a generalization of \textit{guidance image based super-resolution} (\cref{sec:MultiSensorSuperResolutionUsingGuidanceImages}) and employs \textit{iteratively re-weighted minimization} (\cref{sec:RobustMultiFrameSuperResolutionWithSparseRegularization}) for numerical optimization. We present several potential applications of both methodologies ranging from color and multispectral imaging to hybrid 3-D range imaging.

\pref{sec:SuperResolutionInMedicalImaging} is focused on novel applications of super-resolution to facilitate diagnostic and interventional medical imaging. In \cref{sec:RetinalFundusVideoImaging}, we address \textit{super-resolution for retinal fundus video imaging}. This chapter is concerned with a new method to reconstruct high-resolution fundus images from multiple low-resolution video frames by exploiting natural human eye movements. This method is presented in the field of low-cost imaging to gain fundus images at high spatial resolution by means of low-priced and mobile video camera systems. \cref{sec:HybridRangeImagingForImageGuidedSurgery} covers \textit{super-resolution for image-guided surgery} using hybrid range imaging. This chapter adopts the previously introduced \textit{guidance image based super-resolution} (\cref{sec:MultiSensorSuperResolutionUsingGuidanceImages}) and presents the system calibration schemes to make it accessible for the desired application. The proposed framework is presented in the context of hybrid 3-D endoscopy and image-guided open surgery.

In \cref{sec:09_Summary}, we draw a conclusion and summarize the main findings of this thesis. Finally, \cref{sec:10_Outlook} provides an outlook regarding promising directions for future research.

\glsresetall

\chapter{Multi-Frame Super-Resolution and the Sampling Theorem}
\label{sec:MultiFrameSuperResolutionAndTheSamplingTheorem}

\myminitoc

\noindent
This chapter is devoted to the relationship between the Nyquist-Shannon sampling theorem \cite{Shannon1948} as fundamental principle underlying the acquisition of digital signals and super-resolution reconstruction. In order to formulate this theoretical framework, single-channel sampling is generalized to the multi-channel case, where a continuous signal is sampled multiple times to capture a set of discrete signals. Based on a Fourier domain analysis, signal reconstruction to recover the original, continuous signal from multiple sampled channels is formulated as a linear inverse problem. It is shown that a solution of this linear problem yields a super-resolved signal to overcome the constraints stated by the sampling theorem. Finally, inherent limitations of super-resolution regarding the maximum magnification and the uniqueness of the reconstruction are derived.

The analysis of multi-channel sampling presented in this chapter is based on the pioneering work on image super-resolution algorithms \cite{Tsai1984,Kim1990,Tekalp1992a} formulated in the frequency domain. A similar analysis is also presented in the work of Vandewalle \cite{Vandewalle2006a,Vandewalle2007}, where the more general concept of finite dimensional Hilbert spaces is used as mathematical tool.

\section{Introduction}

In digital imaging, a continuous description of the real world, \ie geometry or texture of objects, is discretized to provide a digital representation. In terms of signal processing, one major parameter of such a system is how this sampling is performed and whether the resulting image is sampled appropriately. In case of a digital camera, the number of pixels on a sensor array as well as the spacing between the pixels are relevant system parameters related to the sampling process, see \sref{sec:ResolutionOfDigitalImagingSystems}. The Nyquist-Shannon sampling theorem \cite{Shannon1948} states inherent requirements regarding an appropriate sampling in order to capture a digital representation without loss of information. These requirements concern the sampling frequency as well as the spectral properties of the continuous signal. Violating the sampling theorem such that the sampling frequency is chosen too small with regard to the signal's spectral properties leads to \textit{aliasing}. In case of aliasing, low frequency components of the continuous signal are superimposed by its high frequency components resulting in signal distortions. 

In presence of aliasing, a further analysis of the sampled signal is prone to errors and restoration techniques are required to overcome undersampling. In this context, super-resolution aims at reconstructing a digital signal that is free of aliasing artifacts from an undersampled signal. This methodology is based on the concept of \textit{multi-channel} sampling, where a continuous signal is sampled multiple times as opposed to classical \textit{single-channel} sampling. Super-resolution can be seen as a fusion of multiple channels in order to overcome the limitations stated by the sampling theorem for a single channel. This is feasible by exploiting complementary information across the channels. In case of digital imaging, these channels correspond to different frames of an image sequence taken from the same scene, whereas each frame contains a complementary view.

The remainder of this chapter is organized as follows. \sref{sec:02_SinglechannelSamplingAndSamplingTheorem} covers the fundamentals of sampling under ideal and real conditions along with the Nyquist-Shannon theorem. In \sref{sec:02_MultichannelSamplingTheory}, we extend single-channel sampling to the multi-channel case. Accordingly, super-resolution is formulated as an inverse linear problem based on the multi-channel sampling theory. \sref{sec:02_FromMultichannelSamplingToSuperResolution} presents a numerical algorithm to solve this inverse problem in the frequency domain. \sref{sec:02_LimitsOfSuperResolution} covers fundamental properties of super-resolution regarding the effective magnification factor and the uniqueness of the signal reconstruction. Finally, \sref{sec:02_Conclusion} provides a summary and a conclusion of these concepts. 

\section{Single-Channel Sampling Theory}
\label{sec:02_SinglechannelSamplingAndSamplingTheorem}

The sampling of a continuous signal is first modeled for single channels, where one set of discrete samples is obtained from the original signal \cite{Mallat1999}. For convenience, but without loss of generality, this process is modeled for one-dimensional signals only. As the different dimensions of multidimensional signals are separable, the underlying theory can be extended and applied to each individual dimension. Since common physical measurements such as digital images are real-valued, we limited the  following analysis to real-valued signals.

Let $x: \Real \rightarrow \Real$ be a real-valued, continuous signal denoted by $x(t)$\label{notation:realValuedSignal}. For discretization in the domain $t \in \Real$, $x(t)$ is sampled in equidistant steps with \textit{sampling pitch} $T$. The sampled signal defined as continuous function is denoted by $y(t) = \mathcal{D}_T \{ x(t) \}$,\label{notation:sampledSignal} where $\mathcal{D}_T\{\cdot\}$\label{notation:samplingPitch}\label{notation:samplingOperator} denotes the sampling operator. Then, the discretization $y[n]$\label{notation:discretizedSignal} associated with $x(t)$ is obtained according to $y[n] \defeq x(nT)$, where $n \in \Integer$ denotes the sample index. This process is examined for two different situations including \textit{ideal sampling} as well as \textit{real sampling} as a reasonable model in the context of digital imaging.

\subsection{Ideal Single-Channel Sampling}
\label{sec:02_IdealSampling}

In case of ideal sampling, the sampling operator is modeled by Dirac delta impulses. Then, $y(t)$ is obtained from a product of the continuous signal $x(t)$ and a Dirac comb \cite{Mallat1999} as depicted in \fref{fig:02_singleChannelSampling}. More formally, single-channel sampling is modeled by:
\begin{equation}
	\label{eqn:02_sampling_diracSampling}
	y(t) = \mathcal{D}_T \{ x(t) \} \defeq \sum_{m = -\infty}^\infty x(t) \delta(t - m T),
\end{equation}
where $m \in \Integer$ and the discrete Dirac delta\label{notation:diracDelta} is defined as:
\begin{equation}
	\delta(t) \defeq 
	\begin{cases}
		1 	& \text{if } t = 0 \\
		0		& \text{otherwise}
	\end{cases}.
\end{equation}
The resulting signal $y(t)$ is continuous and represents the discrete values of $y[n]$ for $t = nT$. In order to present the sampling theorem, we model the sampling process in the frequency domain. Let $X(f) = \mathcal{F} \{ x(t) \}$\label{notation:cftOp} be the \gls{cft} \cite{Rahman2011} of the signal $x(t)$ defined as:
\begin{equation}
	X(f) = \mathcal{F} \! \left\{ x(t) \right\} \defeq \int_{- \infty}^\infty x(t) \cdot \exp(-j 2\pi f t) dt,
\end{equation}
where $j$ is the imaginary unit. Using the Fourier transform\label{notation:cft}\label{notation:imagUnit}, linearity, and the convolution theorem, \eref{eqn:02_sampling_diracSampling} can be written in the frequency domain according to:
\begin{equation}
	\begin{split} 
		Y(f) &= \mathcal{F} \! \left\{ \sum_{m = -\infty}^\infty x(t) \delta(t - mT)  \right\} \\
				 &= \sum_{m = -\infty}^\infty \mathcal{F} \! \left\{ x(t)  \delta(t - mT) \right\} \\
				 &= \sum_{m = -\infty}^\infty \mathcal{F} \! \left\{ x(t) \right\} \conv \mathcal{F} \! \left\{ \delta(t - mT) \right\}
				 = \frac{1}{T} \sum_{m = -\infty}^\infty X(f) \conv \Delta \! \left( f - \frac{m}{T} \right)
	\end{split} 
\end{equation}
where $\conv$\label{notation:conv} denotes the convolution, $Y(f)$ denotes the \gls{cft} associated with the sampled signal $y(t)$, and $\Delta ( f )$\label{notation:diracDeltaCft} is the \gls{cft} of the Dirac delta $\delta(t)$ \cite{Mallat1999}. According to the definition of the sampling pitch, we define the \textit{sampling frequency} $f_s = \frac{1}{T}$\label{notation:samplingRate}. Thus, ideal sampling can be described as:
\begin{equation}
	\label{eqn:02_sampling_fourierPeriodicSummation}
	\begin{split} 
		Y(f)	& = \sum_{m = -\infty}^\infty X(f) \conv f_s \Delta \! \left( f - m f_s \right) \\
					& = f_s \sum_{m = -\infty}^\infty X(f - m f_s).
	\end{split} 
\end{equation}
Accordingly, the sampling of $x(t)$ by the frequency $f_s$ corresponds to a periodic summation of $X(f)$, where the periodic length is given by the sampling frequency $f_s$. One key question is under which conditions the continuous signal $x(t)$ can be fully characterized by its discrete samples $y[n]$ without loss of information. The possibility of this reconstruction depends on the Fourier properties of $x(t)$ and is studied for \textit{band-limited} signals that are defined as follows.
\begin{restatable}[Band-limited signal]{definition}{02_bandLimitedSignal}
	A continuous signal $x(t)$ is band-limited if there exists a cut-off frequency $f_0$\label{notation:bandLimit} such that $X(f) = \mathcal{F} \{ x(t) \}$ fulfills $X(f) = 0$ for $|f| \geq f_0$. 
\end{restatable}
Let us assume that $x(t)$ is band-limited. Depending on the sampling frequency $f_s$, three situations for the sampling process can be distinguished \cite{Vandewalle2010}, see \fref{fig:02_aliasing}. In order to verify if $x(t)$ can be fully reconstructed from $y[n]$, the properties of the periodic summation in \eref{eqn:02_sampling_fourierPeriodicSummation} are analyzed.

\begin{figure}[!t]
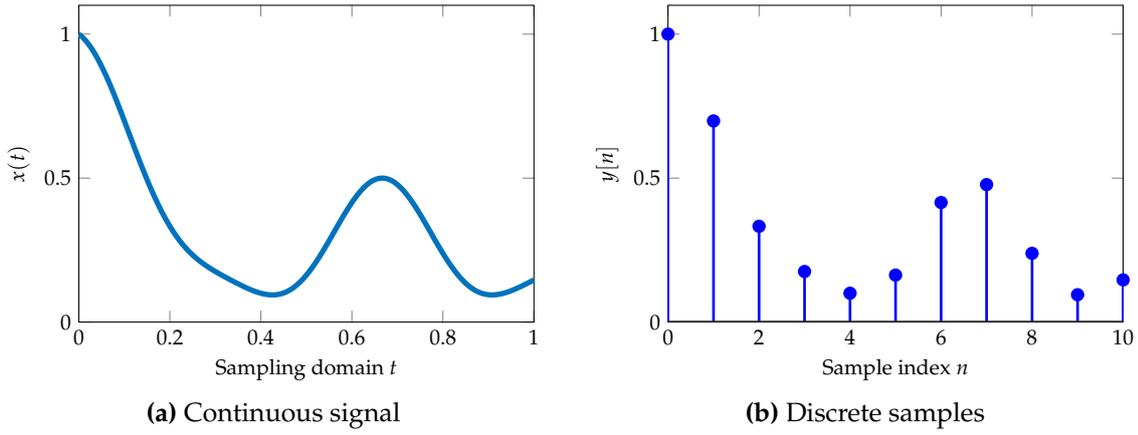

	\scriptsize  
	\centering 
	\setlength \figureheight{4.2cm} 
	\setlength \figurewidth{6.0cm} 
	\subfloat[Continuous signal]{\input{images/chapter2/singleChannelSampling_signal.tikz}}\qquad
	\subfloat[Discrete samples]{\input{images/chapter2/singleChannelSampling_samples.tikz}}
	\caption[Ideal sampling to obtain discrete samples from a continuous signal]{Ideal sampling to obtain discrete samples $y[n]$ from a continuous signal $x(t)$.}
	\label{fig:02_singleChannelSampling}
\end{figure}

\paragraph{Nyquist Sampling.} 
If the sampling frequency $f_s$ is chosen according to $f_s \geq 2 f_0$, adjacent parts of the spectrum $Y(f)$ calculated by the periodic summation of $X(f)$ in \eref{eqn:02_sampling_fourierPeriodicSummation} are non-overlapping. For this situation that is depicted in \fref{fig:02_aliasing:nyquist} and termed as \textit{Nyquist sampling}, it is feasible to recover $x(t)$ from its discrete samples $y[n]$. This fact is explained by the Nyquist-Shannon sampling theorem for band-limited signals \cite{Shannon1948}.
\begin{restatable}[Nyquist-Shannon sampling theorem]{thm}{02_samplingTheorem}
	Let $x(t)$ be a band-limited continuous signal with cut-off frequency $f_0$. Then, $x(t)$ is completely described by its discrete samples $y[n]$ obtained by the sampling frequency $f_s$ if $f_s \geq 2 f_0$.
\end{restatable}

In case of Nyquist sampling, the signal $x(t)$ can be reconstructed from $y[n]$ using low-pass filtering to remove the periodic parts of $Y(f)$. We can obtain $x(t)$ based on the inverse \gls{cft} $\mathcal{F}^{-1}\{\cdot\}$\label{notation:cftOpInv} according to:
\begin{equation}
	x(t) = \mathcal{F}^{-1} \Big\{ Y(f) \cdot H_{\text{reco}}(f) \Big\},
\end{equation}
where $H_{\text{reco}}(f)$ denotes the \gls{cft} of a reconstruction low-pass filter to suppress the periodic parts. For example, one can employ the ideal low-pass filter:
\begin{equation}
	H_{\text{reco}}(f) =
	\begin{cases}
		1		& \text{if}~ |f| \leq f_0 \\
		0		& \text{otherwise}
	\end{cases}.
\end{equation}
\begin{figure}[!p]
	\captionsetup[subfloat]{labelformat=parens}
	\scriptsize  
	\centering 
	\setlength \figureheight{3.0cm} 
	\setlength \figurewidth{14.0cm} 
	\subfloat[\gls{cft} of the continuous signal $x(t)$]{\input{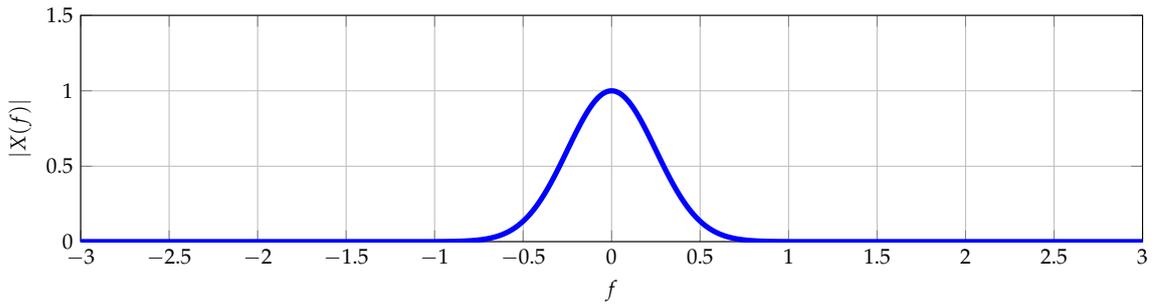}} \\
	\subfloat[\gls{cft} of the sampled signal $y(t)$ under Nyquist sampling ($f_s \geq 2 f_0$)]{\input{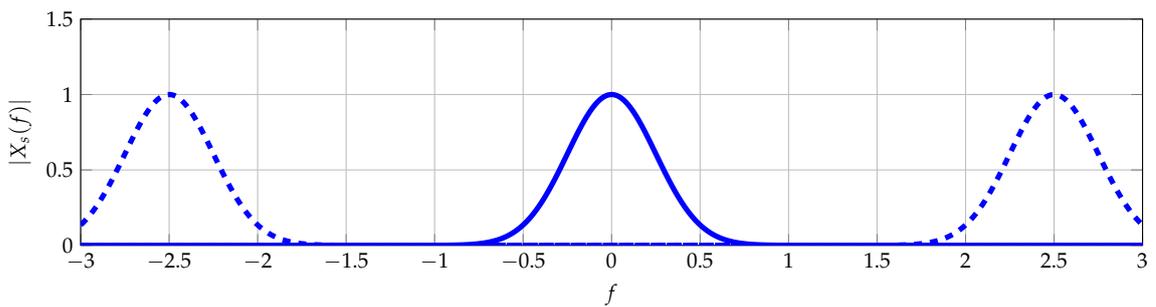}\label{fig:02_aliasing:nyquist}} \\
	\subfloat[\gls{cft} of the sampled signal $y(t)$ for undersampling with partial aliasing ($f_0 \leq f_s < 2 f_0$)]{\input{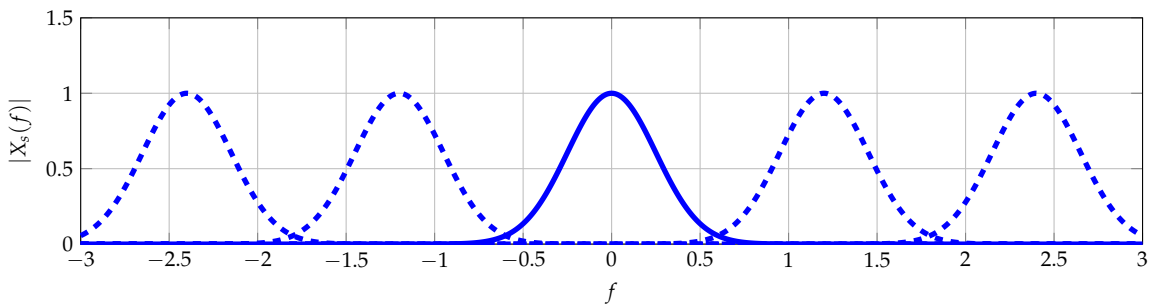}\label{fig:02_aliasing:partial}} \\
	\subfloat[\gls{cft} of the sampled signal $y(t)$ for undersampling with total aliasing ($f_s < f_0$)]{\input{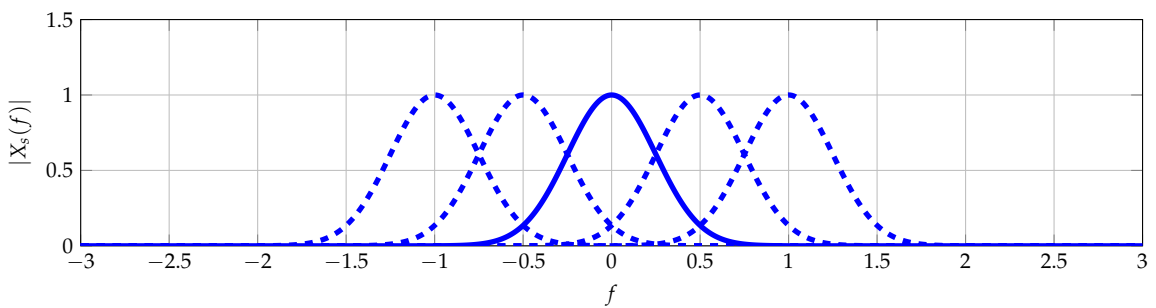}\label{fig:02_aliasing:total}}
	\caption[Sampling of a continuous, band-limited signal]{The sampling of the continuous, band-limited signal $x(t)$ with frequency $f_s$ corresponds to a periodic summation of the \gls{cft} $X(f)$. For the sake of visualization, the frequencies $f$ are normalized \wrt the band-limitation $f_0$ of $x(t)$. According of \cite{Vandewalle2010}, three situations for the sampling process can be distinguished. Depending on $f_s$, the samples $y[n]$ are acquired at the Nyquist rate, partial aliased or total aliased.}
	\label{fig:02_aliasing}
\end{figure}

\paragraph{Undersampling with Partial Aliasing.} While $x(t)$ can be fully reconstructed if the sampling theorem holds true, this is no longer feasible in case of undersampling. Let us assume that $f_0 < f_s < 2 f_0$. Then, in the periodic summation in \eref{eqn:02_sampling_fourierPeriodicSummation}, parts of $X(f)$ are superimposed as depicted in \fref{fig:02_aliasing:partial}. In this situation, the samples $y[n]$ are considered as \textit{partial aliased}. 

In digital imaging, a violation of the sampling theorem leads to aliasing artifacts that are visible as Moir\'{e} pattern. This effect is visualized on the ISO 12233:2000 resolution chart \cite{ISO2000} in \fref{fig:02_reschart} that was acquired with a Basler acA2000-50gm \gls{cmos} camera\footnote{\url{http://www.baslerweb.com/en}}. The resolution chart was captured at the full pixel resolution of the camera (\fref{fig:02_reschart:full}) as well as at a reduced resolution using 4$\times$4 hardware binning \cite{Kohler2017} of pixels on the sensor array (\fref{fig:02_reschart:binning}). Here, line pairs that have small spacings compared to the sampling frequency are distorted by aliasing.

Notice that a reconstruction of $x(t)$ using low-pass filtering would be distorted due to undersampling if aliasing is not considered. One can remove aliasing in the design of the reconstruction low-pass filter $\tilde{H}_{\text{reco}}(f)$:
\begin{equation}
	\tilde{H}_{\text{reco}}(f) =
	\begin{cases}
		1		& \text{if}~ |f| \leq f_s - f_0 \\
		0		& \text{otherwise}
	\end{cases}.
\end{equation}  
Unfortunately, removing the signal distortions caused by undersampling, one also loses the high-frequency content present in the original signal $x(t)$. Another strategy to overcome aliasing is an artificial increase of the sampling frequency by means of super-resolution reconstruction as discussed below.
\begin{figure}[!t]
	\centering
	\subfloat[Full resolution]{\includegraphics[width=0.485\textwidth]{images/chapter2/resolutionChart_bin1}\label{fig:02_reschart:full}}~	
	\subfloat[4$\times$4 hardware binning]{\includegraphics[width=0.485\textwidth]{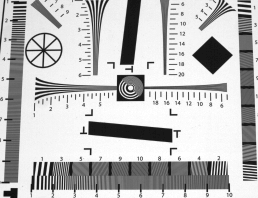}\label{fig:02_reschart:binning}}
	\caption[Aliasing in digital imaging on a resolution chart]{Aliasing in digital imaging depicted on the ISO 12233:2000 resolution chart \cite{ISO2000}. \protect\subref{fig:02_reschart:full} Acquisition of the resolution chart at the full pixel resolution (2048$\times$1088\,px) of a Basler acA2000-50gm camera. \protect\subref{fig:02_reschart:binning} Acquisition of the same chart with 4$\times$4 hardware binning relative to the full pixel resolution. Notice that structures with high spatial frequencies relative to the sampling frequency, \eg line pairs with small spacings, are distorted by a Moir\'{e} pattern caused by undersampling.}
	\label{fig:02_reschart}
\end{figure}

\paragraph{Undersampling with Total Aliasing.}
A situation that is even more severe appears if the sampling frequency is chosen as $f_s < f_0$. In this case, the interference in the spectrum $Y(f)$ results in a \textit{total aliased} discretization $y[n]$ as depicted in \fref{fig:02_aliasing:total}. As opposed to the aforementioned partial aliasing, all parts of the spectrum $Y(f)$ and thus the entire signal $y[n]$ are distorted. Notice that a simple reconstruction of the continuous signal $x(t)$ by a removal of the aliasing artifacts using low-pass filtering is no longer possible in this situation. However, similar as for partial aliased signals, we will show how super-resolution can be utilized to perform a reconstruction of $x(t)$.

\subsection{Real Single-Channel Sampling}
\label{sec:02_RealSampling}

Up to now, the sampling process was considered to be ideal, such that a Dirac delta can be used to model the sampling operator $\mathcal{D}_T$. Theoretically, this would result in an arbitrarily high resolution as long as the sampling frequency is chosen such that the sampling theorem is fulfilled. However, this simplistic assumption is never feasible in practice. In case of real sampling, the sampled signal is a blurred version of the original one due to the fact that the impulse response of the acquisition device deviates from the Dirac delta, see \sref{sec:ResolutionOfDigitalImagingSystems}. Recently, non-ideal sampling models gained attention for signal reconstruction. This results in a deconvolution problem rather than interpolation as in the case of ideal sampling \cite{Eldar2006, Ramani2008, Guevara2010}. Modeling of real sampling becomes important for imaging systems, where diffraction, manufacturing uncertainties of lenses, and the summation of light photons on a finite pixel area of the sensor array restrict an ideal sampling and introduce blur. An illustration of this issue is depicted for the ISO 12233:2000 resolution chart in \fref{fig:02_reschartBlur}. In this example, steep edges on the resolution chart appear blurred in digital images.

Mathematically, blur in the sampling model is considered by replacing the Dirac delta in \eref{eqn:02_sampling_diracSampling} by a blur kernel $h(t, t_0)$. Using a linear blur model, the sampled signal $y(t)$ is given by:
\begin{equation}
	y(t) = \sum_{m = -\infty}^\infty x(t) h(t, m T).
\end{equation}
This models the blur at position $t$, whereas the kernel is evaluated at the sample positions $m T$. If the blur kernel is assumed to be \gls{lsi}, \ie $h(t, mT) = h(t - mT)$, the amount of blurring depends only on the distance $t - m T$. Then, the sampling process can be described according to: 
\begin{equation}
	\label{eqn:02_sampling_realSamplingDefinition}
	\begin{split}
		y(t)	&= \sum_{m = -\infty}^\infty \underbrace{\left( x(t) \conv h(t) \right)}_{z(t)} \delta(t - m T) \\
					&= \mathcal{D}_T \! \left\{ x(t) \conv h(t)  \right\},
	\end{split}
\end{equation}
Thus, the sampling process can be described in two steps. First, a filtered version $z(t)$ of the continuous signal $x(t)$ according to the underlying kernel $h(t)$\label{notation:blurKernel1D} is determined. Since $h(t)$ is the impulse response of a low-pass filter, the convolution to determine $z(t)$ corresponds to an averaging of $x(t)$ over the support of the kernel, which explains the blurring of $x(t)$. Finally, to obtain a sampled signal $y(t)$ and thus the discretization $y[n]$, $z(t)$ is sampled with ideal Dirac deltas.
\begin{figure}[!t]
	\centering
	\begin{tikzpicture}[spy using outlines={rectangle, blue, magnification=13.0, height=3.72cm, width=5.1cm, connect spies, every spy on node/.append style={thick}}] 
		\node {\pgfimage[width=0.53\linewidth]{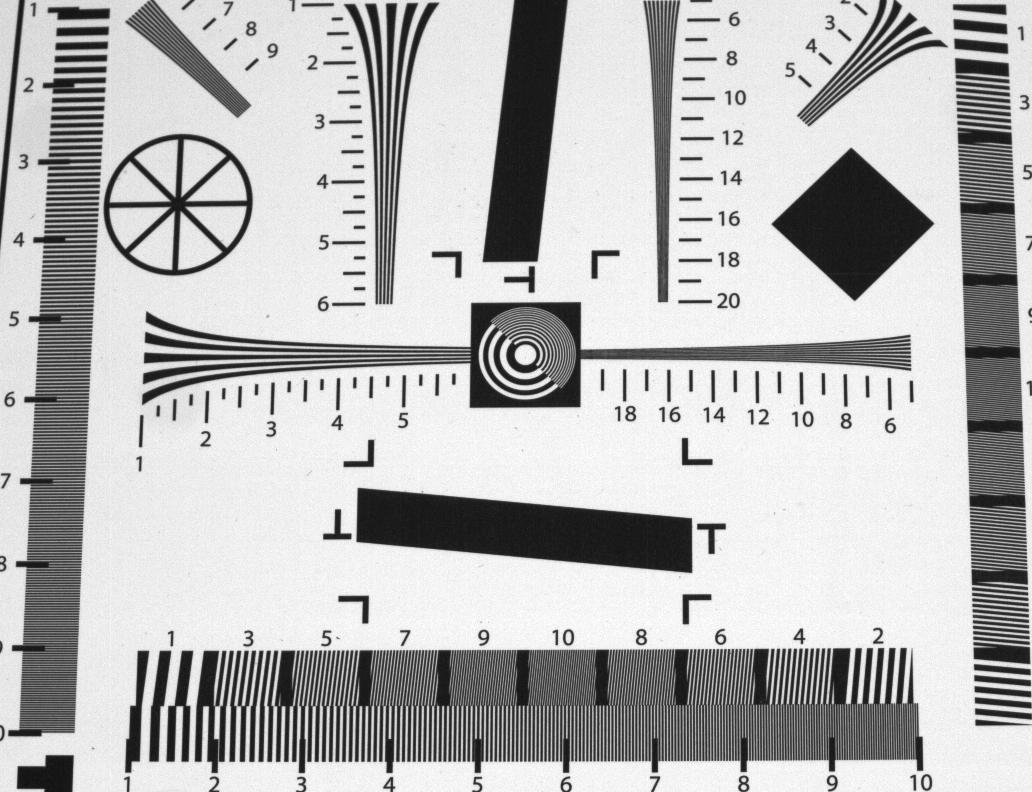}}; 
    \spy on (1.41, -0.44) in node [left] at (9.7, 0);
  \end{tikzpicture}
	\caption[Sampling in digital imaging on a resolution chart]{Illustration of sampling in digital imaging using the ISO 12233:2000 resolution chart \cite{ISO2000}. Steep edges on the resolution chart are blurred in digital images due to diffraction, non-ideal lenses, and the finite pixel size of the camera.}
	\label{fig:02_reschartBlur}
\end{figure}

\section{Multi-Channel Sampling Theory}
\label{sec:02_MultichannelSamplingTheory}

In the previous section, the analysis was restricted to single-channel sampling, where the continuous signal $x(t)$ is sampled once to acquire a discrete signal $y[n]$. One situation interesting for the development of super-resolution algorithms is the case of multi-channel sampling \cite{Papoulis1977,Unser1997}, where $x(t)$ is sampled multiple times. This appears in digital imaging, where multiple images of a scene can be captured by moving the camera to different viewpoints while acquiring video data. This section presents multi-channel sampling based on the formulation of Tsai and Huang \cite{Tsai1984}. In addition to the theoretical insights, this introduces the historically first multi-frame super-resolution algorithm proposed in \cite{Tsai1984}.

Multi-channel sampling of the continuous signal $x(t)$ can be modeled by a sequence of complementary channels determined from $x(t)$. If the sampling of these channels is not synchronized, the $k$-th channel $\FrameIdx{x}{k}(t)$, $k = 1, \ldots, K$\label{notation:continousChannel} can be defined as a shifted version of $x(t)$ according to $\FrameIdx{x}{k}(t) = x(t - t_k)$, where $t_k \in \Real$\label{notation:channelOffset} denotes the \textit{channel offset}. Without loss of generality, the first channel is defined as reference, \ie $t_1 = 0$. The sampled signals obtained from the different channels are denoted as $\FrameIdx{y}{k}(t)$\label{notation:sampledChannel} and the corresponding discrete samples are given by $\FrameIdx{y}{k}[n] \defeq \FrameIdx{y}{k}(n T_k)$\label{notation:discreteChannel}, where $T_k$\label{notation:channelSamplingPitch} denotes the sampling pitch of the $k$-th channel. For the sake of notational brevity, let us assume that the sampling pitch is fixed, \ie $T_k = T$ for all channels. The samples $\FrameIdx{y}{k}[n]$ contain complementary information about the underlying continuous signal $x(t)$. Consequently, $x(t)$ is sampled by a non-uniform scheme parametrized by the sampling frequency and the channel offsets as illustrated in \fref{fig:02_multichannelSampling}. We analyze this process for two situations.
\begin{figure}[!t]
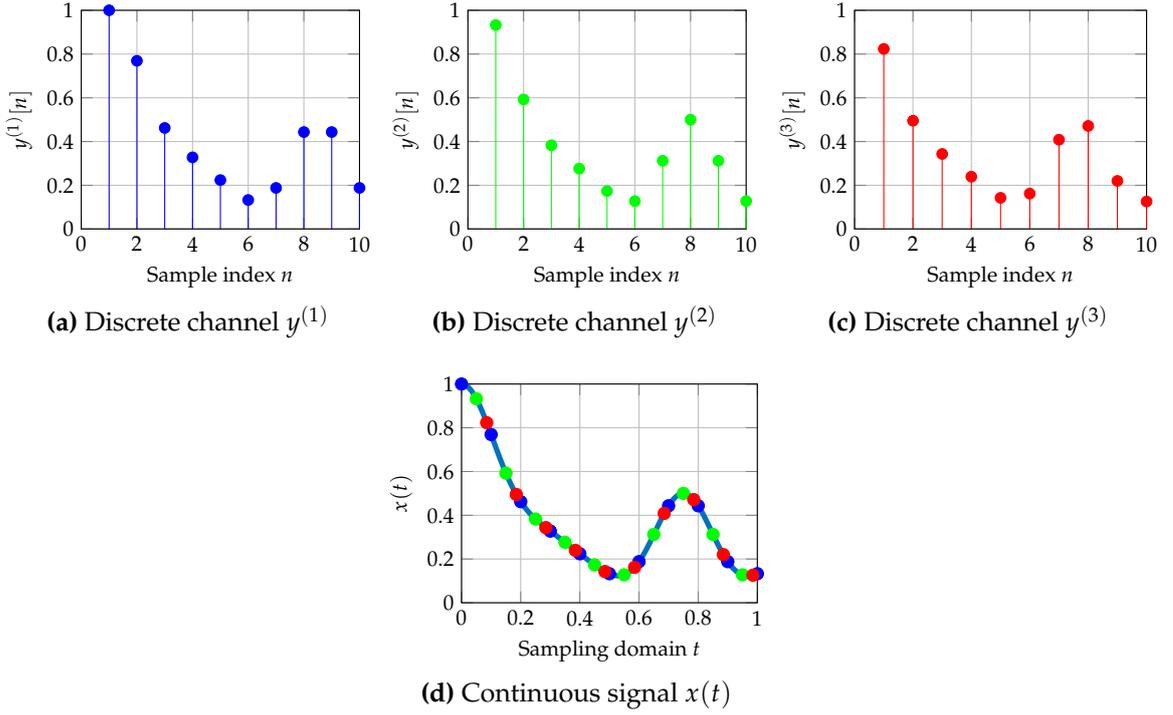

	\centering
	\scriptsize 
	\setlength \figureheight{2.9cm} 
	\setlength \figurewidth{3.85cm} 
	\subfloat[{Discrete channel $\FrameIdx{y}{1}$}]{\input{images/chapter2/multiChannelSampling_channel1.tikz}\label{fig:02_multichannelSampling:1}}~
	\subfloat[{Discrete channel $\FrameIdx{y}{2}$}]{\input{images/chapter2/multiChannelSampling_channel2.tikz}\label{fig:02_multichannelSampling:2}}~
	\subfloat[{Discrete channel $\FrameIdx{y}{3}$}]{\input{images/chapter2/multiChannelSampling_channel3.tikz}\label{fig:02_multichannelSampling:3}}\\
	\setlength \figureheight{2.9cm} 
	\setlength \figurewidth{3.9cm} 
	\subfloat[Continuous signal $x(t)$]{\input{images/chapter2/multiChannelSampling_signal.tikz}\label{fig:02_multichannelSampling:signal}}
	\caption[Multi-channel sampling with constant sampling pitch]{Illustration of multi-channel sampling with constant sampling pitch for all channels. The continuous signal $x(t)$ is sampled $K$ times ($K = 3$) according to the channel offsets $t_k$ with $k = 1, \ldots, K$ as shown in \protect\subref{fig:02_multichannelSampling:1} - \protect\subref{fig:02_multichannelSampling:3}. The fusion of the resulting discrete channels $\FrameIdx{y}{k}[n]$ leads to a non-uniform sampling in the domain $t$ as shown in \protect\subref{fig:02_multichannelSampling:signal}.}
	\label{fig:02_multichannelSampling}
\end{figure}
	
\subsection{Ideal Multi-Channel Sampling}
\label{sec:02_MultichannelSamplingForIdealKernel}

Let us first examine ideal multi-channel sampling, where the sampling operator $\mathcal{D}_T\{ \cdot \}$ is modeled by the Dirac comb. Thus, the sampled version of the $k$-th channel is given by:
\begin{equation}
	\begin{split} 
		\FrameIdx{y}{k}(t)	
			&= \sum_{m = -\infty}^\infty \FrameIdx{x}{k}(t) \delta(t - m T_k) \\
			&= \sum_{m = -\infty}^\infty x(t - t_k) \delta(t - m T_k).
	\end{split}
\end{equation}
Notice that the \gls{cft} $\FrameIdx{X}{k}(f) = \mathcal{F}\{\FrameIdx{x}{k}(t)\}$ for $k > 1$ is related to the \gls{cft} of the first channel $X(f) = \mathcal{F}\{\FrameIdx{x}{1}(t)\}$ using the shift property of the Fourier transform according to:
\begin{equation}
	\label{eqn:02_sampling_shiftProperty}
	\FrameIdx{X}{k}(f) = \exp(-j 2\pi f t_k) X(f).
\end{equation}
In order to derive a relationship between $\FrameIdx{y}{k}[n]$ and $x(t)$, it is assumed that all discrete channels are defined by a finite number of $N$ samples acquired over a finite interval. Then, the discrete samples $\FrameIdx{y}{k}[n]$ are represented by $N$ complex-valued frequency coefficients using the \gls{dft}\label{notation:dft} \cite{Oppenheim1999}:
\begin{equation}
	\label{eqn:02_sampling_defDFT}
	\FrameIdx{\mathcal{Y}}{k}[n] = \sum_{m = 0}^{N-1} \exp \left( -j 2\pi n \frac{m}{N} \right) \FrameIdx{y}{k}[m],
\end{equation}
where $n = 0, \ldots, N-1$. The \gls{dft} coefficients $\FrameIdx{\mathcal{Y}}{k}[n]$ are related to the \gls{cft} $\FrameIdx{X}{k}(f)$ of the $k$-th channel. According to the aliasing property of the Fourier transform in \eref{eqn:02_sampling_fourierPeriodicSummation} and the fact that the \gls{cft} and the associated \gls{dft} of a real-valued signal $x(t)$ are symmetric \cite{Cooley1969}, this relationship is given by:
\begin{equation}
	\label{eqn:02_sampling_aliasingPropertyInfty}
	\FrameIdx{\mathcal{Y}}{k}[n] = f_s \sum_{m = -\infty}^{\infty} \FrameIdx{X}{k} \! \left( \frac{n}{N} f_s - m f_s \right).
\end{equation}

Since the continuous signal $x(t)$ is assumed to be band-limited at a certain cut-off frequency, we have $X(f) = 0$ for $|f| \geq L f_s$ with a finite integer $L$. Hence, the aliasing property can be formulated for a finite interval in the Fourier domain instead of considering the infinite summation in \eref{eqn:02_sampling_aliasingPropertyInfty}. Moreover, using the Fourier shift theorem in \eref{eqn:02_sampling_shiftProperty}, the aliasing property is formulated as:
\begin{equation}
	\begin{split} 
		\FrameIdx{\mathcal{Y}}{k}[n]	
			&= f_s \sum_{m = -L}^{L-1} \FrameIdx{X}{k} \! \left( \frac{n}{N} f_s - m f_s \right) \\
			&= f_s \sum_{m = -L}^{L-1} \exp \! \left( -j 2\pi \left( \frac{n}{N} f_s - m f_s \right) t_k \right) X \! \left( \frac{n}{N} f_s - m f_s \right).
	\end{split} 
\end{equation}
Following the derivation of Kim \etal \cite{Kim1990}, this condition is written in terms of the linear system:
\begin{equation}
	\label{eqn:02_sampling_aliasingEqSystem}
	\underbrace{
		\begin{pmatrix}
			\FrameIdx{\mathcal{Y}}{k}[1] \\ \FrameIdx{\mathcal{Y}}{k}[2] \\ \vdots \\ \FrameIdx{\mathcal{Y}}{k}[N]
		\end{pmatrix}
	}_{\FrameIdx{\vec{\mathcal{Y}}}{k}}
	=
	\underbrace{
		\begin{pmatrix}
			\FrameIdx{\vec{w}_1}{k}	& \vec{0} 								& \cdots 	& \vec{0}	\\
			\vec{0}				 					& \FrameIdx{\vec{w}_2}{k}	& \cdots	& \vec{0} \\
			\vdots 									& \vdots									& \ddots	& \vdots \\
			\vec{0}									&	\vec{0}									& \cdots	&	\FrameIdx{\vec{w}_N}{k} \\
		\end{pmatrix}
	}_{\FrameIdx{\vec{W}}{k}}
	\underbrace{
		\begin{pmatrix}
			X_{1,-L} \\ X_{1,-L+1} \\ \vdots \\ X_{N, L-1}
		\end{pmatrix}
	}_{\vec{X}},
\end{equation}
where $\FrameIdx{\vec{\mathcal{Y}}}{k} \in \ComplexN{N}$\label{notation:dftCoeff} comprises the \gls{dft} coefficients of the $k$-th channel, $\vec{X} \in \ComplexN{2LN}$\label{notation:cftCoeff} comprises the $2L$ samples of $X(f)$ abbreviated as $X_{n,m} \defeq f_s X(n/N f_s - m f_s)$ and $\FrameIdx{\vec{w}_n}{k} \defeq ( \FrameIdx{W_{n,-L}}{k}, \FrameIdx{W_{n,-L+1}}{k}, \ldots, \FrameIdx{W_{n,L-1}}{k} )$ are the row vectors containing the non-zero elements of the system matrix $\FrameIdx{\vec{W}}{k} \in \ComplexN{N \times 2LN}$. These elements can be computed according to:
\begin{equation}
	\label{eqn:02_sampling_systemMatrix} 
	\FrameIdx{W_{n,m}}{k}	= \exp \! \left( -j 2\pi \left( \frac{n}{N} f_s - m f_s \right) t_k \right). 
\end{equation}

The sampling process for a single channel with $N$ samples is modeled by $\FrameIdx{\vec{W}}{k}$, which is fully determined by the channel offset $t_k$ and the sampling frequency $f_s$. After concatenating the system matrices to $\vec{W} = ( \FrameIdx{\vec{W}}{1}, \ldots, \FrameIdx{\vec{W}}{K} )^\top \in \ComplexN{KN \times 2LN}$\label{notation:systemMatrixFourier} and the \gls{dft} coefficients to $\vec{\mathcal{Y}} = ( \FrameIdx{\vec{\mathcal{Y}}}{1}, \ldots, \FrameIdx{\vec{\mathcal{Y}}}{K} )^\top$, the aliasing property is given by the linear system:
\begin{equation}
	\label{eqn:02_sampling_aliasingEqSystemTotal}
	\vec{\mathcal{Y}} = \vec{W} \vec{X}.
\end{equation}
This linear system explains how the \gls{cft} of the continuous signal $x(t)$ is related to the \gls{dft} of the sampled channels $\FrameIdx{y}{1}[n], \ldots, \FrameIdx{y}{K}[n]$. Thus, it represents a generative model for multi-channel sampling in the Fourier domain. In order to reconstruct $x(t)$ from its sampled channels, the linear problem in \eref{eqn:02_sampling_aliasingEqSystemTotal} must be solved \wrt the \gls{cft} coefficients in $\vec{X}$. Note that, since $X(f)$ and $\FrameIdx{\mathcal{Y}}{k}[n]$ are complex-valued in general, \eref{eqn:02_sampling_aliasingEqSystemTotal} provides one condition for the real part and one for the imaginary part and thus $2 \cdot NK$ equations in total. However, for real-valued signals, $X(f)$ as well as $\FrameIdx{\mathcal{Y}}{k}[n]$ are symmetric \cite{Cooley1969} and the linear system provides $NK$ independent constraints. In this situation, it is sufficient to formulate the aliasing property for one half of the spectrum only.

\subsection{Real Multi-Channel Sampling}
\label{sec:02_MultichannelSamplingForRealKernel}

In order to model non-ideal sampling, the blur kernel for each channel is assumed to be linear and shift invariant in accordance to the analysis of single-channel sampling in \sref{sec:02_RealSampling}. Let us consider the general case of different blur kernels $\FrameIdx{h}{k}(t)$ associated with the set of channels. Then, the aliasing property introduced in \eref{eqn:02_sampling_aliasingPropertyInfty} can be formulated as:
\begin{equation}
	\begin{split}
		\FrameIdx{\mathcal{Y}}{k}[n] = f_s \sum_{m = -L}^{L-1} 
		&\exp \! \left( -j 2\pi \left( \frac{n}{N} f_s - m f_s \right) t_k \right) 
		\FrameIdx{H}{k} \! \left( \frac{n}{N} f_s - m f_s \right) \\
		& X \! \left( \frac{n}{N} f_s - m f_s \right),
	\end{split}
\end{equation}
where $\FrameIdx{H}{k}(f) = \mathcal{F} \{ \FrameIdx{h}{k}(t) \}$ is the \gls{cft} of the blur kernel associated with the $k$-th channel. Reformulating the aliasing property as a linear system yields the generalized version of \eref{eqn:02_sampling_aliasingEqSystemTotal}:
\begin{equation}
	\label{eqn:02_sampling_aliasingEqSystemTotalBlurKernel}
	\vec{\mathcal{Y}} = \vec{H} \vec{X}.
\end{equation}
The system matrix is assembled as \smash{$\vec{H} = (\FrameIdx{\vec{H}}{1}, \ldots, \FrameIdx{\vec{H}}{K})^\top \in \ComplexN{KN \times 2LN}$} according to the block structure in \eref{eqn:02_sampling_aliasingEqSystem} with the non-zero elements:
\begin{equation}
	\FrameIdx{H_{n,m}}{k} = \exp \! \left( -j 2\pi \left( \frac{n}{N} f_s - m f_s \right) t_k \right) \FrameIdx{H}{k} \left( \frac{n}{N} f_s - m f_s \right),
\end{equation}
This linear system states the relationship between the \gls{cft} of a continuous signal and the \gls{dft} of the corresponding discrete channels with consideration of a blur kernel.  

\section{From Multi-Channel Sampling to Super-Resolution}
\label{sec:02_FromMultichannelSamplingToSuperResolution}

The analysis of multi-channel sampling presented above yields a super-resolution algorithm in the frequency domain. This is the historically first multi-frame approach as proposed by Tsai and Huang \cite{Tsai1984}. In this framework, the target of super-resolution is the reconstruction of the continuous signal $x(t)$ from the sampled channels $\FrameIdx{y}{k}[n]$. However, for practical implementation purposes, the solution of this inverse problem is restricted to the reconstruction of discrete samples $x[n]$ using an artificial sampling frequency $f_s^\prime$ that should be higher than the original frequency $f_s$. Note that if $x[n]$ is determined such that the aliasing present in $\FrameIdx{y}{k}[n]$ is removed and thus sampled at the Nyquist rate, the continuous signal $x(t)$ can be recovered from $x[n]$. Therefore, $f_s^\prime$ must be chosen such that it fulfills the Nyquist-Shannon sampling theorem.

In order to define a super-resolution algorithm, the gain in resolution provided by the algorithm needs to be quantified. For this purpose, the \textit{magnification factor} denotes the enhancement of the sampling frequency achieved by super-resolution relative to the original sampling frequency. This parameter is defined as follows.
\begin{restatable}[Magnification factor]{definition}{magnificationFactor}
	Let $f_s$ be the sampling frequency that is used to obtain the discrete channels $\FrameIdx{y}{k}[n]$ and $f_s^\prime \geq f_s$ be the sampling frequency provided by a super-resolution algorithm. Then, the super-resolution magnification factor is given by:
	\begin{equation}
		s = \frac{f_s^\prime}{f_s}.
	\end{equation}
\end{restatable}
In a computational super-resolution approach, $s N$ discrete samples $x[n]$ associated with the continuous signal $x(t)$ are reconstructed from the sampled channels $\FrameIdx{y}{k}[n]$, $k = 1, \ldots, K$ consisting of $KN$ samples. In the Fourier domain formulation, $x[n]$ is represented by the $s N$ discrete frequency coefficients of $X(f)$. The key idea behind super-resolution is that the different channels $\FrameIdx{y}{k}[n]$ contain complementary information about $x(t)$. In summary, the main computational steps of super-resolution in the Fourier domain are as follows:
\begin{enumerate}
	\item \textbf{Fourier transform:} For all discrete channels $\FrameIdx{y}{k}[n]$, $k = 1, \ldots, K$ given by $NK$ samples, the associated \gls{dft} coefficients are determined efficiently by means of a \gls{fft} \cite{Cooley1965}. These coefficients form a complex-valued observation vector $\vec{\mathcal{Y}} \in \ComplexN{KN}$.
		
	\item \textbf{Reconstruction:} Once the offsets $t_k$ for all channels are known, the samples of the \gls{cft} $X(f)$ reorganized in $\vec{X} \in \ComplexN{2L \cdot KN}$ for $s = 2L$ are determined for the magnification factor $s$. This is done by solving the linear systems in \eref{eqn:02_sampling_aliasingEqSystemTotal} (in case of ideal sampling) or in \eref{eqn:02_sampling_aliasingEqSystemTotalBlurKernel} (in case of real sampling).

	\item \textbf{Inverse Fourier transform:} Finally, the discrete samples $x[n]$ associated with the continuous signal $x(t)$ are recovered from $\vec{X}$ using an inverse \gls{fft}.
\end{enumerate}

Existing frequency domain based super-resolution methods mainly differ in the implementation of the second step. In \cite{Tsai1984}, the most basic situation of ideal sampling and the absence of observation noise is considered. This is a simplistic assumption, particularly in digital imaging, where an optical system acts as a blur kernel in the sampling process and observation noise is caused by non-ideal sensors. Kim \etal \cite{Kim1990} have proposed a generalization of the Tsai and Huang algorithm, where observation noise is taken into account. In this case, an estimate for $\vec{X}$ can be obtained by solving the linear problem given in \eref{eqn:02_sampling_aliasingEqSystemTotal} in a least square manner. Later, Tekalp \etal \cite{Tekalp1992a} have introduced a frequency domain approach that considers observation noise as well as a blur kernel involved in the sampling process. Super-resolution is then based on the more general linear problem given in \eref{eqn:02_sampling_aliasingEqSystemTotalBlurKernel}.

\section{Limits of Super-Resolution}
\label{sec:02_LimitsOfSuperResolution}

This section analyzes the linear problems in \eref{eqn:02_sampling_aliasingEqSystemTotal} and \eref{eqn:02_sampling_aliasingEqSystemTotalBlurKernel} in case of ideal and non-ideal sampling, respectively. The properties of these relationships between discrete channels and the underlying continuous signal indicate if super-resolution is applicable and how well the behavior of an algorithm can be. 

The following analysis covers two aspects. First, it is examined under which conditions super-resolution is profitable in order to reconstruct an aliasing-free signal from undersampled ones. This states that under certain conditions super-resolution cannot provide resolution enhancement beyond a given limit. In particular, upper bounds regarding the effective magnification factor that depends on the sampling parameters are derived. Second, necessary and sufficient conditions for the existence of unique solutions for \eref{eqn:02_sampling_aliasingEqSystemTotal} and \eref{eqn:02_sampling_aliasingEqSystemTotalBlurKernel} are studied. This states inherent theoretical limitations for super-resolution in the presence of degenerate situations, where the underlying linear system is underdetermined.

\subsection{Effective Magnification Factor}
\label{sec:02_EffectiveMagnificationFactor}

We are interested in an effective sampling frequency $f^*$\label{notation:effSamplingFreq} that can be achieved by means of super-resolution. In this context, the term \textit{effective} means that super-resolution cannot recover aliased parts of the spectrum beyond $f^*$, which states an upper bound of the artificial sampling frequency as well as an effective magnification factor $s^*$\label{notation:effMagFac}. Applying super-resolution beyond this limit does not yield additional gains compared to a simple interpolation of discrete samples. The key idea regarding the following analysis is that even if aliasing is usually considered as undesirable effect, it is a prerequisite for super-resolution. This is due to the fact that super-resolution exploits aliased components in undersampled signals according to \eref{eqn:02_sampling_aliasingPropertyInfty}. We examine this property for two different situations.

\paragraph{Magnification Factor for Ideal Sampling.}
Let us first consider ideal sampling of $x(t)$ with band-limitation $f_0$ below the Nyquist rate, \ie $f_s \geq 2 f_0$. Thus, there is no aliasing present in the sampled channels $\FrameIdx{y}{k}[n]$. In this case the infinite summation in \eref{eqn:02_sampling_aliasingPropertyInfty} comprises only of one non-zero term, as there are no superimpositions of periodically shifted versions $\FrameIdx{X}{k}(f)$. Then, it follows for the aliasing property:
\begin{equation}
	\FrameIdx{\mathcal{Y}}{k}[n] = \FrameIdx{X}{k} \! \left( \frac{n}{N} f_s \right),
\end{equation}
for all $k = 1, \ldots, K$. That is, each \gls{dft} $\FrameIdx{\mathcal{Y}}{k}[n]$ is a sampled version of the corresponding \gls{cft} without loss of information as the sampling theorem is fulfilled. It is obvious that a solution of the linear system  in \eref{eqn:02_sampling_aliasingEqSystemTotal} cannot recover additional information. Consequently, it follows for the effective sampling frequency $f^* = f_s$. Moreover, it is obvious that super-resolution cannot reconstruct frequencies beyond the band-limitation $f_0$. By combining these findings, the effective sampling frequency that can be achieved by super-resolution is given by:
\begin{equation}
	f^* = 
	\begin{cases}
		f_s			& \text{if}~ f_s \geq 2 f_0 \\
		2 f_0		& \text{otherwise}.
	\end{cases}
\end{equation}
This yields an upper bound regarding the effective magnification factor:
\begin{equation}
	s^* = 
	\begin{cases}
		1										& \text{if}~ f_s \geq 2 f_0 \\
		2\frac{f_0}{f_s}		& \text{otherwise}.
	\end{cases}
\end{equation}

\paragraph{Magnification Factor for Real Sampling.}
If real sampling is considered, the presence of the blur kernel $\FrameIdx{h}{k}(t)$ limits the effective magnification factor. As derived in \sref{sec:02_RealSampling}, $\FrameIdx{h}{k}(t)$ acts as a low-pass filter for the channel $\FrameIdx{x}{k}(t)$. Unfortunately, this blur kernel also performs anti-aliasing depending on its cut-off frequency $f_h$. If the cut-off frequency is above the band-limitation, \ie $f_h \geq f_0$, the blur kernel does not affect the sampling process. However, if $f_h < f_0$, spectral components of $\FrameIdx{X}{k}(f)$ affected by aliasing are suppressed by the blur kernel. In the worst case where $f_h < f_0 - f_s$, aliasing is fully removed. However, these are exactly the signal components exploited by super-resolution. Hence, the cut-off frequency $f_h$ limits the effective sampling frequency to: 
\begin{equation}
	f^* = 
	\begin{cases}
		\min(f_s, f_h)			& \text{if}~ f_s \geq 2 f_0 \\
		\min(2 f_0, f_h)		& \text{otherwise}.
	\end{cases}
\end{equation}
Note that $f_h$ is now an upper bound for the sampling frequency $f^*$ that can be smaller than the original sampling frequency $f_s$. The effective sampling frequency $f^*$ yields an upper bound regarding the magnification factor: 
\begin{equation}
	s^* = 
	\begin{cases}
		\frac{1}{f_s} \min(f_s, f_h)			& \text{if}~ f_s \geq 2 f_0 \\
		\frac{1}{f_s} \min(2 f_0, f_h)		& \text{otherwise}.
	\end{cases}
\end{equation}

This barrier needs to be considered in digital imaging, where the optical \gls{psf} and the finite size of pixels on the sensor array act as a low-pass filter. This has the consequence that super-resolution cannot provide effective magnifications beyond the system band-limitation related to these properties. In \cref{sec:ComputationalFrameworkForMultiFrameSuperResolution} and \ref{sec:RobustMultiFrameSuperResolutionWithSparseRegularization}, we study different regularization techniques in conjunction with super-resolution reconstruction to alleviate this limitation in practical applications.   

\subsection{Uniqueness of the Reconstruction}
\label{sec:02_SingularityOfSuperResolution}

The uniqueness of super-resolution based on the linear problems in \eref{eqn:02_sampling_aliasingEqSystemTotal} in case of ideal sampling and \eref{eqn:02_sampling_aliasingEqSystemTotalBlurKernel} in case of real sampling depends on several parameters of the sampling process. For this analysis, a reconstruction is called \textit{unique} iff the associated system matrix involved in the linear problem is non-singular. In this case, the complementary information encoded by multiple channels is sufficient to provide a super-resolved signal. If the system matrix is singular, super-resolution becomes underdetermined and does not enable a unique reconstruction. In general, non-uniqueness might be caused by degenerate settings in terms of the sampling parameters resulting in a major limitation of super-resolution in practical applications. 

This sections analyzes the uniqueness of the underlying inverse problem and derives conditions for a unique reconstruction. These derivations lead to two relevant classes of super-resolution algorithms applicable in digital imaging.

\paragraph{Uniqueness for Ideal Sampling.}
The uniqueness of \eref{eqn:02_sampling_aliasingEqSystemTotal} is first studied for ideal sampling. For this purpose, the channel offsets are used as a cue to provide complementary information and to guarantee a unique reconstruction. Let us study the case that the magnification factor is given by $s = K$ for $K$ channels. If we consider the real and imaginary parts of the complex-valued Fourier transforms in \eref{eqn:02_sampling_aliasingEqSystemTotal}, the system matrix is quadratic and an exact solution of this inverse problem can be obtained\footnote{For $s < K$, an approximation can be obtained by means of least-squares estimation \cite{Tekalp1992a}.}. The conditions regarding uniqueness of super-resolution in this situation are summarized in the following theorem.
\begin{restatable}[Uniqueness for ideal sampling]{thm}{uniqunessIdealSampling}
	Let $s = K$ be the super-resolution magnification factor and $K$ be the number of channels in a multi-channel sampling process, where $t_i$ with $i = 1, \ldots, K$ and $t_1 = 0$ are the corresponding channel offsets and $T$ is the sampling pitch. Then, the solution of the linear inverse problem in \eref{eqn:02_sampling_aliasingEqSystemTotal} is unique if and only if:
	$t_j \neq c_1 t_i + c_2 T ~\text{for all}~ 1 \leq i < j \leq K ~\text{and}~ c_1, c_2 \in \Integer$.
	\label{theorem:sampling_uniqunessIdealSampling}
	\label{theo:02_uniqunessIdealSampling}
\end{restatable}
\begin{proof}
	The proof of this theorem is given in \appref{sec:A_UniquenessForIdealSampling}.
\end{proof}
These conditions provide an intuitive approach to perform super-resolution reconstruction. The distinct and non-integer offsets \wrt the sampling pitch enable a non-uniform sampling at a higher frequency compared to single-channel sampling. Thus, a solution of \eref{eqn:02_sampling_aliasingEqSystemTotal} can be seen as a fusion of the complementary information encoded by single channels, see \fref{fig:02_multichannelSampling}. In fact, choosing distinct and non-integer channel offsets are necessary and sufficient conditions for a unique reconstruction. This is a popular strategy for resolution enhancement in digital imaging, where channel offsets can be related to subpixel displacements among multiple images \cite{Park2003}. The offsets required for this \textit{motion-based} super-resolution can be provided by capturing a set of images of the underlying scene while moving a camera to slightly different viewpoints.

\paragraph{Uniqueness for Real Sampling.}
\theoref{theorem:sampling_uniqunessIdealSampling} states necessary conditions for a unique reconstruction if only the channel offsets are exploited. However, in case of real sampling according to \eref{eqn:02_sampling_aliasingEqSystemTotalBlurKernel}, the blur kernel can be used as cue to achieve uniqueness even in the absence of channel offsets. The following theorem summarizes the conditions to achieve a unique reconstruction in this situation.
\begin{restatable}[Uniqueness for real sampling]{thm}{uniqunessRealSampling}
	Let $s = K$ be the super-resolution magnification factor and $K$ be the number of channels in multi-channel sampling with offsets $t_i = 0$ for all $i = 1, \ldots, K$ and sampling pitch $T$. Each channel $\FrameIdx{x}{i}(t)$ is affected by a blur kernel $\FrameIdx{H}{i}(f)$ denoted by $\FrameIdx{\vec{H}}{i}$ in matrix notation. Then, the solution of the linear inverse problem in \eref{eqn:02_sampling_aliasingEqSystemTotalBlurKernel} is unique if and only if:
	\begin{enumerate}
		\item $\sum_{i = 1}^K c_i \FrameIdx{\vec{H}}{i} \neq \vec{0}$ ~for all~ $c_i \neq 0$ ~and~ $i = 1, \ldots, K$
		(linear independent blur kernels)
		\item $\sum_{i = 1}^K \left| \FrameIdx{H}{i} \! \left(\frac{n}{N} f_s + m f_s \right) \right| \neq 0$ for all $m = -L, \ldots, L-1$
		(kernel cut-off frequency)
	\end{enumerate}
	\label{theo:02_uniqunessRealSampling}
\end{restatable}
\begin{proof}
	The proof of this theorem is given in \appref{sec:A_UniquenessForRealSampling}.
\end{proof}
Besides the use of distinct, non-integer channel offsets to make super-resolution reconstruction unique, an alternative approach is to exploit the properties of the blur kernel. Complementary information required for super-resolution is gained by utilizing independent kernels for multiple channels. Moreover, the kernel cut-off frequency needs to be above the Nyquist rate and at least one kernel needs to span the entire frequency range that should be super-resolved. In this situation, a unique solution can be provided according to \theoref{theo:02_uniqunessRealSampling}. This approach has been widely studied in digital imaging. In \cite{Elad1997}, Elad and Feuer investigated \textit{motion-free} spatial domain super-resolution, which shows that this approach is feasible. Rajagopalan and Kiran \cite{Rajagopalan2003a} proposed a Fourier domain method to perform super-resolution reconstruction from multiple defocused images corresponding to varying levels of blur in a set of channels.

\section{Conclusion}
\label{sec:02_Conclusion}

This chapter presented single- and multi-channel sampling as theoretical framework for super-resolution. This theory considered ideal sampling as well as real sampling in the presence of a blur kernel. Super-resolution was formulated as linear inverse problem that states the relation between a continuous signal and discrete channels that are captured by sampling the continuous signal multiple times. In this context, super-resolution aims at reconstructing an aliasing-free signal from multiple undersampled channels.

In order to derive fundamental limits of super-resolution, the properties of the underlying inverse problem were analyzed. First, the relationship between super-resolution and the Nyquist-Shannon sampling theorem was discussed. It was shown that the effective magnification, achievable by super-resolution, is bounded by the band-limitation of the continuous signal as well as the cut-off frequency of the blur kernel in case of real sampling. Second, the uniqueness of super-resolution reconstruction was examined. This analysis shows that super-resolution requires complementary information in multiple channels to provide a unique solution. The necessary and sufficient conditions to gain complementary information involve properties of the channel offsets or the blur kernel.

\part{Numerical Methods for Multi-Frame Super-Resolution}
\label{sec:NumericalMethodsForMultiFrameSuperResolution}

\chapter{Computational Framework for Multi-Frame Super-Resolution}
\label{sec:ComputationalFrameworkForMultiFrameSuperResolution}

\myminitoc

\noindent
This chapter presents the common computational framework that is employed in the remainder of this thesis. This introduction comprises three parts. First, a literature survey on different paradigms in the field of super-resolution is presented including a review of state-of-the art frequency domain and spatial domain algorithms. Second, a spatial domain model for optical imaging is derived. Third, based on generative modeling, multi-frame super-resolution is approached from a Bayesian perspective and different parameter estimation schemes are presented.

\section{Introduction and Literature Survey}

In \cref{sec:MultiFrameSuperResolutionAndTheSamplingTheorem}, super-resolution was introduced as a multi-channel signal reconstruction problem. A continuous signal was assumed to be sampled by multiple channels to obtain a set of discrete signals. In case of linearly independent offsets among the channels, each of these discrete signals contains complementary information about the underlying continuous one. In terms of \textit{motion-based} super-resolution as the major scope of this work, channel offsets are explained by subpixel displacements across multiple images showing the same scene from slightly different perspectives. Mathematically, this motion is described by image-to-image transformations on the image plane and can be induced by camera motion, object motion, or a combination of both. Given a sequence of undersampled images along with their associated subpixel motion, the goal of image super-resolution is to obtain a high-resolution image from low-resolution ones. This reconstruction is unique if the channel offsets related to the subpixel motion are independent and distinct to multiples of the sampling rate, see \sref{sec:02_SingularityOfSuperResolution}. 

Let us first present a survey on super-resolution paradigms including frequency and spatial domain methods, see \fref{fig:srAlgorithmsClassification}. For a more comprehensive overview, we refer to the review articles by Park \etal \cite{Park2003}, Farsiu \etal \cite{Farsiu2004}, and Nasrollahi and Moeslund \cite{Nasrollahi2014}, as well as the book by Milanfar \cite{Milanfar2010}.

\begin{figure}[!t]
	\centering
		\includegraphics[width=1.00\textwidth]{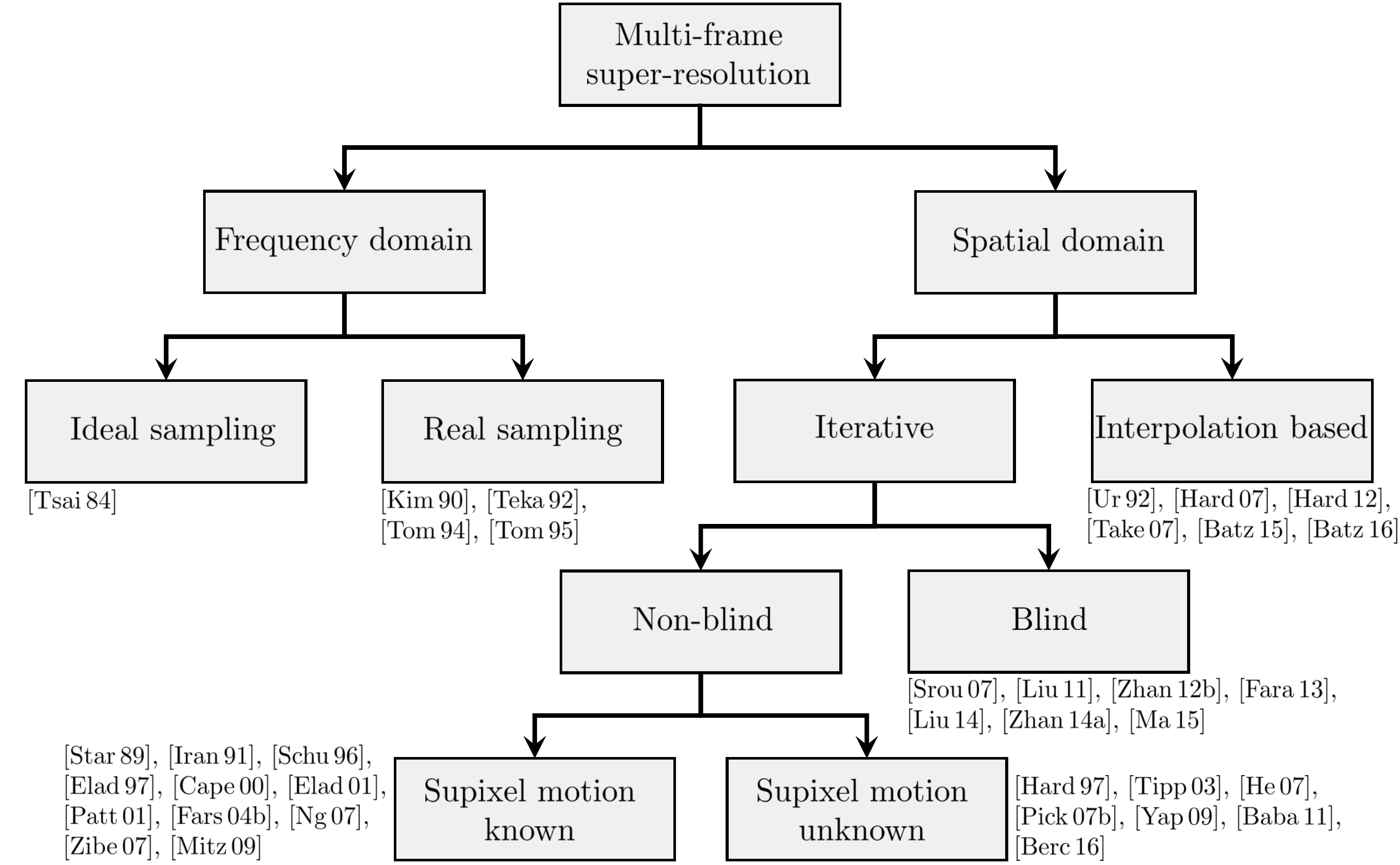}
	\caption[Classification of multi-frame super-resolution algorithms]{Classification of multi-frame super-resolution algorithms including references to seminal works and some of the most recent publications in the different domains.}
	\label{fig:srAlgorithmsClassification}
\end{figure}

\subsection{Frequency Domain Reconstruction}

Early approaches describe super-resolution reconstruction in the frequency domain based on the multi-channel sampling theory presented in \sref{sec:02_MultichannelSamplingTheory}. In this context, the method of Tsai and Huang \cite{Tsai1984} employs the Fourier shift theorem to exploit translational subpixel motion in a sequence of low-resolution images. This yields a generative model for super-resolution in the Fourier domain. Translational motion is first determined by means of image registration, which can be performed in the Fourier domain using phase correlation. Then, super-resolution is implemented as inversion of the underlying model equations parametrized by the estimated motion, see \sref{sec:02_FromMultichannelSamplingToSuperResolution}. In the approaches of Kim \etal \cite{Kim1990} and Tekalp \etal \cite{Tekalp1992a}, this concept has been further extended to tackle sensor noise as well as blurring in the image formation. Tom \etal \cite{Tom1994,Tom1995} have proposed simultaneous super-resolution and translation estimation in the Fourier domain. In \cite{Rhee1999}, Rhee and Kang have employed the \gls{dct} for super-resolution as alternative to the Fourier transform.  

The frequency domain formulation provides valuable theoretical insights to super-resolution and the use of the \gls{fft} as a computational tool enables efficient implementations of these algorithms. However, only simple motion models can be used. For instance, the Fourier shift theorem in \cite{Tsai1984} enables the description of translational motion but cannot model arbitrary displacements. In particular, it is not feasible to handle non-rigid motion. Additionally, blur needs to be described by \gls{lsi} kernels to be tractable by means of the Fourier transform. This restricts the flexibility of these algorithms in terms of the underlying image formation model. 

\subsection{Interpolation-Based Spatial Domain Reconstruction}

Spatial domain reconstruction can be seen as a complementary trend in the design of super-resolution algorithms. These methods have the goal to enhance the flexibility regarding the choice of the image formation model. Ur and Gross \cite{Ur1992} proposed interpolation-based reconstruction based on the multi-channel sampling theory of Papoulis \cite{Papoulis1977} that has later been adopted in other spatial domain methods \cite{Alam2000,Pham2006a,Hardie2007,Hardie2012,Takeda2007,Batz2015}. Such algorithms differ in their implementations but share a similar conceptual structure. Their concept is described by a multi-stage procedure with the following steps, see \fref{fig:02_interpolationScheme}:
\begin{enumerate}
	\item Motion estimation by means of image registration determines the subpixel motion between multiple low-resolution frames.
	\item Motion compensation transforms all low-resolution frames into a common high-resolution grid according to the motion estimate.
	\item Non-uniform interpolation determines a super-resolved image based on the motion-compensated frames.
\end{enumerate}
Some of the well known methods in this area include normalized convolution \cite{Pham2006a}, kernel regression \cite{Takeda2007}, and adaptive Wiener filtering \cite{Hardie2007, Hardie2012}.  More recently, adaptive weighting schemes \cite{Batz2016} as well as hybrid multi-frame and single-image reconstruction \cite{Batz2015} have been proposed to alleviate the impact of motion estimation inaccuracies. These schemes can also be augmented by image deblurring \cite{Patrizio2016} to remove blur after the interpolation.

A closely related class of algorithms employs \textit{deep learning} that became a popular alternative to classical interpolation-based frameworks. These methods learn parts of the multi-stage procedure in \fref{fig:02_interpolationScheme} from pairs of low-resolution and high-resolution images. For instance, non-uniform interpolation \cite{Kappeler2016,Li2017} or motion compensation \cite{Tao2017} can be learned via convolutional neural networks.

\begin{figure}[!t]
	\centering
		\includegraphics[width=1.00\textwidth]{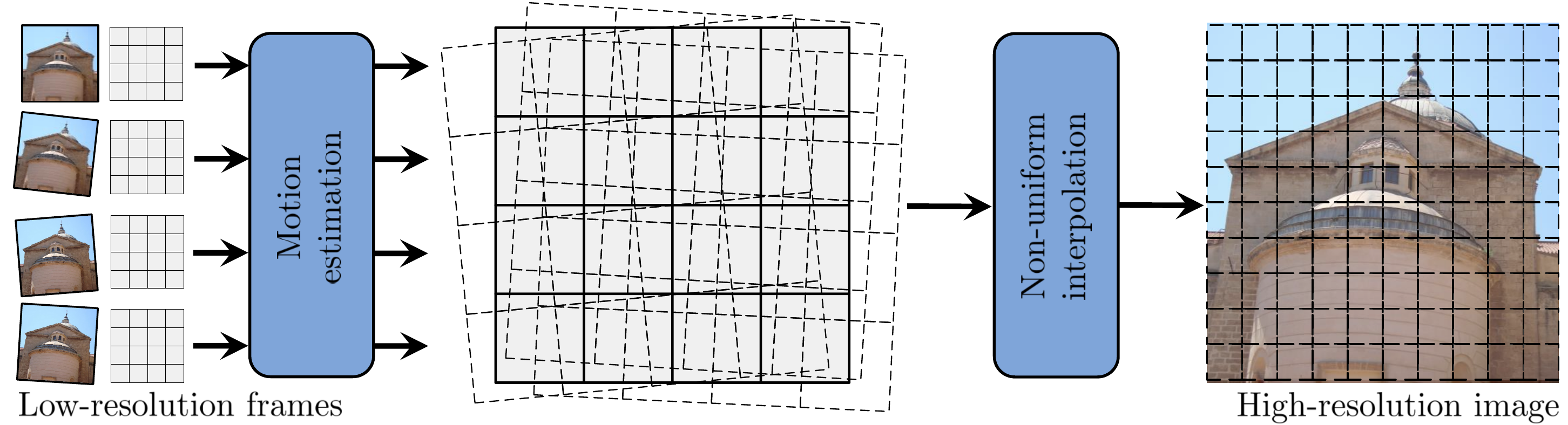}
	\caption[Spatial domain super-resolution using non-uniform interpolation]{Spatial domain super-resolution using non-uniform interpolation. First, subpixel motion between multiple low-resolution frames is estimated and compensated. Then, the super-resolved image is interpolated from the motion-compensated frames.}
	\label{fig:02_interpolationScheme}
\end{figure}

Unlike the aforementioned frequency domain methods that utilize a generative model, non-uniform interpolation aims at a direct reconstruction without considering such a model. These approaches can be implemented efficiently in terms of computational complexity. Moreover, the motion compensation is flexible and can be applied under various types of subpixel motion. However, due to the sequential design, errors in one stage are propagated to following stages, \eg from motion estimation to interpolation. This might lead to suboptimal reconstructions in terms of a global quality criterion \cite{Park2003}. It is also difficult to model prior knowledge regarding the appearance of super-resolved images.

\subsection{Iterative Spatial Domain Reconstruction}

The vast majority of the state-of-art algorithms as well as the methods investigated in this work are formulated as iterative spatial domain reconstructions. In the same way as for the frequency domain methods, a generative model to describe the image formation is utilized. However, this model is formulated in the spatial domain to increase its flexibility. The basic idea is to iteratively refine an estimate for a super-resolved image such that it best explains the observed low-resolution data under the generative model. In literature, this concept has been first formalized in the \textit{iterated backprojection} algorithm proposed by Irani and Peleg \cite{Irani1991}. This approach is illustrated in \fref{fig:02_iterativeRecoScheme} and we consider two realizations.

\begin{figure}[!t]
	\centering
		\includegraphics[width=0.99\textwidth]{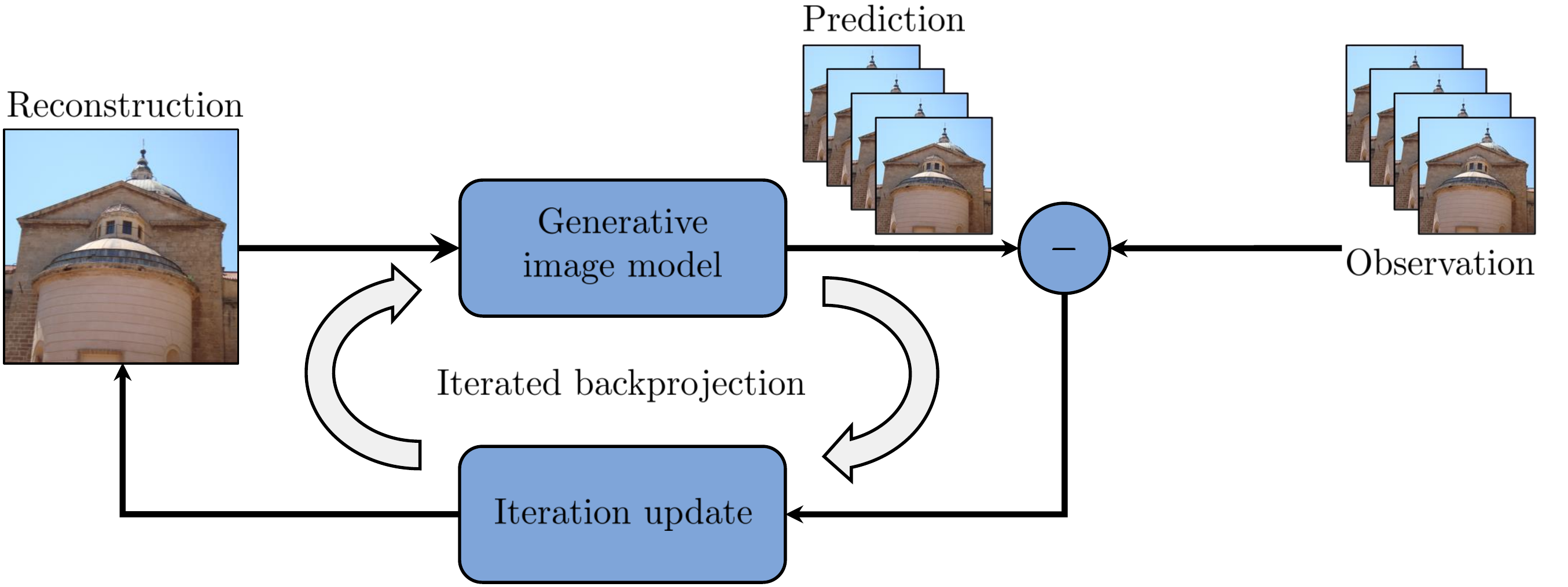}
	\caption[Spatial domain super-resolution using iterated backprojection]{Spatial domain super-resolution using iterated backprojection \cite{Irani1991}. The deviation between the acquired low-resolution observations and a prediction obtained from the current reconstruction under a generative image model is iteratively optimized.}
	\label{fig:02_iterativeRecoScheme}
\end{figure}

\paragraph{Non-Blind Reconstruction.}
In \textit{non-blind} super-resolution, it is assumed that the parameters of the generative model are known a priori. In particular, it is assumed that the \gls{psf} is known either by system calibration, by automatic parameter selection \cite{Nguyen2001}, or by simply modeling it empirically with a realistic blur kernel.

One important class of these algorithms approaches super-resolution from a Bayesian statistics point of view that has also become a common tool in image denoising \cite{Chen2007}, restoration \cite{Besag1991,Bioucas-Dias2006}, and single-image expansion \cite{Schultz1994a}. In \cite{Elad1997}, Elad and Feuer have proposed \gls{ml} estimation that has been used as a probabilistic framework  in several follow-up works \cite{Capel2000,Elad2001,Zibetti2007,Mitzel2009a}. The subpixel motion that is exploited for this point estimation is determined via registration of low-resolution frames. Then, the most probable high-resolution image associated with the subpixel displaced frames is reconstructed under a generative model. \Gls{map} estimation generalizes the \gls{ml} approach by exploiting prior knowledge to regularize super-resolution. For this purpose, a Gaussian prior distribution to model the statistical appearance of images has been introduced in \cite{Elad1997}. Later, other priors have been proposed \cite{Schultz1996,Farsiu2004a,Ng2007} that can better model the characteristics of natural images. These probabilistic models lead to energy minimization problems that can be solved iteratively. For details on these Bayesian methods, we refer to \sref{sec:02_BayesianModelingOfSuperResolution}. A closely related technique is the \gls{pocs} \cite{Stark1989,Patti2001}. \gls{pocs} methods formulate prior knowledge by set theoretic constraints as opposed to probability distributions. Super-resolution is performed by iterative projections under these constraints.
 
Contrary to algorithms that estimate subpixel motion prior to super-resolution, there are also methods that treat the motion as hidden information. In a seminal work, Hardie \etal \cite{Hardie1997} proposed \textit{joint} \gls{map} estimation for both parameter sets based on alternating minimization. This avoids motion estimation on low-resolution data, which is error-prone due to undersampling \cite{Vandewalle2006}. Notice that many algorithms that are formulated in the fashion of iterative spatial domain reconstruction, \eg Gauss-Newton \cite{He2007,Bercea2016} or linear programming \cite{Yap2009} schemes, fall into this category even if they are not explicitly derived in Bayesian frameworks. A related approach is to make use of Bayesian marginalization in the absence of a proper motion estimate. In \cite{Tipping2003}, Tipping and Bishop proposed marginalization over the domain of high-resolution images. A different approach has been developed by Pickup \etal \cite{Pickup2007}, where marginalization is performed over motion parameters to integrate them out from super-resolution reconstruction. As another technique in the field of Bayesian statistics, variational inference \cite{Babacan2011} has proven to be a valuable tool. As opposed to the \gls{ml} and \gls{map} schemes that provide point estimates, variational inference aims at determining full posterior probability distributions. This enables a joint estimation of the super-resolved image along with latent model parameters.   

\paragraph{Blind Reconstruction.}
In terms of \textit{blind} super-resolution, the blur kernel related to the camera \gls{psf} is assumed to be unknown, and so are the parameters of the generative image model. This class of algorithms is closely related to blind deconvolution \cite{Patrizio2016} and treats the blur as a latent variable. Blind super-resolution is commonly implemented as an interlacing of non-blind reconstruction and blur estimation in joint optimization frameworks.

Super-resolution under an unknown \gls{psf} has been investigated by Sroubek \etal \cite{Sroubek2007} and Faramarzi \etal \cite{Faramarzi2013}. Such methods combine multi-frame resolution enhancement and blind deconvolution to unified frameworks. These are formulated via iterative spatial domain reconstruction with known subpixel motion. Later, Zhang \etal \cite{Zhang2012} as well as Liu and Sun \cite{Liu2011,Liu2014} have proposed to treat blur, subpixel motion and the latent high-resolution image as triple coupled variables to determine them simultaneously. This circumvents a direct motion estimation on low-resolution frames. More recently, the handling of motion blur has been studied for situations where optical and sensor blur are no appropriate models, \eg fast camera-shake \cite{Zhang2014a,Ma2015}. 

\section{Modeling the Image Formation Process}
\label{sec:03_ModelingTheImageFormationProcess}

One of the key components of multi-frame super-resolution is an appropriate mathematical modeling of image formation implemented by a digital imaging system. This section describes a spatial domain model that is widely applicable and employed to develop the super-resolution algorithms presented in the remainder of this work. This model can be seen as a spatial domain analog to the Fourier domain model derived in \cref{sec:MultiFrameSuperResolutionAndTheSamplingTheorem}.

\subsection{Continuous Image Formation Model}

The image formation model used in this thesis is based on the work of Elad and Feuer \cite{Elad1997} that has later been extended by Capel and Zisserman \cite{Capel2003,Capel2004}. In recent years, this model has been utilized for the vast majority of super-resolution algorithms. It describes the physics of image acquisition in a forward process to explain how a digital image is obtained from a real-world scene.

For the derivation from a continuous point of view, let $\HRSym: \RealN{2} \rightarrow \Real$ be the \textit{irradiance light field} \cite{Lin2004} obtained by an ideal projection of a 3-D scene onto the 2-D image plane of a digital camera. Mathematically, this projection can be described by a pinhole camera \cite{Hartley2004}. Since we limit ourselves to single-channel images, $\HRFun{\Point}$\label{notation:hrFun} denotes an intensity at the 2-D position $\Point \in \RealN{2}$\label{notation:point}. Using the assumption of an ideal projection to the image plane, $\HRFun{\Point}$ can be seen as a ground truth and is referred to as \textit{ideal image} \cite{Hardie2007}. In particular, as no sampling is modeled and $\HRFun{\Point}$ is given as a continuous irradiance signal, it can be considered as a signal of infinite spatial resolution. In terms of an imaging system that acquires video data, one observes a set of $\NumFrames$\label{notation:numFrames} degraded frames given as continuous functions $\LRFrameFun{\Point}{1}, \ldots, \LRFrameFun{\Point}{\NumFrames}$\label{notation:lrFun}\label{notation:frameIdx} associated with the ideal image $\HRFun{\Point}$. In order to describe the image formation process mathematically, the following operations are analyzed.

\paragraph{Motion Model.} 
The ideal image $\HRFun{\Point}$ is assumed to be warped by a geometric transformation \wrt a certain coordinate reference for each frame to describe the acquisition of an image sequence. This geometric transformation encodes camera motion, object motion or a combination of both. For the sake of convenience, motion is described on the 2-D image plane instead of describing it by a 3-D transformation. The $k$-th warped version of $\HRFun{\Point}$ denoted by $\HRFrameFun{\Point}{k}$ is given by:
\begin{equation}
	\label{eqn:03_contMotionModel}
	\HRFrameFun{\Point}{k} = \MotionOpFrame{x(\Point)}{k},
\end{equation}
where $\MotionOpFrameEmpty{k} \{ \cdot \}$\label{notation:motionOpFrame} denotes the motion model for the $k$-th frame. Without any further assumption on the type of motion, this model is described by:
\begin{equation}
	\MotionOpFrame{\HRFun{\Point}}{k} \defeq x \big( \Point + \DispVecFieldFrame{\Point}{k} \big),
\end{equation}
where $\DispVecFieldFrameSym{k}: \RealN{2} \rightarrow \RealN{2}$\label{notation:dispField} denotes the displacement of $\HRFun{\Point}$ at position $\Point$ relative to the reference frame. For a given reference frame $\HRFrameFun{\Point}{r}$ with $r \in \{1, \ldots, \NumFrames\}$, $\MotionOpFrameEmpty{r}$ is the identity and $\HRFrameFun{\Point}{r}$ coincides with $\HRFun{\Point}$.
	
\paragraph{Sampling Model.} 
Next, we need to describe the sampling process that explains how the imaging system discretizes the ideal image. In the most basic formulation of the image formation model, we limit ourselves to two common aspects. 

First, each frame is affected by the \gls{psf} of the imaging system. This \gls{psf} is the impulse response of the entire system and describes how an ideal point object is captured on the image plane. We assume a \gls{lsi} model that is characterized by a low-pass filter and define the \gls{psf} by a blur kernel $\BlurKernel{\Point}$. Then, a blurred version $\tilde{x}^{(k)}(\Point)$ of the warped image $\HRFrameFun{\Point}{k}$ is obtained by the convolution:
\begin{equation}
	\begin{split}
		\FrameIdx{\tilde{\HRSym}}{k}(\Point)	
			&= \HRFrameFun{\Point}{k} \conv \BlurKernelFrame{\Point}{k} \\
			&= \int_{\RealN{2}} \HRFrameFun{\PointV}{k} \BlurKernelFrame{\Point - \PointV}{k} d \PointV.
	\end{split}
\end{equation}
Note that we assume the general case of a time variant blur kernel, \ie $\BlurKernelFrame{\Point}{k}$\label{notation:blurKernel} can be different for each frame. Following the analysis in \cite{Baker2002}, this blur kernel can be decomposed according to:
\begin{equation}
	\BlurKernelFrame{\Point}{k} = \big( \BlurKernelFrameSym[optics]{k} \, \conv \, \BlurKernelSym[sensor] \big) (\Point),
\end{equation}
where \smash{$\BlurKernelFrame[optics]{\Point}{k}$} models time variant blur caused by optical effects and $\BlurKernel[sensor]{\Point}$ describes the time invariant blur caused the integration of light over a finite area on the sensor array corresponding to a pixel of the detector. 
	
Finally, each frame is discretized on the sensor array that is modeled by two operations. First, each frame is sampled on the center positions of rectangular pixels. Since the integration of light on the sensor array is included in the \gls{psf} model, we describe the sampling by ideal Dirac impulses. Second, the sampled frame is disturbed by random measurement noise caused by an imperfect sensor. Mathematically, the sensor array is described by:
\begin{equation}
	\LRFrameFun{\Point}{k} = \SamplingOp{ \FrameIdx{\tilde{\HRSym}}{k}(\Point) } + \NoiseFrame{\Point}{k},
\end{equation}
where $\SamplingOp{\cdot}$\label{notation:samplingOp} denotes the sampling at the pixel positions and $\NoiseFrame{\Point}{k}$\label{notation:noiseSignal} is a stochastic signal to model measurement noise. Assuming a pixel pitch $\Delta i$ in coordinate direction $i \in \{\CoordU, \CoordV\}$, the sampling operator is given by:
\begin{equation}
	\SamplingOp{\tilde{x}^{(k)}(\Point)} \defeq \sum_{m = -\infty}^\infty \sum_{n = -\infty}^\infty 
	\FrameIdx{\tilde{\HRSym}}{k}(\CoordU, \CoordV) \delta(\CoordU - m \Delta \CoordU, \CoordV - n \Delta \CoordV),
\end{equation}
where $\delta(\Point)$ denotes the 2-D Dirac delta impulse.

\paragraph{Joint Motion and Sampling Model.} 
The different physical effects that occur when acquiring a digital image from a real-world scene are combined in sequential order, see \fref{fig:02_imageFormationModel}. In summary, the $k$-th frame $\LRFrameFun{\Point}{k}$ out of a set of $\NumFrames$ frames is related to the ideal image $\HRFun{\Point}$ according to:
\begin{equation}
	\label{eqn:03_contImageModel}
	\LRFrameFun{\Point}{k} =
	\SamplingOp{
		\MotionOpFrame{\HRFun{\Point}}{k}
		\conv
		\BlurKernelFrame{\Point}{k}
	}
	+ \NoiseFrame{\Point}{k}.
\end{equation}
This model is appropriate if the optical components and the sensor array have a dominant influence to the image formation while atmospheric effects need to be negligible. As shown in the following section, this scheme can be efficiently discretized to make it applicable for super-resolution. 

Notice that there also exist related approaches with the order of the operations in \eref{eqn:03_contImageModel} is reversed. One example is the blur-warping formulation with a reversed order of the motion and blur operators, which is superior to \eref{eqn:03_contImageModel} in case of a high uncertainty regarding the motion as studied by Wang and Qi \cite{Wang2004a}.
\begin{figure}[!t]
	\captionsetup[subfloat]{labelformat=parens}
	\centering  
	\includegraphics[width=0.975\textwidth]{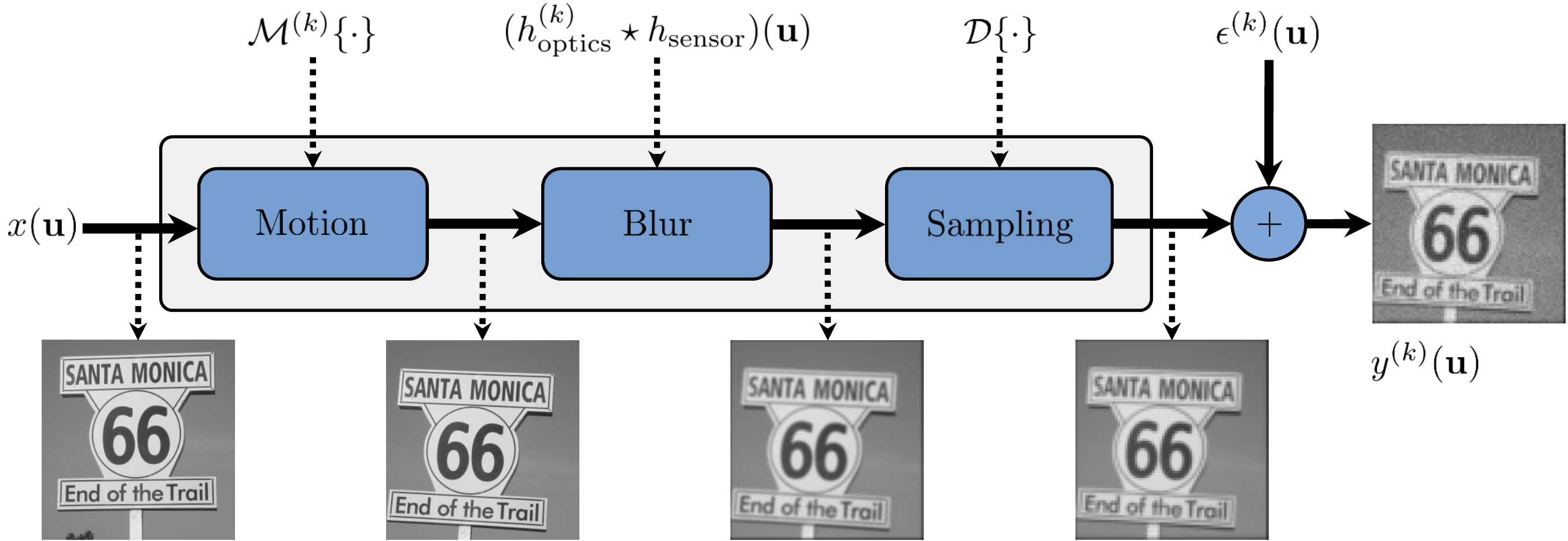}
	\caption[Image formation model employed in this work.]{Illustration of the image formation model employed in this work.}
	\label{fig:02_imageFormationModel}
\end{figure}

\subsection{Discretization of the Image Formation Model}
\label{sec:03_DiscretizationOfTheImageFormationModel}

In order to employ the image formation model defined in \eref{eqn:03_contImageModel} in a computational super-resolution algorithm, a discretization of the model equation is required. For this discretization, two common assumptions must be met. 

First, the only accessible information regarding a set of acquired images are the sampled intensity values at discrete pixel positions $\Point$ in the domain of the input images denoted by $\DomainLR \subset \RealN{2}$\label{notation:domainLR}. For the sake of convenience, the intensity values at the pixel positions for each frame \smash{$\LRFrameFun{\Point}{k}$} of size $\LRDimU \times \LRDimV$\label{notation:lrDim} are reorganized into a vector using line-by-line scanning defined by:
\vspace{-0.5em}
\begin{equation}
	\begin{split}
		\LRFrame{k} \label{notation:lrFrame} &\defeq 
		\begin{pmatrix} 
				\LRFrameFun{\Delta u_y, \Delta v_y}{k}
			& \LRFrameFun{\Delta u_y, 2 \Delta v_y}{k}
			& \ldots
			& \LRFrameFun{\LRDimU \Delta u_y, \LRDimV \Delta v_y}{k} 
		\end{pmatrix}^\top \\ 
		&\in \RealN{\LRDimU \cdot \LRDimV},
	\end{split}
\end{equation}
where $\Delta u_y$ and $\Delta v_y$ denote the pixel pitch in $\CoordU$- and $\CoordV$-direction, respectively. Since \smash{$\LRFrame{k}$} is defined on the pixel grid of the acquired frames, \smash{$\LRFrame{k}$} is referred to as a \textit{low-resolution} frame. We denote by $\LR$\label{notation:lr} the set of $\NumFrames$ low-resolution frames.

Second, as it is not feasible to reconstruct a continuous representation of the ideal image $\HRFun{\Point}$, we limit ourselves to the reconstruction of a digital image with finer spatial sampling in the domain $\DomainHR \subset \RealN{2}$\label{notation:domainHR}. The samples at pixel positions $\Point \in \DomainHR$ in the image $\HRFun{\Point}$ of size $\HRDimU \times \HRDimV$\label{notation:hrDim} are reorganized to a vector according to:
\vspace{-0.5em}
\begin{equation}
	\begin{split}
		\HR \label{notation:hrImage} &\defeq 
		\begin{pmatrix} 
				\HRFun{\Delta u_x, \Delta v_x}
			& \HRFun{\Delta u_x, 2 \Delta v_x}
			& \ldots
			& \HRFun{\HRDimU \Delta u_x, \HRDimV \Delta v_x}
		\end{pmatrix}^\top \\
		&\in \RealN{\HRDimU \cdot \HRDimV},
	\end{split}
\end{equation}
where $\Delta u_x$ and $\Delta v_x$ denote the pixel pitch in $\CoordU$- and $\CoordV$-direction, respectively. As these samples are defined on a finer pixel grid compared to the low-resolution frames, $\HR$ is referred to as \textit{high-resolution} image. Similar to the notation of low-resolution frames, the $k$-th warped version of the reference high-resolution image is denoted by \smash{$\HRFrame{k}$}. Assuming isotropic magnification, the magnification factor $\MagFac \in \Real$\label{notation:magFac} is given by $\MagFac  = \sqrt{\HRSize / \LRSize}$, where $\HRSize = \HRDimU \cdot \HRDimV$\label{notation:numHRPixels} and $\LRSize = \LRDimU \cdot \LRDimV$\label{notation:numLRPixels} denote sizes of high-resolution and low-resolution images, respectively.

\subsubsection{Discretization of the Motion Model}

Let us next discretize the motion model. For this purpose, we consider discrete pixel positions \smash{$\Point = (\CoordU, \CoordV)^\top \in \DomainHR$} in a warped high-resolution frame \smash{$\HRFrame{k}$}. Each point \smash{$\Point$ in $\HRFrame{k}$} is related to a transformed point \smash{$\PointTrans = (\CoordU^\prime, \CoordV^\prime)^\top \in \DomainHR$ in $\HR$}, where \smash{$\PointTrans = \DispVecFieldFrame{\Point}{k}$}. In order to describe subpixel motion, we introduce two motion models widely used within the algorithms developed in this work.

\paragraph{Parametric Motion.}

In case of a parametric model, $\Point$ is assumed to be transformed by a global transformation that is characterized by a small number of parameters to describe image warping of \smash{$\HRFrame{k}$} \wrt $\HR$. We define this model via a \textit{projective homography} in homogeneous coordinates \cite{Hartley2004} according to:
\begin{equation}
	\label{eqn:03_parametricMotion}
	\begin{split}
		\begin{pmatrix} \CoordU^\prime \\ \CoordV^\prime \\ 1 \end{pmatrix}
		&\cong
		\begin{pmatrix} 
			p_{11} & p_{12} & p_{13} \\
			p_{21} & p_{22} & p_{23} \\
			p_{31} & p_{32} & 1 \\
		\end{pmatrix}
		\begin{pmatrix} \CoordU \\ \CoordV \\ 1 \end{pmatrix} = \Homography \begin{pmatrix} \Point \\ 1 \end{pmatrix},
	\end{split}
\end{equation}
where $\cong$ denotes equality up to a scale factor. In general, the homography $\Homography$\label{notation:homography} has full rank and is parametrized by eight degrees of freedom given by nine matrix elements minus a scale factor. In particular cases, the degrees of freedom can be reduced, which leads to a hierarchy of transformations \cite{Hartley2004}. The following cases are widely considered in literature. A homography $\Homography[affine]$ is called \textit{affine} if it can be parametrized by six degrees of freedom according to:
\begin{equation}
	\Homography[affine] = \begin{pmatrix} \vec{A} & \TransVec \\ \Zeros^\top & 1 \end{pmatrix},
\end{equation}
where $\vec{A} \in \RealMN{2}{2}$, $\Zeros \in \RealN{2}$\label{notation:zeros} is an all-zero vector and $\TransVec \in \RealN{2}$\label{notation:transVec} denotes a translation vector. An affine homography describes subpixel motion by rotation, anisotropic scaling and shearing as well as translation. This transformation does not preserve angles between lines and ratio of distances in an image but the parallelism of lines is invariant under an affine homography.
		
An affine homography $\Homography[rigid]$ is called \textit{rigid} if it can be parametrized by three degrees of freedom according to:
\begin{equation}
	\Homography[rigid] = \begin{pmatrix} \RotMat(\RotAngle) & \TransVec \\ \Zeros^\top & 1 \end{pmatrix},
\end{equation}
where $\RotMat(\RotAngle)$\label{notation:rotMat} is an orthogonal rotation matrix parametrized by the rotation angle $\RotAngle$\label{notation:rotAngle}. A rigid homography describes subpixel motion by rotation and translation without considering scaling or shearing. Notice that this transformation preserves angles between lines as well as the ratio of distances in an image.

These models are completely described by a few degrees of freedom but can only model 3-D motion under certain assumptions. One of their major limitations is that they are only applicable under rigid body motion. Moreover, even for rigid body motion, it can be shown that a homography is only valid for pure rotational camera motion or general camera motion in case of planar scenes due to occlusions of objects  \cite{Hartley2004}. Nevertheless, this model is a reasonable approximation to describe rigid body motion for many applications of practical interest. One example is the acquisition of a static and non-planar scene from large distances such that it can be described as approximately planar, \eg in remote sensing.

\paragraph{Non-Parametric Motion.}

A more flexible approach is the description by a dense displacement vector field according to:
\begin{equation}
	\label{eqn:03_nonParametricMotion}
	\begin{pmatrix} \CoordU^\prime \\ \CoordV^\prime \end{pmatrix}
	=
	\begin{pmatrix} \CoordU + \DispVecFieldSym_{\CoordU}(\Point) \\ \CoordV + \DispVecFieldSym_{\CoordV}(\Point) \end{pmatrix}.
\end{equation}
The displacements $\DispVecField{\Point} = (\DispVecFieldSym_{\CoordU}(\Point), \DispVecFieldSym_{\CoordV}(\Point))^\top$ with $\DispVecFieldSym_i: \DomainLR \rightarrow \Real$, $i \in \{\CoordU, \CoordV\}$ describe the motion of $\HRFrame{k}$ towards the reference image $\HR$ at the pixel positions $\Point$. Under the brightness constancy assumption, camera and object motion can be related to displacements on the image plane using the notion of \textit{optical flow} \cite{Horn1981}.

This approach has the advantage that non-rigid deformations can be modeled along with rigid camera motion. In comparison to the parametric approach, it is also able to describe more general types of camera motion under projective distortions. However, as opposed to the parametric transformation in \eref{eqn:03_parametricMotion}, the non-parametric model in \eref{eqn:03_nonParametricMotion} might be not bijective, \eg due to occlusion. For this reason, one cannot obtain the motion of $\HR$ relative to $\HRFrame{k}$ by simply inverting the displacements $\DispVecField{\Point}$ that describe the motion in the opposite direction.

\subsubsection{Discretization of the Sampling Model}

Next, we examine the sampling process and discretize \eref{eqn:03_contImageModel}. In this continuous equation, a single low-resolution frame $\LRFrameFun{\Point}{k}$ with blur kernel $\BlurKernel{\Point}$ and subpixel displacements $\DispVecField{\Point}$ is related to $\HRFun{\Point}$ according to:
\begin{equation}
	\label{eqn:03_contImageModelNoiseFree}
	\LRFrameFun{\Point}{k} = \SamplingOp{ \HRFun{\Point + \DispVecField{\Point}} \conv \BlurKernel{\Point} } + \NoiseFrame{\Point}{k}.
\end{equation}
The discretization of this relationship is derived in terms of the transformed image $\HRFun{\Point + \DispVecField{\Point}}$ and we set $\PointTrans = \Point + \DispVecField{\Point}$. It is important to note that \eref{eqn:03_contImageModelNoiseFree} defines a complicated non-linear relationship between $\HRFun{\Point}$ and $\LRFrameFun{\Point}{k}$ as we do not limit the motion model $\DispVecField{\Point}$ to a simple linear transformation. To simplify the continuous formulation, we approximate \eref{eqn:03_contImageModelNoiseFree} as:
\begin{equation}
	\begin{split}
		\LRFrameFun{\Point}{k}	&\approx \SamplingOp{ \HRFun{\PointTrans} \conv \BlurKernel{\PointTrans} } + \NoiseFrame{\Point}{k}\\
														&= \SamplingOp{ \int_{\RealN{2}} \HRFun{\PointV} \BlurKernel{\PointTrans - \PointV} d\PointV } + \NoiseFrame{\Point}{k},
	\end{split}
\end{equation}
where we assumed that $\BlurKernel{\PointTrans}$ describing the blur kernel transformed according to $\DispVecField{\Point}$ fulfills $\BlurKernel{\PointTrans} \approx \BlurKernel{\Point}$. Notice that under pure translational motion and an arbitrary blur kernel or under rigid motion and a radially symmetric blur kernel, it follows that $\BlurKernel{\PointTrans} = \BlurKernel{\Point}$. Then, the motion and blur operations of the image formation model commutes \cite{Faramarzi2013}. In case of more general transformations, \eg affine motion, the approximation $\BlurKernel{\PointTrans} \approx \BlurKernel{\Point}$ can be justified by the fact that $\BlurKernel{\Point}$ is typically a spatially smooth kernel. In particular, this approximation is sensible under small subpixel motion. In case of local or non-rigid motion, one needs to assure that these deformations are small compared to global motion.  

In order to discretize the relationship between \smash{$\HRFun{\Point}$} and a single frame \smash{$\LRFrameFun{\Point}{k}$}, we consider the vectorized versions of these images given by $\HR$ and \smash{$\LRFrame{k}$}, respectively. Then, we implement \eref{eqn:03_contImageModel} in matrix/vector notation as:
\begin{equation}
	\LRFrame{k} = \SystemMatFrame{k} \HR + \FrameIdx{\NoiseVec}{k},
\end{equation}
where $\SystemMatFrame{k} \in \RealMN{\LRSize}{\HRSize}$\label{notation:systemMatFrame} denotes the \textit{system matrix} that comprises a discrete version of the motion model associated with the $k$-th frame, the blurring caused by the camera \gls{psf} as well as sampling on the sensor array. $\FrameIdx{\NoiseVec}{k} \in \RealN{\LRSize}$ is a random vector to model additive noise. Given a sequence of $\NumFrames$ frames yields:
\begin{equation}
	\label{eqn:03_discreteImageModel}
	\begin{split}
		\underbrace{\begin{pmatrix}
			\LRFrame{1} \\
			\vdots \\
			\LRFrame{\NumFrames}
		\end{pmatrix}}_{\LR}
		&= 
		\underbrace{\begin{pmatrix}
			\SystemMatFrame{1} \\
			\vdots \\
			\SystemMatFrame{\NumFrames}
		\end{pmatrix}}_{\SystemMat}
		\HR 
		+ 
		\underbrace{\begin{pmatrix}
			\FrameIdx{\NoiseVec}{1} \\
			\vdots \\
			\FrameIdx{\NoiseVec}{\NumFrames}
		\end{pmatrix}}_{\NoiseVec},
	\end{split}
\end{equation}
where $\SystemMat \in \RealMN{\NumFrames \LRSize}{\HRSize}$\label{notation:systemMat}, $\LR \in \RealN{\NumFrames \LRSize}$ and $\NoiseVec \in \RealN{\NumFrames \LRSize}$\label{notation:noiseVec} denote the combined versions of the system matrices, the low-resolution frames and the noise vectors, respectively. In the sequel, we introduce two approaches to implement this relationship.

\paragraph{Implementation using Filter Operations.}
To avoid an explicit computation of the system matrix, the image formation can be implemented by discrete filter operations \cite{Zomet2000}. For this approach, the system matrix of the $k$-th frame is decomposed as:
\begin{equation}
	\SystemMatFrame{k} = \SamplingMat \BlurMat \FrameIdx{\MotionMat}{k},
\end{equation}
where $\SamplingMat \in \RealMN{\LRSize}{\HRSize}$\label{notation:samplingMat} denotes subsampling by Dirac impulses, $\BlurMat \in \RealMN{\HRSize}{\HRSize}$\label{notation:blurMat} models the blur kernel, and \smash{$\FrameIdx{\MotionMat}{k} \in \RealMN{\HRSize}{\HRSize}$}\label{notation:motionMatFrame} encodes the motion for the $k$-th frame in matrix notation. These operations are discrete versions of their continuous counterparts in \fref{fig:02_imageFormationModel} and can be implemented as follows. The motion operator \smash{$\FrameIdx{\MotionMat}{k}$} is modeled by geometric warping of $\HR$ according to the underlying motion model. The blur operator $\BlurMat$ is implemented by means of a discrete convolution, whereas the filter kernel corresponds to a discrete version of $\BlurKernel{\Point}$. The subsampling operator $\SamplingMat$ is implemented by a nearest-neighbor interpolation.

This approach is computationally efficient in terms of memory management as the system matrix does not need to be stored. However, due to the concatenation of the three operations interpolation artifacts in these stages are propagated. Here, image warping to implement \smash{$\FrameIdx{\MotionMat}{k}$} might result in aliasing artifacts due to resampling that is required to obtain the intermediate image \smash{$\FrameIdx{\MotionMat}{k} \HR$}. These artifacts are propagated to blurring and subsampling but are not physically meaningful.

\paragraph{Implementation using Matrix Operations.}

The approach that is employed in this thesis is based on the work of Tipping and Bishop \cite{Tipping2003}. For this implementation, the system matrix is constructed without decomposition in discrete filters. This requires that the \gls{psf} is modeled by a narrow kernel $\BlurKernel{\PointTrans}$ and its continuous version is used to determine the matrix elements. This is reasonable since the integration of photons per pixel on the sensor array is performed over a finite area and one does not need to consider the entire detector surface. Hence, the convolution can be replaced by an integration over a circular neighborhood $\Neighbor_{\text{PSF}}(\PointTrans)$ centered at $\PointTrans$, where $\Neighbor_{\text{PSF}}(\PointTrans) = \{\PointV : ||\PointV - \PointTrans||_2 \leq \PSFKernelSize \}$ and $\PSFKernelSize$\label{notation:neighborPSF}\label{notation:psfKernelSize} denotes the \gls{psf} radius. Then, the frame $\LRFrameFun{\Point}{k}$ is related to the high-resolution image $\HRFun{\Point}$ according to:
\begin{equation}
	\begin{split}
		\LRFrameFun{\Point}{k}
			&\approx \SamplingOp { \int_{\Neighbor_{\text{PSF}}(\PointTrans)} \HRFun{\PointV} \BlurKernel{\PointTrans - \PointV} d\PointV } 
			+ \NoiseFrame{\Point}{k} \\
			&= \SamplingOp{ \sum_{\PointV \in \Neighbor_{\text{PSF}}(\PointTrans)} \HRFun{\PointV} \BlurKernel{\PointTrans - \PointV} }
			+ \NoiseFrame{\Point}{k}.
	\end{split}
\end{equation}
Thus, the intensity at the pixel position $\Point$ is described by a weighted sum of the intensities in $\Neighbor_{\text{PSF}}(\PointTrans)$ and the weights are expressed in terms of $\BlurKernel{\PointTrans}$. For the $k$-th frame, these weights are encoded in the system matrix \smash{$\SystemMatFrame{k}$}. The matrix element at position $(m, n)$ is determined with normalized row sums according to:
\begin{equation}
	\label{eqn:systemMatrixElements}
	W_{mn} =
	\begin{cases}
		\frac{1}{\sum_{i=1}^\HRSize h \left( \PointTrans_m - \PointV_i \right) } h \left( \PointTrans_m - \PointV_n \right)	
			& \PointV_n \in \Neighbor_{\text{PSF}}(\PointTrans_m)  \\
		0	
			& \text{otherwise}
	\end{cases},
\end{equation}
where $\PointTrans_m$ are the coordinates of the $m$-th pixel in $\LRFrameFun{\Point}{k}$ transformed to the coordinate grid of $\HRFun{\Point}$, and $\PointV_n$ are the coordinates of the $n$-th pixel in $\HRFun{\Point}$. This construction exploits the fact that the \gls{psf} is described by a narrow blur kernel and we set $W_{mn} = 0$ if $\PointTrans_m$ and $\PointV_n$ does not affect each other. 

One common assumption is to model the \gls{psf} by an isotropic Gaussian kernel $\BlurKernel{\Point} = \exp(-\frac{1}{2} ||\Point||_2^2 / (\MagFac^2 \PSFWidth^2))$ and to truncate $\BlurKernel{\Point}$ for $||\Point||_2 > 3 \PSFWidth$\label{notation:psfWidth} \cite{Pickup2007a} as depicted in \fref{fig:02_systemMatrixComputation}. Notice that $\PSFWidth$ characterizes the \gls{psf} size in units of low-resolution pixels. For efficient storage, the joint system matrix $\SystemMat$ is assembled as a sparse matrix since most of the elements are zero due to the finite support of the blur kernel. The size of the matrix depends on the number of frames $\NumFrames$, the image dimension $\LRSize$ as well as the \gls{psf} radius $\PSFKernelSize$. In the asymptotic case, the number of non-zero elements in $\SystemMat$ is $\BigO{\NumFrames \LRSize \PSFKernelSize}$.
\begin{figure}[!t]
	\centering
		\includegraphics[width=1.00\textwidth]{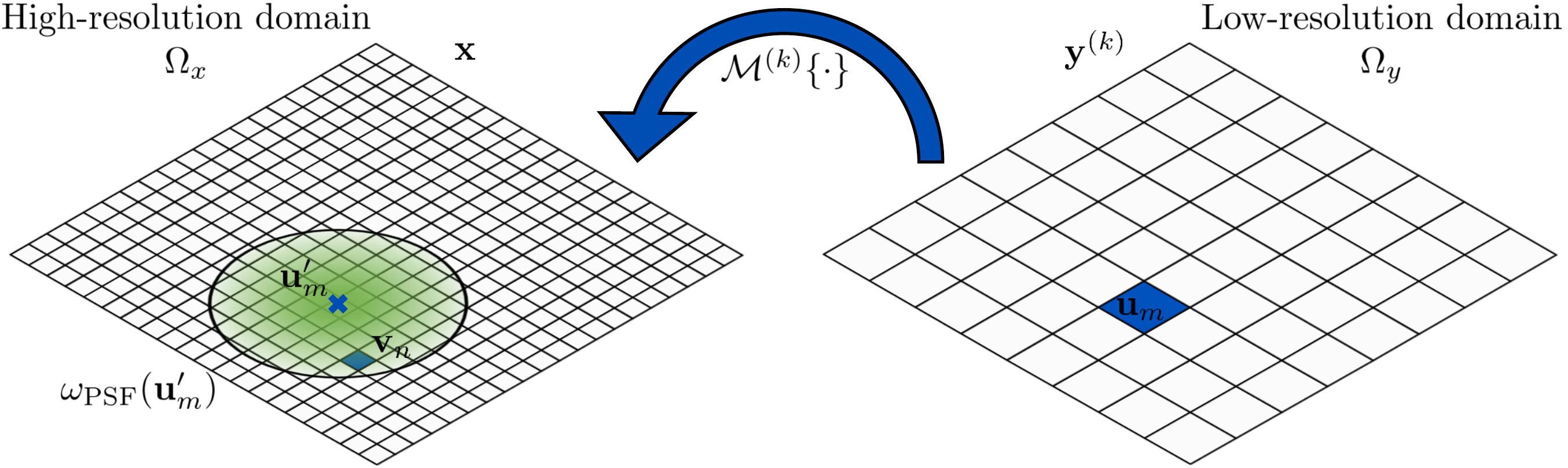}
	\caption[Construction of the system matrix in an element-wise scheme]{Construction of the system matrix $\SystemMatFrame{k}$ in an element-wise scheme. Each low-resolution pixel $\Point_m$ is warped towards the high-resolution image $\HR$ resulting in the transformed pixel $\PointTrans_m$. The element $W_{mn}$ is computed from $\PointTrans_m$ and $\PointV_n$ assuming a radial symmetric \gls{psf} that is non-zero in the neighborhood $\Neighbor_{\text{PSF}}(\PointTrans_m)$ but approaches zero otherwise.}
	\label{fig:02_systemMatrixComputation}
\end{figure}

\subsection{Discussion and Limitations of the Model}
\label{sec:03_DiscussionAndLimitationsOfTheModel}

The benefit of the presented image formation model is that low-resolution data $\LR$ can be simulated in an efficient way from a high-resolution image $\HR$ by means of the system matrix $\SystemMat$, which is precomputed from motion and imaging parameters. This simulation only involves matrix-vector operations and is used to implement super-resolution via iterative energy minimization. For an efficient implementation, the matrix elements can be calculated in a parallel way \cite{Wetzl2013}.

It is important to note that the proposed discretization approximates the continuous model in \eref{eqn:03_contImageModel} by assuming that the \gls{psf} blur kernel is the same in the reference frame and a subpixel warped frame. This is reasonable under subpixel motion that is approximately rigid. In prior work, other discretization schemes have been proposed, see \eg \cite{Pickup2007a} and \cite{Capel2004} as well as the references therein. These are more accurate under certain types of motion, \eg affine motion, but are computationally more demanding.

There are also several practical limitations of the model that need to be considered. First, it is assumed that the characteristics of the imaging system in terms of the \gls{psf} are space invariant. In many practical applications, this assumption might be violated. Common examples are motion blur \cite{Ma2015} or atmospheric blur that might be space variant. In order to model such effects, the construction of the system matrix needs to be implemented spatially adaptive to take a varying blur kernel into account. Another class of limitations is related to internal signal processing performed by a camera after capturing raw data. One issue is white balancing, which results in photometric variations over the low-resolution frames. In this chapter, such effects are not considered but the model can be extended to allow spatially and temporally varying photometric conditions, see \cref{sec:RetinalFundusVideoImaging}. Such shortcomings of the model haven been quantitatively studied in \cite{Pickup2007a}.

In the above derivation, we limited ourselves to single-channel images, where each pixel represents one discrete measurement. One interesting extension is to model the acquisition of color images in the RGB space. In today's low-cost cameras, color images are mosaiced since each pixel can only measure one spectral band according to a \gls{cfa} \cite{Farsiu2006}. This \gls{cfa} needs to be considered in color image super-resolution by extending the image formation model.

\section{Bayesian Modeling of Super-Resolution}
\label{sec:02_BayesianModelingOfSuperResolution}

This section introduces multi-frame super-resolution from a Bayesian point of view. Let us describe the latent high-resolution image $\HR$ along with the set of low-resolution observations $\LR$ as random variables. The \textit{observation model} that describes the probability of observing a single frame \smash{$\LRFrame{k}$} from the high-resolution image $\HR$ is denoted by the \gls{pdf} \smash{$\PdfCondT{\LRFrame{k}}{\HR}$}\label{notation:pdf}\label{notation:pdfCond}. This conditional probability is defined in terms of the discrete image formation model in \eref{eqn:03_discreteImageModel}. If the observation noise $\NoiseVec$ is space invariant and follows a normal distribution with zero mean and covariance $\NoiseStd^2 \Id$\label{notation:normalDist}\label{notation:noiseStd}, the observation model is given by:
\begin{equation}
	\label{eqn:03_MGGD}
	\begin{split}
		\PdfCond{\LRFrame{k}}{\HR} 
			&= \NormalDist{\LRFrame{k} - \SystemMatFrame{k} \HR}{\Zeros}{\NoiseStd^2 \Id} \\
			&\defeq \frac{1}{\NoiseStd \sqrt{2 \pi}}
			\exp \left\{ 
				- \frac{ \big(\LRFrame{k} - \SystemMatFrame{k} \HR \big)^\top \big(\LRFrame{k} - \SystemMatFrame{k} \HR \big) }
				{2 \NoiseStd^2 } 
			\right\},
	\end{split}
\end{equation}
Notice that the observation model can be tailored to different noise characteristics. In \sref{sec:04_BayesianModelForRobustSuperResolution}, we present a model that accounts for space variant noise in the low-resolution observations.

Below, we review two commonly used approaches to infer the high-resolution image $\HR$ from the set of low-resolution observations $\LR$ under this distribution. Both approaches yield a formulation of super-resolution as unconstrained energy minimization and provide point estimates for the latent high-resolution image.

\subsection{Maximum Likelihood Estimation}

Let us assume that the sequence of low-resolution frames \smash{$\LRSequence{1}{\NumFrames}$} are independent and identically distributed (\iid) random variables. Then, one can derive the joint distribution to describe the probability of observing the low-resolution data $\LR$ from the high-resolution image $\HR$ according to the factorization:
\begin{equation}
	\PdfCond{\LR}{\HR} = \prod_{k = 1}^\NumFrames \NormalDist{\LRFrame{k} - \SystemMatFrame{k} \HR}{\Zeros}{\NoiseStd^2 \Id}.
\end{equation}

The objective of \gls{ml} estimation is to infer a high-resolution image that best explains the set of low-resolution observations. If there is no prior knowledge about the high-resolution image available, it can be directly inferred from the point estimation:
\begin{equation}
	\label{eqn:03_MLEstimator}
	\HR_{\text{ML}} = \argmax_{\HR} \PdfCond{\LR}{\HR}.
\end{equation}
Taking the negative log-likelihood $\DataTerm{\HR} \propto -\log \PdfCondT{\LR}{\HR}$\label{notation:dataTerm} of \eref{eqn:03_MLEstimator}, this is equivalent to the unconstrained minimization problem:
\begin{equation}
	\label{eqn:03_MLEstimatorEnergyMin}
	\HR_{\text{ML}}	= \argmin_{\HR} \DataTerm{\HR},
\end{equation}
where:
\begin{equation}
	\begin{split}
		\DataTerm{\HR} 
		&= \sum_{k = 1}^\NumFrames \big| \big| \ResidualSym \big(\HR, \LRFrame{k} \big) \big| \big|_2^2 \\
		&= \big| \big| \ResidualFun{\HR}{\LR} \big| \big|_2^2\enspace,
	\end{split}
\end{equation}
and \smash{$\ResidualFun{\HR}{\LR} = \LR - \SystemMat \HR$}\label{notation:residualError} denotes the \textit{residual error} of the estimate $\HR$ \wrt the observations $\LR$. This links the Gaussian observation model to least-square optimization.

The convex minimization problem in \eref{eqn:03_MLEstimatorEnergyMin} can be solved in closed form. Unfortunately, this requires an inversion of the system matrix $\SystemMat$, which is computationally prohibitive for real-world problem sizes. For the purpose of a practical implementation, energy minimization is performed by means of iterative numerical optimization to avoid a direct inversion of the system matrix. Several approaches that are widely used in literature include steepest descent iterations with fixed \cite{Elad1997,Li2010} or adaptive step size \cite{Hardie1997,Lee2003} as well as \gls{cg} based iteration schemes \cite{Nguyen2001a,Zibetti2007}. In case of pure translational motion and a space invariant generative model, \gls{ml} estimation can also be decomposed into a non-iterative interpolation and an iterative deblurring stage \cite{Elad2001}. For details on the numerical optimization, we refer to \sref{sec:04_BayesianModelForRobustSuperResolution}.

\subsection{Maximum A-Posteriori Estimation}
\label{sec:03_MaximumAPosterioriEstimation}

In terms of \gls{map} estimation, prior knowledge regarding the occurrence of high-resolution images is exploited instead of using a uniform prior. The motivation of this approach is that super-resolution is a highly ill-posed problem under practical conditions \cite{Borman2004}. Thus, \gls{ml} estimation that does not consider prior knowledge on the desired high-resolution image needs to be regularized to steer the reconstruction algorithm to a reasonable solution. 

\Fref{fig:03_srExample_ml} depicts this issue on a simulated dataset with known subpixel motion, where $\NumFrames = 16$ low-resolution frames are obtained from a ground truth according to the proposed image formation model. This example considers low-resolution observations in the intensity range $[0, 1]$ that are corrupted by Gaussian noise at different standard deviations $\NoiseStd$. In the corresponding \gls{ml} estimates, image noise in the input frames is severely amplified. Notice that in addition to the influence of image noise, super-resolution based on \gls{ml} estimation is also ill-conditioned in case of uncertainties of model parameters \cite{Pickup2007a}. 
\begin{figure}[t]
	\centering
	\subfloat[Ground truth]{\includegraphics[width=0.236\textwidth]{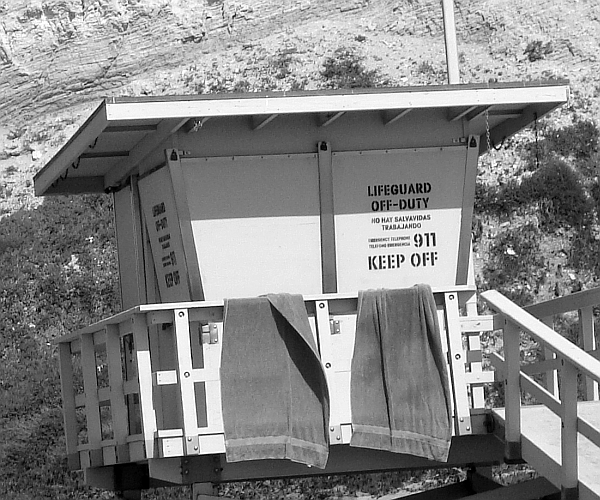}}\\[-0.8ex]
	\subfloat{\includegraphics[width=0.236\textwidth]{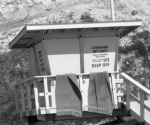}}~
	\subfloat{\includegraphics[width=0.236\textwidth]{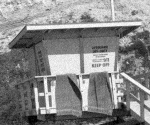}}~
	\subfloat{\includegraphics[width=0.236\textwidth]{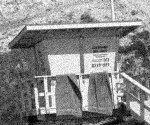}}~
	\subfloat{\includegraphics[width=0.236\textwidth]{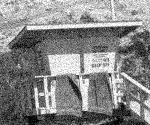}}\\[-0.8ex]
	\setcounter{subfigure}{1}
	\subfloat[$\NoiseStd = 0$]{\includegraphics[width=0.236\textwidth]{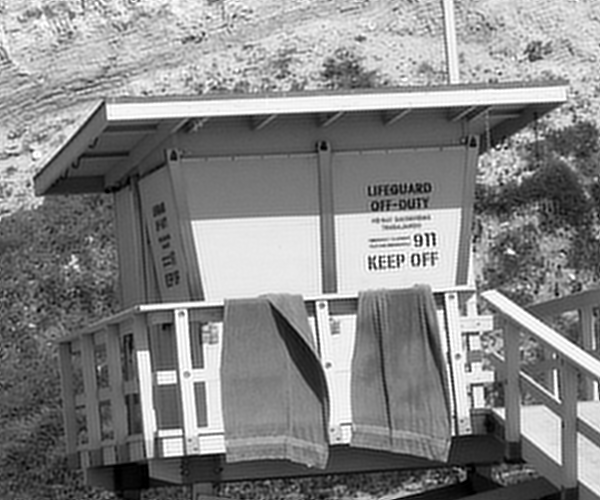}}~
	\subfloat[$\NoiseStd = 0.025$]{\includegraphics[width=0.236\textwidth]{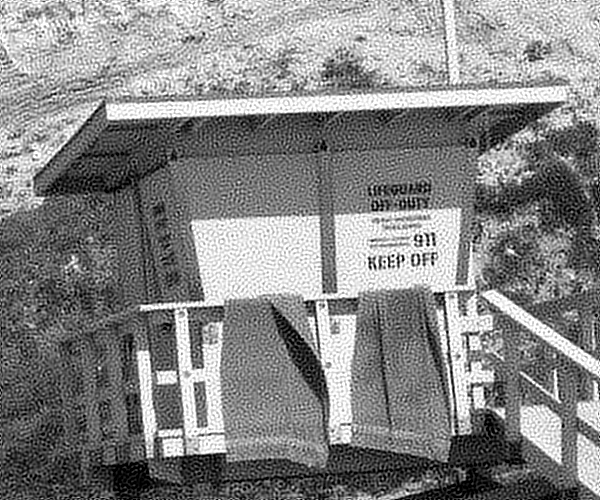}}~
	\subfloat[$\NoiseStd = 0.05$]{\includegraphics[width=0.236\textwidth]{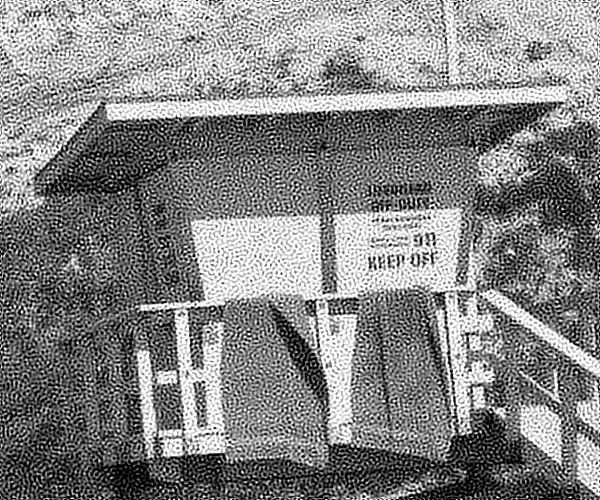}}~
	\subfloat[$\NoiseStd = 0.075$]{\includegraphics[width=0.236\textwidth]{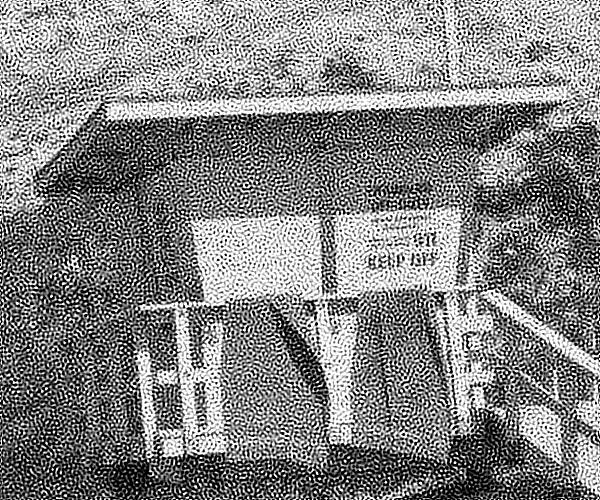}}
	\caption[Super-resolution using \gls{ml} estimation on simulated data]{Super-resolution using \gls{ml} estimation on simulated data. First row: ground truth image used for this example. Second row: low-resolution frames simulated from the ground truth at different levels of Gaussian noise with standard deviation $\NoiseStd$. Third row: \gls{ml} estimates using $\NumFrames = 16$ low-resolution frames with magnification $\MagFac = 4$.}
	\label{fig:03_srExample_ml}
\end{figure}

Let \smash{$\PdfT{\HR}$} be a prior distribution on the latent high-resolution image to model its appearance in Bayesian way. Similarly, let $\PdfT{\LR}$ be the distribution of the low-resolution frames. Then, using Bayes' rule, the posterior distribution $\PdfCondT{\HR}{\LR}$ is given by:
\begin{equation}
	\label{eqn:03_bayesRule}
	\begin{split}
		\PdfCond{\HR}{\LR}	
			&= \frac{\PdfCond{\LR}{\HR} \cdot \Pdf{\HR}}{\Pdf{\LR}} \propto \PdfCond{\LR}{\HR} \cdot \Pdf{\HR}.
	\end{split}	
\end{equation}
The goal of \gls{map} estimation is to reconstruct a high-resolution image that maximizes the posterior $\PdfCondT{\HR}{\LR}$ according to:
\begin{equation}
	\HR_{\text{MAP}} = \argmax_{\HR} \left\{ \PdfCond{\LR}{\HR} \Pdf{\HR} \right\}.
\end{equation}
In order to derive this point estimation as an energy minimization problem, we use the negative log-likelihood associated with the posterior according to:
\begin{equation}
	\label{eqn:03_mapEstimator}
	\HR_{\text{MAP}}	
	= \argmin_{\HR} 
		\left\{ 
			\DataTerm{\HR} + \RegWeight \RegTerm{\HR}
		\right\},
\end{equation}
where \smash{$\DataTerm{\HR}$} denotes the negative log-likelihood from \gls{ml} estimation referred to as \textit{data fidelity term}. \smash{$\RegTerm{\HR} \propto - \log \PdfT{\HR}$}\label{notation:regTerm} denotes a \textit{regularization term} with the regularization weight $\RegWeight \geq 0$\label{notation:regWeight} to weight this term relative to the data fidelity. 

To define the prior distribution $\PdfT{\HR}$, various approaches emerged in literature. A wide class of these general-purpose models exploits smoothness, piecewise smoothness or sparsity of natural images in transform domains. We consider image priors that are expressed by the Boltzmann distribution \cite{Bishop2006}:
\begin{equation}
	\label{eqn:03_expFormPrior}
	\Pdf{\HR} = \frac{1}{\PartFun(\PriorStd)} \exp \left\{- \frac{\RegTerm{\HR}}{\PriorStd} \right\},
\end{equation}
where $\PriorStd > 0$ is a distribution scale parameter and $\PartFun(\PriorStd)$\label{notation:partFun} is the partition function used for normalization. Below, we review some of the most commonly used priors that are employed in this thesis. For the design of prior distributions that are tailored to the characteristics of natural images, we refer to \cref{sec:RobustMultiFrameSuperResolutionWithSparseRegularization}.

\paragraph{Gaussian Prior.}
A Gaussian prior as the most basic and commonly used approach \cite{Elad1997,Hardie1997,Capel2003} considers a high-resolution image $\HR$ as a spatially smooth signal. This should avoid noise amplification by \gls{ml} estimation. In doing so, the prior $\PdfT{\HR}$ defined by \eref{eqn:03_expFormPrior} is parametrized by:\label{notation:regTermGauss}
\begin{equation}
	\label{eqn:03_gaussianPrior}
	\RegTerm[Gauss]{\HR} 
	= \left|\left| \HPMat \HR \right|\right|_2^2,
\end{equation}
where $\HPMat \in \RealMN{\HRSize}{\HRSize}$\label{notation:hpMat} is a circulant matrix to write a discrete convolution with a high-pass filter $\HPKernel$\label{notation:hpKernel} as a matrix-vector product, \ie $\HPMat \HR \equiv \HR \conv \HPKernel$. Typical choices for $\HPMat$ are the gradient determined by finite differences \cite{Capel2003} or the discrete Laplacian \cite{Hardie1997,Kondi2006a}. This prior yields a Tikhonov regularized optimization problem also known as ridge regression \cite{Hastie2009}. The main benefits of this model are that it is feasible to use the prior for analytical computations and that it yields a convex regularization term that is easy to minimize by numerical optimization.

\Fref{fig:03_srExample_map} depicts the influence of the image prior on the simulated dataset in \fref{fig:03_srExample_ml}. In this example, the \gls{map} estimation is based on a discrete Laplacian and a constant regularization weight ($\RegWeight = 0.05$). In contrast to \gls{ml} estimation, noise amplification is reduced by regularization with the Gaussian prior. This stabilizes the reconstruction and improves the visual quality of super-resolved images.
\begin{figure}[t]
	\centering
	\subfloat[$\NoiseStd = 0$]{\includegraphics[width=0.236\textwidth]{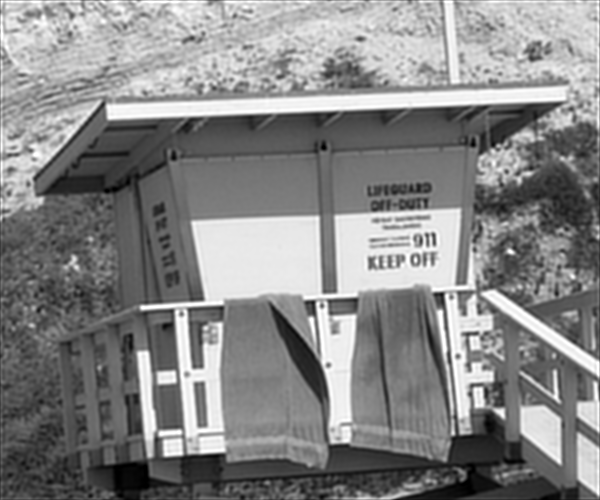}}~
	\subfloat[$\NoiseStd = 0.025$]{\includegraphics[width=0.236\textwidth]{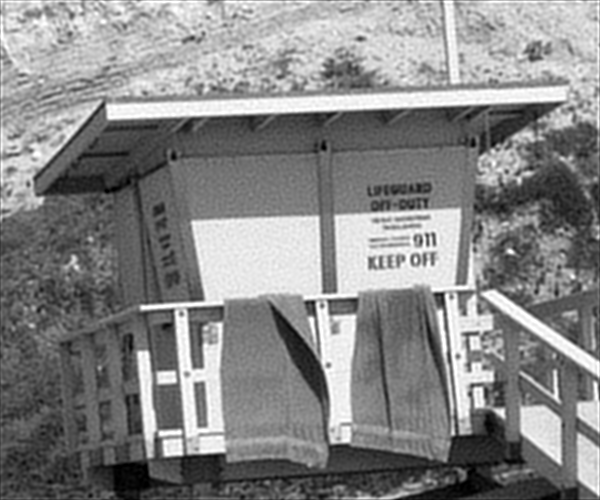}}~
	\subfloat[$\NoiseStd = 0.05$]{\includegraphics[width=0.236\textwidth]{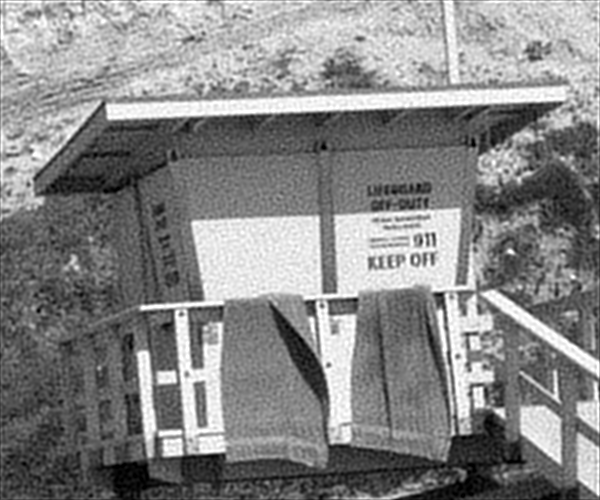}}~
	\subfloat[$\NoiseStd = 0.075$]{\includegraphics[width=0.236\textwidth]{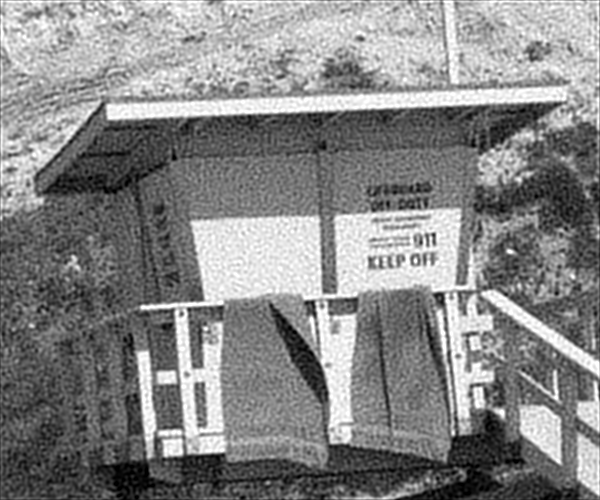}}
	\caption[\Gls{map} estimation on the simulated data in \fref{fig:03_srExample_ml}]{\Gls{map} estimation on the simulated data in \fref{fig:03_srExample_ml} with a Gaussian prior modeled by a discrete Laplacian \cite{Hardie1997} and a constant regularization weight ($\RegWeight = 0.05$). Super-resolved images are depicted at different noise standard deviations $\NoiseStd$.}
	\label{fig:03_srExample_map}
\end{figure}

\paragraph{Huber Prior.}
The major shortcoming of the Gaussian prior is that discontinuities in an image are penalized in the same manner as noise. This reduces the ability of edge reconstruction since sharp edges appear blurred due to the smoothness assumption. One approach to enhance edge reconstruction is to replace the \LTwo norm of the Gaussian prior by a robust loss function to obtain a distribution with heavier tails. A common choice is to employ the Huber prior \cite{Schultz1996,Pickup2007}:\label{notation:regTermHuber}
\begin{equation}
	\label{eqn:03_huberPrior}
	\RegTerm[Huber]{\HR} 
	= \sum_{i = 1}^\HRSize \HuberLoss{\VecEl{\HPMat \HR}{i}},
\end{equation} 
where $\HPMat \in \RealMN{\HRSize}{\HRSize}$ is a circulant matrix and $\VecEl{\vec{z}}{i}$\label{notation:vecEl} denotes the $i$-th element of the vector $\vec{z}$. The function $\HuberLoss{z}$ denotes the Huber loss applied element-wise to the high-pass filtered image $\HPMat \HR$. In this work, we use the smooth approximation of the Huber loss that has continuous first- and second-order derivatives \cite{Hartley2004}:
\begin{equation}
	\label{eqn:huber}
	\HuberLoss{z} =  \HuberThresh \sqrt{1 + \left( \frac{z}{\HuberThresh} \right)^2} - \HuberThresh,
\end{equation}
where $\HuberThresh$ is a scale parameter. This function behaves like the Gaussian prior for small $z$ ($z \ll \HuberThresh$) and penalizes $z$ quadratically. In case of large $z$ ($z \gg \HuberThresh$), it is proportional to $|z|$. Hence, it features piecewise smooth regularization in order to enhance the reconstruction of discontinuities. Similar to the Gaussian prior, \eref{eqn:03_huberPrior} yields a convex regularization term. However, in contrast to the Gaussian prior, it requires non-linear optimization techniques for energy minimization.

\paragraph{Total Variation.}

The Rudin, Osher and Fatemi (ROF) model \cite{Rudin1992} also known as \gls{tv} has been originally introduced for image denoising. Later, it has also been employed for blind deconvolution \cite{Chan1998a} and super-resolution \cite{Ng2007}. Unlike the Huber prior that provides piecewise smooth regularization, the \gls{tv} prior explains an image as a piecewise constant signal. 

The isotropic version of this prior introduced in \cite{Rudin1992} is defined by:\label{notation:regTermTV}
\begin{equation}
	\RegTerm[TV]{\HR}	
	= \sum_{i = 1}^\HRSize \sqrt{ \VecEl{ \nabla_u \HR }{i} ^2 + \VecEl{ \nabla_v \HR }{i} ^2 },
\end{equation}
where $\nabla_u \HR$ and $\nabla_v \HR$\label{notation:imageGrad} denote the discrete image gradient in $\CoordU$- and $\CoordV$-direction, respectively. This convex regularization term exploits the sparsity of natural images in the gradient domain. Besides its application in image restoration problems, this prior has also great importance for regularization of ill-posed problems in the theory of compressed sensing \cite{Donoho2006,Candes2008}. 

Isotropic \gls{tv} has the limitation that it considers the image gradient in horizontal and vertical direction only. For this reason, super-resolution is prone to staircasing artifacts in image regions with small gradient magnitudes. One common generalization of this approach is the use of \gls{btv} \cite{Farsiu2004a}. The \gls{btv} prior is inspired from bilateral filtering \cite{Tomasi1998} and is given by:\label{notation:regTermBTV}
\begin{equation}
	\label{eqn:03_btvPrior}
	\RegTerm[BTV]{\HR} 
	= \sum_{m = -\BTVSize}^\BTVSize \sum_{n = -\BTVSize}^\BTVSize 
	\BTVWeight^{|m| + |n|} \left| \left| \HR - \vec{S}_u^m \vec{S}_v^n \HR \right| \right|_1,
\end{equation}
where $\vec{S}_u^m$\label{notation:shiftMat} and $\vec{S}_v^n$ denote shifts of $\HR$ by $m$ pixel in $\CoordU$-direction and $n$ pixel in $\CoordV$-direction, respectively. The shifts are performed in a $(2 \BTVSize + 1) \times (2 \BTVSize + 1)$\label{notation:btvSize} window, where $\BTVWeight \in ]0, 1]$\label{notation:btvWeight} weights the difference between $\HR$ and its shifted version according to the shift magnitude. This prior performs a multiscale analysis of the image gradient and yields convex regularization similar to isotropic \gls{tv}. 

\section{Conclusion}

This chapter introduced the computational framework that is utilized for the algorithm development in the remainder of this work. In the first part, a literature survey served as an overview regarding different paradigms of multi-frame super-resolution including frequency domain, interpolation-based as well as iterative spatial domain approaches. The algorithms proposed in this thesis are based on the iterative spatial domain formulation due to its flexibility in terms of the motion model and the ability to integrate prior knowledge. The second part covered the derivation of an image formation model to describe image acquisition in digital imaging from a mathematical viewpoint. This model was discretized in order to make it applicable for multi-frame super-resolution algorithms. Finally, super-resolution was formulated from a Bayesian perceptive as a statistical parameter estimation based on the discrete image formation model. For this purpose, \gls{ml} and \gls{map} estimation were discussed. These formulations state super-resolution as energy minimization problem that provides the basis for the computational methods introduced in the subsequent chapters.


\chapter{Robust Multi-Frame Super-Resolution with Sparse Regularization}
\label{sec:RobustMultiFrameSuperResolutionWithSparseRegularization}

\myminitoc

\noindent
The computational framework for multi-frame super-resolution as previously presented in \cref{sec:ComputationalFrameworkForMultiFrameSuperResolution} relies on a simplistic approximation of the true physics of image acquisition and accurate mathematical modeling of this process. For instance, it requires an accurate subpixel motion estimate and prior knowledge about the distribution of measurement noise. Uncertainty regarding these aspects limits the robustness of super-resolution in real-world applications. This chapter introduces a new algorithm for robust super-resolution imaging derived from a Bayesian point of view. The proposed method employs a confidence-aware observation model along with a sparse image prior and is implemented as iteratively re-weighted minimization. Unlike previous work, this approach features robust and edge preserving image reconstruction with small amount of parameter tuning, is flexible in terms of imaging models and computationally efficient. 

Parts of this chapter have been originally published in \cite{Kohler2015c} and have been later extended in \cite{Bercea2016}.

\section{Introduction}
\label{sec:04_Introduction}

Robustness is one of the main design criteria for the development of multi-frame super-resolution algorithms. In this context, the term \textit{robustness} refers to the ability of an algorithm to reconstruct a reasonable high-resolution image even in the presence of degenerated information employed in the reconstruction procedure. Conversely, an algorithm can be considered as not robust if it is severely affected by a small uncertainty of this information. In practice, such uncertainties are unavoidable and deteriorate super-resolution reconstruction. Let us discuss several aspects of practical relevance. These refer to motion estimation, image formation models, numerical optimization, and regularization.

\paragraph{Motion Estimation Uncertainty.}
An important aspect for robust super-resolution is the uncertainty of subpixel motion. If this information is unknown, super-resolution requires an accurate motion estimate to provide reliable results. Unfortunately, this requirement is hard to fulfill using motion estimation on low-resolution images due to systematic artifacts like aliasing or blurring as well as random measurement noise \cite{Vandewalle2006}. These issues cause uncertainties in the estimated subpixel motion. For this reason, motion estimation and super-resolution can be considered as a chicken-or-egg dilemma. In particular, this is the case if independently moving objects in a scene must be taken into account. Another challenging situation is non-rigid motion that causes ambiguities due to occlusions. In these cases, motion estimation may be affected by local outliers. In order to simplify motion estimation, some existing super-resolution algorithms are limited to simple parametric motion, \eg globally rigid motion. However, parametric models are inappropriate in many applications, \eg video upscaling \cite{Keller2011}.
\begin{figure}[t]
	\centering
		\subfloat[Low-resolution frame]{
		\begin{tikzpicture}[spy using outlines={rectangle,red,magnification=2.5,height=2.25cm, width=2.25cm, connect spies, every spy on node/.append style={thick}}] 
			\node {\pgfimage[width=0.32\linewidth]{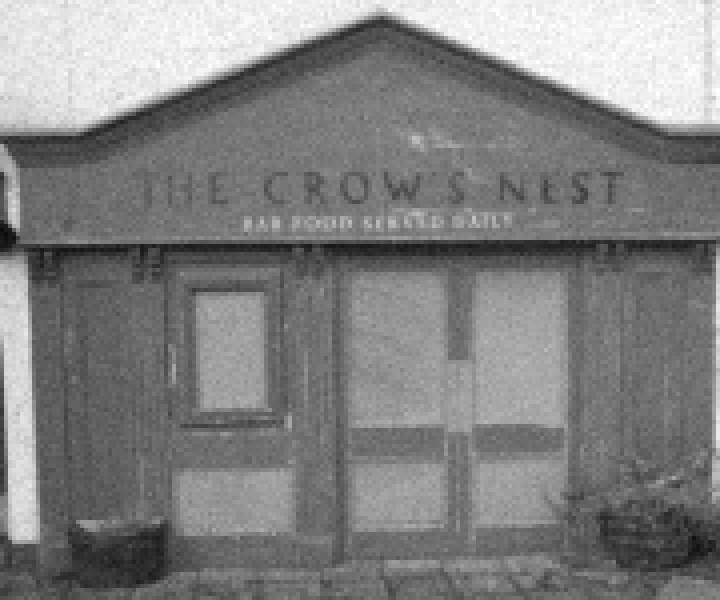}}; 
			\spy on (-0.4,0.5) in node [left] at (-2.5,-0.9);
		\end{tikzpicture}
		\label{fig:04_robustnessMotionUncertainty:lr}
	}
	\subfloat[Ground truth]{
		\begin{tikzpicture}[spy using outlines={rectangle,red,magnification=2.5,height=2.25cm, width=2.25cm, connect spies, every spy on node/.append style={thick}}] 
			\node {\pgfimage[width=0.32\linewidth]{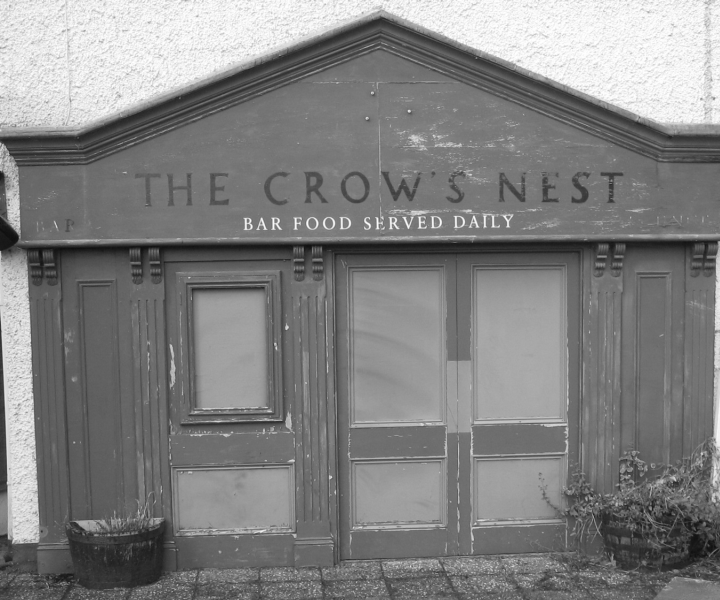}}; 
			\spy on (-0.4,0.5) in node [left] at (-2.5,-0.9);
		\end{tikzpicture}
		\label{fig:04_robustnessMotionUncertainty:gt}
	}
	\\
	\subfloat[Super-resolution with exact motion]{
		\begin{tikzpicture}[spy using outlines={rectangle,red,magnification=2.5,height=2.25cm, width=2.25cm, connect spies, every spy on node/.append style={thick}}] 
			\node {\pgfimage[width=0.32\linewidth]{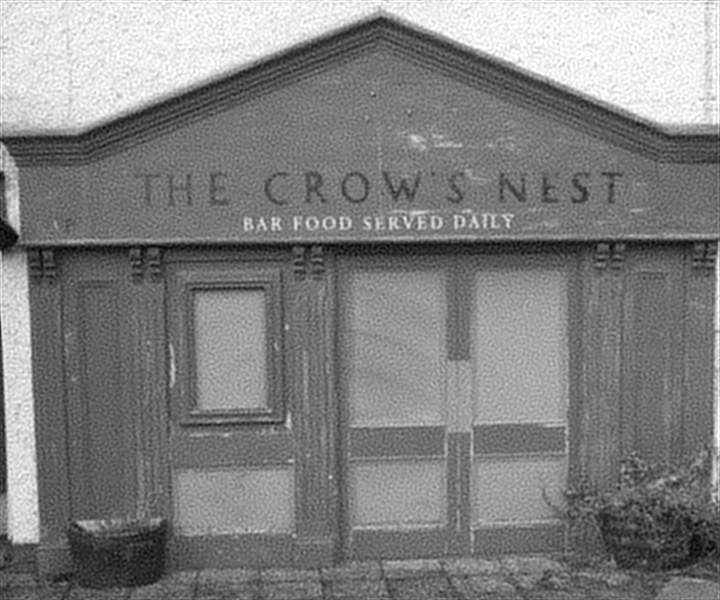}}; 
			\spy on (-0.4,0.5) in node [left] at (-2.5,-0.9);
		\end{tikzpicture}
		\label{fig:04_robustnessMotionUncertainty:exactMotion}
	}
	\subfloat[Super-resolution with inaccurate translation]{
		\begin{tikzpicture}[spy using outlines={rectangle,red,magnification=2.5,height=2.25cm, width=2.25cm, connect spies, every spy on node/.append style={thick}}] 
			\node {\pgfimage[width=0.32\linewidth]{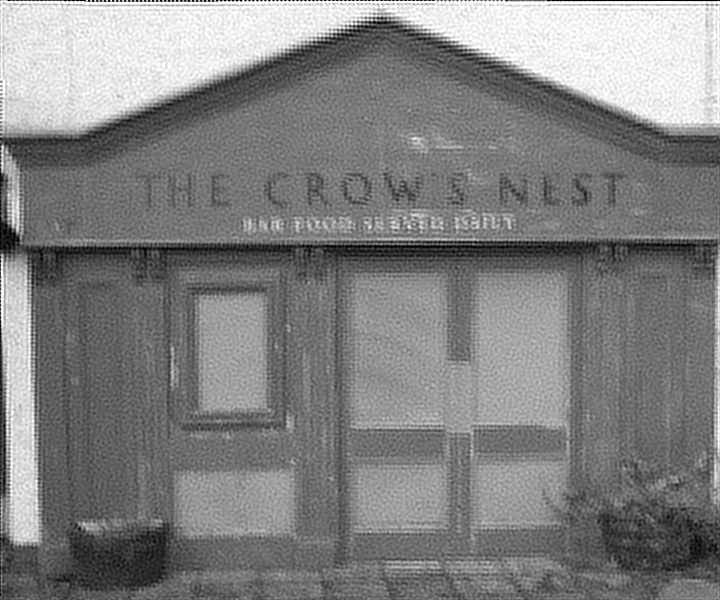}}; 
			\spy on (-0.4,0.5) in node [left] at (-2.5,-0.9);
		\end{tikzpicture}
		\label{fig:04_robustnessMotionUncertainty:inexactMotion}
	}
	\caption[Influence of motion estimation uncertainty to super-resolution]{Super-resolution under motion estimation uncertainty. 
		\protect\subref{fig:04_robustnessMotionUncertainty:lr} Simulated low-resolution frame. 
		\protect\subref{fig:04_robustnessMotionUncertainty:gt} High-resolution ground truth. 
		\protect\subref{fig:04_robustnessMotionUncertainty:exactMotion} and \protect\subref{fig:04_robustnessMotionUncertainty:inexactMotion} Super-resolved images ($4 \times$ magnification) using an \LTwo norm data fidelity term and Tikhonov regularization \cite{Elad1997} with exact subpixel motion and inaccurate translations, respectively. The inaccurate motion estimate leads to ghosting artifacts due to the non-robust observation model.
	}
	\label{fig:04_robustnessMotionUncertainty}
\end{figure}

The impact of motion estimation is demonstrated in \fref{fig:04_robustnessMotionUncertainty}, where $\NumFrames = 16$ low-resolution frames (\fref{fig:04_robustnessMotionUncertainty:lr}) were generated from a ground truth (\fref{fig:04_robustnessMotionUncertainty:gt}) by simulating rigid motion and Gaussian noise. Super-resolution is based on an \LTwo norm data fidelity and Tikhonov regularization using Laplacian filtering \cite{Elad1997}. The super-resolved image using exact subpixel motion is depicted in \fref{fig:04_robustnessMotionUncertainty:exactMotion}. Next, the translations $\vec{t} \in \RealN{2}$ of every second frame were corrupted by uniform distributed errors $\vec{t}_{\epsilon} \in \RealN{2}$ of random directions with $|| \vec{t}_{\epsilon} ||_2 = 1.0$. Super-resolution under this inaccurate motion is depicted in \fref{fig:04_robustnessMotionUncertainty:inexactMotion}. Notice that even under these small uncertainties, the reconstruction is severely affected by ghosting artifacts.

\paragraph{Model and Optimization Parameter Uncertainty.}
In addition to motion estimation, the uncertainties of parameters employed in the image formation model have a considerable impact on super-resolution. Compared to conditions in many real-world imaging setups, super-resolution usually approximates the true physics of image acquisition with simplified mathematical models. One aspect that is rarely considered is internal processing of image data in the camera system, \eg image compression or white-balancing, see \sref{sec:03_DiscussionAndLimitationsOfTheModel}. Another issue is measurement noise that does not follow a simple space invariant normal distribution, \eg due to invalid pixels related to impulse noise \cite{Chan2005} or mixed noise \cite{Xiao2011b}. The influence of these aspects is demonstrated by corrupting low-resolution data by \textit{salt-and-pepper} noise using a fraction of 0.5\,\% invalid pixels as depicted in \fref{fig:robustnessInvalidPixels:lr_invalidPixels1} and \fref{fig:robustnessInvalidPixels:lr_invalidPixels2}. Super-resolution on the corrupted low-resolution frames is not able to compensate for invalid pixels, see \fref{fig:robustnessInvalidPixels:sr}.
\begin{figure}[!t]
	\centering
	\subfloat[1st frame with salt-and-pepper noise]{
		\begin{tikzpicture}[spy using outlines={rectangle,red,magnification=2.5,height=2.25cm, width=2.25cm, connect spies, every spy on node/.append style={thick}}] 
			\node {\pgfimage[width=0.32\linewidth]{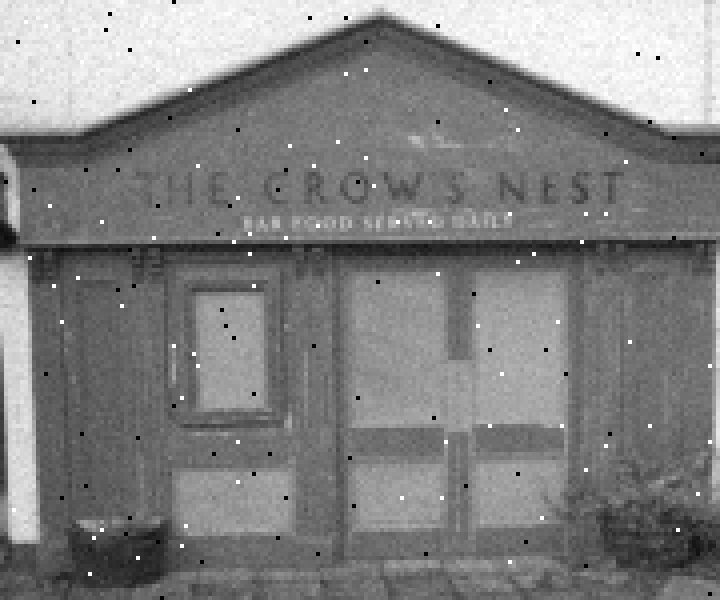}}; 
			\spy on (-0.4,0.5) in node [left] at (-2.5,-0.9);
		\end{tikzpicture}
		\label{fig:robustnessInvalidPixels:lr_invalidPixels1}
	}
	\subfloat[2nd frame with salt-and-pepper noise]{
		\begin{tikzpicture}[spy using outlines={rectangle,red,magnification=2.5,height=2.25cm, width=2.25cm, connect spies, every spy on node/.append style={thick}}] 
			\node {\pgfimage[width=0.32\linewidth]{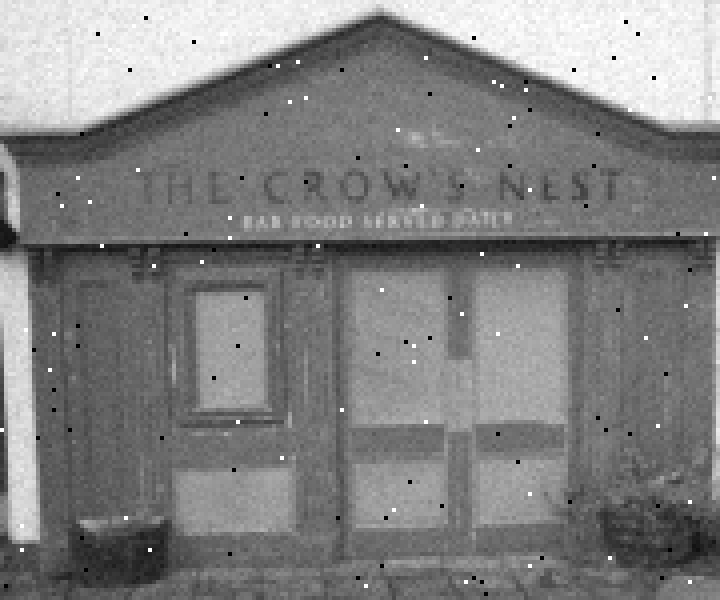}}; 
			\spy on (-0.4,0.5) in node [left] at (-2.5,-0.9);
		\end{tikzpicture}
		\label{fig:robustnessInvalidPixels:lr_invalidPixels2}
	}
	\\
	\subfloat[Super-resolution under salt-and-pepper noise]{
		\begin{tikzpicture}[spy using outlines={rectangle,red,magnification=2.5,height=2.25cm, width=2.25cm, connect spies, every spy on node/.append style={thick}}] 
			\node {\pgfimage[width=0.32\linewidth]{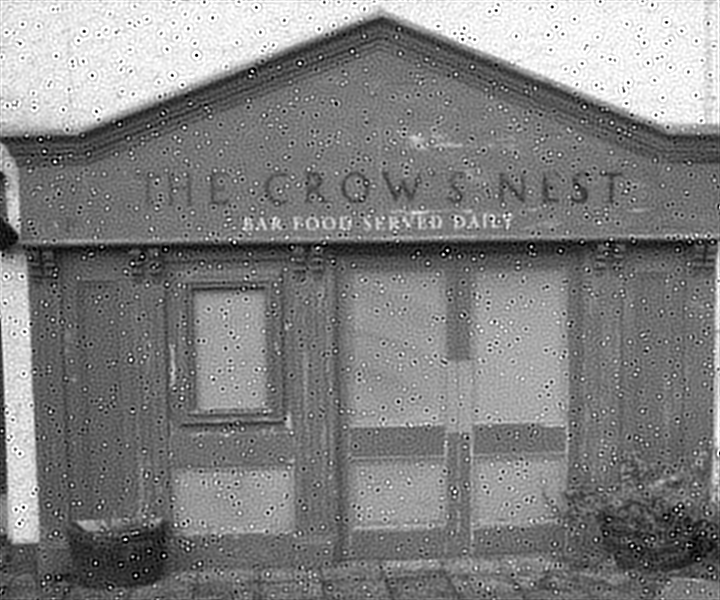}}; 
			\spy on (-0.4,0.5) in node [left] at (-2.5,-0.9);
		\end{tikzpicture}
		\label{fig:robustnessInvalidPixels:sr}
	}
	\subfloat[Ground truth]{
		\begin{tikzpicture}[spy using outlines={rectangle,red,magnification=2.5,height=2.25cm, width=2.25cm, connect spies, every spy on node/.append style={thick}}] 
			\node {\pgfimage[width=0.32\linewidth]{images/chapter4/robustnessExample_gt.png}}; 
			\spy on (-0.4,0.5) in node [left] at (-2.5,-0.9);
		\end{tikzpicture}
		\label{fig:robustnessInvalidPixels:gt}
	}
	\caption[Influence of invalid pixels to super-resolution]{Influence of invalid pixels to super-resolution. 
		\protect\subref{fig:robustnessInvalidPixels:lr_invalidPixels1} and \protect\subref{fig:robustnessInvalidPixels:lr_invalidPixels2} Frames from \fref{fig:04_robustnessMotionUncertainty} corrupted by salt-and-pepper noise. 
		\protect\subref{fig:robustnessInvalidPixels:sr} and \protect\subref{fig:robustnessInvalidPixels:gt} Super-resolved image using an \LTwo norm data fidelity term with Tikhonov regularization \cite{Elad1997} and the ground truth. Notice that the non-robust observation model is unable to compensate for invalid pixels.
	}
	\label{fig:04_robustnessInvalidPixels}
\end{figure}

Besides model parameters, there are also optimization parameters related to the formulation of super-resolution as energy minimization problem. One example are regularization weights that are selected prior to optimization. This is cumbersome as parameter selection is often performed off-line by trial-and-error or by automatic parameter selection schemes \cite{Nguyen2001}. However, in both cases super-resolution is affected by an inappropriate selection and cannot compensate for the uncertainty of the parameters. This issue is demonstrated for the regularization weight $\RegWeight$ in \fref{fig:04_robustnessRegularizationWeight}. In case of an underestimated regularization weight (\fref{fig:robustnessRegularizationWeight:underest}), super-resolution is affected by residual noise. As opposed to underestimation, an overestimated weight (\fref{fig:robustnessRegularizationWeight:overest}) results in oversmoothing. The optimal weight (\fref{fig:robustnessRegularizationWeight:optimal}) results in a suitable tradeoff between residual noise and sharpness.

\paragraph{Ill-Posedness and Regularization.}
Super-resolution is known to be an ill-posed problem \cite{Borman2004} and prior knowledge regarding the appearance of the images to be reconstructed is required to alleviate ill-posedness, see \sref{sec:03_MaximumAPosterioriEstimation}. This prior knowledge is leveraged by regularization techniques, where most parametric regularization terms are derived from Gaussian, Huber or \gls{tv} priors. This has a crucial impact on the performance of super-resolution but most of these general-purpose priors are inadequate to model natural images \cite{Baker2002}. For instance, discontinuities related to texture or edges are not explained appropriately due to the assumption of smooth or piecewise smooth images used to design these priors. This becomes crucial in presence of image noise as there is an inherent tradeoff between denoising and the preservation of edges or texture.
\begin{figure}[!t]
	\centering
	\subfloat[Underestimated reg. weight ($\RegWeight = 5 \cdot 10^{-5}$)]{
		\begin{tikzpicture}[spy using outlines={rectangle,red,magnification=2.5,height=2.25cm, width=2.25cm, connect spies, every spy on node/.append style={thick}}] 
			\node {\pgfimage[width=0.32\linewidth]{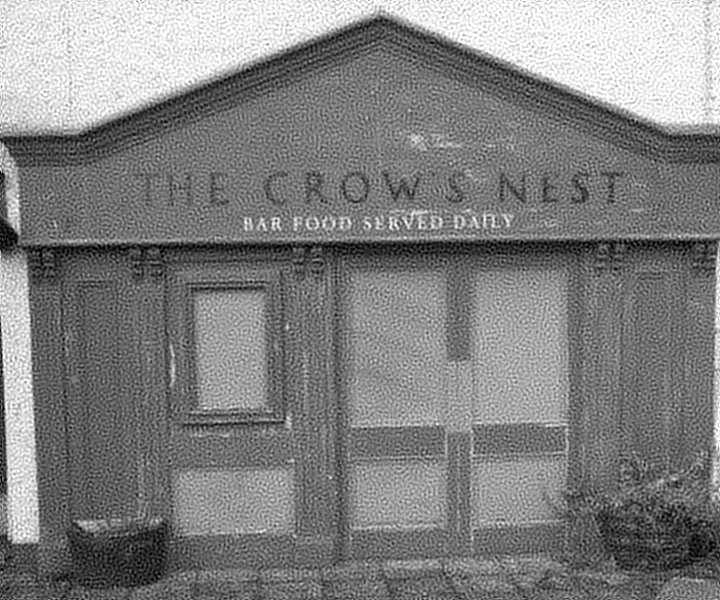}}; 
			\spy on (-0.4,0.5) in node [left] at (-2.5,-0.9);
		\end{tikzpicture}
		\label{fig:robustnessRegularizationWeight:underest}
	}
	\subfloat[Optimal reg. weight ($\RegWeight = 5 \cdot 10^{-3}$)]{
		\begin{tikzpicture}[spy using outlines={rectangle,red,magnification=2.5,height=2.25cm, width=2.25cm, connect spies, every spy on node/.append style={thick}}] 
			\node {\pgfimage[width=0.32\linewidth]{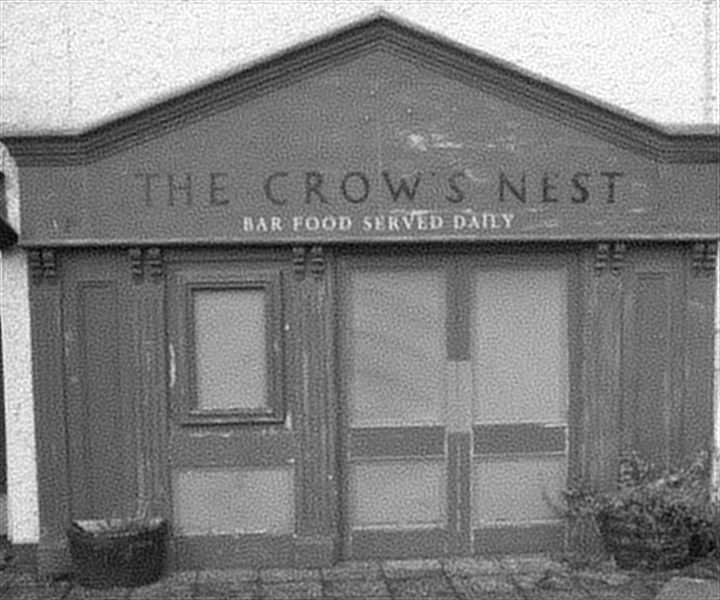}}; 
			\spy on (-0.4,0.5) in node [left] at (-2.5,-0.9);
		\end{tikzpicture}
		\label{fig:robustnessRegularizationWeight:optimal}
	}
	\\
	\subfloat[Overestimated reg. weight ($\RegWeight = 5 \cdot 10^{-1}$)]{
		\begin{tikzpicture}[spy using outlines={rectangle,red,magnification=2.5,height=2.25cm, width=2.25cm, connect spies, every spy on node/.append style={thick}}] 
			\node {\pgfimage[width=0.32\linewidth]{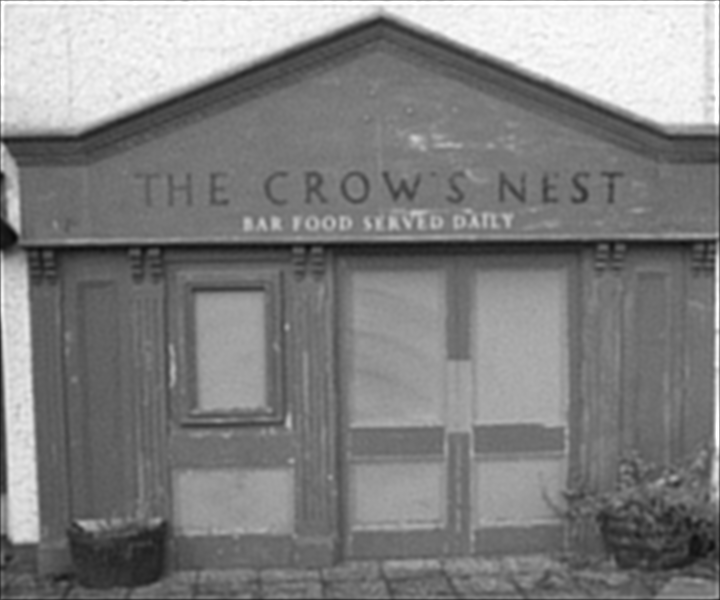}}; 
			\spy on (-0.4,0.5) in node [left] at (-2.5,-0.9);
		\end{tikzpicture}
		\label{fig:robustnessRegularizationWeight:overest}
	}
	\subfloat[Ground truth]{
		\begin{tikzpicture}[spy using outlines={rectangle,red,magnification=2.5,height=2.25cm, width=2.25cm, connect spies, every spy on node/.append style={thick}}] 
			\node {\pgfimage[width=0.32\linewidth]{images/chapter4/robustnessExample_gt.png}}; 
			\spy on (-0.4,0.5) in node [left] at (-2.5,-0.9);
		\end{tikzpicture}
		\label{fig:robustnessRegularizationWeight:gt}
	}
	\caption[Influence of the regularization weight $\RegWeight$ to super-resolution]{Influence of the regularization weight $\RegWeight$ to super-resolution. 
		\protect\subref{fig:robustnessRegularizationWeight:underest} - \protect\subref{fig:robustnessRegularizationWeight:gt} Super-resolution using an \LTwo norm data fidelity term and Tikhonov regularization \cite{Elad1997} with an underestimated $\RegWeight$ ($\RegWeight = 5 \cdot 10^{-5}$), an optimal $\RegWeight$ ($\RegWeight = 5 \cdot 10^{-3}$), an overestimated $\RegWeight$ ($\RegWeight = 5 \cdot 10^{-1}$) as well as the ground truth. Notice that an over- or underestimation lead to oversmoothing or noise amplification, respectively.
	}
	\label{fig:04_robustnessRegularizationWeight}
\end{figure}

The remainder of this chapter is organized as follows. \sref{sec:04_RelatedWork} presents a literature survey on related work in the area of robust super-resolution. \sref{sec:04_BayesianModelForRobustSuperResolution} introduces a confidence-aware Bayesian model for robust super-resolution reconstruction. \sref{sec:04_RobustSuperResolutionReconstruction} introduces an iterative estimation scheme based on this model. \sref{sec:04_ExperimentsAndResults} presents a comprehensive evaluation of this algorithm with comparisons to the state-of-the-art. Finally, \sref{sec:04_Conclusion} concludes this chapter.

\section{Related Work}
\label{sec:04_RelatedWork}

The algorithms most relevant to this work are based on Bayesian models. In particular, we are interested in the formulation of super-resolution as \gls{map} estimation as well as related probabilistic methods. The focus in the design of robust algorithms in this area lies in outlier detection, optimization, and regularization, see \fref{fig:04_relatedWork}.

\paragraph{Outlier Detection.}
The goal of outlier detection is to identify and downweight invalid observations termed \textit{outliers}. According to \eref{eqn:03_mapEstimator}, this is done by defining the data term:
\begin{equation}
	\label{eqn:04_outlierDetectionDataTerm}
	\DataTerm{\HR} = \sum_{i = 1}^{\NumFrames \LRSize} \WeightsBFun_i \big| \VecEl{\LR - \SystemMat \HR}{i} \big|^p,
\end{equation}
where $p \in [1, 2]$ and $\WeightsB = (\WeightsBFun_1, \ldots, \WeightsBFun_{\NumFrames \LRSize})^\top$ denotes a confidence map to indicate valid observations referred to as \textit{inliers}. The confidence $\WeightsBFun_i$ can be either a binary or a continuous variable, where $\WeightsBFun_i = 1$ corresponds to an inlier $\LRSym_i$.

\begin{figure}[!t]
	\centering
		\includegraphics[width=1.00\textwidth]{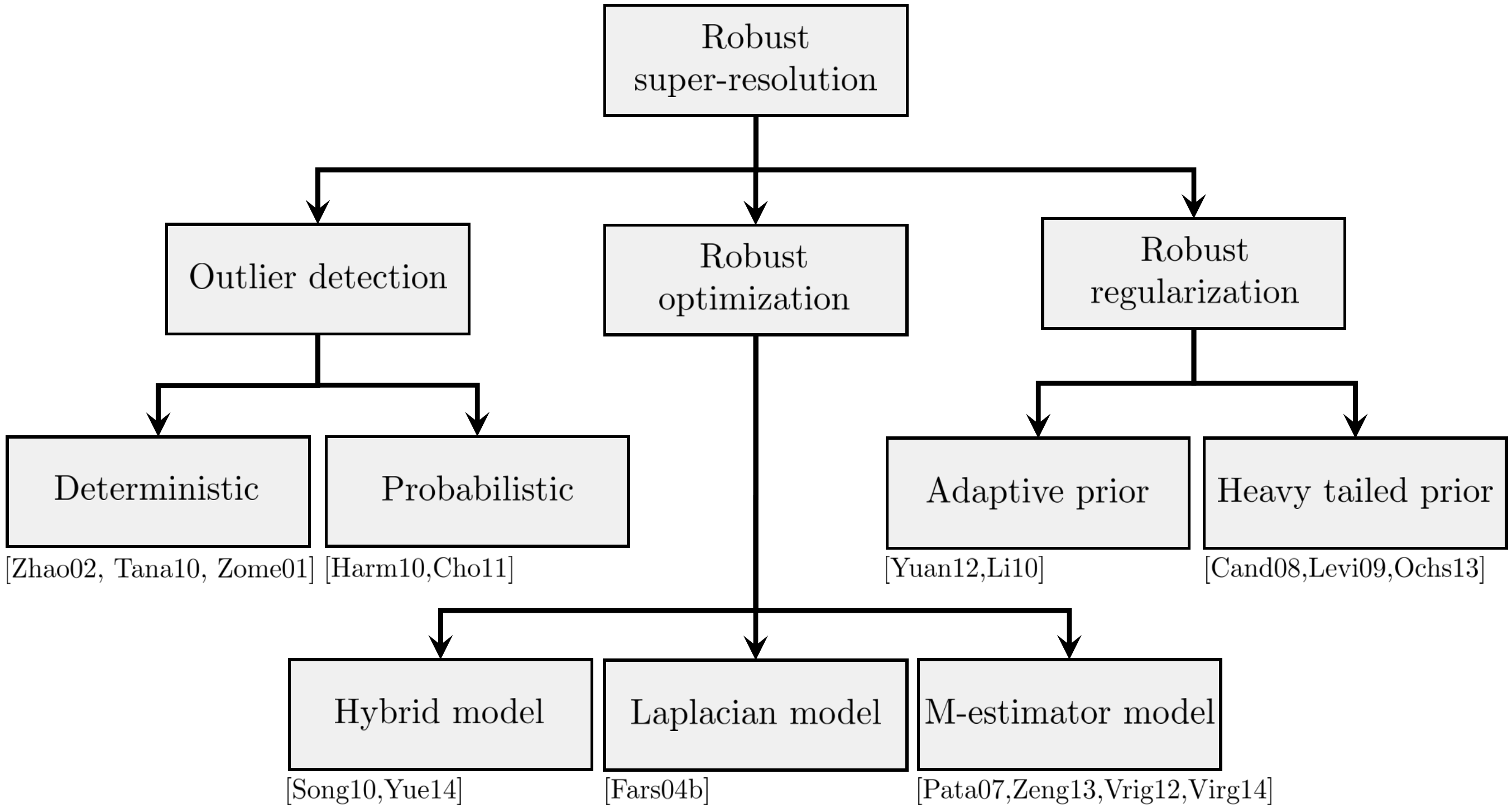}
	\caption[Related work on robust super-resolution techniques]{Overview of related work on robust super-resolution techniques.}
	\label{fig:04_relatedWork}
\end{figure}

Deterministic approaches detect outliers in low-resolution observations. To this end, Zhao and Sawhney \cite{Zhao2002} have proposed local image similarity assessment on displacement vector fields. In \cite{Tanaka2010}, Tanaka and Okutomi have proposed displacement estimation to construct confidence maps. This enables the detection of local outliers in optical flow. One drawback is that outlier detection does not exploit the presence of super-resolved data. Probabilistic strategies aggregate outlier detection and image reconstruction by \gls{em} \cite{Harmeling2010,Cho2011}. These methods also focus on the detection of specific types of outliers, \eg oversaturated pixels, by describing them in a probabilistic way. 

In a different approach, Zomet \etal \cite{Zomet2001} have proposed robust gradient descent optimization. Instead of an explicit construction of the confidence map, outliers in the gradient descent update equations are filtered during iterative minimization. To remove outliers, median filtering of the gradient of the energy function is performed. This approach does not focus on specific types of outliers but has no proper theoretical justification in terms of convergence \cite{Farsiu2004a}.  

\paragraph{Robust Optimization.}
In contrast to explicit outlier detection, outlier observations can be removed implicitly by defining the data fidelity as: 
\begin{equation}
	\label{eqn:04_robustOptimizationDataTerm}
	\DataTerm{\HR} = \sum_{i = 1}^{\NumFrames \LRSize} \LossFun[\text{data}]{\VecEl{\LR - \SystemMat \HR}{i}},
\end{equation}
where $\LossFun[\text{data}]{\cdot}$ is a robust loss function. Robustness refers to the property that outliers are not penalized disproportionately as in case of the \LTwo norm. 

Farsiu \etal \cite{Farsiu2004a} have introduced robust \gls{map} estimation based on the \LOne norm, where the loss function in \eref{eqn:04_robustOptimizationDataTerm} is given by $\LossFun[\text{data}]{z} = |z|$. This measure is statistically optimal in case of Laplacian noise. However, Laplacian noise is rarely the optimal model for inliers, where Gaussian noise is much more common. Inlier and outlier observations are also uniformly weighted and their influences are proportional to the magnitude of their residual errors. From a statistical point of view, this is not necessarily optimal. Besides the \LOne norm, \textit{re-descending M-estimators} have been examined to design the loss function $\LossFun[\text{data}]{z}$. The use of non-convex functions should further reduce the influence of outliers by rejecting them in the data fidelity term. In \cite{Patanavijit2007}, Patanavijit and Jitapunkul have proposed the Lorentzian loss. Other widely used M-estimators are the Gaussian \cite{Pham2008} or Tukey's biweight \cite{Anastassopoulos2009}. For the selection of the scale parameters of these functions, different adaptive schemes have been introduced \cite{El-Yamany2008a,El-Yamany2008}. Zeng and Yang \cite{Zeng2013} have proposed an adaptive Huber function to consider a varying reliability of model parameters associated with the low-resolution frames. A similar approach is to employ hybrid error norms \cite{Song2010,Yue2014}, where $\LossFun[\text{data}]{z}$ is an aggregation of the \LOne and the \LTwo norm to combine their advantages.

One common issue of the aforementioned methods is that they rely on additional model parameters. For instance, regularization weights or scale parameters of M-estimators need to be specified. This task often requires user supervision or ad-hoc methods based on empirical knowledge. Numerical methods for adaptive regularization weight selection have been developed by He and Kondi \cite{Kondi2006a} and Vrigkas \etal \cite{Vrigkas2012,Vrigkas2014}. However, these are based on Tikhonov regularization limiting the ability of edge reconstruction as discussed below.

\paragraph{Robust Regularization.}
Besides recognition-based image priors \cite{Baker2002}, most algorithms employ parametric prior distributions including Huber \cite{Pickup2007a} or \gls{tv} models \cite{Farsiu2004a,Ng2007}. These smoothness priors are characterized by convex regularization terms $\RegTerm{\HR}  \propto -\log\PdfT{\HR}$ related to a distribution $\PdfT{\HR}$ and are not spatially adaptive. While this often leads to efficient algorithms, one inherent limitation is the ability to represent the characteristics of natural images in terms of sparsity. As we show in the derivation of the proposed prior, natural images are typically sparse and need to be represented by \textit{heavy-tailed} distributions \cite{Huang1999}. Such priors have been widely investigated for deblurring, where the Hyper-Laplacian distribution is a common choice \cite{Levin2009,Krishnan2009a,Kotera2013a}. In \cite{Patanavijit2007}, Patanavijit and Jitapunkul have proposed non-convex Lorentzian-Laplacian regularization that implements a heavy-tailed prior for super-resolution. 

Another class of priors aims at enhancing the parametric models to make them spatially adaptive. Yuan \etal \cite{Yuan2012,Yuan2013} and Li \etal \cite{Li2010} have presented spatially adaptive versions of \gls{tv} and \gls{btv}, respectively. The idea is to decrease the impact of the regularization on discontinuities compared to the impact in homogenous regions. This may improve edge reconstruction compared to the unweighted counterparts of these priors. However, their benefit is highly dependent on additional feature extraction algorithms, \eg the computation of second-order derivatives \cite{Yuan2012} or entropy-based measures \cite{Li2010}.  

Regularization of ill-posed problems has also been widely investigated in the theory of compressed sensing \cite{Donoho2006}. Here, sparse regularization is achieved by iteratively re-weighted \LOne norm optimization to approximate \Lp{0} norm minimization. This scheme has been studied by Candes \etal \cite{Candes2008} and Daubechies \etal \cite{Daubechies2010} for sparse signal recovery, where it leads to sparser solutions compared to unweighted priors. This property makes these techniques attractive also for regularization in low-level vision problems \cite{Ochs2013}.

\section{Bayesian Model for Robust Super-Resolution}
\label{sec:04_BayesianModelForRobustSuperResolution}

This section introduces the mathematical model of the proposed super-resolution algorithm from a Bayesian perspective. This requires the definition of an observation model as well as a reasonable image prior. For both components, we propose space variant distributions to enhance space invariant modeling.

\subsection{Space Variant Observation Model}
\label{sec:04_SpaceVariantObservationModel}

In the most basic formulation of multi-frame super-resolution, the observation model $\PdfCondT{\LR}{\HR}$ is defined by a family of parametric distributions. For instance, one can employ a normal distribution \cite{Elad1997} assuming additive Gaussian noise or a Laplacian distribution \cite{Farsiu2004a} assuming additive Laplacian noise in the image formation process, see \sref{sec:02_BayesianModelingOfSuperResolution}. The main motivation behind this approach lies in its simplicity as noise can be fully described by a small number of parameters, \eg its standard deviation, and the corresponding data fidelity term is convex. However, it has the shortcoming of being sensitive to outliers and cannot model spatially varying uncertainties, see \sref{sec:04_Introduction}.  

We follow the assumption that the observation model can be reasonably described \textit{locally} by a parametric distribution. Unlike Gaussian noise with fixed standard deviation for all observations, the proposed model employs a normal distribution with spatially varying standard deviation. This property is enforced by assigning pixel-wise confidence weights in spirit of outlier detection in \eref{eqn:04_outlierDetectionDataTerm}. The observation model is given by the zero-mean weighted normal distribution $\NormalDistWeightedT{\LR - \SystemMat \HR}{\Zeros}{\NoiseStd^2 \Id}{\WeightsB}$\label{notation:normalDistWeighted}\label{notation:weightedObservationModel} defined by:
\begin{equation}
	\label{eqn:04_defObservationModel}
	\begin{split}
		\PdfCond{ \LR }{ \HR , \WeightsB }
		&= \NormalDistWeighted{\LR - \SystemMat \HR}{\Zeros}{\NoiseStd^2 \Id}{\WeightsB} \\
		&\defeq \frac{1}{\PartFun(\NoiseStd, \WeightsB)} \exp 
			\left\{- \frac{1}{2 \NoiseStd^2} (\LR - \SystemMat \HR)^\top \WeightsBMat (\LR - \SystemMat \HR) \right\},
	\end{split}
\end{equation}
with normalization constant $\PartFun(\NoiseStd, \WeightsB)$, noise standard deviation $\NoiseStd$, and non-negative confidence weights $\WeightsB \in \RealNonNegN{\NumFrames \LRSize}$\label{notation:weightsB} that are assembled to the diagonal matrix $\WeightsBMat = \Diag{\WeightsBFun_1, \ldots, \WeightsBFun_{\NumFrames \LRSize}}$\label{notation:diag}\label{notation:weightsBMat}. In fact, assuming \iid~ observations, \eref{eqn:04_defObservationModel} defines the normal distribution with spatially varying standard deviation:
\begin{equation}
	\PdfCond{ \LR } { \HR, \WeightsB } 
	\propto 
	\prod_{m = 1}^{\NumFrames \LRSize} \exp \left\{- \frac{1}{2 \sigma_{m}^2} \VecEl{\LR - \SystemMat \HR}{m}^2 \right\},
\end{equation}
where $\sigma_m = \NoiseStd / \sqrt{\WeightsBFun_m}$ with $\WeightsBFun_m \neq 0$ is the standard deviation at the $m$-th pixel.

This observation model shares several conceptual similarities with deterministic outlier detection. However, deterministic outlier detection is designed as a two-stage procedure, where confidence weights are determined on low-resolution data followed by super-resolution. This might lead to the selection of suboptimal confidence weights. In this chapter, we model the weights $\WeightsB$ as latent variables similar to probabilistic outlier detection \cite{Harmeling2010,Cho2011} and estimate them simultaneously to the super-resolved image. This has the inherent advantage that we can gradually refine the confidence weights in an iterative algorithm. 

\subsection{Space Variant Image Prior}
\label{sec:04_SparseImagePriorModel}

Similar to the observation model, most related works on image priors focused on parametric distributions that yield convex regularization terms in certain transform domains. In general, this transform domain is given by $\SparseTransDomain \subset \RealN{\SparseTransDim}$ and a linear sparsifying transform of an image $\HR$ is described by
$\SparseTransSym: \DomainHR \rightarrow \SparseTransDomain$\label{notation:sparseTransFun} with $\SparseTransFun{\HR} = \SparseTransMat \HR$\label{notation:sparseTransMat}, where $\SparseTransMat \in \RealMN{\SparseTransDim}{\HRSize}$\label{notation:sparseTransDomain}\label{notation:sparseTransDim} denotes the transform in matrix notation. The prior distribution exploits the sparse representation of an image under such transforms to regularize super-resolution reconstruction.

Next, different realization for $\SparseTransMat$ and its properties are compared. Based on these findings, a new sparsity-promoting prior for super-resolution is introduced.

\paragraph{Analysis of Natural Image Statistics.}
The choice for the sparsifying transform $\SparseTransMat$ is based on an analysis of natural image statistics. This analysis studies different realizations for $\SparseTransMat$ based on high-pass filtering. These transforms include first-order methods based on the image gradient implemented by the Sobel operator, the Roberts operator as well as \gls{btv} ($L = 2$ and $\BTVWeight = 0.5)$. Moreover, the discrete Laplacian is examined as a second-order transform. For the sake of comparison, we also analyze the identity given by $\SparseTransMat = \IdMN{\HRSize}{\HRSize}$, which is considered as a zeroth-order transform. All transforms are compared by evaluating the sparsity of a transformed image $\vec{z} = \SparseTransMat \HR$. In this context, sparsity refers to the amount of zero elements in the transform domain $\SparseTransDomain$. 

To measure the sparsity of a transformed image $\vec{z}$ quantitatively, the Gini index $s_{\text{Gini}}(\vec{z})$\label{notation:giniIndex}, the Hoyer index $s_{\text{Hoyer}}(\vec{z})$\label{notation:hoyerIndex} as well as the entropy $s_{\text{Entropy}}(\vec{z})$\label{notation:entropyIndex} are used \cite{Hurley2009}. The higher $s_{\text{Gini}}(\vec{z})$ and $s_{\text{Hoyer}}(\vec{z})$ are, the higher the degree of sparsity of $\vec{z}$ obtained by the underlying transform. Conversely, a small entropy $s_{\text{Entropy}}(\vec{z})$ expresses higher sparsity. All measures are analyzed for 29 reference images of natural scenes that are available in the LIVE database \cite{Sheikh2016}, see \tref{tab:04_sparsityMeasures}. The implementation by the identity does not yield a sparse signal as the intensities itself do not follow a sparse distribution. Notice that the highest degree of sparsity is obtained by \gls{btv} indicating its efficiency to design sparse priors. 
\begin{table}[!t]
	\small
	\centering
		\begin{tabular}{lccccc}
			\toprule
			\textbf{Sparsity measure} & \multicolumn{5}{c}{\textbf{Sparsifying transform $\SparseTransFun{\HR}$}} \\
			\cline{2-6}
			& \multicolumn{1}{c}{\textbf{0th order}} & \multicolumn{3}{c}{\textbf{1st order}} &	\multicolumn{1}{c}{\textbf{2nd order}} \\
			& Identity & Roberts & Sobel & \gls{btv} & Laplacian \\
			\midrule
			$s_{\text{Gini}}(\vec{z})$		& 0.26 $\pm$ 0.06  & 0.59 $\pm$ 0.06 & 0.59 $\pm$ 0.07 & \textbf{0.68 $\pm$ 0.05} & 0.61 $\pm$ 0.05 \\
			$s_{\text{Hoyer}}(\vec{z})$		& 0.09 $\pm$ 0.04  & 0.41 $\pm$ 0.08 & 0.41 $\pm$ 0.08 & \textbf{0.53 $\pm$ 0.06} & 0.43 $\pm$ 0.07 \\
			$s_{\text{Entropy}}(\vec{z})$	& 7.33 $\pm$ 0.34 & 5.46 $\pm$ 0.74 & 7.44 $\pm$ 0.76 & \textbf{2.87 $\pm$ 0.58} & 5.64 $\pm$ 0.69 \\
			\bottomrule
		\end{tabular}
	\caption[Sparsity of natural images in different transform domains]{Sparsity of natural images from the LIVE database \cite{Sheikh2016} in different transform domains. The sparsity is measured by mean $\pm$ standard deviation of the Gini index, the Hoyer index, and the entropy. The domains cover zeroth-order (identity), first-order (Roberts gradient, Sobel gradient, \gls{btv}) as well as second-order (Laplacian) transforms.}
	\label{tab:04_sparsityMeasures}
\end{table}

To prove the benefits of \gls{btv} to design the transform $\SparseTransMat$, we analyze the statistical distribution of $\vec{z}$. \Fref{fig:04_btvSparsity} shows the discrete histogram assembled from all transformed samples $z_i$ that are obtained from the 29 reference images using the \gls{btv} model. This demonstrates the clustering of the samples close to zero, which indicates sparsity of the transformed images. In order to model the histogram statistically, we employ the family of Hyper-Laplacian distributions \cite{Krishnan2009a}: 
\begin{equation}
	\begin{split}
		\Pdf{z_i}	
			&= \HyperLapDist{z_i}{\mu_{\text{Prior}}}{\PriorStd}{\nu}\\
			&\defeq \frac{1}{\PartFun(\PriorStd, \nu)} 
			\exp{ \left\{ - \frac{1}{\nu} \left( \frac{| z_i - \mu_{\text{Prior}} |}{\PriorStd} \right)^\nu \right\} },
	\end{split}
\end{equation}
with shape parameter $\mu_{\text{Prior}}$, scale parameters $\PriorStd$ and $\nu$, and normalization constant $\PartFun(\PriorStd, \nu)$\label{notation:hyperLapDist}. This distribution is used to establish the image prior:
\begin{equation}
	\Pdf{\HR} = \prod_{i = 1}^{\SparseTransDim} \HyperLapDist{z_i}{\mu_{\text{Prior}}}{\PriorStd}{\nu},
\end{equation}
where $z_1, \ldots, z_{\SparseTransDim}$ are assumed to be \iid~variables. The Hyper-Laplacian is fitted to the discrete samples by \gls{ml} estimation using $\nu = 2$ corresponding to a Gaussian $\NormalDistT{z_i}{\mu_{\text{Prior}}}{\PriorStd^2}$, $\nu = 1$ corresponding to a Laplacian $\LapDistT{z_i}{\mu_{\text{Prior}}}{\PriorStd}$, as well as $\nu = 0.6$ corresponding to a heavy-tailed Hyper-Laplacian distribution. 

Note that except for small $z_i$, Gaussian and Laplacian distributions provide poor fits to the statistical appearance of natural images. In \fref{fig:04_btvSparsity}, this is visible by the poor approximation of the histogram tails resulting in an inappropriate modeling of discontinuities. For $\nu < 1$, $\PdfT{z_i}$ follows a heavy-tailed distribution that is able to characterize the histogram tails in a reasonable way. This provides a better fit to the appearance of natural images than the Laplacian, which is consistent with recent findings in the area of natural scene statistics \cite{Huang1999,Srivastava2003a}. For this reason, heavy-tailed priors became a common tool for image restoration \cite{Levin2009,Krishnan2009a,Kotera2013a} and compressed sensing \cite{Candes2008,Daubechies2010}. 
\begin{figure}[!t]
	\scriptsize 
	\centering
		\begin{minipage}[c]{0.39\textwidth}
			\centering
			\includegraphics[width=1.00\textwidth]{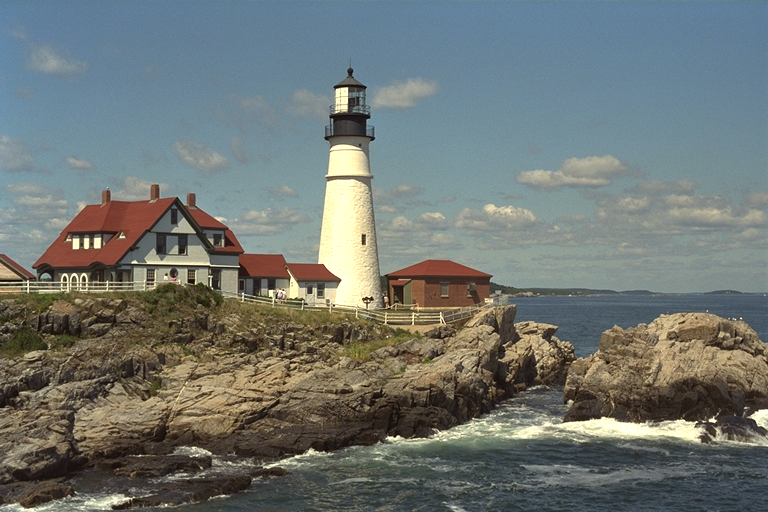}
		\end{minipage}
		~
		\begin{minipage}[c]{0.59\textwidth}
			\setlength \figurewidth{0.78\textwidth}
			\setlength \figureheight{0.65\figurewidth} 
			\centering
			\input{images/chapter4/btvSparsityHistogram.tikz}
		\end{minipage}
	\vspace{-0.5em}
	\caption[Analysis of the \gls{btv} model on natural images]{Analysis of the distribution $\PdfT{z_i}$ on 29 natural images \cite{Sheikh2016} (left) using the \gls{btv} model. The discrete histogram (right) represents the empirical distribution of $z_i$ in the reference images. The histogram is approximated by Hyper-Laplacian distributions using $\nu = 2$ (Gaussian), $\nu = 1$ (Laplacian), as well as $\nu = 0.6$ (heavy-tailed Hyper-Laplacian). Note the fit of the histogram tails for $\nu = 0.6$ compared to the Gaussian with $\nu = 2$.}
	\label{fig:04_btvSparsity}
\end{figure}

\paragraph{Weighted Bilateral Total Variation.}
The design of the proposed image prior is motivated by these findings and exploits the sparsity of natural images in a transform domain $\SparseTransDomain$. However, instead of modeling the prior directly as a Hyper-Laplacian distribution, it is defined by the zero-mean weighted Laplacian distribution $\LapDistWeightedT{\SparseTransMat \HR}{\Zeros}{\PriorStd \Id}{\WeightsA}$\label{notation:lapDistWeighted}\label{notation:adaptPrior}:
\begin{equation}
	\label{eqn:04_defAdaptiveImagePrior}
	\begin{split}
		\PdfCond{\HR}{\WeightsA} 
			&= \LapDistWeighted{\SparseTransMat \HR}{\Zeros}{\PriorStd \Id}{\WeightsA} \\
			&\defeq \frac{1}{\PartFun(\PriorStd, \WeightsA)} 
			\exp \left \{- \frac{\left| \left| \WeightsAMat \SparseTransMat \HR \right| \right|_1}{\PriorStd} \right \},
	\end{split}
\end{equation}
where \smash{$\PriorStd$} denotes a distribution scale parameter, \smash{$\WeightsA \in \RealNonNegN{\SparseTransDim}$}\label{notation:weightsA} are confidence weights of the distribution in the transform domain assembled as the diagonal matrix $\WeightsAMat = \Diag{\WeightsAFun_1, \ldots, \WeightsAFun_{\SparseTransDim}}$\label{notation:weightsAMat}, and $\PartFun(\PriorStd, \WeightsA)$ is a normalization constant.

The regularization term associated with the distribution in \eref{eqn:04_defAdaptiveImagePrior} termed \gls{wbtv} is based on the unweighted \gls{btv} in \eref{eqn:03_btvPrior} due to the performance of this approach to yield a sparse transform. Since \smash{$\BTVWeight^{|m| + |n|} > 0$}, we can reformulate the unweighted \gls{btv} according to:
\begin{equation}
	\begin{split}
		\RegTerm[BTV]{\HR}
			&= \sum_{m = -\BTVSize}^\BTVSize \sum_{n = -\BTVSize}^\BTVSize 
			\left| \left| \BTVWeight^{|m| + |n|} \left( \IdMN{\HRSize}{\HRSize} - \vec{S}_v^m \vec{S}_h^n \right) \HR \right| \right|_1 \\
			&= \sum_{m = -\BTVSize}^\BTVSize \sum_{n = -\BTVSize}^\BTVSize \big| \big| \vec{S}^{m,n} \HR \big| \big|_1 
			= \big| \big| \SparseTransMat \HR \big| \big|_1,
	\end{split}
\end{equation}
where $\vec{S}^{m,n} = \BTVWeight^{|m| + |n|} (\IdMN{\HRSize}{\HRSize} - \vec{S}_v^m \vec{S}_h^n) \in \RealMN{\HRSize}{\HRSize}$ denotes the transform associated with the shift $(m, n)$. The overall transform $\SparseTransMat \in \RealMN{\SparseTransDim}{\HRSize}$ with $\SparseTransDim = (2 \BTVSize + 1)^2\HRSize$ for all shifts is assembled as:
\begin{equation}
	\label{eqn:04_btvSparsifyingTransform}
	\SparseTransMat = 
	\begin{pmatrix}
		\vec{S}^{-\BTVSize,-\BTVSize}~
		& \vec{S}^{-\BTVSize+1,-\BTVSize}~
		& \hdots
		& \vec{S}^{\BTVSize-1,\BTVSize}~
		& \vec{S}^{\BTVSize,\BTVSize}~ 
	\end{pmatrix}^\top.
\end{equation}
Then, \gls{wbtv} regularization is conditioned on the weights $\WeightsA$ according to:
\begin{equation}
	\RegTerm[WBTV]{\HR\,|\,\WeightsA}	
		\defeq \left| \left| \WeightsAMat \SparseTransMat \HR \right| \right|_1
		= \sum_{m = -\BTVSize}^\BTVSize \sum_{n = -\BTVSize}^\BTVSize 
			\sum_{i = 1}^\HRSize \alpha^{m,n}_i \left| \VecEl{\vec{S}^{m,n} \HR}{i} \right|,
\end{equation}
where \smash{$\WeightsA = (\WeightsA^{-\BTVSize,-\BTVSize}, \ldots, \WeightsA^{\BTVSize,\BTVSize})^\top$} denotes the joint weight vector over all shifts and \smash{$\WeightsA^{m,n} = (\alpha^{m,n}_1, \ldots, \alpha^{m,n}_N)^\top$} are weights associated with the shift $(m, n)$. 

This term allows us to locally adapt the prior $\PdfCondT{\HR}{\WeightsA}$ by controlling the weights $\WeightsA$ similar to the locally adaptive \gls{btv} introduced by Li \etal \cite{Li2010}. In particular, the impact of the regularization needs to be reduced on discontinuities compared to the behavior in flat regions. However, unlike \cite{Li2010}, the weights are handled as latent variables in the same way as those of the observation model. This avoids their explicit computation by means of feature detection in a preprocessing step.

\subsection{Inference of the Model Confidence Weights}
\label{sec:04_BayesianInferenceOfTheConfidenceWeights}

Our goal is to reconstruct the high-resolution image that best explains a set of low-resolution observations. If one knows the confidence weights employed in the Bayesian model, the high-resolution image could be inferred by the \gls{map} framework presented in \sref{sec:03_MaximumAPosterioriEstimation}. However, if one does not know the  weights, they need to be treated as latent variables in the estimation of the high-resolution image. For this purpose, three alternative approaches are examined.

\paragraph{Bayesian Marginalization.}

One approach is to marginalize over the latent variables $\WeightsA$ and $\WeightsB$. Then, the high-resolution image is estimated from the marginal distribution. This Bayesian marginalization is formulated by:
\begin{equation}
	\hat{\HR} = \argmax_{\HR} \int_{\RealN{\NumFrames \LRSize}} \int_{\RealN{\SparseTransDim}}
		\PdfCond{\LR, \WeightsB}{\HR} \Pdf{\HR, \WeightsA} d\WeightsA d\WeightsB,
\end{equation}
where the integration is performed over all configurations of the confidence maps. Using the Bayes rule, marginalization over $\WeightsA$ and $\WeightsB$ yields:
\begin{equation}
	\hat{\HR}	= \argmax_{\HR} \int_{\RealNonNegN{\NumFrames \LRSize}}
		\PdfCond{\LR}{\WeightsB, \HR} \Pdf{\WeightsB} \int_{\RealNonNegN{\SparseTransDim}} 
		\PdfCond{\HR}{\WeightsA} \Pdf{\WeightsA} d\WeightsA d\WeightsB,
\end{equation}
where $\Pdf{\WeightsA}$ and $\Pdf{\WeightsB}$ are the prior distributions assigned to the confidence weights.

Although this approach provides a theoretical basis for super-resolution under unknown confidence weights, there exist several practical limitations. One important restriction is that analytic marginalization is possible only for few relatively simple priors\footnote{For instance, conjugate priors \cite{Bishop2006} to the data likelihood can be used. These priors enable an analytic calculation of the marginal distribution to infer latent hyperparameters \cite{Oliveira2009}.} or with simplistic approximations of the integration. An exact solution would require integration in a $\NumFrames \LRSize + \SparseTransDim$ dimensional space. This is computationally prohibitive for real-world applications, where the size of the parameter space lies in the range $\NumFrames \LRSize + \SparseTransDim \approx 10^6$. Another limitation is the parameter tuning that is required to define $\PdfT{\WeightsA}$ and $\PdfT{\WeightsB}$ as these distributions would comprise additional hyperparameters.

\paragraph{Alternating MAP Estimation.}

As an alternative to marginalization, one can jointly estimate the super-resolved image and the confidence weights:
\begin{equation}
	\label{eqn:04_alternatingMapEstimationWithConfidenceWeights}
	(\hat{\HR}, \hat{\WeightsA}, \hat{\WeightsB})	
	= \argmax_{\HR, \WeightsA, \WeightsB} 
	\left \{ 
		\PdfCond{\HR}{\WeightsA} \prod_{k = 1}^{\NumFrames} \PdfCond{\LRFrame{k}}{\HR, \FrameIdx{\WeightsB}{k}} 
		\Pdf{\WeightsA} \Pdf{\WeightsB}
	\right \},
\end{equation}
where $\PdfT{\WeightsA}$ and $\PdfT{\WeightsB}$ are priors for the confidence weights to obtain meaningful solutions of this underdetermined problem. Taking the negative log-likelihood of \eref{eqn:04_alternatingMapEstimationWithConfidenceWeights} leads to the joint energy minimization problem:
\begin{equation}
	\begin{split}
		(\hat{\HR}, \hat{\WeightsA}, \hat{\WeightsB})
		&= \argmin_{\HR, \WeightsA, \WeightsB} 
		\Bigg\{ 
			\sum_{i = 1}^{\NumFrames \LRSize} \WeightsBFun_i \left| \VecEl{\LR - \SystemMat \HR }{i} \right|^2
			+ \RegWeight \sum_{i = 1}^{\SparseTransDim} \WeightsAFun_i \left| \VecEl{\SparseTransMat \HR}{i} \right| \\
			&+ \log \PartFun(\NoiseStd, \WeightsB) + \log \PartFun(\PriorStd, \WeightsA)
			- \log \Pdf{\WeightsB} - \log \Pdf{\WeightsA} 
		\Bigg\}.
	\end{split}
\end{equation}

This minimization problem can be solved by alternating \gls{map} estimation for $\WeightsA$, $\WeightsB$, and $\HR$. Hamada \etal \cite{Hamada2013} investigated this approach for a simplified model, where only confidence weights of a data fidelity term are taken into account. However, similar to Bayesian marginalization, the performance is highly dependent on the priors $\PdfT{\WeightsA}$ and $\PdfT{\WeightsB}$ and only tractable for simplistic models. 

\paragraph{Majorization-Minimization.}

Another approach to infer the confidence weights is to treat them as hidden information within an \gls{em} algorithm \cite{Dempster1977}. Related schemes have been successfully applied for probabilistic outlier removal \cite{Cho2011}. However, similar to the aforementioned approaches, this concept requires a pure probabilistic formulation, \ie an explicit definition of distributions $\PdfT{\WeightsA}$ and $\PdfT{\WeightsB}$. These distributions are difficult to model for real-world problems, which limits the flexibility. For this reason, the proposed method is formulated as \gls{mm} algorithm \cite{Hunter2004} as generalization of \gls{em} \cite{Ba2014}. This does not require explicit modeling of $\PdfT{\WeightsA}$ and $\PdfT{\WeightsB}$, and yields a computationally efficient approach for the inference of the confidence weights via \textit{iteratively re-weighted minimization} \cite{Candes2008,Scales1988}.

The iteratively re-weighted minimization framework comprises two steps, which results in a sequence of iterations $\{ (\IterationIdx{\HR}{t}, \IterationIdx{\WeightsA}{t}, \IterationIdx{\WeightsB}{t}): t = 1, \ldots, \NumIter \}$: 
\begin{enumerate}
	\item Given an estimate $\IterationIdx{\HR}{t-1}$ for the latent high-resolution image, we first determine $\IterationIdx{\WeightsA}{t}$ and $\IterationIdx{\WeightsB}{t}$ according to\label{notation:weightsAFun}\label{notation:weightsBFun}:
\begin{align}
	\IterationIdx{\WeightsA}{t} &= \WeightsAFun \left( \IterationIdx{\HR}{t-1} \right), \\
	\IterationIdx{\WeightsB}{t} &= \WeightsBFun \left( \IterationIdx{\HR}{t-1}, \LRFrame{1}, \ldots, \LRFrame{\NumFrames} \right),
\end{align}
where \smash{$\WeightsAFun: \RealN{\SparseTransDim} \rightarrow \RealNonNegN{\SparseTransDim}$ and $\WeightsBFun: \RealN{\NumFrames \LRSize} \rightarrow \RealNonNegN{\NumFrames \LRSize}$} are \textit{weighting functions} to compute confidence weights based on $\IterationIdx{\HR}{t-1}$ and $\LR$.

	\item Given the weights $\IterationIdx{\WeightsA}{t}$ and $\IterationIdx{\WeightsB}{t}$, we determine $\IterationIdx{\HR}{t}$ according to the solution of the weighted minimization problem:
\begin{equation}
	\label{eqn:04_updateEquationHRImage}
	\IterationIdx{\HR}{t}
	= \argmax_{\HR} 
	\left\{ \PdfCond{\HR}{\IterationIdx{\WeightsA}{t}} \PdfCond{\LR}{\HR, \IterationIdx{\WeightsB}{t}} \right\}.
\end{equation}
\end{enumerate}
Both steps are alternated until convergence. For a detailed analysis of the relationship between this scheme and \gls{mm} algorithms, we refer to \sref{sec:04_AlgorithmAnalysis}.		

\section{Robust Super-Resolution Reconstruction}
\label{sec:04_RobustSuperResolutionReconstruction}

This section introduces a robust super-resolution algorithm based on iteratively re-weighted minimization. For the derivation of this algorithm, the basic computational steps for numerical optimization are outlined. Eventually, a theoretical study of the underlying Bayesian model and the proposed iteration scheme is provided by explicitly deriving this method as \gls{mm} algorithm.

\subsection{Iteratively Re-Weighted Minimization Algorithm}
\label{sec:04_NumericalOptimization}

The general iteratively re-weighted minimization framework has some degrees of freedom that need to be adjusted for robust super-resolution. First, it utilizes weighting functions that need to be specified. Second, it assumes a fixed regularization weight $\RegWeight$ that is adjusted prior to the iterative procedure. Hence, this approach is not adaptive regarding the characteristics of the low-resolution data.

The proposed algorithm is developed as adaptive iteration scheme. The weights $\IterationIdx{\WeightsA}{t}$ and $\IterationIdx{\WeightsB}{t}$ are inferred by:
\begin{align}
	\label{eqn:04_defWeightFunctionObservation}
	\IterationIdx{\WeightsA}{t} &= \WeightsAFun \left( \IterationIdx{\HR}{t-1}, \IterationIdx{\NoiseStd}{t} \right), \\
	\label{eqn:04_defWeightFunctionPrior}
	\IterationIdx{\WeightsB}{t} &= 
	\WeightsBFun \left( \IterationIdx{\HR}{t-1}, \LRFrame{1}, \ldots, \LRFrame{\NumFrames} , \IterationIdx{\PriorStd}{t} \right),
\end{align}
where \smash{$\WeightsAFun: \RealN{\SparseTransDim} \rightarrow \RealNonNegN{\SparseTransDim}$ and $\WeightsBFun: \RealN{\NumFrames \LRSize} \rightarrow \RealNonNegN{\NumFrames \LRSize}$} are adaptive weighting functions that exploit \smash{$\IterationIdx{\NoiseStd}{t}$ and $\IterationIdx{\PriorStd}{t}$} as scale parameters of the observation and the prior model. These scale parameters are adjusted at each iteration and characterize the data uncertainty in the underlying Bayesian model. To avoid manual parameter tuning, automatic hyperparameter estimation is used to determine the regularization weight \smash{$\IterationIdx{\RegWeight}{t}$} at iteration $t$. Finally, the super-resolved image $\IterationIdx{\HR}{t}$ is reconstructed using the confidence weights $\IterationIdx{\WeightsA}{t}$ and $\IterationIdx{\WeightsB}{t}$ as well as the regularization weight $\IterationIdx{\RegWeight}{t}$.

\paragraph{Weight Estimation.}

To determine the confidence weights $\IterationIdx{\WeightsB}{t}$ of the observation model, the residual error $\ResidualFun{\HR}{\LR} = \LR - \SystemMat \HR$ is analyzed. In this work, the weighting function in \eref{eqn:04_defWeightFunctionObservation} is defined element-wise according to: 
\begin{equation}
	\WeightsBFun \big( \HR, \LR, \NoiseStd \big) 
	\defeq 
	\begin{pmatrix} 
		\WeightsBFun_1 \left( \Residual, \NoiseStd \right) &
		\ldots &
		\WeightsBFun_{\NumFrames \LRSize} \left( \Residual, \NoiseStd \right)
	\end{pmatrix}^\top \in \RealNonNegN{\NumFrames \LRSize},	
\end{equation}
where \smash{$\Residual = \ResidualFun{\IterationIdx{\HR}{t-1}}{\LR}$} denotes the residual error associated with the estimate $\IterationIdx{\HR}{t-1}$ obtained at the previous iteration, and $\WeightsBFun_i: \RealN{\NumFrames \LRSize} \rightarrow \RealNonNeg$ determines the weight for the $i$-th observation. The confidence weights are computed by considering frame-wise (global) outliers as well pixel-wise (local) outliers via the decomposition:
\begin{equation}
	\label{eqn:04_dataFidelityWeights}
	\WeightsBFun_i \big(\Residual, \NoiseStd \big) 
	\defeq 
	\underbrace{\WeightsBFun_{i, \text{bias}} \left( \Residual \right)}_{\text{frame-wise}} 
	\cdot 
	\underbrace{\WeightsBFun_{i, \text{local}} \left( \Residual, \NoiseStd \right)}_{\text{pixel-wise}}.
\end{equation}

In order to detect outlier frames, we assume that the residual errors associated with the different frames need to be symmetric and zero-mean according to \eref{eqn:04_defObservationModel}. Individual frames that violate this assumption are considered as outliers. Potential reasons for a violation of this assumption could be systematic errors like global photometric differences between the frames. We perform a bias detection \cite{Zomet2001} to identify such frames using the binary weighting function:
\begin{equation}
	\label{eqn:04_biasDataFidelityWeights}
	\begin{split}
		\WeightsBFun_{i, \text{bias}} \big( \Residual \big)
		=
		\begin{cases}
			1		& \text{if}~ \big| \text{median} \big( \FrameIdx{\Residual}{k} \big) \big| \leq c_{\text{bias}} \\
			0		& \text{otherwise}
		\end{cases}
	\end{split},
\end{equation}
where \smash{$\FrameIdx{\Residual}{k}$} is the residual error of the $k$-th frame associated with the $i$-th observation, and \smash{$\Median{\cdot}$} denotes the sample median as robust estimator of the mean residual error \cite{Zoubir2012}.

In addition to the detection of outlier frames, local outliers are detected pixel-wise using the bi-weight function:
\begin{equation}
	\label{eqn:04_localDataFidelityWeights}
	\WeightsBFun_{i,\text{local}} \big( \Residual, \NoiseStd \big) =
	\begin{cases}
		1																										
			& \text{if}~ \left| \ResidualSym_i \right| \leq c_{\text{local}} \NoiseStd \\
		\frac{c_{\text{local}} \NoiseStd}{ \left| \ResidualSym_i \right|}	
			& \text{otherwise}
	\end{cases},
\end{equation}
where $\NoiseStd$ denotes an estimate of the standard deviation of the weighted normal distribution $\NormalDistWeighted{\Residual}{\Zeros}{\NoiseStd^2 \Id}{\IterationIdx{\WeightsB}{t-1}}$ and $c_{\text{local}}$ is a tuning constant. Notice that a constant confidence is assigned to observations classified as inliers of a normal distribution, whereas outliers are weighted by their inverse residual errors. This function can downweight outliers related to non-Gaussian noise, \eg impulsive noise, or locally inaccurate motion estimation.    

The estimation of the prior weights $\IterationIdx{\WeightsA}{t}$ in \eref{eqn:04_defWeightFunctionPrior} follows a similar motivation and is done under the transform $\vec{z} = \SparseTransMat \IterationIdx{\HR}{t-1}$ according to:
\begin{equation}
	\WeightsAFun( \HR, \PriorStd) 
	\defeq
	\begin{pmatrix} 
		\WeightsAFun_1 \big( \vec{z}, \PriorStd \big) & \ldots & \WeightsAFun_{\SparseTransDim} \left( \vec{z}, \PriorStd \right) 
	\end{pmatrix}^\top \in \RealNonNegN{\SparseTransDim}.
\end{equation}
The $i$-th weight is computed by the weighting function:
\begin{equation}
	\label{eqn:04_regularizationWeights}
	\WeightsAFun_i \big(\vec{z}, \PriorStd \big) =
	\begin{cases}
		1	
			& \text{if}~ \left| \VecEl{Q (\vec{z})}{i} \right| \leq c_{\text{prior}} \PriorStd \\
		\SparsityParam \frac{(c_{\text{prior}} \PriorStd)^{1-\SparsityParam}}{\left| \VecEl{Q (\vec{z})}{i} \right|^{1-\SparsityParam}}
			& \text{otherwise}
	\end{cases},
\end{equation}
where $\SparsityParam \in [0, 1]$\label{notation:sparsityParam} is referred to as \textit{sparsity parameter}, \smash{$\PriorStd$} is an estimate of the scale parameter of the weighted Laplacian distribution $\LapDistWeighted{\vec{z}}{\Zeros}{\PriorStd \Id}{\IterationIdx{\WeightsA}{t-1}}$ and $c_{\text{prior}}$ is a tuning constant. Note that in order to reduce the influence of isolated noisy pixels, these weights are inferred from a locally filtered version of the spatial information denoted by $Q ( \vec{z} )$. In this work, $Q(\cdot)$ is implemented by a $3 \times 3$ median filtering\footnote{In \cite{Yuan2013}, a related method has been proposed, where median filtering is used to extract edge information. This avoids the origination of false edges in spatially adaptive \gls{tv} regularization.}. This scheme explains an image as a mixture of flat regions and discontinuities by assigning spatially adaptive weights. Accordingly, higher weights are assigned to flat regions while the influence of discontinuities is downweighted.

\begin{figure}[!t]
	\centering
	\subfloat{
		\hspace{-0.7em}
		\begin{tikzpicture}[spy using outlines={rectangle,red,magnification=3.25, height=2.9cm, width = 2.9cm, connect spies, every spy on node/.append style={thick}}] 
			\node {\pgfimage[width=0.256\linewidth]{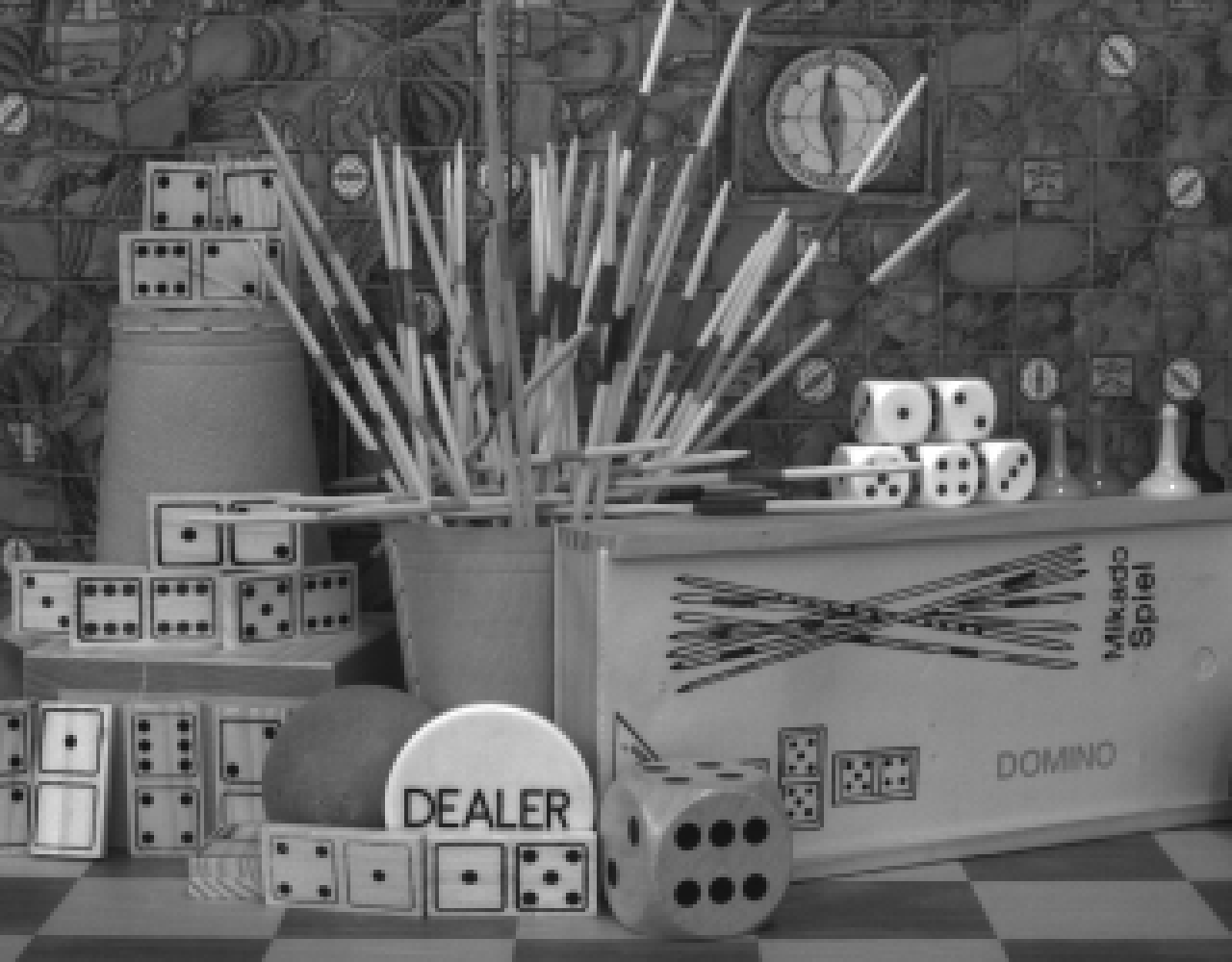}}; 
      \spy on (0.5, 0.4) in node [left] at (4.92, 0); 
    \end{tikzpicture}
	}~
	\subfloat{\includegraphics[width=0.256\textwidth]{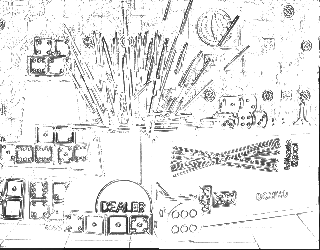}}~
	\subfloat{\includegraphics[width=0.256\textwidth]{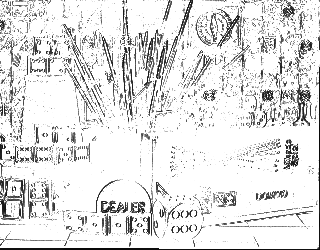}}\\[-0.65ex]
	\setcounter{subfigure}{0}
	\subfloat[]{
		\hspace{-0.7em}
		\begin{tikzpicture}[spy using outlines={rectangle,red,magnification=3.25, height=2.9cm, width = 2.9cm, connect spies, every spy on node/.append style={thick}}] 
			\node {\pgfimage[width=0.256\linewidth]{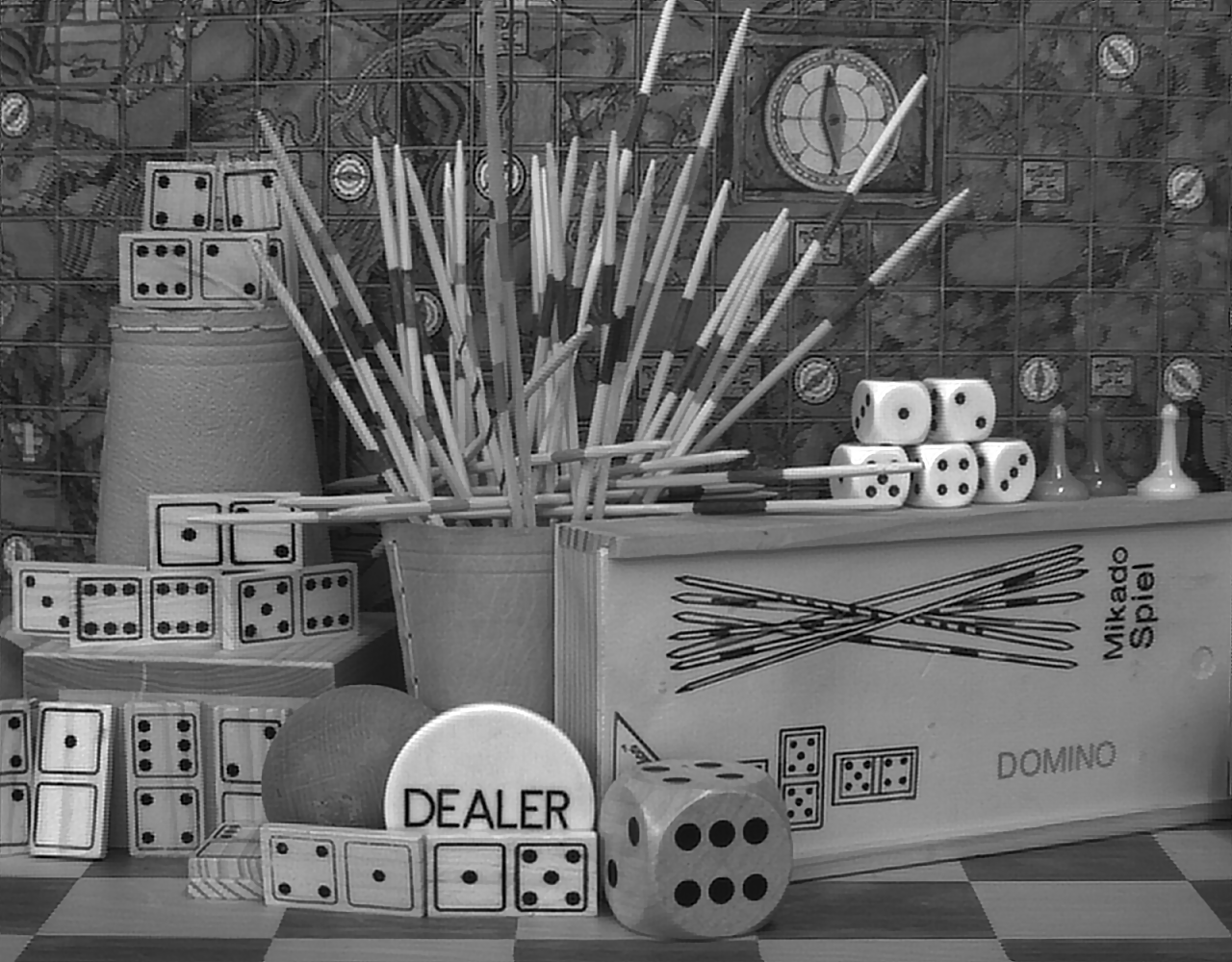}}; 
      \spy on (0.5, 0.4) in node [left] at (4.92, 0); 
    \end{tikzpicture}
		\label{fig:confidenceWeightsExample:image}
	}~
	\subfloat[]{\includegraphics[width=0.256\textwidth]{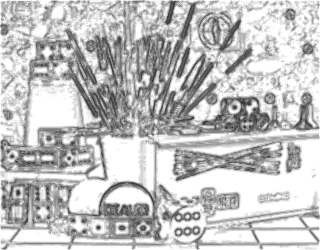}\label{fig:confidenceWeightsExample:iter1}}~
	\subfloat[]{\includegraphics[width=0.256\textwidth]{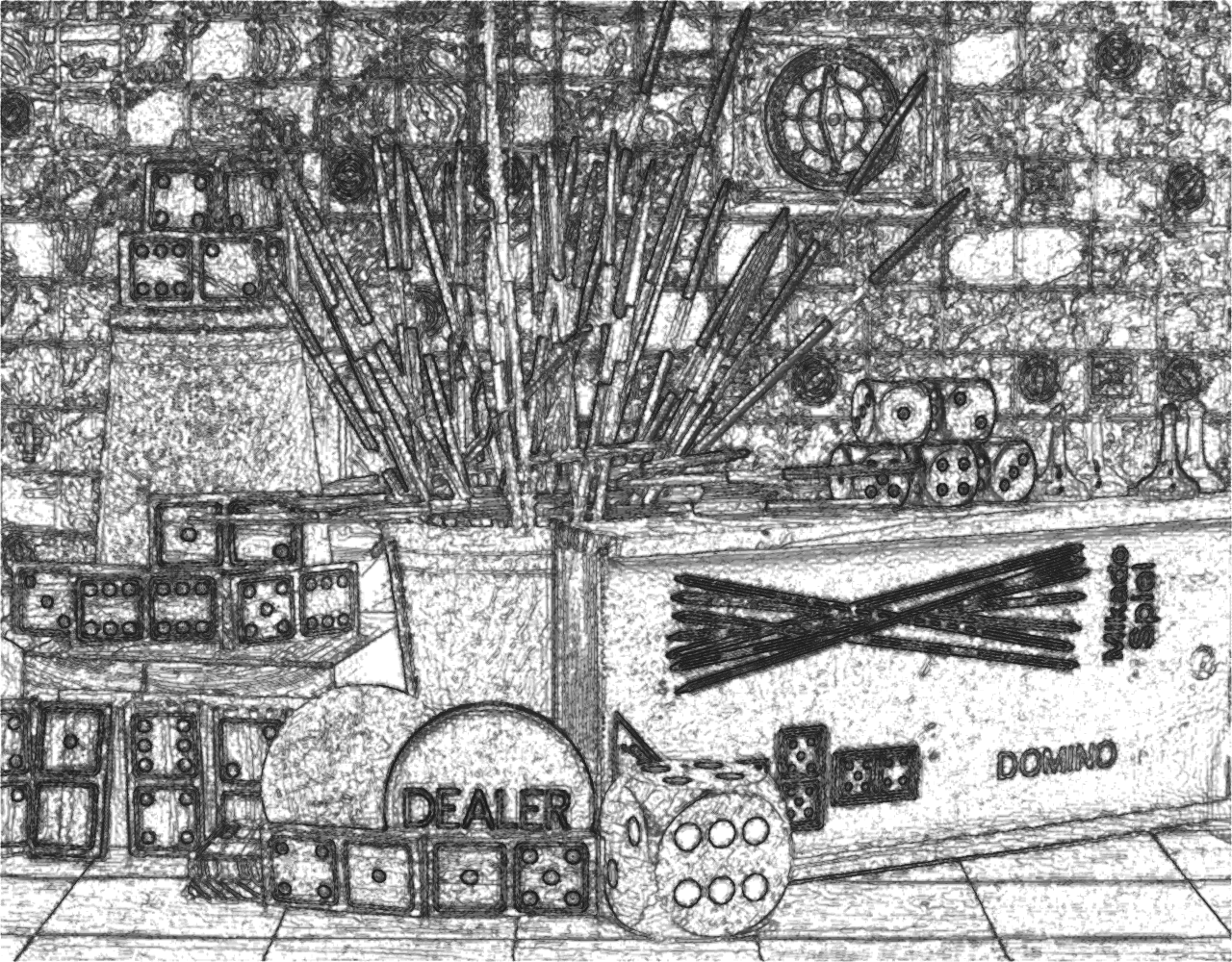}\label{fig:confidenceWeightsExample:iter10}}
	\caption[Confidence weighting on an example image sequence]{Illustration of the proposed confidence weighting on an example image sequence. \protect\subref{fig:confidenceWeightsExample:image} Low-resolution image (top row) and super-resolved image with $4 \times$ magnification (bottom row) along with a zoom-in. \protect\subref{fig:confidenceWeightsExample:iter1} - \protect\subref{fig:confidenceWeightsExample:iter10} Gray-scale visualizations of the observation confidence weights (top row) and the prior weights (bottom row) after the 1st and the 10th iteration, respectively (bright regions denote higher weights). The observation weights identify outlier observations (\eg due to inaccurate motion estimation) while the prior weights extract image structures for adaptive regularization.}
	\label{fig:04_confidenceWeightsExample}
\end{figure}

In \fref{fig:04_confidenceWeightsExample}, we illustrate the proposed weighting functions employed for iteratively re-weighted minimization. \Fref{fig:04_confidenceWeightsExample} (top row) depicts the observation confidence weights associated with a single low-resolution frame affected by insufficient motion estimation. The weights are iteratively refined and model the low-resolution observations by mixed noise. In this example, mixed noise is related to the superposition of measurement noise and motion estimation uncertainty. \Fref{fig:04_confidenceWeightsExample} (bottom row) depicts the prior weights in the domain of the super-resolved image. These weights are gradually refined over the iterations in order to make regularization spatially adaptive \wrt image structures. More specifically, lower weights are assigned to sharp edges to enhance their reconstruction. 

\paragraph{Scale Parameter Estimation.}

The weighting functions in \eref{eqn:04_localDataFidelityWeights} and \eref{eqn:04_regularizationWeights} require the knowledge of the scale parameters $\NoiseStd$ and $\PriorStd$, respectively. In order to avoid manual parameter tuning, both parameters are determined in an optimal way at each iteration. Given $\IterationIdx{\HR}{t-1}$ and $\IterationIdx{\WeightsB}{t-1}$ obtained at the previous iteration and assuming a uniform prior $\PdfT{\NoiseStd}$, we determine \smash{$\IterationIdx{\NoiseStd}{t}$} via the \gls{ml} estimator:  
\begin{equation}
	\label{eqn:04_observationNoiseStd}
	\IterationIdx{\NoiseStd}{t} = \argmax_{\NoiseStd} \PdfCond{\LR}{\IterationIdx{\HR}{t-1}, \IterationIdx{\WeightsB}{t-1}, \NoiseStd}.
\end{equation}
For outlier-insensitive estimation in \eref{eqn:04_observationNoiseStd}, the scale parameter is computed from the \gls{mad} \cite{Scales1988} of \smash{$\IterationIdx{\Residual}{t-1} = \LR - \SystemMat \IterationIdx{\HR}{t-1}$}. In order to take different confidence weights associated with the low-resolution observations into account, the \gls{mad} is computed in a weighted version using the weighted median \cite{Gabbouj1996,Zhang2014}\label{notation:wMedian}\label{notation:wMad}:
\begin{equation}
	\label{eqn:04_observationNoiseStdRobust}
	\begin{split}
		\IterationIdx{\NoiseStd}{t}	
			&= \sigma_0 \cdot \WMad { \IterationIdx{\Residual}{t-1} } { \IterationIdx{\WeightsB}{t-1} } \\
			&= \sigma_0 \cdot 
					\WMedian{ 
						\begin{pmatrix}
							\big| \IterationIdx{\ResidualSym_1}{t-1} - \WMedian{\IterationIdx{\Residual}{t-1}}{\IterationIdx{\WeightsB}{t-1}} \big| \\
							\vdots \\
							\big| \IterationIdx{\ResidualSym_{\NumFrames \LRSize}}{t-1} - \WMedian{\IterationIdx{\Residual}{t-1}}{\IterationIdx{\WeightsB}{t-1}} \big|
						\end{pmatrix}
					}
					{
						\begin{pmatrix}
							\IterationIdx{\WeightsBFun}{t-1}_1 \\ \vdots \\ \IterationIdx{\WeightsBFun}{t-1}_{\NumFrames \LRSize}
						\end{pmatrix}
					},
	\end{split}
\end{equation}
where we set \smash{$\sigma_0 = 1.4826$} to obtain a consistent estimate for the standard deviation of a normal distribution \cite{Scales1988}. The quantities $\WMad{\Residual}{\WeightsB}$ and $\WMedian{\Residual}{\WeightsB}$ denote the weighted \gls{mad} and the weighted median of the residual error $\Residual$ under the confidence weights $\WeightsB$, respectively. The weighted median $\tilde{\ResidualSym} = \WMedian{\Residual}{\WeightsB}$ generalizes the sample median and is defined as the point $\tilde{\ResidualSym}$, where the sum of the weights $\WeightsBFun_i$ associated with residuals $\ResidualSym_i$ above and below $\tilde{\ResidualSym}$ fulfills:  
\begin{equation}
	\sum_{i : \ResidualSym_i < \tilde{\ResidualSym}} \WeightsBFun_i < \frac{1}{2} \sum_{i} \WeightsBFun_i
	\quad\text{and}\quad
	\sum_{i : \ResidualSym_i \geq \tilde{\ResidualSym}} \WeightsBFun_i \leq \frac{1}{2} \sum_{i} \WeightsBFun_i.
\end{equation}

Similarly, the \gls{ml} estimate for the scale parameter $\IterationIdx{\PriorStd}{t}$ is obtained from the distribution of \smash{$\SparseTransMat \IterationIdx{\HR}{t-1}$}. Given the weights \smash{$\IterationIdx{\WeightsA}{t-1}$} determined at the previous iteration, the scale parameter at iteration $t$ is determined by:
\begin{equation}
	\label{eqn:04_btvNoiseStd}
	\IterationIdx{\PriorStd}{t}	= \sigma_0 \cdot \WMad { Q \big( \SparseTransMat \IterationIdx{\HR}{t-1} \big) }{ \IterationIdx{\WeightsA}{t-1} },
\end{equation}
where $\sigma_0 = 1$ for the Laplacian distribution.

\paragraph{Hyperparameter Estimation.}

The selection of the regularization parameter $\RegWeight$ has to deal with the following inherent tradeoff. On the one hand, if $\RegWeight$ is underestimated, super-resolution is ill-conditioned and the reconstructed images are affected by residual noise. On the other hand, in case of an overestimate, the super-resolved images get blurred as illustrated in \fref{fig:04_robustnessRegularizationWeight}. In general, an optimal regularization weight is unknown and manual tuning based on trial-and-error procedures is time-consuming and error prone. In the proposed approach, an optimal $\RegWeight$ also depends on the estimated confidence weights. Fully automatic approaches to select $\RegWeight$ use, e.\,g. \gls{gcv} \cite{Nguyen2001}, the discrepancy principle \cite{Wen2012a}, or Bayesian methods \cite{Oliveira2009,Babacan2011}. Typically these methods deal with simplistic prior distributions, \eg Gaussian priors \cite{Vrigkas2014}, or use approximative schemes to determine a closed-form solution for the prior partition function \cite{Oliveira2009} to make parameter selection tractable. 

In this work, a data-driven parameter selection that generalizes fairly well to different forms of the image prior is used. This approach is inspired by the work of Pickup \etal \cite{Pickup2007} and is based on a two-fold cross validation like procedure that estimates the regularization parameter $\RegWeight$ jointly with the super-resolved image. The advantage of this approach is that $\RegWeight$ is adjusted at each iteration $t$ as $\IterationIdx{\RegWeight}{t}$ and the parameter selection exploits the model confidence weights as opposed to parameter selection prior to super-resolution. The key idea is to determine $\IterationIdx{\RegWeight}{t}$ based on training observations such that it minimizes a cross validation error on a disjoint set of validation observations. For this purpose, the low-resolution observations $\LR$ are decomposed into two disjoint subsets, where a fraction of $\delta$, $0 < \delta < 1$ observations are used for parameter training and the remaining observations are hold back for validation. This is achieved by assembling a binary diagonal matrix $\Id_{\delta} \in \{0, 1 \}^{\NumFrames \LRSize \times \NumFrames \LRSize}$\label{notation:trainingObservations}, where the $i$-th element is $I_{\delta,i} = 1$ with probability $\delta$ to specify the training subset and $I_{\delta,i} = 0$ with probability $1 - \delta$ to specify the validation subset. Given a regularization weight $\RegWeight$, the super-resolved image reconstructed with this setting from the training observations is denoted by:
\begin{equation}
	\HR(\RegWeight) 
	= \argmin_{\HR} 
	\left\{ \big( \LR - \SystemMat \HR \big)^\top \Id_{\delta} \IterationIdx{\WeightsBMat}{t} \big( \LR - \SystemMat \HR \big) 
	+ \RegWeight \big| \big| \IterationIdx{\WeightsAMat}{t} \SparseTransMat \HR \big| \big|_1 \right\},
\end{equation}
where $\IterationIdx{\WeightsAMat}{t} = \Diag{\IterationIdx{\WeightsA}{t}}$ and $\IterationIdx{\WeightsBMat}{t} = \Diag{\IterationIdx{\WeightsB}{t}}$. The optimal weight $\IterationIdx{\RegWeight}{t}$ is determined from the validation observations according to:
\begin{equation}
	\label{eqn:04_cvObjectiveFunction}
	\IterationIdx{\RegWeight}{t} = \argmin_{\RegWeight} \DataTerm[\text{cv}]{\RegWeight, \overline{\Id_{\delta}} }. 
\end{equation}
The cross validation error measures the fidelity of $\HR(\RegWeight)$ on the validation set and is given by:
\begin{equation}
	\label{eqn:04_cvError}
	\DataTerm[\text{cv}]{\RegWeight, \overline{\Id_{\delta}} } 
	= \big( \LR - \SystemMat \HR(\RegWeight) \big)^\top 
	\overline{\Id_{\delta}} \IterationIdx{\WeightsBMat}{t} 
	\big( \LR - \SystemMat \HR(\RegWeight) \big),
\end{equation}
where $\overline{\Id_{\delta}}$\label{notation:validationObservations}\label{notation:cvError} is obtained from $\Id_{\delta}$ by flipping the diagonal elements. The behavior of this cross validation error is visualized in \fref{fig:04_cvError} in the log-transformed range $\log \RegWeight$. Unlike the error on the training observations $\DataTerm[\text{cv}]{\RegWeight, \Id_{\delta} }$ that is strictly monotonic increasing in \fref{fig:04_cvError:train}, the optimal regularization weight is the minimum of $\DataTerm[\text{cv}]{\RegWeight, \overline{\Id_{\delta}} }$ in \fref{fig:04_cvError:valid}.
\begin{figure}[!t]
	\scriptsize 
	\centering
	\setlength \figurewidth{0.39\textwidth}
	\setlength \figureheight{0.78\figurewidth} 
	\subfloat[Error on the validation observations]{\input{images/chapter4/cvError_validation.tikz}\label{fig:04_cvError:valid}}\quad
	\subfloat[Error on the training observations]{\input{images/chapter4/cvError_training.tikz}\label{fig:04_cvError:train}}
	\caption[Cross validation error for the selection of the regularization weight]{Behavior of the cross validation error for the selection of an optimal regularization weight. \protect\subref{fig:04_cvError:valid} Cross validation error $\DataTerm[\text{cv}]{\RegWeight, \overline{\Id_{\delta}} }$ in the log-transformed range $\log \RegWeight$ on the validation observations. \protect\subref{fig:04_cvError:train} Error $\DataTerm[\text{cv}]{\RegWeight, \Id_{\delta} }$ on the training observations. Note that $\DataTerm[\text{cv}]{\RegWeight, \Id_{\delta} }$ is strictly monotonic increasing while $\DataTerm[\text{cv}]{\RegWeight, \overline{\Id_{\delta}} }$ has a unique minimum.}
	\label{fig:04_cvError}
\end{figure}

Notice that the minimization problem in \eref{eqn:04_cvObjectiveFunction} itself depends on an optimization problem. This makes the application of gradient-based optimization \cite{Pickup2007} difficult as the gradient $\nabla \DataTerm[\text{cv}]{\RegWeight, \overline{\Id_{\delta}} }$ is not well-defined and its numerical approximation would be computationally expensive. To this end, the proposed parameter selection utilizes an adaptive grid search to solve \eref{eqn:04_cvObjectiveFunction}. This search is performed in the domain of the log-transformed regularization weight $\log \RegWeight$ instead of using the linear space for $\RegWeight$. For the first iteration of the proposed algorithm, $\IterationIdx{\RegWeight}{1}$ is selected as the global minimum of $\DataTerm[\text{cv}]{\RegWeight, \overline{\Id_{\delta}} }$. For this task, a grid search is performed over the initial search range $[\log \RegWeight_l, \log \RegWeight_u]$\label{notation:cvSearchRange}. In the subsequent iterations ($t > 1$), $\IterationIdx{\RegWeight}{t-1}$ is used as initial guess to refine it to $\IterationIdx{\RegWeight}{t}$ with search range:
\begin{equation}
	\left[
		\log \IterationIdx{\RegWeight}{t-1} - \frac{1}{t},~ 
		\log \IterationIdx{\RegWeight}{t-1} + \frac{1}{t} 
	\right].
\end{equation}
The number of iterations is adaptively adjusted at each iteration of iteratively re-weighted minimization. For $t = 1$, it is initialized by \smash{$\IterationIdx{\NumIter[cv]}{1} = \NumIter[cv]$}\label{notation:cvNumIter}. Then, it is gradually reduced to \smash{$\IterationIdx{\NumIter[cv]}{t} = \lceil  0.5 \cdot \IterationIdx{\NumIter[cv]}{t-1} \rceil$}. This enables a parameter selection of moderate computational effort while avoiding the limitation of a gradient-based search.

\paragraph{Image Reconstruction.}

Once the confidence weights $\IterationIdx{\WeightsA}{t}$ and $\IterationIdx{\WeightsB}{t}$ as well as the regularization weight $\IterationIdx{\RegWeight}{t}$ are determined, the super-resolved image $\IterationIdx{\HR}{t}$ is estimated via the weighted energy minimization problem: 
\begin{equation}
	\label{eqn:04_srObjectiveFunctionImage}
	\IterationIdx{\HR}{t} = \argmin_{\HR} \IterationIdx{\EnergyFunSym}{t}(\HR).
\end{equation}
The energy function that is minimized at iteration $t$ is given by\label{notation:energyFunIter}:
\begin{equation}
	\IterationIdx{\EnergyFunSym}{t}(\HR) =
	\left(\LR - \SystemMat \HR \right)^\top \IterationIdx{\WeightsBMat}{t} \left( \LR - \SystemMat \HR \right)
	+ \IterationIdx{\RegWeight}{t} \left| \left| \IterationIdx{\WeightsAMat}{t} \SparseTransMat \HR \right| \right|_1,
\end{equation}
where $\IterationIdx{\WeightsAMat}{t} = \Diag{\IterationIdx{\WeightsA}{t}}$ and $\IterationIdx{\WeightsBMat}{t} = \Diag{\IterationIdx{\WeightsB}{t}}$. This convex and unconstrained minimization problem provides an \gls{map} estimate for $\IterationIdx{\HR}{t}$ under the given parameters and is numerically solved by means of gradient-based techniques. In this thesis, we employ \gls{scg} iterations \cite{Nabney2002} to solve for $\IterationIdx{\HR}{t}$ and to enhance the rate of convergence compared to steepest descent schemes. \Gls{scg} iterations seek a stationary point:
\begin{equation}
	\label{eqn:04_srObjectiveFunctionImageGradient}
	\nabla_{\HR} \IterationIdx{\EnergyFunSym}{t}(\HR) 
	= -2 \IterationIdx{\WeightsBMat}{t} \SystemMat^\top \left(\LR - \SystemMat \HR \right)
	+ \IterationIdx{\RegWeight}{t} 
	\IterationIdx{\WeightsAMat}{t} \SparseTransMat^\top \Sign{\IterationIdx{\WeightsAMat}{t} \SparseTransMat\HR}
	\mbeq \Zeros.
\end{equation}

Numerical optimization requires a smooth and continuous differentiable regularization term to facilitate gradient-based iterations. Therefore, the regularization term is approximated by the convex Charbonnier function\label{notation:charbonnierLoss} \cite{Charbonnier1994}:
\begin{equation}
	\LossFun[\text{Char}]{\vec{z}} \defeq \sum_{i = 1}^{\SparseTransDim} \sqrt{z_i^2 + \tau}.
\end{equation}
For small $\tau$ ($\tau = 10^{-4}$), this provides a reasonable approximation of the \LOne norm while avoiding the non-differentiability. Consequently, the gradient of the regularization term is given by:
\begin{equation}
	\IterationIdx{\WeightsAMat}{t} \SparseTransMat^\top \Sign{\IterationIdx{\WeightsAMat}{t} \SparseTransMat\HR}
	\approx \IterationIdx{\WeightsAMat}{t} \SparseTransMat^\top \cdot \psi_{\text{Char}}(\IterationIdx{\WeightsAMat}{t} \SparseTransMat\HR),
\end{equation}
where $\psi_{\text{Char}}(\vec{z}) = \nabla_{\vec{z}} \LossFun[\text{Char}]{\vec{z}}$ is the gradient of the Charbonnier function:
\begin{align}
	\psi_{\text{Char}}(\vec{z}) 
	&= \begin{pmatrix} 
		\psi_{\text{Char}}(z_1)
		& \psi_{\text{Char}}(z_2)
		& \ldots
		& \psi_{\text{Char}}(z_{\SparseTransDim}) 
	\end{pmatrix}^\top \\
	\psi_{\text{Char}}(z_i) &= z_i \left( \sqrt{z_i^2 + \tau} \right)^{-1}.
\end{align}

\paragraph{Coarse-to-Fine Optimization.}
Although re-weighted minimization according to \eref{eqn:04_srObjectiveFunctionImage} is convex, it is important to note that the overall optimization problem solved by iteratively re-weighted minimization is non-convex. Intuitively, this is caused by the fact that the convergence is affected by the initialization of the confidence weights and thus it may converge to different local minimums. For this reason, iteratively re-weighted minimization is implemented in a \textit{coarse-to-fine} scheme as shown in \aref{alg:04_srAlgorithm}. In this approach, the initial confidence weights are set to $\IterationIdx{\WeightsA}{0} = \Ones$ and $\IterationIdx{\WeightsB}{0} = \Ones$, where $\vec{1}$ is an all-one vector. The super-resolved image $\IterationIdx{\HR}{0}$ is initialized by the temporal median of the motion compensated low-resolution frames that severs as an outlier-insensitive initial guess \cite{Farsiu2003}. The magnification factor is initialized by a lower value than the desired magnification starting with $\IterationIdx{\MagFac}{1} = 1$. Then, it is gradually increased by $\Delta \MagFac $ per iteration such that $\IterationIdx{\MagFac}{t} = \IterationIdx{\MagFac}{t-1} + \Delta \MagFac$ until the desired magnification is reached. In order to solve \eref{eqn:04_srObjectiveFunctionImage}, $\IterationIdx{\HR}{t-1}$ is propagated as initial guess to determine $\IterationIdx{\HR}{t}$.

We perform a maximum number of $\NumIter[irwsr]$\label{notation:irwsrNumIter} iterations in the outer optimization loop, a maximum number of $\NumIter[scg]$\label{notation:scgNumIter} iterations for \gls{scg} in the inner loop, and an initial number of $\NumIter[cv]$ iterations for cross validation based parameter selection. As a termination criterion we use the absolute difference between $\IterationIdx{\HR}{t}$ and $\IterationIdx{\HR}{t-1}$ according to:
\begin{equation}
	\label{eqn:04_convergenceCriteria}
	\max_{i = 1, \ldots, \HRSize} \left( \big| \IterationIdx{\HR_i}{t - 1} - \IterationIdx{\HR_i}{t} \big| \right) < \TermTolerance,
\end{equation}
where $\TermTolerance$\label{notation:irwsrTerminationTol} denotes the termination tolerance.

This approach has two benefits compared to single-scale optimization. First, it reduces the risk of getting stuck in local minimums as the non-convexity of the energy function in a lower dimensional space is less crucial. Second, it reduces the computational costs as more iterations of the computational demanding hyperparameter estimation are done more efficiently for smaller magnification factors. 
\begin{algorithm}[!t]
	\caption{Super-resolution using iteratively re-weighted minimization}
	\small
	\label{alg:04_srAlgorithm}
	\begin{algorithmic}[1]
		\Require Initial guess for $\IterationIdx{\HR}{0}$ (high-resolution image), $\IterationIdx{\WeightsA}{0}$ and $\IterationIdx{\WeightsB}{0}$ (confidence weights), $\IterationIdx{\MagFac}{0}$ (magnification factor), and $[\log \RegWeight_l, \log \RegWeight_u]$ (regularization weight search range)
		\Ensure Final high-resolution image $\HR$, confidence weights $\WeightsA$ and $\WeightsB$, and regularization weight $\RegWeight$
		\While{Convergence criterion in \eref{eqn:04_convergenceCriteria} not fulfilled and $t \leq \NumIter[irwsr]$}
			\State Select magnification factor $\IterationIdx{\MagFac}{t} = \min(\IterationIdx{\MagFac}{t-1} + \Delta \MagFac, \MagFac)$
			\State Propagate $\IterationIdx{\HR}{t-1}$ in coarse-to-fine scheme using magnification factor $\IterationIdx{\MagFac}{t}$
			
			\State Compute scale parameters $\IterationIdx{\NoiseStd}{t}$ and $\IterationIdx{\PriorStd}{t}$ according to \eref{eqn:04_observationNoiseStdRobust} -- \eqref{eqn:04_btvNoiseStd}
			
			\State Compute confidence weights $\IterationIdx{\WeightsA}{t}$ and $\IterationIdx{\WeightsB}{t}$ according to \eref{eqn:04_biasDataFidelityWeights} -- \eqref{eqn:04_regularizationWeights}
			
			\State Compute regularization weight $\IterationIdx{\RegWeight}{t}$ according to \eref{eqn:04_cvObjectiveFunction} with $\IterationIdx{\NumIter[cv]}{t}$ iterations
			
			\State $t_{\text{scg}} \gets 1$
			\While{Convergence criterion in \eref{eqn:04_convergenceCriteria} not fulfilled and $t_{\text{scg}} \leq \NumIter[scg]$}
				\State Update $\IterationIdx{\HR}{t}$ by \gls{scg} iteration for \eref{eqn:04_srObjectiveFunctionImageGradient}
				\State $t_{\text{scg}} \gets t_{\text{scg}} + 1$
			\EndWhile

			\State $t \gets t + 1$
		\EndWhile
	\end{algorithmic}
\end{algorithm}

\subsection{Algorithm Analysis}
\label{sec:04_AlgorithmAnalysis}

In this section, we analyze \aref{alg:04_srAlgorithm} regarding the following aspects. First and foremost, we discuss the relationship of iteratively re-weighted minimization to \gls{mm} algorithms. This links the proposed weighted optimization to the solution of a non-convex energy minimization problem. Afterwards, based on this relationship to the \gls{mm} theory, we prove the convergence of the underlying iteration scheme.

\paragraph{Relationship to Majorization-Minimization Algorithms.}

The proposed super-resolution method can be considered as \gls{mm} algorithm, see \sref{sec:04_BayesianModelForRobustSuperResolution}. The basic notion of this class of algorithms is to replace the direct minimization of a difficult -- potentially non-convex function -- with the minimization of a surrogate function. Compared to the original non-convex function, this surrogate function is easier to optimize. A surrogate function that can be employed in this context is referred to as \textit{majorizing function} \cite{Hunter2004} and is defined as follows.  
\begin{restatable}[Majorizing function]{definition}{04_majorizingFunction}
	Let $\EnergyFun{\HR}$ be a real-valued function. Then, the real-valued function $\EnergyFunMajorize{\HR, \IterationIdx{\HR}{t-1}}$\label{notation:majorizingFun} is called a majorizing function for $\EnergyFun{\HR}$ at $\IterationIdx{\HR}{t-1} \in \RealN{\HRSize}$ if:
	\begin{enumerate}
		\item $\EnergyFunMajorize{\HR, \IterationIdx{\HR}{t-1}} \geq \EnergyFun{\HR}$ for all $\HR \in \RealN{\HRSize}$, and
		\item $\EnergyFunMajorize{\IterationIdx{\HR}{t-1}, \IterationIdx{\HR}{t-1}} = \EnergyFun{\IterationIdx{\HR}{t-1}}$.
	\end{enumerate}
\end{restatable}

Let us now consider the robust and sparse reconstruction given by the minimum of the non-convex energy function:
\begin{equation}
	\label{eqn:04_nonConvexEnergy}
	\EnergyFun{\HR} = \sum_{i=1}^{\NumFrames \LRSize} \HuberLoss{ \VecEl{\LR - \SystemMat \HR}{i} } 
	+ \RegWeight \sum_{i = 1}^{\SparseTransDim} \LossFun[\SparsityParam]{ \VecEl{\SparseTransMat \HR}{i} },
\end{equation}
where the data fidelity is given by the Huber loss with scale parameter $\NoiseStd$:
\begin{equation}
	\label{eqn:04_nonConvexEnergy_huber}
	\HuberLoss{z} =
	\begin{cases}
		z^2        											& \text{if}~~ |z| \leq \NoiseStd \\
		2 \NoiseStd |z| - \NoiseStd^2, 	& \text{otherwise}.
	\end{cases},
\end{equation}
and the regularization term is defined by the mixed \LOne/\Lp{\SparsityParam} norm with $\SparsityParam \in [0, 1]$ and scale parameter $\PriorStd$\label{notation:mixedL1Lp}:
\begin{equation}
	\label{eqn:04_nonConvexEnergy_l1lp}
	\LossFun[\SparsityParam]{z} = 
	\begin{cases}
		|z|											& \text{if}~|z| \leq \PriorStd \\
		\PriorStd^{1-\SparsityParam} |z|^\SparsityParam		& \text{otherwise},
	\end{cases}.
\end{equation}
This function comprises an outlier-insensitive data fidelity term and sparse regularization. Moreover, let us define the convex energy: 
\begin{equation}
	\label{eqn:04_majorizingFunction}
	\EnergyFunMajorize{\HR, \IterationIdx{\HR}{t-1}} =
	\IterationIdx{\EnergyFunSym}{t}(\HR, \IterationIdx{\HR}{t-1}) 
	+ \sum_{i = 1}^{\NumFrames\LRSize} \rho \left( \VecEl{\LR - \SystemMat \IterationIdx{\HR}{t-1}}{i} \right)
	+ \RegWeight \sum_{i = 1}^{\SparseTransDim} \tau \left( \VecEl{\SparseTransMat \IterationIdx{\HR}{t-1}}{i} \right),
\end{equation}
where:
\begin{align}	
	\label{eqn:04_majorizingFunctionRho}
	\rho(z) &= 
		\begin{cases}
			0 																									&	\text{if}~z \leq \NoiseStd \\
			\NoiseStd^2 \left( \frac{z}{\NoiseStd} - 1 \right)	& \text{otherwise}
		\end{cases}, \\
	\label{eqn:04_majorizingFunctionTau}
	\tau(z) &=
		\begin{cases}
			0																& \text{if}~|z| \leq \PriorStd \\
			(1 - \SparsityParam) \PriorStd^{1-\SparsityParam} |z|^\SparsityParam		& \text{otherwise}
		\end{cases},
\end{align}
and $\IterationIdx{\EnergyFunSym}{t}(\HR, \IterationIdx{\HR}{t-1})$ is the energy function in \eref{eqn:04_srObjectiveFunctionImage} as optimized by \aref{alg:04_srAlgorithm} with regularization weight $\RegWeight$. Notice that \smash{$\EnergyFunMajorize{\HR, \IterationIdx{\HR}{t-1}}$} and \smash{$\IterationIdx{\EnergyFunSym}{t}(\HR, \IterationIdx{\HR}{t-1})$} are equal up to the non-negative terms $\rho(\cdot)$ and $\tau(\cdot)$ that are independent of $\HR$. Thus, \smash{$\EnergyFunMajorize{\HR, \IterationIdx{\HR}{t-1}}$} is an upper bound for \smash{$\IterationIdx{\EnergyFunSym}{t}(\HR, \IterationIdx{\HR}{t-1})$} and the minimizer of these functions \wrt $\HR$ are equivalent. 

The relation of iteratively re-weighted minimization to \gls{mm} algorithms is established by the following theorem.
\begin{restatable}{thm}{mmTheorem}
	The convex energy function $\EnergyFunMajorize{\HR, \IterationIdx{\HR}{t-1}}$ in \eref{eqn:04_majorizingFunction} is a majorizing function for the non-convex energy function $\EnergyFun{\HR}$ in \eref{eqn:04_nonConvexEnergy} at $\HR = \IterationIdx{\HR}{t-1}$.
	\label{theo:04_mmTheorem}
\end{restatable}
\begin{proof}
	The proof of this theorem is given in \appref{sec:A_RelationshipToMajorizationMinimizationAlgorithms}.
\end{proof}

If the scale parameters $\NoiseStd$ and $\PriorStd$ as well as the regularization weight $\RegWeight$ are assumed to be constant, the proposed algorithm can be considered as an \gls{mm} algorithm to minimize the non-convex energy in \eref{eqn:04_nonConvexEnergy}. The basic principle of this scheme is to successively construct majorizing functions $\EnergyFunMajorize{\HR, \IterationIdx{\HR}{t-1}}$ at $\IterationIdx{\HR}{t-1}$ to obtain a refined estimate $\IterationIdx{\HR}{t}$, see \fref{fig:04_mmPrinciple}. Thus, direct optimization of a non-convex energy function is casted to a sequence of weighted but convex optimizations. This relationship also clarifies the properties of the proposed algorithm regarding robustness as minimization of the confidence-aware observation model is related to minimizing the Huber loss. Similarly, minimization based on the \gls{wbtv} prior is related to minimizing the sparsity-promoting \LOne/\Lp{\SparsityParam} regularization term.  
\begin{figure}[!t]
	\centering
		\includegraphics[width=0.96\textwidth]{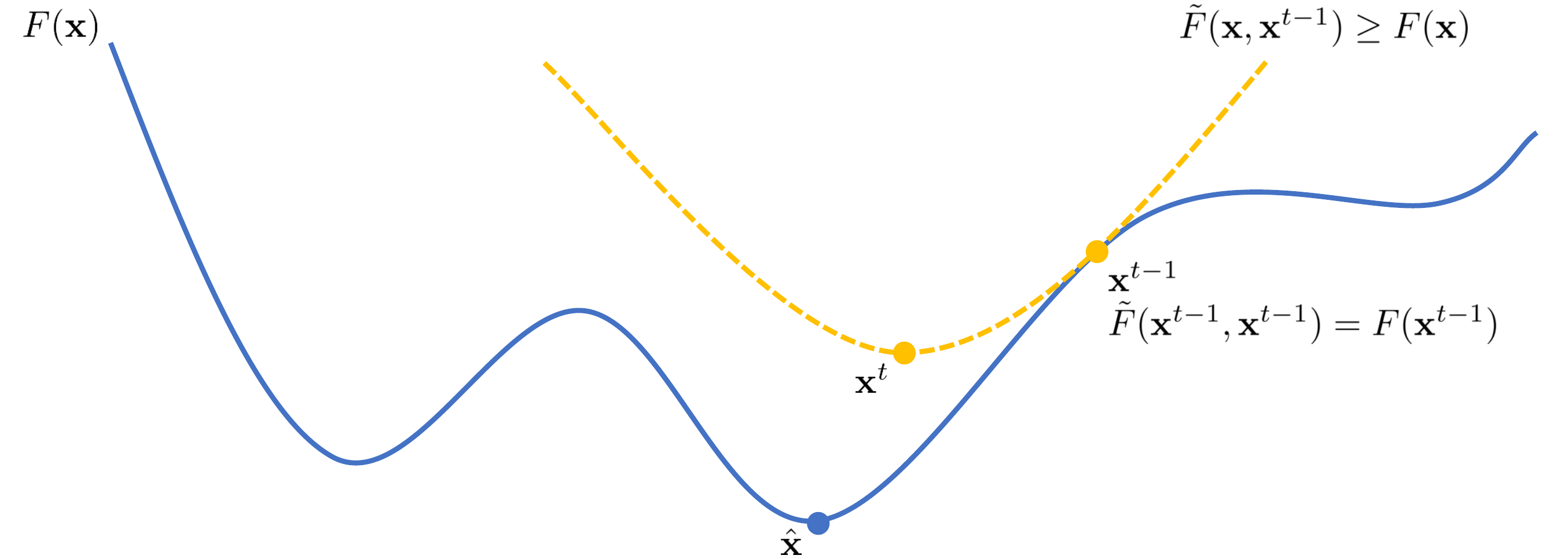}
	\caption[Illustration of the \gls{mm} principle]{Illustration of the \gls{mm} principle. The minimization of the non-convex function $\EnergyFun{\HR}$ is casted to the iterative minimization of convex majorizing functions $\EnergyFunMajorize{\HR, \IterationIdx{\HR}{t-1}}$.}
	\label{fig:04_mmPrinciple}
\end{figure}

\paragraph{Convergence Analysis.}

Based on this relationship, we establish a convergence proof of iteratively re-weighted minimization. In order to study the convergence, let \smash{$\IterationIdx{\HR}{0}$} be the initial guess and $\NoiseStd$, $\PriorStd$ as well as $\RegWeight$ constant parameters over the iterations. Then, the objective value $\EnergyFun{\HR}$ in \eref{eqn:04_nonConvexEnergy} converges within a finite number of iterations, which is stated by the following theorem.
\begin{restatable}{thm}{convergenceTheorem}
	Let $\IterationIdx{\HR}{1}, \ldots, \IterationIdx{\HR}{\NumIter}$ be an iteration sequence obtained by iteratively re-weighted minimization. Then, for all $t = 2, \ldots, \NumIter$ there exists a strict positive $\underline{\WeightsBFun}$ such that:
	\begin{equation}
		\EnergyFun{\IterationIdx{\HR}{t-1}} - \EnergyFun{\IterationIdx{\HR}{t}} 
		\geq \underline{\WeightsBFun} 
		\big| \big| \SystemMat \IterationIdx{\HR}{t-1} - \SystemMat \IterationIdx{\HR}{t} \big| \big|_2 ^2.
	\end{equation}
	\label{theorem:04_convergenceTheorem}
\end{restatable}
\begin{proof}
	The proof of this theorem is given in \appref{sec:A_ConvergenceAnalysis}.
\end{proof}
This theorem states that the objective value $\EnergyFun{\IterationIdx{\HR}{t}}$ is monotonically decreasing. Since $\EnergyFun{\HR}$ is a lower-bounded function, $\EnergyFun{\IterationIdx{\HR}{t}}$ converges to an extreme value, and so does Algorithm\,\ref{alg:04_srAlgorithm}. For a detailed experimental convergence study with adaptive scale and regularization parameters, we refer to \sref{sec:04_ConvergenceAndParameterSensitivity}. 

\section{Experiments and Results}
\label{sec:04_ExperimentsAndResults}

The experimental results reported in this section provide a comparison of the proposed method to several state-of-the-art algorithms as well as an in-depth analysis of its numerical properties. This includes quantitative evaluations on simulated data and qualitative assessment of super-resolution on real images. We focus on super-resolution under challenging conditions in real-world applications, including motion estimation uncertainty or image noise with space variant properties.

\subsection{Experiments on Simulated Data}
\label{sec:04_QuantitativeEvaluationOnSimulatedData}

To enable quantitative evaluations, simulated low-resolution data with a known ground truth for the desired high-resolution data was used. The ground truth images were projected by the image formation model to obtain their low-resolution counterparts. This mapping was described by rigid motion with uniform distributed translation $\TransVec = (\TransSym_\CoordU, \TransSym_\CoordV)^\top$, $\TransSym_\CoordU, \TransSym_\CoordV \in [-3, 3]$ and rotation angle $\varphi \in [-1, 1]$, a Gaussian \gls{psf} ($\PSFWidth = 0.5$), and subsampling according to the desired magnification factor $\MagFac$. Each frame was corrupted by a superimposition of intensity-dependent Poisson noise, Gaussian noise with standard deviation $\NoiseStd$, and salt-and-pepper noise at level $\InvPixelFraction$\label{notation:invPixelFraction} that specifies the amount of invalid pixels. For each ground truth, the simulation was performed ten times and all results were averaged over these randomized realizations. Grayscale converted reference images were taken from the LIVE database \cite{Sheikh2016} consisting of color photographs of natural scenes. The \gls{psnr} and \gls{ssim} \cite{Wang2004} were used to compare super-resolution to a ground truth.

Iteratively re-weighted minimization was used with $\NumIter[irwsr] = 10$, $\NumIter[scg] = 10$, $\NumIter[cv] = 20$, and termination tolerance $\TermTolerance = 0.001$. The \gls{wbtv} parameters were set to $\BTVSize = 2$ and $\BTVWeight = 0.7$ with sparsity parameter $\SparsityParam = 0.5$. The tuning constants of the underlying weighting functions were set to $c_{\text{bias}} = 0.02$ and $c_{\text{local}} = c_{\text{prior}} = 2.0$ for images given in the intensity range $[0, 1]$ according to \cite{Kohler2015c}. 

The proposed approach was compared to several related spatial domain reconstruction algorithms, namely \LTwo norm minimization coupled with Tikhonov regularization (\LTwo-TIK) \cite{Elad1997}, \LOne norm minimization coupled with \gls{btv} regularization (\LOne-BTV) \cite{Farsiu2004a}, Lorentzian M-estimator based super-resolution (LOR) \cite{Patanavijit2007}, and adaptive super-resolution with bilateral edge preserving regularization (BEP) \cite{Zeng2013}. For a fair evaluation of these algorithms, their regularization weights were selected for each dataset individually using a grid search on a training sequence and maximization of the \gls{psnr}. Notice that the proposed algorithm does not require off-line regularization parameter selections.
 
\begin{figure}[!t]
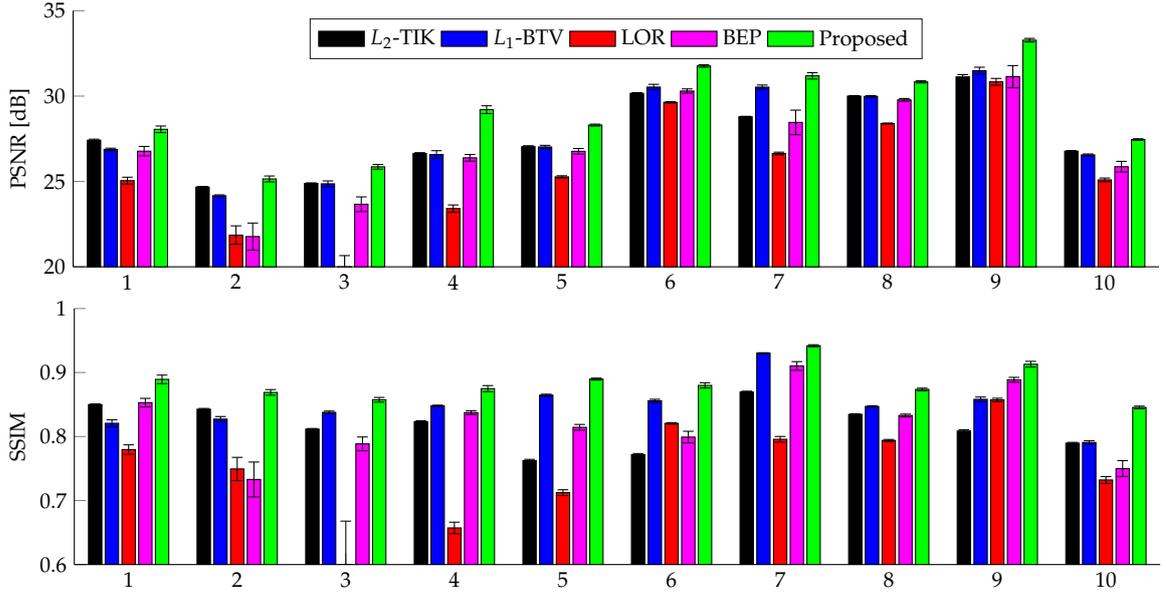

	\scriptsize 
	\centering
	\setlength \figurewidth{0.94\textwidth}
	\setlength \figureheight{0.89\figurewidth}
	\subfloat{\input{images/chapter4/barPlot_exactMotion_psnr.tikz}}\\[-0.25em]
	\subfloat{\input{images/chapter4/barPlot_exactMotion_ssim.tikz}}
	\caption[\Gls{psnr} and \gls{ssim} of super-resolution with exact motion estimation]{Mean $\pm$ standard deviation of the \gls{psnr} and \gls{ssim} achieved by the competing super-resolution algorithms under exact subpixel motion estimation. The benchmark includes ten simulated datasets with ten randomly generated image sequences per dataset.}
	\label{fig:04_barPlotsExactMotion}
\end{figure}

\paragraph{Effect of Image Noise.}
\Fref{fig:04_barPlotsExactMotion} depicts a benchmark of the competing super-resolution algorithms in a baseline experiment by utilizing the exact subpixel motion from the simulation process. In this experiment, ten datasets with sequences of $K = 8$ low-resolution frames were employed for super-resolution with magnification $\MagFac = 2$. Each frame was degraded by a fixed level of Gaussian noise ($\NoiseStd = 0.02$). Note that the proposed algorithm consistently outperformed the state-of-the-art in terms of both measures. In comparison to \LOne-BTV, the \gls{psnr} and \gls{ssim} measures were improved by 1.2\,\gls{db} and 0.03, respectively. A qualitative comparison is shown in \fref{fig:04_gaussianNoiseExample}, where the proposed algorithm achieved decent results in terms of the reconstruction of image structures while the competing methods were prone to oversmoothing or residual noise.
\begin{figure}[!t]
	\centering
	\subfloat[Original]{\includegraphics[width=0.322\textwidth]{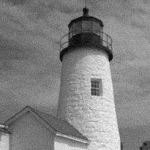}}~
	\subfloat[\LTwo-TIK \cite{Elad1997}]{\includegraphics[width=0.322\textwidth]{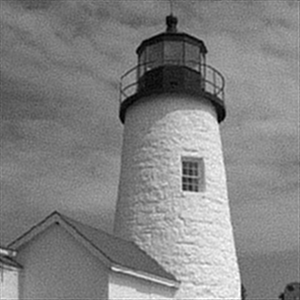}}~
	\subfloat[\LOne-BTV \cite{Farsiu2004a}]{\includegraphics[width=0.322\textwidth]{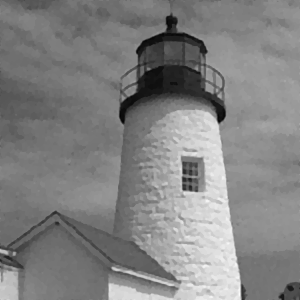}}\\
	\subfloat[LOR \cite{Patanavijit2007}]{\includegraphics[width=0.322\textwidth]{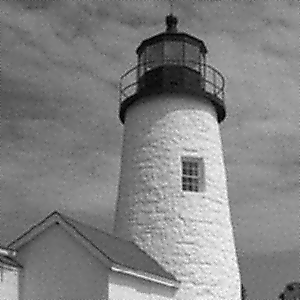}}~
	\subfloat[BEP \cite{Zeng2013}]{\includegraphics[width=0.322\textwidth]{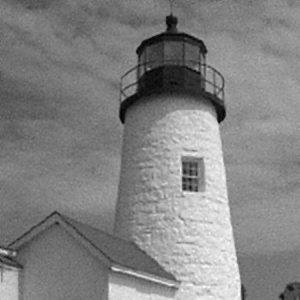}}~
	\subfloat[Proposed]{\includegraphics[width=0.322\textwidth]{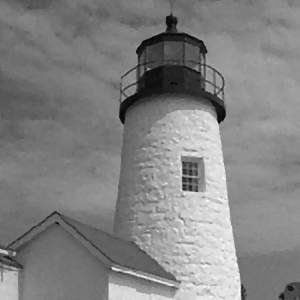}}
	\caption[\textit{Lighthouse} dataset with exact motion estimation]{Super-resolution ($\NumFrames = 8$ frames, magnification $\MagFac = 2$) with exact subpixel motion and additive Gaussian noise ($\NoiseStd = 0.02$) on the simulated \textit{lighthouse} dataset with a comparison of different combinations of observation models and prior distributions.}
	\label{fig:04_gaussianNoiseExample}
\end{figure}

The influence of image noise was investigated by varying the levels of Gaussian noise (\smash{$\NoiseStd \in [0, 0.04]$}) and salt-and-pepper noise (\smash{$\InvPixelFraction \in [0, 0.15]$}). The averaged \gls{psnr} and \gls{ssim} measures over ten realization of these experiments are plotted in \fref{fig:04_imageNoiseErrorMeasures}. In the presence of invalid pixels, the \LTwo-TIK method failed to reconstruct reliable high-resolution data as this approach does not compensate for outliers. The different robust models (\LOne-BTV \cite{Farsiu2004a}, LOR \cite{Patanavijit2007}, BEP \cite{Zeng2013}, and the proposed method) were less sensitive. Moreover, the proposed method quantitatively outperformed the competing robust models for both noise types. See \fref{fig:04_saltAndPepperNoiseExample} for a comparison on example data with salt-and-pepper noise.
\begin{figure}[!t]
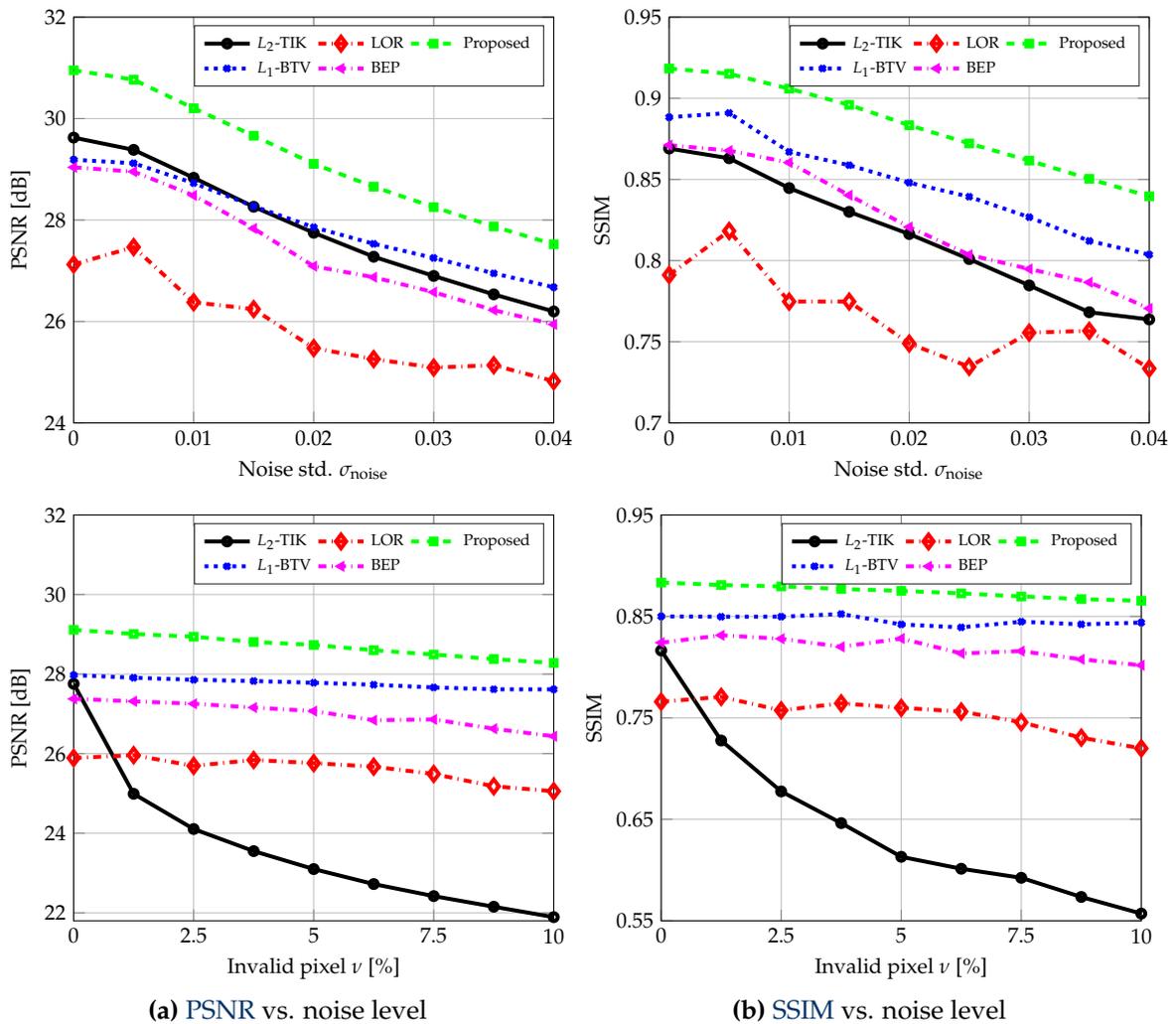

	\scriptsize
	\centering
	\setlength \figurewidth{0.44\textwidth}
	\setlength \figureheight{0.81\figurewidth} 
	\subfloat{\input{images/chapter4/gaussianNoise_psnr.tikz}}~
	\subfloat{\input{images/chapter4/gaussianNoise_ssim.tikz}}\\[-0.7em]
	\setcounter{subfigure}{0}
	\subfloat[\gls{psnr} vs. noise level]{\input{images/chapter4/saltAndPepperNoise_psnr.tikz}}~
	\subfloat[\gls{ssim} vs. noise level]{\input{images/chapter4/saltAndPepperNoise_ssim.tikz}}
	\hfill
	\caption[\gls{psnr} and \gls{ssim} of super-resolution with image noise]{\gls{psnr} and \gls{ssim} of super-resolution with image noise. Top row: performance of the competing algorithms under Gaussian noise of varying standard deviations $\NoiseStd$. Bottom row: influence of salt-and-pepper noise at different levels $\InvPixelFraction$.}
	\label{fig:04_imageNoiseErrorMeasures}
\end{figure}
\begin{figure}[!t]
	\centering
	\subfloat[Original]{\includegraphics[width=0.322\textwidth]{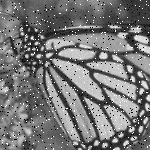}}~
	\subfloat[\LTwo-TIK \cite{Elad1997}]{\includegraphics[width=0.322\textwidth]{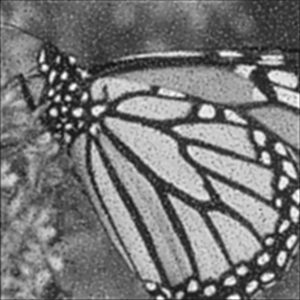}}~
	\subfloat[\LOne-BTV \cite{Farsiu2004a}]{\includegraphics[width=0.322\textwidth]{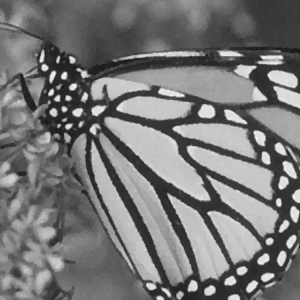}}\\
	\subfloat[LOR \cite{Patanavijit2007}]{\includegraphics[width=0.322\textwidth]{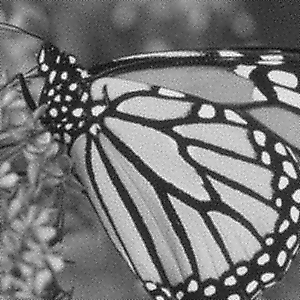}}~
	\subfloat[BEP \cite{Zeng2013}]{\includegraphics[width=0.322\textwidth]{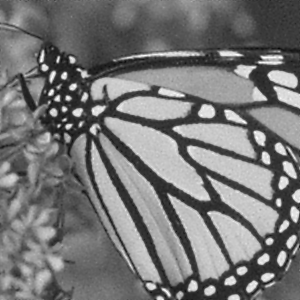}}~
	\subfloat[Proposed]{\includegraphics[width=0.322\textwidth]{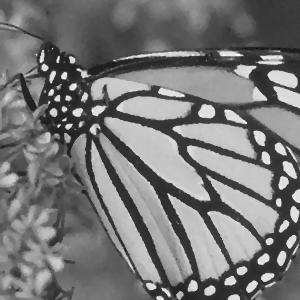}}
	\caption[\textit{Monarch} dataset with mixed Gaussian and salt-and-pepper noise]{Super-resolution ($\NumFrames = 8$ frames, magnification $\MagFac = 2$) on the simulated \textit{monarch} dataset with mixed Gaussian noise ($\NoiseStd = 0.02$) and salt-and-pepper noise with a comparison of different combinations of observation models and prior distributions.}
	\label{fig:04_saltAndPepperNoiseExample}
\end{figure}

\paragraph{Effect of Motion Estimation Uncertainty.}
The effect of motion estimation uncertainty was studied by simulating deviations between the true and the actual motion model. For this purpose, for two out of eight frames, an isotropic scaling factor to simulate camera zoom was considered such that the motion associated with these frames deviated from the rigid motion model. Motion estimation was performed using the \gls{ecc} optimization framework proposed by Evangelidis and Psarakis \cite{Evangelidis2008} assuming rigid motion. Hence, the frames affected by scaling can be considered as outliers. 

\Fref{fig:04_barPlotsInaccurateMotion} depicts a benchmark of super-resolution on ten simulated datasets in this situation, where the scaling factor followed a normal distribution $\NormalDistT{c}{1}{\sigma_c^2}$ with standard deviation $\sigma_c = 0.05$. Notice that the \LTwo-TIK method was prone to inaccurate motion estimation while the different robust methods were less sensitive. In this benchmark, the proposed method outperformed the state-of-the-art on most of the datasets. Compared to \LOne-BTV, the \gls{psnr} and \gls{ssim} measures were enhanced by 0.7\,\gls{db} and 0.04, respectively. See \fref{fig:04_motionEstimationUncertaintyExample} for a qualitative comparison among the competing algorithms. The effect of motion estimation uncertainty is visible by ghosting artifacts that were avoided by the proposed method.
\begin{figure}[!t]
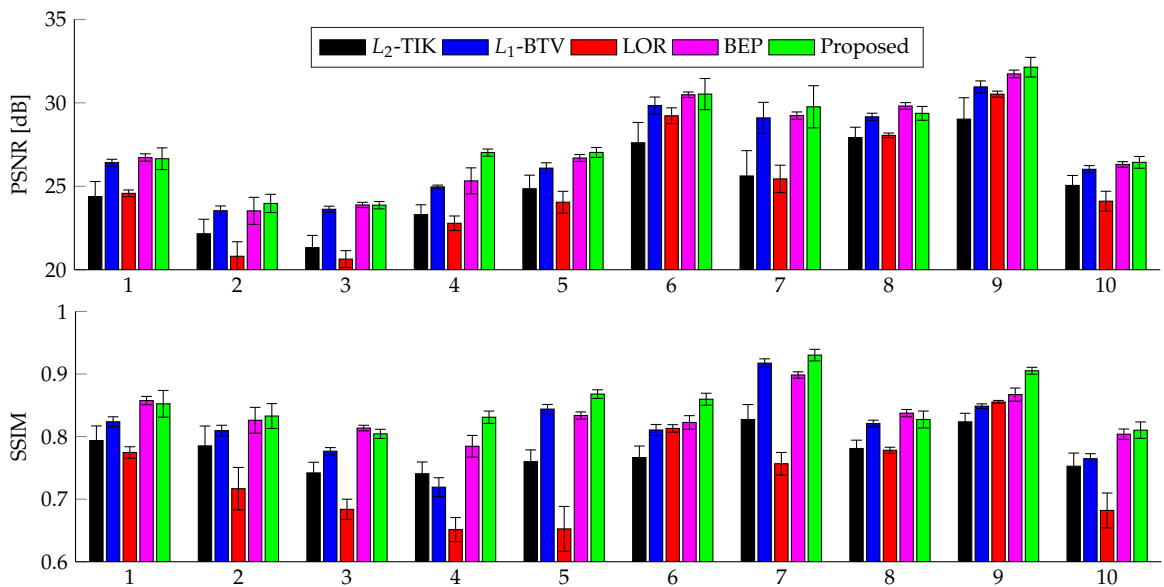

	\scriptsize 
	\centering
	\setlength \figurewidth{0.94\textwidth}
	\setlength \figureheight{0.87\figurewidth}
	\subfloat{\input{images/chapter4/barPlot_inaccurateMotion_psnr.tikz}}\\[-0.25em]
	\subfloat{\input{images/chapter4/barPlot_inaccurateMotion_ssim.tikz}}
	\caption[\gls{psnr} and \gls{ssim} of super-resolution with inaccurate motion estimation]{Mean $\pm$ standard deviation of the \gls{psnr} and \gls{ssim} measures in \fref{fig:04_barPlotsExactMotion} in the presence of inaccurate motion estimation.}
	\label{fig:04_barPlotsInaccurateMotion}
\end{figure}

\begin{figure}[!t]
	\centering
	\subfloat[Original]{\includegraphics[width=0.322\textwidth]{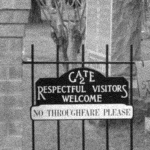}}~
	\subfloat[\LTwo-TIK \cite{Elad1997}]{\includegraphics[width=0.322\textwidth]{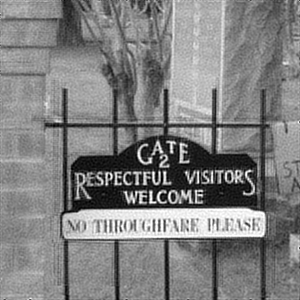}}~
	\subfloat[\LOne-BTV \cite{Farsiu2004a}]{\includegraphics[width=0.322\textwidth]{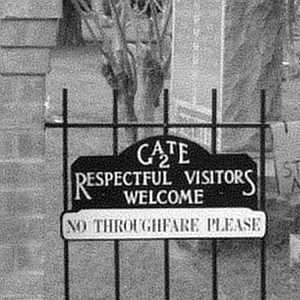}}\\
	\subfloat[LOR \cite{Patanavijit2007}]{\includegraphics[width=0.322\textwidth]{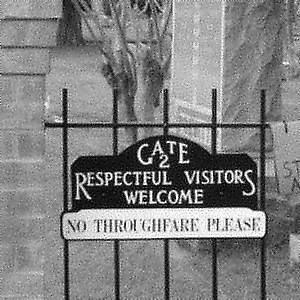}}~
	\subfloat[BEP \cite{Zeng2013}]{\includegraphics[width=0.322\textwidth]{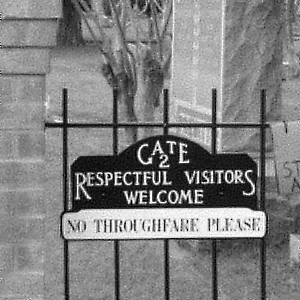}}~
	\subfloat[Proposed]{\includegraphics[width=0.322\textwidth]{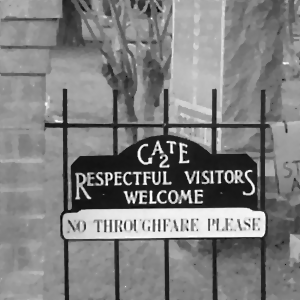}}
	\caption[\textit{Cemetery} dataset with inaccurate motion estimation]{Super-resolution ($\NumFrames = 8$ frames, magnification $\MagFac = 2$) on the simulated \textit{cemetery} dataset in the presence of inaccurate motion estimation. The uncertainty of motion parameters led to ghosting artifacts in case of a non-robust observation model (\LTwo-TIK \cite{Elad1997}), while robust models (\LOne-BTV \cite{Farsiu2004a}, LOR \cite{Patanavijit2007}, BEP \cite{Zeng2013}, and the proposed method) compensated for this uncertainty.}
	\label{fig:04_motionEstimationUncertaintyExample}
\end{figure}

\paragraph{Effect of Photometric Variations.}
The image formation models widely used in literature ignore several effects of digital imaging, see \sref{sec:03_DiscussionAndLimitationsOfTheModel}. This includes varying photometric conditions during image acquisition caused by camera white balancing or time variant lighting conditions. If varying photometric conditions should be taken into account, the image formation model needs to be extended and photometric registration has to be employed to estimate photometric parameters \cite{Capel2003, Capel2004}. However, photometric registration might be error prone and its uncertainty leads to outliers in super-resolution reconstruction. To evaluate the impact of photometric variations, an original low-resolution frame \smash{$\LRFrame{k}$} is corrupted according to \smash{$\FrameIdx{\vec{z}}{k} = \FrameIdx{\gamma_m}{k} \LRFrame{k} + \FrameIdx{\gamma_a}{k} \Ones$} to obtain a  distorted frame \smash{$\FrameIdx{\vec{z}}{k}$} \cite{Capel2004}. For two randomly selected frames, photometric variations were simulated by choosing uniform distributed parameters in \smash{$[-\frac{1}{2} \sigma_p, +\frac{1}{2} \sigma_p ]$} for \smash{$\FrameIdx{\gamma_a}{k}$} and in \smash{$[1 - \frac{1}{2} \sigma_p, 1 + \frac{1}{2} \sigma_p ]$} for \smash{$\FrameIdx{\gamma_m}{k}$}, where $\sigma_p$ reflects the parameter uncertainty.

\Fref{fig:04_photometricRegistrationErrorMeasures} depicts the impact of photometric variations at different levels $\sigma_p$. In this situation, the \LTwo-TIK approach was affected by an intensity bias as captured by the \gls{psnr}. The robust algorithms were less sensitive to photometric variations. In particular, the proposed method consistently achieved the highest quality measures since photometric variations were successfully compensated by bias detection. See \fref{fig:04_photometricRegistrationExample} for a visual comparison of this behavior on the \textit{lighthouse} dataset. Here, photometric variations in the input frames caused an intensity bias that is apparent in the \LTwo-TIK reconstruction but compensated by the proposed method.
\begin{figure}[!t]
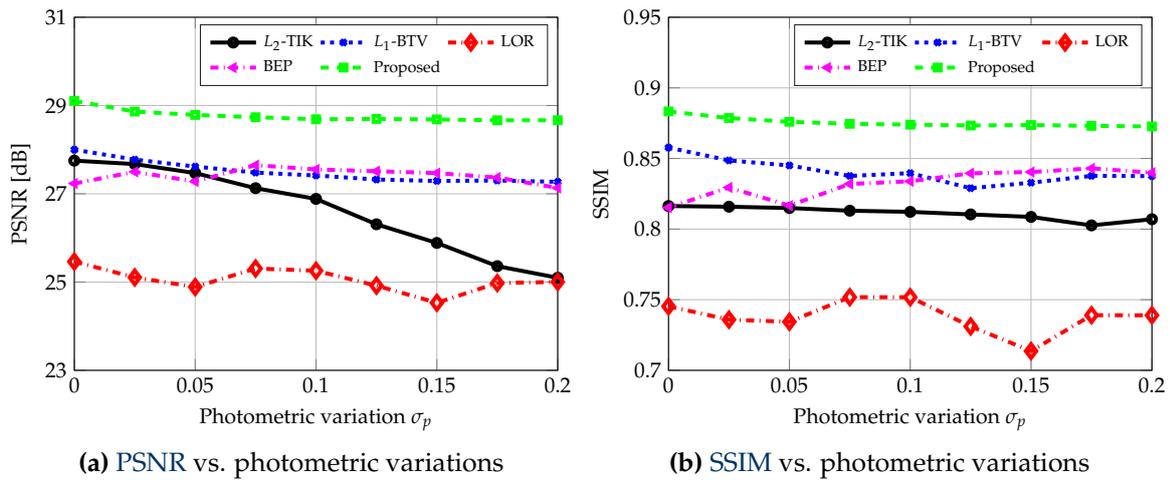

	\scriptsize 
	\centering
	\setlength \figurewidth{0.44\textwidth}
	\setlength \figureheight{0.70\figurewidth}  
	\subfloat[\gls{psnr} vs. photometric variations]{\input{images/chapter4/photometricChanges_psnr.tikz}}~
	\subfloat[\gls{ssim} vs. photometric variations]{\input{images/chapter4/photometricChanges_ssim.tikz}}
	\caption[\gls{psnr} and \gls{ssim} of super-resolution with photometric variations]{\gls{psnr} and \gls{ssim} of super-resolution at different levels of photometric variations. All photometric variations were simulated by uniform distributed global contrast and brightness changes with standard deviation $\sigma_p$ relative to a reference image.}
	\label{fig:04_photometricRegistrationErrorMeasures}
\end{figure}
\begin{figure}[!t]
	\centering
	\subfloat[Original]{\includegraphics[width=0.322\textwidth]{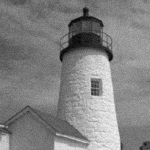}}~
	\subfloat[\LTwo-TIK \cite{Elad1997}]{\includegraphics[width=0.322\textwidth]{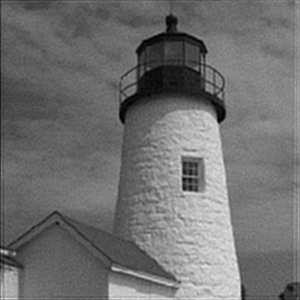}}~
	\subfloat[\LOne-BTV \cite{Farsiu2004a}]{\includegraphics[width=0.322\textwidth]{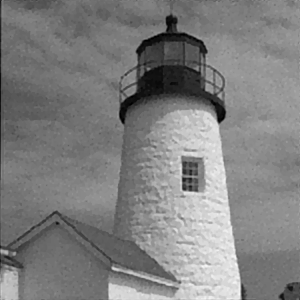}}\\
	\subfloat[LOR \cite{Patanavijit2007}]{\includegraphics[width=0.322\textwidth]{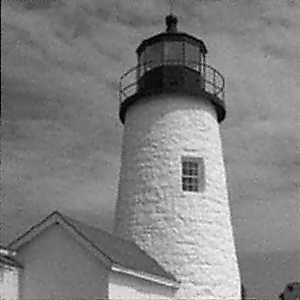}}~
	\subfloat[BEP \cite{Zeng2013}]{\includegraphics[width=0.322\textwidth]{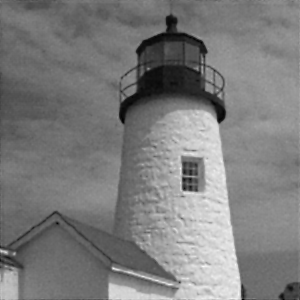}}~
	\subfloat[Proposed]{\includegraphics[width=0.322\textwidth]{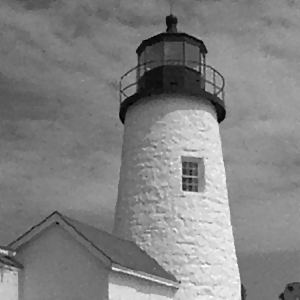}}
	\caption[\textit{Lighthouse} dataset with photometric variations]{Super-resolution ($\NumFrames = 8$ frames, magnification $\MagFac = 2$) on the \textit{lighthouse} dataset with photometric variations ($\sigma_p = 0.15$). Photometric variations in input frames led to an intensity bias under non-robust models (\LTwo-TIK \cite{Elad1997}), while robust models (\LOne-BTV \cite{Farsiu2004a}, LOR \cite{Patanavijit2007}, BEP \cite{Zeng2013}, and the proposed method) were less sensitive.}
	\label{fig:04_photometricRegistrationExample}
\end{figure}

\paragraph{Effect of the Sequence Length.}
One relevant parameter for super-resolution is the number of low-resolution input frames. This parameter was investigated for the magnification factor $\MagFac = 3$ as larger magnifications typically require more input frames. Throughout this experiment, the fraction of invalid pixels was set to $\InvPixelFraction = 0.01$ to simulate outliers and the exact subpixel motion was utilized. 

In \fref{fig:04_sequenceLengthErrorMeasures}, we depict the quality measures versus the number of low-resolution input frames. As expected, a larger number of frames resulted in more accurate reconstructions indicated by an increasing \gls{psnr} and \gls{ssim}. However, even in the case of long input sequences, the performance of \LTwo-TIK was limited due to the presence of outliers. In comparison to the competing algorithms, the proposed method performed best in terms of both quality measures. Notice that the proposed method with $\NumFrames = 8$ provided competitive results to \LOne-BTV and BEP with $\NumFrames = 20$ frames. Hence, it is more economical regarding the number of input frames. This study also considered the important use case of underdetermined super-resolution, which is the case for $\NumFrames < \MagFac^2$. Even in this challenging situation that appeared for $\NumFrames < 9$, the proposed algorithm provided reliable reconstructions \wrt the ground truth and outperformed the state-of-the-art.
\begin{figure}[!t]
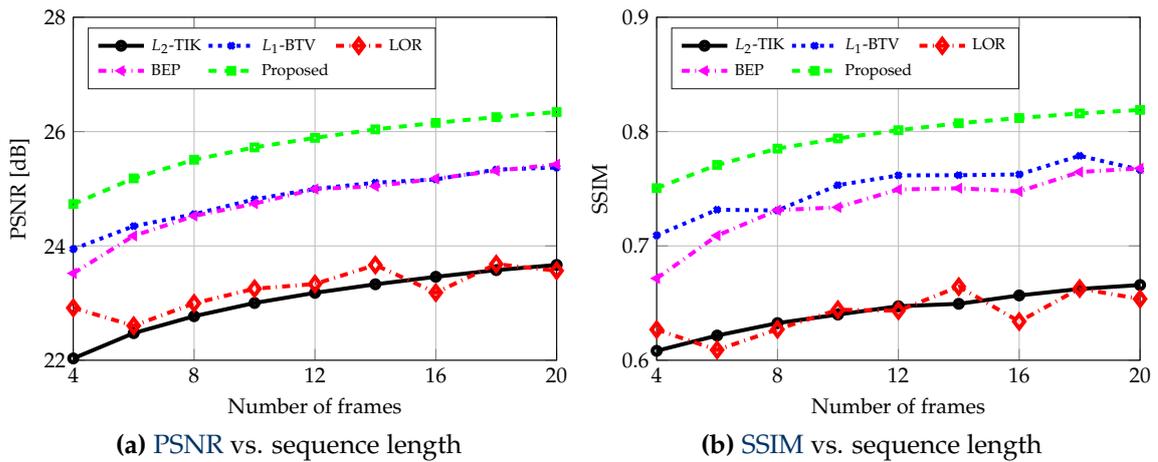

	\scriptsize 
	\centering
	\setlength \figurewidth{0.44\textwidth}
	\setlength \figureheight{0.68\figurewidth} 
	\subfloat[\gls{psnr} vs. sequence length]{\input{images/chapter4/sequenceLength_psnr.tikz}}~
	\subfloat[\gls{ssim} vs. sequence length]{\input{images/chapter4/sequenceLength_ssim.tikz}}
	\caption[\gls{psnr} and \gls{ssim} of super-resolution for different numbers of input frames]{\gls{psnr} and \gls{ssim} of super-resolution for different numbers of input frames.}
	\label{fig:04_sequenceLengthErrorMeasures}
\end{figure}

\subsection{Experiments on Real Data}
\label{sec:04_ExperimentsOnRealData}

The proposed method was qualitatively evaluated on real image data in two different applications. First, experiments with natural images were conducted. These are challenging due to the uncertainty of subpixel motion estimation. Second, experimental results in the field of 3-D range imaging are presented. Here, super-resolved range data was reconstructed from low-resolution range images captured with a \gls{tof} sensor that is affected by space variant noise. 

\paragraph{Evaluation on Natural Images.}

For the experiments on natural scenes, image sequences with different types of subpixel motion were used. \Fref{fig:04_mdspResults} compares the different super-resolution algorithms on the \textit{car} sequence taken from the MDSP database \cite{Farsiu2014}. This experiment aims at super-resolving a license plate using $\NumFrames = 12$ frames. Super-resolution was applied with magnification $\MagFac = 3$ and a Gaussian \gls{psf} ($\PSFWidth = 0.4$). The subpixel motion across these frames followed an affine model with a substantial amount of scaling related to out-of-plane movements of the car. The motion estimation for this sequence was performed by \gls{ecc} optimization \cite{Evangelidis2008}. Note that large car movements made motion estimation difficult and resulted in outliers due to misregistrations for individual frames. Consequently, \LTwo-TIK was affected by ghosting artifacts, while the robust algorithms were less sensitive. In terms of the recovery of the license plate, the proposed method provided an artifact-free and sharp reconstruction.

\Fref{fig:04_globeResults} depicts super-resolution on the \textit{globe} sequence \cite{Kohler2017} acquired with a Basler acA2000-50gm \gls{cmos} camera. For this experiment, $K = 17$ low-resolution frames captured by 4$\times$4 hardware binning on the sensor array relative to the maximum pixel resolution were used. Super-resolution was performed with magnification $\MagFac = 4$ and a Gaussian \gls{psf} ($\PSFWidth = 0.4$). The subpixel motion was related to a superposition of rigid camera movements and an independent rotation of the globe. In order to handle this non-rigid model, the variational optical flow algorithm proposed by Liu \cite{Liu2009} was employed for motion estimation. Notice that optical flow computation was error-prone due to occlusions that were caused by large rotations of the globe. Such outliers resulted in artifacts on the globe surface in the \LTwo-TIK reconstruction. The proposed method was robust against these outliers and achieved a decent recovery of text on the globe surface.
\begin{figure}[!t]
	\centering
	\subfloat[Original]{\includegraphics[width=0.322\textwidth]{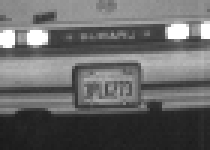}}~
	\subfloat[\LTwo-TIK \cite{Elad1997}]{\includegraphics[width=0.322\textwidth]{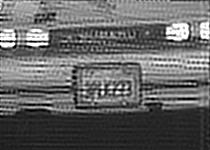}}~
	\subfloat[\LOne-BTV \cite{Farsiu2004a}]{\includegraphics[width=0.322\textwidth]{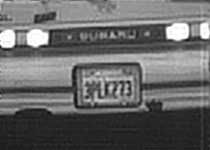}}\\
	\subfloat[LOR \cite{Patanavijit2007}]{\includegraphics[width=0.322\textwidth]{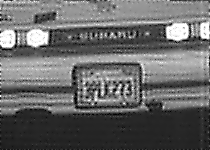}}~
	\subfloat[BEP \cite{Zeng2013}]{\includegraphics[width=0.322\textwidth]{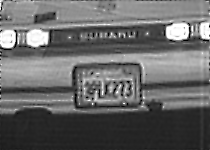}}~
	\subfloat[Proposed]{\includegraphics[width=0.322\textwidth]{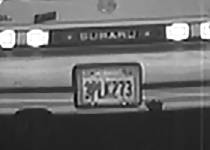}}
	\caption[Super-resolution on the \textit{car} dataset to identify a license plate]{Super-resolution on the \textit{car} dataset to identify a license plate ($\NumFrames = 12$ frames, magnification $\MagFac = 3$). The subpixel motion is related to out-of-plane movements of the car. Figure reused from \cite{Kohler2015c} with the publisher's permission \copyright2016 IEEE.}
	\label{fig:04_mdspResults}
\end{figure}
\begin{figure}[!t]
	\centering
	\hspace{-0.7em}
	\subfloat[Original]{
	\begin{tikzpicture}[spy using outlines={rectangle,red,magnification=3.75, height=3.95cm, width=2.6cm, connect spies, every spy on node/.append style={thick}}] 
			\node {\pgfimage[width=0.487\linewidth]{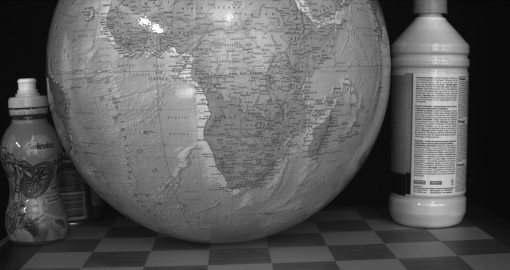}}; 
			\spy on (-0.3, 0.60) in node [left] at (3.7, 0);
		\end{tikzpicture}
	}
	\hspace{-1.2em}
	\subfloat[\LTwo-TIK \cite{Elad1997}]{
	\begin{tikzpicture}[spy using outlines={rectangle,red,magnification=3.75, height=3.95cm, width=2.6cm, connect spies, every spy on node/.append style={thick}}] 
			\node {\pgfimage[width=0.487\linewidth]{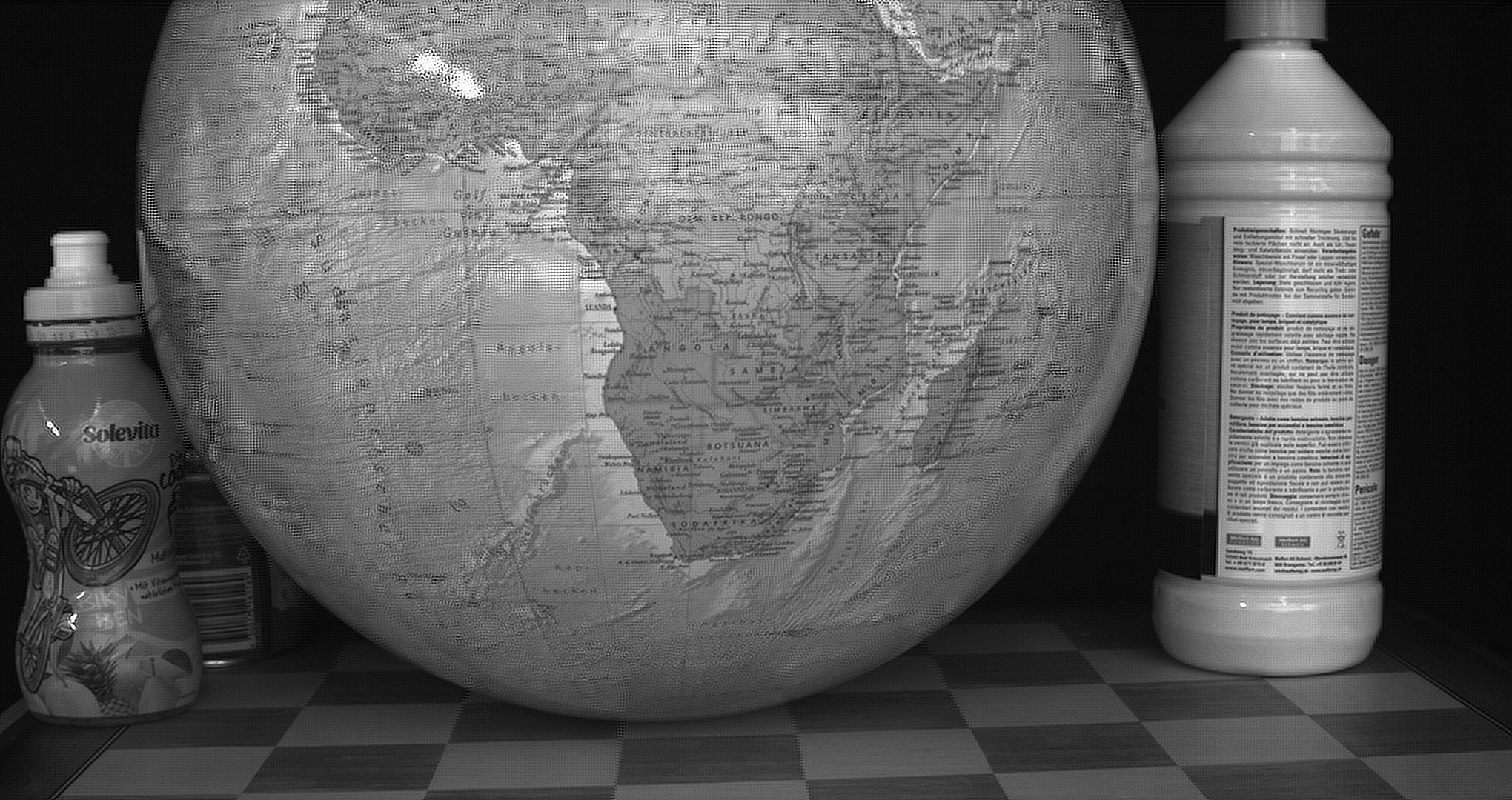}}; 
			\spy on (-0.3, 0.60) in node [left] at (3.7, 0);
		\end{tikzpicture}
	}\\[-0.6em]
	\hspace{-0.7em}
	\subfloat[\LOne-BTV \cite{Farsiu2004a}]{
	\begin{tikzpicture}[spy using outlines={rectangle,red,magnification=3.75, height=3.95cm, width=2.6cm, connect spies, every spy on node/.append style={thick}}] 
			\node {\pgfimage[width=0.487\linewidth]{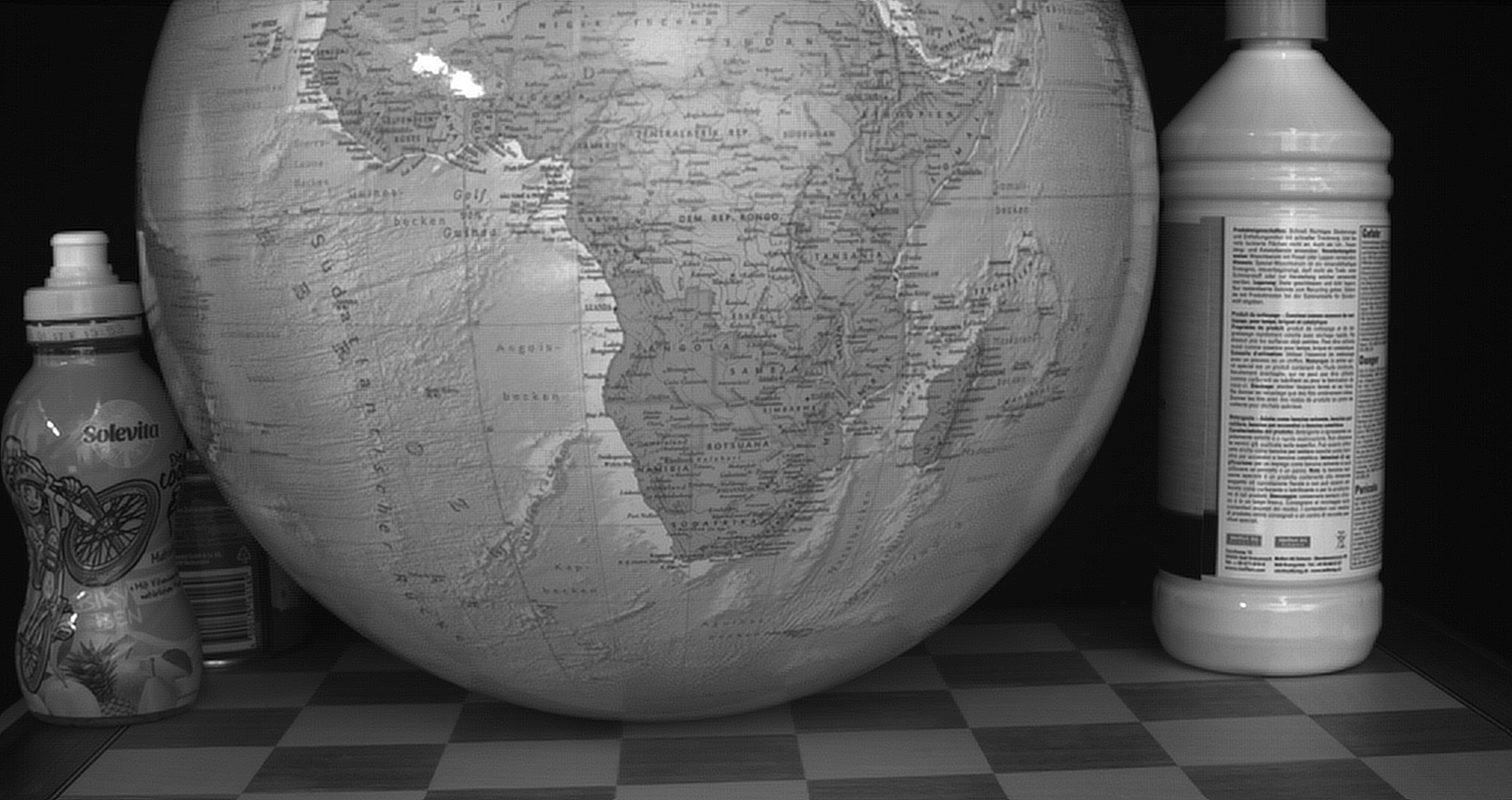}}; 
			\spy on (-0.3, 0.60) in node [left] at (3.7, 0);
		\end{tikzpicture}
	}
	\hspace{-1.2em}
	\subfloat[LOR \cite{Patanavijit2007}]{
	\begin{tikzpicture}[spy using outlines={rectangle,red,magnification=3.75, height=3.95cm, width=2.6cm, connect spies, every spy on node/.append style={thick}}] 
			\node {\pgfimage[width=0.487\linewidth]{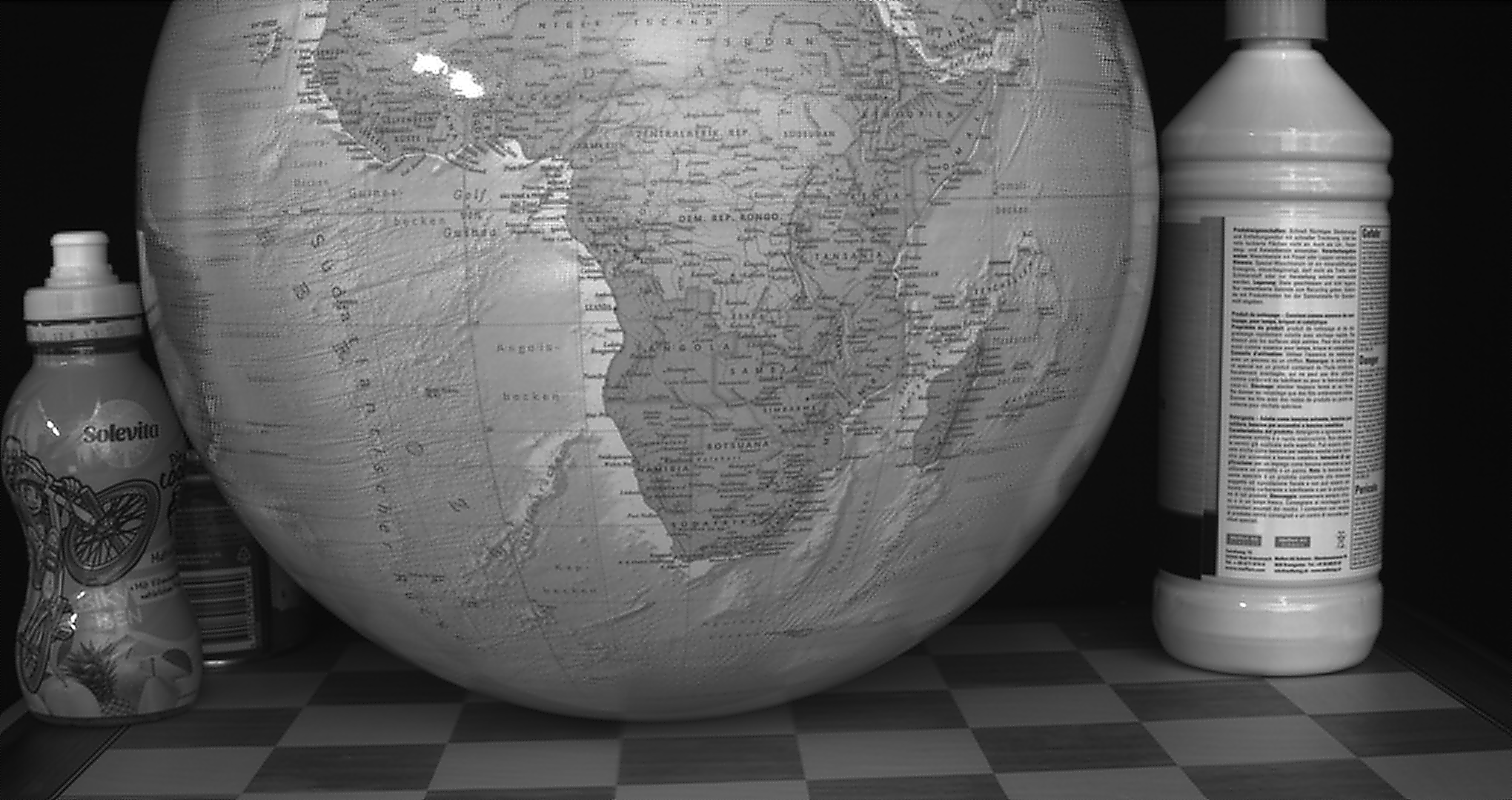}}; 
			\spy on (-0.3, 0.60) in node [left] at (3.7, 0);
		\end{tikzpicture}
	}\\[-0.6em]
	\hspace{-0.7em}
	\subfloat[BEP \cite{Zeng2013}]{
	\begin{tikzpicture}[spy using outlines={rectangle,red,magnification=3.75, height=3.95cm, width=2.6cm, connect spies, every spy on node/.append style={thick}}] 
			\node {\pgfimage[width=0.487\linewidth]{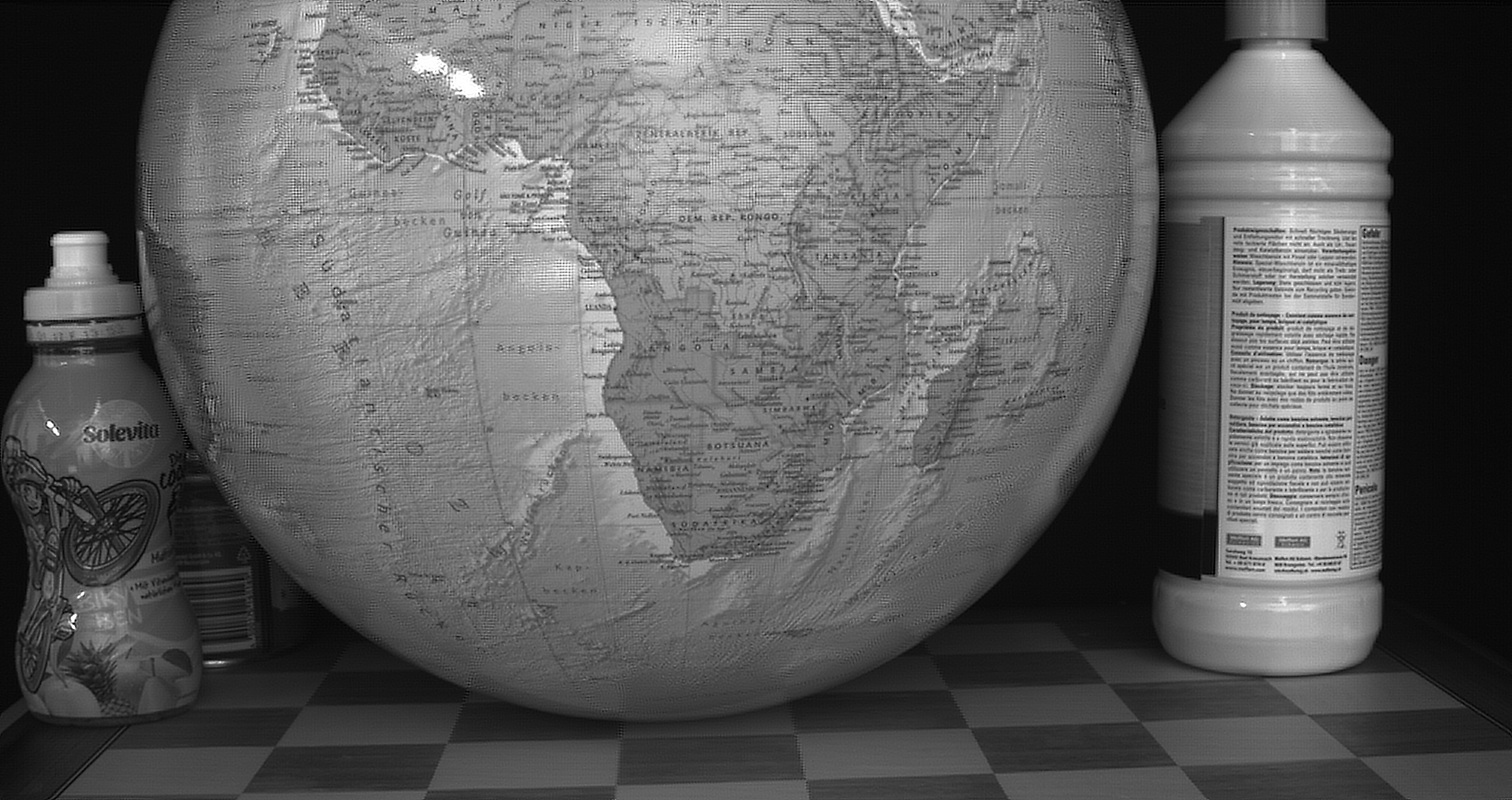}}; 
			\spy on (-0.3, 0.60) in node [left] at (3.7, 0);
		\end{tikzpicture}
	}
	\hspace{-1.2em}
	\subfloat[Proposed]{
	\begin{tikzpicture}[spy using outlines={rectangle,red,magnification=3.75, height=3.95cm, width=2.6cm, connect spies, every spy on node/.append style={thick}}] 
			\node {\pgfimage[width=0.487\linewidth]{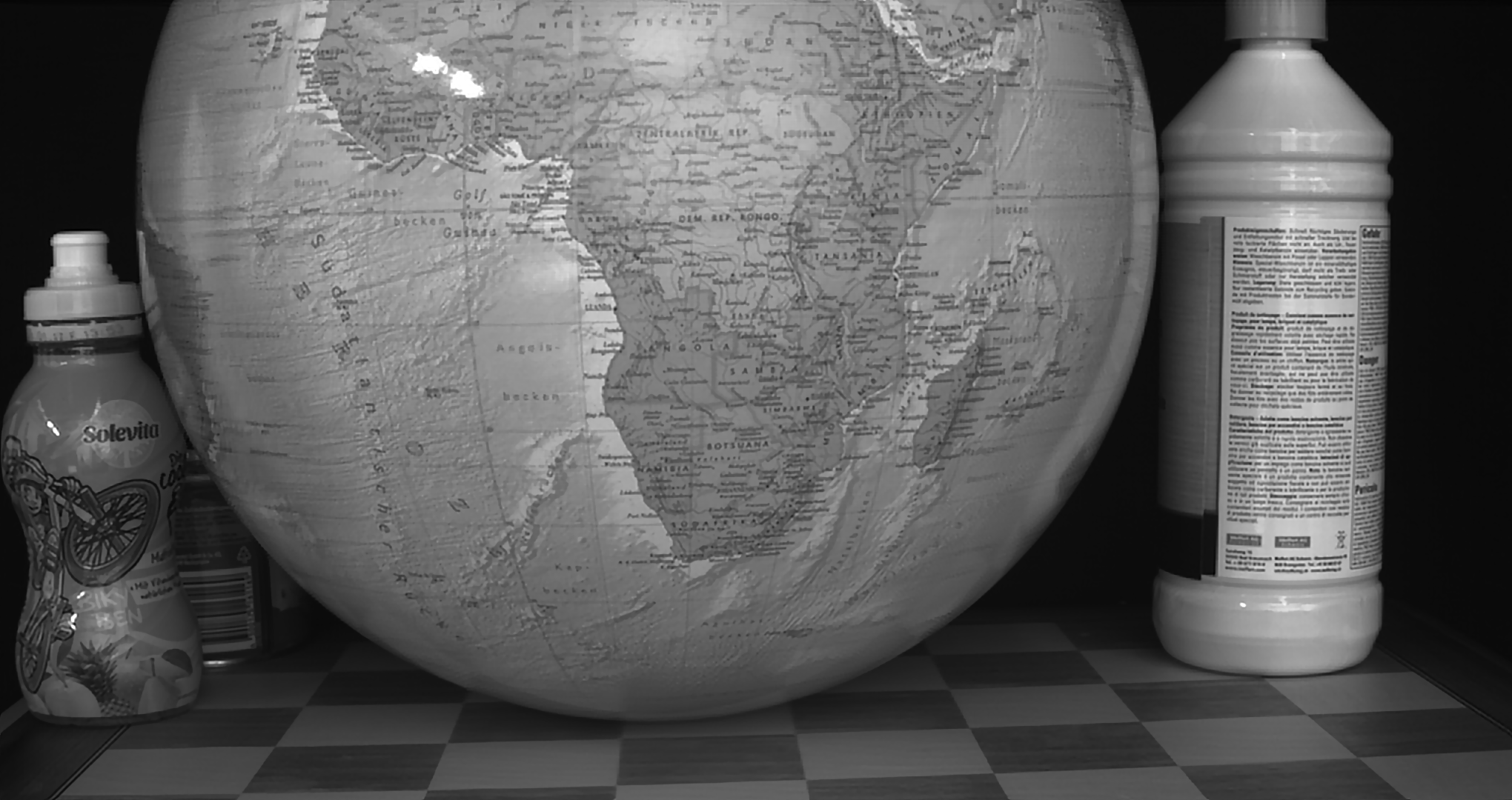}}; 
			\spy on (-0.3, 0.60) in node [left] at (3.7, 0);
		\end{tikzpicture}
	}
	\caption[Super-resolution on the \textit{globe} dataset]{Super-resolution on the \textit{globe} dataset ($\NumFrames = 17$ frames, magnification $\MagFac = 4$). The subpixel motion is a mixture of camera movements and a rotation of the globe.}
	\label{fig:04_globeResults}
\end{figure}

\paragraph{Evaluation on Range Images.}

For the experiments in range imaging, a PDM CamCube 3.0 \gls{tof} camera was used to measure the 3-D scene in \fref{fig:04_tofResult}. Range data was acquired with $200 \times 200$ px at a frame rate of 30\,Hz and super-resolution was applied to sets of range images. In addition to the low spatial resolution, the reliability of the \gls{tof} sensor was affected by intensity-dependent errors on the black surface of the punch. This resulted in space variant noise, \ie larger uncertainties of range data on the punch surface compared to regions with brighter illumination, which is a common issue in \gls{tof} imaging. Super-resolution was applied with $\NumFrames = 16$ frames, a Gaussian \gls{psf} ($\PSFWidth = 0.5$) and magnification $\MagFac = 3$. Motion estimation was performed by \gls{ecc} optimization with an affine model. 

In this example, the proposed method achieved the best behavior under space variant noise as shown by the reconstruction of flat surfaces and object edges. Range data with lower confidence, \ie measurements affected by higher noise levels, was successfully determined as a by-product of iteratively re-weighted minimization. This is visible in the visualization of the observation confidence map \smash{$\FrameIdx{\WeightsB}{1}$} associated with the first frame. Here, lower weights were assigned to surfaces affected by intensity-dependent noise. Similarly, the confidence map $\WeightsA$ of \gls{wbtv} steers the regularization to improve the reconstruction of depth discontinuities.
\begin{figure}[!t]
	\centering
		\subfloat[Original]{\includegraphics[width=0.231\textwidth]{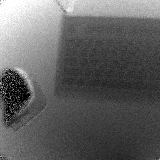}}~
		\subfloat[\LTwo-TIK \cite{Elad1997}]{\includegraphics[width=0.231\textwidth]{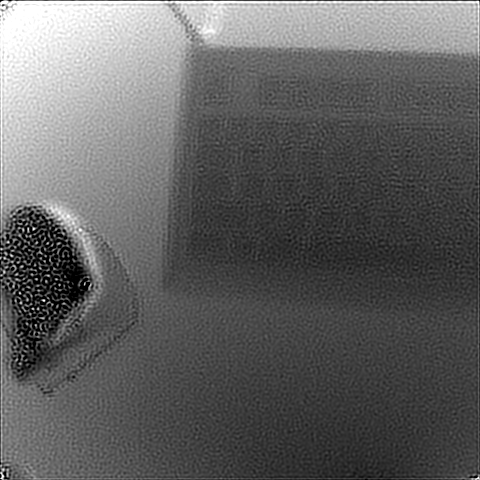}\label{fig:tofResult:l2}}~
		\subfloat[\LOne-BTV \cite{Farsiu2004a}]{\includegraphics[width=0.231\textwidth]{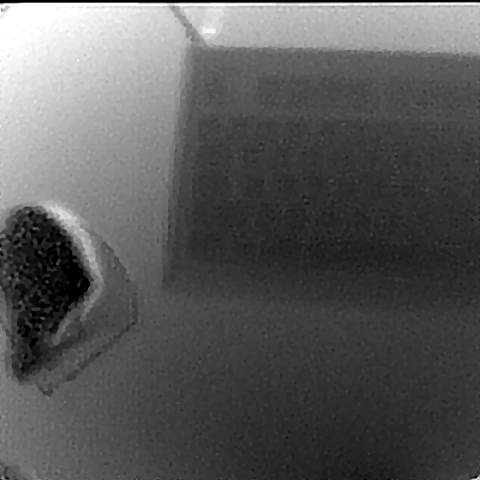}}\quad
		\setcounter{subfigure}{6}
		\subfloat[Weights $\FrameIdx{\WeightsB}{1}$]{\includegraphics[width=0.231\textwidth]{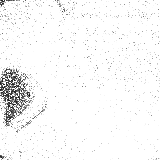}\label{fig:tofResult:observationWeights}}\\
		\setcounter{subfigure}{3}
		\subfloat[LOR \cite{Patanavijit2007}]{\includegraphics[width=0.231\textwidth]{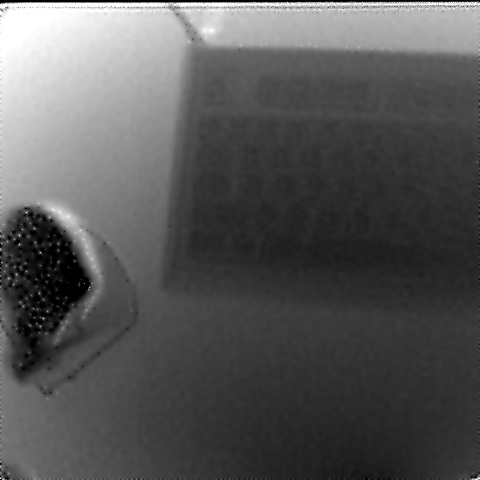}}~
		\subfloat[BEP \cite{Zeng2013}]{\includegraphics[width=0.231\textwidth]{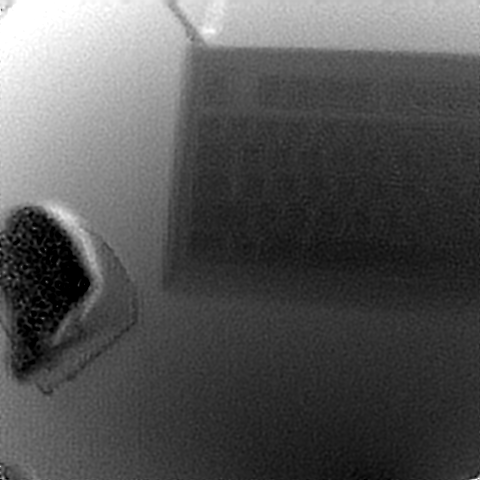}}~
		\subfloat[Proposed]{\includegraphics[width=0.231\textwidth]{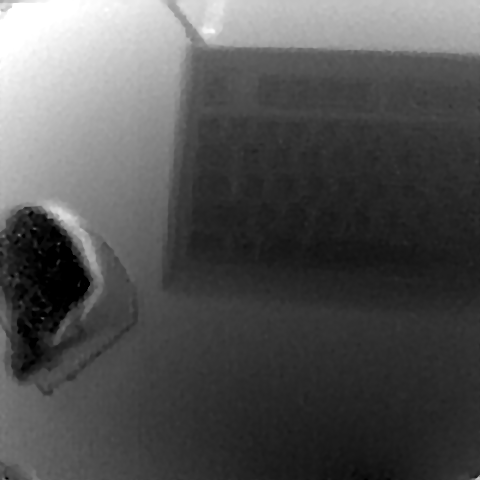}\label{fig:tofResult:irwsr}}\quad 
		\setcounter{subfigure}{7}
		\subfloat[Weights $\WeightsA$]{\includegraphics[width=0.231\textwidth]{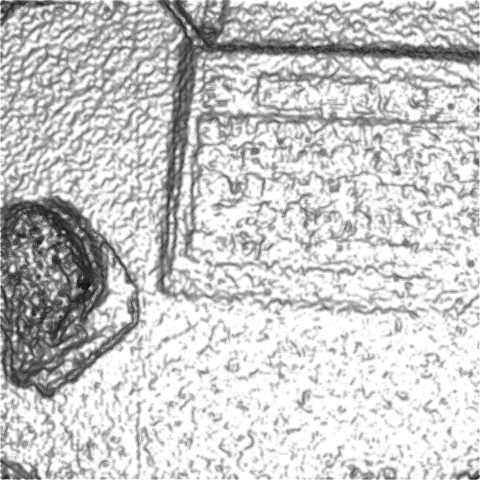}\label{fig:tofResult:priorWeights}}
	\caption[Super-resolution for \gls{tof} range images]{Super-resolution for \gls{tof} range images in the presence of space variant noise ($\NumFrames = 16$ frames, magnification $\MagFac = 3$). \protect\subref{fig:tofResult:l2} - \protect\subref{fig:tofResult:irwsr} Super-resolved images obtained by the competing algorithms. \protect\subref{fig:tofResult:observationWeights} - \protect\subref{fig:tofResult:priorWeights} Observation weights $\FrameIdx{\WeightsB}{1}$ associated with the first frame and prior weights $\WeightsA$ visualized in grayscale (brighter regions denote higher weights). Figure reused from \cite{Kohler2015c} with the publisher's permission \copyright2016 IEEE.}
	\label{fig:04_tofResult}
\end{figure}

\subsection{Convergence and Parameter Sensitivity}
\label{sec:04_ConvergenceAndParameterSensitivity}

To confirm the convergence of iteratively re-weighted minimization experimentally, the parameter estimates provided by the proposed algorithm were traced over the iterations. The convergence is studied on simulated images under a mixture of Gaussian noise ($\NoiseStd = 0.02$) and salt-and-pepper noise ($\InvPixelFraction \in [0, 0.1]$). To evaluate the sensitivity of the algorithm regarding the initial guess, the proposed initialization computed by the motion-compensated temporal median was compared to an initialization computed by bicubic upsampling of a single frame. Super-resolved images at different iterations with magnification factor $\MagFac = 3$ using $\NumFrames = 12$ frames and the temporal median as initial guess are shown in \fref{fig:04_convergenceExampleImages}. 

\Fref{fig:04_convergenceBehavior} depicts the average \gls{psnr} and \gls{ssim} measures of the super-resolved images at different iterations and different amounts of invalid pixels over ten random realizations of the experiment. Independently of the noise level and the initial guess, iteratively re-weighted minimization converged within the first five iterations. This also appeared in case of a large amount of outliers and confirms the convergence of the iteration scheme. In addition, the behavior of the adaptive scale and regularization parameter estimation is depicted. Similar to the latent high-resolution image, these estimates converged within a few iterations.

\begin{figure}[!t]	
	\subfloat[Original]{
	\begin{tikzpicture}[spy using outlines={rectangle,red,magnification=2.75, height=1.5cm, width=1.5cm, connect spies, every spy on node/.append style={thick}}] 
			\node {\pgfimage[width=0.236\linewidth]{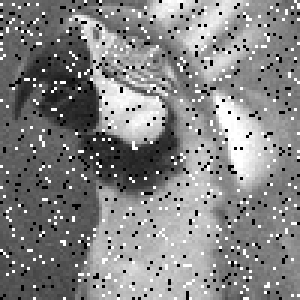}}; 
			\spy on (-0.1, 1.1) in node [left] at (-0.3, -1.032);
		\end{tikzpicture}
	}
	\hspace{-1.1em}
	\subfloat[Iteration 1]{
	\begin{tikzpicture}[spy using outlines={rectangle,red,magnification=2.75, height=1.5cm, width=1.5cm, connect spies, every spy on node/.append style={thick}}] 
			\node {\pgfimage[width=0.236\linewidth]{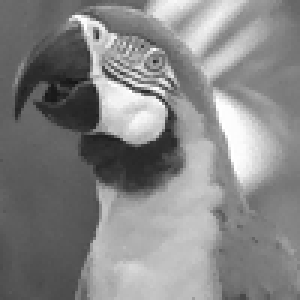}}; 
			\spy on (-0.1, 1.1) in node [left] at (-0.3, -1.032);
		\end{tikzpicture}
	}
	\hspace{-1.1em}
	\subfloat[Iteration 2]{
	\begin{tikzpicture}[spy using outlines={rectangle,red,magnification=2.75, height=1.5cm, width=1.5cm, connect spies, every spy on node/.append style={thick}}] 
			\node {\pgfimage[width=0.236\linewidth]{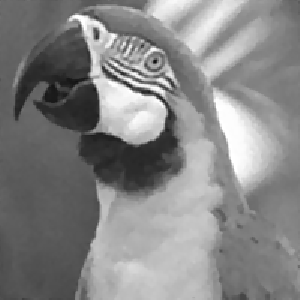}}; 
			\spy on (-0.1, 1.1) in node [left] at (-0.3, -1.032);
		\end{tikzpicture}
	}
	\hspace{-1.1em}
	\subfloat[Iteration 10]{
	\begin{tikzpicture}[spy using outlines={rectangle,red,magnification=2.75, height=1.5cm, width=1.5cm, connect spies, every spy on node/.append style={thick}}] 
			\node {\pgfimage[width=0.236\linewidth]{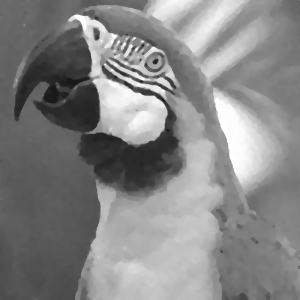}}; 
			\spy on (-0.1, 1.1) in node [left] at (-0.3, -1.032);
		\end{tikzpicture}
	}
	\caption[Convergence of iteratively re-weighted minimization on the \textit{parrots} dataset]{Illustration of the convergence of iteratively re-weighted minimization. The example depicts super-resolved images ($\NumFrames = 12$ frames, magnification $\MagFac = 3$) at different iterations for the \textit{parrots} dataset. The low-resolution input images are affected by a mixture of Gaussian noise ($\NoiseStd = 0.02$) and salt-and-pepper noise ($\InvPixelFraction = 0.1$).}
	\label{fig:04_convergenceExampleImages}
\end{figure}
\begin{figure}[!t]
	\centering
	\scriptsize 
	\setlength \figurewidth{0.41\textwidth}
	\setlength \figureheight{0.61\figurewidth} 
	\subfloat[\gls{psnr} over the iterations]{\input{images/chapter4/convergence_psnr.tikz}\label{fig:04_convergenceBehavior:psnr}}\quad
	\subfloat[\gls{ssim} over the iterations]{\input{images/chapter4/convergence_ssim.tikz}\label{fig:04_convergenceBehavior:ssim}}\\
	\subfloat[$\NoiseStd$ over the iterations]{\input{images/chapter4/convergence_noiseStd.tikz}\label{fig:04_convergenceBehavior:noiseStd}}\quad
	\subfloat[$\PriorStd$ over the iterations]{\input{images/chapter4/convergence_priorStd.tikz}\label{fig:04_convergenceBehavior:priorStd}}\\
	\subfloat[$\RegWeight$ over the iterations]{\input{images/chapter4/convergence_regularizationWeight.tikz}\label{fig:04_convergenceBehavior:regWeight}}
	\caption[Convergence analysis of iteratively re-weighted minimization]{Convergence analysis of iteratively re-weighted minimization. \protect\subref{fig:04_convergenceBehavior:psnr} - \protect\subref{fig:04_convergenceBehavior:ssim} \gls{psnr} and \gls{ssim} depicted for different amounts of invalid pixels on the dataset in \fref{fig:04_convergenceExampleImages}. The iterations were initialized by the temporal median (med) and bicubic upsampling of a single frame (bic). \protect\subref{fig:04_convergenceBehavior:noiseStd} - \protect\subref{fig:04_convergenceBehavior:regWeight} adaptive estimates of $\NoiseStd$, $\PriorStd$, and $\RegWeight$.}
	\label{fig:04_convergenceBehavior}
\end{figure}

One relevant parameter of iteratively re-weighted minimization is the sparsity parameter $\SparsityParam$ of the underlying prior weighting function. This parameter controls how strong sparsity is enforced and $\SparsityParam < 1$ implements a heavy-tailed prior distribution, see \sref{sec:04_AlgorithmAnalysis}. The impact of this parameter is studied at different noise levels. To this end, low-resolution images were corrupted by a mixture of Gaussian noise ($\NoiseStd \in [0, 0.04]$) and salt-and-pepper noise ($\InvPixelFraction = 0.01$). \Fref{fig:04_sparsifyingParameterExample} compares \gls{btv} ($\SparsityParam = 1.0$) to the proposed \gls{wbtv} ($\SparsityParam < 1$) on an example dataset. Notice that \gls{wbtv} regularization contributed to an improved reconstruction of fine textures. Furthermore, it was less sensitive to staircasing in homogenous image regions. The means and the standard deviations of the \gls{psnr} and \gls{ssim} measures for ten random realizations of this experiment are plotted in \fref{fig:04_sparsifyingParameterInfluence} for different noise levels. In these experiments, $\SparsityParam < 1$ enhanced the accuracy of super-resolution due to the edge-aware reconstruction compared to the unweighted \gls{btv} prior. The contributions of the sparse prior were more substantial for larger noise levels. In this work, $\SparsityParam$ is chosen in the range $[0.3, 0.8]$ as a too small $\SparsityParam$ decreases the numerical stability of weight computation and increases the degree of non-convexity of the underlying optimization problem. Conversely, a too large $\SparsityParam$ limits the benefit of the \gls{wbtv} prior. In summary, $\SparsityParam = 0.5$ is a reasonable choice for natural images and generalizes fairly well to scenes with different content.

\begin{figure}[!t]
	\subfloat[Original]{
	\begin{tikzpicture}[spy using outlines={rectangle,red,magnification=2.75, height=1.5cm, width=1.5cm, connect spies, every spy on node/.append style={thick}}] 
			\node {\pgfimage[width=0.236\linewidth]{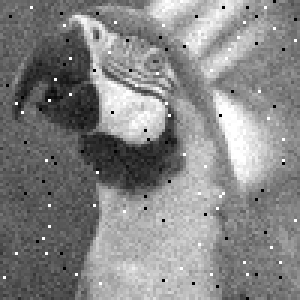}}; 
			\spy on (-0.1, 1.1) in node [left] at (-0.3, -1.032);
		\end{tikzpicture}
	}
	\hspace{-1.1em}
	\subfloat[\gls{btv} ($\SparsityParam = 1.0$)]{
	\begin{tikzpicture}[spy using outlines={rectangle,red,magnification=2.75, height=1.5cm, width=1.5cm, connect spies, every spy on node/.append style={thick}}] 
			\node {\pgfimage[width=0.236\linewidth]{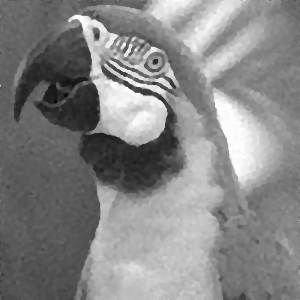}}; 
			\spy on (-0.1, 1.1) in node [left] at (-0.3, -1.032);
		\end{tikzpicture}
	}
	\hspace{-1.1em}
	\subfloat[\gls{wbtv} ($\SparsityParam = 0.5$)]{
	\begin{tikzpicture}[spy using outlines={rectangle,red,magnification=2.75, height=1.5cm, width=1.5cm, connect spies, every spy on node/.append style={thick}}] 
			\node {\pgfimage[width=0.236\linewidth]{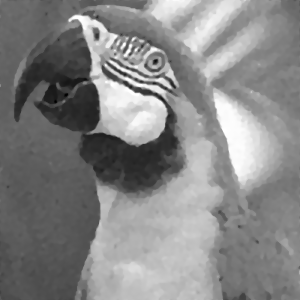}}; 
			\spy on (-0.1, 1.1) in node [left] at (-0.3, -1.032);
		\end{tikzpicture}
	}
	\hspace{-1.1em}
	\subfloat[\gls{wbtv} ($\SparsityParam = 0.3$)]{
	\begin{tikzpicture}[spy using outlines={rectangle,red,magnification=2.75, height=1.5cm, width=1.5cm, connect spies, every spy on node/.append style={thick}}] 
			\node {\pgfimage[width=0.236\linewidth]{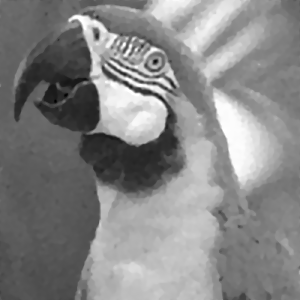}}; 
			\spy on (-0.1, 1.1) in node [left] at (-0.3, -1.032);
		\end{tikzpicture}
	}
	\caption[Impact of the sparsity parameter on the \textit{parrots} dataset]{Impact of the sparsity parameter $\SparsityParam$ of the proposed prior weighting function on the \textit{parrots} dataset. The low-resolution frames are affected by a mixture of Gaussian noise ($\NoiseStd = 0.04$) and salt-and-pepper noise ($\InvPixelFraction = 0.01$). This example compares the \gls{btv} prior ($\SparsityParam = 1.0$) to the proposed \gls{wbtv} prior using different settings of the sparsity parameter ($\SparsityParam = 0.3$ and $\SparsityParam = 0.5$). Notice the recovery of the texture for $\SparsityParam < 0.5$.}
	\label{fig:04_sparsifyingParameterExample}
\end{figure}
\begin{figure}[!t]
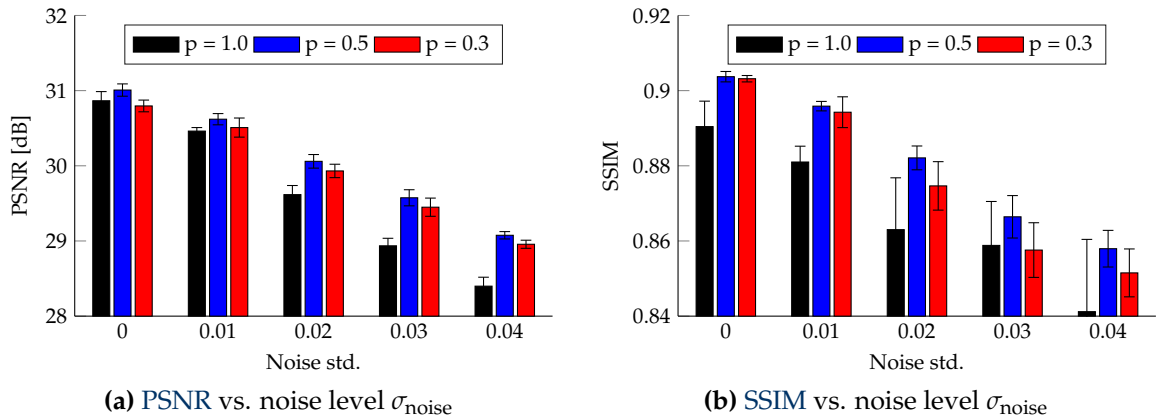

	\centering
	\scriptsize 
	\setlength \figurewidth{0.43\textwidth}
	\setlength \figureheight{0.61\figurewidth}
	\subfloat[\gls{psnr} vs. noise level $\NoiseStd$]{\input{images/chapter4/sparsityParameter_psnr.tikz}}\qquad
	\subfloat[\gls{ssim} vs. noise level $\NoiseStd$]{\input{images/chapter4/sparsityParameter_ssim.tikz}}
	\caption[Impact of the sparsity parameter to the performance of iteratively re-weighted minimization]{Impact of the sparsity parameter $\SparsityParam$ of the prior weighting function to the performance of iteratively re-weighted minimization. The parameter sensitivity was assessed on the dataset in \fref{fig:04_sparsifyingParameterExample} at different Gaussian noise levels. The proposed choice $\SparsityParam = 0.5$ led to superior results for all noise levels.}
	\label{fig:04_sparsifyingParameterInfluence}
\end{figure}

\subsection{Computational Complexity}
\label{sec:ComputationalComplexity}

This section reports the computational complexity of the proposed algorithm in terms of computation time as well as the number of energy function evaluations for numerical optimization. In \tref{tab:04_runTimes}, we compare these  performance characteristics on image sequences of different sizes\footnote{These experiments were performed on an Intel Xeon CPU E5-1630v4 with 3.7\,GHz and 64\,GB RAM using a non-parallelized MATLAB implementation.}. To this end, the \textit{car} and the \textit{globe} sequences (see \sref{sec:04_ExperimentsOnRealData}) were used. In both experiments, \LTwo-TIK converged quite fast and required the lowest computation time while the different robust algorithms required more iterations. 

To examine the complexity of the different computational stages, iteratively re-weighted minimization was evaluated with the proposed hyperparameter selection (adaptive $\RegWeight$) and with a bypass of this stage (constant $\RegWeight$). Here, the adaptive algorithm increased the computation time compared to the competing methods that do not provide an automatic parameter selection. Notice that a bypass of this stage considerably reduced the complexity. In this case, the computation time was comparable to those of the state-of-the-art. Moreover, the proposed coarse-to-fine optimization was compared to a single-scale implementation. Note that by-passing the coarse-to-fine optimization increased the computation time, which reveals the benefit of the proposed iteration scheme.
\begin{table}[!t]
	\centering
	\small
	\caption[Computational complexity of iteratively re-weighted minimization and several state-of-the-art algorithms]{Computational complexity of iteratively re-weighted minimization and several state-of-the-art algorithms. This analysis includes the computation times and the number of energy function evaluations for numerical optimization. The \LTwo-TIK method is considered as baseline and the numbers in brackets denote the relative increase of the computation time and the number of function evaluations compared to \LTwo-TIK. Iteratively re-weighted minimization was evaluated with (w/) and without (w/o) coarse-to-fine optimization as well as with adaptive regularization and with constant regularization weight.}
	\begin{tabular}{p{13.05em} cccc}
		\toprule
		\textbf{Super-resolution algorithm} & \multicolumn{2}{c}{\textit{Globe} sequence} & \multicolumn{2}{c}{\textit{Car} sequence} \\
		& \multicolumn{2}{c}{(510\,$\times$\,270, $\NumFrames = 17$ frames)} & \multicolumn{2}{c}{(70\,$\times$\,50, $\NumFrames = 12$ frames)} \\
		\scriptsize
		& Time [s] & \# Fun. eval. & Time [s] & \# Fun. eval. \\
		\midrule
		\textbf{State-of-the-art} 						& & & & \\
		~\LTwo-TIK \cite{Elad1997}		& 174 ($\times$1.0) & 50 ($\times$1.0) & 2 ($\times$1.0) & 48 ($\times$1.0) \\
		~\LOne-BTV \cite{Farsiu2004a}	& 383 ($\times$2.2) & 50 ($\times$\,1.0) & 4 ($\times$2.0) & 50 ($\times$\,1.0) \\
		~LOR \cite{Patanavijit2007}		& 323 ($\times$1.9) & 50 ($\times$1.0) & 4 ($\times$2.0) & 50 ($\times$\,1.0) \\
		~BEP \cite{Zeng2013}					& 2291 ($\times$13.2) & 50 ($\times$1.0) & 18 ($\times$9.0) & 50 ($\times$\,1.0) \\
		\midrule
		\textbf{Proposed} 						& & & & \\
		~w/ coarse-to-fine (adapt. $\RegWeight$) 	& 1198 ($\times$6.9) & 75 ($\times$\,1.5) & 15 ($\times$7.5) & 82 ($\times$1.7) \\
		~w/ coarse-to-fine (const. $\RegWeight$) 	& 914 ($\times$5.3) & 50 ($\times$\,1.0) & 8 ($\times$4) & 50 ($\times$1.0) \\
		~w/o coarse-to-fine (adapt. $\RegWeight$) & 3098 ($\times$17.8) & 80 ($\times$1.6) & 26 ($\times$13.0) & 90 ($\times$1.9) \\
		\bottomrule
	\end{tabular}
	\label{tab:04_runTimes}
\end{table}

\section{Conclusion}
\label{sec:04_Conclusion}

This chapter introduced a robust super-resolution algorithm from a Bayesian perspective. This approach is based on adaptive confidence weighting to define space variant observation and prior distributions. The confidence weights are treated as latent variables in the Bayesian model and are inferred simultaneously to the super-resolved image in an adaptive scheme by means of iteratively re-weighted minimization. Mathematically, this technique can be derived as an \gls{mm} algorithm. Iteratively re-weighted minimization combines the advantages of robustness regarding outliers in image formation with sparse regularization to enhance the reconstruction of edges and texture. As an additional merit, it does not require an extensive manual parameter tuning and provides an automatic parameter selection in a computationally efficient way.

In a baseline benchmark with mixed Gaussian and Poisson noise, iteratively re-weighted minimization achieved average gains of $1.2$\,\gls{db} and $0.03$ in terms of PSNR and SSIM over related robust algorithms. In a benchmark that considered inaccurate motion estimation, it improved the PSNR and SSIM by $0.7$\,\gls{db} and $0.04$, respectively. The iteration scheme showed fast convergence and converged to stationary points within five iterations regardless of the initialization and the fraction of outliers in the image formation.

Notice that throughout this chapter, subpixel motion is initially estimated on low-resolution frames prior to super-resolution. However, the proposed framework is extensible to treat subpixel motion as hidden information. In \cite{Bercea2016}, iteratively re-weighted minimization has been formulated via confidence-aware Levenberg-Marquardt optimization \cite{Marquardt1963}. This enables joint motion estimation and super-resolution to enhance the accuracy of an initial motion estimate.    

\part{Multi-Sensor Super-Resolution for Hybrid Imaging}
\label{sec:MultiSensorSuperResolutionForHybridImaging}

\chapter{Multi-Sensor Super-Resolution using Guidance Images}
\label{sec:MultiSensorSuperResolutionUsingGuidanceImages}

\myminitoc

\noindent
The algorithms investigated in \pref{sec:NumericalMethodsForMultiFrameSuperResolution} of this work provide super-resolution of a single modality only. \pref{sec:MultiSensorSuperResolutionForHybridImaging} of this thesis examines the extension of super-resolving images of one modality in the presence of a complementary modality. The key idea of this approach termed \textit{multi-sensor} super-resolution is to steer image reconstruction by guidance images. For this purpose, we present a computational framework that employs guidance data to enhance motion estimation and regularization compared to conventional algorithms dealing with a single modality only. Moreover, we present an outlier detection scheme for multi-sensor super-resolution as an extension of this framework. As an important application of practical relevance, the proposed method is evaluated in the field of \textit{hybrid range imaging}. In this application, high-resolution photometric data is used as guidance to super-resolve low-resolution 3-D range data. The presented experimental evaluation reveals that multi-sensor super-resolution outperforms conventional reconstruction algorithms that work solely on range data.

This methodology has been introduced by K\"ohler \etal \cite{Kohler2013a,Kohler2014a,Kohler2015a} and later presented by Haase \cite{Haase2016} for interventional imaging.

\section{Introduction}

Over the past decades, the vast majority of super-resolution algorithms has been designed to handle image data of a single modality. As these traditional approaches exploit information acquired with a single sensor, they are referred to as \textit{single-sensor} super-resolution. While this concept has the benefit of great flexibility regarding its applicability in different imaging systems, it suffers from the inherent drawback that only information present in low-resolution data is utilized. For instance, motion estimation as one of the most essential prerequisites of super-resolution needs to be carried out on low-resolution images, which might be error prone in practical applications. In addition, all algorithmic stages of these approaches are formulated for a single modality, \eg to design image priors in Bayesian methods. If super-resolution is applied in \textit{hybrid imaging} as the major goal of this chapter, this basic concept essentially ignores the presence of additional information captured by different modalities. In this context, hybrid imaging refers to a class of techniques that combines a set of complementary modalities in a common system by means of \textit{sensor data fusion}. 

\paragraph{Hybrid Imaging Technologies.}
Let us first review several popular hybrid imaging technologies that have emerged in literature along with their basic characteristics. All of these techniques have in common that the involved modalities are complementary in terms of their properties. Hence, their fusion enables a comprehensive representation of the underlying scene.

Some of the most popular hybrid imaging systems have been developed for healthcare. This includes the fusion of functional nuclear imaging such as \gls{pet} with structural imaging modalities such as \gls{ct} or \gls{mri} that are widely used in radiology. Combinations of these technologies have been engineered in \gls{pet}/\gls{ct} \cite{Beyer2000} or \gls{pet}/\gls{mri} \cite{Judenhofer2008} scanners. For instance in this context, structural imaging features the acquisition of anatomical information with high spatial resolution while nuclear imaging provides an acquisition of functional processes in lower resolution. Hybrid imaging has also been proposed for 3-D range imaging as the primary application in this chapter. In this field, measurements of 3-D surface information can be gained by means of active sensor technologies such as \gls{tof} \cite{Kolb2010} or structured light \cite{Scharstein2003}. In addition to the surface information encoded by range images that are acquired with these technologies, other optical techniques are used to capture photometric information of the same scene simultaneously \cite{Han2013}. The modalities are complementary, as photometric data provides high-resolution color and texture information while active range sensors acquire the corresponding surface information. 

One common observation in most of these systems is that some modalities feature a high spatial resolution while others are available in lower resolution. In practice, these gaps of the sensor characteristics are caused by technological or economic reasons. This initiates the development of novel super-resolution algorithms that exploit multiple modalities and their complementary natures in order to enhance the traditional single-sensor approaches.

\paragraph{Multi-Sensor Super-Resolution.}
The target of the proposed super-resolution approach for hybrid imaging is to identify one modality that is available at high spatial resolution as a \textit{guidance} modality. Accordingly, in contrast to the conventional single-sensor approaches, this chapter shows how resolution enhancement for one modality can be steered by such guidance images. Guidance images are exploited in various ways including 1) motion estimation, 2) spatially adaptive regularization, and 3) outlier detection as vital parts of multi-sensor super-resolution. This method is driven by the hypothesis that guidance images of high quality in terms of their spatial resolution and \gls{snr} hold the potential to enhance super-resolution of another modality.

The remainder of this chapter is organized as follows. \sref{sec:05_RelatedWork} provides a literature survey on related super-resolution and filtering techniques. In \sref{sec:05_MultiSensorSuperResolutionFramework}, we introduce a multi-sensor framework that employes guidance data for motion estimation and spatially adaptive regularization. Afterwards, \sref{sec:05_OutlierDetectionForRobustMultiSensorSuperResolution} extends this method by outlier detection that is driven by guidance data. \sref{sec:05_ApplicationToHybridRangeImaging} studies the application of this framework in hybrid 3-D range imaging, where low-resolution range images are super-resolved under the guidance of high-resolution photometric data. Finally, \sref{sec:05_Conclusion} presents a summary of this chapter.

\section{Related Work}
\label{sec:05_RelatedWork}

Compared to the great number of algorithms for single-sensor resolution enhancement,  there are only a few approaches that deal with the problem of multi-sensor super-resolution. Prior work in this field typically addresses specific imaging setups. An early method has been introduced in the pioneering work of Zomet and Peleg \cite{Zomet2002}. This method addresses super-resolution of multi-channel images and is driven by the strategy that super-resolution of one of the channels can be guided by the remaining channels. For this purpose, it exploits statistical redundancies across the channels to derive an observation model with a virtual prediction error. Instead of minimizing the residual error as done in single-sensor algorithms, this virtual prediction error is minimized. Super-resolution for color and infrared data have been considered as example applications but the experiments are limited to single-image upsampling. One limitation in comparison to the approach presented in this chapter is that it does not consider motion estimation as an integral part of super-resolution reconstruction. 

Methods that are conceptually closely related to the approach of Zomet and Peleg \cite{Zomet2002} include local image filters. Some of the well known techniques are guided upsampling \cite{He2013,He2010} or joint bilateral upsampling \cite{Kopf2007} to upsample an image under the guidance of a second one. Here, range imaging is one important application, where range data upsampling is guided by high-resolution color images. More recently, these filters have been extended by various approaches, \eg non-local means regularization \cite{Park2011}, anisotropic total generalized variation regularization \cite{Ferstl2013} or photometric and range co-sparse analysis \cite{Kiechle2013}. However, despite their success, these approaches were designed for single-image upsampling and use image formation models of limited flexibility compared to the models proposed for multi-frame super-resolution. In particular, effects of the camera \gls{psf} are seldom modeled by these local filters. For a generalization of such filters towards multi-frame super-resolution, we refer to \cref{sec:SuperResolutionForMultiChannelImages}.

Another mentionable approach that employs the concept of guidance images has been proposed by Kennedy \etal \cite{Kennedy2007} for \gls{pet}/\gls{ct} scanners in medicine. This method super-resolves \gls{pet} scans by using anatomical information gained from \gls{ct} data. Even if this approach is conceptually interesting, it is limited to \gls{pet} resolution enhancement and does not generalize to other imaging setups.  

\section{Multi-Sensor Super-Resolution Framework}
\label{sec:05_MultiSensorSuperResolutionFramework}

In this section, we present the basis of the proposed multi-sensor super-resolution framework that processes low-resolution images under the guidance of a complementary modality. The main novelty of this methodology is two-fold. First, a filter-based technique is presented that uses high-resolution guidance data to obtain a reliable motion estimate for super-resolution. In addition, a regularization technique is introduced that exploits high-resolution guidance images to adaptively regularize super-resolution on a second modality. Both techniques have the goal to improve robustness and accuracy of the framework compared to a single-sensor approach that does not exploit guidance data.

\subsection{Framework Overview}

The proposed framework aims at reconstructing a high-resolution image $\HR$ from a set of low-resolution frames \smash{$\LRSequence{1}{\NumFrames}$} termed \textit{input images}. For each input image \smash{$\LRFrame{k}$}, there exists a corresponding frame \smash{$\GuideFrame{k}$}\label{notation:guideFrame} that is acquired with another modality and termed \textit{guidance image}\footnote{Each guidance image $\GuideFrame{k}$ is synchronized in time with the respective input image $\LRFrame{k}$.}. Each guidance image \smash{$\GuideFrame{k}$} is encoded as a $\GuideDimU \times \GuideDimV$\label{notation:guideDim} image. In fact, the pixel resolution $\GuideSize = \GuideDimU \GuideDimV$\label{notation:guideSize} can be much higher than those of the associated input image \smash{$\LRFrame{k}$} given by $\LRSize = \LRDimU \LRDimV$ to take advantage of the guidance data.

We assume that each pair \smash{$(\LRFrame{k}, \GuideFrame{k})$} is aligned to each other by means of sensor data fusion. This alignment is described by the pixel-wise mapping\label{notation:senFusMapping}:
\begin{equation}
	\label{eqn:sensorDataFusionPointMapping}
	\Point_{\LRSym} = \SenFusMapping{\Point_{\GuideSym}},
\end{equation}
where $\Point_{\LRSym} \in \DomainLR$\label{notation:pointLr} denotes a pixel in an input image and $\Point_{\GuideSym} \in \DomainGuide$\label{notation:pointGuide}\label{notation:domainGuide} denotes the corresponding pixel position in the guidance image that encodes the same position in the captured scene but with a different modality. Note that the mapping $\SenFusMappingSym: \DomainGuide \rightarrow \DomainLR$\label{notation:domainInput} needs not be bijective. In particular, one important situation is the case of a surjective mapping. In this situation, a set of pixel coordinates in a guidance image maps to the same pixel in the corresponding input image, which appears if the pixel resolution of the guidance data is higher than those of the input images. In the most common setup, a set of pixels $\PointV_{\GuideSym} \in \SenFusNeighbor(\Point_{\GuideSym})$\label{notation:senFusNeighbor} is mapped to the same position $\Point_{\LRSym}$, see \fref{fig:sensorFusion}. Conversely, $\SenFusMappingSym^{-1}(\Point_{\LRSym})$ denotes a set of pixels in the guidance image that are associated with $\Point_{\LRSym}$ under the assumption that the mapping is invertible\footnote{Notice that depending on the implementation of the sensor data fusion, the mapping might not be invertible and occlusions needs to be considered. For instance, this is the case in stereo vision setups used for hybrid range imaging \cite{Kohler2015a}.}. 
If the underlying mapping is applied to fuse guidance and input images, both domains are aligned up a scale factor to preserve their pixel resolutions. For the sake of convenience, we limit ourselves to static systems, \ie the sensors involved in the system have a fixed relative orientation. Hence, the mapping is constant over all frames. In this case, sensor data fusion can be either achieved by a software-based calibration involving image registration or can be directly implemented by the imaging system. 
\begin{figure}[!t]
	\centering
		\includegraphics[width=0.98\textwidth]{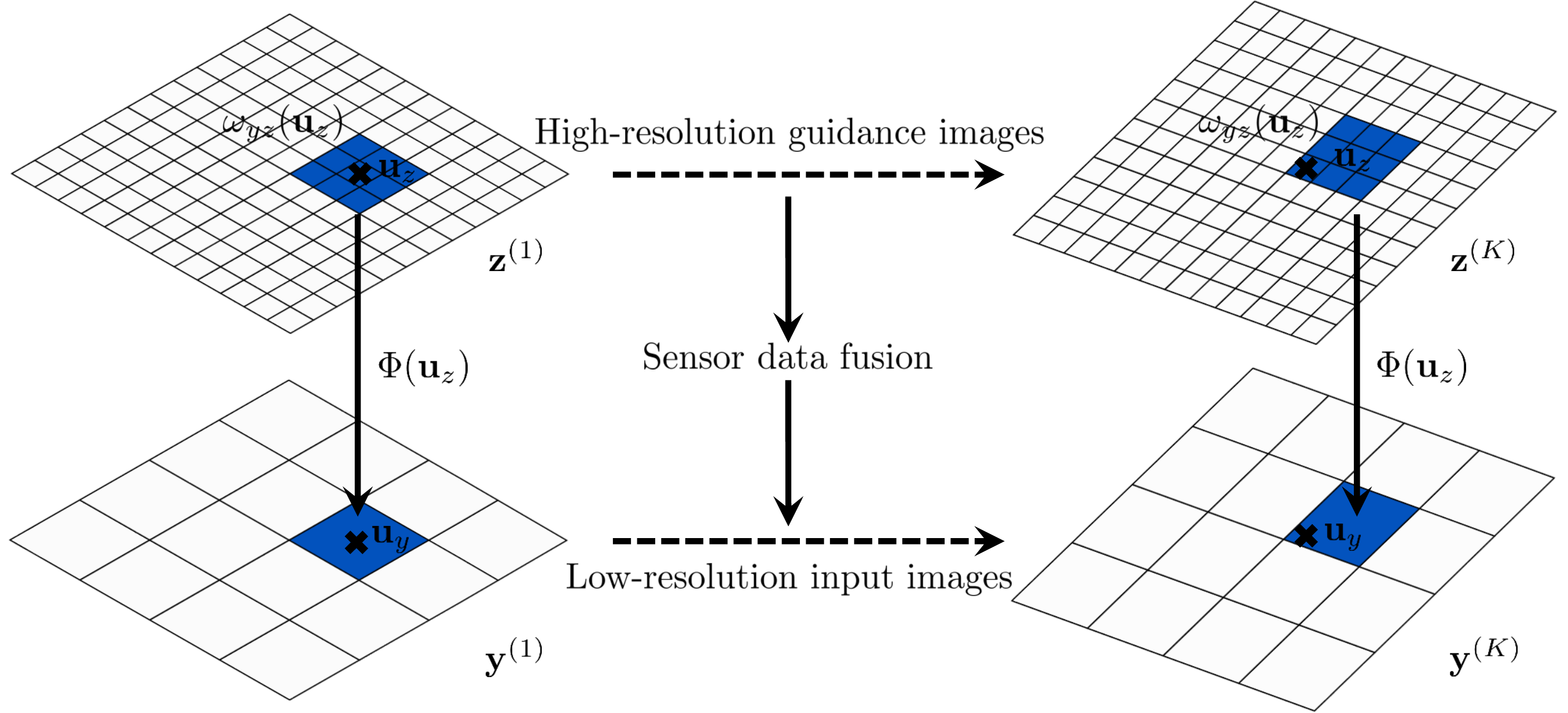}
	\caption[Sensor data fusion for multi-sensor super-resolution]{Illustration of sensor data fusion between low-resolution input images and the associated high-resolution guidance data for multi-sensor super-resolution. Each pair $(\LRFrame{k}, \GuideFrame{k})$ is geometrically aligned up to a scale factor. In the proposed framework, the local neighborhood $\SenFusNeighbor(\Point_{\GuideSym})$ centered at the pixel position $\Point_{\GuideSym}$ in a guidance image is mapped to the pixel position $\Point_{\LRSym}$ in an input image according to the mapping $\SenFusMapping{\Point_{\GuideSym}}$. We assume a fixed mapping $\SenFusMapping{\Point_{\GuideSym}}$ over the sequence of input and guidance images.}
	\label{fig:sensorFusion}
\end{figure}

Once the guidance data $\Guide$ is fused with the low-resolution input data $\LR$, the proposed framework builds on the \gls{map} estimator for the high-resolution image $\hat{\HR}$ according to:
\begin{equation}
	\label{eqn:msrMapEstimator}
	\HR_{\text{MAP}} = \argmin_{\HR} \left\{  \DataTerm[MSR]{\HR, \Guide} + \RegWeight \RegTerm[MSR]{\HR, \Guide} \right\}.
\end{equation}
The guidance data $\Guide$ is involved in two components. In terms of the observation model defined by the data fidelity term $\DataTerm[MSR]{\HR, \Guide}$\label{notation:msrDataFidelity}, guidance images are used to estimate subpixel motion. This avoids a direct motion estimation on low-resolution frames with the goal to enhance the accuracy of super-resolution. In terms of the image prior, a spatially adaptive regularization term $\RegTerm[MSR]{\HR, \Guide}$\label{notation:msrRegTerm} weighted by $\RegWeight \geq 0$ is used. This term exploits both, a super-resolved image $\HR$ as well as the guidance data $\Guide$ with the goal of taking advantage of structural correlation across both modalities. Both novelties of the multi-sensor framework over the single-sensor counterpart are introduced in the following subsections.

\subsection{Motion Estimation using Guidance Images}
\label{sec:05_MotionEstimationUsingGuidanceImages}

The proposed framework explicitly employs displacement vector fields as the most flexible approach to model subpixel motion. For this purpose, motion estimation is realized by means of optical flow computation \cite{Liu2009}. In \cite{Zhao2002}, Zhao and Sawhney suggested that reconstruction-based super-resolution is feasible with this kind of motion estimation under the prerequisite of small noise in the estimated flow. However, the accuracy of optical flow is limited by noise, blur or aliasing present in low-resolution data. For this reason, many attempts have been made to treat image reconstruction and optical flow estimation in a joint framework, \eg by probabilistic methods \cite{Fransens2007}. This has the goal to compensate for inaccurate optical flow. 

To meet the requirement regarding precise optical flow and to circumvent its direct estimation on low-resolution data, the proposed motion estimation is driven by guidance images and implemented as computationally efficient local filtering of dense displacement vector fields. In this \textit{filter-based} approach, we determine displacements fields $\DispVecFieldFrame[\GuideSym]{\Point_{\GuideSym}}{k}$ of each guidance image $\GuideFrame{k}$ relative to a fixed reference frame $\GuideFrame{r}$ with $r \neq k$. In order to obtain the associated displacements in the domain of the input images, we take advantage of the sensor data fusion with the guidance data. For the realization of motion estimation, we assume that motion present in input data is also encoded by the guidance data if both modalities are co-aligned. Intuitively this means that both sensors involved in this setup "`see"' the same scene and describe the same motion. 

Based on sensor data fusion, a displacement vector field $\DispVecField[\LRSym]{\Point_{\LRSym}}$\label{notation:dispVecFieldInput} for a single input frame $\LR$ relative to the reference frame is obtained from $\DispVecField[\GuideSym]{\Point_{\GuideSym}}$\label{notation:dispVecFieldGuide} estimated on guidance images as depicted in \fref{fig:displacementFieldFiltering}. This process consists of two steps:
\begin{enumerate}
	\item The displacement field $\DispVecField[\GuideSym]{\Point_{\GuideSym}}$ that is given in terms of pixel units in the domain of the guidance images is first rescaled element-wise to:
		\begin{equation}
			\tilde{\DispVecFieldSym}_{\GuideSym}(\Point_{\GuideSym})
			= 
			\begin{pmatrix} 
				\frac{\LRDimU}{\GuideDimU} \cdot \DispVecFieldSym_{\GuideSym, \CoordU}(\Point_{\GuideSym}) \\[1.0ex]
				\frac{\LRDimV}{\GuideDimV} \cdot \DispVecFieldSym_{\GuideSym, \CoordV}(\Point_{\GuideSym})
			\end{pmatrix}.	
		\end{equation}
		Thus, the displacements $\tilde{\DispVecFieldSym}_{\GuideSym}(\Point_{\GuideSym})$ are defined in units of low-resolution pixels.
		
	\item The displacements on the input frames are determined from the intermediate displacements obtained in the first stage by resampling described by the filter operation $\DispVecField[\LRSym]{\Point} = \MotionResamplingOp{ \tilde{\DispVecFieldSym}_{\GuideSym}(\Point)}$\label{notation:motionResamplingOp}. The filtering of the displacement field is performed element-wise according to:
	\begin{equation}
			\label{eqn:subpixelDisplacementResampling}
			\begin{split}
				\DispVecField[\LRSym]{\Point_{\LRSym}} 
					&= \MotionResamplingOp{ \tilde{\DispVecFieldSym}_{\GuideSym}(\Point_{\GuideSym})} \\
					&\defeq 
					\begin{pmatrix}
						 \MotionResamplingOp[\SenFusNeighbor(\Point_{\GuideSym}), \CoordU]{ \tilde{\DispVecFieldSym}_{\GuideSym, \CoordU}(\Point_{\GuideSym}) }  \\[1.0ex]
						 \MotionResamplingOp[\SenFusNeighbor(\Point_{\GuideSym}), \CoordV]{ \tilde{\DispVecFieldSym}_{\GuideSym, \CoordV}(\Point_{\GuideSym}) }
					\end{pmatrix},	
			\end{split}
		\end{equation}
		where $\SenFusNeighbor(\Point_{\GuideSym})$ denotes the set of pixels in a local window centered at position $\Point_{\GuideSym}$ that corresponds to a single pixel position $\Point_{\LRSym}$ in the low-resolution input image, see \fref{fig:displacementFieldFiltering}.
\end{enumerate}

This filter-based technique enables noise suppression to deal with single erroneously estimated displacements. In addition to noise suppression, it needs to preserve motion discontinuities in the original displacement fields. To this end, the resampling used in the second stage of the proposed technique is formulated as local median filtering:
\begin{equation}
	\MotionResamplingOp[\SenFusNeighbor(\Point_{\GuideSym}), i]{ \tilde{\DispVecFieldSym}_{\GuideSym, i}(\Point_{\GuideSym}) }
	= \Median[\PointV_{\GuideSym} \in \SenFusNeighbor(\Point_{\GuideSym})]{\tilde{\DispVecFieldSym}_{\GuideSym, i}(\PointV_{\GuideSym})},
\end{equation}
where $\Median[\PointV \in \SenFusNeighbor(\Point)]{\cdot}$ computes the median of the displacements estimated in the local neighborhood $\SenFusNeighbor(\Point)$ in the coordinate direction $i \in \{\CoordU, \CoordV \}$.

\begin{figure}[!t]
	\centering
		\includegraphics[width=0.90\textwidth]{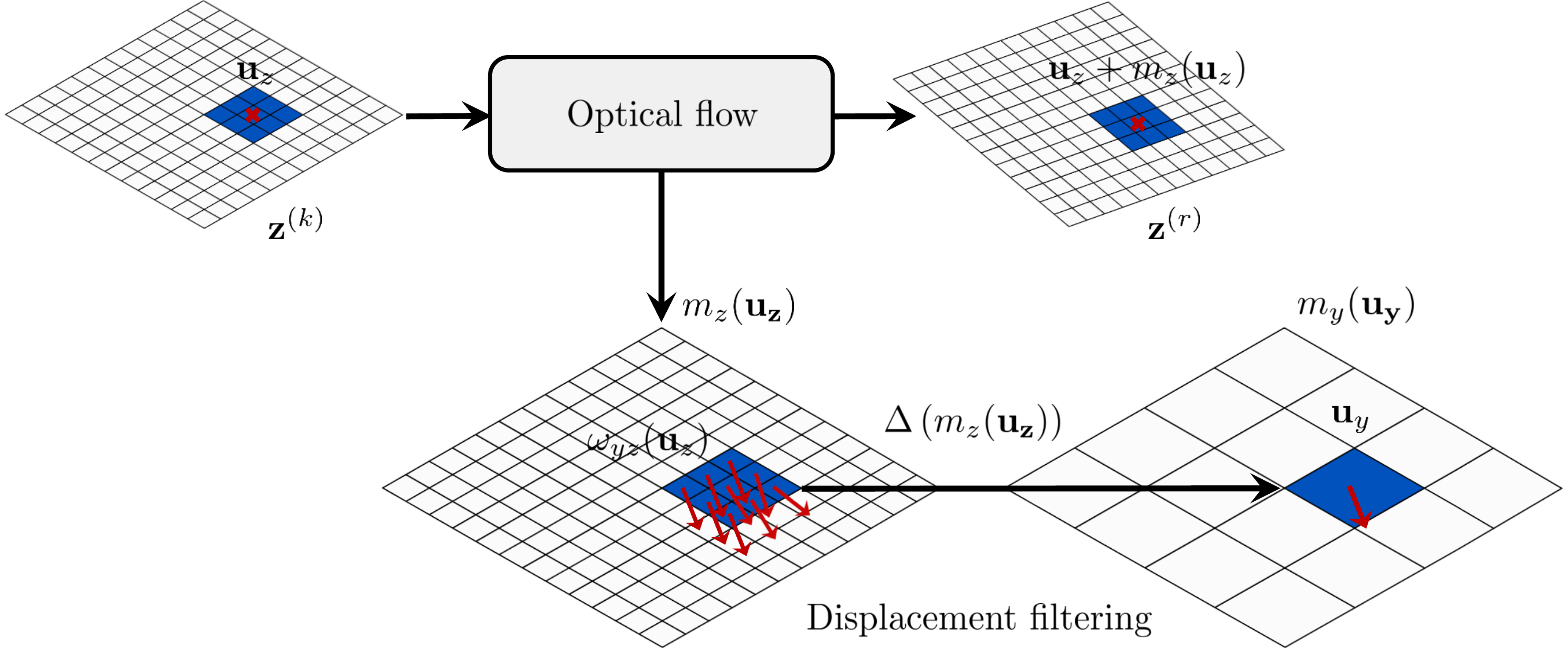}
	\caption[Filter-based motion estimation using guidance images]{Flowchart of filter-based motion estimation using guidance images. First, the displacement field $\DispVecField[\GuideSym]{\Point_{\GuideSym}}$ is gained by optical flow estimation of the frame $\GuideFrame{k}$ towards the reference frame $\GuideFrame{r}$. Then, the displacement field $\DispVecField[\LRSym]{\Point_{\LRSym}}$ is determined from $\DispVecField[\GuideSym]{\Point_{\GuideSym}}$ using patch-wise filtering with the neighborhood $\SenFusNeighbor(\Point_{\GuideSym})$.}
	\label{fig:displacementFieldFiltering}
\end{figure}

\subsection{Spatially Adaptive Regularization using Guidance Images}
\label{sec:05_SpatiallyAdaptiveRegularizationUsingGuidanceImages}

Spatially adaptive regularization exploits the fact that input and guidance images capture the same structural content but are encoded by different modalities. In particular, we make use of the assumption that there exist correlations between both image types due to common structures. While this is of course not completely true over the entire image, one can assume correlations in terms of a set of structural features, \eg areas or edges. This idea takes the same line as related concepts that became a standard in hybrid image processing, \eg color-guided range upsampling \cite{Park2011,Ferstl2013,Kiechle2013} or \gls{ct}-guided \gls{pet} reconstruction \cite{Kennedy2007}.

In this chapter, correlations among modalities are modeled by a weighting function $\WeightsAFun: \DomainHR \times \DomainGuide \rightarrow \RealNonNegN{N}$\label{notation:amsrWeights} that exploits the image $\HR$ and a corresponding guidance image $\vec{z}$. This function is used to control the regularization term:
\vspace{-0.5em}
\begin{equation}
	\label{eqn:spatialDomainRegularizationTerm}
	\RegTerm[MSR]{\HR, \Guide}	= \WeightsAFun \left( \HR, \Guide \right)^\top \LossFun[\text{MSR}]{\HPMat \HR},
\end{equation}
where $\LossFun[\text{MSR}]{\cdot}$\label{notation:msrRegLossFun} denotes a loss function applied on a high-pass filtered version of $\HR$ to penalize residual noise\footnote{Since the proposed framework is flexible regarding the choice of the loss function $\LossFun[\text{MSR}]{\cdot}$, we tailor the regularization to the characteristics of specific applications. For instance, the Huber loss can be used to implement piecewise-smooth regularization, see \cite{Kohler2013a}.}. This is done using a discrete filter modeled by the circulant matrix $\HPMat \in \RealMN{N}{N}$. Similar to the sparse regularization proposed in \cref{sec:RobustMultiFrameSuperResolutionWithSparseRegularization}, the adaptive weights $\WeightsAFun(\HR, \Guide)$ are controlled in such a way that the regularizer does not penalize discontinuities in $\HR$. In contrast to \cref{sec:RobustMultiFrameSuperResolutionWithSparseRegularization}, the selection of the weights is steered by guidance data. For this purpose, the guidance images are used to extract discontinuities that are considered as relevant structural features. This process is driven by edge detection on the guidance image $\Guide$ to obtain an edge map $\tau: \DomainGuide \rightarrow \{ 0, 1 \}^L$\label{notation:edgeMap}, where $\tau(\Point) = 1$ indicates an edge at position $\Point$. 

The adaptive weights are first computed in the domain of the guidance data as\label{notation:amsrWeightsGuide}:
\vspace{-0.5em}
\begin{equation}
	\tilde{\WeightsAFun} \left( \tilde{\HR}, \Guide \right) 
	= 
	\begin{pmatrix}
		\tilde{\WeightsAFun}_1 \left(\tilde{\HR}, \Guide \right)
		& \tilde{\WeightsAFun}_2 \left(\tilde{\HR}, \Guide \right)
		& \ldots
		& \tilde{\WeightsAFun}_L \left(\tilde{\HR}, \Guide \right) 
	\end{pmatrix}^\top \in \RealNonNegN{L},
\end{equation}
where $\tilde{\HR}$ is the image $\HR$ resampled to the size of the guidance image $\Guide$ using bicubic interpolation. The weight at the $i$-th pixel position $\Point_i$ is computed by:
\begin{equation}
	\label{eqn:spatiallyAdaptiveMSRWeightingFunction}
	\tilde{\WeightsAFun}_i \left(\tilde{\HR}, \Guide \right) =
	\begin{cases}
		\exp \left\{- \frac{ \SimMeasureLocal{\tilde{\HR}}{\Guide}{\AMSRNeighbor(\Point_i)} }{ \AMSRContrastFactor } \right\} 
			& \text{if\,\,} \tau(\Point_i) = 1 \\
		1
			& \text{otherwise}
	\end{cases},
\end{equation}
where $\SimMeasureLocal{\tilde{\HR}}{\Guide}{\AMSRNeighbor(\Point_i)}$\label{notation:simMeasureLocal} reflects the degree of correlation computed by a similarity measure between the image $\tilde{\HR}$ and guidance data $\Guide$ in a $\AMSRPatchSize \times \AMSRPatchSize$\label{notation:amsrPatchSize} local neighborhood $\AMSRNeighbor(\Point_i)$\label{notation:amsrNeighbor} centered at $\Point_i$. The parameter $\AMSRContrastFactor$\label{notation:amsrContrastFactor} denotes a contrast factor to map the local image similarity at positions corresponding to an edge to a weight $\tilde{\WeightsAFun}_i(\tilde{\HR}, \Guide) \in [0, 1]$. Finally, the weights $\WeightsAFun( \HR, \Guide) \in \RealNonNegN{N}$ for the regularization term in \eref{eqn:spatialDomainRegularizationTerm} are obtained by bicubic interpolation of $\tilde{\WeightsAFun} (\tilde{\HR}, \Guide) \in \RealNonNegN{L}$ to the domain of super-resolved data. This reduces the impact of regularization for image regions associated with edges in the guidance data according to the similarity measure.

In \eref{eqn:spatiallyAdaptiveMSRWeightingFunction}, the similarity measure $\SimMeasureLocal{\tilde{\HR}}{\Guide}{\AMSRNeighbor(\Point)}$ indicates how reasonable the assumption of correlations in terms of discontinuities actually is. In particular, it needs to downweight the impact of guidance data if this assumption does not hold true for certain image regions. Multi-modal measures are utilized to analyze these similarities. This can be achieved by cross correlation as used in mutual-structure filters \cite{Shen2015} or by information theoretic measures. In this work, the similarity is assessed by \gls{lmi} that has been also successfully applied in related fields of image enhancement like adaptive \gls{tv} denoising \cite{Guo2008}. The \gls{lmi} is computed following the definition of Pluim \etal \cite{Pluim2003}\label{notation:miMeasureLocal}:
\begin{equation}
	\label{eqn:defLMI}
		\MiMeasureLocal{\tilde{\HR}}{\Guide}{\AMSRNeighbor(\Point)}
		= \sum \limits_{\tilde{\HRSym_i}, \GuideSym_i \in \AMSRNeighbor(\Point)} 
		\Pdf{\tilde{\HRSym_i}, \GuideSym_i} \log 
			\left( 
				\frac{
					\Pdf{\tilde{\HRSym_i}, \GuideSym_i}
				}
				{
					\Pdf{\tilde{\HRSym_i}} \Pdf{\GuideSym_i}
				} 
			\right),
\end{equation}
where $\PdfT{\tilde{\HRSym_i}, \GuideSym_i}$ is an estimate of the joint \gls{pdf} of the samples extracted from the local neighborhood $\AMSRNeighbor(\Point)$ in the images $\tilde{\HR}$ and $\Guide$. Similarly, $\PdfT{\tilde{\HRSym_i}}$ and $\PdfT{\GuideSym_i}$ are the corresponding marginals. To define $\SimMeasureLocal{\tilde{\HR}}{\Guide}{\AMSRNeighbor(\Point)}$, we use the normalized \gls{lmi}:
\begin{equation}
	\label{eqn:defNormLMI}
	\SimMeasureLocal{\tilde{\HR}}{\Guide}{\AMSRNeighbor(\Point)}
	= -\frac
		{
			\MiMeasureLocal{\tilde{\HR}}{\Guide}{\AMSRNeighbor(\Point)}
		}
		{
			\sum \limits_{\tilde{\HRSym_i}, \GuideSym_i \in \AMSRNeighbor(\Point)} 
			\Pdf{\tilde{\HRSym_i}, \GuideSym_i} \log \Pdf{\tilde{\HRSym_i}, \GuideSym_i}
		}.
\end{equation}
The size of $\AMSRNeighbor(\Point)$ in \eref{eqn:defNormLMI} is adjusted to balance between too many empty bins in the joint histogram $\PdfT{\tilde{\HRSym_i}, \GuideSym_i}$ and a too low resolution of \gls{lmi}.

\subsection{Numerical Optimization}

In this basic approach to multi-sensor super-resolution, we solve \eref{eqn:msrMapEstimator} with a Gaussian observation model. This leads to the data fidelity term:
\begin{equation}
	\DataTerm[MSR]{\HR, \Guide} 
	= \sum_{k = 1}^K \left| \left| \LRFrame{k} - \SystemMatFrameGuide{k} \HR \right| \right|_2^2,
\end{equation}
where the system matrix \smash{$\SystemMatFrameGuide{k}$} is parametrized with the displacement fields obtained from the optical flow on the guidance images, see \sref{sec:05_MotionEstimationUsingGuidanceImages}. The regularization term is driven by the adaptive weights $\WeightsA$, see \sref{sec:05_SpatiallyAdaptiveRegularizationUsingGuidanceImages}.

The optimization of \eref{eqn:msrMapEstimator} is performed in an alternating fashion to jointly estimate the super-resolved image and the weights $\WeightsA$. We limit this scheme to two stages corresponding to two optimization loops as outlined in \aref{alg:msrTwoStageAlgorithm}. 

In the first stage, we reconstruct an intermediate solution for the super-resolved image using \gls{scg} iterations \cite{Nabney2002} given uniform weights in the regularization term. The initial guess for these iterations is obtained by bicubic interpolation of the reference input image. In the second stage, we compute the spatially adaptive weights $\WeightsA$ using the super-resolved image of the first stage and the respective guidance image. Instead of using a single image, the spatially adaptive weights are obtained from the motion-compensated temporal median of the sequence $\GuideSequence{1}{K}$ to enhance the accuracy of edge detection. Finally, given the weights $\WeightsA$, the image reconstructed in the first stage is iteratively refined by \gls{scg} and spatially adaptive regularization.
\begin{algorithm}[!t]
	\small
	\caption{Two-stage multi-sensor super-resolution}
	\label{alg:msrTwoStageAlgorithm}
	\begin{algorithmic}[1]
		\Require Initial guess for high-resolution image $\HR$
		\Ensure Final high-resolution image $\HR$ and spatially adaptive regularization weights $\WeightsA$
		\For{$k = 1, \ldots, K$}
			\State Compute optical flow $\DispVecFieldFrame[\GuideSym]{\Point_{\GuideSym}}{k}$ of $\GuideFrame{k}$ towards the reference $\GuideFrame{r}$
			\State Compute $\DispVecFieldFrame[\LRSym]{\Point_{\GuideSym}}{k}$ by local filtering of $\DispVecFieldFrame[\GuideSym]{\Point_{\GuideSym}}{k}$ according to \eref{eqn:subpixelDisplacementResampling}
		\EndFor
		\While{\gls{scg} convergence criteria not fulfilled}
			\State Update $\HR$ by \gls{scg} iteration for \eref{eqn:msrMapEstimator} using uniform weights $\WeightsA = \Ones$
		\EndWhile
		\State Set $\Guide$ to the temporal motion-compensated median of $\GuideSequence{1}{K}$
		\State Compute spatially adaptive weights $\WeightsA$ according to \eref{eqn:spatiallyAdaptiveMSRWeightingFunction}
		\While{\gls{scg} convergence criteria not fulfilled}
			\State Update $\HR$ by \gls{scg} iteration for \eref{eqn:msrMapEstimator} using the adaptive weights $\WeightsA$
		\EndWhile		
	\end{algorithmic}	
\end{algorithm}

\section{Outlier Detection for Robust Multi-Sensor Super-Resolution}
\label{sec:05_OutlierDetectionForRobustMultiSensorSuperResolution}

The framework presented in the previous section is based on several idealizing assumptions limiting its robustness in real-world applications. First and foremost, it neglects outliers in the optical flow estimated on guidance data. While noise in the displacement fields can be compensated by the proposed filter-based technique, motion estimation is prone to occlusions or inaccurate flows in texture-less regions. In addition, we derived \aref{alg:msrTwoStageAlgorithm} under a Gaussian observation model that neglects outliers in low-resolution data. 

In order to deal with the aforementioned issues, this section presents an outlier detection scheme as extension of the two-stage framework. This approach is divided into two separate detection schemes applied on guidance and input data. These techniques yield confidence maps associated with the input data and their displacement fields, which are combined for robust super-resolution.  

\subsection{Outlier Detection on Guidance Images}

Similar to the filter-based technique in \sref{sec:05_MotionEstimationUsingGuidanceImages}, the detection of outliers in terms of motion estimation is driven by guidance images. The proposed outlier detection is inspired by the image similarity based method of Zhao and Sawhney \cite{Zhao2002} but adopted in such a way that it is applicable on guidance images instead of using the low-resolution input frames directly. This approach determines the reliability of the estimated displacement fields that affects the robustness of super-resolution. 

For outlier detection, the reference frame in the domain of the guidance images denoted as \smash{$\GuideFrame{r}$} is warped towards each of the remaining frames \smash{$\GuideFrame{k}$, $k \neq r$} according to the estimated displacements. Then, we assess the consistency of each target frame \smash{$\GuideFrame{k}$} relative to the warped reference \smash{$\FrameIdx{\Warp{\Guide}}{k}$} as shown in \fref{fig:guidanceImageOutlierDetection} for one of these pairs. Similar to the approach in \cite{Zhao2002}, this consistency is assessed by the \gls{ncc} that is used as a local similarity measure. The local \gls{ncc} at the $i$-th pixel position $\Point_i$ in the guidance images is computed for the local window $\SenFusNeighbor(\Point_i)$ centered at $\Point_i$ according to\label{notation:nccMeasureLocal}:
\begin{equation}
	\label{eqn:localNCC}
	\NccMeasureLocal{\GuideFrame{k}}{\FrameIdx{\Warp{\Guide}}{k}}{\SenFusNeighbor(\Point_i)}
	=
	\frac
	{ \displaystyle\sum_{\PointV_j \in \SenFusNeighbor(\Point_i)} 
		\Big( \FrameIdx{\GuideSym_j}{k} - \mu_i \Big) 
		\Big( \FrameIdx{\tilde{\GuideSym}_j}{k} - \Warp{\mu_i} \Big) 
	}
	{ 
		\sqrt{ 
			\displaystyle\sum_{\PointV_j \in \SenFusNeighbor(\Point_i)} \Big( \FrameIdx{\GuideSym_j}{k} - \mu_i \Big)^2 
			\displaystyle\sum_{\PointV_j \in \SenFusNeighbor(\Point_i)} \Big( \FrameIdx{\tilde{\GuideSym}_j}{k} - \Warp{\mu_i} \Big)^2 		} 
	}, 
\end{equation}
where \smash{$\FrameIdx{\GuideSym_j}{k}$} and \smash{$\FrameIdx{\tilde{\GuideSym}_j}{k}$} are the elements in \smash{$\GuideFrame{k}$} and \smash{$\FrameIdx{\Warp{\Guide}}{k}$} at the position $\PointV_j \in \SenFusNeighbor(\Point_i)$, and \smash{$\mu_i$} and \smash{$\Warp{\mu_i}$} are the local means of \smash{$\GuideFrame{k}$} and \smash{$\FrameIdx{\Warp{\Guide}}{k}$} in $\SenFusNeighbor(\Point_i)$, respectively. In order to transform this local similarity to the domain of the low-resolution images, it is computed by means of patch-wise processing as shown in \fref{fig:guidanceImageOutlierDetection}. 

The local similarity measures the fidelity of the image warping according to the optical flow and is used for outlier detection. In the proposed detection scheme, the local \gls{ncc} computed in the range $[-1,+1]$ is used to determine the confidence weight\label{notation:confWeightGuidePixel}\label{notation:nccThreshold}:
\begin{align}
	\WeightsBFun_{\GuideSym, i} \left( \GuideFrame{k} \right) =
	\begin{cases}
		\frac{1}{2} \NccMeasureLocal{\GuideFrame{k}}{\FrameIdx{\Warp{\Guide}}{k}}{\SenFusNeighbor(\Point_i)} + \frac{1}{2}	
			& \text{if}~ \NccMeasureLocal{\GuideFrame{k}}{\FrameIdx{\Warp{\Guide}}{k}}{\SenFusNeighbor(\Point_i)} \geq \SimMeasureSym_{0} \\
		0				
			& \text{otherwise}
	\end{cases},
\end{align}
where $\SimMeasureSym_{0} \in [-1, +1]$ is a fixed threshold to classify observations associated with a poor local similarity as an outlier. For the remaining observations that are not classified as outliers, a higher local image similarity due to more accurate optical flow estimation indicates a higher confidence. The confidence map for the $k$-th frame is assembled as\label{notation:confWeightGuideFrame}:  
\begin{equation}
	\label{eqn:guidanceDataConfidenceSingleFrame}
	\WeightsBFun_{\GuideSym} \left(\GuideFrame{k} \right) =
	\begin{pmatrix}
			\WeightsBFun_{\GuideSym, 1} \left(\GuideFrame{k} \right)~
			& \WeightsBFun_{\GuideSym, 2} \left(\GuideFrame{k} \right)~
			& \ldots
			& \WeightsBFun_{\GuideSym, M} \left(\GuideFrame{k} \right)
	\end{pmatrix}^\top.
\end{equation}
Then, the joint confidence map for the entire image sequence is constructed as\label{notation:confWeightGuide}:
\begin{equation}
	\label{eqn:guidanceDataConfidence}
	\WeightsBFun_{\GuideSym} \left(\Guide \right) =
	\begin{pmatrix}
			\WeightsBFun_{\GuideSym} \left(\GuideFrame{1} \right)~
			& \WeightsBFun_{\GuideSym} \left(\GuideFrame{2} \right)~
			& \ldots
			& \WeightsBFun_{\GuideSym} \left(\GuideFrame{K} \right)
	\end{pmatrix}^\top.
\end{equation}

\begin{figure}[!t]
	\centering
		\includegraphics[width=0.97\textwidth]{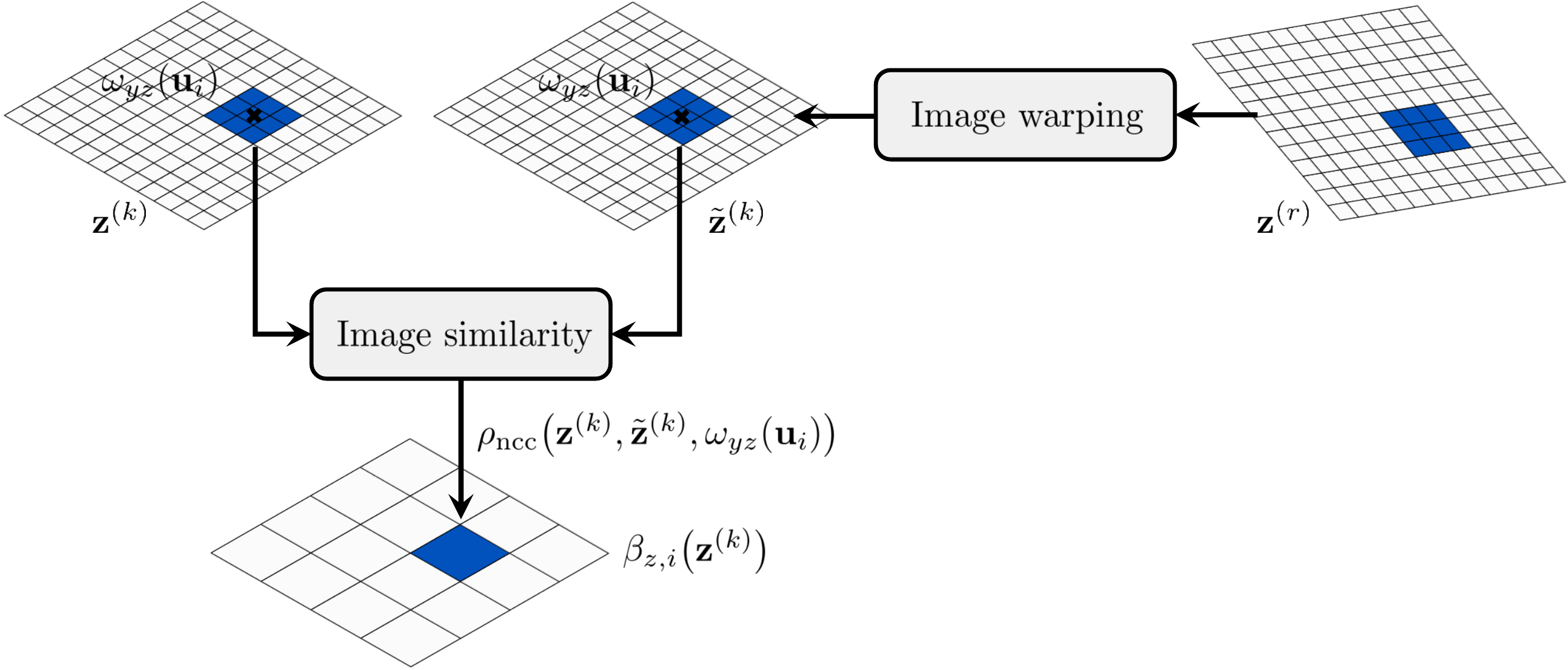}
	\caption[Outlier detection on guidance images]{Illustration of outlier detection on guidance images. The local image similarity between the $k$-th frame $\GuideFrame{k}$ and the warped reference $\FrameIdx{\Warp{\Guide}}{k}$ according to the estimated optical flow is used to calculate the confidence weight $\WeightsBFun_{\GuideSym,i} (\GuideFrame{k} )$. The resulting confidence map $\WeightsBFun_{\GuideSym} \big(\GuideFrame{k} \big)$ is constructed by patch-wise processing with the neighborhood $\SenFusNeighbor(\Point_i)$ and defined in the domain of the input images.}
	\label{fig:guidanceImageOutlierDetection}
\end{figure}

\subsection{Outlier Detection on Input Images}

In addition to outlier detection in the estimated displacement fields based on the guidance images, the low-resolution input frames themselves are assessed to remove outliers. This takes outliers due to non-Gaussian noise into account that cannot be detected on the guidance images. This outlier detection is formulated in an implicit way in accordance to the algorithm presented in \sref{sec:04_RobustSuperResolutionReconstruction}. 

Let $\HR$ be an estimate for the super-resolved image. Then, we define the confidence weight associated with the $i$-th pixel in the $k$-th  frame \smash{$\LRFrame{k}$} according to:\label{notation:confWeightInputPixel}
\begin{equation}
		\WeightsBFun_{\LRSym, i} \left( \HR, \LRFrame{k} \right) = \WeightsBFun_{\text{bias}, i} \left( \HR, \LRFrame{k} \right)	\WeightsBFun_{\text{local}, i} \left( \HR, \LRFrame{k} \right),
\end{equation}
where the weighting functions are given by:
\begin{align}
	\WeightsBFun_{\text{bias}, i} \left(\HR, \LRFrame{k} \right) &=
	\begin{cases}
		1
			& \text{if}~ \left| \Median{\LRFrame{k} - \SystemMatFrameGuide{k} \HR} \right| \leq c_{\text{bias}} \\
		0
			& \text{otherwise}
	\end{cases},\\
	\WeightsBFun_{\text{local}, i} \left(\HR, \LRFrame{k} \right) &=
	\begin{cases}
		1																										
			& \text{if}~ \left| \VecEl{\LRFrame{k} - \SystemMatFrameGuide{k} \HR}{i} \right| \leq c_{\text{local}}\NoiseStd \\
		\frac{c_{\text{local}} \NoiseStd}{ \left| \VecEl{\LRFrame{k} - \SystemMatFrameGuide{k} \HR}{i} \right|}	
			& \text{otherwise}
	\end{cases},
\end{align}
with noise level $\NoiseStd$ and tuning constants $c_{\text{bias}}$ and $c_{\text{local}}$ to detect biased frames along with local outliers as shown in \sref{sec:04_NumericalOptimization}. The confidence map for the $k$-th low-resolution frame is assembled according to\label{notation:confWeightInput}:
\begin{equation}
	\label{eqn:lrDataConfidenceFrame}
	\WeightsBFun_{\LRSym} \left(\HR, \LRFrame{k} \right) =
	\begin{pmatrix}
		\WeightsBFun_{\LRSym, 1} \left(\HR, \LRFrame{k} \right)~
		& \WeightsBFun_{\LRSym, 2} \left(\HR, \LRFrame{k} \right)~
		& \ldots
		& \WeightsBFun_{\LRSym, \LRSize} \left(\HR, \LRFrame{k} \right)
	\end{pmatrix}^\top.
\end{equation}
Then, the joint confidence map for all low-resolution observations is given by:
\begin{equation}
	\label{eqn:lrDataConfidence}
	\WeightsBFun_{\LRSym} \left(\HR, \LR \right) =
	\begin{pmatrix}
		\WeightsBFun_{\LRSym} \left(\HR, \LRFrame{1} \right)~
		& \WeightsBFun_{\LRSym} \left(\HR, \LRFrame{2} \right)~
		& \ldots
		& \WeightsBFun_{\LRSym} \left(\HR, \LRFrame{\NumFrames} \right)
	\end{pmatrix}^\top.
\end{equation}

\subsection{Numerical Optimization}

For an outlier-aware numerical optimization, super-resolution is performed by means of iteratively re-weighted minimization. We estimate the super-resolved image iteratively via a sequence of weighted minimization problems:
\begin{equation}
	\label{eqn:msrRobustEstimator}
	\IterationIdx{\HR}{t} = \argmin_{\HR} 
	\left\{ 
		\left(\LR - \SystemMatGuide \HR \right)^\top \IterationIdx{\WeightsBMat}{t} \left(\LR - \SystemMatGuide \HR \right)
		+ \RegWeight \RegTerm[MSR]{\HR, \Guide} 
	\right\},
\end{equation}
where $\RegTerm[MSR]{\HR, \Guide}$ denotes the spatially adaptive regularization term defined in \eref{eqn:spatialDomainRegularizationTerm} with constant regularization weight\footnote{An adaptive version of this algorithm, where the regularization weight $\RegWeight$ is re-computed per iteration can be developed using the cross validation scheme presented in \sref{sec:04_NumericalOptimization}.} $\RegWeight \geq 0$. In this weighted minimization, $\IterationIdx{\WeightsBMat}{t}$\label{notation:jointConfidence} denotes the confidence map at iteration $t$. In order to take outliers in the displacement fields and the low-resolution observations into account, $\IterationIdx{\WeightsBMat}{t}$ is constructed according to:
\begin{equation}
	\label{eqn:jointConfidenceMap}
	\IterationIdx{\WeightsBMat}{t} = \Diag{
	\WeightsBFun_{\GuideSym} \left( \Guide \right) 
	\odot 
	\WeightsBFun_{\LRSym} \left( \IterationIdx{\HR}{t-1}, \LR \right)},
\end{equation}
where $\odot$ is the Hadamard (element-wise) product and $\WeightsBFun_{\GuideSym}(\Guide)$ is the confidence map determined from the local image similarity according to \eref{eqn:guidanceDataConfidence}. Notice that these confidence weights can be pre-computed and kept fixed over the iterations. The confidence map $\WeightsBFun_{\LRSym} (\IterationIdx{\HR}{t-1}, \LR)$ is updated dynamically according to \eref{eqn:lrDataConfidence} based on the estimate $\IterationIdx{\HR}{t-1}$ obtained at the previous iteration. In this dynamic weighting scheme, we use an adaptive estimate of the noise standard deviation $\NoiseStd$ at each iteration according to the \gls{mad} rule, see \sref{sec:04_NumericalOptimization}. 

This iterative procedure is initialized by the super-resolved image $\IterationIdx{\HR}{0}$ and the corresponding adaptive weights $\WeightsA$ for the regularization term. These initializations are obtained by the two-stage approach introduced in \aref{alg:msrTwoStageAlgorithm}. Afterwards, the super-resolved image is gradually refined using \gls{scg} iterations. The overall optimization scheme is outlined in \aref{alg:msrOutlierDetectionAlgorithm}.

\begin{algorithm}[!t]
	\small
	\caption{Robust multi-sensor super-resolution using outlier detection}
	\label{alg:msrOutlierDetectionAlgorithm}
	\begin{algorithmic}[1]
		\Require Initial guess for high-resolution image $\IterationIdx{\HR}{0}$ and adaptive regularization weights $\WeightsA$
		\Ensure Final high-resolution image $\HR$ and joint confidence map $\WeightsBMat$
		\For{$k = 1 \dots K$}
			\State Construct confidence map $\WeightsBFun_{\GuideSym} (\GuideFrame{k} )$ according to \eref{eqn:guidanceDataConfidenceSingleFrame}
		\EndFor
		\State Construct confidence map $\WeightsBFun_{\GuideSym} (\Guide)$ for all guidance images according to \eref{eqn:guidanceDataConfidence}
		\State $t \gets 1$
		\While{Convergence criterion not fulfilled}
			\State Update confidence map $\WeightsBFun_{\LRSym}(\IterationIdx{\HR}{t-1}, \LR)$ according to \eref{eqn:lrDataConfidence}
			\State Update joint confidence map $\IterationIdx{\WeightsBMat}{t}$ according to \eref{eqn:jointConfidenceMap}
			\While{\gls{scg} convergence criterion not fulfilled}
				\State Update $\IterationIdx{\HR}{t}$ by \gls{scg} iteration for \eref{eqn:msrRobustEstimator} using adaptive weights $\WeightsA$
			\EndWhile
			\State $t \gets t + 1$
		\EndWhile
	\end{algorithmic}	
\end{algorithm}

\section{Application to Hybrid Range Imaging}
\label{sec:05_ApplicationToHybridRangeImaging}

This section validates the proposed multi-sensor framework in the field of hybrid range imaging that is considered as example application. In this area, we aim at reconstructing high-resolution surface information from sequences of low-resolution range images to overcome resolution limitations of current range sensors. Prior work in this field approached this task solely on range images as shown by Schuon \etal \cite{Schuon2008, Schuon2009} or Bhavsar and Rajagopalan \cite{Bhavsar2012}. Unlike these single-sensor approaches, we employ high-resolution color images as a guidance to super-resolve low-resolution range data.

We first introduce a tailor-made image formation model for range imaging. Subsequently, this model is adopted in the multi-sensor framework to formulate range super-resolution reconstruction. Eventually, we present a proof-of-concept evaluation for \gls{tof} imaging based on simulated datasets. For a thorough experimental evaluation on real image data within the scope of interventional medical imaging, we refer to \cref{sec:HybridRangeImagingForImageGuidedSurgery}.

\subsection{Image Formation Model for Range Imaging}
\label{sec:05_ImageFormationModelForRangeImaging}

In terms of the image formation model, we need to adopt the model presented in \cref{sec:ComputationalFrameworkForMultiFrameSuperResolution} to describe the formation of low-resolution range images from high-resolution ones. Here, one crucial aspect is the formulation of the motion model. Following the concept of multi-sensor super-resolution, the motion associated with the $k$-th range image \smash{$\LRFrame{k}$} is first estimated by means of optical flow from the color image \smash{$\GuideFrame{k}$} and subsequently projected to the domain of the range data. Then, the motion on the range images encoded by dense displacement fields should ideally represent the motion that appears in the 3-D space. 

However, under a general type of camera motion, the actual motion that appears in range images cannot be described solely based on 2-D displacement fields. One prominent example is camera motion orthogonal to the measured surface and the sensor image plane, which is referred to as \textit{out-of-plane} motion. In this case, the motion in color images and hence the estimated displacements on range images provide only a 2-D view on the actual scene motion. For this reason, it does not explain the out-of-plane component appropriately. A similar situation appears if there is a tilting of a surface across two frames. Notice that related range super-resolution techniques \cite{Schuon2008, Schuon2009,Bhavsar2012} ignore this limitation of the motion model. However, neglecting this aspect is only reasonable for situations that allow an accurate description of the actual motion by 2-D displacement fields, whereas more general types of motion lead to a bias in super-resolution reconstruction \cite{Kohler2015a}. Therefore, we extend the image formation model to better explain the actual motion. 

\paragraph{Formulation of the Model.}
To enhance the modeling of scene motion and to take out-of-plane movements into account, a transformation of the measured range values is used in addition to 2-D displacement fields. For the derivation of the motion model, let \smash{$\LRFrame{r}$} be the reference range image and \smash{$\LRFrame{k}$} be the frame that needs to be explained under this model. The most simplest but non-trivial transformation between \smash{$\LRFrame{r}$} and \smash{$\LRFrame{k}$} as a better approximation to the actual 3-D motion is given by:
\begin{equation}
	\LRFrame{k} = \FrameIdx{\gamma_m}{k} \MotionOp{\LRFrame{r}} + \FrameIdx{\gamma_a}{k},
\end{equation}
where the motion operator $\MotionOp{\LRFrame{r}}$ describes the subpixel motion in the domain of the range data, and \smash{$\FrameIdx{\gamma_m}{k} \in \Real$ and $\FrameIdx{\gamma_a}{k} \in \Real$} are called the \textit{range correction} parameters. While the former describes motion on the image plane, the latter account for more general types of 3-D motion. The additive parameter \smash{$\FrameIdx{\gamma_a}{k}$} describes a global shift of the range values and can roughly explain out-of-plane motion. Similarly, the multiplicative parameter \smash{$\FrameIdx{\gamma_m}{k}$} describes a shearing of the range values. 

The range correction parameters are used to formulate the image formation model: 
\begin{equation}
	\label{eqn:rangeImageFormationModel}
	\LRFrame{k} = \FrameIdx{\gamma_m}{k} \SystemMatFrameGuide{k} \HR + \FrameIdx{\gamma_a}{k} + \FrameIdx{\NoiseVec}{k},
\end{equation}
which describes the formation of the $k$-th range image from the high-resolution range data $\HR$ according to the system matrix \smash{$\SystemMatFrameGuide{k}$} and the observation noise \smash{$\FrameIdx{\NoiseVec}{k}$}. The system matrix is defined by the motion information on the image plane given by displacement vector fields as well as the underlying sampling model.

\paragraph{Model Parameter Estimation.}
The range correction parameters need to be estimated from the low-resolution range images \smash{$\LRFrame{k}$}, $k \neq r$ relative to the reference frame \smash{$\LRFrame{r}$}. From a conceptual point of view, this is equivalent to a photometric registration of intensity images acquired under varying photometric conditions as shown in the work of Capel and Zisserman \cite{Capel2003}. Let $(y_i, \tilde{y}_i)$ be a pair of range values that are obtained from $\LRFrame{r}$ and $\FrameIdx{\Warp{\LR}}{k}$, where $\FrameIdx{\Warp{\LR}}{k}$ is the $k$-th frame $\LRFrame{k}$ warped towards the reference frame according to the estimated displacement fields. Then, the range correction parameters can be determined by pair-wise registration. The registration associated with the $k$-th frame is formulated as the line fitting problem:
\begin{equation}
	\label{eqn:rangeCorrectionObjective}
	(\hat{\gamma}_m, \hat{\gamma}_a) = \argmin_{\gamma_m, \gamma_a} \sum_{i = 1}^M \LossFun[\text{range}]{\tilde{y}_i - \gamma_m y_i - \gamma_a},
\end{equation}
where $\LossFun[\text{range}]{r}$ denotes a loss function applied to the residual errors for $M$ pairs of range values. A simulated example is depicted in \fref{fig:rangeCorrection}.

Estimating the range correction parameters is challenging due to random measurement noise, systematic errors in the range data, or outliers in optical flow. To deal with these issues, the range correction needs to be performed by robust parameter estimation and is applied on a median filtered version of original range values. In this work, \eref{eqn:rangeCorrectionObjective} is solved by probabilistic optimization using the \gls{msac} algorithm \cite{Torr2000}. The loss function that is used for \gls{msac} is given by the truncated least-squares term:
\begin{equation}
	\label{eqn:rangeCorrectionLossFunction}
	\LossFun[\text{range}]{r}  = \min \left( \delta_{\text{msac}}^2, r^2 \right).
\end{equation}
This loss function assigns a constant penalty $\delta_{\text{msac}}^2$ to residual errors that exceed the threshold $\delta_{\text{msac}}^2$, which are referred to as outliers. Residual errors that fall below this threshold are penalized quadratically and are considered as inliers. In this work, the threshold is adaptively set to $\delta_{\text{msac}} = 1.96 \NoiseStd$ to achieve a correct classification of $95\,\%$ of the true inliers under the assumption that the range values are affected by zero-mean Gaussian noise with standard deviation $\NoiseStd$. 

For a probabilistic optimization of \eref{eqn:rangeCorrectionObjective}, the initial parameter values for \gls{msac} are set to $\gamma_m = 1$ and $\gamma_a =  0$, and only parameter settings that result in lower objective values are accepted within estimation to avoid unreliable solutions. Then, at each iteration, two pairs of range values $(y^1, \tilde{y}^1)$ and $(y^2, \tilde{y}^2)$ are randomly drawn from the images $\LRFrame{r}$ and $\FrameIdx{\Warp{\LR}}{k}$. From these pairs, $(\gamma_m, \gamma_a)$ is computed in closed form. Accordingly, the pairs $(y_i, \tilde{y}_i)$ for $i = 1, \ldots, M$ are classified either as inliers or outliers in accordance to the objective value, which yields an inlier set associated with the current iteration. This procedure is repeated for $T_{\text{msac}}$ iterations to detect the optimal inlier set  $\mathcal{Y}_{\text{min}}$ that leads to a minimum objective value. Finally, the inlier set $\mathcal{Y}_{\text{min}}$ is used to gain $(\hat{\gamma}_m, \hat{\gamma}_a)$ by linear least-squares estimation, see \fref{fig:rangeCorrection}. \aref{alg:msac} summarizes the overall procedure to determine the range correction parameters for one pair of range images.

\begin{figure}[!t]
	\centering
	\scriptsize 
	\begin{minipage}[b]{0.36\textwidth}
		\includegraphics[width=0.82\textwidth]{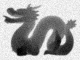}\\
		\includegraphics[width=0.82\textwidth]{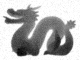}
	\end{minipage}\quad
	\setlength \figurewidth{0.36\textwidth}
	\setlength \figureheight{1.00\figurewidth} 
	\subfloat{\input{images/chapter5/rangeCorrection.tikz}}
	\caption[Range correction in presence of out-of-plane motion]{Range correction for a pair of range images in presence of out-of-plane motion. Left: Reference frame $\LRFrame{r}$ and $k$-th frame $\LRFrame{k}$, $k \neq r$ warped towards the reference according to the displacement field on the image plane. Right: Scatter plot of the range values $(y_i, \tilde{y}_i)$ drawn from $\LRFrame{r}$ and $\LRFrame{k}$ shown in blue. Inliers detected by \gls{msac} ($\NoiseStd = 0.025$, $T_{\text{msac}} = 200$) are marked in green and the fitted model is visualized by the black line.}
	\label{fig:rangeCorrection}
\end{figure}

\begin{algorithm}[!t]
	\small
	\caption{M-estimator sample consensus (MSAC) based range correction}
	\label{alg:msac}
	\begin{algorithmic}[1]
		\Require Pair of range images $\LRFrame{r}$ (reference) and $\LRFrame{k}$ (template)
		\Ensure Range correction parameters $\hat{\gamma}_m$ and $\hat{\gamma}_a$
		\State Determine $\FrameIdx{\Warp{\LR}}{k}$ by warping $\LRFrame{k}$ towards the reference $\LRFrame{r}$
		\State Initialize $\gamma_m \gets 1$, $\gamma_a \gets 0$, and $\phi_{\text{min}} \gets \sum_{i = 1}^M \phi(\tilde{y}_i - \gamma_m y_i - \gamma_a)$
		\For{$t = 1, \ldots, T_{\text{msac}}$}
			\State Draw randomly selected pairs $(y^1, \tilde{y}^1)$ and $(y^2, \tilde{y}^2)$ from $\LRFrame{r}$ and $\FrameIdx{\Warp{\LR}}{k}$
			\State Estimate range correction parameters $(\gamma_m, \gamma_a)$ from $(y^1, \tilde{y}^1)$ and $(y^2, \tilde{y}^2)$
			\State Initialize objective value $\IterationIdx{\phi}{t} \gets 0$ and inlier set $\IterationIdx{\mathcal{Y}}{t} \gets \{ \}$
			\For{$i = 1, \ldots, M$}
				\State Determine residual error $r_i = \tilde{y}_i - \gamma_m y_i - \gamma_a$
				\If{$r_i^2 < \delta_{\text{msac}}^2$}
					\State Update inlier set according to 
					$\IterationIdx{\mathcal{Y}}{t} \leftarrow \IterationIdx{\mathcal{Y}}{t} \cup \{ (y_i, \tilde{y}_i) \}$
				\EndIf
				\State Update objective value according to 
				$\IterationIdx{\phi}{t} \leftarrow \IterationIdx{\phi}{t} + \min(\delta_{\text{msac}}^2, r_i^2)$
			\EndFor
			\If{$\IterationIdx{\phi}{t} < \phi_{\text{min}}$}
				\State $\phi_{\text{min}} \gets \IterationIdx{\phi}{t}$ and $\mathcal{Y}_{\text{min}} \gets \IterationIdx{\mathcal{Y}}{t}$
			\EndIf
		\EndFor
		\State Determine $\hat{\gamma}_m$ and $\hat{\gamma}_a$ by least-squares estimation on the optimal inlier set $\mathcal{Y}_{\text{min}}$
	\end{algorithmic}
\end{algorithm}

\subsection{Range Super-Resolution Reconstruction}
\label{sec:05_ReconstructionAlgorithms}

Let us next adopt \aref{alg:msrTwoStageAlgorithm} and \aref{alg:msrOutlierDetectionAlgorithm} to the desired application. This requires custom observation and prior models.

The observation model for range data is described by a weighted normal distribution to account for space variant noise characteristics of current range sensors. Hence, we employ the confidence-aware data fidelity term:
\begin{equation}
	\DataTerm[MSR]{\HR} = \sum_{i = 1}^K 
	\left( \LRFrame{k} - \FrameIdx{\gamma_m}{k} \SystemMatFrameGuide{k} \HR - \FrameIdx{\gamma_a}{k} \right)^\top
	\FrameIdx{\WeightsBMat}{k}
	\left( \LRFrame{k} - \FrameIdx{\gamma_m}{k} \SystemMatFrameGuide{k} \HR - \FrameIdx{\gamma_a}{k} \right),
\end{equation}
which is derived from the image formation model in \eref{eqn:rangeImageFormationModel}. $\FrameIdx{\WeightsBMat}{k}$ denotes the joint confidence map of the $k$-th frame constructed from the range and color data. In the two-stage approach according to \aref{alg:msrTwoStageAlgorithm}, this confidence map is set to $\FrameIdx{\WeightsBMat}{k} = \Id$. For robust super-resolution according to \aref{alg:msrOutlierDetectionAlgorithm}, the confidence map is computed dynamically.

The image prior is formulated via piecewise smooth regularization to describe the appearance of smooth surfaces and depth discontinuities captured in range data. This type of regularization can be achieved by the Huber prior, see \sref{sec:03_MaximumAPosterioriEstimation}. Combined with the proposed spatially adaptive scheme, the regularization term for range super-resolution is defined as: 
\begin{equation}
	\label{eqn:huberAMSR}
	\RegTerm[MSR]{\HR, \Guide} 
	= \sum_{i = 1}^N \WeightsAFun_i(\HR, \Guide) \cdot  
	\left( \HuberThresh \sqrt{1 + \left( \frac{[\HPMat \HR]_i}{\HuberThresh} \right)^2} - \HuberThresh \right),
\end{equation}
where $\HuberThresh$ denotes the Huber threshold parameter. $\HPMat \in \RealMN{N}{N}$ denotes the filter kernel of a discrete Laplacian expressed as a circulant matrix to exploit the curvature of range data for regularization. We define this circulant matrix according to:
\begin{equation}
	\HPMat \HR \equiv 
	\frac{1}{4}
	\begin{pmatrix}
		0 &	1 & 0 \\
		1 & -4 & 1 \\
		0 &	1 & 0 \\
	\end{pmatrix} 
	\conv \vec{X},
\end{equation}
where $\vec{X} \in \RealMN{\LRDimU}{\LRDimV}$ is a representation of $\HR \in \RealN{\LRDimU \LRDimV}$ in matrix notation and $\conv$ denotes the discrete 2-D convolution.

\subsection{Experiments and Results}

Let us now present an experimental evaluation of the proposed multi-sensor super-resolution techniques in hybrid range imaging. The goal of this study is two-fold. On the one hand, we aim at comparing multi-sensor super-resolution to the conventional single-sensor approach as implemented by state-of-the-art algorithms \cite{Schuon2008,Schuon2009,Bhavsar2012}. On the other hand, the influence of the different multi-sensor techniques including motion estimation, spatially adaptive regularization and outlier detection to the performance of the proposed framework is studied.

In order to conduct a quantitative evaluation, we limited ourselves to experiments on artificial datasets with known ground truth. Experiments on real range data corrupted with systematic errors within the scope of a medical application are presented in \cref{sec:HybridRangeImagingForImageGuidedSurgery}. This simulation addresses the conditions of commercially available \gls{tof} sensors that are characterized by a low spatial resolution compared to color sensors. \Fref{fig:hybridRangeDataExample} depicts the \textit{Stanford Bunny} and the \textit{Dragon} scenes that were taken form the Stanford 3-D Scanning Repository\footnote{\url{http://graphics.stanford.edu/data/3Dscanrep}} for this study. Geometrically aligned range and color images were obtained from 3-D mesh representations of these scenes using the \gls{ritk} \cite{Wasza2011b}. The ground truth data was captured in a pixel resolution of 640$\times$480\,px and is available online\footnote{\url{https://www5.cs.fau.de/research/data/multi-sensor-super-resolution-datasets}}. For the simulation of a realistic color sensor, the color images were encoded in a pixel resolution of 640$\times$480\,px but blurred according to a Gaussian \gls{psf} ($\PSFWidth = 0.5$) and disturbed by zero-mean Gaussian noise ($\NoiseStd = 0.002$). The corresponding range images were simulated in a pixel resolution of 80$\times$60\,px with a Gaussian \gls{psf} ($\PSFWidth = 0.5$) and additive Gaussian noise ($\NoiseStd = 0.025$). This setup was used to generate four datasets showing the artificial scenes from different perspectives and with different textures. The displacements across the frames of these image sequences were related to rigid camera movements. 
\begin{figure}[!t]
	\centering
	\small
	\subfloat{\includegraphics[width=0.228\textwidth]{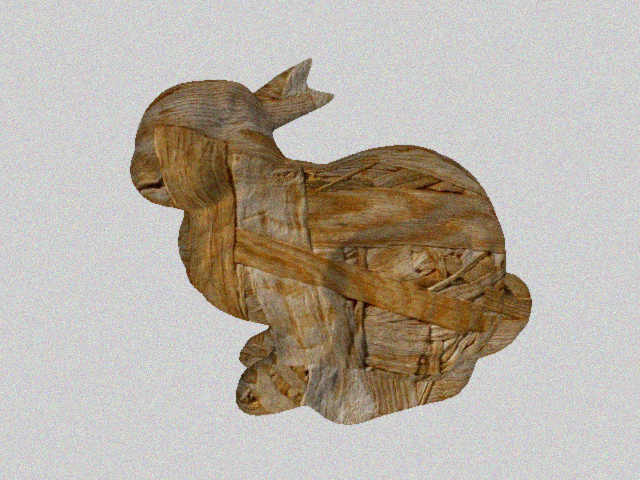}}~
	\subfloat{\includegraphics[width=0.228\textwidth]{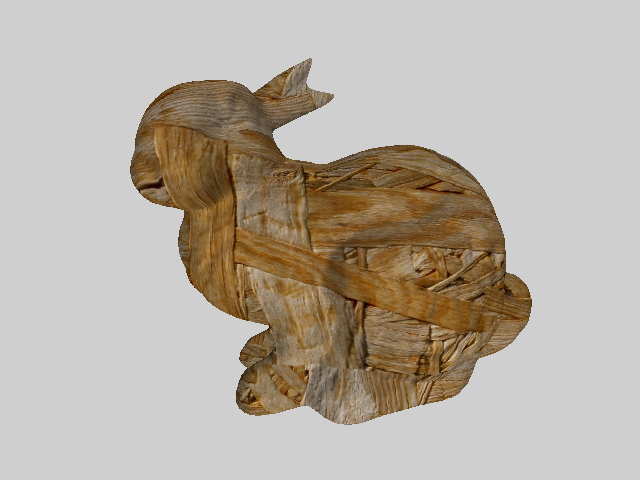}}\qquad
	\subfloat{\includegraphics[width=0.228\textwidth]{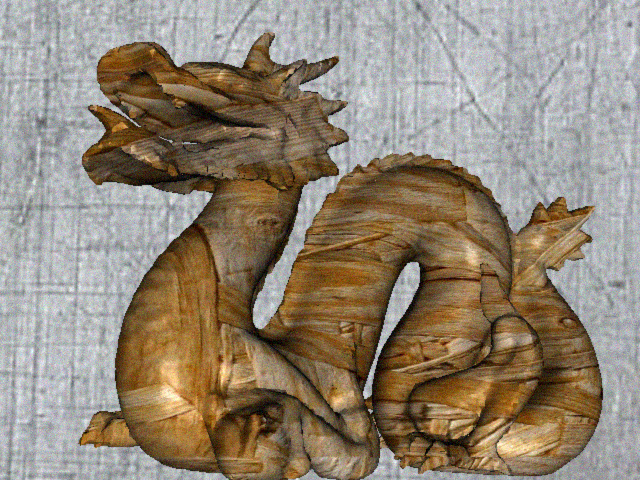}}~
	\subfloat{\includegraphics[width=0.228\textwidth]{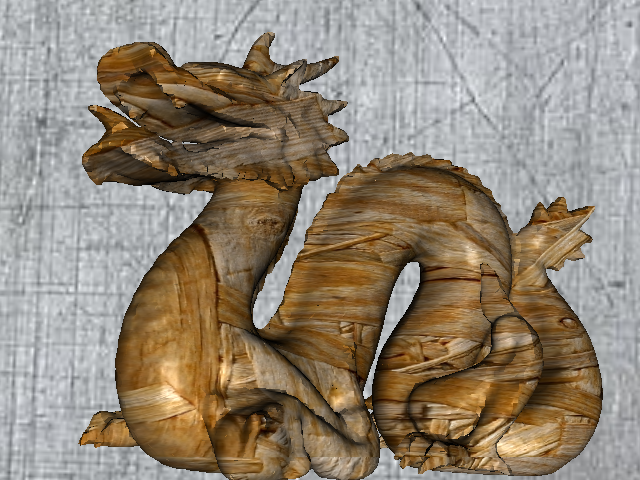}}\\
	\setcounter{subfigure}{0}
	\subfloat[Simulated data]{\includegraphics[width=0.228\textwidth]{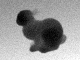}}~
	\subfloat[Ground truth]{\includegraphics[width=0.228\textwidth]{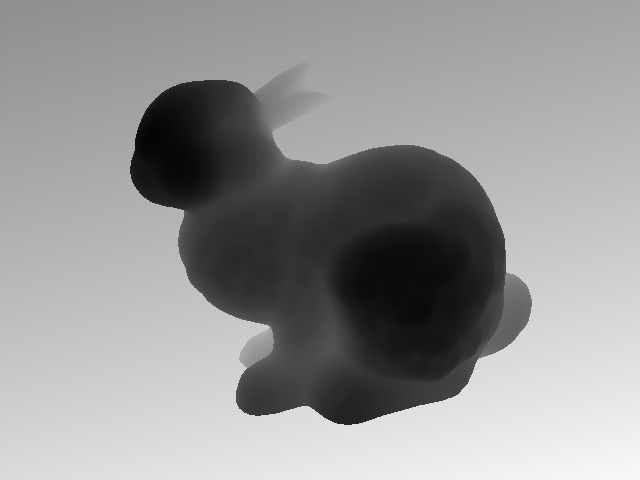}}\qquad
	\subfloat[Simulated data]{\includegraphics[width=0.228\textwidth]{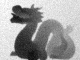}}~
	\subfloat[Ground truth]{\includegraphics[width=0.228\textwidth]{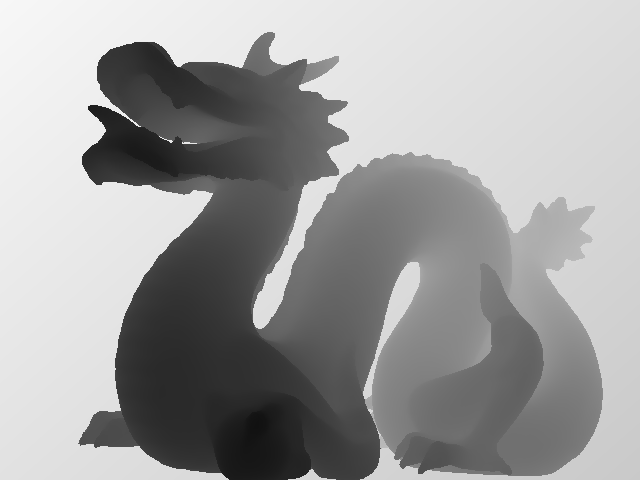}}
	\caption[Range and color data for the \textit{Bunny} and the \textit{Dragon} scenes]{Simulated range and color data along with the ground truth data obtained from the artificial \textit{Stanford Bunny} and the \textit{Dragon} scenes.}
	\label{fig:hybridRangeDataExample}
\end{figure}

In the following experiments, the algorithms summarized in \tref{tab:msrReconstructionAlgos} along with their model parameters were analyzed. All reconstruction algorithms are based on a Huber prior with $\HuberThresh = 5 \cdot 10^{-4}$ and $\RegWeight = 0.08$. To evaluate the impact of the techniques proposed in this chapter, the algorithms differ in the way of how color images are exploited. \Gls{ssr} that works solely on the range data is considered as the baseline. \Gls{msr} utilizes color images in \aref{alg:msrTwoStageAlgorithm} for filter-based motion estimation using uniform weights for regularization. The \gls{amsr} uses the color images also for spatially adaptive regularization. \Gls{amsrod} augments \gls{amsr} with the outlier detection scheme in \aref{alg:msrOutlierDetectionAlgorithm}. 

\paragraph{Direct vs. Filter-Based Motion Estimation.}
Let us first present a comparison of the different strategies for motion estimation as a prerequisite for super-resolution. In these experiments, the computation of displacement fields was performed by the variational optical flow algorithm introduced by Liu \cite{Liu2009}. In the single-sensor approach, optical flow was obtained directly on the range images, whereas the multi-sensor approaches used the proposed filter-based technique on color images. A qualitative comparison among these strategies is shown in \fref{fig:displacementFieldFilteringExample}. While direct motion estimation (\fref{fig:displacementFieldFilteringExample:rangeData}) was error prone and resulted in noisy displacement fields, the filter-based technique (\fref{fig:displacementFieldFilteringExample:filtered}) accurately recorded camera motion. As shown in the experiments reported below, this substantially affects the accuracy of super-resolved range information. 

In the context of motion estimation, the proposed outlier detection that is driven by color images provides a confidence map associated with the estimated displacement fields (\fref{fig:displacementFieldFilteringExample:confidenceMap}). This can further enhance the robustness of super-resolution compared to a reconstruction without proper outlier detection.
\begin{figure}[!t]
	\centering
	\subfloat[Direct]{\includegraphics[width=0.313\textwidth]{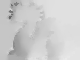} 
	\label{fig:displacementFieldFilteringExample:rangeData}}~
	\subfloat[Filter-based]{\includegraphics[width=0.313\textwidth]{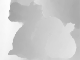}
	\label{fig:displacementFieldFilteringExample:filtered}}~
	\subfloat[Confidence map]{\includegraphics[width=0.313\textwidth]{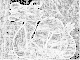}
	\label{fig:displacementFieldFilteringExample:confidenceMap}}
	\caption[Comparison of motion estimation strategies for range images]{Comparison of motion estimation strategies to obtain displacement fields on range images. \protect\subref{fig:displacementFieldFilteringExample:rangeData} Displacement field magnitudes obtained by the \textit{direct} approach using optical flow estimation on range images. \protect\subref{fig:displacementFieldFilteringExample:filtered} Displacement field magnitudes obtained by the proposed filter-based approach that exploits color images as guidance (bright regions denote higher magnitudes). \protect\subref{fig:displacementFieldFilteringExample:confidenceMap} Confidence map computed on the color data for the displacement field in \protect\subref{fig:displacementFieldFilteringExample:filtered} (bright regions denote higher weights).}
	\label{fig:displacementFieldFilteringExample}
\end{figure}

\paragraph{Single-Sensor vs. Multi-Sensor Super-Resolution.}
In \fref{fig:simulatedDatasetResults}, we compare the different reconstruction algorithms on the \textit{Dragon-1} dataset using $K = 25$ frames and a magnification factor $\MagFac = 4$. In comparison to \gls{ssr} that is affected by inaccurate motion estimation, the different multi-sensor approaches (\gls{msr}, \gls{amsr}, \gls{amsrod}) improved the accuracy of range information. More specifically, the reconstruction algorithms that implement spatially adaptive regularization enhanced the reconstruction of depth discontinuities compared to the non-adaptive methods. 

These properties are confirmed by a quantitative comparison based on the \gls{psnr} and \gls{ssim} measures of super-resolved range data relative to the ground truth. \Fref{fig:simulatedDatasetBoxplots} summarizes the statistics of both measures on four datasets, where each dataset comprises 15 randomly generated image sequences. This reveals that the different multi-sensor approaches outperformed the single-sensor approach in terms of both measures. Among the different multi-sensor algorithms, the highest accuracy was obtained by \gls{amsrod} that uses color images for motion estimation, spatially adaptive regularization and outlier detection. In comparison to the \gls{ssr} reconstruction, \gls{amsrod} improved the mean \gls{psnr} and \gls{ssim} by $0.9$\,\gls{db} and $0.02$, respectively.

\begin{table}[!t]
  \centering
  \caption[Super-resolution reconstruction approaches for hybrid range imaging along with their parameters]{Overview of the reconstruction approaches that are examined for their use in hybrid range imaging along with their optimization parameters. The multi-sensor framework is employed in different versions (\gls{msr}, \gls{amsr}, \gls{amsrod}) using filter-based motion estimation, spatially adaptive regularization, and outlier detection for range data based on color images. The single-sensor algorithm (\gls{ssr}) that is not guided by color images is considered as the baseline.}
	\small 
	\begin{tabular}{l ccc}
		\toprule
		\textbf{Reconstruction algorithm} & \multicolumn{3}{c}{\textbf{Algorithm properties}} \\
		\cmidrule{2-4}
		& \textbf{Motion} & \textbf{Adaptive} & \textbf{Outlier} \\
		& \textbf{estimation} & \textbf{regularization} & \textbf{detection} \\
		\midrule
		Single-sensor super-resolution & direct & \xmark & \xmark \\
		(SSR) & & & \\
		\midrule
		Multi-sensor super-resolution & filter-based & \xmark & \xmark \\
		(MSR) & & & \\
		\midrule
		Adaptive multi-sensor super-resolution & filter-based & \cmark & \xmark \\
		(AMSR) & & $\AMSRContrastFactor = 0.06$, $\AMSRPatchSize = 5$ & \\
		\midrule
		Adaptive multi-sensor super-resolution & filter-based & \cmark & \cmark \\
		with outlier detection (AMSR-OD) & & $\AMSRContrastFactor = 0.06$, $\AMSRPatchSize = 5$ & $\rho_{0} = 0.5$ \\
		\bottomrule
	\end{tabular}
	\label{tab:msrReconstructionAlgos}
\end{table}
\begin{figure}[!t]
	\centering
	\small
	\subfloat[Original]{	
		\begin{tikzpicture}[spy using outlines={rectangle, green, magnification=2.0, height=1.3cm, width=1.3cm, connect spies, every spy on node/.append style={thick}}] 
			\node {\pgfimage[width=0.31\linewidth]{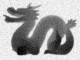}}; 
			\spy on (-1.9, 0.9) in node [left] at (2.35, 1.15);
		\end{tikzpicture}
	}
	\subfloat[\gls{ssr}]{	
		\begin{tikzpicture}[spy using outlines={rectangle, green, magnification=2.0, height=1.3cm, width=1.3cm, connect spies, every spy on node/.append style={thick}}] 
			\node {\pgfimage[width=0.31\linewidth]{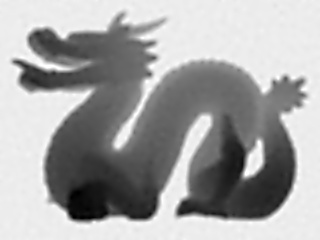}}; 
			\spy on (-1.9, 0.9) in node [left] at (2.35, 1.15);
		\end{tikzpicture}
	}
	\subfloat[\gls{msr}]{	
		\begin{tikzpicture}[spy using outlines={rectangle, green, magnification=2.0, height=1.3cm, width=1.3cm, connect spies, every spy on node/.append style={thick}}] 
			\node {\pgfimage[width=0.31\linewidth]{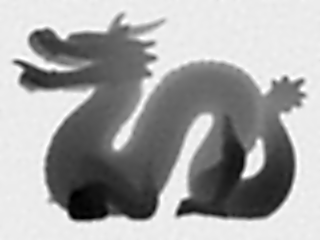}}; 
			\spy on (-1.9, 0.9) in node [left] at (2.35, 1.15);
		\end{tikzpicture}
	}\\
	\subfloat[\gls{amsr}]{	
		\begin{tikzpicture}[spy using outlines={rectangle, green, magnification=2.0, height=1.3cm, width=1.3cm, connect spies, every spy on node/.append style={thick}}] 
			\node {\pgfimage[width=0.31\linewidth]{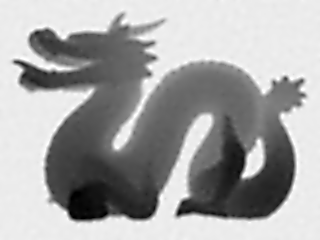}}; 
			\spy on (-1.9, 0.9) in node [left] at (2.35, 1.15);
		\end{tikzpicture}
	}
	\subfloat[\gls{amsrod}]{	
		\begin{tikzpicture}[spy using outlines={rectangle, green, magnification=2.0, height=1.3cm, width=1.3cm, connect spies, every spy on node/.append style={thick}}] 
			\node {\pgfimage[width=0.31\linewidth]{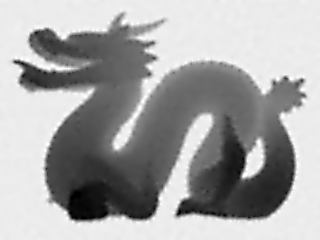}}; 
			\spy on (-1.9, 0.9) in node [left] at (2.35, 1.15);
		\end{tikzpicture}
	}
	\subfloat[Ground truth]{	
		\begin{tikzpicture}[spy using outlines={rectangle, green, magnification=2.0, height=1.3cm, width=1.3cm, connect spies, every spy on node/.append style={thick}}] 
			\node {\pgfimage[width=0.31\linewidth]{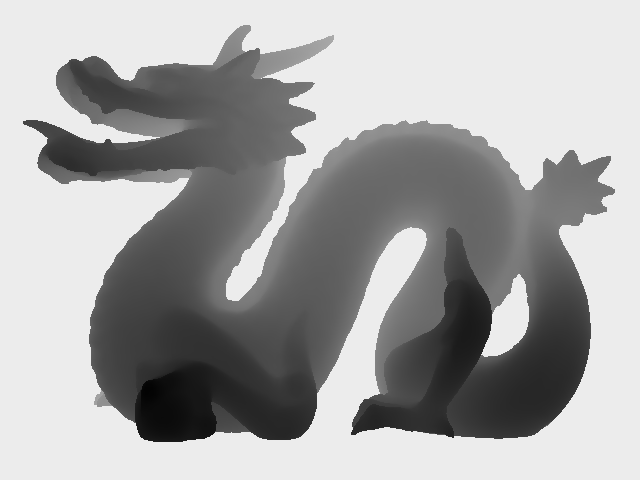}}; 
			\spy on (-1.9, 0.9) in node [left] at (2.35, 1.15);
		\end{tikzpicture}
	}
	\caption[Comparison of super-resolution approaches on the \textit{Dragon} dataset]{Comparison of original, low-resolution range data, \acrfull{ssr} and the different multi-sensor approaches (\gls{msr}, \gls{amsr} and \gls{amsrod}) to the ground truth range data on the \textit{Dragon-1} dataset. The reconstructions were obtained from $K = 25$ frames with magnification factor $\MagFac = 4$.}
	\label{fig:simulatedDatasetResults}
\end{figure}
\begin{figure}[!t]
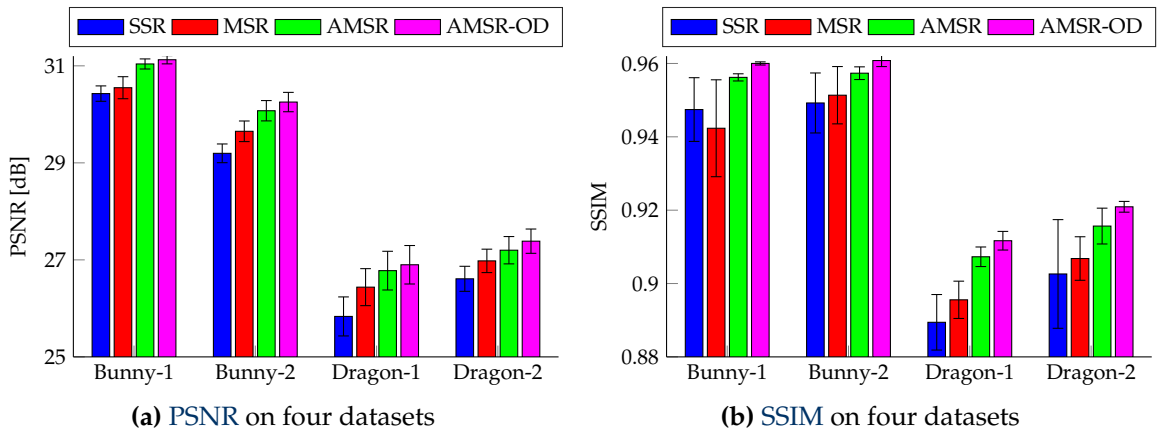

	\centering
	\scriptsize 
	\setlength \figurewidth{0.42\textwidth}
	\setlength \figureheight{0.65\figurewidth} 
	\subfloat[\gls{psnr} on four datasets]{\input{images/chapter5/psnr_barPlot.tikz}}\quad
	\subfloat[\gls{ssim} on four datasets]{\input{images/chapter5/ssim_barPlot.tikz}}
	\caption[\gls{psnr} and \gls{ssim} on the \textit{Bunny} and the \textit{Dragon} datasets]{Mean $\pm$ standard deviation of the \gls{psnr} and \gls{ssim} measures on four datasets obtained from the artificial \textit{Stanford Bunny} and \textit{Dragon} scenes. Both measures were evaluated for 15 randomly generated image sequences per dataset.}
	\label{fig:simulatedDatasetBoxplots}
\end{figure}

\paragraph{Influence of the Model Parameters.}
\Fref{fig:regularizationWeight} reports the behavior of the competing methods on the \textit{Dragon-2} dataset regarding the choice of the regularization weight $\RegWeight$ on a logarithmic scaled axis. It is worth noting that the different multi-sensor algorithms considerably outperformed the single-sensor algorithm regardless of the choice of the regularization weight over several orders of magnitude ($-2 \leq \log \RegWeight \leq 0.5$). In the case of an overestimation of this parameter ($\log \RegWeight \geq 0.5$), which resulted in oversmoothing of the range data, the different approaches showed a similar behavior. 
\begin{figure}[!t]
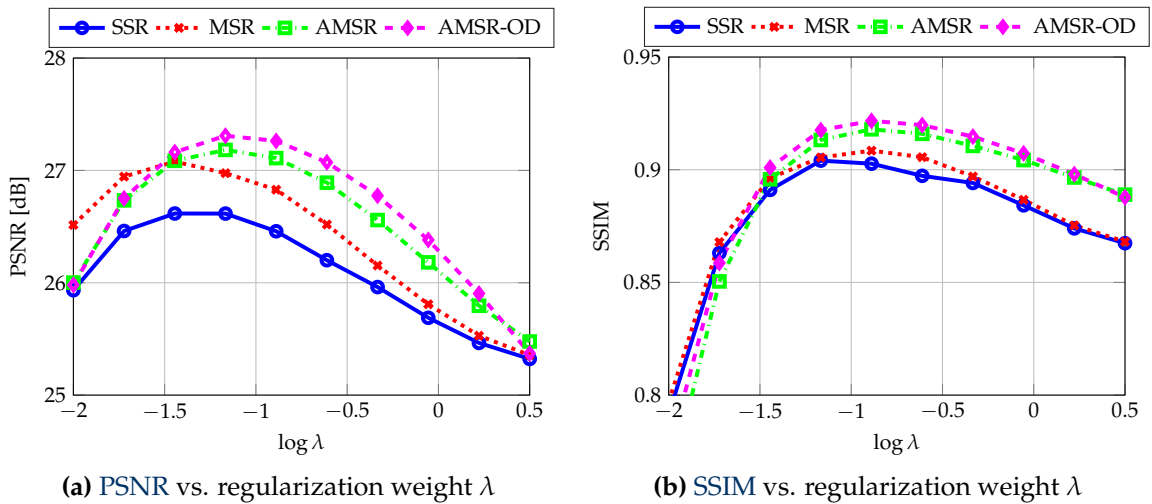

	\centering
	\scriptsize
	\setlength \figurewidth{0.395\textwidth}
	\setlength \figureheight{0.392\textwidth} 
	\subfloat[\gls{psnr} vs. regularization weight $\RegWeight$]{\input{images/chapter5/regularizationWeight_psnr.tikz}}\quad
	\setlength \figurewidth{0.395\textwidth}
	\setlength \figureheight{0.37\textwidth}
	\subfloat[\gls{ssim} vs. regularization weight $\RegWeight$]{\input{images/chapter5/regularizationWeight_ssim.tikz}}
	\caption[Parameter sensitivity study for the regularization weight $\RegWeight$]{Parameter sensitivity study for the regularization weight $\RegWeight$. The \gls{psnr} and \gls{ssim} measures at each parameter setting are averaged over 15 random realizations of the experiment on the \textit{Dragon-2} dataset. The classical single-sensor approach (\gls{ssr}) is compared to the different proposed multi-sensor approaches (\gls{msr}, \gls{amsr}, \gls{amsrod}).}
	\label{fig:regularizationWeight}
\end{figure}

Another relevant model parameter is the contrast factor $\AMSRContrastFactor$ that is used for spatially adaptive regularization. \Fref{fig:contrastFactor} depicts the influence of this parameter on the \textit{Dragon-2} dataset. If the contrast factor was overestimated ($\log \AMSRContrastFactor \geq -1.0$), one can observe that the adaptive approaches (\gls{amsr} and \gls{amsrod}) behave like the non-adaptive algorithms (\gls{ssr} and \gls{msr}) that can be considered as the baseline. In case of an underestimation ($\log \AMSRContrastFactor \leq -1.5$), the adaptive approaches were prone to texture copying artifacts as texture in the color images was erroneously transferred to the super-resolved range data. This behavior was captured quantitatively by the \gls{ssim}. Notice that this texture copying is comparable to related color-guided range filtering and upsampling techniques \cite{Park2011,Kiechle2013,Ferstl2013} but is controllable by limiting the contrast factor to a reasonable range ($-1.5 \leq \log \AMSRContrastFactor \leq -1.0$).
\begin{figure}[!t]
	\centering
	\scriptsize
	\setlength \figurewidth{0.395\textwidth}
	\setlength \figureheight{0.37\textwidth} 
	\subfloat[\gls{psnr} vs. contrast factor $\AMSRContrastFactor$]{\input{images/chapter5/contrastFactor_psnr.tikz}}\quad
	\subfloat[\gls{ssim} vs. contrast factor $\AMSRContrastFactor$]{\input{images/chapter5/contrastFactor_ssim.tikz}}\\
	\subfloat[$\log \AMSRContrastFactor = -2.0$]{
		\begin{tikzpicture}[spy using outlines={rectangle, green, magnification=2.0, height=1.3cm, width=1.3cm, connect spies, every spy on node/.append style={thick}}] 
			\node {\pgfimage[width=0.31\linewidth]{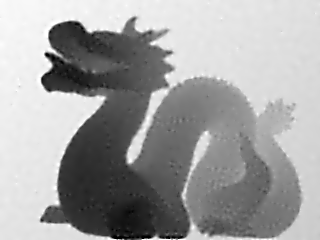}}; 
			\spy on (-1.4, 1.15) in node [left] at (2.35, 1.15);
		\end{tikzpicture}
	}
	\subfloat[$\log \AMSRContrastFactor = -1.3$]{
		\begin{tikzpicture}[spy using outlines={rectangle, green, magnification=2.0, height=1.3cm, width=1.3cm, connect spies, every spy on node/.append style={thick}}] 
			\node {\pgfimage[width=0.31\linewidth]{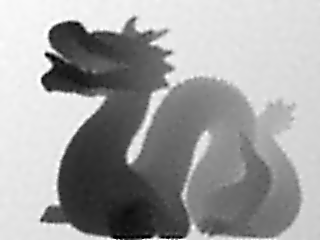}}; 
			\spy on (-1.4, 1.15) in node [left] at (2.35, 1.15);
		\end{tikzpicture}
	}
	\subfloat[$\log \AMSRContrastFactor = -0.5$]{
		\begin{tikzpicture}[spy using outlines={rectangle, green, magnification=2.0, height=1.3cm, width=1.3cm, connect spies, every spy on node/.append style={thick}}] 
			\node {\pgfimage[width=0.31\linewidth]{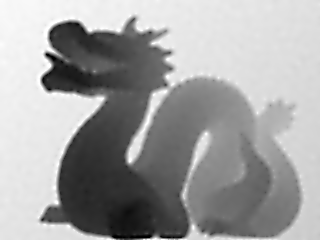}}; 
			\spy on (-1.4, 1.15) in node [left] at (2.35, 1.15);
		\end{tikzpicture}
	}
	\caption[Parameter sensitivity study for the contrast factor $\AMSRContrastFactor$]{Parameter sensitivity study for the contrast factor $\AMSRContrastFactor$ of spatially adaptive regularization. Top row: \gls{psnr} and \gls{ssim} for different $\AMSRContrastFactor$. Both measures are averaged over 15 random realizations of the experiment on the \textit{Dragon-2} dataset. The non-adaptive algorithms (\gls{ssr} and \gls{msr}) are the baseline and are compared to the adaptive algorithms (\gls{amsr} and \gls{amsrod}). Bottom row: \gls{amsrod} for an underestimated, an optimal, and an overestimated $\AMSRContrastFactor$, respectively. Note the texture copying artifacts in case of a too low $\AMSRContrastFactor$ ($\log \AMSRContrastFactor = -2.0$). For an appropriate parameter setting ($\log \AMSRContrastFactor = -1.3$), spatially adaptive regularization shows good tradeoffs between unwanted texture copying and the reconstruction of depth discontinuities.}
	\label{fig:contrastFactor}
\end{figure}

\section{Conclusion}
\label{sec:05_Conclusion}

This chapter studied super-resolution for a single modality under the guidance of a complementary modality. In contrast to the algorithms presented in the first part of this thesis, the proposed multi-sensor framework takes advantage of additional guidance data. This aims at enhancing accuracy and robustness of super-resolution reconstruction. The computational stages that are steered by guidance images include motion estimation, spatially adaptive regularization as well as outlier detection. These concepts yield two algorithms: In the two-stage algorithm, a filter-based technique to obtain displacement fields in the domain of low-resolution data from guidance images and spatially adaptive regularization steered by the guidance images is utilized for image reconstruction. In the iteratively re-weighted minimization algorithm, confidence maps for outlier detection are constructed by exploiting guidance data.

In order to prove the benefit of multi-sensor super-resolution over the single-sensor counterpart, hybrid range imaging was considered as example application. The goal was to super-resolve range data acquired with low-cost sensors under the guidance of color images fused with the range data. Multi-sensor super-resolution was tailored to range imaging by extending the underlying image formation model. We demonstrated in a simulation study that the proposed multi-sensor approach is able to take advantage of color images. The combination of the multi-sensor techniques provided superior surface reconstructions compared to the single-sensor approach and led to improvements of the \gls{psnr} and \gls{ssim} of 0.9\,\gls{db} and 0.02, respectively.


\chapter{Multi-Sensor Super-Resolution using Locally Linear Regression}
\label{sec:SuperResolutionForMultiChannelImages}

\myminitoc

\noindent
This chapter introduces a generalization of multi-sensor super-resolution that is applicable for an arbitrary number of modalities in hybrid imaging but does not require reliable guidance information. For this problem formulation, the modalities are represented by \textit{multi-channel} images. This framework builds on a Bayesian model that features a novel image prior to exploit sparsity of the channels in a transform domain as well as locally linear regressions across them. Super-resolution is then derived via joint estimation of high-resolution image channels along with latent hyperparameters of the Bayesian model. In order to solve this non-convex optimization problem efficiently, an alternating minimization algorithm is developed. The proposed methodology is validated for resolution enhancement in various applications in computer vision, including color- and multispectral imaging as well as 3-D range imaging.

Parts of this chapter have been originally published in \cite{Ghesu2014} and \cite{Kohler2015b}.

\section{Introduction}

In \cref{sec:MultiSensorSuperResolutionUsingGuidanceImages}, a first multi-sensor super-resolution method for hybrid imaging has been presented. In principle, this concept relies on two fundamental prerequisites. First and foremost, the underlying reconstruction algorithm super-resolves only a single modality. In addition, it relies on the existence of a reliable guidance modality that is used to steer super-resolution. Despite the advantages over a single-sensor formulation, these requirements limit the use of this approach for many target applications. Some prominent examples are color or multispectral imaging, where current systems feature the acquisition of multiple spectral bands ranging from three to several hundreds. Since it is inadequate to super-resolve single spectral bands only, this initiated the development of color \cite{Gotoh2004,Farsiu2006} and multispectral \cite{Akgun2005,Aguena2006,Zhang2012b} super-resolution techniques. The second shortcoming is that the required guidance data might not be available or it requires unjustifiable efforts to provide them. This situation appears in range imaging with devices that lack reliable high-resolution color sensors to gain guidance data \cite{Ghesu2014}. This is connected with the feature extraction from the guidance, \eg in terms of spatially adaptive regularization \cite{Kohler2015a} or outlier detection \cite{Kohler2014a}, that deteriorates in case of insufficient image quality.

\begin{figure}[!t]
	\centering
	\subfloat[Guidance image framework (\cref{sec:MultiSensorSuperResolutionUsingGuidanceImages})]{\includegraphics[width=0.465\textwidth]{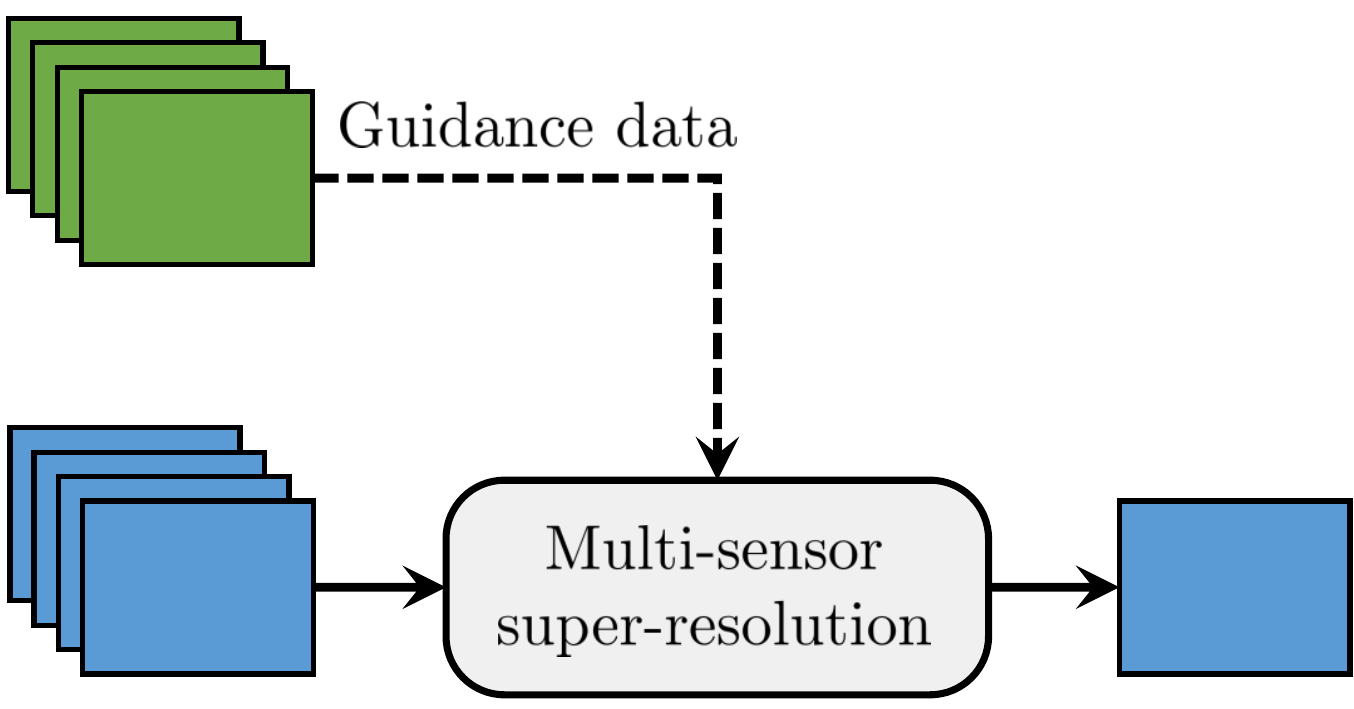} 
	\label{fig:06_multisensorIllustration:guidance}}\qquad
	\subfloat[Multi-channel framework (\cref{sec:SuperResolutionForMultiChannelImages})]{\includegraphics[width=0.465\textwidth]{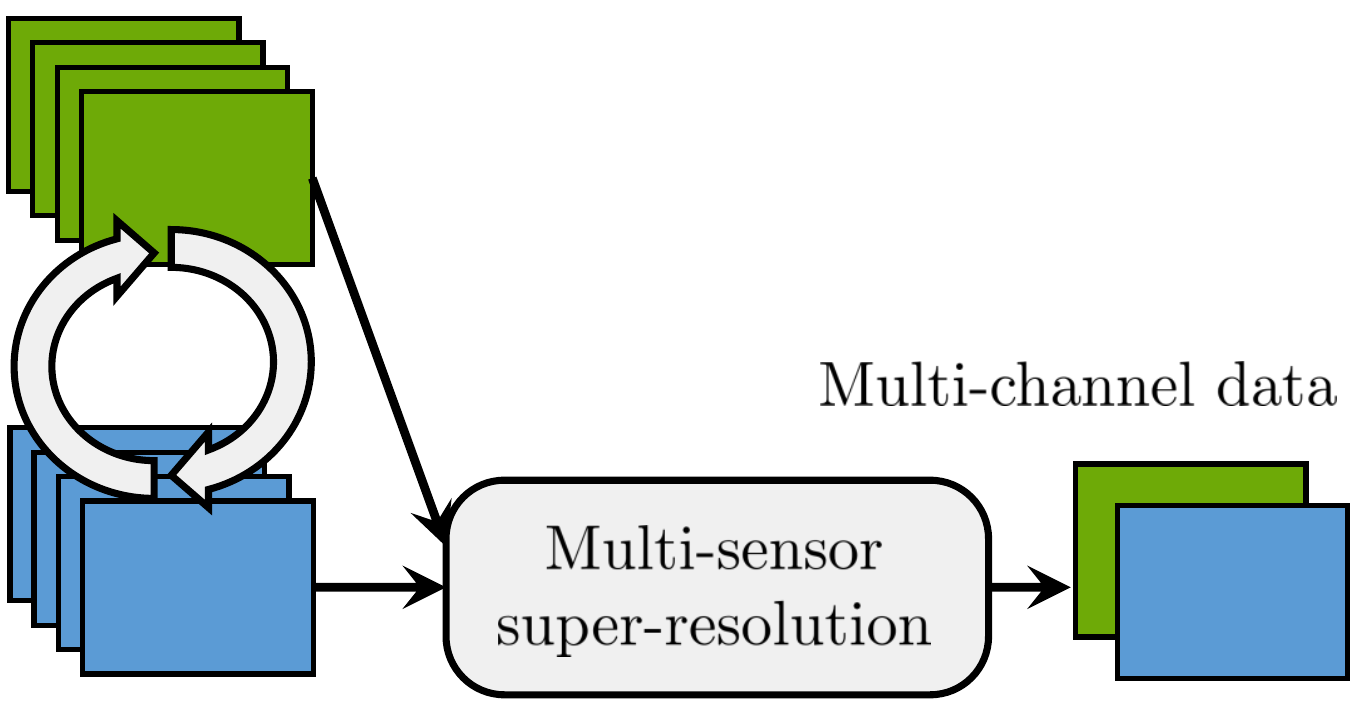}
	\label{fig:06_multisensorIllustration:multichannel}}
	\caption[Comparison of multi-sensor super-resolution approaches]{Comparison of the different approaches to multi-sensor super-resolution. The guidance image framework in \protect\subref{fig:06_multisensorIllustration:guidance} uses static guidance information of a single modality to steer super-resolution for another modality. The multi-channel framework in \protect\subref{fig:06_multisensorIllustration:multichannel} exploits dependencies among $\NumChannels \geq 2$ modalities (image channels) to jointly super-resolve them.}
	\label{fig:06_multisensorIllustration}
\end{figure}

In this chapter, we sacrifice the concept of guidance images to circumvent the aforementioned limitations and approach multi-sensor super-resolution from a generalized perspective. For this purpose, a set of modalities is represented by \textit{multi-channel} images in the underlying mathematical framework. In contrast to processing the individual channels one after another, super-resolution is performed jointly for the entire set of channels. The basic assumption of this approach is the existence of \textit{inter-channel dependencies}, such as geometrical structures that are visible in multiple channels. A well known example are color images that exhibit a high degree of correlation among their spectral bands as widely studied in color image processing \cite{Galatsanos1991,Katsaggelos1993a,Schultz1995}. In the context of multi-sensor super-resolution, inter-channel dependencies can be exploited in a Bayesian formulation as prior knowledge for the reconstruction of high-resolution multi-channel images. We capture these dependencies by a novel \gls{llr} image prior that is flexible with regard to the number of channels and does not rely on additional guidance information. This prior steers super-resolution \textit{dynamically} as opposed to the static, feature-based techniques in \cref{sec:MultiSensorSuperResolutionUsingGuidanceImages}. Based on this Bayesian model, we derive the simultaneous estimation of the unknown high-resolution channels along with latent prior hyperparameters as a joint energy minimization problem that is solved by confidence-aware optimization. In \fref{fig:06_multisensorIllustration}, we illustrate this methodology in comparison to the guidance image framework. 

The remainder of this chapter is structured as follows. \sref{sec:06_RelatedWork} provides a literature review regarding related methods. \sref{sec:06_BayesianModelingOfMultiChannelImages} introduces a Bayesian model of multi-channel images that is used in \sref{sec:06_BayesianApproachToMultiChannelSuperResolution} to formulate multi-sensor super-resolution via joint energy minimization. \sref{sec:06_AlgorithmAnalysis} presents an in-depth analysis of this model and theoretical comparisons to related methods. In \sref{sec:06_ExperimentsAndResults}, we report an experimental evaluation by studying multiple target applications including color, multispectral and range imaging along with comparisons to the state-of-the-art in these domains. Finally, \sref{sec:06_Conclusion} draws a conclusion.

\section{Related Work}
\label{sec:06_RelatedWork}

The proposed approach to multi-sensor super-resolution exploits mutual dependencies among image channels to jointly super-resolve them. In particular, as shown in this chapter, we are interested in modeling statistical dependencies across the channels as a prior distribution for Bayesian estimation. Below, we provide a survey on similar concepts that have been successfully applied in related areas.

\paragraph{Color Imaging.}
In color imaging, dependencies among spectral bands have been widely investigated. Here, most commercially available cameras acquire red (R), green (G) and blue (B) spectral bands that form the RGB space. However, due to economic reasons, the sensor array is usually equipped with a \gls{cfa} and is made sensitive to a single color per pixel. The interpolation of full RGB measurements referred to as \textit{demosaicing} \cite{Kimmel1999} can be considered as some sort of super-resolution. To avoid inconsistencies among interpolated color channels, inter-channel dependencies are exploited for demosaicing, which can be done via color ratios \cite{Kimmel1999} or color correlation terms \cite{Keren1999}.

Later, correlations among color bands have been studied for various tasks such as denoising \cite{Keren1998}, deconvolution \cite{Molina2003,Vega2006}, and sparse representation of color images \cite{Mairal2008}, among others. In the area of image restoration, Ono and Yamada \cite{Ono2016} have proposed local color nuclear norm regularization based on the color-line property \cite{Omer2004}, which states that color bands in local image regions are linearly dependent. In \cite{Fattal2015}, Fattal employed this property for color image dehazing. Moreover, demosaicing of color images has been augmented with the notion of multi-frame super-resolution as proposed in the work of Gotoh and Okutomi \cite{Gotoh2004}. This approach considers dependencies between color channels by a transformation of the highly correlated RGB space into luminance and chrominance components that are modeled by different prior distributions. In \cite{Farsiu2006}, Farsiu \etal proposed a related method based on an inter-channel regularization in the RGB space. This regularization enforces consistency in terms of locations and orientations of edges captured in the different channels. Such techniques avoid color artifacts, \eg color bleeding, and serve as a strong prior for image super-resolution.

\paragraph{Multi- and Hyperspectral Imaging.}
In the area of multi- and hyperspectral image processing, different attempts have been made at extending color image restoration to a larger number of spectral bands. In \cite{Akgun2005}, Akgun \etal proposed a generalized image formation model for hyperspectral images. Similar to color image restoration, such methods benefit from incorporating dependencies among the spectral bands to their underlying model. In the method of Zhang \etal \cite{Zhang2012b}, statistical dependencies are considered by applying a \gls{pca} on the original spectral bands. Then, super-resolution is performed on \gls{pca} compressed hyperspectral data. A different notion is to employ correlations with a high spatial but a low spectral resolution image \cite{Aguena2006,Akhtar2015,Lanaras2015} to steer hyperspectral super-resolution. Notice that this concept is closely related to guidance image based super-resolution in \cref{sec:MultiSensorSuperResolutionUsingGuidanceImages}.

\paragraph{Joint and Mutual Structure Filtering.}
\textit{Joint} image filters process a single input image driven by a guidance image. In this area, the guidance is assumed to be static following the same line of thought as guidance image based super-resolution. Local filters related to this concept include joint bilateral \cite{Kopf2007}, guided \cite{He2013,Horentrup2014} or weighted median filtering \cite{Ma2013,Zhang2014}, see \sref{sec:05_RelatedWork}. Such filters can also be learned from example data using convolutional neural networks \cite{Li2016}. Global filters that are formulated via regularized energy minimization have been developed for range image upsampling \cite{Park2011,Ferstl2013,Kiechle2013} as well as cross-field image restoration \cite{Yan2013}. These approaches impose properties of the filter output based on the given guidance image using implicit regularization terms. Contrary to the method proposed in this chapter, joint filtering has the common prerequisite of high-quality guidance data similar to guided image based super-resolution. Another limitation is that joint filtering is prone to erroneously transfer image structures to the filter output that are only present in the guidance image.

In contrast to the aforementioned techniques, Shen \etal \cite{Shen2015} introduced \textit{mutual structure} filtering to simultaneously filter input and guidance data with consideration of structural inconsistencies. This inconsistency-aware and dynamic formulation alleviates the erroneous structure transfer compared to filters with pure static guidance. In \cite{Ham2015}, Ham \etal proposed with a similar motivation static and dynamic guided filtering that combines regularization gained by static guidance data and the filter input. Although these methods improve flexibility and robustness of joint filtering, they ignore reasonable models of the image formation process and address denoising problems rather than super-resolution. Contrary to the proposed super-resolution approach, these filters are developed as single-image methods without considering multi-frame processing.

\section{Bayesian Modeling of Multi-Channel Images}
\label{sec:06_BayesianModelingOfMultiChannelImages}

In this chapter, an unknown, high-resolution multi-channel image is represented as the composite of $\NumChannels$ disjoint channels denoted by $\HR = (\HRChannel{1}^\top, \ldots, \HRChannel{\NumChannels}^\top)^\top$\label{notation:hrChannelComposite}, where each channel $\HRChannel{i}$\label{notation:hrChannel}, $i = 1, \ldots, \NumChannels$\label{notation:numChannels} is represented by a $\ChannelIdx{\HRSize}{i} \times 1$\label{notation:hrChannelSize} vector. For the sake of notational brevity, we limit ourselves to channels with consistent dimensions, i.\,e. $\ChannelIdx{\HRSize}{i} = \HRSize$ for all $i = 1, \ldots, \NumChannels$. Each high-resolution channel $\HRChannel{i}$ is related to a sequence of $\NumFrames$ low-resolution frames \smash{$\LRChannel{i} = (\LRChannelFrame{i}{1}, \ldots, \LRChannelFrame{i}{\NumFrames})^\top$}\label{notation:lrChannel}. Here, \smash{$\LRChannelFrame{i}{k}$} is the $k$-th frame associated with the $i$-th channel and is represented by a $\ChannelIdx{\LRSize}{i} \times 1$\label{notation:lrChannelSize} vector, where we again assume the same dimension in all channels, i.\,e. $\ChannelIdx{\LRSize}{i} = \LRSize$ for all $i = 1, \ldots, \NumChannels$. Furthermore, we denote the composite of $\NumChannels$ channels by $\LR = (\LRChannel{1}^\top, \ldots, \LRChannel{\NumChannels}^\top)^\top$\label{notation:lrChannelComposite} that represents the entire set of low-resolution observations as $(\NumChannels \cdot \NumFrames \cdot \LRSize) \times 1$ vector.

Let us consider the low-resolution observations $\LR$ along with the unknown multi-channel image $\HR$ as random variables. We aim at determining the joint posterior probability over all channels:
\begin{equation}
	\label{eqn:06_posterior}
	\begin{split}
		\PdfCond{\HR, \LLRParams}{\LR}
			&= \PdfCond{\HRChannel{1}, \ldots, \HRChannel{\NumChannels}, \LLRParams}{\LRChannel{1}, \ldots, \LRChannel{\NumChannels}}\\
			&= \frac{\PdfCond{\LRChannel{1}, \ldots, \LRChannel{\NumChannels}}{\HRChannel{1}, \ldots, \HRChannel{\NumChannels}}
			\cdot 
			\PdfCond{\HRChannel{1}, \ldots, \HRChannel{\NumChannels}}{\LLRParams} \cdot \Pdf{\LLRParams}}
			{\Pdf{\LRChannel{1}, \ldots, \LRChannel{\NumChannels}}},
	\end{split}
\end{equation}
where $\PdfCondT{\LR}{\HR} = \PdfCondT{\LRChannel{1}, \ldots, \LRChannel{\NumChannels}}{\HRChannel{1}, \ldots, \HRChannel{\NumChannels}}$ is the conditional probability of obtaining the entire set of observations $\LR$ from the unknown multi-channel image $\HR$ and $\PdfCondT{\HR}{\LLRParams} = \PdfCondT{\HRChannel{1}, \ldots, \HRChannel{\NumChannels}}{\LLRParams}$ is the prior probability for $\HR$. In \eref{eqn:06_posterior}, $\LLRParams$ are latent hyperparameters of the imaging process with the assigned distribution $\PdfT{\LLRParams}$.

This section proceeds with the definition of these distributions in a hierarchical way as follows. First, the observation model $\PdfCondT{\LR}{\HR}$ is developed. Accordingly, a prior distribution $\PdfCondT{\HR}{\LLRParams}$ is assigned to the multi-channel image $\HR$ to model its statistical appearance. Eventually, we introduce a prior distribution $\PdfT{\LLRParams}$ that is employed for an inference of the hyperparameters $\LLRParams$.

\subsection{Multi-Channel Observation Model}

In order to calculate the posterior probability in \eref{eqn:06_posterior}, the following assumptions are made to derive the conditional probability $\PdfCondT{\LR}{\HR}$ that represents the observation model: 1) The low-resolution channels $\LRChannel{1}, \ldots, \LRChannel{\NumChannels}$ are mutually independent assuming statistically independent noise among the channels, and 2) the formation of a low-resolution channel $\LRChannel{i}$ depends only on the corresponding high-resolution channel $\HRChannel{i}$ but is independent on the remaining channels, see \fref{fig:multiChannelImageFormation}. Hence, the conditional probability $\PdfCondT{\LR}{\HR}$ can be factorized according to:   
\begin{equation}
	\label{eqn:06_observationModelFactorization}
	\PdfCond{\LRChannel{1}, \ldots, \LRChannel{\NumChannels}}{\HRChannel{1}, \ldots, \HRChannel{\NumChannels}}
	= \prod_{i = 1}^\NumChannels \PdfCond{\LRChannel{i}}{\HRChannel{i}}
	= \prod_{i = 1}^\NumChannels \prod_{k = 1}^\NumFrames \PdfCond{\FrameIdx{\LRChannel{i}}{k}}{\HRChannel{i}},
\end{equation}
where $\PdfCondT{\LRChannel{i}}{\HRChannel{i}}$ is the joint conditional probability of observing the frames of the low-resolution channel $\LRChannel{i}$ from the high-resolution channel $\HRChannel{i}$ and \smash{$\PdfCondT{\FrameIdx{\LRChannel{i}}{k}}{\HRChannel{i}}$} is the conditional probability of observing a single frame \smash{$\FrameIdx{\LRChannel{i}}{k}$}. The formation of the low-resolution observations $\LRChannel{i}$ from the high-resolution channel $\HRChannel{i}$ is described by:
\begin{equation}
	\label{eqn:06_imageFormationModel}
	\LRChannel{i} = \SystemMatChannel{i} \HRChannel{i} + \ChannelIdx{\NoiseVec}{i}
	= \begin{pmatrix}
		\SamplingMat \ChannelIdx{\BlurMat}{i} \ChannelIdx{\FrameIdx{\MotionMat}{1}}{i} \HRChannel{i} \\
		\vdots \\
		\SamplingMat \ChannelIdx{\BlurMat}{i} \ChannelIdx{\FrameIdx{\MotionMat}{\NumFrames}}{i} \HRChannel{i}
	\end{pmatrix}
	+ 
	\begin{pmatrix}
		\FrameIdx{\ChannelIdx{\NoiseVec}{i}}{1}\\
		\vdots\\
		\FrameIdx{\ChannelIdx{\NoiseVec}{i}}{\NumFrames}
	\end{pmatrix},
\end{equation}
where $\SystemMatChannel{i}$\label{notation:systemMatChannel} is the system matrix and $\ChannelIdx{\NoiseVec}{i}$ is additive measurement noise for this channel. The system matrix $\SystemMatChannel{i}$ comprises subsampling modeled by $\SamplingMat$, which is assumed to be constant over the channels. The circulant matrix $\ChannelIdx{\BlurMat}{i}$ denotes space invariant blur associated with the \gls{psf} of the $i$-th channel that might be varying over the channels. \smash{$\ChannelIdx{\FrameIdx{\MotionMat}{k}}{i}$} models subpixel motion relative to the reference coordinate grid for the $k$-th frame associated with the $i$-th channel.

\begin{figure}[!t]
	\centering
		\includegraphics[width=1.00\textwidth]{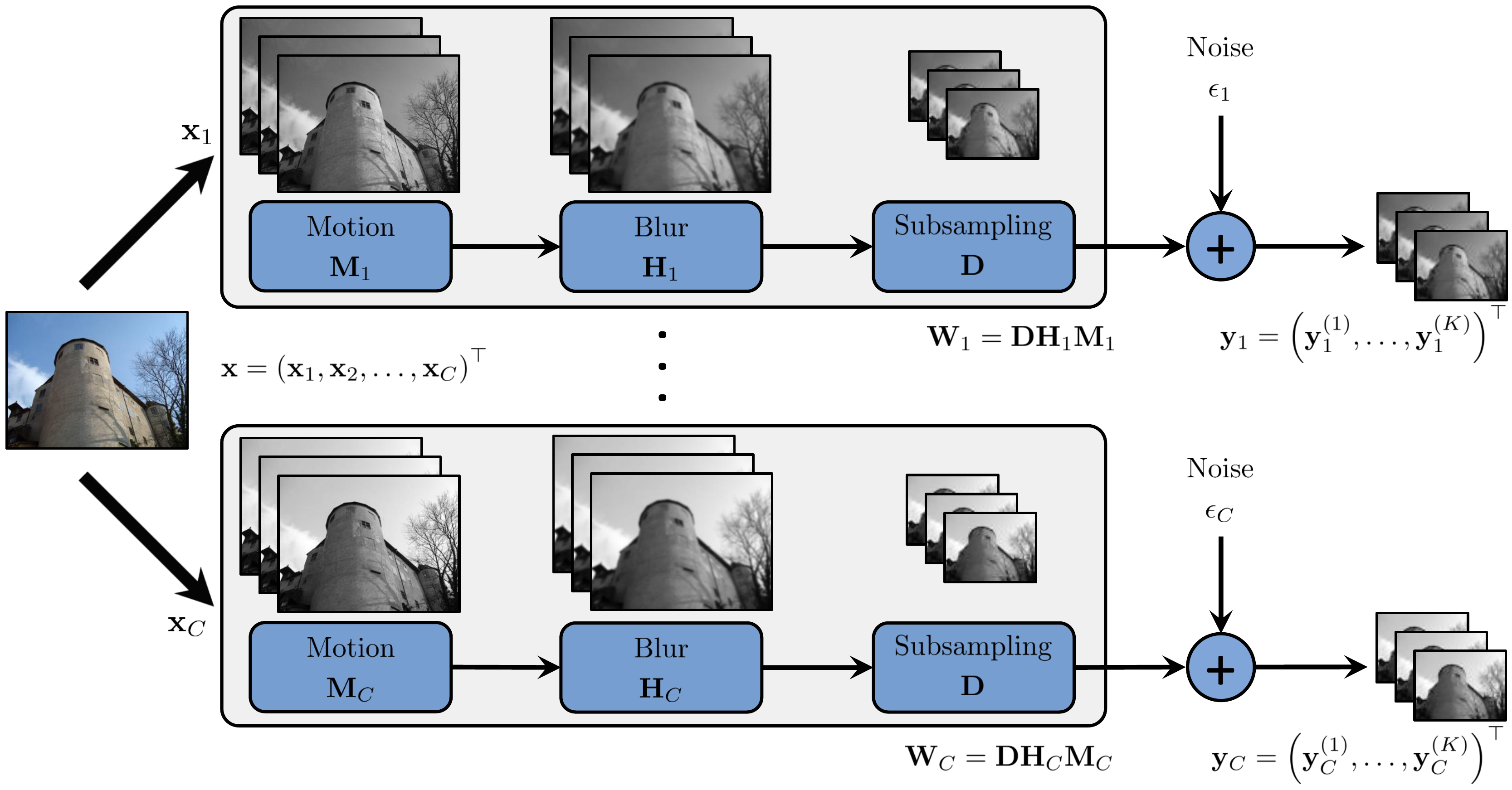}
	\caption[Observation model for multi-channel images]{Formation of low-resolution multi-channel observations $\LR$ encoded in $\NumChannels$ channels and $\NumFrames$ frames from the high-resolution multi-channel image $\HR$.}
	\label{fig:multiChannelImageFormation}
\end{figure}

Based on the factorization in \eref{eqn:06_observationModelFactorization} and the image formation in \eref{eqn:06_imageFormationModel}, the observation model is given by the distribution:
\begin{equation}
	\label{eqn:06_observationModel}
	\PdfCond{\LRChannel{1}, \ldots, \LRChannel{\NumChannels}}{\HRChannel{1}, \ldots, \HRChannel{\NumChannels}}
	\propto \exp 
	\left\{ -  
		\sum_{i = 1}^\NumChannels \sum_{m = 1}^{\NumFrames \LRSize} 
		\LossFun[\LRChannel{i}]{ \VecEl{ \LRChannel{i} - \SystemMatChannel{i} \HRChannel{i}}{m} }
	\right\},
\end{equation}
where $\LossFunSym[\LRChannel{i}]: \Real \rightarrow \RealNonNeg$ denotes a loss function to model the noise distribution for the $i$-th channel. In spirit of the weighted normal distribution proposed in \cref{sec:RobustMultiFrameSuperResolutionWithSparseRegularization} and to make the model robust to space variant noise and outliers, we define \eref{eqn:06_observationModel} based on the Huber loss: 
\begin{equation}
	\LossFun[\LRChannel{i}]{z} =
	\begin{cases}
		z^2
			& \text{if}~ |z| \leq \NoiseStdChannel{i} \\
		2 \NoiseStdChannel{i} |z| - \NoiseStdChannel{i}^2
			& \text{otherwise}
	\end{cases},
\end{equation}
where $\NoiseStdChannel{i}$ denotes the distribution scale parameter that characterizes the noise level associated with the $i$-th channel.

\subsection{Multi-Channel Image Prior Model}
\label{sec:06_MultiChannelImagePriorModel}

Let us now define the image prior $\PdfCondT{\HR}{\LLRParams}$ employed in the posterior probability in \eref{eqn:06_posterior}. In general, this prior needs to be factorized according to:
\begin{equation}
	\label{eqn:06_priorFactorization}
	\begin{split}
		\PdfCond{\HR}{\LLRParams} 
		&= 
			\PdfCond{\HRChannel{1}}{\LLRParams} \PdfCond{\HRChannel{2}}{\HRChannel{1},\LLRParams} \ldots 
			\PdfCond{\HRChannel{\NumChannels}}{\HRChannel{1}, \ldots, \HRChannel{\NumChannels-1},\LLRParams} \\
		&= 
			\PdfCond{\HRChannel{1}}{\LLRParams} 
			\prod_{i = 2}^\NumChannels \PdfCond{\HRChannel{i}}{\HRChannel{1}, \ldots, \HRChannel{i-1}, \LLRParams}.
	\end{split}
\end{equation}
Contrary to the observation model,  this factorization considers dependencies among the high-resolution channels as the key notion of the Bayesian formulation. 

The prior distribution $\PdfCondT{\HR}{\LLRParams}$ considers two complementary aspects. On the one hand, it models the statistical appearance of each individual channel $\HRChannel{i}$, which is related to an \textit{intra}-channel prior. On the other hand, it accounts for statistical dependencies of each channel $\HRChannel{i}$ relative to all other channels $\HRChannel{j}$, $i \neq j$ that is considered by an \textit{inter}-channel prior. Using a pair-wise approach to consider these dependencies, the joint distribution $\PdfCondT{\HR}{\LLRParams}$ in \eref{eqn:06_priorFactorization} is given by:  
\begin{equation}
	\PdfCond{\HR}{\LLRParams} = \prod_{i = 1}^\NumChannels \PdfCond{\HRChannel{i}}{ \ChannelIdx{\mathcal{X}}{i}, \LLRParams},
\end{equation}
and the prior distribution $\PdfCondT{\HRChannel{i}}{\ChannelIdx{\mathcal{X}}{i}, \LLRParams}$ associated with the channel $\HRChannel{i}$ is written as: 
\begin{equation}
	\label{eqn:06_prior}
	\PdfCond{\HRChannel{i}}{\ChannelIdx{\mathcal{X}}{i}, \LLRParams} \propto \exp 
	\left\{ 
		- \ChannelIdx{\RegWeight}{i} \RegTerm[intra]{\HRChannel{i}}
		- \sum_{j = 1, j \neq i}^\NumChannels \ChannelIdx{\mu}{ij} \RegTerm[inter]{\HRChannel{i}, \HRChannel{j}, \ChannelIdx{\LLRParams}{ij}}
	\right\},
\end{equation}
where\label{notation:intraChannelRegTerm}\label{notation:interChannelRegTerm}\label{notation:intraChannelRegWeight}\label{notation:interChannelRegWeight}\label{notation:llrParams} $\ChannelIdx{\mathcal{X}}{i} = \{ \HRChannel{1}, \ldots, \HRChannel{i-1}, \HRChannel{i+1}, \ldots, \HRChannel{\NumChannels} \}$. In \eref{eqn:06_prior}, $\RegTerm[intra]{\HRChannel{i}}$ denotes the regularization term for the intra-channel prior distribution associated with $\HRChannel{i}$. Similarly, $\RegTerm[inter]{\HRChannel{i}, \HRChannel{j}, \ChannelIdx{\LLRParams}{ij}}$ denotes the regularization term of the inter-channel prior for the channel pair $(\HRChannel{i}, \HRChannel{j})$ parametrized by a set of hyperparameters $\ChannelIdx{\LLRParams}{ij}$ as shown below. The regularization weight $\ChannelIdx{\RegWeight}{i} \geq 0$ denotes the contribution of the intra-channel prior to model the statistical appearance of the multi-channel image solely based on the individual channels. The regularization weight $\ChannelIdx{\mu}{ij} \geq 0$ denotes the contribution of the inter-channel prior between $\HRChannel{i}$ and $\HRChannel{j}$. Hence, it considers statistical dependencies according to \eref{eqn:06_priorFactorization}, whereas in case of $\ChannelIdx{\mu}{ij} = \ChannelIdx{\mu}{ji} = 0$ the channels $\HRChannel{i}$ and $\HRChannel{j}$ are treated as independent. We call the prior distribution \textit{symmetric}, if the regularization weights fulfill the property $\ChannelIdx{\mu}{ij} = \ChannelIdx{\mu}{ji}$ for all $i, j = 1, \ldots, \NumChannels$.

\begin{figure}[!t]
	\centering
		\includegraphics[width=1.00\textwidth]{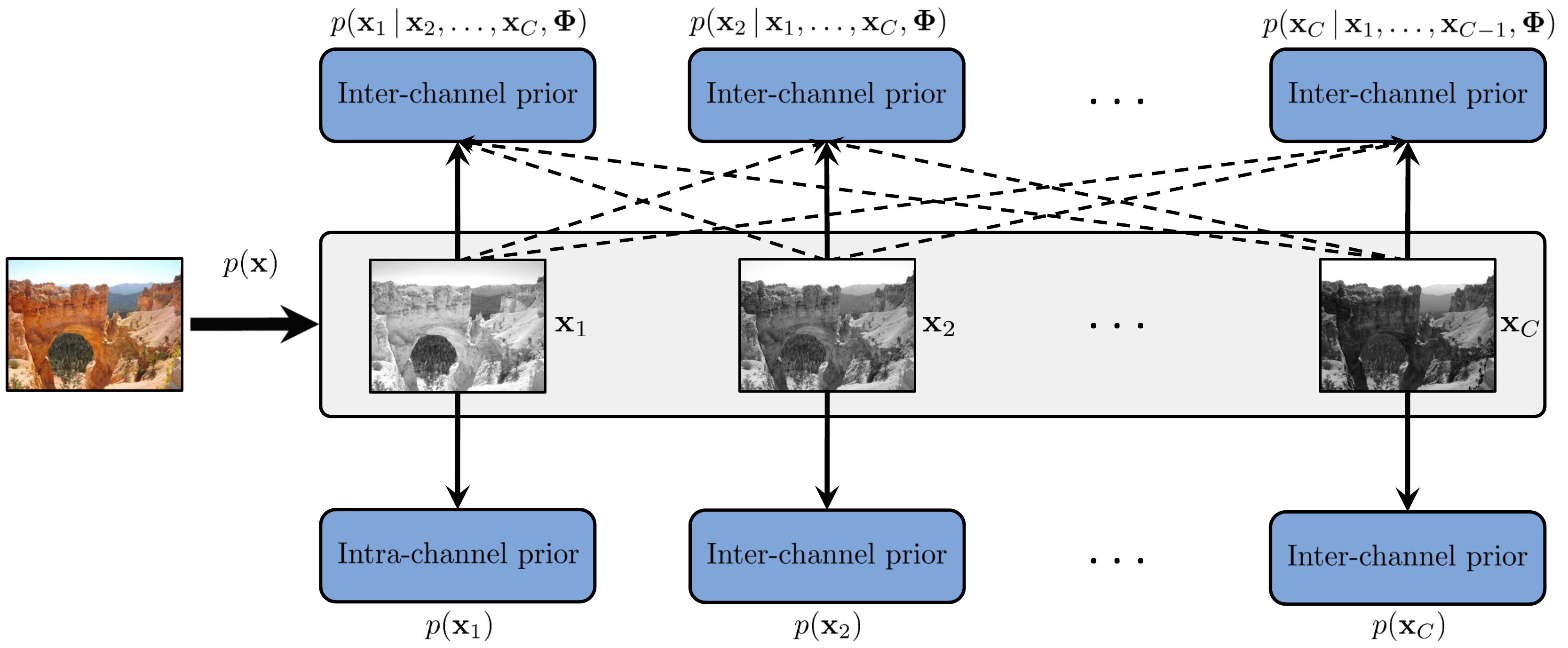}
	\caption[Prior distribution for multi-channel images]{Schematic representation of the prior distribution $\PdfT{\HR}$. The proposed model consists of an inter-channel prior to describe statistical dependencies among pairs of channels and an intra-channel prior to describe the appearance of the individual channels.}
	\label{fig:multiChannelPriorModel}
\end{figure}

\paragraph{Intra-Channel Prior.} 
Following the state-of-the-art in image restoration for single-channel images \cite{Krishnan2009a}, the intra-channel prior needs to account for the sparsity of the individual channels in a certain transform domain. The corresponding regularization term adopts \gls{wbtv} \cite{Kohler2015c} as previously introduced in \cref{sec:RobustMultiFrameSuperResolutionWithSparseRegularization} and is given by:
\begin{equation}
	\label{eqn:06_intraPrior}
	R_\text{intra}(\HRChannel{i}) \defeq \sum_{n = 1}^{\SparseTransDim} \LossFun[\HRChannel{i}] { \VecEl{\SparseTransMat \HRChannel{i}}{n} },
\end{equation}
where $\SparseTransMat \in \RealMN{\SparseTransDim}{\HRSize}$ with $\SparseTransDim = (2 \BTVSize + 1)^2\HRSize$ denotes the linear sparsifying transform:
\begin{equation}
	\SparseTransMat = 
	\begin{pmatrix}
		\BTVWeight^{|-\BTVSize| + |-\BTVSize|} 
		\left( \Id_{\HRSize \times \HRSize} - \SparseTransMat[v]^{-\BTVSize} \SparseTransMat[h]^{-\BTVSize} \right) \\
		\vdots \\
		\BTVWeight^{|+\BTVSize| + |+\BTVSize|} 
		\left( \Id_{\HRSize \times \HRSize} - \SparseTransMat[v]^{+\BTVSize} \SparseTransMat[h]^{+\BTVSize} \right)
	\end{pmatrix}.
\end{equation}
Here, $\BTVWeight \in ]0, 1]$ is the \gls{btv} weighting factor, $\BTVSize \geq 1$ is the \gls{btv} window size, and $\SparseTransMat[v]^m$ and $\SparseTransMat[h]^n$ model vertical and horizontal shifts of $\HRChannel{i}$ by $m$ and $n$ pixels, respectively. The loss function $\LossFunSym[\HRChannel{i}]: \Real \rightarrow \RealNonNeg$ in \eref{eqn:06_intraPrior} is given by the mixed \LOne/\Lp{\SparsityParam} norm:
\begin{equation}
	\LossFun[\HRChannel{i}]{z} =
	\begin{cases}
		|z|
			& \text{if}~ |z| \leq \PriorStdChannel{i} \\
		\PriorStdChannel{i}^{1 - \ChannelIdx{\SparsityParam}{i}} \cdot |z|^{\ChannelIdx{\SparsityParam}{i}}
			& \text{otherwise}
	\end{cases},
\end{equation}
where $\PriorStdChannel{i}$ and $\ChannelIdx{\SparsityParam}{i} \in [0, 1]$ are the prior distribution scale parameter and the sparsity parameter for the channel $\HRChannel{i}$, respectively.

\paragraph{Inter-Channel Prior.}
In \eref{eqn:06_prior}, the inter-channel prior accounts for pair-wise statistical dependencies among the image channels. For this purpose, we follow the assumption that such dependencies can be described \textit{locally} by means of linear regressions on a patch-wise basis. Let $\HRChannel{i}$ and $\HRChannel{j}$ be a pair of disjoint image channels. Then, the regression of $\HRChannel{i}$ towards $\HRChannel{j}$ at the $n$-th pixel position is given by:   
\begin{equation}
	\label{eqn:06_llrPixelwise}
	\ChannelIdx{\HRSym}{j, m} = \ChannelIdx{C}{ij, n} \ChannelIdx{\HRSym}{i, m} + \ChannelIdx{b}{ij, n},
	\qquad
	\text{for all}~ m \in \LLRPatch(n), 
\end{equation}
where the parameters $\ChannelIdx{C}{ij,n}$\label{notation:llrCoeffsMultLocal} and $\ChannelIdx{b}{ij,n}$\label{notation:llrCoeffsAddLocal} denote the local regression coefficients for a $(2 \LLRPatchSize + 1) \times (2 \LLRPatchSize + 1)$\label{notation:llrPatchSize} image patch $\LLRPatch(n)$\label{notation:llrPatch} centered at the $n$-th pixel. These coefficients are assumed to be constant for all pixel positions $m \in \LLRPatch(n)$, see \fref{fig:06_llrModel}. Based on this patch-wise relationship, the inter-channel prior for the channel $\HRChannel{i}$ is defined via the fidelity of a regression towards each of the remaining channels $\HRChannel{j}$, $j \neq i$ over all local patches. This fidelity is stated by the \gls{llr} model:
\begin{equation}
	\label{eqn:06_interPrior}
	\RegTerm[inter]{\HRChannel{i}, \HRChannel{j}, \ChannelIdx{\LLRParams}{ij}} 
	\defeq \sum_{n = 1}^\HRSize \LossFun[\HRChannel{ij}] { \VecEl{ \ChannelIdx{\vec{C}}{ij} \HRChannel{i} + \ChannelIdx{\vec{b}}{ij} - \HRChannel{j}}{n} },
\end{equation}
where \smash{$\ChannelIdx{\vec{C}}{ij} = \Diag{\ChannelIdx{C}{ij, 1}, \ldots, \ChannelIdx{C}{ij, \HRSize}} \in \RealMN{\HRSize}{\HRSize}$}\label{notation:llrCoeffsMultGlobal} and \smash{$\ChannelIdx{\vec{b}}{ij} = \left(\ChannelIdx{b}{ij, 1}, \ldots, \ChannelIdx{b}{ij, \HRSize} \right)^\top \in \RealN{\HRSize}$}\label{notation:llrCoeffsAddGlobal} are regression coefficients over the entire image assembled from the pixel-wise coefficients $\ChannelIdx{C}{ij, n}$ and $\ChannelIdx{b}{ij, n}$ in \eref{eqn:06_llrPixelwise}, respectively. We denote by $\ChannelIdx{\LLRParams}{ij} = \{ \ChannelIdx{\vec{C}}{ij}, \ChannelIdx{\vec{b}}{ij} \}$ the set of coefficients that are treated as hyperparameters of the prior distribution. In \eref{eqn:06_interPrior}, $\LossFunSym[\HRChannel{ij}]: \Real \rightarrow \RealNonNeg$ denotes a loss function to measure the regression fidelity. In order to tolerate outliers regarding the linear regression assumption, we define the \gls{llr} model according to Tukey's biweight function \cite{Meer1991}:
\begin{equation}
	\label{eqn:06_interChannelLossFun}
	\LossFun[\HRChannel{ij}]{z} =
	\begin{cases}
		\frac{1}{6} \LLRStd{i}{j}^2 \left( 1 - \left(1 - \frac{z^2}{\LLRStd{i}{j}^2} \right)^3 \right)
			& \text{if}~ |z| \leq \LLRStd{i}{j} \\
		\frac{1}{6} \LLRStd{i}{j}^2
			& \text{otherwise}
	\end{cases},
\end{equation}
where $\LLRStd{i}{j}$ is the distribution scale parameter for the channels $\HRChannel{i}$ and $\HRChannel{j}$.

\begin{figure}[!t]
	\centering
		\includegraphics[width=0.98\textwidth]{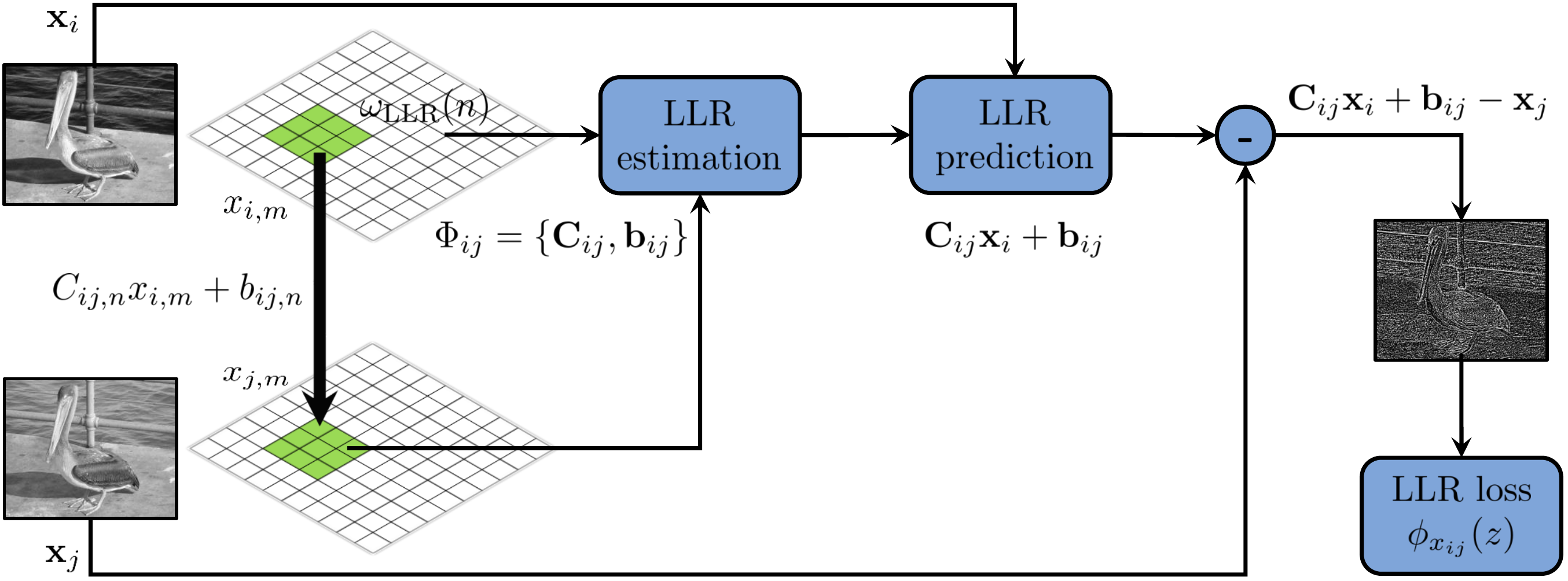}
	\caption[\Gls{llr} model for pairs of color channels]{Schematic representation of the \acrfull{llr} model of the channel $\HRChannel{i}$ towards the channel $\HRChannel{j}$ depicted for pairs of color channels. We establish the \gls{llr} model for image patches $\LLRPatch(n)$ and define the corresponding prior distribution based on the regression residual error and an outlier-insensitive loss function.}
	\label{fig:06_llrModel}
\end{figure}

It is worth noting that similar regression models have been proposed previously for multi-sensor super-resolution \cite{Zomet2001,Ghesu2014} as well as joint filtering \cite{He2013,Shen2015}. In particular, the regression in \eref{eqn:06_interPrior} generalizes the concept of guided filtering as proposed by He \etal \cite{He2013}. However, the key novelty is that the proposed prior is applicable to an arbitrary number of image channels and represents a Bayesian interpretation of mutual dependencies, while guided filtering considers only dependencies of a filter input relative to a static guidance. Moreover, since \eref{eqn:06_interPrior} is formulated via an outlier-insensitive loss function, it is spatially adaptive and features robustness regarding image regions that violate the linear regression assumption. We elaborate on these properties with comparisons to related methods in more detail in \sref{sec:06_AlgorithmAnalysis}.

\paragraph{Prior on the Hyperparameters.}

The \gls{llr} prior distribution relies on knowledge regarding the regression coefficients but in general these parameters are unknown. For this reason, they are treated as latent hyperparameters and in order to estimate them, we need to assign a meaningful prior distribution $\PdfT{\LLRParams}$. Let us assume that the coefficients associated with the different channels are mutually independent random variables. Then, the joint prior distribution $\PdfT{\LLRParams} = \PdfT{\ChannelIdx{\LLRParams}{11}, \ldots, \ChannelIdx{\LLRParams}{\NumChannels \NumChannels}}$ with $\ChannelIdx{\LLRParams}{ij} = \{ \ChannelIdx{\vec{C}}{ij}, \ChannelIdx{\vec{b}}{ij} \}$ can be factorized to:
\begin{equation}
	\Pdf{\LLRParams} = \prod_{i = 1}^\NumChannels \prod_{j = 1, j \neq i}^\NumChannels \Pdf{ \ChannelIdx{\vec{C}}{ij} } \Pdf{ \ChannelIdx{\vec{b}}{ij}}.
\end{equation}

In this work, $\PdfT{\ChannelIdx{\vec{b}}{ij}}$ is assumed to be a uniform distribution. In addition, $\PdfT{\ChannelIdx{\vec{C}}{ij}}$ is adopted from ridge regression \cite{He2013} and is given by the normal distribution:
\begin{equation}
	\Pdf{\ChannelIdx{\vec{C}}{ij}} \propto \exp \left\{ - \ChannelIdx{\epsilon}{ij} 
	\left|\left| \ChannelIdx{\vec{C}}{ij} \right|\right|_F^2 \right\},
\end{equation}
where $|| \cdot ||_F$ is the Frobenius norm that is given by $|| \ChannelIdx{\vec{C}}{ij} ||_F = || \Diag{\ChannelIdx{\vec{C}}{ij}} ||_2$ for the diagonal matrix $\ChannelIdx{\vec{C}}{ij}$ and $\ChannelIdx{\epsilon}{ij} \geq 0$ denotes a hyperparameter regularization weight for the channels $\HRChannel{i}$ and $\HRChannel{j}$. Intuitively, this prior distribution penalizes large coefficients $\ChannelIdx{\vec{C}}{ij}$. As shown in \sref{sec:06_BayesianApproachToMultiChannelSuperResolution}, the benefit of this prior is that the regression coefficients can be estimated in closed-form within the proposed algorithm\footnote{This prior leads to a ridge regression problem to determine the regression coefficients \cite{He2013}.}. 
 
\section{Bayesian Multi-Channel Super-Resolution}
\label{sec:06_BayesianApproachToMultiChannelSuperResolution}

This section aims at the development of parameter estimation techniques based on the proposed Bayesian model of multi-channel images. More specifically, we are interested in obtaining point estimates of the high-resolution channels $\HR$ given the low-resolution, multi-channel observations $\LR$. For this purpose, two approaches can be distinguished.

\subsection{Sequential Maximum A-Posteriori Estimation}

Let us first discuss the \textit{sequential} estimation of the unknown high-resolution image channels, which serves as a baseline approach in this chapter. Therefore, let the inter-channel prior $\PdfCondT{\HRChannel{i}}{\HRChannel{j}, \ChannelIdx{\LLRParams}{ij}}$ for each pair of channels be a uniform distribution. Accordingly, statistical dependencies among the channels can be ignored and the \gls{map} estimate for the high-resolution image $\HR$ is given by:  
\begin{equation}
	\label{eqn:06_sequentialEstimationMAP}
	\HR_{\text{MAP}} = \argmax_{\HR} \prod_{i = 1}^\NumChannels \PdfCond{\LRChannel{i}}{\HRChannel{i}} \prod_{i = 1}^\NumChannels \Pdf{\HRChannel{i}}.
\end{equation}
Based on this simplifying assumption, the unknown high-resolution channels $\HRChannel{i}$ for $i = 1,\ldots,C$ are reconstructed independently by minimizing the negative log-likelihood of \eref{eqn:06_sequentialEstimationMAP}:
\begin{equation}
	\label{eqn:06_sequentialEstimationOjectiveFunction}
	{\HRChannel{i}}_{,\,\text{MAP}} = \argmin_{\HRChannel{i}} 
	\left\{ \DataTerm{\HRChannel{i}} + \ChannelIdx{\RegWeight}{i} \RegTerm[intra]{\HRChannel{i}} \right\}.
\end{equation}

This approach is based solely on the observation model related to the data fidelity term $\DataTerm{\HRChannel{i}} \propto - \log \PdfCondT{\LRChannel{i}}{\HRChannel{i}} $ and the intra-channel regularization term $\RegTerm[intra]{\HRChannel{i}}$ associated with the considered channel. The advantage of sequential estimation is its conceptual simplicity, as \eref{eqn:06_sequentialEstimationOjectiveFunction} can be solved straightforwardly using super-resolution for single-channel images. However, its limitation is that statistical dependencies are ignored, which might cause inconsistencies among the super-resolved image channels. 

\subsection{Joint Maximum A-Posteriori Estimation}

Our aim is to jointly estimate the image channels under consideration of inter-channel dependencies. Using the joint posterior distribution in \eref{eqn:06_posterior}, the goal is to determine the high-resolution image $\HR$ along with the latent regression coefficients $\LLRParams$ by the joint \gls{map} estimation:
\begin{equation}
	\label{eqn:06_jointMAP}
	\left( \HR_{\text{MAP}} , \LLRParams_{\text{MAP}}  \right) 
	= \argmax_{\HR, \LLRParams} 
	\PdfCond{\LR}{\HR}
	\PdfCond{\HR}{\LLRParams} 
	\Pdf{\LLRParams}.
\end{equation}
Notice that \eref{eqn:06_jointMAP}  is non-convex due to the non-convexity of the regularization terms related to the prior distribution. Moreover, the dimension of the parameter space $(\HR, \LLRParams)$ is $\NumChannels \HRSize + \NumChannels(\NumChannels - 1) \HRSize = \NumChannels^2 \HRSize$ and the number of observations is $\NumChannels \NumFrames \LRSize$. Thus, the dimension of the parameter space grows quadratically as a function of the number of channels and joint \gls{map} estimation is underdetermined if $\NumFrames \LRSize < \NumChannels \HRSize$. These properties make iterative optimization challenging and a joint numerical optimization would be computationally prohibitive due to the high dimensionality of the parameter space. 

For an efficient numerical solution, we alternatively solve \eref{eqn:06_jointMAP} \wrt the high-resolution image $\HR$ and the regression coefficients $\LLRParams$ while keeping the other parameters fixed. Starting at an initial guess $(\IterationIdx{\HR}{0}, \IterationIdx{\LLRParams}{0})$, this leads to a sequence of estimates $(\IterationIdx{\HR}{t}, \IterationIdx{\LLRParams}{t})$ according to the iteration scheme:
\begin{align}
	\label{eqn:06_amHyperparameters}
	\IterationIdx{\LLRParams}{t} &= \argmax_{\LLRParams} \PdfCond{\IterationIdx{\HR}{t-1}}{\LLRParams} \Pdf{\LLRParams}, \\
	\label{eqn:06_amImage}
	\IterationIdx{\HR}{t} &= \argmax_{\HR} \PdfCond{\LR}{\HR} \PdfCond{\HR}{\IterationIdx{\LLRParams}{t}},
\end{align}
for $t = 1, \ldots, \NumIter[am]$. Let us now derive efficient solutions for both substeps.

\paragraph{Estimation of the Regression Coefficients.}
Given the estimate $\IterationIdx{\HR}{t-1}$ for the image channels at iteration $t-1$, the latent regression coefficients are obtained by minimizing the negative log-likelihood of \eref{eqn:06_amHyperparameters}:  
\begin{equation}
	\label{eqn:06_hyperparameterSubproblem}
	\IterationIdx{\LLRParams}{t}
	= \argmin_{ \LLRParams } \sum_{i = 1}^\NumChannels \sum_{j = 1, j \neq i}^\NumChannels \IterationIdx{\ChannelIdx{\EnergyFunSym}{ij}}{t}(\ChannelIdx{\vec{C}}{ij}, \ChannelIdx{\vec{b}}{ij}).
\end{equation}
This energy minimization is separable and the regression coefficients among the $i$-th and $j$-th channel are obtained by optimizing the negative log-likelihood of the corresponding prior distributions. In order to solve \eref{eqn:06_hyperparameterSubproblem} efficiently, an upper bound of this log-likelihood term is minimized. For this purpose, the non-convex biweight loss function $\LossFun[\HRChannel{ij}]{z}$ is rewritten by means of an \gls{mm} algorithm \cite{Hunter2004}, see \appref{sec:A_MajorizationMinimizationForTheRegressionCoefficients}. This leads to the confidence-aware energy function:
\begin{equation}
	\label{eqn:06_hyperparameterSubproblemEnergyFunction}
	\IterationIdx{\ChannelIdx{\EnergyFunSym}{ij}}{t}(\ChannelIdx{\vec{C}}{ij}, \ChannelIdx{\vec{b}}{ij}) 
	=  
	\big(\ChannelIdx{\vec{C}}{ij} \IterationIdx{\HRChannel{i}}{t-1} + \ChannelIdx{\vec{b}}{ij} - \IterationIdx{\HRChannel{j}}{t-1} \big)^\top 
	\IterationIdx{\ChannelIdx{\vec{K}}{ij}}{t}
	\big(\ChannelIdx{\vec{C}}{ij} \IterationIdx{\HRChannel{i}}{t-1} + \ChannelIdx{\vec{b}}{ij} - \IterationIdx{\HRChannel{j}}{t-1} \big)
	+ \ChannelIdx{\epsilon}{ij} \left| \left|  \ChannelIdx{\vec{C}}{ij}  \right| \right|_F^2.
\end{equation}
The confidence weights $\IterationIdx{\ChannelIdx{\vec{K}}{ij}}{t}$ used in this convex energy function at iteration $t$ are assembled as the diagonal matrix:
\begin{equation}
	\IterationIdx{\ChannelIdx{\vec{K}}{ij}}{t} = 
	\Diag{
	\kappa_1 \left( \IterationIdx{\ChannelIdx{\vec{C}}{ij}}{t-1}, \IterationIdx{\ChannelIdx{\vec{b}}{ij}}{t-1} \right)~~
	~~\kappa_2 \left( \IterationIdx{\ChannelIdx{\vec{C}}{ij}}{t-1}, \IterationIdx{\ChannelIdx{\vec{b}}{ij}}{t-1} \right)~~
	~~\ldots~~
	~~\kappa_\HRSize \left(\IterationIdx{\ChannelIdx{\vec{C}}{ij}}{t-1}, \IterationIdx{\ChannelIdx{\vec{b}}{ij}}{t-1} \right)
	},
\end{equation}
and the weighting function to obtain the $k$-th weight is given by:
\begin{align}
	\label{eqn:weightingFunction}
	\kappa_k ( \ChannelIdx{\vec{C}}{ij}, \ChannelIdx{\vec{b}}{ij}) &= 
	\begin{cases}
		\left(1 - \left( \frac{ \ResidualSym_{ij,k}( \ChannelIdx{\vec{C}}{ij}, \ChannelIdx{\vec{b}}{ij}) }{ c_{\text{LLR}} \cdot \IterationIdx{\LLRStd{i}{j}}{t} } \right)^2 \right)^2
			&	\text{if}~ \left| \ResidualSym_{ij,k}( \ChannelIdx{\vec{C}}{ij}, \ChannelIdx{\vec{b}}{ij}) \right| \leq c_{\text{LLR}} \IterationIdx{\LLRStd{i}{j}}{t} \\
		0																
			& \text{otherwise}
	\end{cases},\\
	\ResidualSym_{ij}( \ChannelIdx{\vec{C}}{ij}, \ChannelIdx{\vec{b}}{ij}) 
	&= Q_{\LLRPatch} \left( \ChannelIdx{\vec{C}}{ij} \IterationIdx{\HRChannel{i}}{t-1} + \ChannelIdx{\vec{b}}{ij} - \IterationIdx{\HRChannel{j}}{t-1} \right),
\end{align}
where $\ResidualSym_{ij}( \ChannelIdx{\vec{C}}{ij}, \ChannelIdx{\vec{b}}{ij})$ denotes a filtered version of the regression residual error among the given channels. Notice that in order to reduce the influence of isolated pixels with large regression error and to avoid the origination of pseudo-structures by the inter-channel prior, $Q_{\LLRPatch}(\cdot)$ is implemented as median filter with window size $(2 \LLRPatchSize + 1) \times (2 \LLRPatchSize + 1)$. The tuning constant $c_{\text{LLR}}$ for Tukey's biweight loss is set to be \smash{$c_{\text{LLR}} = 4.6851$} to achieve a \smash{95\,\%} asymptotic efficiency under a normal distribution of the regression error \cite{Meer1991}. The unknown distribution scale parameter \smash{$\IterationIdx{\LLRStd{i}{j}}{t}$} is adaptively updated at each iteration based on the weighted \gls{mad} rule under the weights $\IterationIdx{\ChannelIdx{\vec{K}}{ij}}{t-1}$ according to: 
\begin{equation}
	\IterationIdx{\LLRStd{i}{j}}{t}
	= \sigma_0 \cdot \WMad{\ResidualSym_{ij}( \IterationIdx{\ChannelIdx{\vec{C}}{ij}}{t-1}, \IterationIdx{\ChannelIdx{\vec{b}}{ij}}{t-1}) } {\IterationIdx{\ChannelIdx{\vec{K}}{ij}}{t-1}},
\end{equation}
where $\sigma_0 = 1.4826$ to obtain a consistent estimate under a normal distribution for the regression inliers \cite{Scales1988}.  

The minimization of the energy function in \eref{eqn:06_hyperparameterSubproblemEnergyFunction} needs to consider overlapping image patches to establish linear regressions. However, the regression coefficients are defined to be constant within each patch according to the definition of the \gls{llr} model, see \sref{sec:06_MultiChannelImagePriorModel}. It is worth noting that this constraint avoids the trivial solution for the regression coefficients (\smash{$\IterationIdx{\ChannelIdx{\vec{C}}{ij}}{t} = \Zeros$} and \smash{$\IterationIdx{\ChannelIdx{\vec{b}}{ij}}{t} = -\IterationIdx{\HRChannel{i}}{t-1}$}). In order to consider this constraint, we utilize the separability of \eref{eqn:06_hyperparameterSubproblemEnergyFunction} and estimate local regression coefficients associated with the image $\LLRPatch(k)$ centered at the $k$-th pixel position in the image channels $\HRChannel{i}$ and $\HRChannel{j}$ according to:
\begin{equation}
	(\IterationIdx{\ChannelIdx{\tilde{C}}{ij, k}}{t}, \IterationIdx{\ChannelIdx{\tilde{b}}{ij, k}}{t}) 
	= \argmin_{\ChannelIdx{C}{ij, k}, \ChannelIdx{b}{ij, k}} 
	\sum_{l \in \LLRPatch(k)} \ChannelIdx{\kappa}{ij,l} \left( \ChannelIdx{C}{ij, k} \IterationIdx{\ChannelIdx{\HRSym}{i,l}}{t-1} + \ChannelIdx{b}{ij, k} - \IterationIdx{\ChannelIdx{\HRSym}{j,l}}{t-1} \right)^2 + \ChannelIdx{\epsilon}{ij} \ChannelIdx{C}{ij, k}^2\enspace,
\end{equation}
where the confidence weights are computed by $\ChannelIdx{\kappa}{ij,l} = \kappa_l ( \IterationIdx{\ChannelIdx{\vec{C}}{ij}}{t-1}, \IterationIdx{\ChannelIdx{\vec{b}}{ij}}{t-1} )$ according to \eref{eqn:weightingFunction}. This ridge regression problem is equivalent to confidence-aware guided filtering \cite{Horentrup2014}. Hence, the local coefficients are computed in closed-form:
\begin{align}
	\label{eqn:guidedFilterCoefficients_1}
	\IterationIdx{\ChannelIdx{\tilde{C}}{ij, k}}{t} 
		&= \frac{\text{E}_{\LLRPatch(k)} \left( \IterationIdx{\HRChannel{i}}{t-1} \odot \IterationIdx{\HRChannel{j}}{t-1}, \IterationIdx{\ChannelIdx{\vec{K}}{ij}}{t} \right) - \text{E}_{\LLRPatch(k)} \left(\IterationIdx{\HRChannel{i}}{t-1}, \IterationIdx{\ChannelIdx{\vec{K}}{ij}}{t} \right) 
		\cdot \text{E}_{\LLRPatch(k)} \left(\IterationIdx{\HRChannel{j}}{t-1}, \IterationIdx{\ChannelIdx{\vec{K}}{ij}}{t} \right)}
		{\text{E}_{\LLRPatch(k)} \left(\IterationIdx{\HRChannel{i}}{t-1} \odot \IterationIdx{\HRChannel{i}}{t-1}, \IterationIdx{\ChannelIdx{\vec{K}}{ij}}{t} \right) + \ChannelIdx{\epsilon}{ij}}, \\
	\label{eqn:guidedFilterCoefficients_2}
	\IterationIdx{\ChannelIdx{\tilde{b}}{ij, k}}{t} 
		&= \text{E}_{\LLRPatch(k)} \left(\IterationIdx{\HRChannel{j}}{t-1}, \IterationIdx{\ChannelIdx{\vec{K}}{ij}}{t} \right) - \IterationIdx{\ChannelIdx{\tilde{C}}{ij}}{t} \cdot \text{E}_{\LLRPatch(k)} \left(\IterationIdx{\HRChannel{i}}{t-1}, \IterationIdx{\ChannelIdx{\vec{K}}{ij}}{t} \right),
\end{align}
where $\text{E}_{\LLRPatch(k)}(\vec{z}, \vec{K})$ denotes the weighted mean in the image patch $\LLRPatch(k)$ centered at the $k$-th pixel in $\vec{z}$ under the confidence weights $\vec{K}$, see \appref{sec:A_EstimationOfTheRegressionCoefficients}. 

The regression coefficients over the entire image are computed by averaging the local coefficients \smash{$\ChannelIdx{\tilde{\vec{C}}}{ij}$ and $\ChannelIdx{\tilde{\vec{b}}}{ij}$} corresponding to overlapping image patches following related strategies in image filtering \cite{Horentrup2014,He2013}. Thus, the regression coefficients for the image channels $\HRChannel{i}$ and $\HRChannel{j}$ are obtained by:
\begin{align}
	\IterationIdx{\ChannelIdx{\vec{C}}{ij}}{t} 
	&= \mathrm{diag}
	\begin{pmatrix}
	\text{E}_{\LLRPatch(1)} \left( \IterationIdx{\ChannelIdx{\tilde{\vec{C}}}{ij}}{t}, \IterationIdx{\ChannelIdx{\vec{K}}{ij}}{t} \right)
	& \ldots
	& \text{E}_{\LLRPatch(\HRSize)} \left( \IterationIdx{\ChannelIdx{\tilde{\vec{C}}}{ij}}{t}, \IterationIdx{\ChannelIdx{\vec{K}}{ij}}{t} \right)
	\end{pmatrix},
	\\
	\IterationIdx{\ChannelIdx{\vec{b}}{ij}}{t} 
	&= \begin{pmatrix}
	\text{E}_{\LLRPatch(1)} \left( \IterationIdx{\ChannelIdx{\tilde{\vec{b}}}{ij}}{t}, \IterationIdx{\ChannelIdx{\vec{K}}{ij}}{t} \right)
	& \ldots 
	& \text{E}_{\LLRPatch(\HRSize)} \left( \IterationIdx{\ChannelIdx{\tilde{\vec{b}}}{ij}}{t}, \IterationIdx{\ChannelIdx{\vec{K}}{ij}}{t} \right)
	\end{pmatrix}^\top.
\end{align}
It is important to note that this calculation of the regression coefficients for a single pair of image channels can be implemented with a time complexity of $\BigO{\HRSize}$ using box filtering, see \sref{sec:06_ComputationalComplexityAndConvergence}. This considerably accelerates the hyperparameter estimation compared to the use of gradient-based optimization.

\paragraph{Estimation of the Image Channels.}
Given the estimate $\IterationIdx{\LLRParams}{t}$ for the regression coefficients, the high-resolution image $\IterationIdx{\HR}{t}$ is estimated by jointly minimizing the negative log-likelihood of \eref{eqn:06_amImage} \wrt the individual channels $\HRChannel{1}, \ldots, \HRChannel{\NumChannels}$. That is, we obtain $\IterationIdx{\HR}{t}$ as the solution of the energy minimization problem:
\begin{equation}
	\label{eqn:06_jointImageChannelEstimation}
	\IterationIdx{\HR}{t} = \argmin_{\HRChannel{1}, \ldots, \HRChannel{\NumChannels}} \IterationIdx{\EnergyFunSym}{t}(\HR).
\end{equation}

Similar to the estimation of the regression coefficients, the optimization of this non-convex log-likelihood term is rewritten as an \gls{mm} algorithm \cite{Hunter2004}. This leads to a weighted but convex minimization problem, see \sref{sec:04_AlgorithmAnalysis}. Thus, the joint energy function is given by:
\begin{equation}
	\label{eqn:06_imageChannelEstimationEnergyFunction}
	\begin{split}
		\IterationIdx{\EnergyFunSym}{t}(\HR)
		= & \sum_{i = 1}^\NumChannels \left( \LRChannel{i} - \SystemMatChannel{i} \HRChannel{i} \right)^\top 
			\IterationIdx{\ChannelIdx{\WeightsBMat}{i}}{t} 
			\left( \LRChannel{i} - \SystemMatChannel{i} \HRChannel{i} \right) 
		+ \ChannelIdx{\RegWeight}{i} \left| \left| \IterationIdx{\ChannelIdx{\WeightsAMat}{i}}{t} \HRChannel{i} \right| \right|_1 \\
		&+ \sum_{j = 1, j \neq i}^\NumChannels 
			\ChannelIdx{\mu}{ij} \left( \IterationIdx{\ChannelIdx{\vec{C}}{ij}}{t} \HRChannel{i} + \IterationIdx{\ChannelIdx{\vec{b}}{ij}}{t} - \HRChannel{j} \right)^\top
			\IterationIdx{\ChannelIdx{\vec{K}}{ij}}{t}
			\left( \IterationIdx{\ChannelIdx{\vec{C}}{ij}}{t} \HRChannel{i} + \IterationIdx{\ChannelIdx{\vec{b}}{ij}}{t} - \HRChannel{j} \right),
	\end{split}
\end{equation}
where the confidence weights for the $i$-th channel at iteration $t$ are assembled as the diagonal matrices:
\begin{align}
	\IterationIdx{\ChannelIdx{\WeightsAMat}{i}}{t} 
	&= \Diag{\ChannelIdx{\WeightsAFun}{i,1} \left(\IterationIdx{\HR}{t-1}\right)~~ 
	~~\ChannelIdx{\WeightsAFun}{i,2} \left(\IterationIdx{\HR}{t-1}\right)~~
	~~\ldots~~ 
	~~\ChannelIdx{\WeightsAFun}{i, \SparseTransDim} \left(\IterationIdx{\HR}{t-1}\right) }, \\
	\IterationIdx{\ChannelIdx{\WeightsBMat}{i}}{t} 
	&= \Diag{\ChannelIdx{\WeightsBFun}{i,1} \left(\IterationIdx{\HR}{t-1}\right)~~ 
	~~\ChannelIdx{\WeightsBFun}{i,2} \left(\IterationIdx{\HR}{t-1}\right)~~
	~~ \ldots~~ 
	~~\ChannelIdx{\WeightsBFun}{i,\NumFrames \LRSize} \left(\IterationIdx{\HR}{t-1}\right) },
\end{align}
using the weighting functions $\ChannelIdx{\WeightsAFun}{i,k}(\IterationIdx{\HR}{t-1}) \equiv \WeightsAFun_k(\IterationIdx{\ChannelIdx{\HR}{i}}{t-1})$ and $\ChannelIdx{\WeightsBFun}{i,k}(\IterationIdx{\HR}{t-1}) \equiv \WeightsBFun_k(\IterationIdx{\ChannelIdx{\HR}{i}}{t-1})$ with adaptive scale parameter selection as previously introduced in \sref{sec:04_NumericalOptimization}, see \erefmore{eqn:04_dataFidelityWeights}{eqn:04_regularizationWeights}. Note that both weighting functions can be applied channel-wise to determine the corresponding confidence weights and to majorize the negative log-likelihood of \eref{eqn:06_jointMAP}.

Numerical optimization of \eref{eqn:06_jointImageChannelEstimation} is performed by means of \gls{scg} iterations starting from \smash{$\IterationIdx{\HR}{t-1}$} obtained at the previous iteration. This gradient-based iteration scheme seeks a stationary point:
\begin{equation}
	\label{eqn:06_imageChannelEstimationEnergyFunctionGrad}
	\nabla_{\HR} \IterationIdx{\EnergyFunSym}{t}(\HR) =
	\begin{pmatrix}
		\frac{\partial \IterationIdx{\EnergyFunSym}{t}}{\partial \HRChannel{1}}(\HR) &
		\frac{\partial \IterationIdx{\EnergyFunSym}{t}}{\partial \HRChannel{2}}(\HR) &
		\hdots &
		\frac{\partial \IterationIdx{\EnergyFunSym}{t}}{\partial \HRChannel{\NumChannels}}(\HR)
	\end{pmatrix}^\top
	= \Zeros, 
\end{equation}
where the gradient of the joint energy function \wrt the $k$-th channel, $k = 1, \ldots, \NumChannels$ is computed in closed-form:
\begin{equation}
	\begin{split}
		\frac{\partial \IterationIdx{ \EnergyFunSym }{t} }{\partial \HRChannel{k}}
		= &-2 \IterationIdx{\ChannelIdx{\WeightsBMat}{k}}{t} \SystemMatChannel{k}^\top 
			\Big(\LRChannel{k} - \SystemMatChannel{k} \HRChannel{k} \Big)
		+ \ChannelIdx{\RegWeight}{k} \cdot \IterationIdx{\ChannelIdx{\WeightsAMat}{k}}{t} \vec{S}^\top 
			\Sign{ \IterationIdx{\ChannelIdx{\WeightsAMat}{k}}{t} \vec{S} \HRChannel{k} }\\
		&+ \sum_{j = 1, j \neq k}^\NumChannels 2 \ChannelIdx{\mu}{kj} \cdot \IterationIdx{\ChannelIdx{\vec{K}}{kj}}{t} \IterationIdx{\ChannelIdx{\vec{C}}{kj}}{t} \Big( \IterationIdx{\ChannelIdx{\vec{C}}{kj}}{t} \HRChannel{k} + \IterationIdx{\ChannelIdx{\vec{b}}{kj}}{t} - \HRChannel{j} \Big)
		- \sum_{i = 1, i \neq k}^\NumChannels 2 \ChannelIdx{\mu}{ik} \cdot \IterationIdx{\ChannelIdx{\vec{K}}{ik}}{t} \Big(\IterationIdx{\ChannelIdx{\vec{C}}{ik}}{t} \HRChannel{i} + \IterationIdx{\ChannelIdx{\vec{b}}{ik}}{t} - \HRChannel{k} \Big).
	\end{split}	
\end{equation}
To facilitate gradient-based optimization, the derivatives of non-smooth \LOne norm terms are approximated by the Charbonnier function, \ie $\Sign{z} \approx z / (\sqrt{z^2 + \tau})$, where $\tau$ is a small constant ($\tau = 10^{-4}$) to ensure differentiability at $z = 0$ \cite{Charbonnier1994}.
 
\paragraph{Overall Optimization Algorithm.}
We divide the proposed alternating minimization scheme in an outer and two inner optimization loops, see \aref{alg:llrSuperResolutionAlgorithm}. In order to provide an initialization for the iterations, a sequential estimation of the high-resolution image channels is performed. This can be done by minimizing \eref{eqn:06_jointImageChannelEstimation} without inter-channel prior (\ie $\ChannelIdx{\mu}{ij} = 0$ for all $i, j = 1, \ldots, \NumChannels$), which is equivalent to a channel-wise iteratively re-weighted reconstruction as previously introduced in \cref{sec:RobustMultiFrameSuperResolutionWithSparseRegularization} using constant regularization weights $\ChannelIdx{\RegWeight}{i}$. Subsequently, the regression coefficients are computed pair-wise based on \ereftwo{eqn:guidedFilterCoefficients_1}{eqn:guidedFilterCoefficients_2} while the high-resolution channels are estimated jointly by \gls{scg} iterations for \eref{eqn:06_jointImageChannelEstimation}.

In total, we perform a maximum number of $\NumIter[am]$ iterations for alternating minimization and a maximum number of $\NumIter[scg]$ iterations for \gls{scg} to estimate the high-resolution image channels. As a termination criterion we choose the maximum absolute difference among consecutive iterations:
\begin{equation}
	\label{eqn:06_amTerminationCriterion}
	\max_{i = 1, \ldots, C} \left( \max_{k = 1, \ldots, \HRSize} \left( \left| \IterationIdx{\ChannelIdx{\HRSym}{i, k}}{t} - \IterationIdx{\ChannelIdx{\HRSym}{i, k}}{t - 1} \right| \right) \right) < \TermTolerance,
\end{equation}
where $\TermTolerance$ denotes the termination tolerance.
\begin{algorithm}[!t]
	\small
	\caption{Multi-sensor super-resolution using \acrfull{llr}}
	\label{alg:llrSuperResolutionAlgorithm}
	\begin{algorithmic}[1]
		\Require Initial guess for high-resolution multi-channel image $\IterationIdx{\HR}{0}$
		\Ensure Final high-resolution multi-channel image $\HR$ with \gls{llr} coefficients $\ChannelIdx{\vec{C}}{ij}$, $\ChannelIdx{\vec{b}}{ij}$ and $\ChannelIdx{\vec{K}}{ij}$
		\State $t \gets 1$
		\While{Convergence criterion in \eref{eqn:06_amTerminationCriterion} not fulfilled and $t \leq \NumIter[am]$}
			\For{$i = 1,\ldots,\NumChannels$}
				\For{$j = 1, \ldots, \NumChannels$}
					\State Estimate confidence weights $\IterationIdx{\ChannelIdx{\vec{K}}{ij}}{t}$ for \gls{llr} prior by \eref{eqn:weightingFunction}
					\State Estimate \gls{llr} coefficients $\IterationIdx{\ChannelIdx{\vec{C}}{ij}}{t}$ and $\IterationIdx{\ChannelIdx{\vec{b}}{ij}}{t}$ by \ereftwo{eqn:guidedFilterCoefficients_1}{eqn:guidedFilterCoefficients_2}
				\EndFor
			\EndFor
			\State $t_{\text{scg}} \gets 1$
			\While{Convergence criterion in \eref{eqn:06_amTerminationCriterion} not fulfilled and $t_{\text{scg}} \leq \NumIter[scg]$}
				\State Update high-resolution channels $\IterationIdx{\HR}{t}$ by SCG iteration for \eref{eqn:06_jointImageChannelEstimation}
				\State $t_{\text{scg}} \leftarrow t_{\text{scg}} + 1$
			\EndWhile
			\State $t \gets t + 1$
		\EndWhile
	\end{algorithmic}
\end{algorithm}

\section{Model and Algorithm Analysis}
\label{sec:06_AlgorithmAnalysis}

In this section, we present an in-depth analysis of the proposed Bayesian model along with the joint \gls{map} estimation approach. In particular, this covers a study regarding the adaptivity of the prior distribution as well as the computational complexity and the convergence of the algorithm. Eventually, a theoretical comparison of the Bayesian model to related state-of-the-art methods is presented. 

\subsection{Adaptivity of the Regression Model}
\label{sec:06_AdaptivityAndConvergence}

Unlike many of the closely related joint filters \cite{Kopf2007,Zhang2014,He2013}, the proposed inter-channel prior is robust against inconsistent structures among channels. Prominent examples for this issue appear in range imaging, where texture does not necessarily coincide with surface information, or in multispectral restoration with structural inconsistencies among the spectral bands \cite{Shen2015}. One essential property of the inter-channel prior is its adaptivity regarding these inconsistencies to avoid the erroneous transfer of structures from original channels to complementary reconstructed ones.

\Fref{fig:06_paramSensExample} investigates this adaptivity in the context of joint upsampling of single range and color images. In this simulated data example, the inter-channel prior is adopted in two different versions. On the one hand, we analyze the proposed regression based on Tukey's biweight loss that tolerates outliers related to inconsistent structures and is referred to as adaptive \gls{llr}. On the other hand, Tukey's biweight is replaced by the \LTwo norm, which leads to a simplified model that is referred to as non-adaptive \gls{llr}. The inter-channel regularization weight $\mu$ controls the impact of the regressions among range and color data to the upsampled channels\footnote{In this analysis, we limit ourselves to uniform inter-channel weights $\mu$ and \gls{llr} hyperparameter regularization weights $\epsilon$ for the regression between all pairs of channels.}. \Fref{fig:06_paramSensExample:llrNonAdapt_opt} and \fref{fig:06_paramSensExample:llrNonAdapt_overEst} demonstrates that the non-adaptive model is prone to texture-copying artifacts in case of an overestimated regularization weight $\mu$. Moreover, structural inconsistencies cause an oversmoothing of the color data. The adaptive regression depicted in \fref{fig:06_paramSensExample:llrAdapt_opt} and \fref{fig:06_paramSensExample:llrAdapt_overEst} features higher robustness against unwanted texture-copying. This behavior is quantitatively analyzed in \fref{fig:06_paramSensAnalysis} (left) by the \gls{psnr} of upsampled range and color data over a wide range of parameter settings. Here, the adaptive model features a lower sensitivity regarding an overestimation of the regularization weight. Note that the non-adaptive model might converge to a solution that is even inferior to a simple sequential upsampling, which is considered as the baseline. 

In addition, \fref{fig:06_paramSensAnalysis} (right) depicts the parameter sensitivity analysis regarding the hyperparameter regularization weight $\epsilon$. Notice that the adaptive model is stable over a wide range of parameter settings and outperforms the non-adaptive counterpart by a large margin.
\begin{figure}[!t]
	\centering
	\small
	\captionsetup[subfigure]{justification=centering}
	\subfloat{\includegraphics[width=0.187\textwidth]{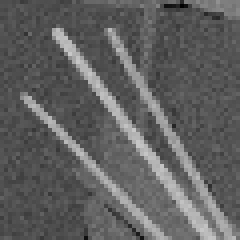}}\hspace{0.55em}
	\subfloat{\includegraphics[width=0.187\textwidth]{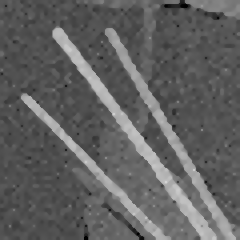}}\hspace{0.0001em}
	\subfloat{\includegraphics[width=0.187\textwidth]{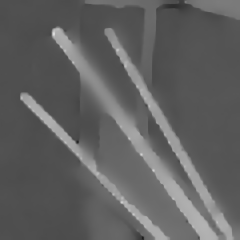}}\hspace{0.55em}
	\subfloat{\includegraphics[width=0.187\textwidth]{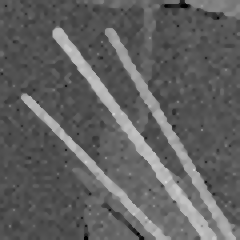}}\hspace{0.0001em}
	\subfloat{\includegraphics[width=0.187\textwidth]{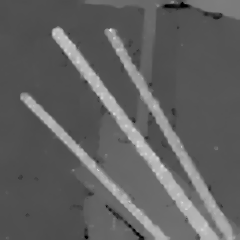}}\\[-0.7em]
	\setcounter{subfigure}{0}
	\subfloat[Original]{\includegraphics[width=0.187\textwidth]{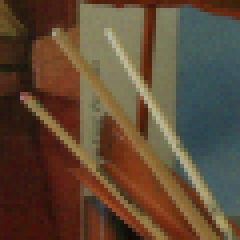}\label{fig:06_paramSensExample:orig}}\hspace{0.55em}
	\subfloat[Non-adaptive ($\mu = 0.01$)][Non-adaptive \\ ($\mu = 0.01$)]{\includegraphics[width=0.187\textwidth]{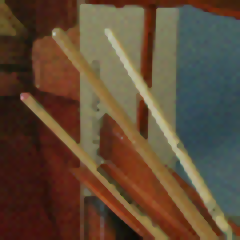}\label{fig:06_paramSensExample:llrNonAdapt_opt}}\hspace{0.0001em}
	\subfloat[Non-adaptive ($\mu = 1.00$)][Non-adaptive \\ ($\mu = 1.00$)]{\includegraphics[width=0.187\textwidth]{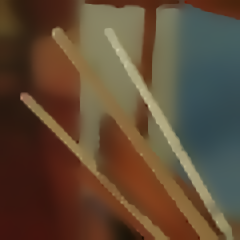}\label{fig:06_paramSensExample:llrNonAdapt_overEst}}\hspace{0.55em}
	\subfloat[Adaptive ($\mu = 0.01$)][Adaptive \\ ($\mu = 0.01$)]{\includegraphics[width=0.187\textwidth]{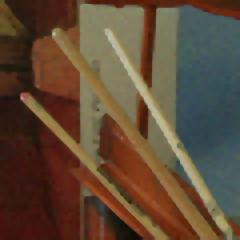}\label{fig:06_paramSensExample:llrAdapt_opt}}\hspace{0.0001em}
	\subfloat[Adaptive ($\mu = 1.00$)][Adaptive \\ ($\mu = 1.00$)]{\includegraphics[width=0.187\textwidth]{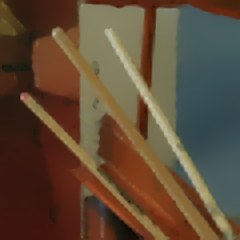}\label{fig:06_paramSensExample:llrAdapt_overEst}}\\
	\caption[\Gls{llr} prior for joint upsampling of range and color data]{Analysis of the \gls{llr} prior for joint upsampling of range and color data ($4 \times$ upsampling). \protect\subref{fig:06_paramSensExample:orig} Simulated range and color data. \protect\subref{fig:06_paramSensExample:llrNonAdapt_opt} and \protect\subref{fig:06_paramSensExample:llrNonAdapt_overEst} Upsampled range and color data using a non-adaptive version of the \gls{llr} prior based on the \LTwo norm with different inter-channel regularization weights $\mu$. \protect\subref{fig:06_paramSensExample:llrAdapt_opt} and \protect\subref{fig:06_paramSensExample:llrAdapt_overEst} Upsampled range and color data with the proposed adaptive \gls{llr} prior based on Tukey's biweight loss. Notice the texture-copying artifacts and the oversmoothing caused by the non-adaptive version of the prior.}
	\label{fig:06_paramSensExample}
\end{figure}
\begin{figure}[!t]
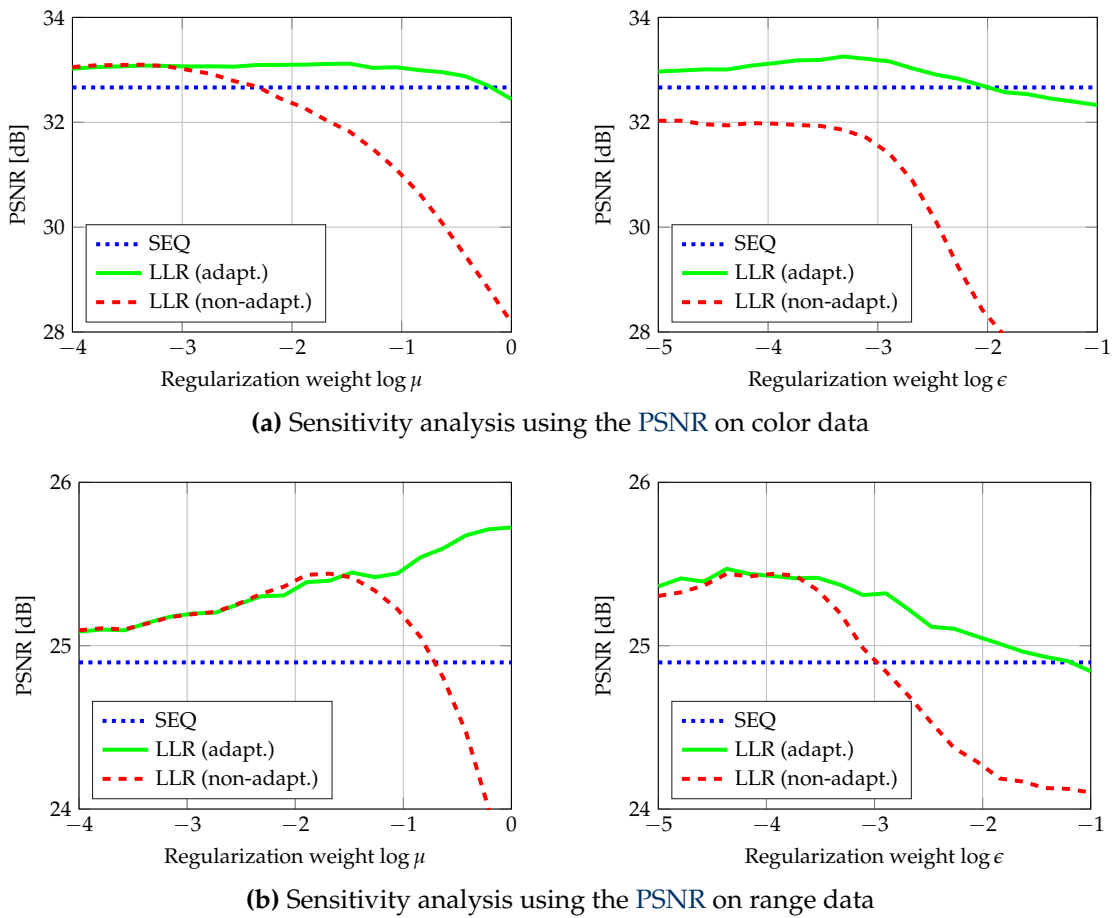

	\centering
	\scriptsize
	\setlength \figurewidth{0.38\textwidth}
	\setlength \figureheight{0.75\figurewidth}
	\subfloat{\input{images/chapter6/paramSens_interChannelWeight_color.tikz}}\quad\qquad
	\subfloat{\input{images/chapter6/paramSens_llrWeight_color.tikz}}\\
	\footnotesize
	\textbf{(a)} Sensitivity analysis using the \gls{psnr} on color data\\
	\scriptsize
	\subfloat{\input{images/chapter6/paramSens_interChannelWeight_range.tikz}}\quad\qquad
	\subfloat{\input{images/chapter6/paramSens_llrWeight_range.tikz}}\\
	\footnotesize
	\textbf{\footnotesize (b)} Sensitivity analysis using the \gls{psnr} on range data
	\caption[Sensitivity analysis of the \gls{llr} prior for joint range and color upsampling]{Sensitivity analysis of the \gls{llr} prior for joint range and color upsampling including a comparison of a non-adaptive version of the prior using the \LTwo norm to the adaptive one based on Tukey's biweight loss. Sequential upsampling of the channels (SEQ) is considered as the baseline. (a) Sensitivity regarding the inter-channel regularization weight $\mu$. (b) Sensitivity regarding the hyperparameter regularization weight $\epsilon$. Notice that adaptive \gls{llr} features a higher stability over a wider range of parameter settings.}
	\label{fig:06_paramSensAnalysis}
\end{figure}

\subsection{Computational Complexity and Convergence}
\label{sec:06_ComputationalComplexityAndConvergence}

The time complexity of \aref{alg:llrSuperResolutionAlgorithm} is related to two phases. First, given $C$ channels, the regression coefficients for $\NumChannels(\NumChannels-1)$ pairs of channels are computed according to \ereftwo{eqn:guidedFilterCoefficients_1}{eqn:guidedFilterCoefficients_2} at each iteration. This involves element-wise vector products as well as confidence-aware box filtering. Given $\HRSize$ pixels per channel, this can be implemented with time complexity $\BigO{\HRSize}$ for each pair by means of integral images and is independent on the regression patch size \cite{Crow1984,He2013}. Hence, the regression coefficient estimation has $\BigO{\NumChannels^2 \HRSize}$ time complexity. Second, the estimation of the high-resolution channels using \gls{scg} requires the computation of the joint energy function and its gradient based on \erefmore{eqn:06_imageChannelEstimationEnergyFunction}{eqn:06_imageChannelEstimationEnergyFunctionGrad}. Given $\NumFrames$ frames of size $\LRSize$ pixels, this can be implemented by sparse matrix-vector products as well as element-wise vector products and has $\BigO{\NumChannels \NumFrames \LRSize + \NumChannels^2 \HRSize}$ time complexity at each inner iteration. For $\NumIter[scg]$ iterations for \gls{scg} in the inner optimization loop, the overall time complexity of a single alternating minimization iteration in \aref{alg:llrSuperResolutionAlgorithm} is $\BigO{\NumChannels^2 \HRSize + \NumIter[scg] (\NumChannels \NumFrames \LRSize + \NumChannels^2 \HRSize)}$. 

In addition to the complexity, let us also investigate the convergence of \aref{alg:llrSuperResolutionAlgorithm} experimentally. In \fref{fig:06_convergence}, the convergence is analyzed for the joint range and color upsampling example in \fref{fig:06_paramSensExample}. This depicts the progress of the \gls{psnr} of the range data as well as the sum of absolute differences between the channels for successive iterations. Furthermore, we show a comparison regarding the influence of the initial guess to the solution of the underlying non-convex energy minimization problem. For this purpose, let us compare the proposed initialization provided by a sequential reconstruction without inter-channel prior to bicubic and nearest-neighbor interpolations. While the former provides a more accurate starting point, the latter are easy to compute. Albeit different initializations are used, alternating minimization converges to comparable solutions. We found that $\NumIter[am] = 10$ iterations for alternating minimization with $\NumIter[scg] = 10$ iterations for \gls{scg} are typically sufficient for convergence.
\begin{figure}[!t]
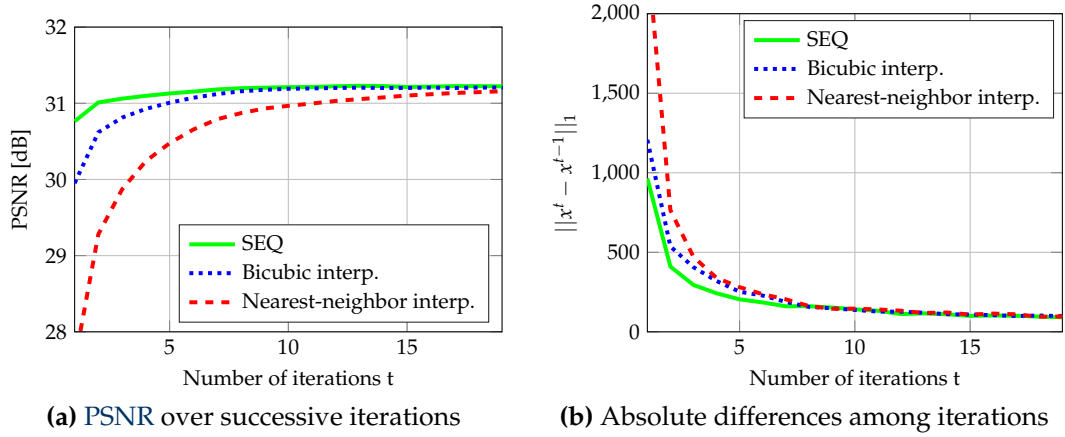

	\centering
	\scriptsize
	\setlength \figurewidth{0.37\textwidth}
	\setlength \figureheight{0.75\figurewidth} 
	\subfloat[\gls{psnr} over successive iterations]{\input{images/chapter6/convergence_psnr.tikz}}\qquad
	\subfloat[Absolute differences among iterations]{\input{images/chapter6/convergence_sad.tikz}}
	\caption[Convergence analysis of alternating minimization for multi-channel super-resolution]{Convergence analysis of alternating minimization for multi-channel super-resolution. We compared different initializations including a sequential reconstruction of the channels (SEQ) as well as channel-wise bicubic and nearest-neighbor interpolation.}
	\label{fig:06_convergence}
\end{figure}

\subsection{Connection to Related Methods}

The Bayesian model introduced in \sref{sec:06_BayesianModelingOfMultiChannelImages} features a combination of two complementary paradigms of image filtering and restoration. On the one hand, this includes local filtering (\eg \cite{Kopf2007,Zhang2014}), where the goal is to obtain a filter output image from an input by considering relationships between both in local patches. Global methods (\eg \cite{Ham2015,Shen2015}) on the other hand aim at an implicit reconstruction of a filter output by optimizing global energy functions. For the sake of notational brevity, let us study the relationship to these paradigms for $\NumChannels = 2$ channels. Here, a single iteration of alternating minimization aims at optimizing the joint energy function:
\begin{equation}
	\label{eqn:06_localGlobalFormulation}
	\begin{split}
		\EnergyFunSym(\HRChannel{1}, \HRChannel{2}, \LLRParams) &= 
		\underbrace{\LossFun[\LRSym] { \LRChannel{1} - \SystemMatChannel{1} \HRChannel{1} }
			+ \LossFun[\LRSym] { \LRChannel{2} - \SystemMatChannel{2} \HRChannel{2} }
			+ \ChannelIdx{\RegWeight}{1} \LossFun[\HRSym]{ \HRChannel{1} } + \ChannelIdx{\RegWeight}{2} \LossFun[\HRSym] { \HRChannel{2} } }
			_{\text{global term}~\EnergyFunSym_{\text{global}}(\HRChannel{1}, \HRChannel{2})}\\
		&+ \underbrace{
			\ChannelIdx{\mu}{12} \LossFun[\ChannelIdx{\HRSym}{12}]{ \ChannelIdx{\vec{C}}{12} \HRChannel{1} + \ChannelIdx{\vec{b}}{12} - \HRChannel{2} }
			+ \ChannelIdx{\mu}{21} \LossFun[\ChannelIdx{\HRSym}{21}] { \ChannelIdx{\vec{C}}{21} \HRChannel{2} + \ChannelIdx{\vec{b}}{21} - \HRChannel{1} }}
			_{\text{local term}~\EnergyFunSym_{\text{local}}(\HRChannel{1}, \HRChannel{2}, \LLRParams)}\\
		&+ \underbrace{\ChannelIdx{\epsilon}{12} \big| \big| \Diag{\ChannelIdx{\vec{C}}{12}} \big| \big|_2^2 
		+ \ChannelIdx{\epsilon}{21} \big| \big| \Diag{\ChannelIdx{\vec{C}}{21}} \big| \big|_2^2}
			_{\text{hyperparameter regularization}}.
	\end{split}
\end{equation}
The observation model and the intra-channel prior are related to the global energy $\EnergyFunSym_{\text{global}}(\HRChannel{1}, \HRChannel{2})$, while the inter-channel prior forms the local energy $\EnergyFunSym_{\text{local}}(\HRChannel{1}, \HRChannel{2}, \LLRParams)$. This local term appears as an additional regularizer in this inverse problem. Similar formulations appear in two closely related filtering techniques.

\paragraph{Relation to Guided Filtering.}
The mixed local/global formulation in \eref{eqn:06_localGlobalFormulation} provides a generalization of the well known guided filter \cite{He2013} as a prominent example for a local operator. For this consideration, let $\ChannelIdx{\RegWeight}{1} = \ChannelIdx{\RegWeight}{2} = 0$ and let us drop the global data fidelity term. Moreover, let $\ChannelIdx{\mu}{21} = \ChannelIdx{\epsilon}{21} = 0$. Then, the optimization of \eref{eqn:06_localGlobalFormulation} can be simplified to the minimization of the energy function:    
\begin{equation}
	\EnergyFunSym_{\text{GF}}(\HRChannel{1}, \HRChannel{2}, \LLRParams)
	= \LossFun[\ChannelIdx{\HRSym}{12}]{ \ChannelIdx{\vec{C}}{12} \HRChannel{1} + \ChannelIdx{\vec{b}}{12} - \HRChannel{2} } 
	+ \ChannelIdx{\epsilon}{12} \big| \big| \Diag{\ChannelIdx{\vec{C}}{12}} \big| \big|_2^2.
\end{equation}

If we keep the channel $\HRChannel{1}$ fixed, optimizing $\EnergyFunSym_{\text{GF}}(\HRChannel{1}, \HRChannel{2}, \LLRParams)$ is a generalized version of guided filtering for $\HRChannel{2}$ under the guidance of $\HRChannel{1}$. Compared to conventional guided filtering, the proposed algorithm provides several valuable extensions. First and foremost, the local model adopts the robust loss function $\LossFun[\ChannelIdx{\HRSym}{ij}]{z}$ to establish linear regressions, while guided filtering relies on simple least-squares estimation making it prone to outliers. This extension is sensible in case of inconsistent structures, see \sref{sec:06_AdaptivityAndConvergence}. In addition, the proposed formulation couples local filtering with a global model. This offers the flexibility to model the image formation for multi-frame super-resolution in contrast to a purely local filtering that might exhibit halo artifacts \cite{He2013}. Similar relations applies in comparison to other local filters, \eg joint bilateral filtering \cite{Kopf2007}.

\paragraph{Relation to Mutual Structure Filtering.}
\eref{eqn:06_localGlobalFormulation} also provides a generalization of mutual structure filtering as proposed by Shen \etal \cite{Shen2015}. To outline this relationship, let us assume that the system matrices of the individual channels are given by the identity, \ie $\SystemMatChannel{1} = \SystemMatChannel{2} = \Id$. In addition, let us drop the intra-channel prior, \ie $\ChannelIdx{\RegWeight}{1} = \ChannelIdx{\RegWeight}{2} = 0$. Then, \eref{eqn:06_localGlobalFormulation} can be simplified to: 
\begin{equation}
	\begin{split}
		\EnergyFunSym_{\text{MS}}(\HRChannel{1}, \HRChannel{2}, \LLRParams) &= 
		\LossFun[\LRSym] { \LRChannel{1} - \HRChannel{1} }
			+ \LossFun[\LRSym] {\LRChannel{2} - \HRChannel{2} } \\
		&+
			\ChannelIdx{\mu}{12} \LossFun[\ChannelIdx{\HRSym}{12}] { \ChannelIdx{\vec{C}}{12} \HRChannel{1} + \ChannelIdx{\vec{b}}{12} - \HRChannel{2} }
			+ \ChannelIdx{\mu}{21} \LossFun[\ChannelIdx{\HRSym}{21}] { \ChannelIdx{\vec{C}}{21} \HRChannel{2} + \ChannelIdx{\vec{b}}{21} - \HRChannel{1} }\\
		 &+ 
			\ChannelIdx{\epsilon}{12} \big| \big| \Diag{\ChannelIdx{\vec{C}}{12}} \big| \big|_2^2 + \ChannelIdx{\epsilon}{21} \big| \big| \Diag{\ChannelIdx{\vec{C}}{21}} \big| \big|_2^2.
	\end{split}
\end{equation}

Joint minimization of $\EnergyFunSym_{\text{MS}}(\HRChannel{1}, \HRChannel{2}, \LLRParams)$ \wrt the regression coefficients $\LLRParams$ and the channels $\HRChannel{1}$ and $\HRChannel{2}$ is conceptually equivalent to mutual structure filtering. However, notice that the algorithm in \cite{Shen2015} considers a denoising problem ($\SystemMatChannel{1} = \SystemMatChannel{2} = \Id$) and employs the \LTwo norm for $\LossFun[\HRSym]{z}$ and $\LossFun[\ChannelIdx{\HRSym}{ij}]{z}$. These simplifications make joint minimization efficient to compute, since closed-form solutions can be derived for the latent image channels and the regression coefficients. In contrast to this approach, the proposed algorithm enables multi-frame super-resolution.   

\section{Experiments and Results}
\label{sec:06_ExperimentsAndResults}

This section presents a detailed experimental evaluation of the proposed multi-channel super-resolution. Several applications that are of great interest in computer vision are being investigated. The main focus of this study lies on resolution enhancement for color and 3-D range images as two classical applications. In addition, further experiments including multispectral image upsampling as well as joint segmentation and super-resolution are presented.

\subsection{Applications in Color Imaging}
\label{sec:06_ApplicationsInColorImaging}

In terms of color image resolution enhancement, we aim at simultaneously super-resolving color channels in the RGB space. For this purpose, the \gls{llr} model is adopted to exploit dependencies among the spectral bands that are common in case of natural images \cite{Omer2004}. The proposed algorithm was compared to the following approaches to color super-resolution. As a baseline approach, a sequential super-resolution of the color channels that served as initial guess for multi-channel super-resolution was evaluated. This is conceptually equivalent to the algorithm introduced by K\"ohler \etal \cite{Kohler2015c} for single-channel images. In addition, multi-channel super-resolution was evaluated using the \gls{idp} proposed by Farsiu \etal \cite{Farsiu2006} as an alternative to the proposed prior\footnote{Chrominance and luminance regularization as well as demosaicing as proposed in \cite{Farsiu2006} were omitted to exclusively evaluate the influence of the inter-channel regularization.}. This regularization term penalizes mismatches in terms of the location and orientation of edges among the color channels. In the sequel, experiments on simulated and real color image sequences are presented.

\paragraph{Simulated Data.}
In order to conduct a quantitative evaluation, simulated color image sequences ($\NumFrames = 10$ frames) with randomly generated rigid motion were generated from the LIVE Database \cite{Sheikh2016}. This simulation comprises a Gaussian \gls{psf} ($\PSFWidth = 0.5$), subsampling ($\MagFac = 3$) as well as additive Gaussian noise ($\NoiseStd = 0.03$) for each color channel. Throughout these experiments, the exact subpixel motion was used for super-resolution to explicitly study the impact of the different prior models. The inter-channel regularization weights $\ChannelIdx{\mu}{ij}$ as well as the hyperparameter weights $\ChannelIdx{\epsilon}{ij}$ were defined in a symmetric way ($\ChannelIdx{\mu}{ij} = \ChannelIdx{\mu}{ji}$ and $\ChannelIdx{\epsilon}{ij} = \ChannelIdx{\epsilon}{ji}$). The intra-channel prior parameters were set to $\ChannelIdx{\RegWeight}{i} = 4 \cdot 10^{-3}$ with \gls{btv} window size $\BTVSize = 1$ and weighting factor $\BTVWeight = 0.5$ for each channel. The inter-channel parameters were set to $\ChannelIdx{\mu}{ij} = 0.5$ and $\ChannelIdx{\epsilon}{ij} = 10^{-4}$ with \gls{llr} patch size $\LLRPatchSize = 3$ for all pairs of channels.
\begin{figure}[!t]
	\centering
	\small
	\mbox{
	\centering
	\hspace{-0.9em}
	\subfloat[Original]{
		\begin{tikzpicture}[spy using outlines={rectangle,red,magnification=2.5, height=1.8cm, width = 3.65cm, connect spies, every spy on node/.append style={thick}}] 
			\node {\pgfimage[width=0.242\linewidth]{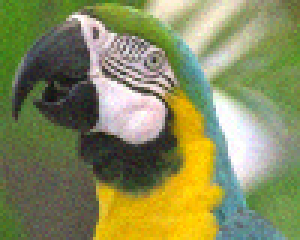}}; 
      \spy on (0.5, 0.8) in node [left] at (1.82, -2.42); 
    \end{tikzpicture}
	}
	\hspace{-1.1em}
	\subfloat[Sequential]{
		\begin{tikzpicture}[spy using outlines={rectangle,red,magnification=2.5, height=1.8cm, width = 3.65cm, connect spies, every spy on node/.append style={thick}}] 
			\node {\pgfimage[width=0.242\linewidth]{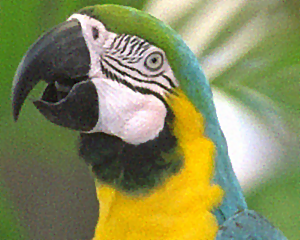}}; 
      \spy on (0.5, 0.8) in node [left] at (1.82, -2.42); 
    \end{tikzpicture}
	}
	\hspace{-1.1em}
	\subfloat[\gls{idp} \cite{Farsiu2006}]{
		\begin{tikzpicture}[spy using outlines={rectangle,red,magnification=2.5, height=1.8cm, width = 3.65cm, connect spies, every spy on node/.append style={thick}}] 
			\node {\pgfimage[width=0.242\linewidth]{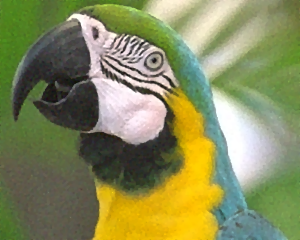}}; 
      \spy on (0.5, 0.8) in node [left] at (1.82, -2.42); 
    \end{tikzpicture}
	}
	\hspace{-1.1em}
	\subfloat[\gls{llr}]{
		\begin{tikzpicture}[spy using outlines={rectangle,red,magnification=2.5, height=1.8cm, width = 3.65cm, connect spies, every spy on node/.append style={thick}}] 
			\node {\pgfimage[width=0.242\linewidth]{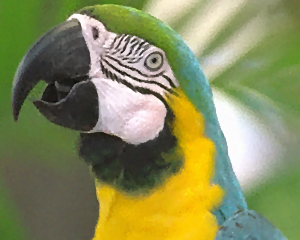}}; 
      \spy on (0.5, 0.8) in node [left] at (1.82, -2.42); 
    \end{tikzpicture}
	}
	}
	\caption[Color super-resolution on simulated data]{Color super-resolution ($\NumFrames = 10$ frames, magnification $\MagFac = 3$) on simulated data with a comparison of a sequential reconstruction of the color channels to a multi-channel reconstruction using \gls{idp} regularization \cite{Farsiu2006} and the \gls{llr} prior.}
	\label{fig:06_liveDatabaseExample}
\end{figure}

\Fref{fig:06_liveDatabaseExample} compares sequential super-resolution of the individual color channels to multi-channel super-resolution. In contrast to the sequential approach, both multi-channel methods avoided inconsistencies between the super-resolved color channels. This resulted in lower noise levels in homogeneous areas while the sequential approach caused color artifacts in these regions. In a quantitative comparison on 20 simulated datasets, the proposed method achieved the highest \gls{psnr} and \gls{ssim} measures, see \fref{fig:06_liveDatabaseErrorMeasures}. On average, compared to the sequential and the \gls{idp} super-resolution, the proposed algorithm improved the \gls{psnr} (\gls{ssim}) by $1.5$\,\gls{db} (0.04) and $0.5$\,\gls{db} (0.01), respectively.

\begin{figure}[!t]
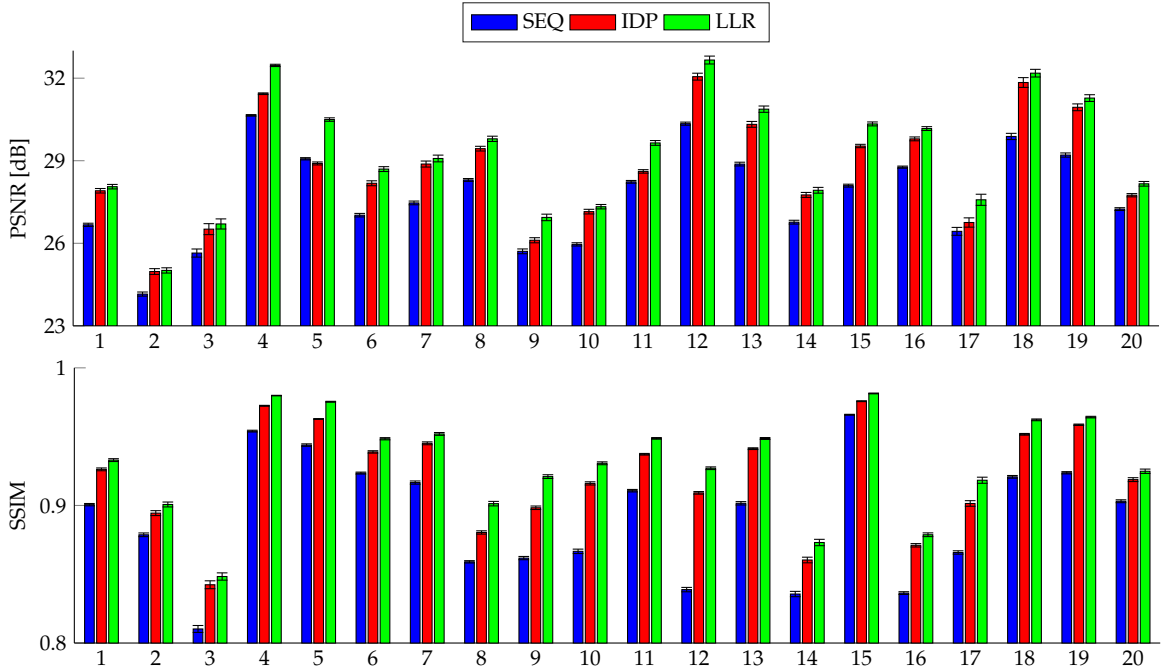

	\centering
	\scriptsize 
	\setlength \figurewidth{0.94\textwidth}
	\setlength \figureheight{0.285\figurewidth} 
	\subfloat{\input{images/chapter6/liveDatabasePSNR.tikz}}\\[-0.5ex]
	\subfloat{\input{images/chapter6/liveDatabaseSSIM.tikz}}
	\caption[\gls{psnr} and \gls{ssim} of super-resolution on simulated color images]{Mean $\pm$ standard deviation of the \gls{psnr} and the \gls{ssim} of 20 simulated color image datasets obtained from the LIVE database \cite{Sheikh2016}. For each dataset, 15 randomly simulated image sequences were generated to determine the quality measure statistics. This benchmark compares sequential super-resolution of the color channels (SEQ) to multi-channel super-resolution using \gls{idp} regularization \cite{Farsiu2006} and the \gls{llr} prior.}
	\label{fig:06_liveDatabaseErrorMeasures}
\end{figure}

\paragraph{Real Data.}
\Fref{fig:06_mdspDatabaseExample} shows a qualitative comparison of the competing methods on the \textit{Bookcase} sequence ($\NumFrames = 30$ frames) taken from the MDSP benchmark dataset \cite{Farsiu2014}. Each pixel in these color images reflects full RGB information to make additional demosaicing unnecessary. Super-resolution was conducted with magnification $\MagFac = 3$ and a Gaussian \gls{psf} ($\PSFWidth = 0.4$). The unknown subpixel motion was described by an affine model and estimated by \gls{ecc} optimization \cite{Evangelidis2008}. Throughout these experiments, we set the intra-channel prior parameters to $\ChannelIdx{\RegWeight}{i} = 5 \cdot 10^{-3}$, $\BTVSize = 1$ and $\BTVWeight = 0.5$ for each channel. The inter-channel parameters were set to $\ChannelIdx{\mu}{ij} = 20$, $\ChannelIdx{\epsilon}{ij} = 10^{-4}$ and $\LLRPatchSize = 1$ for all pairs of channels.
 
Similar to the previous experiments, multi-channel super-resolution gets rid of color artifacts that appeared in the super-resolved color channels obtained by the sequential approach. Examples regarding these artifacts are jagged edges and color bleeding as shown in the highlighted region that contains text. These artifacts were better compensated by multi-channel super-resolution using the \gls{llr} prior.
\begin{figure}[!t]
	\centering
	\small
	\mbox{
	\centering
	\hspace{-0.9em}
	\subfloat[Original]{
		\begin{tikzpicture}[spy using outlines={rectangle,red,magnification=2.5, height=1.8cm, width = 3.65cm, connect spies, every spy on node/.append style={thick}}] 
			\node {\pgfimage[width=0.242\linewidth]{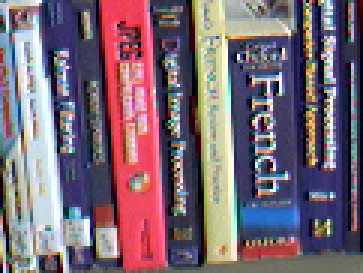}}; 
      \spy on (0.5, 0.55) in node [left] at (1.82, -2.32); 
    \end{tikzpicture}
	}
	\hspace{-1.1em}
	\subfloat[Sequential]{
		\begin{tikzpicture}[spy using outlines={rectangle,red,magnification=2.5, height=1.8cm, width = 3.65cm, connect spies, every spy on node/.append style={thick}}] 
			\node {\pgfimage[width=0.242\linewidth]{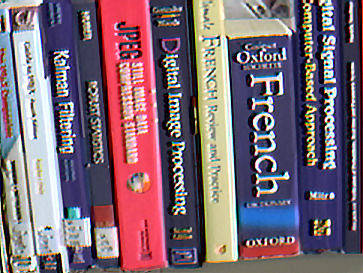}}; 
      \spy on (0.5, 0.55) in node [left] at (1.82, -2.32); 
    \end{tikzpicture}
	}
	\hspace{-1.1em}
	\subfloat[\gls{idp} \cite{Farsiu2006}]{
		\begin{tikzpicture}[spy using outlines={rectangle,red,magnification=2.5, height=1.8cm, width = 3.65cm, connect spies, every spy on node/.append style={thick}}] 
			\node {\pgfimage[width=0.242\linewidth]{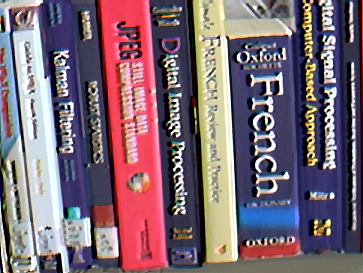}}; 
      \spy on (0.5, 0.55) in node [left] at (1.82, -2.32); 
    \end{tikzpicture}
	}
	\hspace{-1.1em}
	\subfloat[\gls{llr}]{
		\begin{tikzpicture}[spy using outlines={rectangle,red,magnification=2.5, height=1.8cm, width = 3.65cm, connect spies, every spy on node/.append style={thick}}] 
			\node {\pgfimage[width=0.242\linewidth]{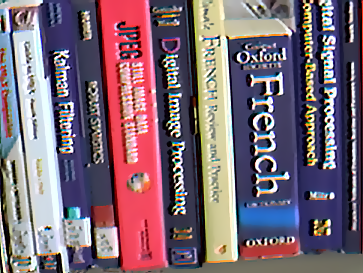}}; 
      \spy on (0.5, 0.55) in node [left] at (1.82, -2.32); 
    \end{tikzpicture}
	}
	}
	\caption[Color super-resolution on the \textit{Bookcase} sequence]{Color super-resolution ($\NumFrames = 30$ frames, magnification $s = 3$) on the \textit{Bookcase} sequence \cite{Farsiu2014} with a comparison of sequential color channel reconstruction to multi-channel reconstructions using \gls{idp} regularization \cite{Farsiu2006} and the \gls{llr} prior.}
	\label{fig:06_mdspDatabaseExample}
\end{figure}

\subsection{Applications in Range Imaging}
\label{sec:06_ApplicationsInRangeImaging}

In \cref{sec:MultiSensorSuperResolutionUsingGuidanceImages}, super-resolution for 3-D range data was investigated by exploiting color images as a static guidance. The multi-channel technique presented in this chapter is applicable to similar setups but does not rely on the existence of accurate guidance data. The following experiments consider single-image upsampling as well as multi-frame super-resolution in the context of range imaging.

\paragraph{Joint RGB-D Upsampling.}
In the domain of single image joint RGB-D upsampling, we aim at simultaneously upsampling $\NumChannels = 4$ channels (RGB color plus depth) from their low-resolution counterparts. This is a highly underdetermined reconstruction problem but exploiting mutual dependencies among range and color data serves as a strong prior to alleviate this issue. 

For the sake of a quantitative evaluation, artificial RGB-D images were obtained from the ground truth color and disparity data of the Middlebury Stereo Datasets \cite{Hirschmuller2007,Scharstein2007}. The formation of low-resolution images considered the conditions of low-cost \gls{tof} sensors and comprises a Gaussian \gls{psf} ($\PSFWidth = 0.3$) and subsampling ($\MagFac = 8$) relative to the ground truth. These degradations were jointly applied to all channels to consider the absence of reliable guidance information. In addition, color and range channels were corrupted by additive Gaussian noise with standard deviations $0.02$ and $0.04$, respectively. We adopted the \gls{llr} prior with a symmetric distribution for joint multi-channel upsampling. The inter-channel parameters were set to $\ChannelIdx{\mu}{ij} = 1.5$, \smash{$\ChannelIdx{\epsilon}{ij} = 10^{-4}$} and $\LLRPatchSize = 5$ for all pairs of channels. For the intra-channel prior a \gls{btv} window size of $\BTVSize = 2$ with weighting factor $\BTVWeight = 0.5$ was used. The intra-channel regularization weights were set to \smash{$\ChannelIdx{\RegWeight}{i} = 5 \cdot 10^{-4}$} for the color channels ($i \in \{1, 2, 3\}$) and \smash{$\ChannelIdx{\RegWeight}{i} = 10^{-2}$} for the range channel ($i = 4$) to reflect the noise levels of both modalities.

\Fref{fig:06_middleburyRangeExample} depicts a comparison of the proposed joint multi-channel upsampling against various state-of-the-art filters for color guided range upsampling using publicly available reference implementations. This includes the guided filter (GF) \cite{He2013}, the weighted median filter (WMF) \cite{Zhang2014}, the mutual structure filter (MSF) \cite{Shen2015} as well as the static and dynamic guidance filter (SDF) \cite{Ham2015}. Among them, GF and WMF employ color data as a static guidance that was obtained by bicubic upsampling of the input color image. As a consequence, these methods gained a limited quality of the upsampled range data due to the absence of reliable color data. In addition to a pure static guidance, MSF and SDF also exploit range data regularization similar to the proposed model. Although these approaches were less sensitive to erroneous texture-copying since structural inconsistencies could be considered, they do not incorporate an appropriate image formation model. Thus, they were inherently limited for upsampling of structures that were lost in low-resolution range \textit{and} color data. Opposed to the state-of-the-art, the proposed method simultaneously upsampled all channels under a reasonable image formation model and profited from enhanced color data for the task of range upsampling. This is depicted in \fref{fig:06_middleburyColorExample} showing the original and bicubic interpolation of the color channels in comparison to the color image reconstructed as a by-product of joint multi-channel upsampling.

In \tref{tab:06_middleburyErrorMeasures}, a benchmark on various Middlebury datasets is summarized. For fair comparisons, the parameters of the different joint image filters were adjusted to each dataset individually to optimize the \gls{psnr}, while the proposed joint multi-channel upsampling was employed with the aforementioned default parameters. In these experiments, joint multi-channel upsampling based on \gls{llr} considerably outperformed a simple sequential upsampling of the channels. Moreover, the joint upsampling considerably improved the accuracy of range information compared to the state-of-the-art filters on most of the evaluated datasets.
\begin{figure}[!t] 
	\centering 
  \subfloat[Range input]{\label{fig:06_middleburyRangeExample:lr}
		\begin{tikzpicture}[spy using outlines={rectangle,red,magnification=3.5,height=1.7cm, width = 1.7cm, connect spies, every spy on node/.append style={thick}}] 
			\node {\pgfimage[width=0.237\linewidth]{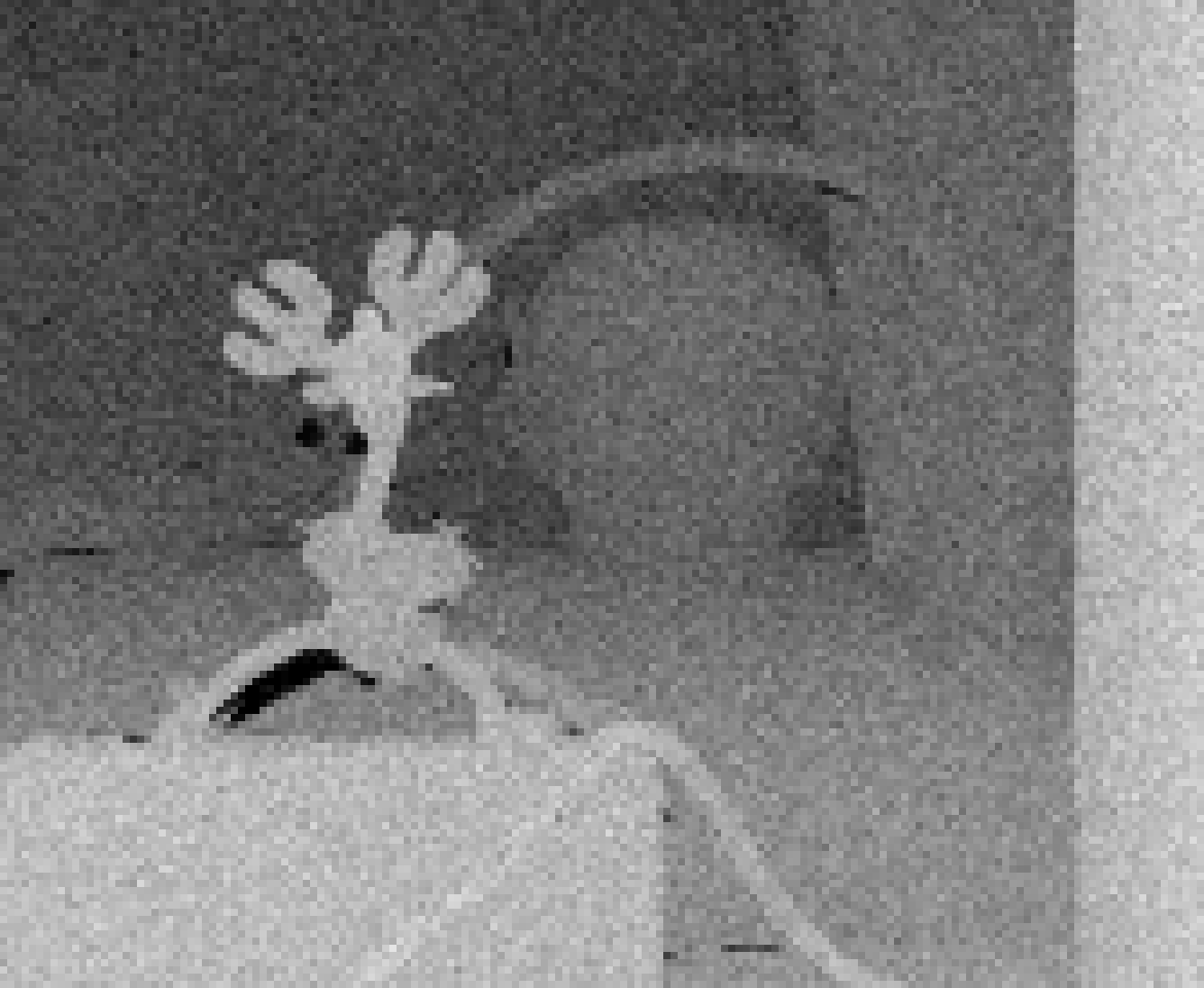}}; 
      \spy on (-0.4,0.5) in node [left] at (-0.1,-2.35); 
			\spy[green] on (0.15,-0.7) in node [left] at (1.8,-2.35); 
    \end{tikzpicture} 
	}\hspace{-1.2em}
	\subfloat[GF \cite{He2013}]{\label{fig:06_middleburyRangeExample:gf}
		\begin{tikzpicture}[spy using outlines={rectangle,red,magnification=3.5,height=1.7cm, width = 1.7cm, connect spies, every spy on node/.append style={thick}}]
			\node {\pgfimage[width=0.237\linewidth]{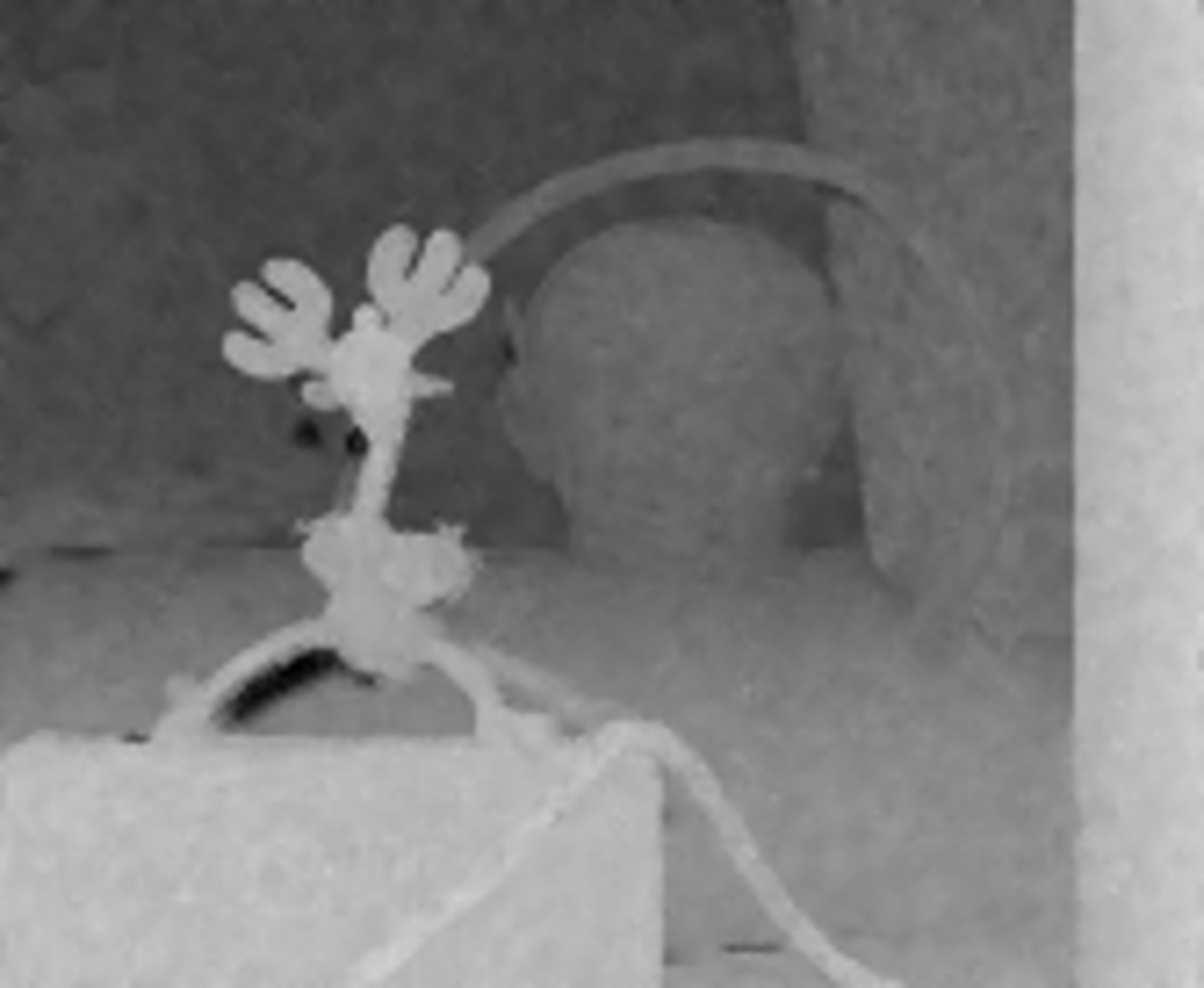}}; 
			\spy on (-0.4,0.5) in node [left] at (-0.1,-2.35);
			\spy[green] on (0.15,-0.7) in node [left] at (1.8,-2.35);
		\end{tikzpicture} 
	}\hspace{-1.2em}
  \subfloat[WMF \cite{Zhang2014}]{\label{fig:06_middleburyRangeExample:wmf}
		\begin{tikzpicture}[spy using outlines={rectangle,red,magnification=3.5,height=1.7cm, width = 1.7cm, connect spies, every spy on node/.append style={thick}}] 
			\node {\pgfimage[width=0.237\linewidth]{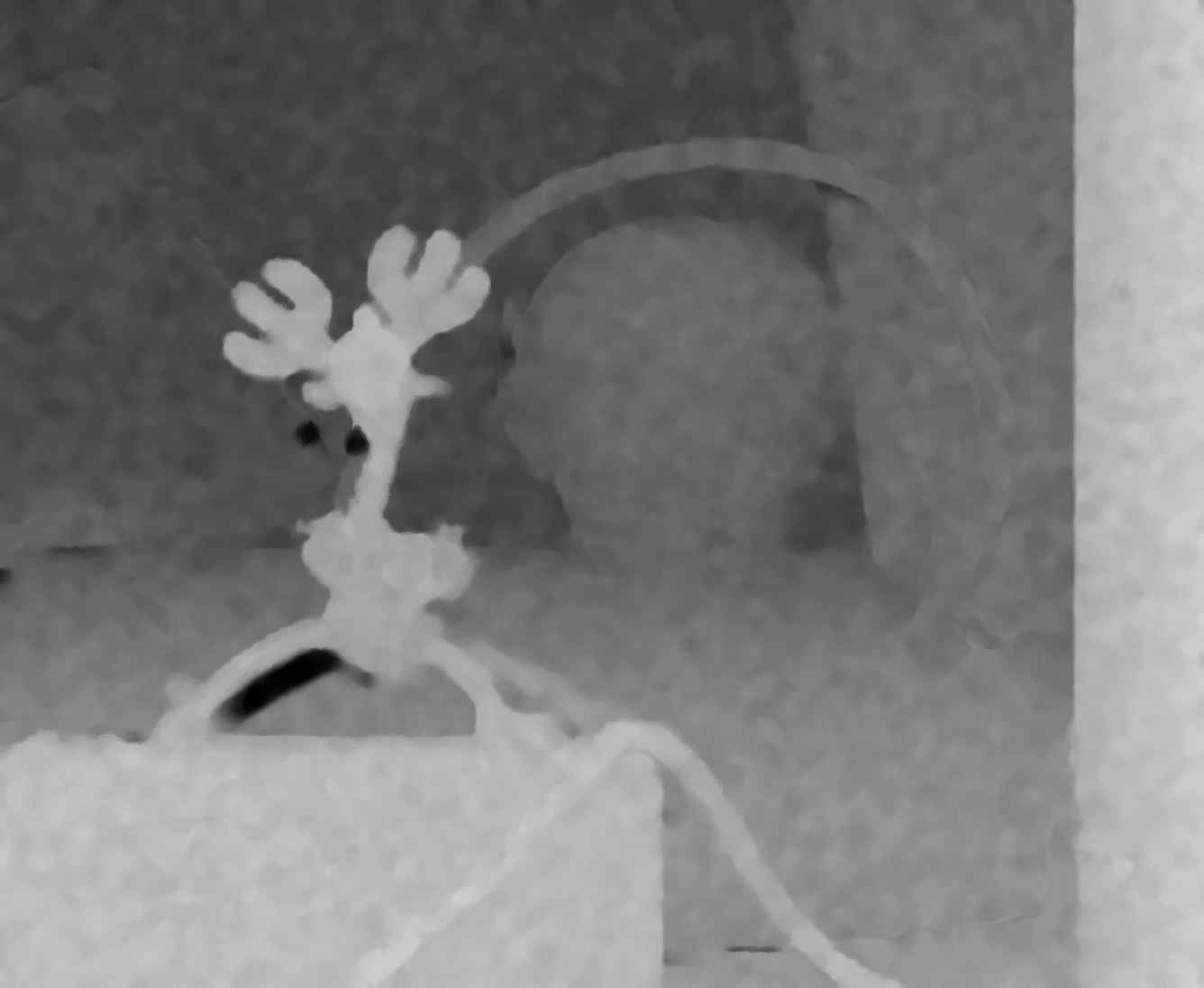}}; 
      \spy on (-0.4,0.5) in node [left] at (-0.1,-2.35);
			\spy[green] on (0.15,-0.7) in node [left] at (1.8,-2.35);
    \end{tikzpicture} 
	}\hspace{-1.2em}
  \subfloat[MSF \cite{Shen2015}]{\label{fig:06_middleburyRangeExample:msf}
		\begin{tikzpicture}[spy using outlines={rectangle,red,magnification=3.5,height=1.7cm, width = 1.7cm, connect spies, every spy on node/.append style={thick}}] 
			\node {\pgfimage[width=0.237\linewidth]{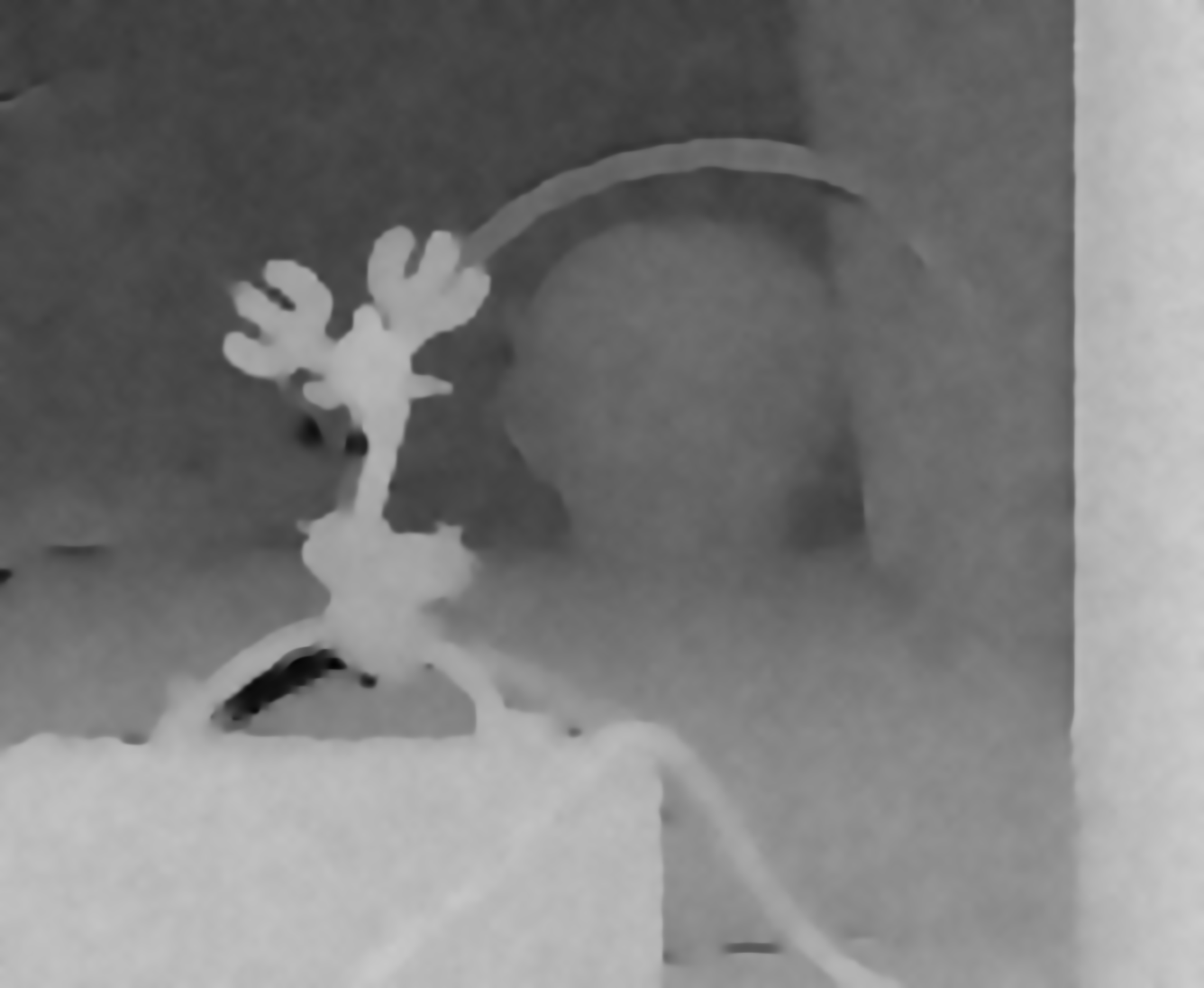}}; 
      \spy on (-0.4,0.5) in node [left] at (-0.1,-2.35);
			\spy[green] on (0.15,-0.7) in node [left] at (1.8,-2.35);
    \end{tikzpicture} 
  }\\
	\subfloat[SDF \cite{Ham2015}]{\label{fig:06_middleburyRangeExample:sdf}
		\begin{tikzpicture}[spy using outlines={rectangle,red,magnification=3.5,height=1.7cm, width = 1.7cm, connect spies, every spy on node/.append style={thick}}]
			\node {\pgfimage[width=0.237\linewidth]{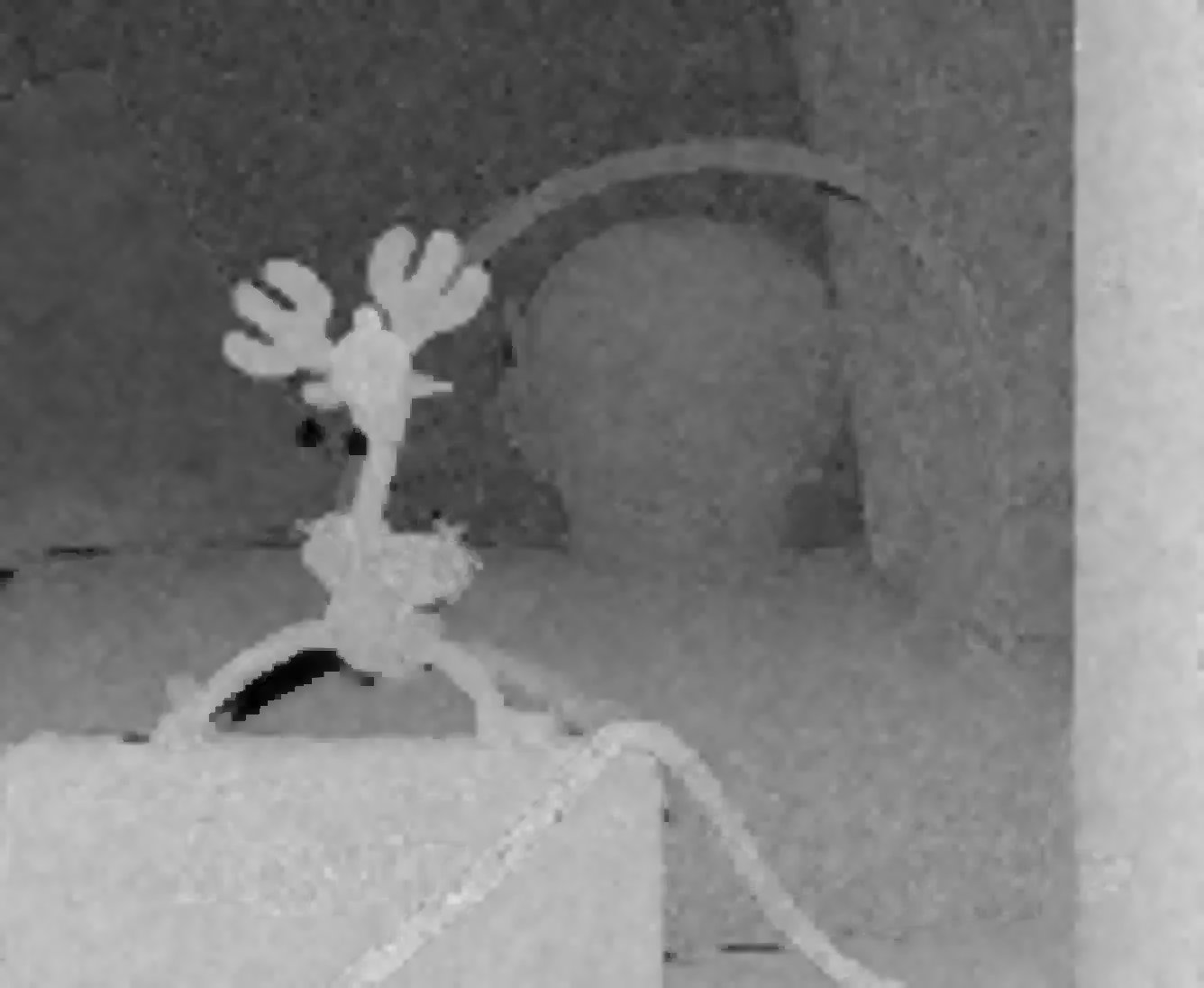}}; 
      \spy on (-0.4,0.5) in node [left] at (-0.1,-2.35);
			\spy[green] on (0.15,-0.7) in node [left] at (1.8,-2.35);
     \end{tikzpicture} 
	}\hspace{-1.2em}
  \subfloat[Sequential]{\label{fig:06_middleburyRangeExample:seq}
		\begin{tikzpicture}[spy using outlines={rectangle,red,magnification=3.5,height=1.7cm, width = 1.7cm, connect spies, every spy on node/.append style={thick}}] 
			\node {\pgfimage[width=0.237\linewidth]{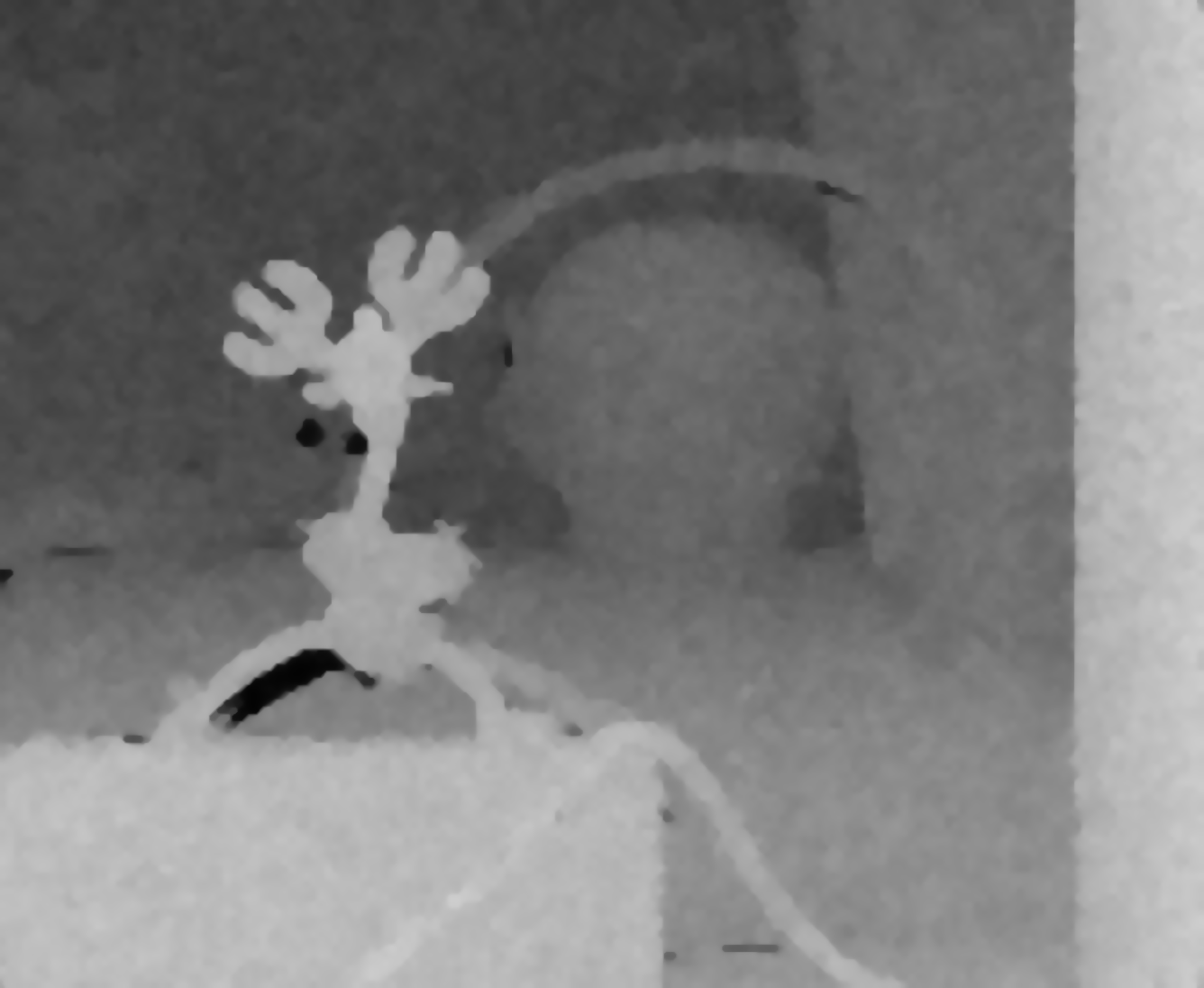}}; 
      \spy on (-0.4,0.5) in node [left] at (-0.1,-2.35);
			\spy[green] on (0.15,-0.7) in node [left] at (1.8,-2.35);
    \end{tikzpicture} 
  }\hspace{-1.2em}
  \subfloat[\gls{llr}]{\label{fig:06_middleburyRangeExample:llr}
		\begin{tikzpicture}[spy using outlines={rectangle,red,magnification=3.5,height=1.7cm, width = 1.7cm, connect spies, every spy on node/.append style={thick}}]
			\node {\pgfimage[width=0.237\linewidth]{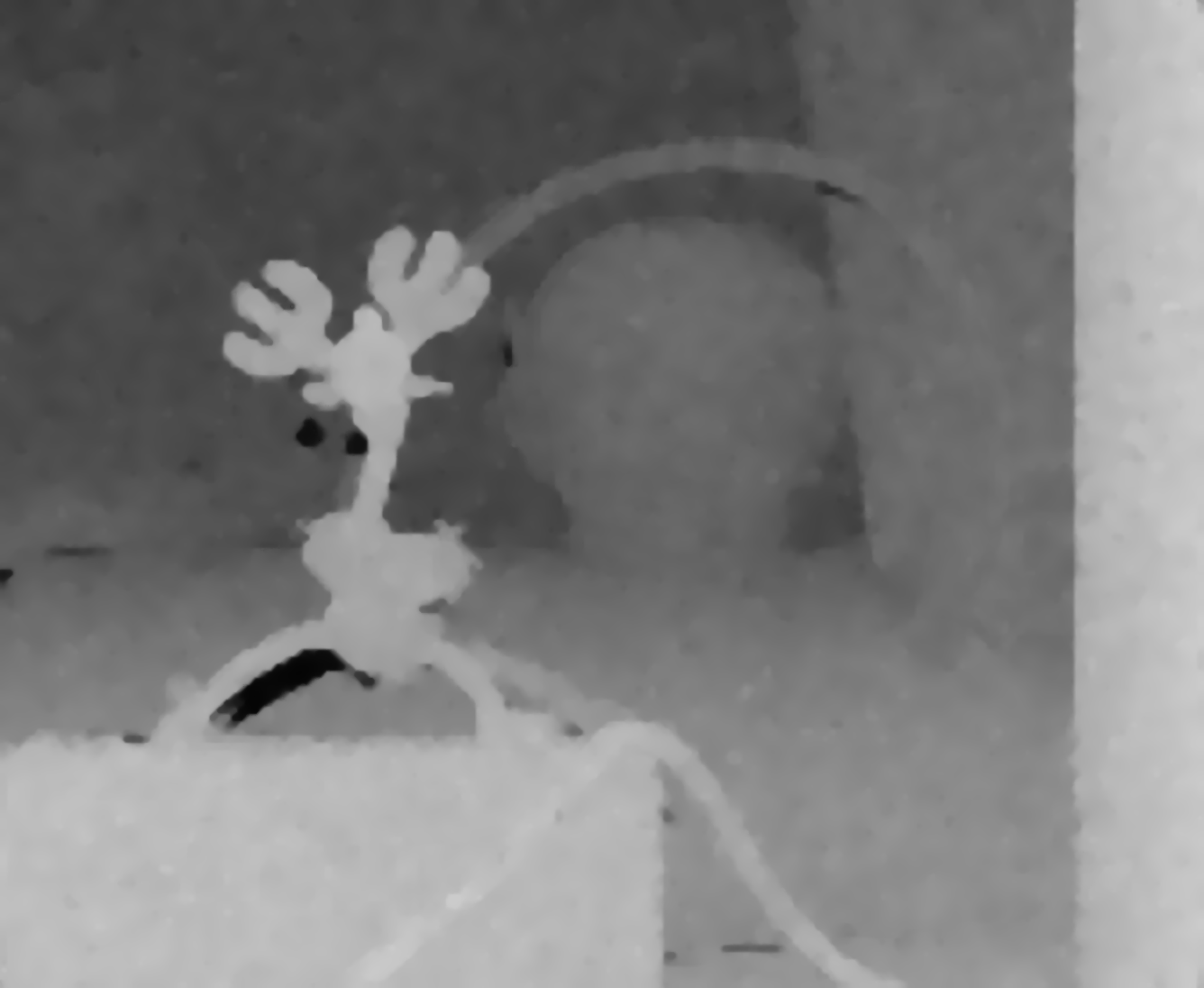}}; 
      \spy on (-0.4,0.5) in node [left] at (-0.1,-2.35);
			\spy[green] on (0.15,-0.7) in node [left] at (1.8,-2.35);
    \end{tikzpicture} 
  }\hspace{-1.2em}
  \subfloat[Ground truth]{\label{fig:06_middleburyRangeExample:gt}
		\begin{tikzpicture}[spy using outlines={rectangle,red,magnification=3.5,height=1.7cm, width = 1.7cm, connect spies, every spy on node/.append style={thick}}] 
			\node {\pgfimage[width=0.237\linewidth]{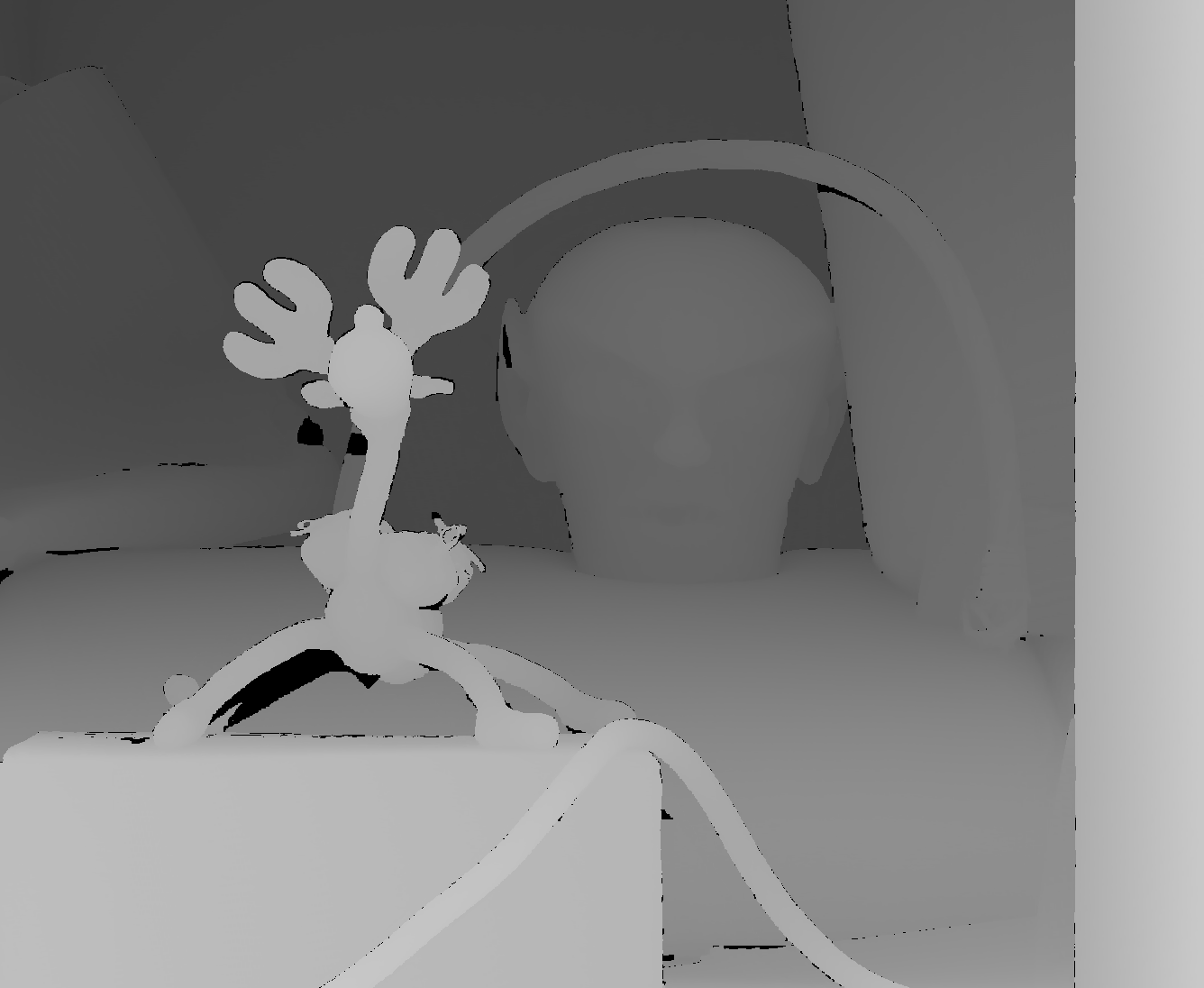}}; 
			\spy on (-0.4,0.5) in node [left] at (-0.1,-2.35);
			\spy[green] on (0.15,-0.7) in node [left] at (1.8,-2.35);
    \end{tikzpicture} 
  }
  \caption[Range images from RGB-D upsampling on the Middlebury Stereo Datasets]{Single image RGB-D upsampling (magnification $\MagFac = 8$) on the Middlebury Stereo Datasets \cite{Hirschmuller2007,Scharstein2007} with visual comparison of the upsampled range images. \protect\subref{fig:06_middleburyRangeExample:lr} Low-resolution range image, \protect\subref{fig:06_middleburyRangeExample:gf} guided filter \cite{He2013}, \protect\subref{fig:06_middleburyRangeExample:wmf} weighted median filter \cite{Zhang2014}, \protect\subref{fig:06_middleburyRangeExample:msf} mutual structure filter \cite{Shen2015}, \protect\subref{fig:06_middleburyRangeExample:sdf} static and dynamic guidance filter \cite{Ham2015}, \protect\subref{fig:06_middleburyRangeExample:seq} sequential upsampling of range and color channels without inter-channel prior, \protect\subref{fig:06_middleburyRangeExample:llr} multi-channel upsampling using the \gls{llr} prior, \protect\subref{fig:06_middleburyRangeExample:gt} ground truth range data.} 
  \label{fig:06_middleburyRangeExample} 
\end{figure}

\begin{figure}[!t]
	\centering
	\subfloat[Color input]{\label{fig:06_middleburyColorExample:lr}
		\begin{tikzpicture}[spy using outlines={rectangle,red,magnification=3.5,height=1.7cm, width = 1.7cm, connect spies, every spy on node/.append style={thick}}] 
			\node {\pgfimage[width=0.237\linewidth]{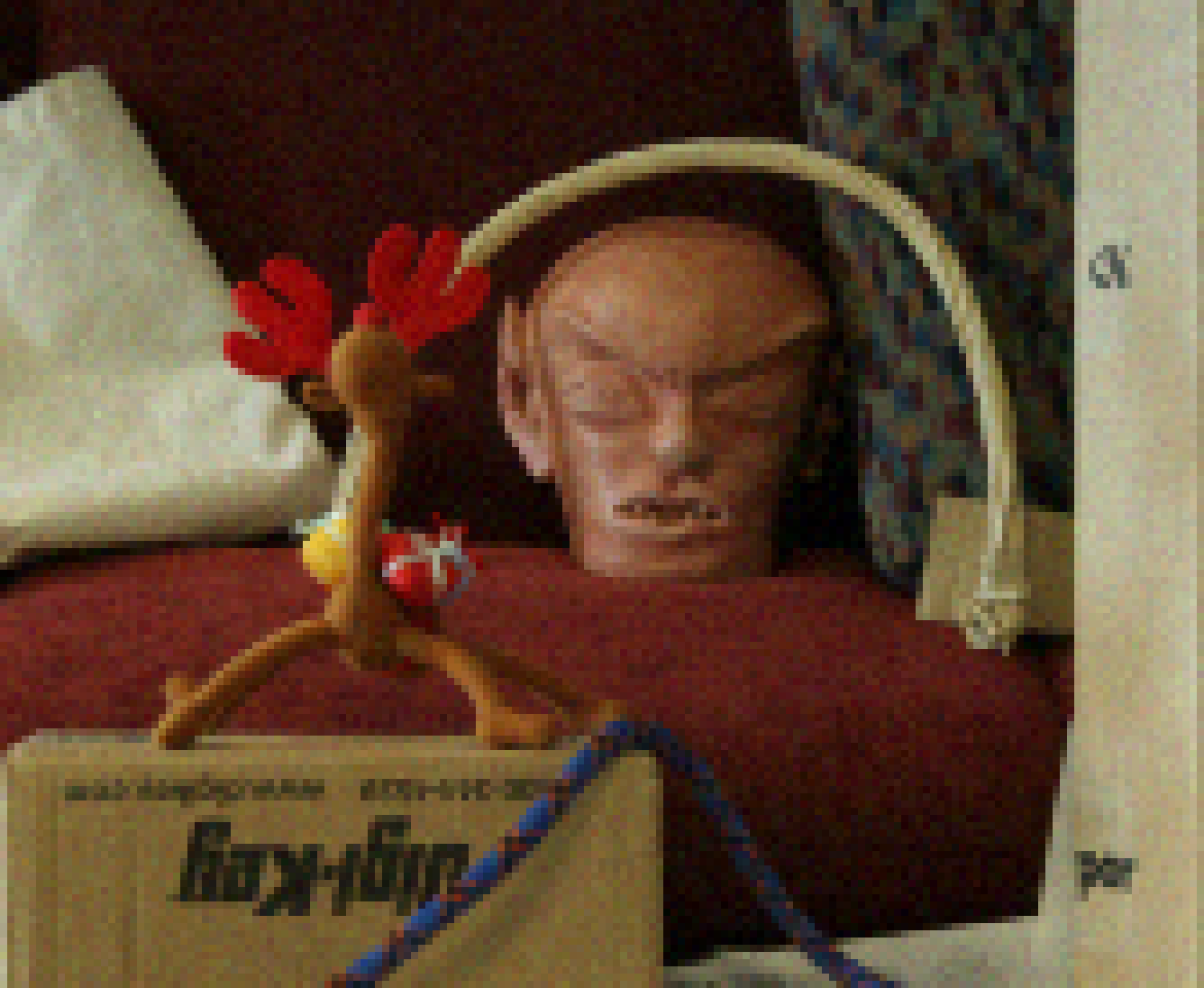}}; 
      \spy on (-0.4,0.5) in node [left] at (-0.1,-2.35);
			\spy[green] on (0.15,-0.7) in node [left] at (1.8,-2.35);
    \end{tikzpicture} 
	}\hspace{-1.2em}
	\subfloat[Bicubic interp.]{\label{fig:06_middleburyColorExample:bc}
		\begin{tikzpicture}[spy using outlines={rectangle,red,magnification=3.5,height=1.7cm, width = 1.7cm, connect spies, every spy on node/.append style={thick}}] 
			\node {\pgfimage[width=0.237\linewidth]{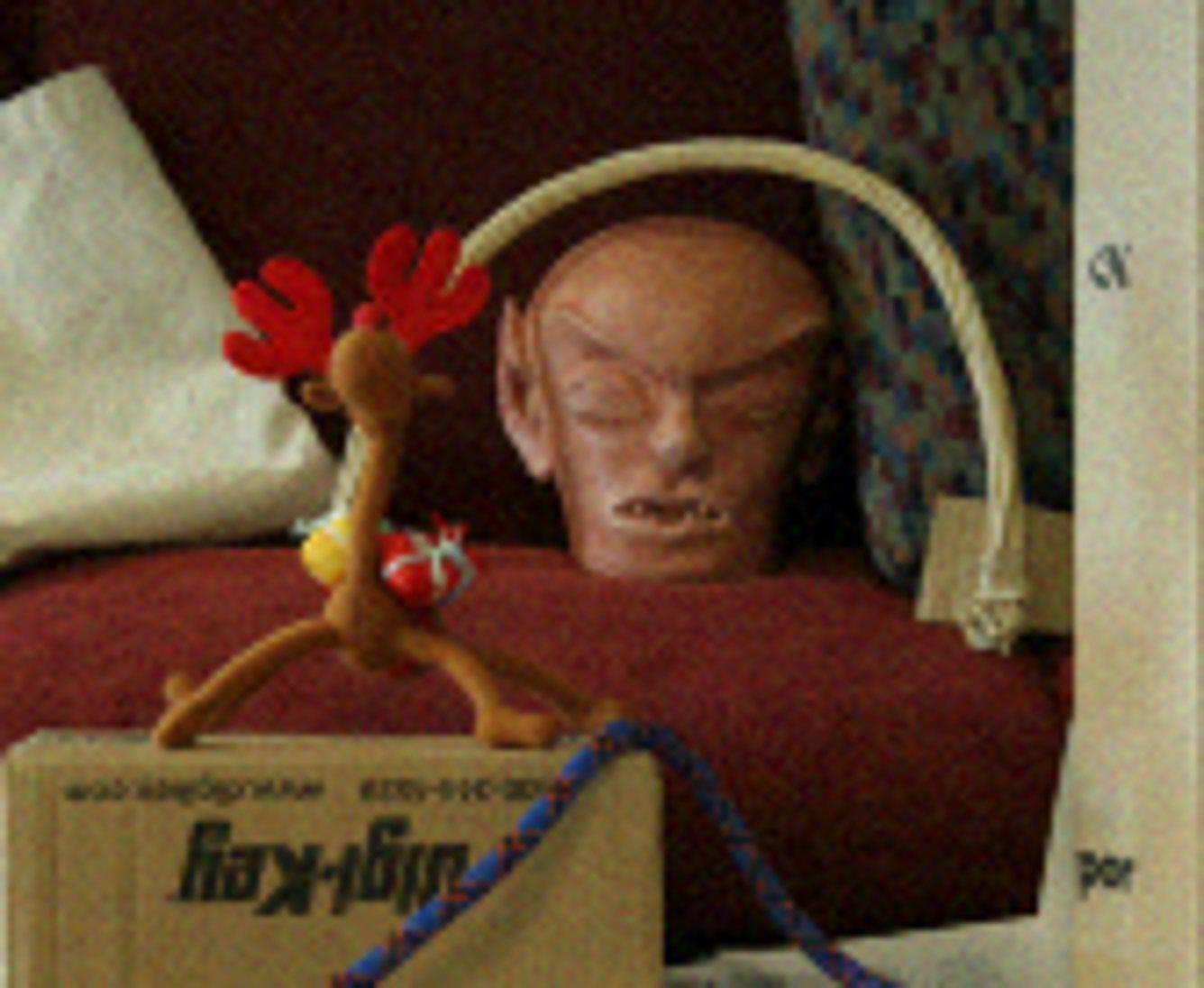}}; 
      \spy on (-0.4,0.5) in node [left] at (-0.1,-2.35);
			\spy[green] on (0.15,-0.7) in node [left] at (1.8,-2.35);
    \end{tikzpicture} 
	}\hspace{-1.2em}
	\subfloat[\gls{llr}]{\label{fig:06_middleburyColorExample:llr}
		\begin{tikzpicture}[spy using outlines={rectangle,red,magnification=3.5,height=1.7cm, width = 1.7cm, connect spies, every spy on node/.append style={thick}}] 
			\node {\pgfimage[width=0.237\linewidth]{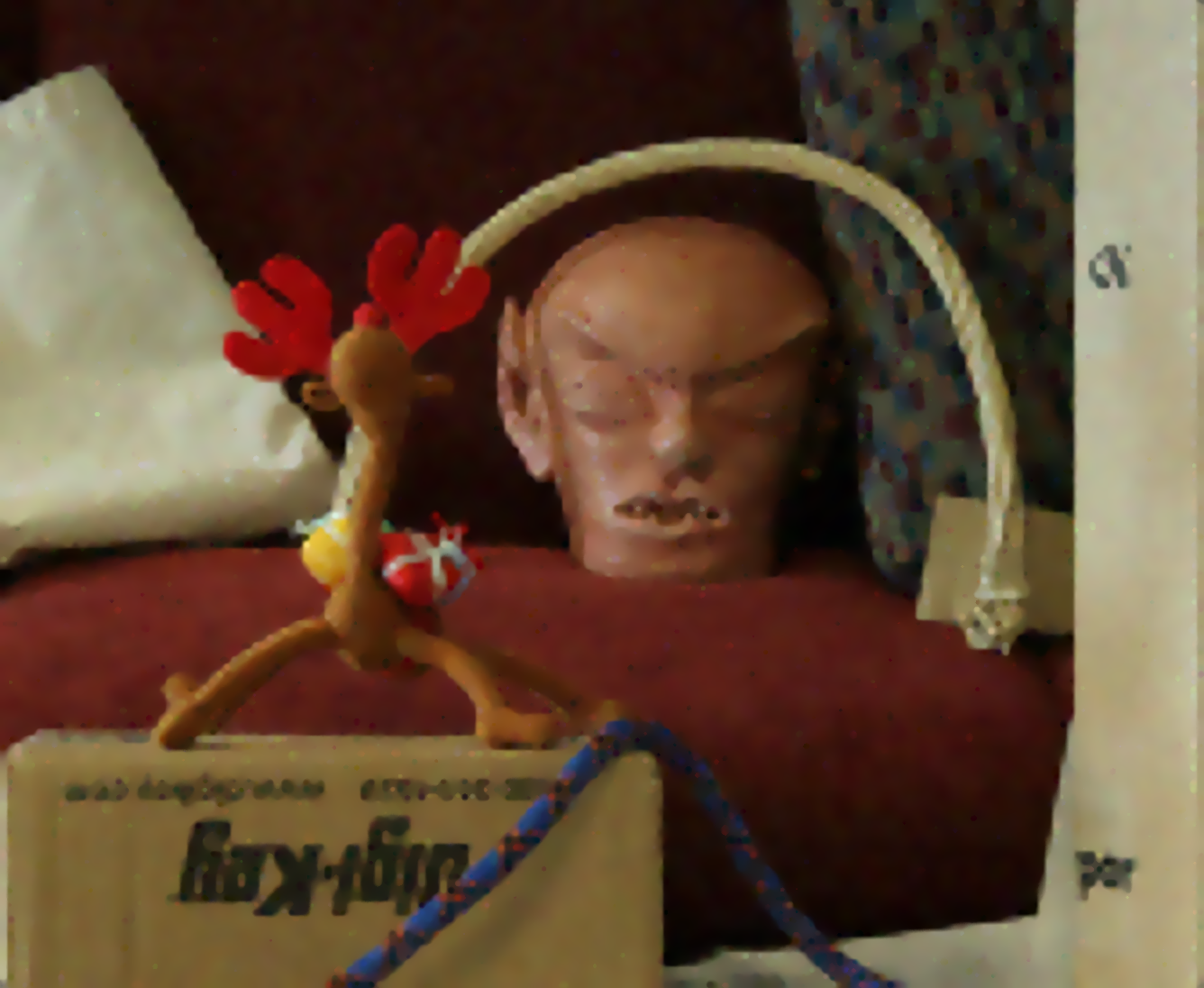}}; 
      \spy on (-0.4,0.5) in node [left] at (-0.1,-2.35);
			\spy[green] on (0.15,-0.7) in node [left] at (1.8,-2.35);
    \end{tikzpicture} 
	}\hspace{-1.2em}
	\subfloat[Ground truth]{\label{fig:06_middleburyColorExample:gt}
		\begin{tikzpicture}[spy using outlines={rectangle,red,magnification=3.5,height=1.7cm, width = 1.7cm, connect spies, every spy on node/.append style={thick}}] 
			\node {\pgfimage[width=0.237\linewidth]{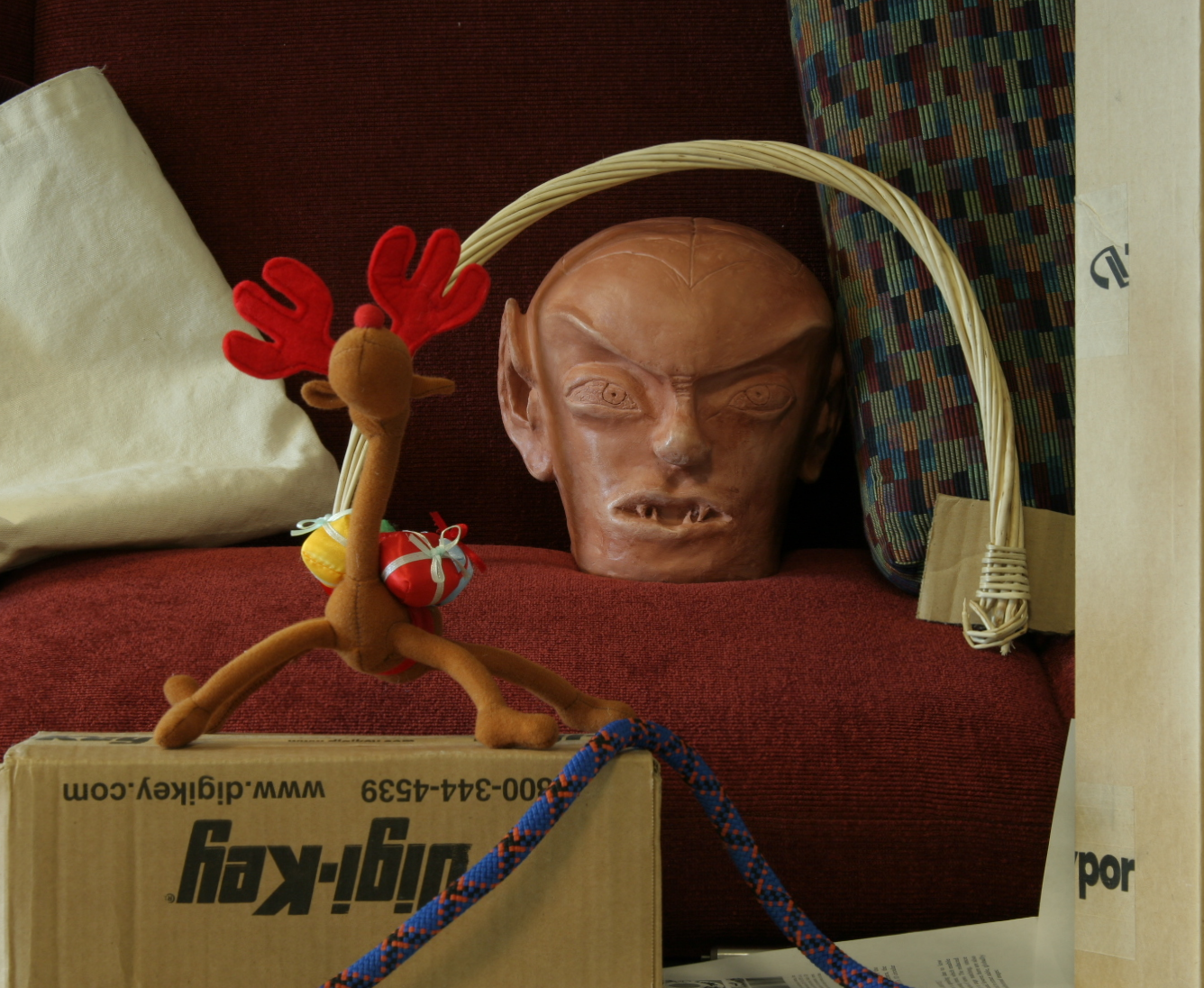}}; 
      \spy on (-0.4,0.5) in node [left] at (-0.1,-2.35);
			\spy[green] on (0.15,-0.7) in node [left] at (1.8,-2.35);
    \end{tikzpicture} 
	}
	\caption[Color images from RGB-D upsampling on the Middlebury Stereo Datasets]{Single image RGB-D upsampling on the Middlebury Stereo Datasets \cite{Hirschmuller2007,Scharstein2007} with visual comparison of the upsampled color data. \protect\subref{fig:06_middleburyColorExample:lr} Low-resolution image, \protect\subref{fig:06_middleburyColorExample:bc} bicubic upsampling, \protect\subref{fig:06_middleburyColorExample:llr} upsampling using the \gls{llr} prior, \protect\subref{fig:06_middleburyColorExample:gt} ground truth color image.}
	\label{fig:06_middleburyColorExample}
\end{figure}

\begin{table}[!t]
	\caption[Quantitative evaluation of joint RGB-D image upsampling on the Middlebury Stereo Datasets]{Quantitative evaluation of joint RGB-D image upsampling on the Middlebury Stereo Datasets \cite{Hirschmuller2007,Scharstein2007} ($8 \times$ upsampling). We compared the \gls{psnr} of upsampled range data using different joint image filters (GF \cite{He2013}, WMF \cite{Zhang2014}, MSF \cite{Shen2015}, SDF \cite{Ham2015}) to multi-channel upsampling using a sequential (channel-wise) approach as well as the proposed \gls{llr} prior. All joint filters used the bicubic upsampled color images as static guidance. For each dataset, the best and the second best results are highlighted.}
	\centering
	\small
	\begin{tabular}{lcccccc}
		\toprule
																											& \textit{Art} 											& \textit{Books} 								& \textit{Dolls} 									& \textit{Laundry} 							& \textit{Moebius} 						& \textit{Reindeer} \\
		\cmidrule{2-7}
		\textbf{Interpolation} 	& & & & & & \\
		~~Nearest-neighbor	& 26.04 & 26.44 & 26.90	& 26.93 & 26.80 & 26.75 \\
		~~Bicubic 					& 27.82 & 27.99 & 28.56 & 28.62 & 28.42 & 28.46 \\
		\midrule
		\textbf{Joint filters} 	& & & & & & \\
		~~GF \cite{He2013} 			& 30.57 & 32.69 & 34.46 & 33.25 & 33.48 & 32.55 \\
		~~WMF \cite{Zhang2014}	& \textbf{30.81}$^{)2}$	& \textbf{32.93}$^{)2}$	& 34.32 & 33.47	& \textbf{33.55}$^{)2}$ & 32.94 \\
		~~MSF \cite{Shen2015} 	& 30.43 & \textbf{33.14}$^{)1}$	& \textbf{34.50}$^{)2}$	& \textbf{33.49}$^{)2}$ & 33.50	& \textbf{33.10}$^{)2}$ \\
		~~SDF \cite{Ham2015} 		& 30.43 & 31.80 & 33.23 & 33.33 & 32.85 & 32.09 \\
		\midrule
		\textbf{Multi-channel} 	& & & & & &\\
		~~Sequential						& 30.71 & 32.36 & 33.98 & 33.30 & 33.23 & 32.96 \\
		~~\gls{llr} 						& \textbf{30.93}$^{)1}$ & 32.75 & \textbf{34.56}$^{)1}$ & \textbf{33.81}$^{)1}$	& \textbf{33.72}$^{)1}$	& \textbf{33.42}$^{)1}$ \\
		\bottomrule
	\end{tabular}
	\label{tab:06_middleburyErrorMeasures}
\end{table}

\paragraph{Photogeometric Super-Resolution.}
Contrary to related joint filters, the proposed model can be directly applied to gain resolution enhancement of range and color data from a multi-frame perspective referred to as \textit{photogeometric super-resolution} \cite{Ghesu2014}. We investigate this methodology for current \gls{tof} cameras that provide geometric information by 3-D range images along with photometric information encoded by amplitude images at the same pixel resolution at a video frame rate. \Fref{fig:06_mesaSRResults} and \fref{fig:06_mesaSRResults2} depict super-resolution on range and amplitude data of two datasets captured with a Mesa SR-4000 camera\footnote{The data acquisition for this study was done in collaboration with Peter Fürsattel at Metrilus GmbH, Erlangen, Germany.}. Both channels were acquired at a resolution of $176 \times 144$\,px and subpixel motion across the sequences was induced by camera translations. Motion estimation was implemented via variational optical flow \cite{Liu2009} on the amplitude images. We used the \gls{llr} prior for $\NumChannels = 2$ channels with symmetric distribution. The inter-channel parameters were set to $\ChannelIdx{\mu}{ij} = 10^2$, \smash{$\ChannelIdx{\epsilon}{ij} = 10^{-4}$} and $\LLRPatchSize = 1$ for all pairs of channels. The intra-channel parameters were set to  \smash{$\ChannelIdx{\RegWeight}{i} = 5 \cdot 10^{-4}$} for the amplitude data ($i = 1$) and \smash{$\ChannelIdx{\RegWeight}{i} = 2 \cdot 10^{-3}$} for the range data ($i = 2$) with $\BTVSize = 1$ and $\BTVWeight = 0.5$ for both channels. 

We compared the proposed technique against \acrfull{amsr} as presented in \sref{sec:05_ApplicationToHybridRangeImaging}. \Gls{amsr} was applied to both channels separately using the amplitude data as static guidance for its feature-based regularizer. In addition, we evaluated sequential super-resolution of both channels under the proposed model without inter-channel prior.

Photogeometric super-resolution simultaneously enhanced geometric and photometric information. In comparison to a sequential reconstruction of both channels, exploiting mutual dependencies between range and amplitude data boosted range super-resolution even further. The \gls{llr} prior could also better capture such dependencies compared to \gls{amsr} that used low-resolution amplitude data directly to steer the underlying regularization technique. Notice that in this setup, amplitude images do not necessarily meet the quality requirements for a static guidance. This improvement of the range regularization resulted in superior reconstructions of depth discontinuities and surfaces.

\begin{figure}[!p] 
	\centering 
  \subfloat{\label{fig:06_tofSRResults:amp_lr}
		\begin{tikzpicture}[spy using outlines={rectangle,red,magnification=2.3,height=3.4cm, width = 2.4cm, connect spies, every spy on node/.append style={thick}}] 
			\node {\pgfimage[width=0.32\linewidth]{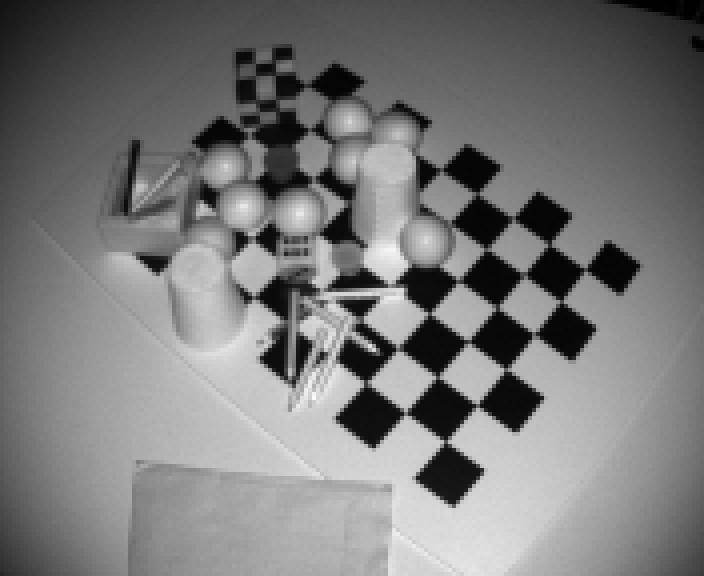}}; 
      \spy on (-0.4,-0.2) in node [left] at (4.88, 0); 
    \end{tikzpicture}
	}
	\subfloat{\label{fig:06_tofSRResults:amp_amsr}
		\begin{tikzpicture}[spy using outlines={rectangle,red,magnification=2.3,height=3.4cm, width = 2.4cm, connect spies, every spy on node/.append style={thick}}] 
			\node {\pgfimage[width=0.32\linewidth]{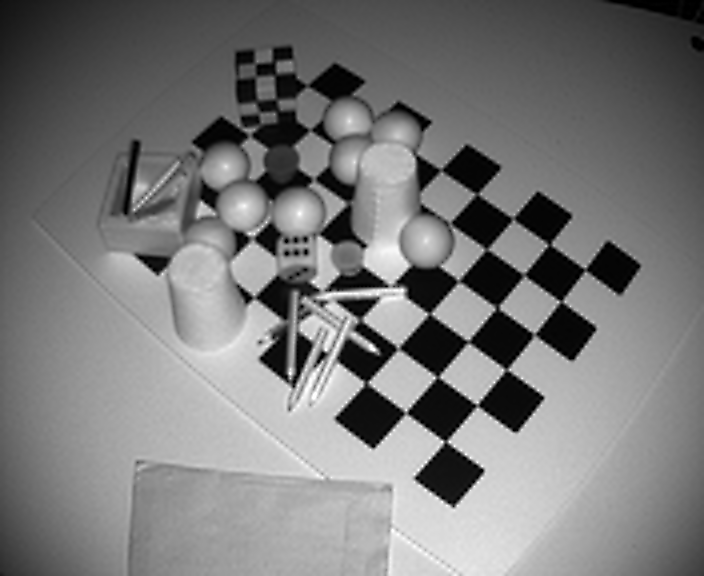}}; 
      \spy on (-0.4,-0.2) in node [left] at (4.88, 0); 
    \end{tikzpicture} 
	}
	\vspace{-1.4em}
	\setcounter{subfigure}{0}
	\subfloat[Original]{\label{fig:06_tofSRResults:range_lr}
		\begin{tikzpicture}[spy using outlines={rectangle,red,magnification=2.3,height=3.4cm, width = 2.4cm, connect spies, every spy on node/.append style={thick}}] 
			\node {\pgfimage[width=0.32\linewidth]{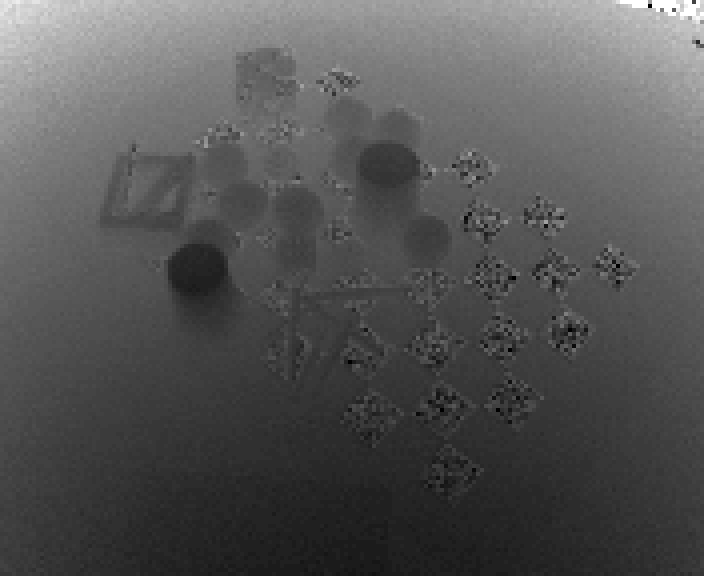}}; 
      \spy[green] on (-1.2,0.6) in node [left] at (4.88, 0); 
    \end{tikzpicture}
	}
	\subfloat[\gls{amsr}]{\label{fig:06_tofSRResults:range_amsr}
		\begin{tikzpicture}[spy using outlines={rectangle,red,magnification=2.3,height=3.4cm, width = 2.4cm, connect spies, every spy on node/.append style={thick}}] 
			\node {\pgfimage[width=0.32\linewidth]{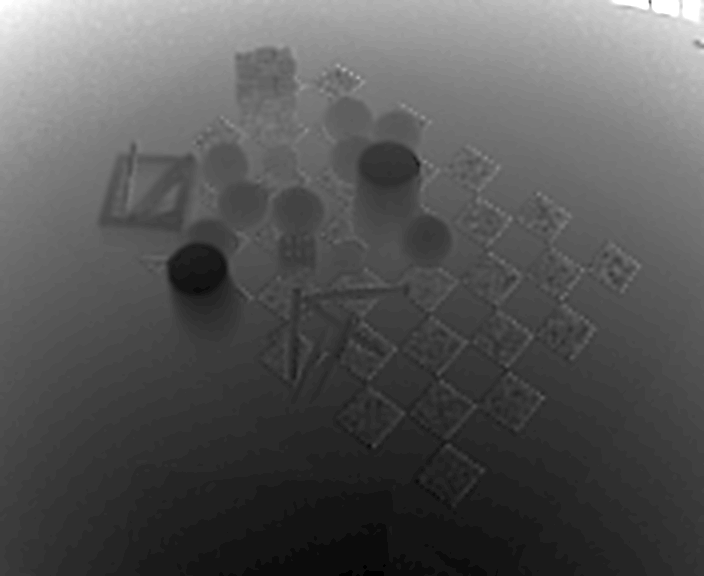}}; 
      \spy[green] on (-1.2,0.6) in node [left] at (4.88, 0); 
    \end{tikzpicture} 
	}
	
	\subfloat{\label{fig:06_tofSRResults:amp_seq}
		\begin{tikzpicture}[spy using outlines={rectangle,red,magnification=2.3,height=3.4cm, width = 2.4cm, connect spies, every spy on node/.append style={thick}}] 
			\node {\pgfimage[width=0.32\linewidth]{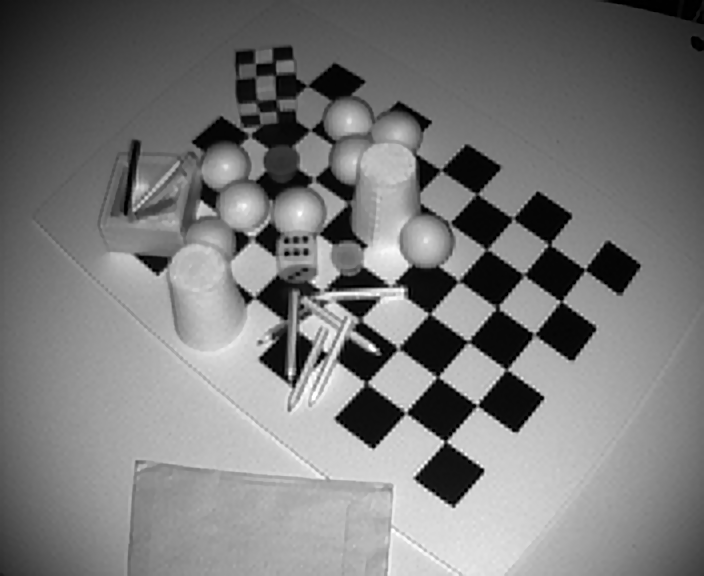}}; 
      \spy on (-0.4,-0.2) in node [left] at (4.88, 0); 
    \end{tikzpicture}
	}
	\subfloat{\label{fig:06_tofSRResults:amp_llr}
		\begin{tikzpicture}[spy using outlines={rectangle,red,magnification=2.3,height=3.4cm, width = 2.4cm, connect spies, every spy on node/.append style={thick}}] 
			\node {\pgfimage[width=0.32\linewidth]{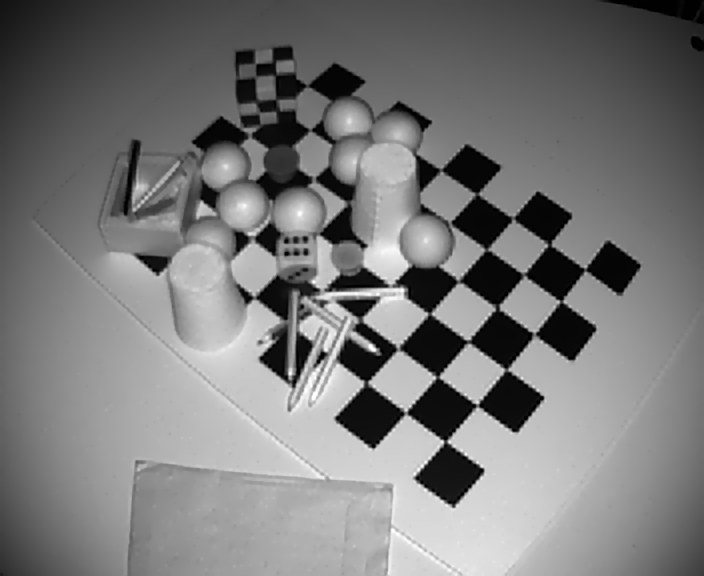}}; 
      \spy on (-0.4,-0.2) in node [left] at (4.88, 0); 
    \end{tikzpicture} 
	}
	\vspace{-1.4em}
	\setcounter{subfigure}{2}
	\subfloat[Sequential]{\label{fig:06_tofSRResults:range_seq}
		\begin{tikzpicture}[spy using outlines={rectangle,red,magnification=2.3,height=3.4cm, width = 2.4cm, connect spies, every spy on node/.append style={thick}}] 
			\node {\pgfimage[width=0.32\linewidth]{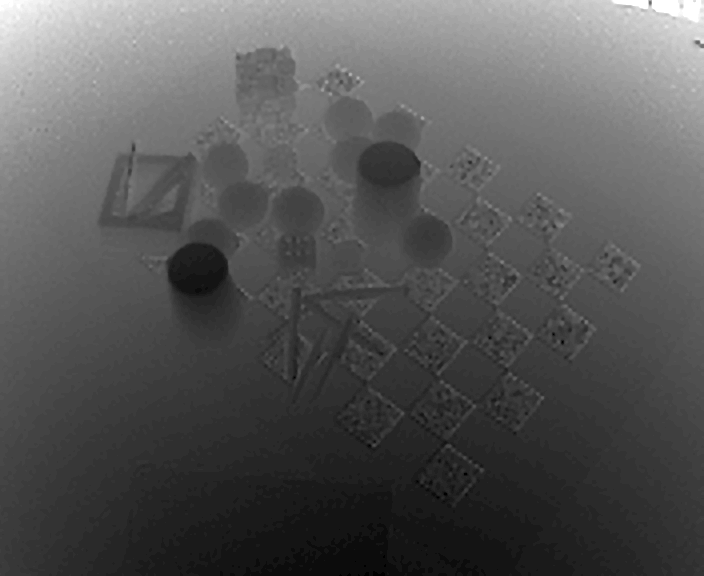}}; 
      \spy[green] on (-1.2,0.6) in node [left] at (4.88, 0); 
    \end{tikzpicture}
	}
	\subfloat[\gls{llr}]{\label{fig:06_tofSRResults:range_llr}
		\begin{tikzpicture}[spy using outlines={rectangle,red,magnification=2.3,height=3.4cm, width = 2.4cm, connect spies, every spy on node/.append style={thick}}] 
			\node {\pgfimage[width=0.32\linewidth]{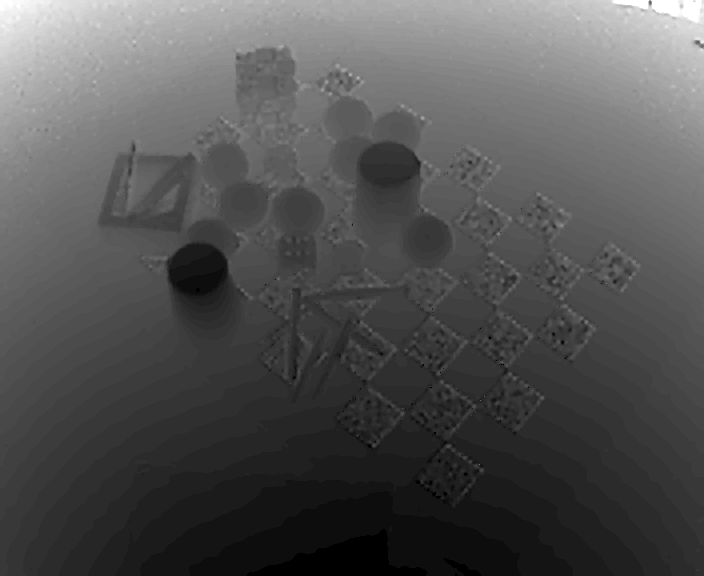}}; 
      \spy[green] on (-1.2,0.6) in node [left] at (4.88, 0); 
    \end{tikzpicture} 
	}
	\caption[Photogeometric super-resolution on the \textit{checkerboard} dataset]{Photogeometric super-resolution on amplitude and range data ($\NumFrames = 16$ frames, magnification $s = 4$) of the \textit{checkerboard} dataset. The image data was captured with a Mesa SR-4000 \gls{tof} camera. \protect\subref{fig:06_tofSRResults:range_lr} Original amplitude (first row) and range data (second row), \protect\subref{fig:06_tofSRResults:range_amsr} super-resolved data using the \acrfull{amsr} presented in \sref{sec:05_ApplicationToHybridRangeImaging}, \protect\subref{fig:06_tofSRResults:range_seq} super-resolved data using sequential processing of both channels and \protect\subref{fig:06_tofSRResults:range_llr} super-resolved data gained by multi-channel processing using the \gls{llr} prior.}
	\label{fig:06_mesaSRResults}
\end{figure}
\begin{figure}[!p] 
	\centering 
  \subfloat{\label{fig:06_mesaSRResults2:amp_lr}
		\begin{tikzpicture}[spy using outlines={rectangle,red,magnification=2.1,height=3.4cm, width = 2.4cm, connect spies, every spy on node/.append style={thick}}] 
			\node {\pgfimage[width=0.32\linewidth]{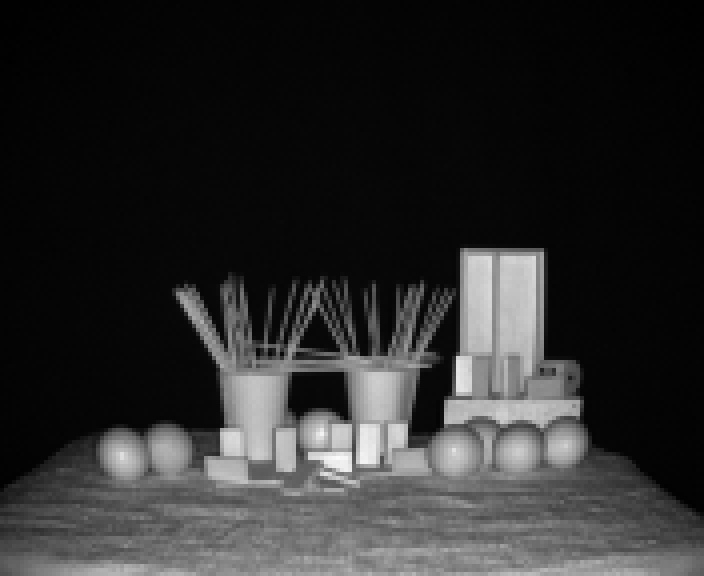}}; 
      \spy on (-0.4,-0.60) in node [left] at (4.88, 0); 
    \end{tikzpicture}
	}
	\subfloat{\label{fig:06_mesaSRResults2:amp_amsr}
		\begin{tikzpicture}[spy using outlines={rectangle,red,magnification=2.1,height=3.4cm, width = 2.4cm, connect spies, every spy on node/.append style={thick}}] 
			\node {\pgfimage[width=0.32\linewidth]{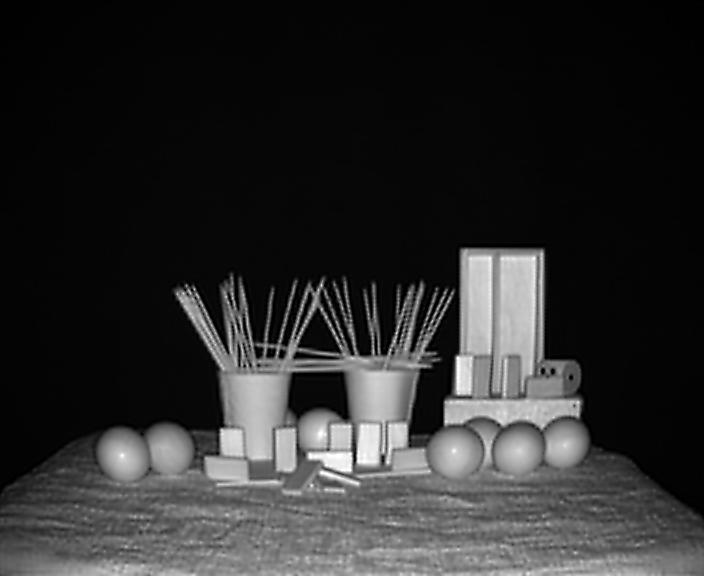}}; 
      \spy on (-0.4,-0.60) in node [left] at (4.88, 0); 
    \end{tikzpicture} 
	}
	\vspace{-1.4em}
	\setcounter{subfigure}{0}
	\subfloat[Original]{\label{fig:06_mesaSRResults2:range_lr}
		\begin{tikzpicture}[spy using outlines={rectangle,red,magnification=2.1,height=3.4cm, width = 2.4cm, connect spies, every spy on node/.append style={thick}}] 
			\node {\pgfimage[width=0.32\linewidth]{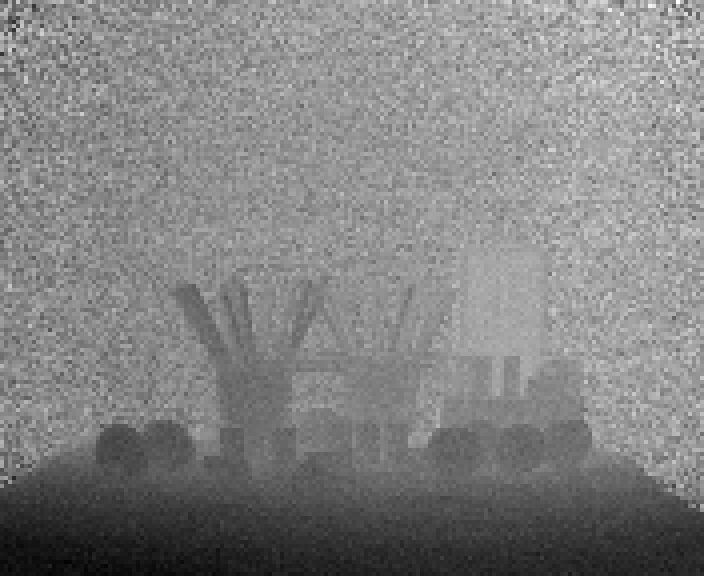}}; 
      \spy[green] on (-0.4,-0.60) in node [left] at (4.88, 0); 
    \end{tikzpicture}
	}
	\subfloat[\gls{amsr}]{\label{fig:06_mesaSRResults2:range_amsr}
		\begin{tikzpicture}[spy using outlines={rectangle,red,magnification=2.1,height=3.4cm, width = 2.4cm, connect spies, every spy on node/.append style={thick}}] 
			\node {\pgfimage[width=0.32\linewidth]{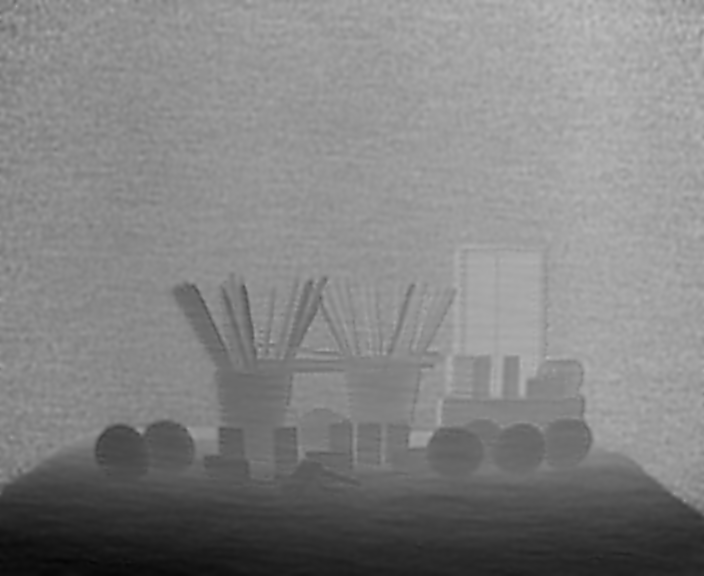}}; 
      \spy[green] on (-0.4,-0.60) in node [left] at (4.88, 0); 
    \end{tikzpicture} 
	}
	
	\subfloat{\label{fig:06_mesaSRResults2:amp_seq}
		\begin{tikzpicture}[spy using outlines={rectangle,red,magnification=2.1,height=3.4cm, width = 2.4cm, connect spies, every spy on node/.append style={thick}}] 
			\node {\pgfimage[width=0.32\linewidth]{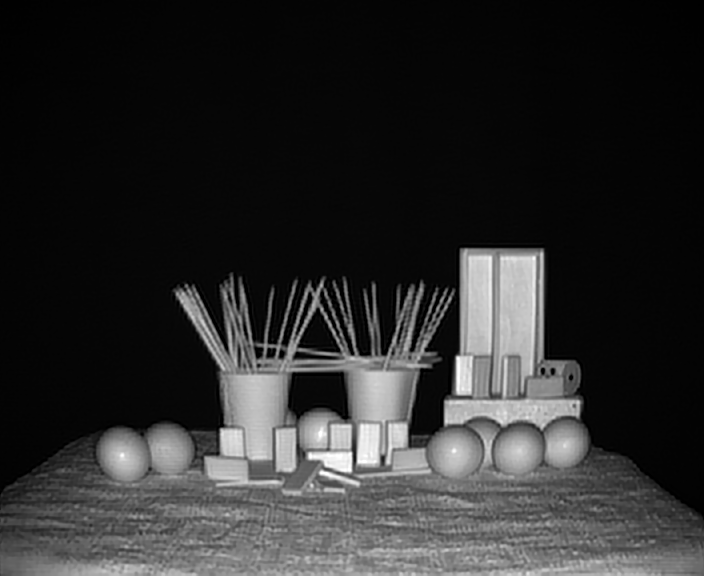}}; 
      \spy on (-0.4,-0.60) in node [left] at (4.88, 0); 
    \end{tikzpicture}
	}
	\subfloat{\label{fig:06_mesaSRResults2:amp_llr}
		\begin{tikzpicture}[spy using outlines={rectangle,red,magnification=2.1,height=3.4cm, width = 2.4cm, connect spies, every spy on node/.append style={thick}}] 
			\node {\pgfimage[width=0.32\linewidth]{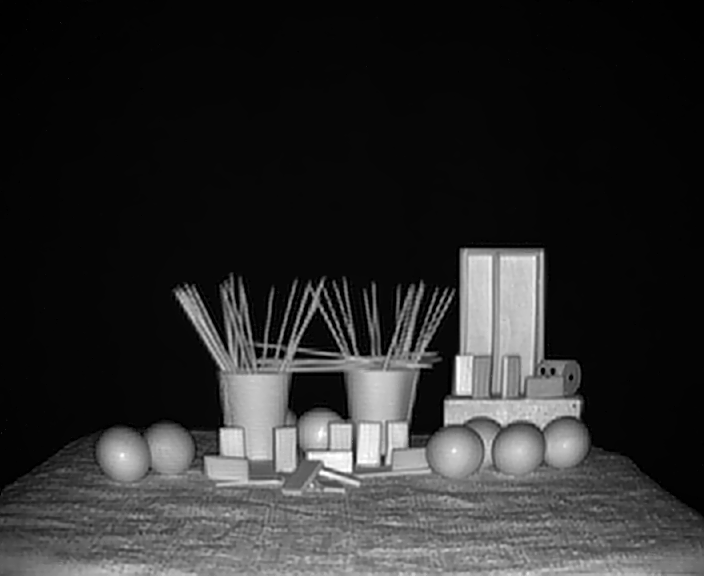}}; 
      \spy on (-0.4,-0.60) in node [left] at (4.88, 0); 
    \end{tikzpicture} 
	}
	\vspace{-1.4em}
	\setcounter{subfigure}{2}
	\subfloat[Sequential]{\label{fig:06_mesaSRResults2:range_seq}
		\begin{tikzpicture}[spy using outlines={rectangle,red,magnification=2.1,height=3.4cm, width = 2.4cm, connect spies, every spy on node/.append style={thick}}] 
			\node {\pgfimage[width=0.32\linewidth]{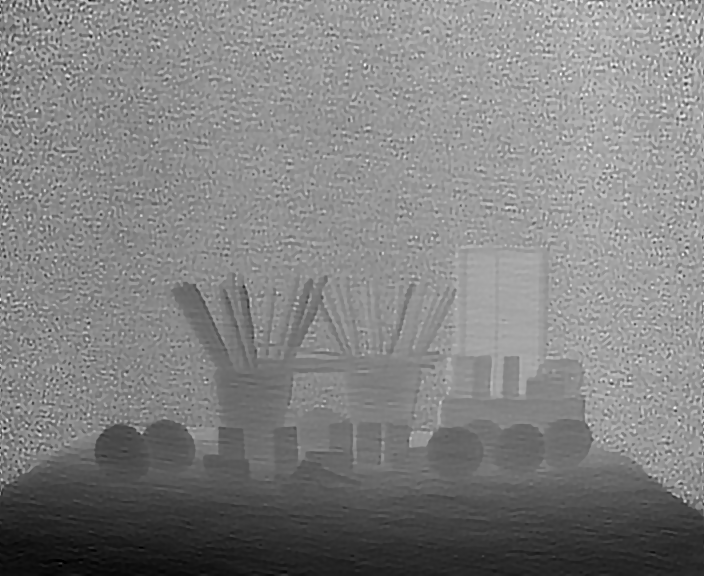}}; 
      \spy[green] on (-0.4,-0.60) in node [left] at (4.88, 0); 
    \end{tikzpicture}
	}
	\subfloat[\gls{llr}]{\label{fig:06_mesaSRResults2:range_llr}
		\begin{tikzpicture}[spy using outlines={rectangle,red,magnification=2.1,height=3.4cm, width = 2.4cm, connect spies, every spy on node/.append style={thick}}] 
			\node {\pgfimage[width=0.32\linewidth]{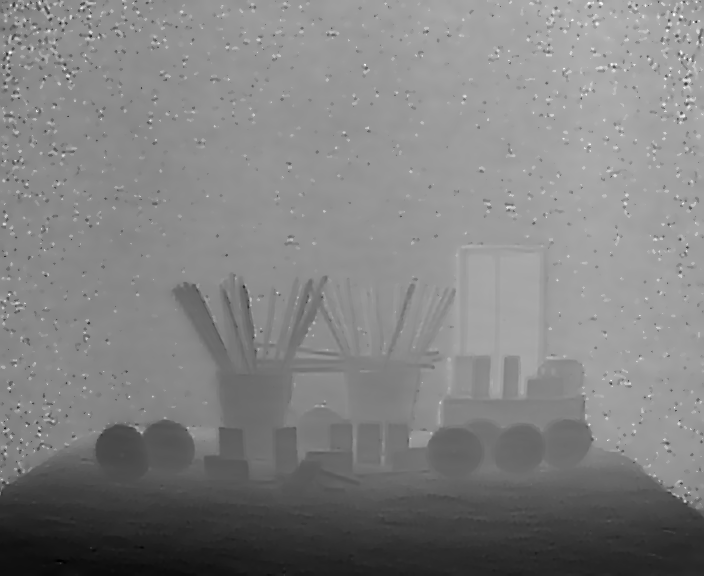}}; 
      \spy[green] on (-0.4,-0.60) in node [left] at (4.88, 0); 
    \end{tikzpicture} 
	}
	\caption[Photogeometric super-resolution on the \textit{games} dataset]{Photogeometric super-resolution on amplitude and range data ($\NumFrames = 21$ frames, magnification $s = 4$) of the \textit{games} dataset. The image data was captured with a Mesa SR-4000 \gls{tof} camera. \protect\subref{fig:06_mesaSRResults2:range_lr} Original amplitude (first row) and range data (second row), \protect\subref{fig:06_mesaSRResults2:range_amsr} super-resolved data using the \acrfull{amsr} presented in \sref{sec:05_ApplicationToHybridRangeImaging}, \protect\subref{fig:06_mesaSRResults2:range_seq} super-resolved data using sequential processing of both channels and \protect\subref{fig:06_mesaSRResults2:range_llr} super-resolved data gained by multi-channel processing using the \gls{llr} prior.}
	\label{fig:06_mesaSRResults2}
\end{figure}

\subsection{Further Applications}
\label{sec:06_FurtherApplications}

The proposed methodology facilitates numerous vision tasks beyond classical resolution enhancement in color and range imaging. Below, we briefly investigate two further example applications of practical relevance.

\paragraph{Multispectral Image Upsampling.}
The proposed multi-channel model generalizes to multispectral imaging for $\NumChannels \gg 3$ channels in a straightforward way as extension of color image processing. Let us consider the application of single-image upsampling, where we target at estimating high-resolution multispectral images from single low-resolution ones\footnote{The reconstruction algorithm can easily be extended to multi-frame reconstruction using motion estimation techniques tailored for multi- and hyperspectral data, see \eg \cite{Zhang2012b}.}. In this highly underdetermined reconstruction problem, we exploit the high degree of correlations among the spectral bands of multispectral images. \Fref{fig:06_resultMultispectral} depicts multispectral upsampling on an example image taken from the Harvard dataset \cite{Chakrabarti2011}. The multispectral data consists of $\NumChannels = 31$ bands that correspond to central wavelengths between 420 and 720\,nm. Here, we depict a false-color visualization using a self organizing map \cite{Jordan2014} (top row) along with a single spectral band centered at $600$\,nm (bottom row).
\begin{figure}[!t]
	\centering
	\mbox{
	\centering
	\hspace{-1.0em}
	\subfloat{
		\begin{tikzpicture}[spy using outlines={rectangle,red,magnification=2.3,height=1.8cm, width = 3.6cm, connect spies, every spy on node/.append style={thick}}] 
			\node {\pgfimage[width=0.242\linewidth]{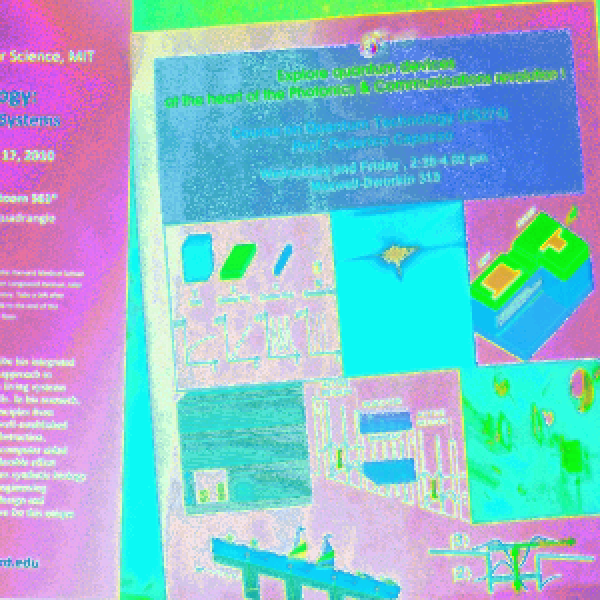}}; 
      \spy on (0.35, 1.2) in node [left] at (1.8, -2.78); 
    \end{tikzpicture}
	}\hspace{-1.1em}
	\subfloat{
		\begin{tikzpicture}[spy using outlines={rectangle,red,magnification=2.3,height=1.8cm, width = 3.6cm, connect spies, every spy on node/.append style={thick}}] 
			\node {\pgfimage[width=0.242\linewidth]{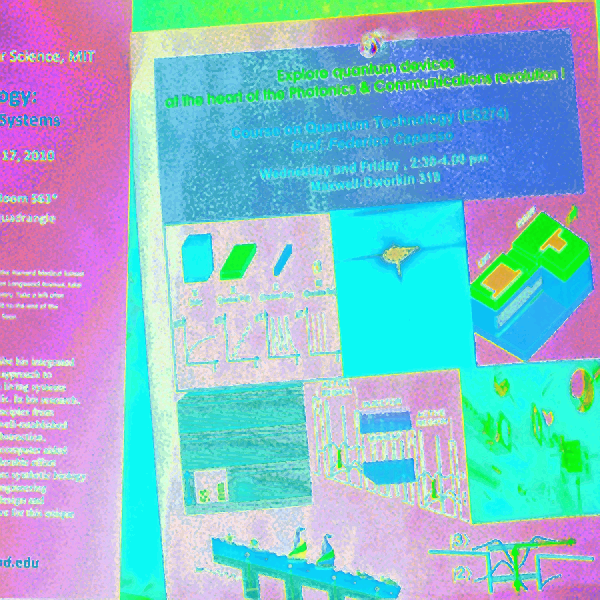}}; 
      \spy on (0.35, 1.2) in node [left] at (1.8, -2.78); 
    \end{tikzpicture}
	}\hspace{-1.1em}	
	\subfloat{
		\begin{tikzpicture}[spy using outlines={rectangle,red,magnification=2.3,height=1.8cm, width = 3.6cm, connect spies, every spy on node/.append style={thick}}] 
			\node {\pgfimage[width=0.242\linewidth]{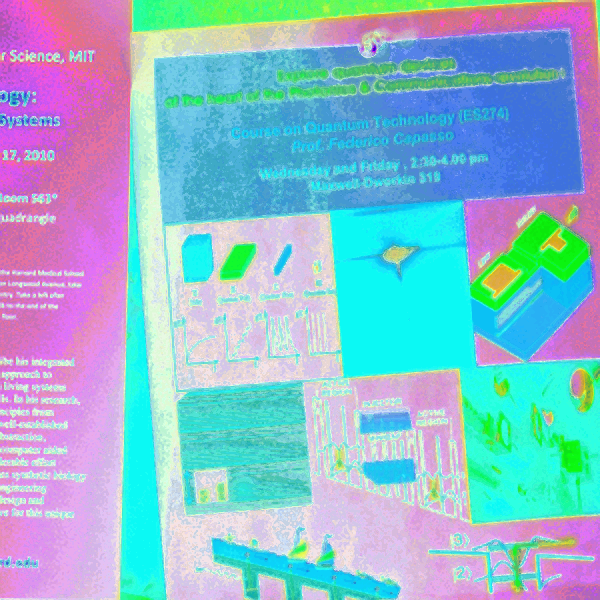}}; 
      \spy on (0.35, 1.2) in node [left] at (1.8, -2.78); 
    \end{tikzpicture}
	}\hspace{-1.1em}	
	\subfloat{
		\begin{tikzpicture}[spy using outlines={rectangle,red,magnification=2.3,height=1.8cm, width = 3.6cm, connect spies, every spy on node/.append style={thick}}] 
			\node {\pgfimage[width=0.242\linewidth]{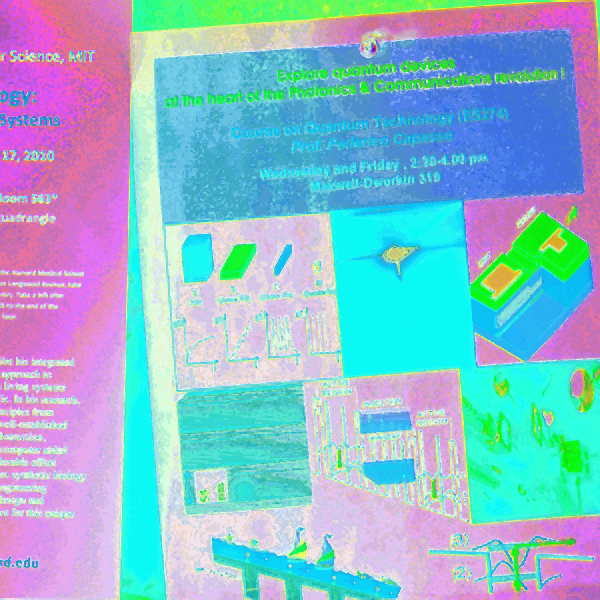}}; 
      \spy on (0.35, 1.2) in node [left] at (1.8, -2.78); 
    \end{tikzpicture}
	}}\\[-1.6ex]
	\setcounter{subfigure}{0}
	\mbox{
	\centering
	\hspace{-1.0em}
	\subfloat[Original]{\label{fig:06_resultMultispectral:orig}
		\begin{tikzpicture}[spy using outlines={rectangle,red,magnification=2.3,height=1.8cm, width = 3.6cm, connect spies, every spy on node/.append style={thick}}] 
			\node {\pgfimage[width=0.242\linewidth]{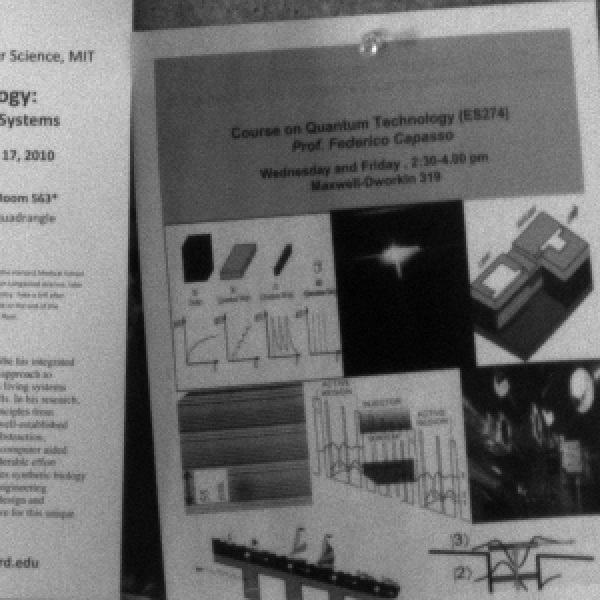}}; 
      \spy on (0.35, 1.2) in node [left] at (1.8, -2.78); 
    \end{tikzpicture}
	}\hspace{-1.1em}
	\subfloat[Sequential]{\label{fig:06_resultMultispectral:seq}
		\begin{tikzpicture}[spy using outlines={rectangle,red,magnification=2.3,height=1.8cm, width = 3.6cm, connect spies, every spy on node/.append style={thick}}] 
			\node {\pgfimage[width=0.242\linewidth]{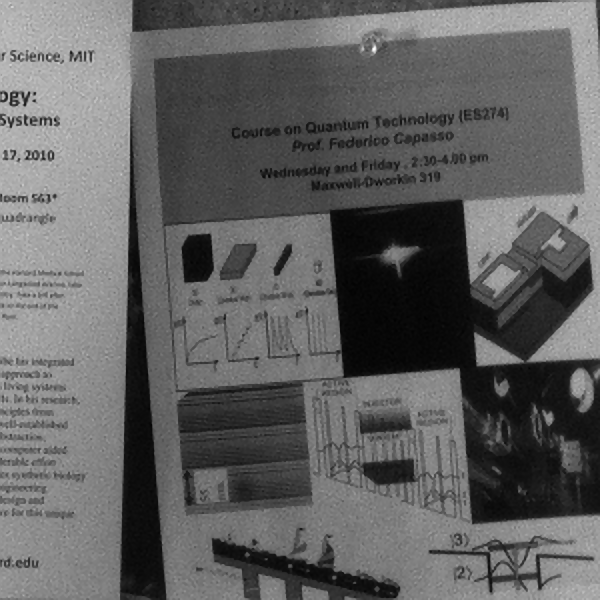}}; 
      \spy on (0.35, 1.2) in node [left] at (1.8, -2.78); 
    \end{tikzpicture}
	}\hspace{-1.1em}	
	\subfloat[\gls{idp} \cite{Farsiu2006}]{\label{fig:06_resultMultispectral:idp}
		\begin{tikzpicture}[spy using outlines={rectangle,red,magnification=2.3,height=1.8cm, width = 3.6cm, connect spies, every spy on node/.append style={thick}}] 
			\node {\pgfimage[width=0.242\linewidth]{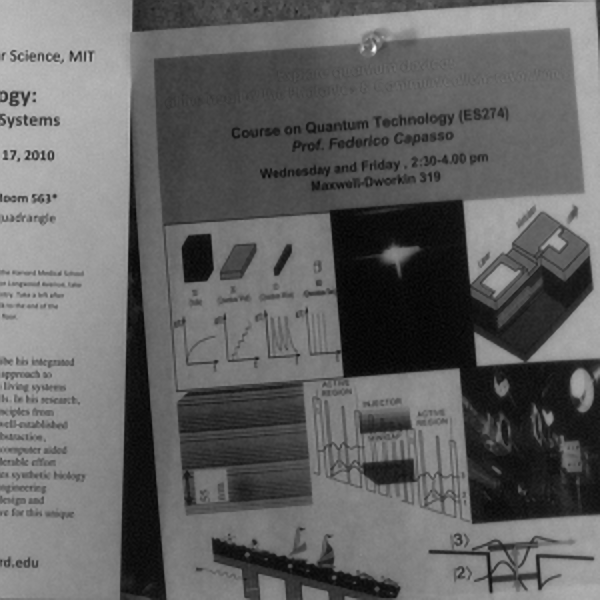}}; 
      \spy on (0.35, 1.2) in node [left] at (1.8, -2.78); 
    \end{tikzpicture}
	}\hspace{-1.1em}	
	\subfloat[\gls{llr}]{\label{fig:06_resultMultispectral:llr}
		\begin{tikzpicture}[spy using outlines={rectangle,red,magnification=2.3,height=1.8cm, width = 3.6cm, connect spies, every spy on node/.append style={thick}}] 
			\node {\pgfimage[width=0.242\linewidth]{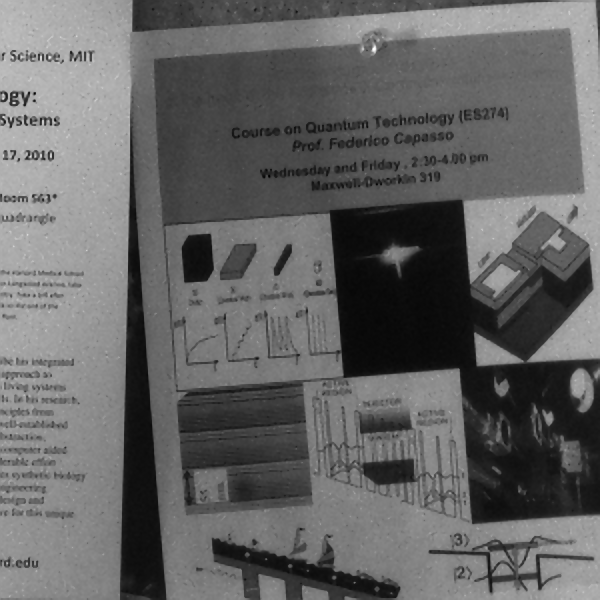}}; 
      \spy on (0.35, 1.2) in node [left] at (1.8, -2.78); 
    \end{tikzpicture}
	}}	
	\caption[Multispectral image upsampling with false-color visualization and single spectral bands]{Multispectral image upsampling ($\NumChannels = 31$ spectral bands, magnification $\MagFac = 2$). Top row: False-color visualization for the original image \protect\subref{fig:06_resultMultispectral:orig}, sequential upsampling of the different channels without inter-channel prior \protect\subref{fig:06_resultMultispectral:seq}, multi-channel upsampling using \gls{idp} regularization \cite{Farsiu2006} \protect\subref{fig:06_resultMultispectral:idp}, and multi-channel upsampling using the \gls{llr} prior \protect\subref{fig:06_resultMultispectral:llr}. Bottom row: Single spectral band at wavelength $600$\,nm. Notice that \gls{idp} produced unwanted structure copying artifacts in the depicted image region.}
	\label{fig:06_resultMultispectral}
\end{figure}

For this application, we employed the \gls{llr} prior with a symmetric distribution. The inter-channel parameters were set to \smash{$\ChannelIdx{\mu}{ij} = 5 \cdot 10^{-3}$}, \smash{$\ChannelIdx{\epsilon}{ij} = 10^{-4}$} and $\LLRPatchSize = 3$ for all pairs of channels. The corresponding intra-channel parameters were set to \smash{$\ChannelIdx{\RegWeight}{i} = 5 \cdot 10^{-4}$}, $\BTVSize = 1$ and $\BTVWeight = 0.5$ for all channels. For the sake of comparison to the \gls{llr} prior, sequential upsampling of the different channels without inter-channel prior as well as multi-channel upsampling using \gls{idp} regularization \cite{Farsiu2006} extended to an arbitrary number of channels was applied. 

In comparison to sequential upsampling, both multi-channel approaches led to a decrease of color artifacts. Similar to color super-resolution, these color artifacts appeared as residual noise due to disregarding mutual dependencies. However, notice that in this example the assumption of strong mutual dependencies was violated for certain pairs of channels due to inconsistent structures. In \fref{fig:06_resultMultispectral}, this assumption was violated for the image region that contains text. This resulted in erroneously copied structure from other, original channels to the upsampled ones in case of \gls{idp} regularization. The proposed outlier-insensitive prior based on Tukey's biweight loss features higher robustness against inconsistencies and avoided these structure copying artifacts.

\paragraph{Joint Segmentation and Super-Resolution.}
One recent trend in image processing is to couple image analysis with retrospective image enhancement via joint frameworks \cite{Shen2007,Lelandais2015}. In \cite{Kohler2015d}, segmentation driven deblurring has been proposed to boost both subtasks compared to their decoupled usage. Following a similar notion, the multi-channel methodology can be adopted to joint segmentation and super-resolution. We can employ consistency among images and a corresponding segmentation as a strong prior to leverage resolution enhancement. This assumption is reasonable for text document images that feature a correlation among intensity values and the appearance of text and symbols.
\begin{figure}[!t]
	\centering
	\hspace{-0.5em}
	\mbox{
	\raisebox{1.5cm}{\rotatebox{90}{\footnotesize Grayscale image}\hspace{-0.4em}}
	\subfloat{
		\begin{tikzpicture}[spy using outlines={rectangle,red,magnification=2.2,height=2.5cm, width=1.3cm, connect spies, every spy on node/.append style={thick}}] 
			\node {\pgfimage[width=0.315\linewidth]{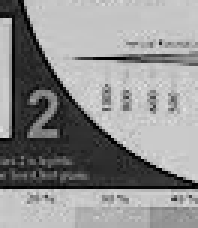}}; 
      \spy on (0.1, 0.1) in node [left] at (-1.075, -1.5); 
    \end{tikzpicture}
	}\hspace{-1.1em}
	\subfloat{
		\begin{tikzpicture}[spy using outlines={rectangle,red,magnification=2.2,height=2.5cm, width=1.3cm, connect spies, every spy on node/.append style={thick}}] 
			\node {\pgfimage[width=0.315\linewidth]{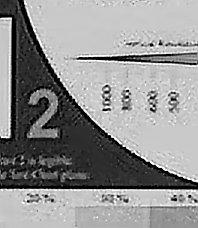}}; 
      \spy on (0.1, 0.1) in node [left] at (-1.075, -1.5); 
    \end{tikzpicture}
	}\hspace{-1.1em}
	\subfloat{
		\begin{tikzpicture}[spy using outlines={rectangle,red,magnification=2.2,height=2.5cm, width=1.3cm, connect spies, every spy on node/.append style={thick}}] 
			\node {\pgfimage[width=0.315\linewidth]{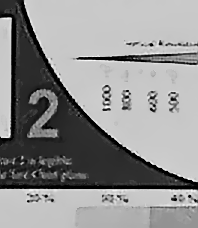}}; 
      \spy on (0.1, 0.1) in node [left] at (-1.075, -1.5); 
    \end{tikzpicture}
	}}\\[-1.6ex]
	\setcounter{subfigure}{0}
	\mbox{
	\raisebox{0.70cm}{\rotatebox{90}{\footnotesize Binarization probability map}\hspace{-0.4em}}
	\subfloat[Original]{\label{fig:06_jbsr_lr}
		\begin{tikzpicture}[spy using outlines={rectangle,red,magnification=2.2,height=2.5cm, width=1.3cm, connect spies, every spy on node/.append style={thick}}] 
			\node {\pgfimage[width=0.315\linewidth]{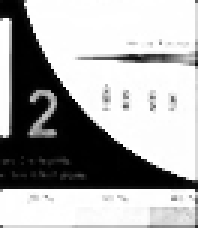}}; 
      \spy on (0.1, 0.1) in node [left] at (-1.075, -1.5); 
    \end{tikzpicture}
	}\hspace{-1.1em}
	\subfloat[Sequential binarization \& SR]{\label{fig:06_jbsr_sr}
		\begin{tikzpicture}[spy using outlines={rectangle,red,magnification=2.2,height=2.5cm, width=1.3cm, connect spies, every spy on node/.append style={thick}}] 
			\node {\pgfimage[width=0.315\linewidth]{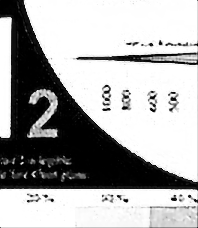}}; 
      \spy on (0.1, 0.1) in node [left] at (-1.075, -1.5); 
    \end{tikzpicture}
	}\hspace{-1.1em}
	\subfloat[Joint binarization \& SR]{\label{fig:06_jbsr_sr_llr}
		\begin{tikzpicture}[spy using outlines={rectangle,red,magnification=2.2,height=2.5cm, width=1.3cm, connect spies, every spy on node/.append style={thick}}] 
			\node {\pgfimage[width=0.315\linewidth]{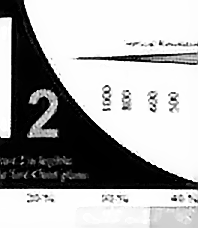}}; 
      \spy on (0.1, 0.1) in node [left] at (-1.075, -1.5); 
    \end{tikzpicture}
	}}
	\caption[Joint binarization and super-resolution on two-tone images]{Joint binarization and super-resolution on two-tone images ($\NumFrames = 20$ frames, magnification $\MagFac = 3$). \protect\subref{fig:06_jbsr_lr} - \protect\subref{fig:06_jbsr_sr} Single grayscale image and the decoupled use of super-resolution (top row) along with the corresponding binarizations modeled as probability maps (bottom row). \protect\subref{fig:06_jbsr_sr_llr} Joint binarization and super-resolution using the \gls{llr} prior. It is worth noting that the joint approach better removed ringing and compression artifacts.}
	\label{fig:06_jbsrExample}
\end{figure}

We demonstrate this concept in \fref{fig:06_jbsrExample} for binarization on compressed two-tone frames of the \textit{Adyoron} sequence \cite{Farsiu2014}. Motion estimation across the sequence ($\NumFrames = 20$ frames) was performed by \gls{ecc} optimization \cite{Evangelidis2008} with an affine model. We employed the \gls{llr} prior with symmetric distribution for $\NumChannels = 2$ channels with $\ChannelIdx{\mu}{ij} = 25$, \smash{$\ChannelIdx{\epsilon}{ij} = 10^{-4}$} and $\LLRPatchSize = 3$. The intra-channel parameters were set to \smash{$\ChannelIdx{\RegWeight}{i} = 10^{-3}$} for the grayscale image ($i = 1$) and \smash{$\ChannelIdx{\RegWeight}{i} = 10^{-1}$} for the binarization ($i = 2$) with $\BTVSize = 2$ and $\BTVWeight = 0.5$ for both channels. 

\Fref{fig:06_jbsr_lr} depicts a single grayscale image along with its binarization using the intensity-based soft-clustering proposed in \cite{Kohler2015d} for text images. We represent a binarization of an image $\HR \in \RealN{\HRSize}$ as a probability map $\vec{s} \in [0, 1]^\HRSize$, where $s_i = 0$ indicates a dark structure, \eg font, at the $i$-th pixel. Note that the binarization failed to detect text and symbols accurately on aliased low-resolution data. For the sake of comparison, \fref{fig:06_jbsr_sr} depicts super-resolution on the grayscale images followed by the binarization in a sequential manner. This served as initialization for the joint approach in \fref{fig:06_jbsr_sr_llr} that considered grayscale data and its binarization as coupled channels via the \gls{llr} prior. Notice that the joint approach better compensated ringing and compression artifacts and reconstructed both channels simultaneously at a super-resolved scale. This can serve as a more reliable basis for subsequent image analysis tasks like text detection and recognition \cite{Espana-Boquera2011} or writer identification \cite{Christlein2015} to name a few.

\section{Conclusion}
\label{sec:06_Conclusion}

This chapter proposed a novel approach to multi-sensor super-resolution that is applicable to a variety of current computer vision applications. As the core idea, we introduced a Bayesian model that accounts for the image formation process of multi-channel images as well as a \gls{llr} prior distribution to consider mutual dependencies among different image channels. Subsequently, we developed Bayesian parameter estimation techniques based on this model, where we proposed a joint multi-channel reconstruction algorithm to take inter-channel dependencies into consideration. We presented a thorough analysis of this model including comparisons to related methods, where we discussed connections to several popular image filters. Eventually, we studied several target applications, where the proposed method generalized fairly well and outperformed the state-of-the-art. In color image super-resolution as a classical application, \gls{llr} improved the \gls{psnr} by $1.5$\,\gls{db} and the \gls{ssim} by $0.04$ compared to conventional channel-wise super-resolution. Unlike related work, \gls{llr} can handle single- and multi-frame resolution enhancement in a unified framework and does neither rely on guidance information nor on a fixed number of image channels. We also presented potential applications beyond classical super-resolution including multispectral image upsampling as well as joint segmentation and super-resolution on two-tone images. 

Albeit its flexible and unified formulation, this approach can be further tailored to specific domains. One extension is to optionally augment the prior distribution by a static guidance as proposed for related image filters \cite{Ham2015}. This might further boost super-resolution for applications, where such guidance data is available. Another extension includes the acceleration of the algorithm. In this chapter, we proposed a brute-force approach to exploit mutual dependencies among all pairs of image channels but a suitable dimensionality reduction could enhance the efficiency in case of a very large number of channels.
 

\part{Super-Resolution in Medical Imaging}
\label{sec:SuperResolutionInMedicalImaging}

\chapter{Applications in Retinal Fundus Video Imaging}
\label{sec:RetinalFundusVideoImaging}

\myminitoc

\noindent
Over the past years, super-resolution has found its entry into various fields of medical imaging and has been examined for diagnostic or interventional workflows, \eg in radiology \cite{Greenspan2008,Robinson2010a}. This chapter presents new applications of super-resolution in retinal video imaging that has recently emerged as a branch of today's ophthalmic imaging technologies. As the primary contribution, a tailored method to reconstruct high-resolution retinal fundus images from video sequences taken from the human eye background is presented. Super-resolution exploits natural human eye movements that occur during an examination and cause subpixel motion across video frames. As an additional contribution, a novel method to assess noise and sharpness characteristics in fundus images in a fully automatic manner is presented. This quality measure is employed in a new automatic hyperparameter selection scheme for super-resolution reconstruction.

The proposed super-resolution framework has been originally published in \cite{Kohler2014} and image quality assessment has been first published in \cite{Kohler2013}.

\section{Introduction and Medical Background}

In today's ophthalmology, retinal \textit{fundus imaging} is one of the most frequently used techniques to gain information about the human eye background non-invasively \cite{Patton2006}. The primary scopes of this structural imaging technology are documentation \cite{Abramoff2010} and computer-aided screening of eye diseases such as  diabetic retinopathy \cite{Abramoff2015} or glaucoma \cite{Bock2010,Kohler2015}, among others. This is mainly because of the cost efficiency and availability of fundus cameras compared to other modalities like \gls{oct} \cite{Huang1991}. The most common approach in clinical practice are single-shot techniques that provide high-resolution color photographs of the human retina. Consequently, they provide static information of the retina. Another trend that has recently emerged is \textit{fundus video imaging} to gain dynamic measurements. The main motivation behind the use of video capable cameras is to measure fast temporal changes of the retina, \eg the cardiac cycle \cite{Tornow2015}. Compared to single-shot imaging, video imaging features a high temporal resolution but single frames are limited in terms of their spatial resolution and \gls{snr}, see \fref{fig:07_compLowCostVideoKowa}. This is mainly caused by technological aspects, \eg increased light exposure times, but also related to economic reasons as the use of mobile and cost-effective hardware is desirable \cite{Hoher2015}. 

Image enhancement and restoration is an emerging field of research in retinal imaging in order to enhance the diagnostic usability of fundus images. One approach are temporal denoising schemes. In \cite{Kohler2012}, K\"ohler \etal have proposed adaptive temporal averaging using a registration based compensation of natural eye movements that appear during video imaging. This approach recovers single denoised images from a set of noisy frames. Another direction of prior work are blind deconvolution algorithms with the goal to recover a sharp retinal image from a blurred one. These methods can be applied in a multi-frame scheme to image pairs acquired in longitudinal examinations as shown by Marrugo \etal \cite{Marrugo2011}. However, one limitation of blind deconvolution is that it does not enhance the spatial resolution in terms of pixel sampling. 

This chapter investigates a different direction and aims at improving the diagnostic usability of retinal images by multi-frame super-resolution. The basic idea behind this approach is to utilize natural eye movements \cite{Rayner1998,Duchowski2007} during video imaging as a cue for super-resolution reconstruction. In \cite{Murillo2011a}, Murillo \etal proposed this idea to enhance scanning laser ophthalmoscopy. In parallel to the work presented in this thesis, single- and multi-frame resolution enhancement for fundus imaging have also been studied by Thapa \etal \cite{Thapa2014}. In general, super-resolution applied to retinal images is challenging due to optical aberrations caused by the optics of the human eye. Other common issues are photometric distortions or specular reflections that lead to oversaturations. These effects are related to the external illumination and the imaging through a small pupil, which makes homogeneous illuminations of the retina difficult to achieve. Moreover, fundus video data might be affected by a poor \gls{snr} as the light exposure during the examination needs to be low to avoid impairments of the patient. Therefore, robust estimation techniques are required to deal with these conditions.

The remainder of this chapter is structured as follows. \sref{sec:07_ImageFormationModelForRetinalImaging} develops a domain-specific image formation model that provides the basis of super-resolution in fundus video imaging. \sref{sec:07_SuperResolutionWithQualitySelfAssessment} presents a super-resolution algorithm that employs a novel hyperparameter selection scheme termed \textit{quality self-assessment} to jointly estimate a super-resolved image along with optimal regularization parameters. \sref{sec:07_ExperimentsAndResults} presents an experimental evaluation of the proposed framework in fundus video imaging. Finally, \sref{sec:07_Conclusion} draws a conclusion.
\begin{figure}[!t]
	\centering
	\subfloat[]{\includegraphics[height=0.343\textwidth]{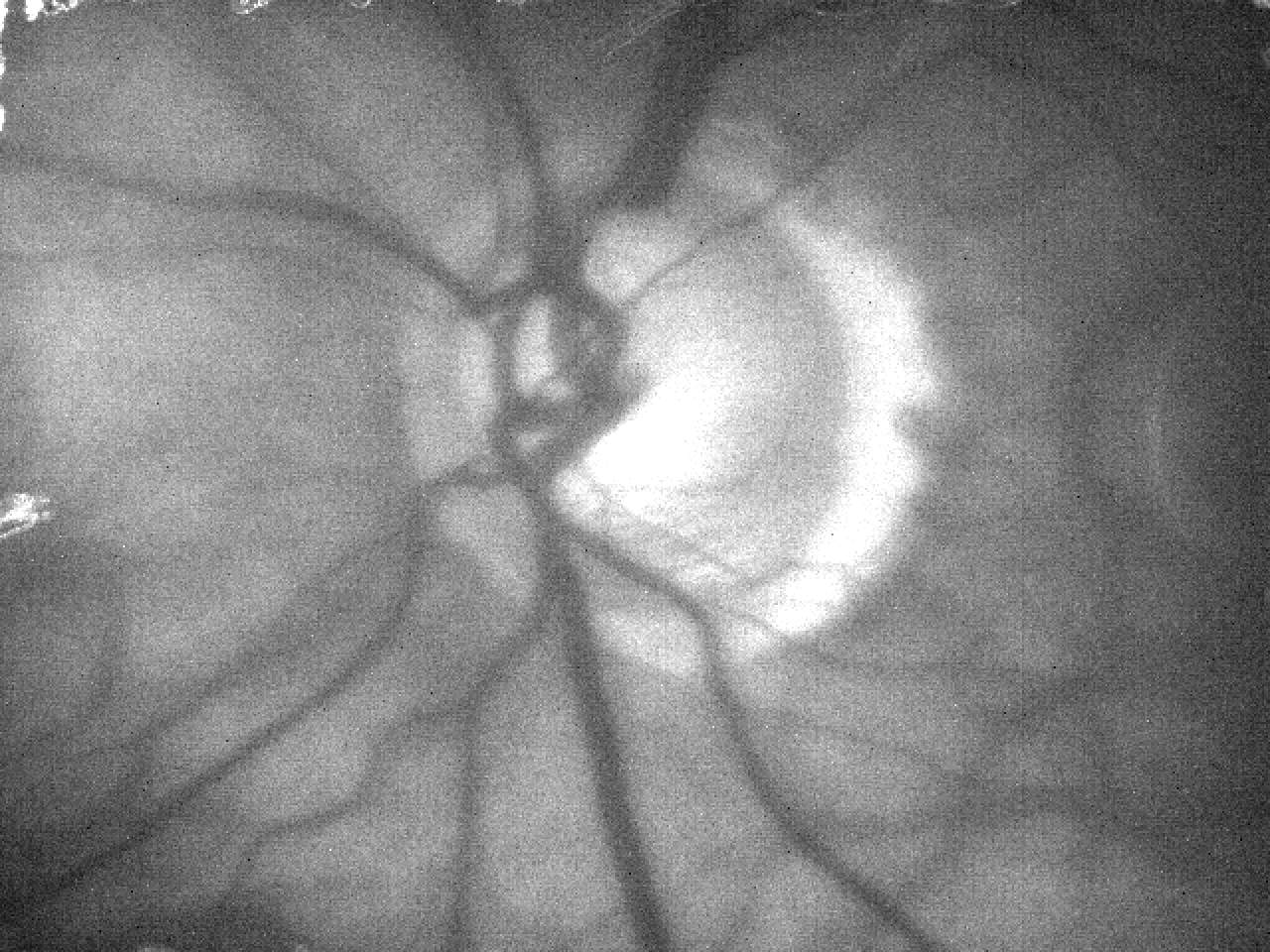}\label{fig:07_compLowCostVideoKowa:lowCost}}~
	\subfloat[]{\includegraphics[height=0.343\textwidth]{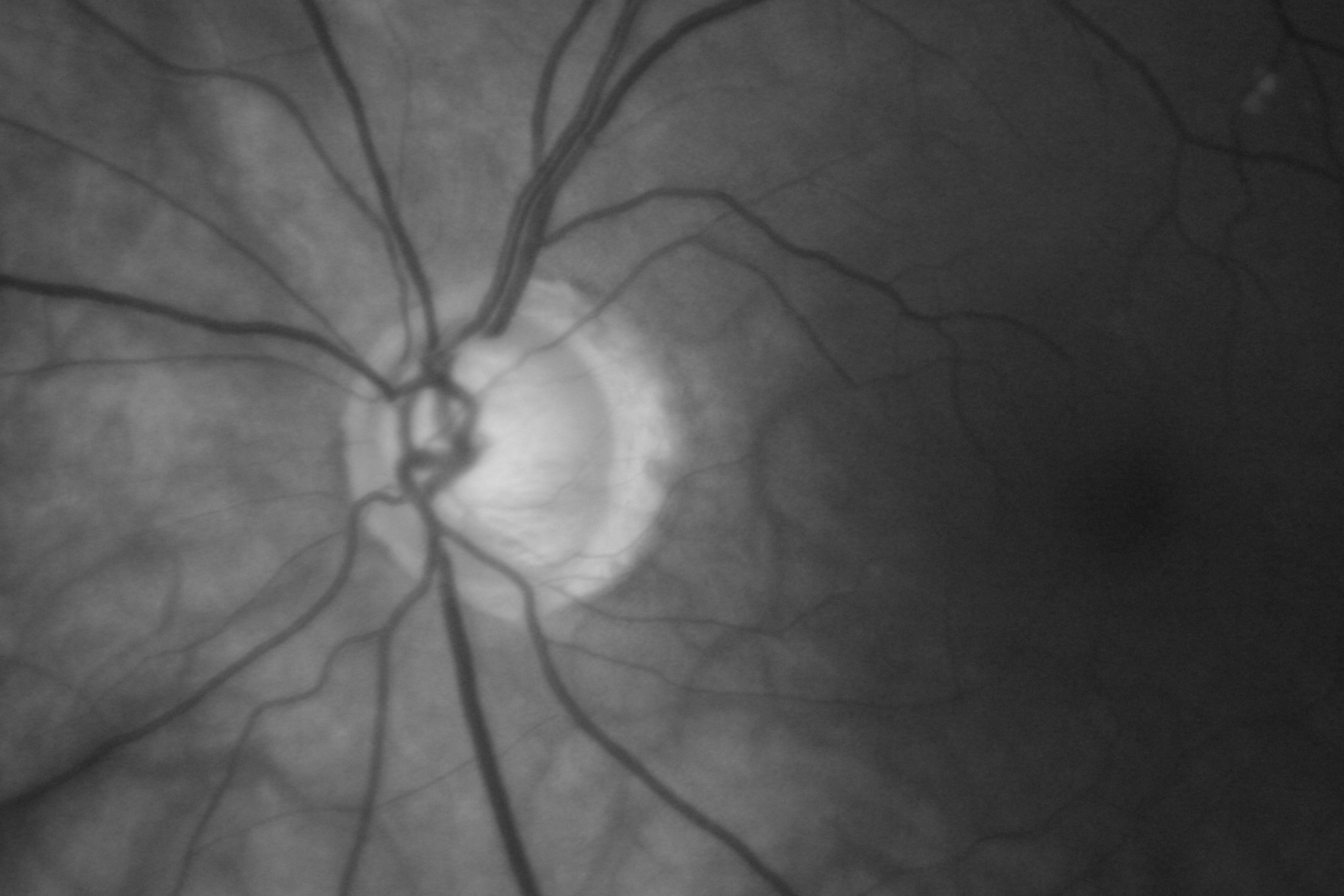}\label{fig:07_compLowCostVideoKowa:kowa}}
	\caption[Single-shot versus video techniques in retinal imaging]{Single-shot versus video techniques in retinal imaging. \protect\subref{fig:07_compLowCostVideoKowa:lowCost} Single frame (640\,$\times$\,480\,px, 15$^\circ$ \gls{fov}) acquired from a glaucoma patient using the video camera system developed by Tornow \etal \cite{Tornow2015}. \protect\subref{fig:07_compLowCostVideoKowa:kowa} Single photograph (1944\,$\times$\,1296\,px, 22.5$^\circ$ \gls{fov}) captured from the same patient with a commercially available Kowa nonmyd fundus camera. For fair comparison to monochromatic video data, the green color channel of the Kowa image is depicted.}
	\label{fig:07_compLowCostVideoKowa}
\end{figure}

\section{Image Formation Model for Retinal Imaging}
\label{sec:07_ImageFormationModelForRetinalImaging}

In order to reconstruct high-resolution retinal images, we exploit sequences of low-resolution video frames. For the sake of notational brevity, we limit ourselves in this chapter to monochromatic video data and hence a single frame represents the luminance of a retinal image. For super-resolution in a Bayesian framework, we need to formulate an image formation model that relates the individual frames to the unknown high-resolution image. In the following subsections, we adopt the model previously presented in \sref{sec:03_ModelingTheImageFormationProcess} to the conditions in retinal imaging.

\subsection{Derivation of the Motion Model}

One of the most important aspects of the proposed model is the fact that subpixel motion across video frames that are captured during an examination is related movements of the eye relative to a fundus camera. According to Rayner \cite{Rayner1998}, such eye movements can be categorized into pursuits, vergence, vestibular motion, and saccades that differ in their causes, amplitudes, and speeds. Pursuit and vergence motion are related to continuously fixating moving targets and nearby, static targets, respectively. Vestibular motion is caused by small motion of the head or the entire body. Saccades refer to abrupt motion with high velocity that occur during the fixation to a new target \cite{Duchowski2007}. In addition to these movements, notice that the human eye is never completely still when fixating a target due to natural eye motion caused by tremors, drifts and microsaccades \cite{Rayner1998,Duchowski2007}. Such movements occur randomly and have small amplitudes. As they are unavoidable during an examination, there is no need to induce camera motion by means of mechanical components to enable super-resolution reconstruction.

To describe eye movements mathematically, we follow the model developed by Can \etal \cite{Can2002} that approximates the human retina as a spherical surface. Let us consider two views of the retina with eye movements among them. These views are associated with two video frames \smash{$\HRFrame{r}$} and \smash{$\HRFrame{k}$} captured with one single camera, where \smash{$\HRFrame{r}$} denotes the reference frame. Furthermore, let \smash{$\FrameIdx{\vec{U}}{r} \in \RealN{3}$} and \smash{$\FrameIdx{\vec{U}}{k}\in \RealN{3}$} be the coordinates of a single point on the retina in the 3-D space transformed to the local camera coordinate systems associated with these frames. According to \cite{Can2002}, we obtain \smash{$\FrameIdx{\vec{U}}{k} = (U, V, \pi(U, V))^\top$} by the quadratic surface model: 
\vspace{-0.4em}
\begin{equation}
	\label{eqn:07_retinaSurfaceModel}
	\pi(U, V) = \pi_1 U^2 + \pi_2 V^2 + \pi_3 U V + \pi_4 U + \pi_5 V + \pi_6, 
\end{equation}
where the parameter set \smash{$\pi = \{ \pi_1, \ldots, \pi_6 \}$} describes the retina shape. Since the curvature of the retina is negligible compared to its distance to the camera center, we can assume a weak-perspective camera model \cite{Hartley2004} that is represented by the projection matrix \smash{$\vec{P} \in \RealMN{3}{4}$} for both views. Then, \smash{$\FrameIdx{\vec{U}}{k}$} and \smash{$\FrameIdx{\vec{U}}{r}$} are mapped to \smash{$\FrameIdx{\vec{u}}{k} = \vec{P} \FrameIdx{\vec{U}}{k}$} and \smash{$\FrameIdx{\vec{u}}{r} = \vec{P} \FrameIdx{\vec{U}}{r}$} on the image plane in the two frames. As eye motion can be considered as rigid \cite{Duchowski2007}, the relationship between \smash{$\FrameIdx{\vec{U}}{k}$} and \smash{$\FrameIdx{\vec{U}}{r}$} is given by $\FrameIdx{\vec{U}}{r} = \RotMat \FrameIdx{\vec{U}}{k} + \TransVec$, where $\RotMat \in \RealMN{3}{3}$ and $\TransVec \in \RealN{3}$ denote a rotation matrix and a translation vector, respectively. This leads to the relationship between the 3-D point \smash{$\FrameIdx{\vec{U}}{r} = (U^\prime, V^\prime, \pi(U^\prime, V^\prime))^\top$} and its 2-D projection \smash{$\FrameIdx{\vec{u}}{k}$} according to: 
\vspace{-0.4em}
\begin{align}
	\label{eqn:07_imageToRetinaRelationship1}
	U^\prime &= R_{11} \cdot \frac{s_0 u - c_u}{f_u} + R_{12} \cdot \frac{s_0 v - c_v}{f_v} + R_{13} \cdot \zeta(u, v) + t_u, \\
	\label{eqn:07_imageToRetinaRelationship2}
	V^\prime &= R_{21} \cdot \frac{s_0 u - c_u}{f_u} + R_{22} \cdot \frac{s_0 v - c_v}{f_v} + R_{23} \cdot \zeta(u, v) + t_v,
\end{align}
where $f_u$ and $f_v$ are the focal lengths, $(c_u, c_v)^\top$ is the camera center, and $s_0$ is a scaling parameter of the weak-perspective camera. $\zeta(u, v)$ is the quadratic equation:
\vspace{-0.4em} 
\begin{equation}
	\zeta(u, v) = \zeta_1 u^2 +  \zeta_2 v^2 +  \zeta_3 uv + \zeta_4 u +  \zeta_5 v +  \zeta_6,
\end{equation}
where the parameters $\zeta = \{\zeta_i, \ldots, \zeta_6 \}$ depend on the shape parameters $\pi$ in \eref{eqn:07_retinaSurfaceModel} and the projection matrix $\vec{P}$. Based on \ereftwo{eqn:07_imageToRetinaRelationship1}{eqn:07_imageToRetinaRelationship2}, we can establish the relationship between the 2-D points \smash{$\FrameIdx{\vec{u}}{r} = (u^\prime, v^\prime)^\top$} and \smash{$\FrameIdx{\vec{u}}{k} = (u, v)^\top$} by the quadratic image-to-image transformation: 
\begin{equation}
	\label{eqn:07_quadraticMotionModel}
	\begin{pmatrix}
		u^\prime \\ v^\prime
	\end{pmatrix}
	=
	\begin{pmatrix}
			\MotionParamsSingle{1}	&	\MotionParamsSingle{2}	&	\MotionParamsSingle{3} & \MotionParamsSingle{4}	&	\MotionParamsSingle{5}	&	\MotionParamsSingle{6} \\
			\MotionParamsSingle{7}	&	\MotionParamsSingle{8}	&	\MotionParamsSingle{9} & \MotionParamsSingle{10}	&	\MotionParamsSingle{11}	&	\MotionParamsSingle{12}
	\end{pmatrix}
	\begin{pmatrix}
			u^2 & v^2 & uv & u & v & 1
	\end{pmatrix}^\top,
\end{equation}
where the transformation parameters $\MotionParamsSingle{i}$ depend on the shape parameters $\pi$, the projection matrix $\vec{P}$, as well as the eye movement characterized by $\RotMat$ and $\TransVec$.

In contrast to the derivation in \cite{Can2002} for general types of rigid eye motion, the proposed model considers video imaging with high frame rates. For super-resolution, we exploit miniature movements when fixating static targets over short time intervals, \ie tremors and microsaccades \cite{Rayner1998}, but neglect movements with larger amplitudes over longer intervals. As this motion is small compared to the \gls{fov} (see \fref{fig:07_eyeMovements}), we assume that $\TransVec \approx \Zeros$ and $\RotMat \approx \Id$. Therefore, we trade the accuracy of the quadratic transformation in \eref{eqn:07_quadraticMotionModel} against improved robustness of parameter estimation using the affine transformation \cite{Fang2006}\label{notation:motionParamsSingle}:
\begin{align}
	\label{eqn:07_affineMotionModel}
	\begin{pmatrix}
		u^\prime \\ v^\prime \\ 1
	\end{pmatrix}
	=
	\begin{pmatrix}
			\MotionParamsSingle{1}	&	\MotionParamsSingle{2}	&	\MotionParamsSingle{3} \\
			\MotionParamsSingle{4}	&	\MotionParamsSingle{5}	&	\MotionParamsSingle{6} \\
			0	&	0	&	1
		\end{pmatrix}
		\begin{pmatrix}
		u \\ v \\ 1
	\end{pmatrix}.
\end{align}
This homography relates eye motion to translation, rotation, scaling and shearing on the image plane. These effects are described by six degrees of freedom given by $\MotionParams = \{ \MotionParamsSingle{1}, \ldots, \MotionParamsSingle{6} \}$\label{notation:motionParams} (see \sref{sec:03_DiscretizationOfTheImageFormationModel}) as opposed to the twelve degrees of freedom in \eref{eqn:07_quadraticMotionModel}. Notice that related eye motion models are based on globally rigid homographies \cite{Kolar2015,Kolar2016a}, which is a specialization of the affine model.

\begin{figure}[!t]
	\centering
	\subfloat[Frame 1]{\includegraphics[width=0.32\textwidth]{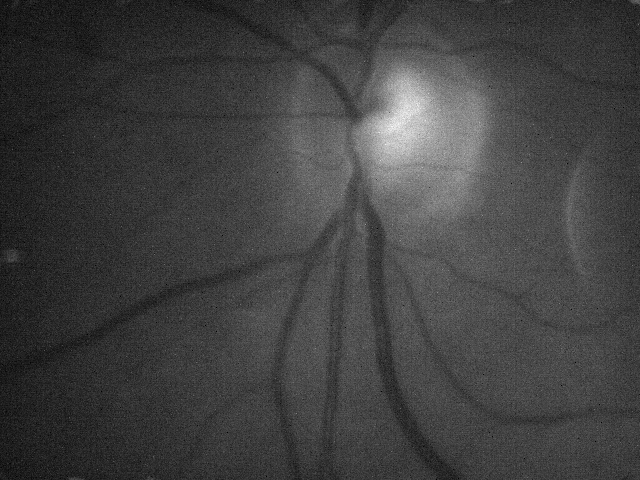}\label{fig:07_eyeMovements:frame1}}~
	\subfloat[Frame 12]{\includegraphics[width=0.32\textwidth]{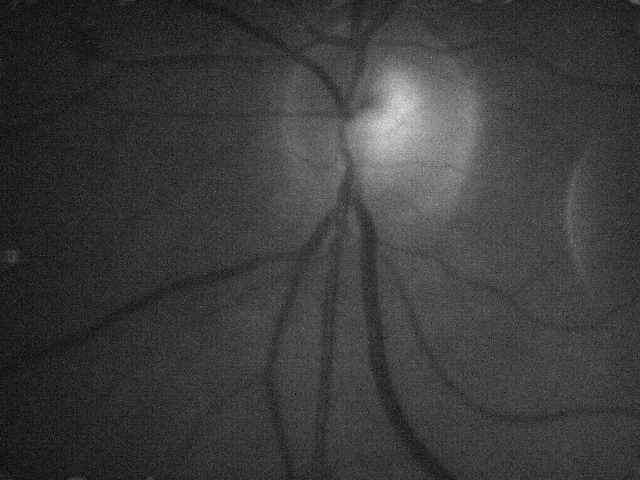}\label{fig:07_eyeMovements:frame12}}~
	\subfloat[Checkerboard overlay]{\includegraphics[width=0.32\textwidth]{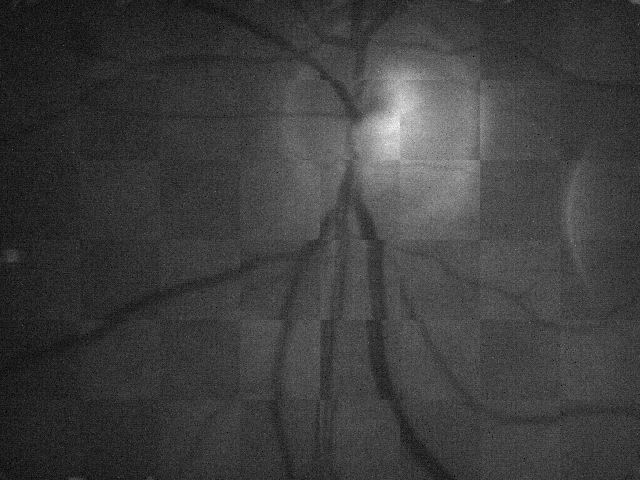}\label{fig:07_eyeMovements:checkerboard}}
	\caption[Natural eye movements in retinal fundus video imaging]{Illustration of natural, miniature eye movements in retinal fundus video imaging. \protect\subref{fig:07_eyeMovements:frame1} - \protect\subref{fig:07_eyeMovements:frame12} Two frames (frames 1 and 12) with eye movements captured at a frame rate of 25\,Hz. \protect\subref{fig:07_eyeMovements:checkerboard} Checkerboard visualization for both frames. Notice that miniature movements can be perceived in the checkerboard visualization but are small compared to the \gls{fov}.}
	\label{fig:07_eyeMovements}
\end{figure}

\subsection{Derivation of the Photometric Model}

In practice, intensity changes among fundus video frames are not exclusively related to eye motion. Another effect that needs to be modeled are photometric variations, which can be distinguished in spatial and temporal ones, see \fref{fig:07_photometricChanges}. Spatial variations are caused by illumination inhomogeneities within a single image \cite{Marrugo2011a}. Such variations occur due to the curved shape of the retina, which makes it difficult to achieve homogenous illumination conditions at the image center and in peripheral regions. Furthermore, the illumination depends on the eye anatomy and the presence of diseases. Temporal variations are caused by brightness changes across multiple frames, which can be caused by eye movements or pulsatile changes \cite{Tornow2015}. Since both types of variations are unavoidable in general, photometric registration is required to compensate for them. This can be achieved by retrospective correction methods \cite{Kolar2011,Zheng2012} that are commonly applied as a preprocessing step in retinal image restoration \cite{Marrugo2011}.

In this work, photometric variations are corrected jointly to super-resolution instead of correcting the input frames in a preprocessing step. The proposed model extends the global photometric model introduced by Capel and Zisserman \cite{Capel2003}. This model describes the relation between an image \smash{$\HRFrame{k}$} with photometric variations and a reference image $\HR$ according to:
\begin{equation}
	\label{eqn:07_retinalImagingPhotometricModel}
	\HRFrame{k} = \BiasFieldMultFrame{k} \odot \HR + \BiasFieldAddFrame{k} \Ones.  
\end{equation}
\smash{$\BiasFieldMult \in \RealN{\HRSize}$}\label{notation:biasFieldMult} denotes a \textit{bias field} associated with \smash{$\HR \in \RealN{\HRSize}$} to consider uneven illumination of the retina relative to the reference, which is formulated pixel-wise in a multiplicative model. $\BiasFieldAdd \in \Real$\label{notation:biasFieldAdd} denotes a global brightness offset to describe intensity variations over time. Note that in this formulation, the set of photometric parameters $\PhotometricParams  = \{ \BiasFieldMult, \BiasFieldAdd \}$\label{notation:photometricParams} is defined in the domain of high-resolution data.

\begin{figure}[!t]
	\centering
	\subfloat[Frame 1]{\includegraphics[width=0.32\textwidth]{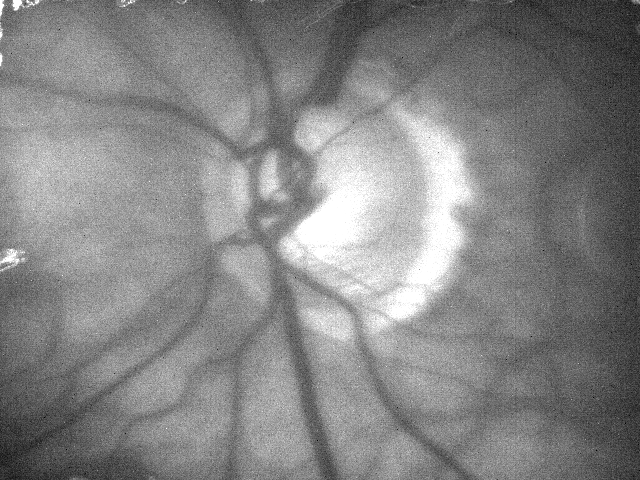}\label{fig:07_photometricChanges:frame1}}~
	\subfloat[Frame 18]{\includegraphics[width=0.32\textwidth]{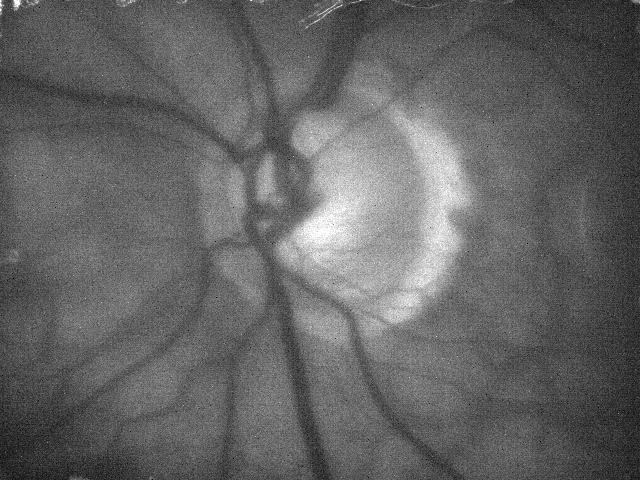}\label{fig:07_photometricChanges:frame18}}~
	\subfloat[Checkerboard overlay]{\includegraphics[width=0.32\textwidth]{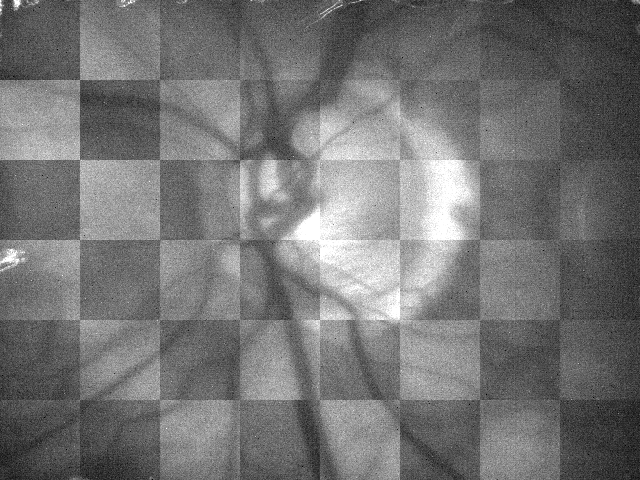}\label{fig:07_photometricChanges:checkerboard}}
	\caption[Photometric variations in retinal fundus video imaging]{Illustration of spatial and temporal photometric variations in fundus video imaging. \protect\subref{fig:07_photometricChanges:frame1} - \protect\subref{fig:07_photometricChanges:frame18} Two frames (frames 1 and 18) with eye movements and varying photometric conditions. \protect\subref{fig:07_photometricChanges:checkerboard} Corresponding checkerboard visualization. Notice the temporal brightness changes and the illumination inhomogeneities within both frames.}
	\label{fig:07_photometricChanges}
\end{figure}

\subsection{Joint Photogeometric and Sampling Model}

The formation of low-resolution video from a high-resolution image is described by combining the motion model in \eref{eqn:07_affineMotionModel} with the photometric one in \eref{eqn:07_retinalImagingPhotometricModel}. To describe this process in a physically appropriate manner, it is assumed that the geometric transformation related to eye movements takes place as the first operation, followed by the photometric transformation and the sampling onto the domain of the low-resolution data. We limit ourselves to space and time invariant modeling of the sampling process as a reasonable assumption for retinal images captured over a short time period with a high frame rate and a small \gls{fov}. Hence, the formation of the $k$-th low-resolution frame is described by:
\begin{equation}
	\LRFrame{k} = \SamplingMat \BlurMat \left( \BiasFieldMultFrame{k} \odot \left( \FrameIdx{\MotionMat}{k} \HR \right)  + \BiasFieldAddFrame{k} \Ones \right)
	+ \FrameIdx{\NoiseVec}{k},
	\label{eqn:07_retinalImageFormationModelAtmosphericEffects}
\end{equation}
where \smash{$\FrameIdx{\MotionMat}{k}$} denotes the subpixel motion related to eye movements, \smash{$\BiasFieldMultFrame{k}$} and \smash{$\BiasFieldAddFrame{k}$} are the photometric parameters, and \smash{$\FrameIdx{\NoiseVec}{k}$} denotes additive noise for the $k$-th frame. $\SamplingMat$ and $\BlurMat$ are time invariant and describe subsampling and a \gls{lsi} blur kernel. The latter approximates the superposition of the camera \gls{psf} and unavoidable optical aberrations in the human eye \cite{Marrugo2011}.

In \eref{eqn:07_retinalImageFormationModelAtmosphericEffects}, photometric variations are modeled as atmospheric effects in the domain of high-resolution data. However, under spatially smooth variations  as a common assumption of retrospective illumination correction \cite{Hou2006}, we can simplify \eref{eqn:07_retinalImageFormationModelAtmosphericEffects} using the approximations\footnote{Similar approximations regarding the photometric parameters have also been proposed for related image formation models in the field of blind deconvolution \cite{Marrugo2011}.}:
\begin{align}
	\SamplingMat \BlurMat \left( \BiasFieldMultFrame{k} \odot \FrameIdx{\MotionMat}{k} \HR \right)
	&\approx \SamplingMat \BiasFieldMultFrame{k} \odot \SamplingMat \BlurMat \FrameIdx{\MotionMat}{k} \HR, \\
	\BiasFieldAddFrame{k} \SamplingMat \BlurMat \Ones
	&\approx \BiasFieldAddFrame{k} \SamplingMat \Ones.
\end{align}
Based on these approximations, \eref{eqn:07_retinalImageFormationModelAtmosphericEffects} can be rewritten to explain the formation of low-resolution frames from a high-resolution image according to:
\begin{equation}
	\LRFrame{k} = \BiasFieldMultFrame{k} \odot \SystemMatFrame{k} \HR + \BiasFieldAddFrame{k} \Ones
	+ \FrameIdx{\NoiseVec}{k},
	\label{eqn:07_retinalImageFormationModel}
\end{equation}
where the system matrix \smash{$\SystemMatFrame{k}$} is assembled element-wise from the blur kernel and motion parameters (see \eref{eqn:systemMatrixElements}). Notice that in \eref{eqn:07_retinalImageFormationModel} the photometric parameters \smash{$\BiasFieldMultFrame{k}$ and $\BiasFieldAddFrame{k}$} are defined in the domain of low-resolution frames.

\section{Super-Resolution with Quality Self-Assessment}
\label{sec:07_SuperResolutionWithQualitySelfAssessment}

The proposed framework aims at reconstructing an eye movement compensated high-resolution fundus image from multiple low-resolution video frames while simultaneously compensating photometric variations. The algorithm developed in this section is divided into a registration and a reconstruction stage as follows: 
\begin{enumerate}
	\item Given a sequence of low-resolution frames, a \textit{photogeometric registration} is accomplished to estimate the latent eye motion as well as the photometric parameters that describe the image formation process according to \eref{eqn:07_retinalImageFormationModel}.
	\item Given the estimate of the photogeometric model, a high-resolution image is reconstructed by \gls{map} estimation from the low-resolution frames. 
\end{enumerate}
First, this section presents photogeometric registration for fundus videos employed in the initial registration stage. Subsequently, an iterative optimization scheme is developed referred to as super-resolution with quality self-assessment to jointly estimate latent Bayesian hyperparameters along with the high-resolution image. Eventually, a tailored quality measure for a fully automatic assessment of image noise and sharpness within quality self-assessment is proposed.

\subsection{Photogeometric Registration Algorithm}

Photogeometric registration is accomplished in two steps. First, the photometric parameters that are related to spatial and temporal illumination variations are estimated from the observed, low-resolution frames. Second, the geometric parameters that describe eye movements are determined under consideration of the photometric parameters.

\paragraph{Photometric Registration.}
Given the set of low-resolution frames $\LRSequence{1}{\NumFrames}$, the multiplicative bias fields that describe spatial photometric variations are estimated for each frame separately. Following state-of-the-art retrospective illumination correction techniques \cite{Hou2006}, a bias field is assumed to be a spatially smooth signal. In this chapter, a parametric approach is employed that represents a bias field \smash{$\BiasFieldMultFrame{k}$} by the superposition of spatially smooth basis functions as a B-spline surface \cite{Kolar2011}. Hence, \smash{$\BiasFieldMultFrame{k}$} is computed by B-spline fitting of the intensities in the corresponding low-resolution frame \smash{$\LRFrame{k}$}, which implicitly enforces the smoothness condition.

The brightness offsets \smash{$\BiasFieldAddFrame{k}$} that describe temporal photometric variations are determined by pair-wise registration. This is done in a robust manner by computing the median temporal brightness of each frame \smash{$\LRFrame{k}$} relative to the reference \smash{$\LRFrame{r}$} under the estimated bias fields for both frames according to:
\begin{equation}
	\BiasFieldAddFrame{k} = 
	\Median{\left(\BiasFieldMultFrame{k}\right)^{-1} \odot \LRFrame{k}} - \Median{\left(\BiasFieldMultFrame{r}\right)^{-1} \odot \LRFrame{r}},
\end{equation}
where $(\BiasFieldMultFrame{k})^{-1}$ and $(\BiasFieldMultFrame{r})^{-1}$ are the pixel-wise inverted bias fields of the $k$-th frame and the reference frame, respectively.

\paragraph{Geometric Registration.}
Once the photometric parameters  are determined, the geometric parameters that describe eye movements by the affine image-to-image homography in \eqref{eqn:07_affineMotionModel} are obtained by pair-wise registration. For the $k$-th low-resolution frame \smash{$\LRFrame{k}$}, these motion parameters are determined as the solution of the intensity-based registration problem:
\begin{equation}
	\label{eqn:07_geometricRegistration}
	\MotionParamsFrame{k} = \argmax_{\MotionParams} 
	\SimMeasure{ \MotionOpEmpty_{\MotionParams} \left\{ \left(\BiasFieldMultFrame{k}\right)^{-1} \odot \LRFrame{k} - \BiasFieldAddFrame{k} \Ones \right\}} 
	{\left(\BiasFieldMultFrame{r}\right)^{-1} \odot \LRFrame{r} },
\end{equation}
where $\MotionOpEmpty_{\MotionParams}\{  \cdot\}$ denotes image warping towards the reference frame according to the motion parameters $\MotionParams$ and $\SimMeasureSym: \RealN{\LRSize} \times \RealN{\LRSize} \rightarrow \RealNonNeg$ is an image similarity measure. 

It is worth noting that \eref{eqn:07_geometricRegistration} compensates for photometric variations using the photometric parameters in the similarity measure. In order to enhance the robustness of the geometric registration regarding residual variations, the normalized cross correlation is used for the similarity measure. This pair-wise registration is implemented by \gls{ecc} optimization \cite{Evangelidis2008}, which iteratively solves \eref{eqn:07_geometricRegistration} for the motion parameters in a coarse-to-fine scheme.

\subsection{Super-Resolution Reconstruction Algorithm}

Given the low-resolution observations $\LR$ and the photogeometric registration parameters \smash{$\{ \FrameIdx{\MotionParams}{k}, \FrameIdx{\PhotometricParams}{k}\}_{k = 1}^\NumFrames$}, the high-resolution image $\HR$ is inferred according to the \gls{map} estimation:
\begin{equation}
	\HR_{\text{MAP}} = \argmax_{\HR}
	\PdfCond{\LR}{\HR, \{ \FrameIdx{\MotionParams}{k}, \FrameIdx{\PhotometricParams}{k} \}_{k = 1}^\NumFrames } \Pdf{\HR},
	\label{eqn:07_retinalImagingSrMapEstimation}
\end{equation}
where $\PdfCondT{\LR}{\HR, \{ \FrameIdx{\MotionParams}{k}, \FrameIdx{\PhotometricParams}{k} \}_{k = 1}^\NumFrames }$ denotes the observation model related to \eref{eqn:07_retinalImageFormationModel}. For the prior distribution, we assign the exponential form $\PdfT{\HR} \propto \exp(- \RegWeight \RegTerm{\HR})$ with regularization term $\RegTerm{\HR}$ and regularization weight $\RegWeight \geq 0$. 

Notice that \gls{map} estimation in \eref{eqn:07_retinalImagingSrMapEstimation} requires prior knowledge regarding the regularization weight $\RegWeight$. However, in the desired application, the choice of this parameter strongly depends on the imaging conditions. Thus, there might be a considerable variance regarding the optimal parameter for video data of different subjects, which makes its off-line selection on training data difficult. To this end, the regularization weight is treated as a latent hyperparameter. In contrast to \textit{data} driven hyperparameter selection (see \sref{sec:04_RobustSuperResolutionReconstruction}), the regularization weight is inferred in a \textit{quality} driven way termed image quality self-assessment. This optimization scheme employs a quality measure that quantifies image noise and sharpness. This enables a fully automatic parameter selection and provides an objective quality measure for super-resolution as a by-product of the optimization algorithm.

\begin{algorithm}[!t]
	\small
	\caption{Super-resolution with image quality self-assessment}
	\label{alg:07_srWithQualitySelfAssessment}
	\begin{algorithmic}[1]
		\Require Initial guess for image $\HR$ and regularization weight search range $[\log\RegWeight_l, \log\RegWeight_u]$
		\Ensure Final high-resolution image $\HR$ with optimal regularization weight $\hat{\RegWeight}$
		\State $\RegWeight \gets \RegWeight_l$ and $Q_{\text{max}} \gets 0$
		\While{$\RegWeight \leq \RegWeight_u$}
			\For{$t = 1, \ldots, T_{\text{scg}}$}
				\State Update $\HR$ by SCG iteration for \eref{eqn:07_retinalImagingSrObjectiveFunction} with current $\RegWeight$
			\EndFor
			\If{$Q(\HR) > Q_{\text{max}}$}
				\State $\hat{\RegWeight} \gets \RegWeight$ and $Q_{\text{max}} \gets Q(\HR)$ 
			\EndIf
			\State $\RegWeight \gets 10^{\log \RegWeight + \Delta \RegWeight}$
		\EndWhile
		\While{SCG convergence criterion not fulfilled}
			\State Update $\HR$ by SCG iteration for \eref{eqn:07_retinalImagingSrObjectiveFunction} with $\RegWeight = \hat{\RegWeight}$
		\EndWhile
	\end{algorithmic}
\end{algorithm}

The estimations of the high-resolution image and the optimal regularization weight are treated as two coupled subproblems. Our goal is to infer the regularization weight $\RegWeight$ according to: 
\begin{equation}
	\label{eqn:07_qualitySelfAssessmentObjective}
	\hat{\RegWeight} = \argmax_{\RegWeight} Q \big( \HR(\RegWeight) \big),
\end{equation}
where $Q: \RealN{\HRSize} \rightarrow \RealNonNeg$ denotes a \textit{no-reference} image quality measure that quantifies the level of noise and sharpness for a given image. Note that higher measures $Q(\HR)$ indicates a favorable image quality. Given the regularization weight $\RegWeight$, we denote by $\HR(\RegWeight)$ the image reconstructed under this parameter according to the minimization:
\begin{equation}
	\HR(\RegWeight) = \argmin_{\HR} \left\{ \DataTerm{\HR} + \RegWeight \RegTerm{\HR} \right\},
	\label{eqn:07_retinalImagingSrObjectiveFunction}
\end{equation}
where:
\begin{equation}
	\DataTerm{\HR} 
	= \sum_{k = 1}^\NumFrames 
	\LossFun[\text{data}]{ \LRFrame{k} - \BiasFieldMultFrame{k} \odot \SystemMatFrame{k} \HR - \BiasFieldAddFrame{k} \Ones },
\end{equation}
and $\LossFunSym[\text{data}]: \RealN{\LRSize} \rightarrow \RealNonNeg$ is a loss function related to the underlying noise model. It is worth noting that this optimization scheme is independent on the implementations of the observation and prior model. Hence, we omit their definitions in this general derivation. 

For the joint estimation of the optimal regularization weight and the high-resolution image with consideration of their interdependence, the proposed algorithm nests the solution of \eref{eqn:07_qualitySelfAssessmentObjective} and \eref{eqn:07_retinalImagingSrObjectiveFunction}, see \aref{alg:07_srWithQualitySelfAssessment}. In order to determine the regularization weight, \eref{eqn:07_qualitySelfAssessmentObjective}  is approximated by a one-dimensional discrete search. The reconstruction of the desired high-resolution image in \eref{eqn:07_retinalImagingSrObjectiveFunction} is accomplished by $T_{\text{scg}}$ iterations of \gls{scg} optimization. This seeks the stationary point:
\begin{equation}
	\nabla_{\HR} \DataTerm{\HR} + \RegWeight \nabla_{\HR} \RegTerm{\HR} = \Zeros,
\end{equation}
where the gradient of the data fidelity term is given by:
\begin{equation}
	\nabla_{\HR} \DataTerm{\HR} =
	\sum_{k = 1}^\NumFrames \BiasFieldMultFrame{k} \odot \SystemMatFrame{k}
	\psi_{\text{data}} \left( \LRFrame{k} - \BiasFieldMultFrame{k} \odot \SystemMatFrame{k} \HR - \BiasFieldAddFrame{k} \Ones \right),
\end{equation}
and $\psi_{\text{data}}(\vec{z}) = \nabla_{\HR} \LossFun[\text{data}]{\vec{z}}$ denotes the gradient of $\LossFun[\text{data}]{\vec{z}}$. For this optimization scheme, the initial guess for the high-resolution image is obtained by the temporal median of the geometrically and photometrically registered low-resolution frames followed by bicubic interpolation according to the desired magnification factor. The regularization weight is initialized by the log-transformed search range $[\log \RegWeight_l, \log \RegWeight_u]$. Once the optimal regularization weight $\hat{\RegWeight}$ under the given quality measure is determined, the high-resolution image $\HR(\hat{\RegWeight})$ is refined by \gls{scg} iterations for \eref{eqn:07_retinalImagingSrObjectiveFunction} until convergence.

\subsection{No-Reference Quality Measure for Retinal Imaging}

Quality self-assessment requires a reliable measure of image quality. In general, such measures can be divided into classification based and continuous scores. The classification based approach aims at predicting a discrete quality measure from discriminative image features. In the simplest case, this is restricted to two classes to discriminate low-quality images from good ones. In retinal imaging, this can be achieved by supervised learning using features of diagnostic significance \cite{Niemeijer2006a,Paulus2010a}. However, these classification methods are not well suited for continuous optimization within the proposed quality self-assessment scheme. 

In contrast to supervised learning, continuous quality measures are inferred from low-level features in an unsupervised way such that the resulting measure correlates with the human visual perception. Some well known features are the image entropy \cite{Gabarda2007}, spatial and spectral properties \cite{Vu2012}, or the image gradient \cite{Zhu2010}. Such features have previously been used for retinal image analysis \cite{Marrugo2011b}. In this work, we focus on continuous measures that are applicable for quality self-assessment.

\paragraph{Derivation of the Quality Measure.}
The quality measure that is employed in this work is based on the \textit{coherence} feature proposed by Zhu \etal \cite{Zhu2010} for natural images that has been later adopted to retinal imaging by K\"ohler \etal \cite{Kohler2013}. To derive the measure for a given image $\HR$, we decompose $\HR$ in disjoint \smash{$N_p \times N_p$} patches \smash{$\vec{p} \in \RealN{N_p^2}$}\label{notation:imagePatch} with domain $\Domain(\vec{p}) \subset \RealN{2}$. We aim at quantifying noise and sharpness based  on two features that are related to the image gradient and the curvature. 

In terms of the gradient information as proposed in \cite{Zhu2010}, a local gradient matrix is constructed for the patch $\vec{p}$ according to:\label{notation:gradientMatrix}
\begin{equation} 
	\vec{G}(\vec{p}) = 
	\begin{pmatrix}
		\VecEl{\HPMat[\CoordU] \vec{p}}{1}	& \VecEl{\HPMat[\CoordV] \vec{p}}{1} \\
		\vdots & \vdots \\
		\VecEl{\HPMat[\CoordU] \vec{p}}{N_p^2}	& \VecEl{\HPMat[\CoordV] \vec{p}}{N_p^2}
	\end{pmatrix},
\end{equation}
where $\HPMat[\CoordU]$ and $\HPMat[\CoordV]$ denote discrete derivative filters for the coordinate directions $\CoordU$ and $\CoordV$, respectively. For this local gradient matrix, we calculate the \gls{svd}:
\begin{equation}
	\vec{G}(\vec{p})	=  \vec{U}(\vec{p}) \begin{pmatrix} s_1(\vec{p}) & 0 \\ 0 & s_2(\vec{p}) \end{pmatrix} \vec{V}(\vec{p})^\top,
\end{equation}
where $\vec{U}(\vec{p})$ and $\vec{V}(\vec{p})$ are orthogonal matrices, and $s_1(\vec{p})$ and $s_2(\vec{p})$\label{notation:gradientMatrixSingularValues} denote the singular values of the gradient matrix associated with the patch $\vec{p}$. In the noise and sharpness measure presented below, the singular values are used as basic features.

\begin{figure}[!t]
	\centering
	\includegraphics[width=1.00\textwidth]{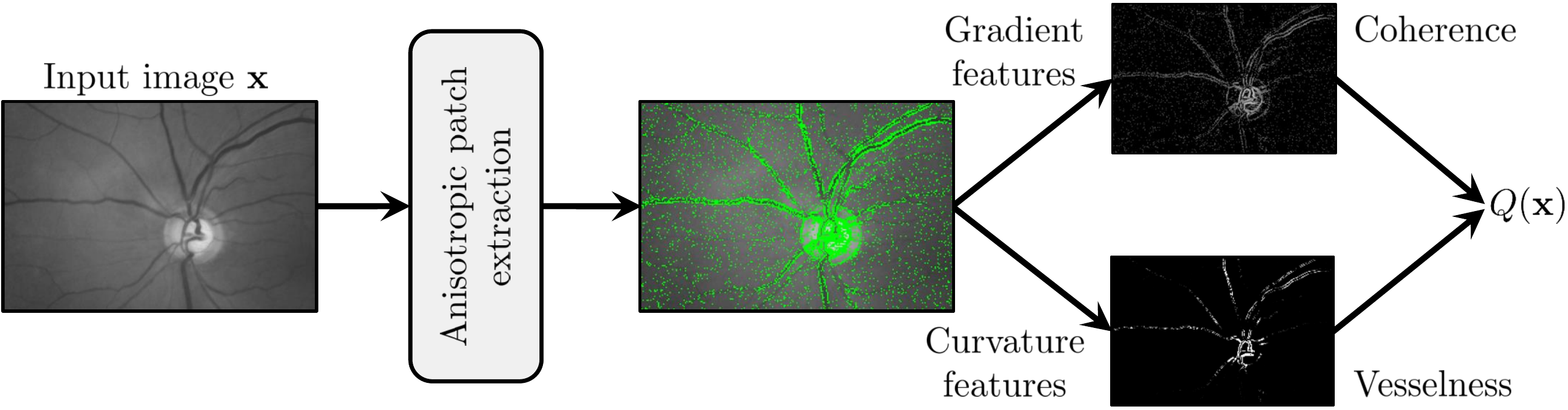}
	\caption[No-reference quality measure for retinal fundus images]{Computation of the no-reference quality measure $Q(\HR)$ for retinal fundus images. For the given input image $\HR$, anisotropic patches are detected. Afterwards, $Q(\HR)$ is determined on a patch level from the coherence and the vesselness features that are obtained from the image gradient and the curvature, respectively.}
	\label{fig:07_qualityAssessmentFlowchart}
\end{figure}

In terms of the curvature information as proposed in \cite{Kohler2013}, we compute the Hessian matrix in a pixel-wise manner according to:\label{notation:hessian}
\begin{equation}
	\vec{H}_i (\sigma_j) =
	\begin{pmatrix}
		\VecEl{\HPMat[\CoordU\CoordU](\sigma_j) \HR}{i} & & \VecEl{\HPMat[\CoordU\CoordV](\sigma_j) \HR}{i} \\[0.5em]
		\VecEl{\HPMat[\CoordU\CoordV](\sigma_j) \HR}{i} & & \VecEl{\HPMat[\CoordV\CoordV](\sigma_j) \HR}{i}
	\end{pmatrix},
\end{equation}
where $\HPMat[\CoordU\CoordU](\sigma_j)$, $\HPMat[\CoordV\CoordV](\sigma_j)$ and $\HPMat[\CoordU\CoordV](\sigma_j)$ denote discrete Laplacian of Gaussian filters with the kernel standard deviation $\sigma_j$\label{notation:logKernelStd} for the coordinate directions $\CoordU$ and $\CoordV$. The Hessian is employed to determine the \textit{vesselness}, which represents a probability map that enables the detection of tubular structures. In retinal imaging, the vesselness provides a blood vessel detection and image quality assessment is steered by the detected vessel tree. The vesselness filter used in this work is based on the approach of Frangi \etal \cite{Frangi1998} to detect dark tubular structures\footnote{Without loss of generality, we limit our consideration to dark tubular structures as these structures correspond to blood vessels in fundus images.} according to:\label{notation:vesselnessSingleScale}
\begin{equation}
	\begin{split}
		&V_i(\sigma_j) = \\
		&\begin{cases}
			\exp{\left( - \frac{1}{2 V_{\beta}^2} \frac{\lambda_{1,i}(\sigma_j)^2}{\lambda_{2,i}(\sigma_j)^2} \right)} 
			\left( 1 - \exp{ \left( - \frac{ \lambda_{1,i}(\sigma_j)^2 + \lambda_{2,i}(\sigma_j)^2 }{2 V_{c}^2} \right) } \right)
				& \text{if}~ \lambda_{1,i}(\sigma_j) \geq 0 \\
			0	
				& \text{otherwise}
		\end{cases},
	\end{split}
\end{equation}
where $\lambda_{1,i}(\sigma_j)$ and $\lambda_{2,i}(\sigma_j)$\label{notation:hessianEigenvalues} with $|\lambda_{1,i}(\sigma_j)| \leq |\lambda_{2,i}(\sigma_j)|$ are the eigenvalues of the Hessian at the $i$-th pixel associated with the kernel standard deviation $\sigma_j$. The parameters $V_{\beta}$ and $V_c$ are thresholds to control the vesselness filter response. The filter responses over a set of $N_{\sigma}$ kernel standard deviations are used to compute the local variance of the vesselness in the patch $\vec{p}$ according to:\label{notation:vesselness}\label{notation:vesselnessVariance}
\begin{align}
	V(\vec{p}) &= \frac{1}{N_p^2} \sum_{i \in \Domain(\vec{p})} \left( V_i^* - \frac{1}{N_p^2} \sum_{i \in \Domain(\vec{p})} V_i^* \right)^2,\\
	V_i^* &= \max_{j = 1, \ldots, N_{\sigma}} V_i(\sigma_j).
\end{align}

The proposed quality measure combines both feature types, \ie the gradient information and the vesselness, to assess the level of noise and sharpness in a given image, see \fref{fig:07_qualityAssessmentFlowchart}. In principle, noise and sharpness in a patch $\vec{p}_i$ are characterized by the singular values $s_1(\vec{p}_i)$ and $s_2(\vec{p}_i)$ of the local gradient matrix \cite{Zhu2010}. The local variance $V(\vec{p}_i)$ is used to guide this measurement based on the hypothesis that patches $\vec{p}_i$ located on the boundaries of tubular structures, \ie blood vessels, should have a higher contribution to the local quality measure. We compute the local quality $q(\vec{p}_i)$ associated with the patch $\vec{p}_i$ according to:\label{notation:coherence}\label{notation:localQualityMeasure}  
\begin{equation}
	q(\vec{p}_i) = V(\vec{p}_i) s_1(\vec{p}_i) c(\vec{p}_i),
\end{equation}
where $c(\vec{p}_i)$ denotes the coherence that is computed from the singular values:
\begin{equation}
	c(\vec{p}_i) = \frac{s_1(\vec{p}_i) - s_2(\vec{p}_i) }{s_1(\vec{p}_i) + s_2(\vec{p}_i)}.
\end{equation}
Then, the global quality measure $Q(\HR)$ for the entire image is given by:\label{notation:globalQualityMeasure}
\begin{equation}
	Q(\HR) = \sum_{\vec{p}_i \in \mathcal{A}(\HR)} q(\vec{p}_i),
\end{equation}
where $\mathcal{A}(\HR)$ denotes a set of \textit{anisotropic} patches. These anisotropic patches are characterized by a dominant orientation of the image gradient and are meaningful to characterize noise and sharpness. In accordance to \cite{Zhu2010}, these patches are detected automatically by statistical significance testing of the local coherence \smash{$c(\vec{p}_i)$}. This leads to the thresholding procedure:\label{notation:anisoSet}
\begin{align}
	\mathcal{A}(\HR) &= \left\{ \vec{p}_i ~:~ c(\vec{p}_i) \geq \tau_c \right\}, \\
	\tau_c &= \sqrt{ \left( 1 - \alpha_c^\frac{1}{{N_p^2 - 1}} \right) \left( 1 + \alpha_c^\frac{1}{N_p^2 - 1} \right)^{-1} },
\end{align}
with threshold $\tau_c$ that is determined from the significance level $\alpha_c$\label{notation:anisoSetLevel}.

\paragraph{Correlation to Full-Reference Quality Assessment.}
Let us now investigate the validity of the proposed measure for image quality self-assessment. For this purpose, the agreement of the no-reference quality measure $Q(\HR)$ to full-reference quality assessment is studied on simulated data. \Fref{fig:07_noReferenceToPSNRCorrelation} shows the progress of the no-reference measure over the search range of the unknown regularization parameter $\RegWeight$ in \aref{alg:07_srWithQualitySelfAssessment} averaged over 40 simulated fundus image sequences. In addition, the \gls{psnr} for the super-resolved images associated with the different parameter settings is depicted as an example full-reference measure. The relationship between both measures confirms a reasonable agreement between no-reference and full-reference assessment. Note that both measures result in comparable solutions in terms of the optimal regularization weight. Furthermore, a Spearman rank correlation of $0.70 \pm 0.34$ averaged over all simulated datasets indicates a reasonable correlation between both measures. This validates the proposed no-reference measure as a surrogate for full-reference quality assessment in the absence of ground truth data. For a comprehensive evaluation of the no-reference measure in retinal fundus imaging and comparisons to other state-of-the-art methods, we refer to \cite{Kohler2013}.
\tikzexternaldisable
\begin{figure}[!t]
	\centering
	\scriptsize 
	\setlength \figurewidth{0.73\textwidth}
	\setlength \figureheight{0.58\figurewidth}
	\subfloat{\input{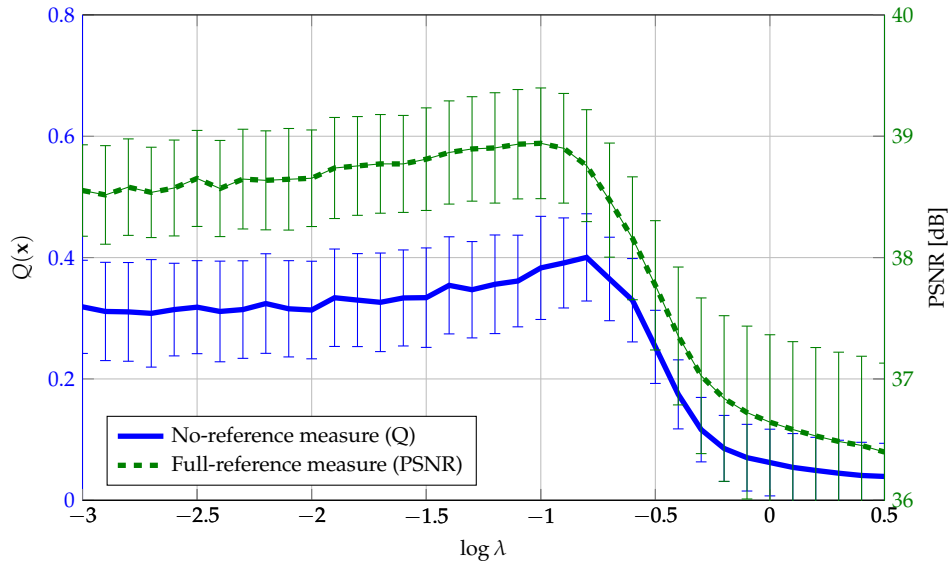}}
	\caption[Correlation between no-reference and full-reference quality assessment]{Correlation analysis between no-reference and full-reference quality assessment. Blue, solid line: mean $\pm$ standard deviation of the no-reference measure $Q(\HR)$ used for quality self-assessment versus the regularization weight $\RegWeight$ on 40 image sequences. Green, dotted line: mean $\pm$ standard deviation of the \gls{psnr} relative to the ground truth. Both measures reach their optimal value within the range $-1.0 \leq \log \RegWeight \leq -0.8$. The Spearman rank correlation between both measures is $0.70 \pm 0.34$.}
	\label{fig:07_noReferenceToPSNRCorrelation}
\end{figure}
\tikzexternalenable

\section{Experiments and Results}
\label{sec:07_ExperimentsAndResults}

The experimental evaluation for the proposed framework is divided into three parts. In the first part, super-resolution is quantitatively evaluated on simulated fundus images to investigate the potential of the proposed framework in retinal imaging. The second part addresses real data experiments with the target to gain high-resolution fundus images from low-resolution video sequences acquired with a low-cost camera. The third part presents \textit{super-resolved mosaicing} \cite{Kohler2016} as a novel application of super-resolution in ophthalmic imaging workflows.  

Throughout all experiments, super-resolution was applied with the \LOne norm error model and a \gls{btv} prior with $\BTVSize = 1$ and $\BTVWeight = 0.4$. The regularization weight selection was performed in the search range given by $\log \RegWeight_l = -3.0$ and $\log \RegWeight_u = 0$ with $\Delta \log \RegWeight = 0.15$ and $T_{\text{scg}} = 50$ \gls{scg} iterations. Quality assessment was performed with patch size $N_p = 8$ and significance level $\alpha_c = 0.001$ to detect anisotropic patches. The vesselness filter parameters were set to $V_{\beta} = 0.5$ and $V_c = 15$ for $N_{\sigma} = 4$ different filter standard deviations $\sigma_i \in \{1, 3, 5, 8\}$.

\subsection{Experiments on Simulated Fundus Images}

For the sake of a quantitative evaluation, simulated images generated from the DRIVE database \cite{Staal2005} were used. For this task, excerpts of 40 reference images of size $360 \times 360$\,px served as ground truth data. The green color channels were used to generate sequences of $\NumFrames = 15$ monochromatic frames of size $120 \times 120$\,px from the reference color images. Eye movements were simulated by uniformly distributed inter-frame translations ($-4$ to $+4$ px) and rotation angles ($-1^\circ$ to $+1^\circ$) relative to a the first frame. The formation of each frame considered a Gaussian \gls{psf} ($\PSFWidth = 0.5$) and additive, zero-mean Gaussian noise ($\NoiseStd = 0.01$). 

The impact of super-resolution was quantitatively assessed by four evaluation measures. On the one hand, \gls{psnr} and \gls{ssim} were used to assess the fidelity of a reconstruction relative to the ground truth. On the other hand, super-resolution was studied in combination with automatic blood vessel segmentation. This was done by applying the proposed framework as preprocessing for the state-of-the-art segmentation method introduced by Budai \etal \cite{Budai2013a}. Super-resolution was assessed by analyzing the sensitivity and specificity of the automatic segmentation relative to a gold standard segmentation provided by a human expert. The statistics of these measures  are summarized in \tref{tab:07_syntheticDataResults} for simulated low-resolution data, the initial guess for the iterative super-resolution algorithm obtained by temporal median filtering as well as the final super-resolved image. On average, super-resolution improved the \gls{psnr} by 3.5\,\gls{db} and the \gls{ssim} by 0.07 compared to the original video data. The sensitivity of vessel segmentation was enhanced by 10\,\% at a comparable specificity in comparison to a direct segmentation on the low-resolution data. This reveals the potential performance boost achieved by super-resolution in this application. 

\Fref{fig:07_syntheticFundusDataExample} compares low-resolution data and super-resolution on one example dataset along with the corresponding vessel segmentations. Notice that the gain of super-resolution is revealed by a recovery of fine structures on the retina, \eg blood vessels, that are barely visible in low-resolution data. Consequently, vessel segmentation achieved a higher sensitivity based on preprocessing by means of super-resolution compared to a segmentation on low-resolution data.
\begin{table}[!t]
	\centering
	\caption[Performance of super-resolution on the DRIVE database]{Performance of super-resolution on the DRIVE database \cite{Staal2005}. The \gls{psnr} and \gls{ssim} statistics were determined for 40 simulated images relative to ground truth data. The sensitivity and specificity statistics were determined for automatic vessel segmentation \cite{Budai2013a} relative to a gold standard. All measures were evaluated for low-resolution data, the initial guess for super-resolution, and the final super-resolved image.}
	\small
	\begin{tabular}{lcccc}
		\toprule
		& \textbf{Original} & \textbf{SR (initial)}	& \textbf{SR (final)}	& \textbf{Ground truth} \\
		\cmidrule {2-5}
		\gls{psnr} [\gls{db}] 	& 35.19 $\pm$ 1.07	& 36.65 $\pm$ 1.55						& \textbf{38.64 $\pm$ 1.00} & - \\
		\gls{ssim} 							& 0.84 $\pm$ 0.01		& 0.89 $\pm$ 0.02							& \textbf{0.91 $\pm$ 0.01}  & - \\
		\midrule
		Sensitivity [\%]	& 59.00 $\pm$ 6.08 	& 66.66 $\pm$ 4.95 						& 69.41 $\pm$ 5.49  				& \textbf{74.96 $\pm$ 5.87} \\
		Specificity [\%]	& 93.13 $\pm$ 1.26 	& \textbf{95.04 $\pm$ 1.02} 	& 94.44 $\pm$ 1.27  				&  94.48 $\pm$ 1.16 \\
		\bottomrule
	\end{tabular}
	\label{tab:07_syntheticDataResults}
\end{table}
\begin{figure}[!t]
	\centering
	\subfloat[Original]{\includegraphics[width=0.236\textwidth]{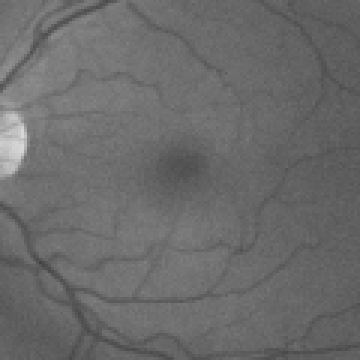}\label{fig:07_syntheticFundusDataExample:lr}}~
	\subfloat[SR (initial guess)]{\includegraphics[width=0.236\textwidth]{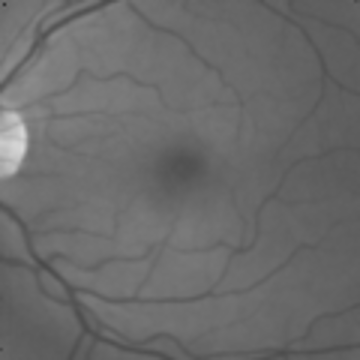}}~
	\subfloat[SR (final)]{\includegraphics[width=0.236\textwidth]{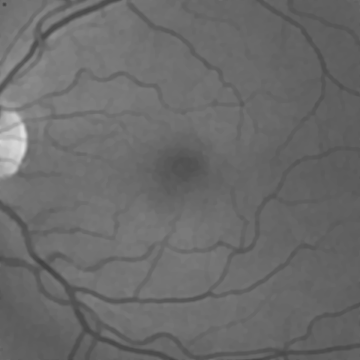}}~
	\subfloat[Ground truth]{\includegraphics[width=0.236\textwidth]{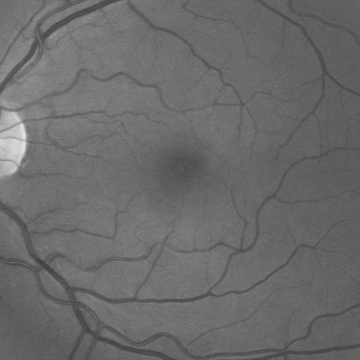}\label{fig:07_syntheticFundusDataExample:gt}}\\[-0.3em]
	\subfloat[Se: 0.64, Sp: 0.93]{\includegraphics[width=0.236\textwidth]{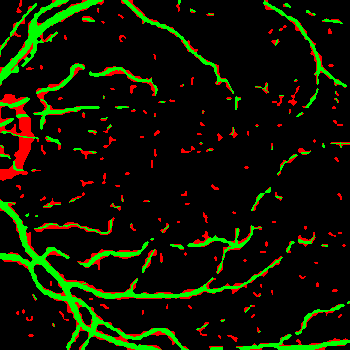}\label{fig:07_syntheticFundusDataExample:lrSeg}}~
	\subfloat[Se: 0.71, Sp: 0.95]{\includegraphics[width=0.236\textwidth]{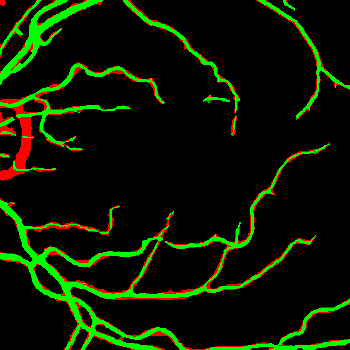}}~
	\subfloat[Se: 0.75, Sp: 0.95]{\includegraphics[width=0.236\textwidth]{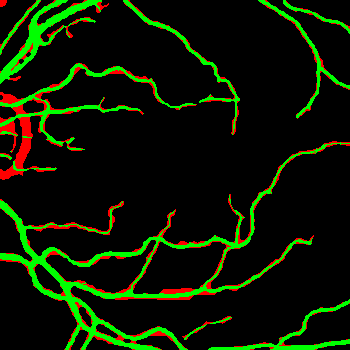}}~
	\subfloat[Se: 0.82, Sp: 0.95]{\includegraphics[width=0.236\textwidth]{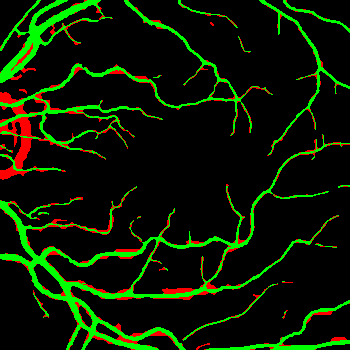}\label{fig:07_syntheticFundusDataExample:gtSeg}}\\
	\caption[Super-resolution and automatic blood vessel segmentation on fundus images]{Super-resolution on simulated fundus images generated from the DRIVE database \cite{Staal2005} ($\NumFrames = 15$ frames, magnification $s = 3$). \protect\subref{fig:07_syntheticFundusDataExample:lr} - \protect\subref{fig:07_syntheticFundusDataExample:gt}  Low-resolution data, the initial guess (temporal median), the super-resolved image, and the ground truth image. \protect\subref{fig:07_syntheticFundusDataExample:lrSeg} - \protect\subref{fig:07_syntheticFundusDataExample:gtSeg} Blood vessel segmentation \cite{Budai2013a} along with the sensitivity (Se) and specificity (Sp). True-positive and false-positive pixels are color-coded in green and red, respectively.}
	\label{fig:07_syntheticFundusDataExample}
\end{figure}

\subsection{Experiments on Real Fundus Videos}

In order to conduct experiments on real images, we used fundus video data captured with the low-cost and mobile camera developed by Tornow \etal \cite{Tornow2015}. This camera system is based on a monochromatic \gls{ccd} sensor and provides a spatial resolution of 640$\times$480\,px, a \gls{fov} of $20^\circ$ in horizontal direction and a temporal resolution of up to $50$\,Hz. In this study, the left eyes of different human subjects including healthy subjects and glaucoma patients were examined. All examinations were done without dilating the pupil, \ie non-mydriatically. The acquired video sequences have durations between 5 and 15\,seconds\footnote{The data acquisition for this study was done in collaboration with Dr.-Ing. Ralf-Peter Tornow at the Department of Ophthalmology, Eye Clinics Erlangen, Germany}. Super-resolution was applied on subsequences extracted from these videos by processing $\NumFrames = 8$ successive frames in a sliding window approach with magnification $\MagFac = 2$. Throughout all experiments, the unknown \gls{psf} was approximated by an isotropic Gaussian kernel ($\PSFWidth = 0.8$).

\paragraph{Comparison to High-Resolution Reference Images.}
For the sake of a qualitative comparison to super-resolved data, a commercially available Kowa nonmyd camera\footnote{\url{http://www.kowamedical.com/}} was employed to capture color fundus images. This single-shot camera features a spatial resolution of 1600$\times$1216\,px with a \gls{fov} of $25^\circ$ and was used to gain high-resolution reference photographs of the same subjects that were examined with the low-cost camera. For fair comparisons to monochromatic video data, the green channels of the color photographs were used in this study.

\tikzexternaldisable
\afterpage{
\begin{landscape}
\begin{figure}[!p]
	\centering	
	\subfloat[Single frame (w/o photometric registration)]{\label{fig:07_lowCostFundus_glaucoma:lr}
		\begin{tikzpicture}[spy using outlines={rectangle,red,magnification=2.0,height=2.8cm, width=2.8cm, connect spies, every spy on node/.append style={thick}}] 
			\node {\pgfimage[height=0.229\linewidth]{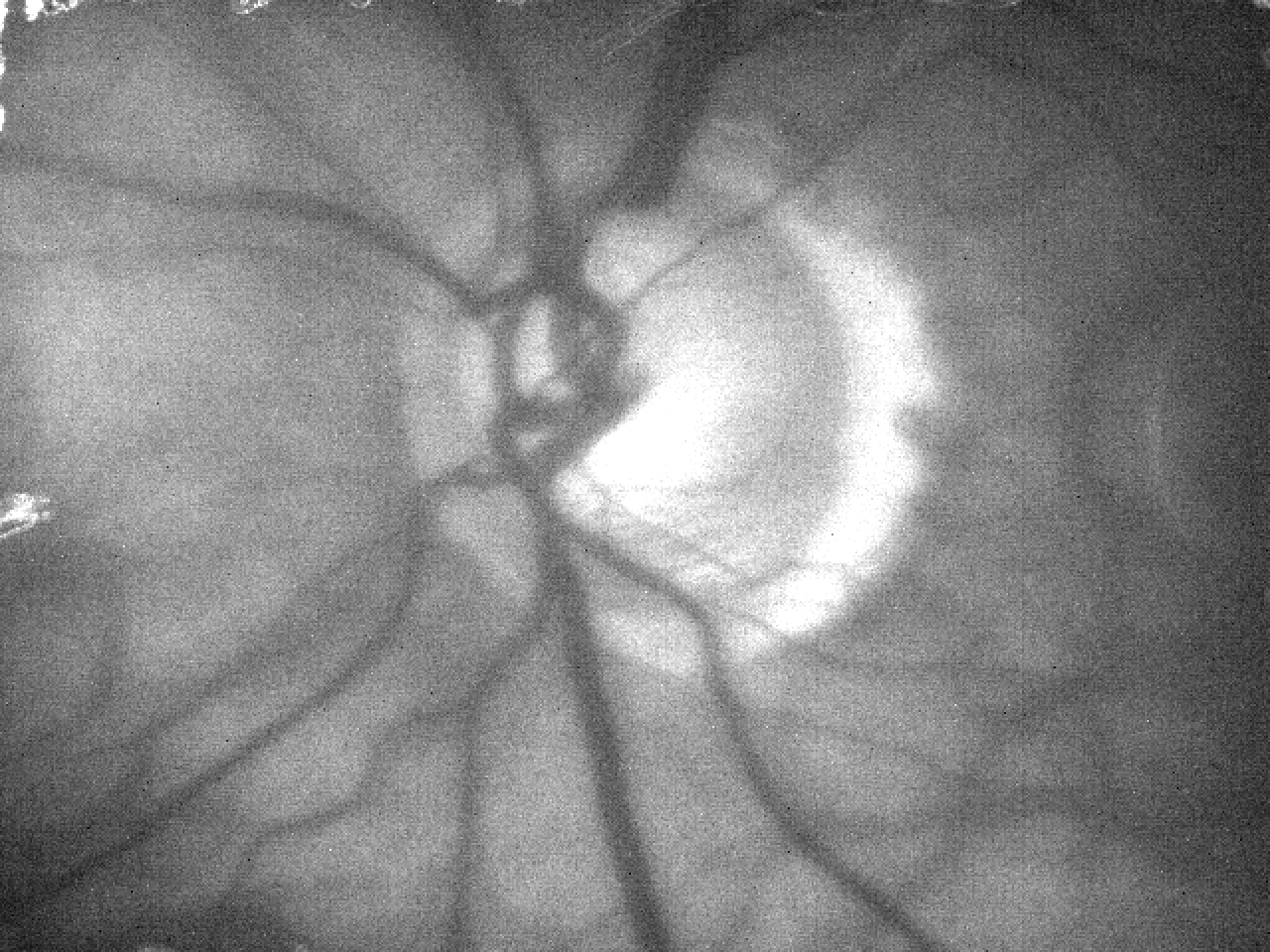}}; 
      \spy on (0.1,-0.3) in node [left] at (3.74, -1.5); 
    \end{tikzpicture} 
	}\hspace{-0.8em}
	\subfloat[Single frame (w/ photometric registration)]{\label{fig:07_lowCostFundus_glaucoma:lrReg}
		\begin{tikzpicture}[spy using outlines={rectangle,red,magnification=2.0,height=2.8cm, width=2.8cm, connect spies, every spy on node/.append style={thick}}] 
			\node {\pgfimage[height=0.229\linewidth]{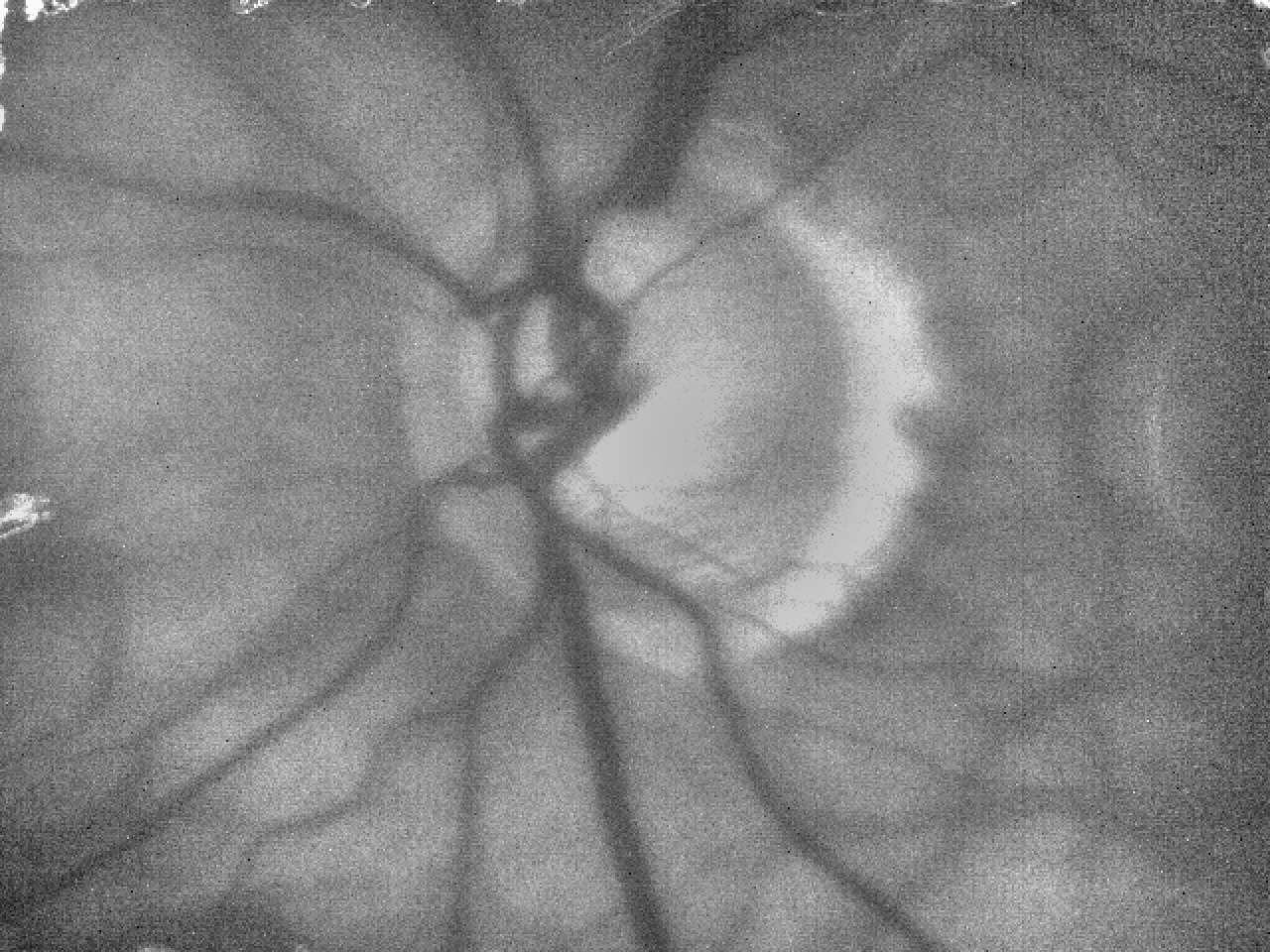}}; 
      \spy on (0.1,-0.3) in node [left] at (3.74, -1.5); 
    \end{tikzpicture} 
	}\\[-0.3em]
	\subfloat[Super-resolution (initial guess)]{\label{fig:07_lowCostFundus_glaucoma:srInit}
		\begin{tikzpicture}[spy using outlines={rectangle,red,magnification=2.0,height=2.8cm, width=2.8cm, connect spies, every spy on node/.append style={thick}}]  
			\node {\pgfimage[height=0.229\linewidth]{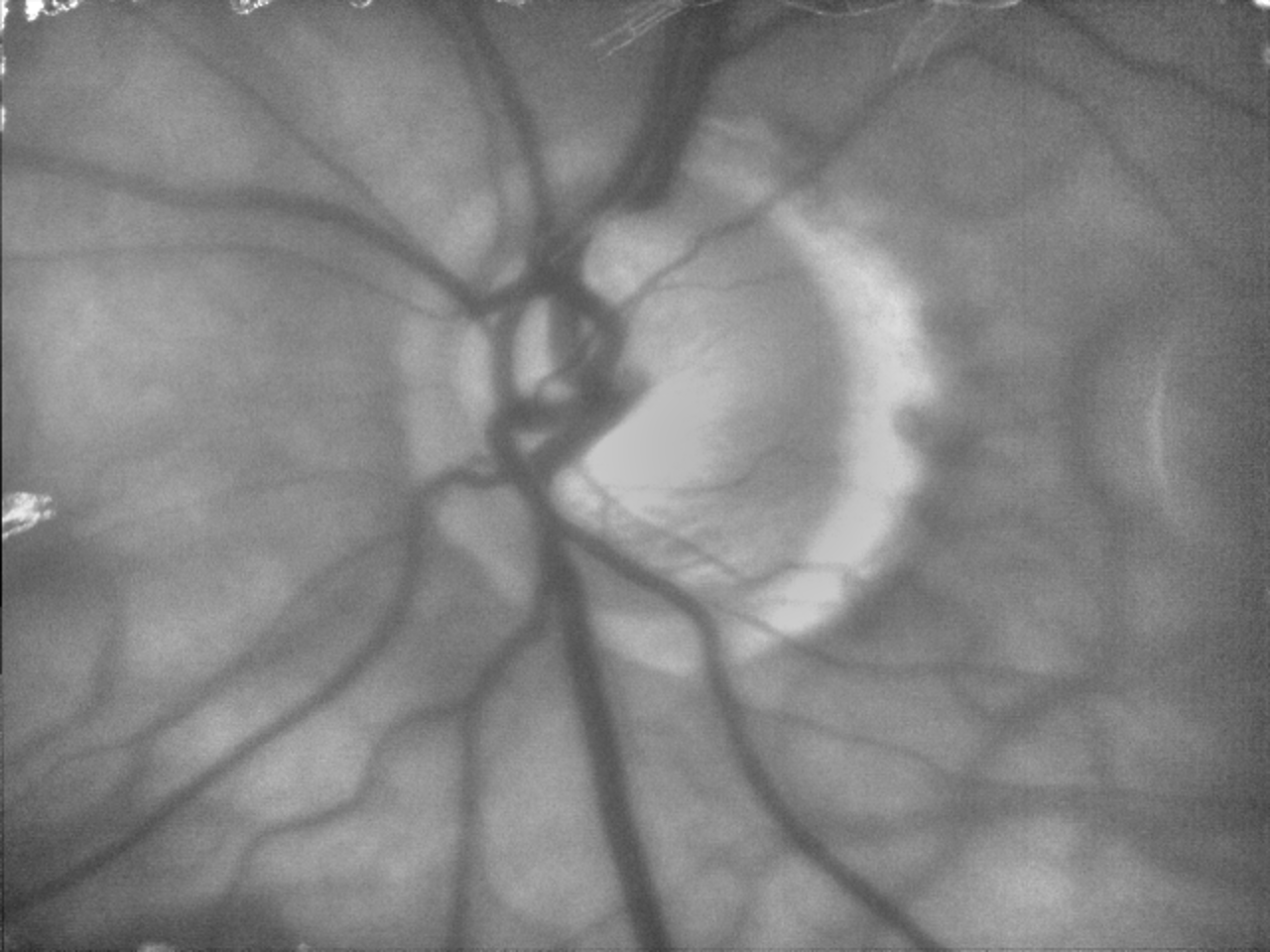}}; 
      \spy on (0.1,-0.3) in node [left] at (3.74, -1.5); 
    \end{tikzpicture} 
	}\hspace{-0.8em}
	\subfloat[Super-resolution (final)]{\label{fig:07_lowCostFundus_glaucoma:sr}
		\begin{tikzpicture}[spy using outlines={rectangle,red,magnification=2.0,height=2.8cm, width=2.8cm, connect spies, every spy on node/.append style={thick}}] 
			\node {\pgfimage[height=0.229\linewidth]{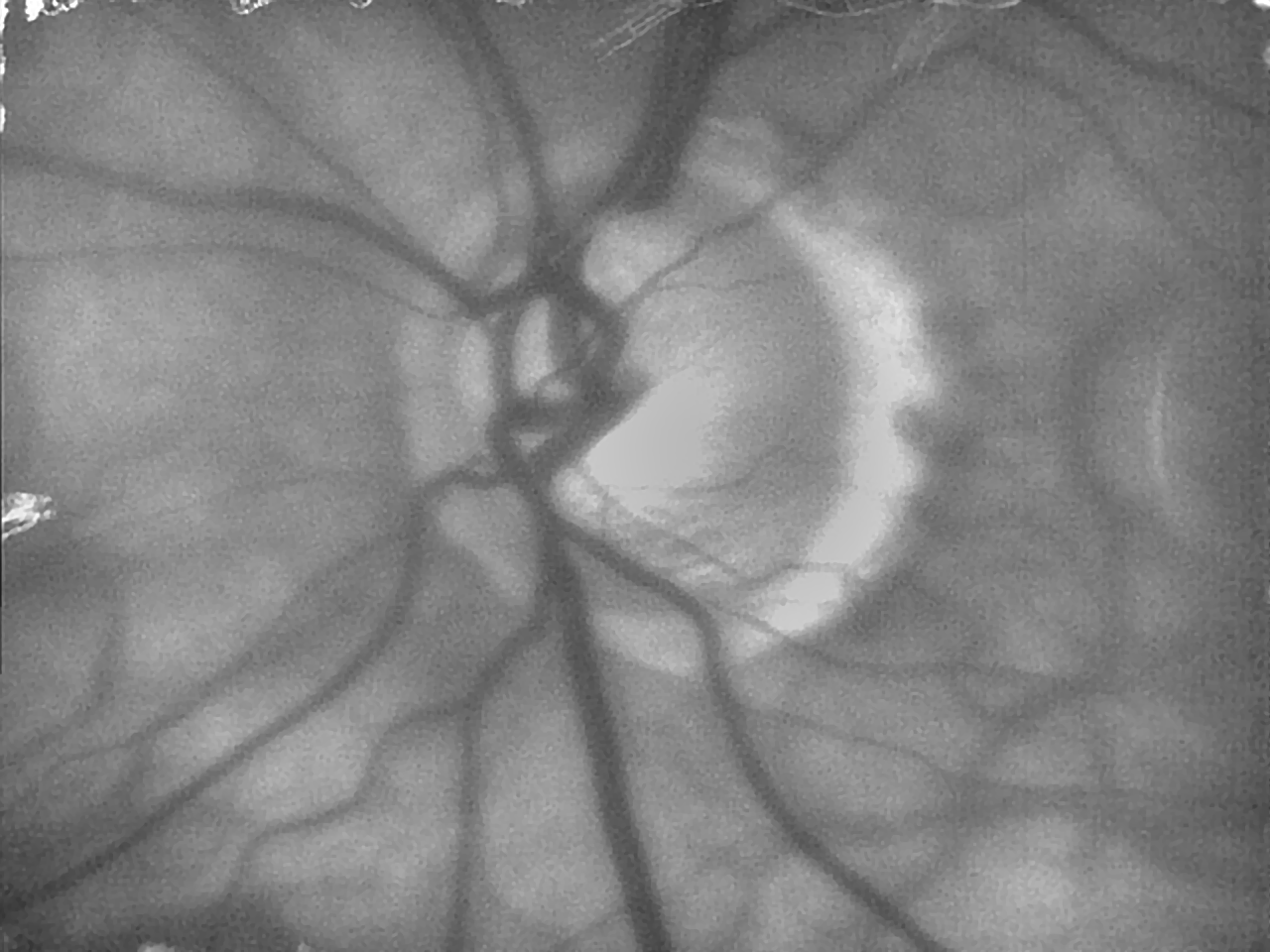}}; 
      \spy on (0.1,-0.3) in node [left] at (3.74, -1.5); 
    \end{tikzpicture} 
	}\hspace{-0.8em}
	\subfloat[Kowa nonmyd reference image]{\label{fig:07_lowCostFundus_glaucoma:kowa}
		\begin{tikzpicture}[spy using outlines={rectangle,red,magnification=3.0,height=2.8cm, width=2.8cm, connect spies, every spy on node/.append style={thick}}] 
			\node {\pgfimage[height=0.229\linewidth]{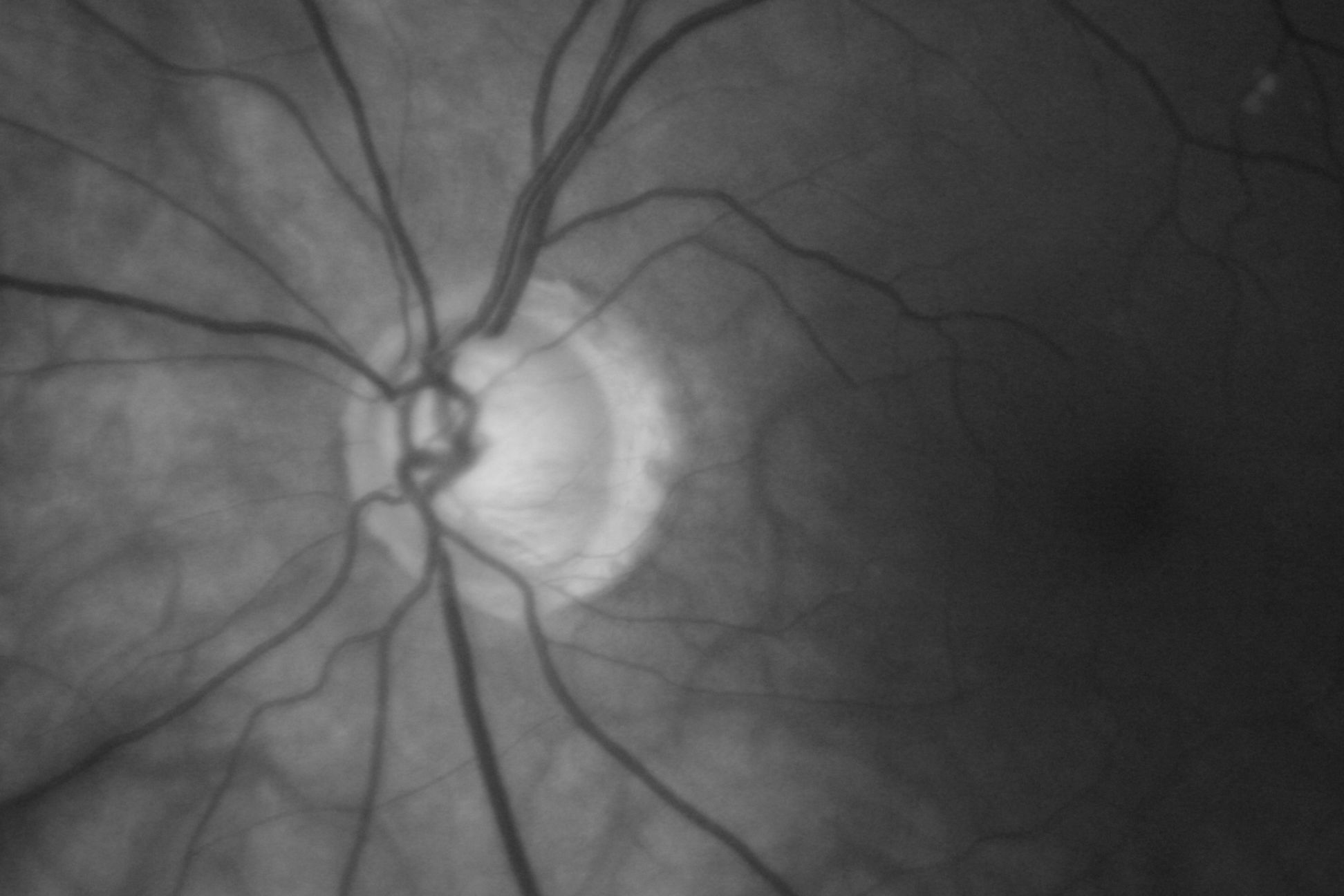}}; 
      \spy on (-1.0,-0.5) in node [left] at (4.2, -1.5); 
    \end{tikzpicture} 
	}
	\caption[Low-cost video data for a glaucoma patient]{Super-resolution on low-cost fundus video frames for a glaucoma patient. \protect\subref{fig:07_lowCostFundus_glaucoma:lr} - \protect\subref{fig:07_lowCostFundus_glaucoma:lrReg} Single low-resolution frames without (w/o) and with (w/) photometric registration used in the proposed framework.  \protect\subref{fig:07_lowCostFundus_glaucoma:srInit} - \protect\subref{fig:07_lowCostFundus_glaucoma:sr} Initial guess determined by the temporal median of the registered low-resolution frames as well as the final super-resolved image.  \protect\subref{fig:07_lowCostFundus_glaucoma:kowa} Reference photograph captured with a Kowa nonymd camera.}
	\label{fig:07_lowCostFundus_glaucoma}
\end{figure}
\end{landscape}
}

\afterpage{
\begin{landscape}
\begin{figure}[!p]
	\centering	
	\subfloat[Single frame (w/o photometric registration)]{\label{fig:07_lowCostFundus_healthy:lr}
		\begin{tikzpicture}[spy using outlines={rectangle,red,magnification=2.0,height=2.8cm, width=2.8cm, connect spies, every spy on node/.append style={thick}}] 
			\node {\pgfimage[height=0.2405\linewidth]{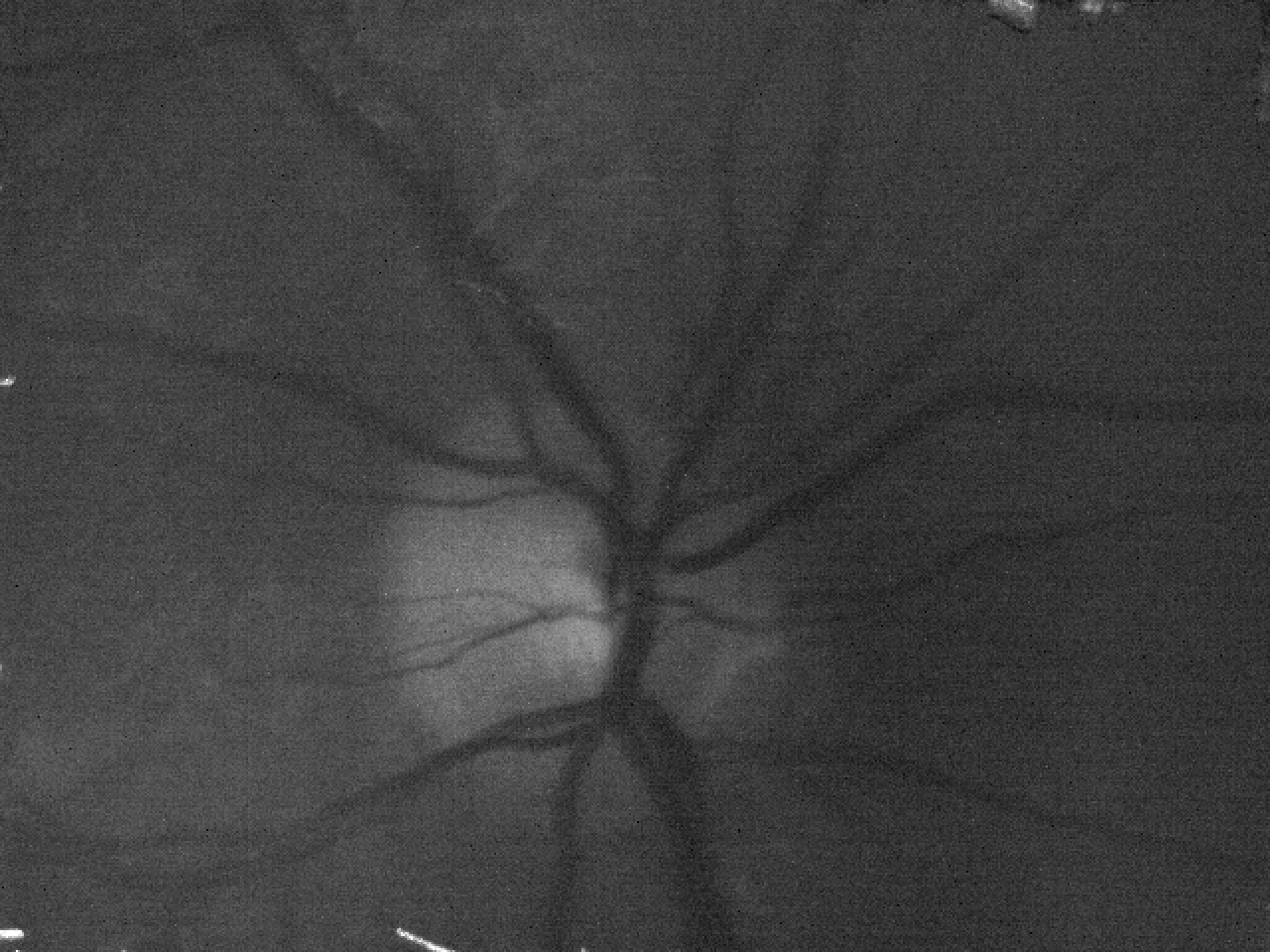}}; 
      \spy on (0.25,-0.4) in node [left] at (-1.1, -1.6); 
    \end{tikzpicture} 
	}\hspace{-0.8em}
	\subfloat[Single frame (w/ photometric registration)]{\label{fig:07_lowCostFundus_healthy:lrReg}
		\begin{tikzpicture}[spy using outlines={rectangle,red,magnification=2.0,height=2.8cm, width=2.8cm, connect spies, every spy on node/.append style={thick}}] 
			\node {\pgfimage[height=0.2405\linewidth]{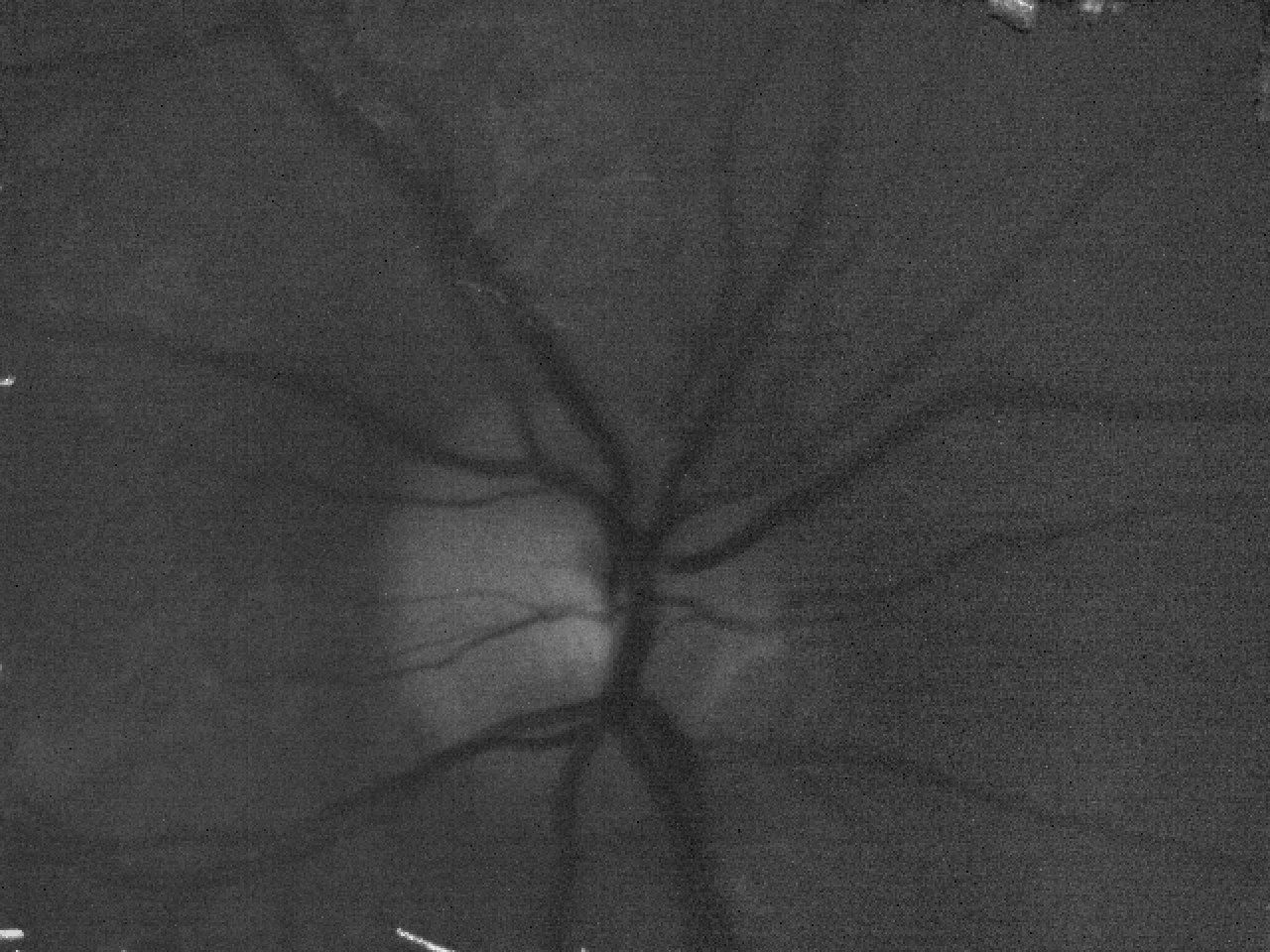}}; 
      \spy on (0.25,-0.4) in node [left] at (-1.1, -1.6); 
    \end{tikzpicture} 
	}\\[-0.3em]
	\subfloat[Super-resolution (initial guess)]{\label{fig:07_lowCostFundus_healthy:srInit}
		\begin{tikzpicture}[spy using outlines={rectangle,red,magnification=2.0,height=2.8cm, width=2.8cm, connect spies, every spy on node/.append style={thick}}]  
			\node {\pgfimage[height=0.2405\linewidth]{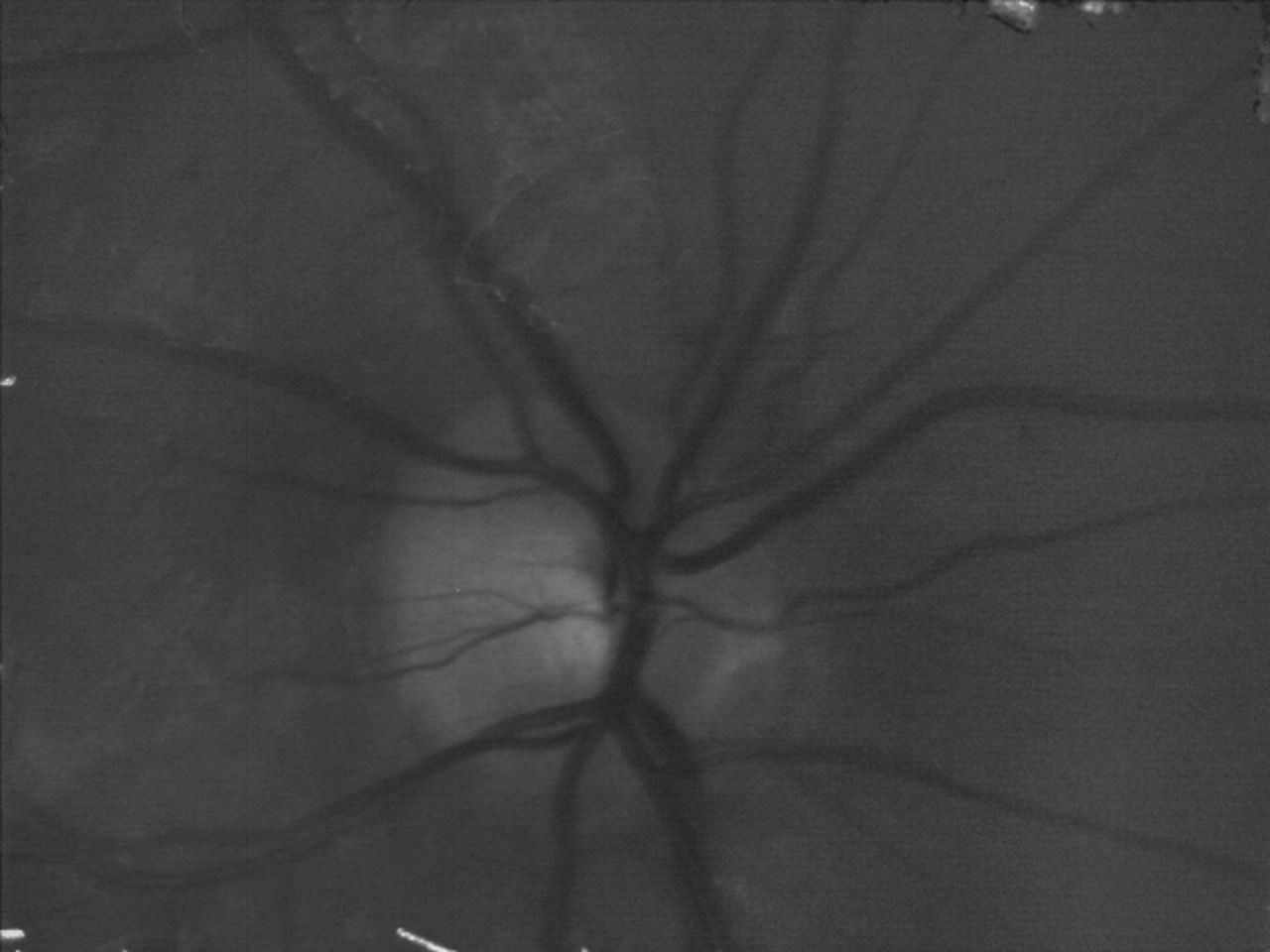}}; 
      \spy on (0.25,-0.4) in node [left] at (-1.1, -1.6); 
    \end{tikzpicture} 
	}\hspace{-0.8em}
	\subfloat[Super-resolution (final)]{\label{fig:07_lowCostFundus_healthy:sr}
		\begin{tikzpicture}[spy using outlines={rectangle,red,magnification=2.0,height=2.8cm, width=2.8cm, connect spies, every spy on node/.append style={thick}}] 
			\node {\pgfimage[height=0.2405\linewidth]{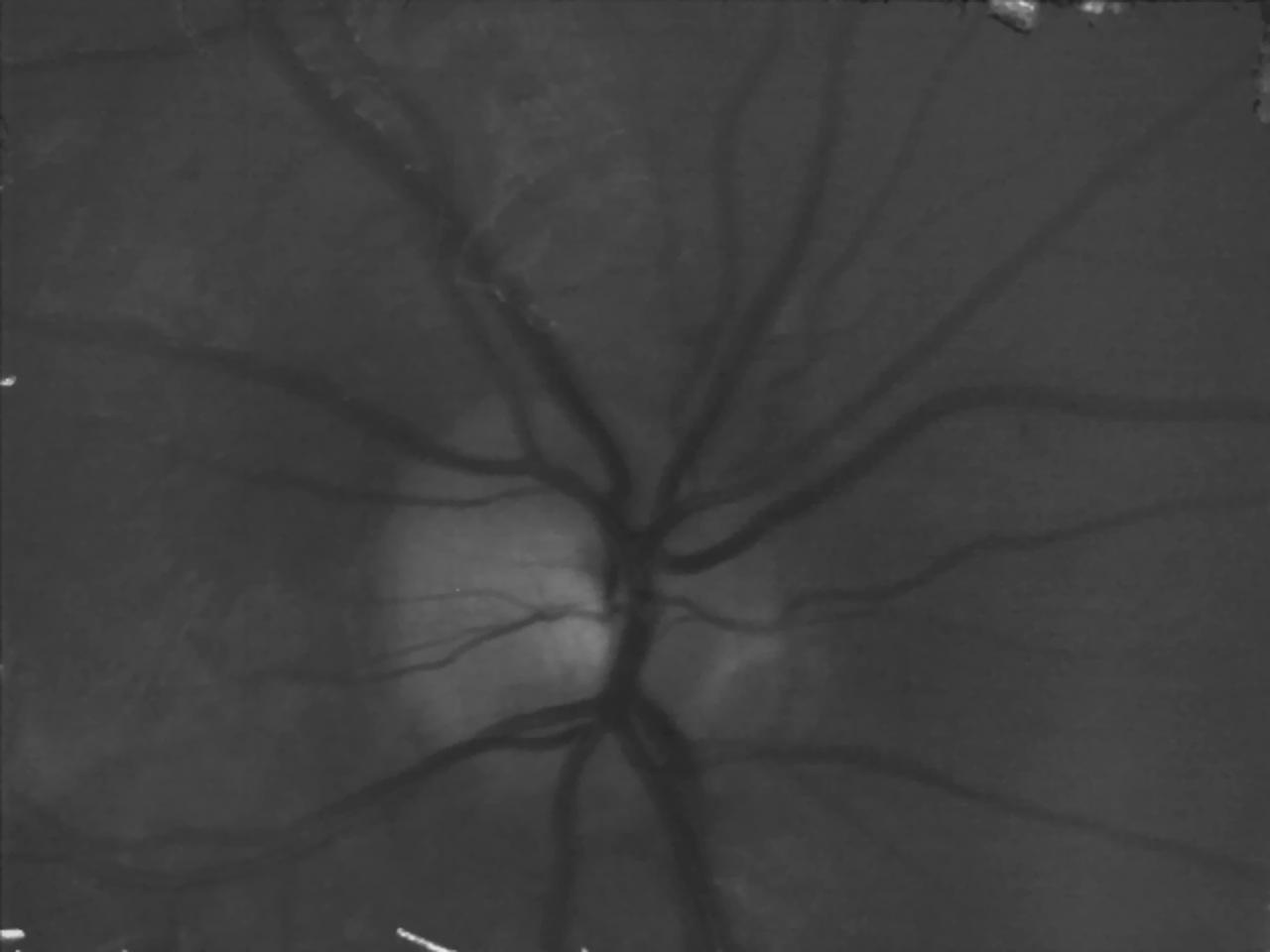}}; 
      \spy on (0.25,-0.4) in node [left] at (-1.1, -1.6); 
    \end{tikzpicture} 
	}\hspace{-0.8em}
	\subfloat[Kowa nonmyd reference image]{\label{fig:07_lowCostFundus_healthy:kowa}
		\begin{tikzpicture}[spy using outlines={rectangle,red,magnification=3.0,height=2.8cm, width=2.8cm, connect spies, every spy on node/.append style={thick}}] 
			\node {\pgfimage[height=0.2405\linewidth]{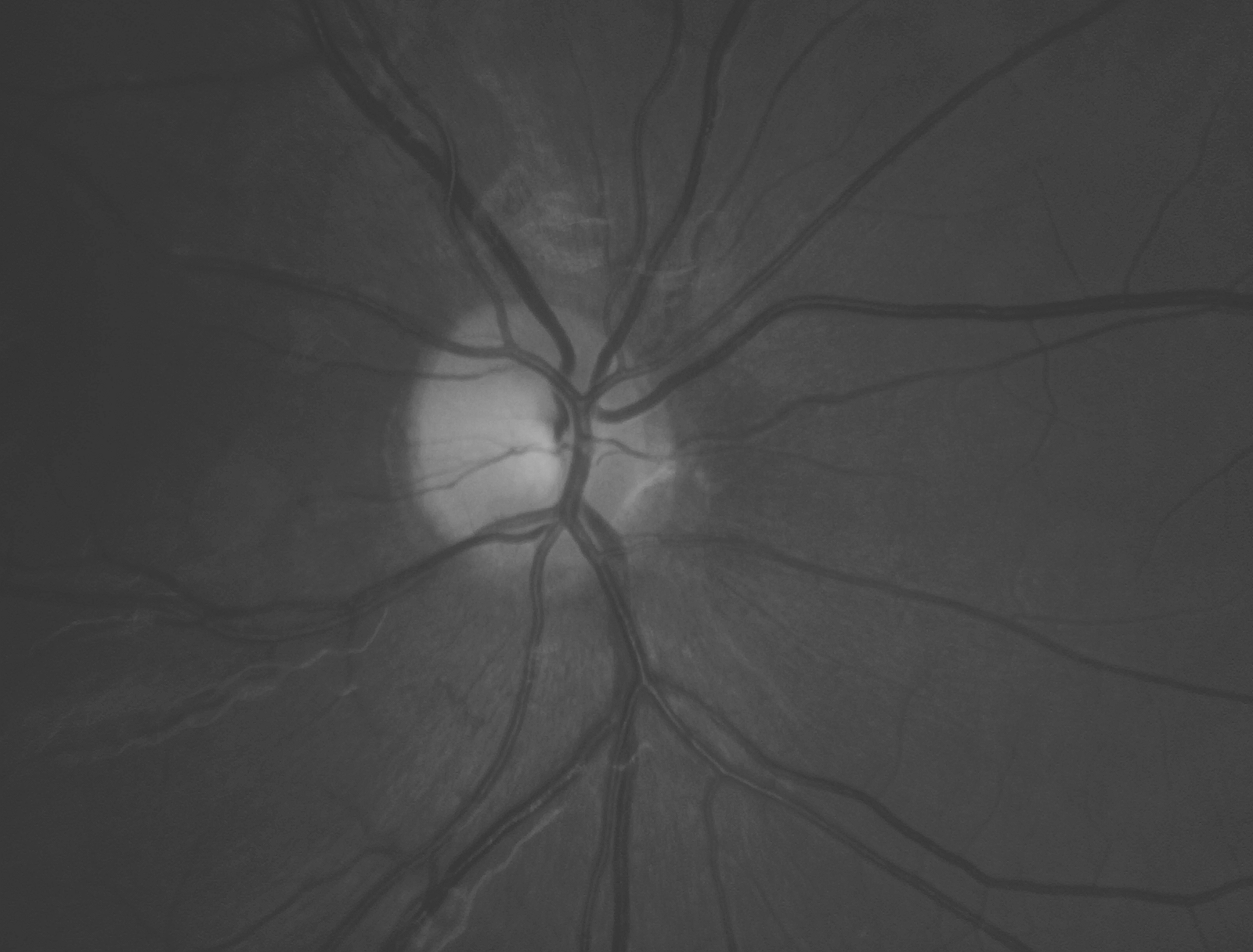}}; 
      \spy on (-0.1,0.4) in node [left] at (-1.06, -1.6);
    \end{tikzpicture} 
	}
	\caption[Low-cost video data for a healthy subject]{Super-resolution on low-cost fundus video frames for a healthy subject. \protect\subref{fig:07_lowCostFundus_healthy:lr} - \protect\subref{fig:07_lowCostFundus_healthy:lrReg} Single low-resolution frames without (w/o) and with (w/) photometric registration used in the proposed framework.  \protect\subref{fig:07_lowCostFundus_healthy:srInit} -  \protect\subref{fig:07_lowCostFundus_healthy:sr} Initial guess determined by the temporal median of the registered low-resolution frames as well as the final super-resolved image. \protect\subref{fig:07_lowCostFundus_healthy:kowa} Reference photograph captured with a Kowa nonymd camera.}
	\label{fig:07_lowCostFundus_healthy}
\end{figure}
\end{landscape}
}

\tikzexternalenable

The reconstruction of high-resolution fundus images from low-resolution video was investigated for anatomical regions that are relevant for diagnostic purposes and contain fine structures to outline the impact of super-resolution. Therefore, the optic nerve head region that captures the optic disk and the cup as two relevant structures for glaucoma detection \cite{Bock2010,Joshi2011} was examined. 

\Fref{fig:07_lowCostFundus_glaucoma} compares original video data and different intermediate results of the proposed framework applied in this region of a glaucoma patient to a reference photograph captured with the Kowa camera. The comparison among a single low-resolution frame in \fref{fig:07_lowCostFundus_glaucoma:lr} and the frame in \fref{fig:07_lowCostFundus_glaucoma:lrReg} depicts the impact of photometric registration as an initial stage of the super-resolution framework. Here, the photometric registration compensated for spatially and temporally varying illumination. \Fref{fig:07_lowCostFundus_glaucoma:srInit} depicts eye movement compensation implemented by the geometric registration and shows the temporal median of $\NumFrames$ registered frames that is used as an initial guess of iterative super-resolution. The final super-resolved image is shown in \fref{fig:07_lowCostFundus_glaucoma:sr}. Note that super-resolution substantially enhanced the appearance of anatomical structures, \eg thin blood vessels, which are barley visible in noisy low-resolution frames. This resulted in a visual appearance that is comparable to the Kowa reference image in \fref{fig:07_lowCostFundus_glaucoma:kowa}. \Fref{fig:07_lowCostFundus_healthy} depicts the same comparison on an example dataset captured from a healthy subject.

\tikzexternaldisable

In order to validate this quality enhancement quantitatively, the gain in terms of the proposed no-reference quality measure was analyzed. The distribution of $Q(\HR)$ normalized by the quality of the reference low-resolution frames is summarized in \fref{fig:07_boxplotsQMeasure} for the optic nerve head regions of six healthy subjects and six glaucoma patients. For each subject, ten consecutive image sequences extracted in a sliding window scheme were analyzed. This comparison among the temporal median and the final super-resolved image confirms that super-resolution improved noise and sharpness characteristics compared to raw video data.
\begin{figure}[!t]
	\centering
	\scriptsize
	\setlength \figurewidth{0.91\textwidth}
	\setlength \figureheight{0.43\figurewidth}
	\subfloat{\input{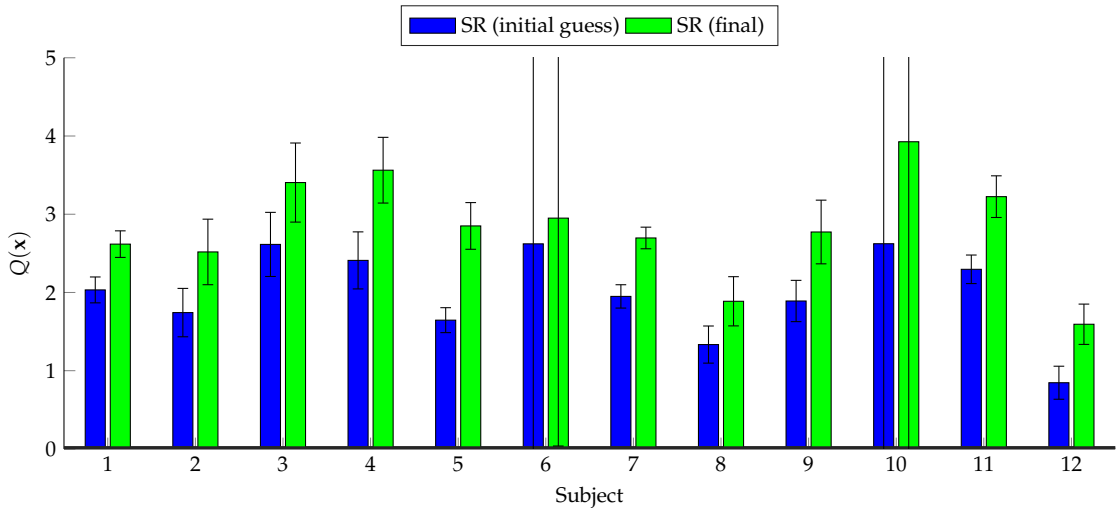}}
	\caption[Quality measure $Q(\HR)$ for healthy subjects and glaucoma patients]{Quality measure $Q(\HR)$ for six healthy subjects (1 - 6) and six glaucoma patients (7 - 12). For each dataset, ten consecutive sequences in a sliding window scheme were processed. The quality assessments were compared for temporal median filtering used as initial guess and the final super-resolved image. Notice that $Q(\HR)$ is normalized by the quality measurement of the corresponding low-resolution reference image.}
	\label{fig:07_boxplotsQMeasure}
\end{figure}
\tikzexternalenable

\paragraph{Super-Resolution Under Photometric Variations.}
Let us now examine the reliability of super-resolution under challenging conditions in retinal imaging. One common issue is a severe photometric variation during an examination that is caused by the light source of the camera and the patient anatomy. Notice that photogeometric registration cannot entirely compensate such variations, \eg in case of oversaturations of the intensities. 

\Fref{fig:07_robusntessIlluminationExample} (top row) shows this issue for a subset of video frames captured from a glaucoma patient with brightness and contrast variations over time. Super-resolution was applied to $\NumFrames = 8$ frames in a sliding window scheme but using the same reference frame for each window. This led to a set of $L$ super-resolved images $\HRFrame{1}, \ldots, \HRFrame{L}$ obtained from $L$ different windows but reconstructed in the same reference coordinate grid. The sensitivity of super-resolution regarding brightness and contrast variations was examined by assessing the reproducibility \cite{Kraus2017} of the super-resolved images $\HRFrame{l}$, $l > 1$ relative to the first image $\HRFrame{1}$. 

In \fref{fig:07_robusntessIlluminationErrorMeasures}, this reproducibility is depicted by the \gls{ssim} and the \gls{nmad} over ten super-resolved images corresponding to ten frame windows. Thus, a \gls{ssim} equal to one and a \gls{nmad} equal to zero indicate a perfect reproducibility on two disjoint input sequences. In terms of both measures, it is noticeable that inconsistencies among super-resolved images increases with shorter temporal overlap and hence a higher variability of the illumination. However, super-resolution achieved a reasonable reproducibility with a \gls{ssim} of above $0.85$ and \gls{nmad} below 0.03. Compared to the initial guess determined by the temporal median, the proposed iterative algorithm resulted a better reproducibility. This behavior is also noticeable by visual comparisons among super-resolved images reconstructed from different frame windows as shown in \fref{fig:07_robusntessIlluminationExample} (bottom row). Notice that severe photometric variations in the input video were successfully compensated in super-resolved data, which confirms the robustness of the proposed framework.  

\begin{figure}[!t]
	\centering
	\small
	\subfloat{\includegraphics[width=0.185\textwidth]{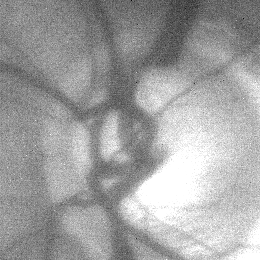}}~
	\subfloat{\includegraphics[width=0.185\textwidth]{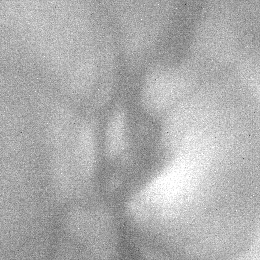}}~
	\subfloat{\includegraphics[width=0.185\textwidth]{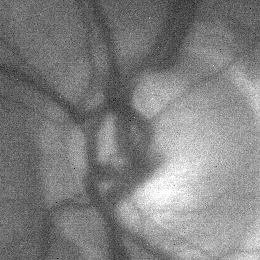}}~
	\subfloat{\includegraphics[width=0.185\textwidth]{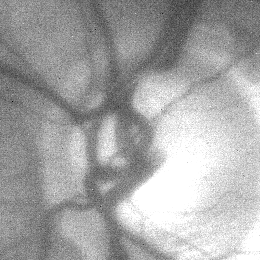}}~
	\subfloat{\includegraphics[width=0.185\textwidth]{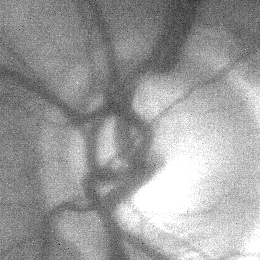}}\\
	\textbf{(a)} Low-resolution video frames with spatial and temporal illumination variations \\
	\subfloat{\includegraphics[width=0.185\textwidth]{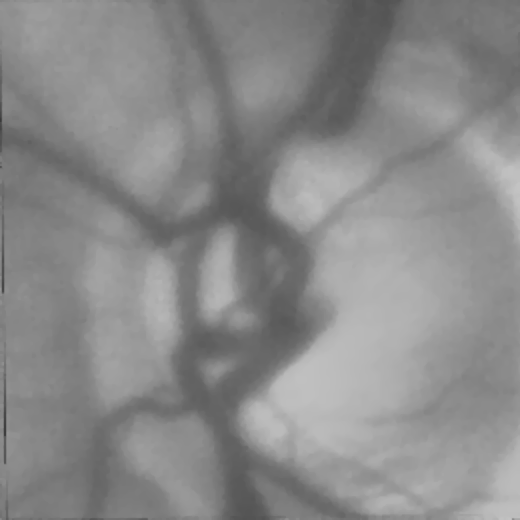}}~
	\subfloat{\includegraphics[width=0.185\textwidth]{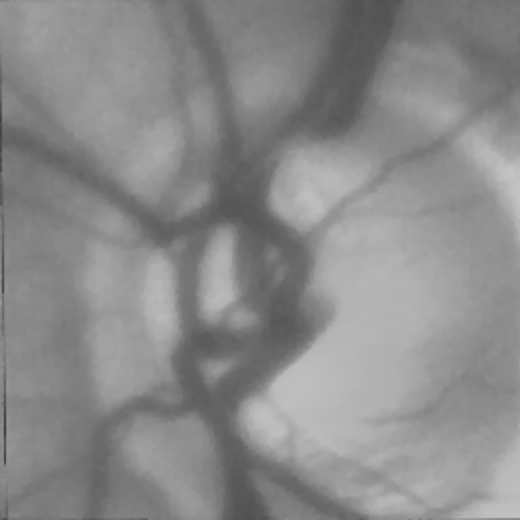}}~
	\subfloat{\includegraphics[width=0.185\textwidth]{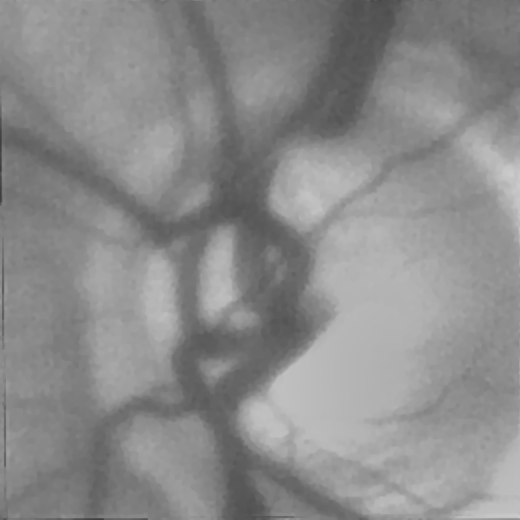}}~
	\subfloat{\includegraphics[width=0.185\textwidth]{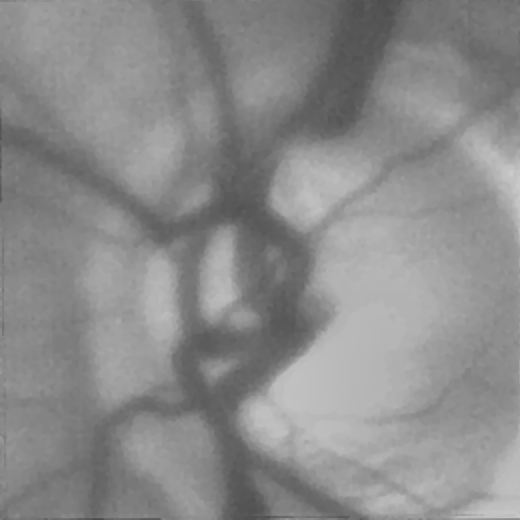}}~
	\subfloat{\includegraphics[width=0.185\textwidth]{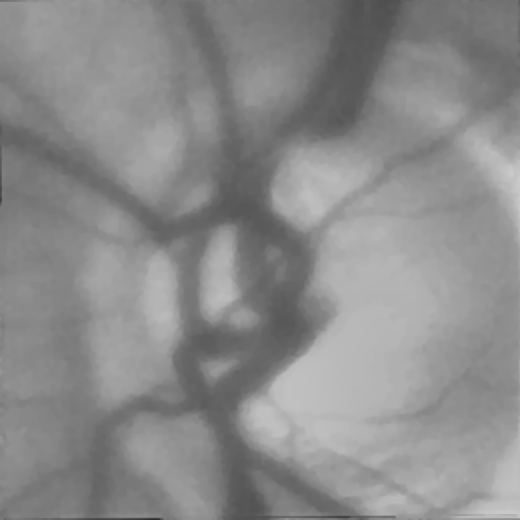}}\\
	\textbf{(b)} Super-resolution under photometric variations \\
	\caption[Super-resolution under photometric variations across video frames]{Super-resolution under spatial and temporal photometric variations across video frames. (a) Different low-resolution frames taken from a sequence with photometric variations. (b) Super-resolution for five different subsequences taken from the input video with $\NumFrames = 8$ frames using the same reference frame.}
	\label{fig:07_robusntessIlluminationExample}
\end{figure}
\begin{figure}[!t]
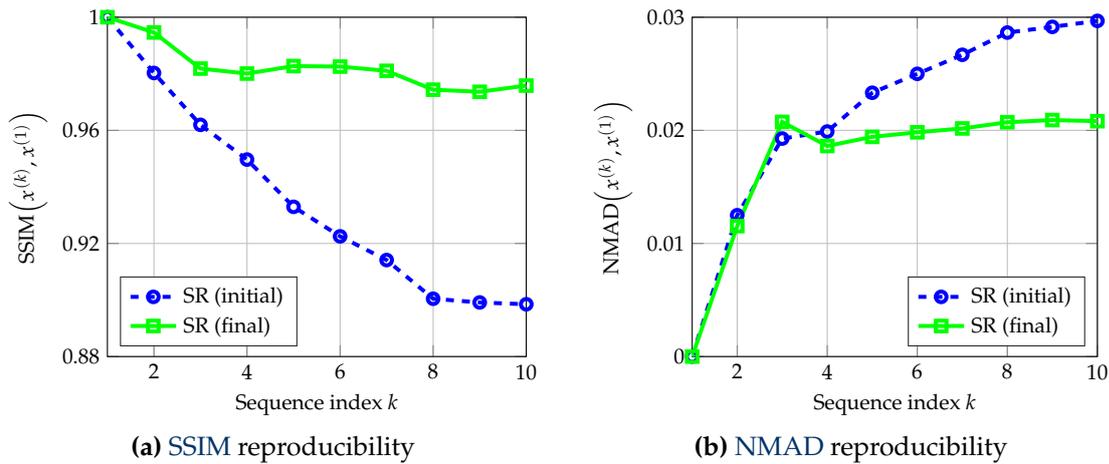

	\centering
	\scriptsize 
	\setlength \figurewidth{0.37\textwidth}
	\setlength \figureheight{0.80\figurewidth} 
	\subfloat[\gls{ssim} reproducibility]{\input{images/chapter7/illumination_ssim.tikz}}\qquad
	\subfloat[\gls{nmad} reproducibility]{\input{images/chapter7/illumination_nmad.tikz}}
	\caption[Sensitivity of super-resolution against photometric variations]{Sensitivity of super-resolution against photometric variations. The sensitivity was assessed by the reproducibility of super-resolved images for ten subsequences with photometric variations relative to the super-resolved image reconstructed from the first (variation-free) sequence. Reproducibility was measured by the \gls{ssim} and the \gls{nmad}.}
	\label{fig:07_robusntessIlluminationErrorMeasures}
\end{figure}

\paragraph{Super-Resolution Under Eye Accommodation.}
Another condition of practical relevance is eye accommodation, which impairs the reliability of super-resolution. 

In \fref{fig:07_robusntessAccommodationExample} (top row), eye accommodation is shown for five video frames captured from a healthy subject. This resulted in out-of-focus blur that is increasing over time. Similar to the previous experiment, super-resolution was performed in a sliding window scheme based on $\NumFrames = 8$ frames but with a fixed reference for each window to study its sensitivity regarding this effect. 

\Fref{fig:07_robusntessAccommodationErrorMeasures} depicts the reproducibility measures for $L$ super-resolved images obtained in this experiment. The reproducibility characterized by these measures was dropped for larger amounts of out-of-focus blur related to accommodation. However, super-resolution provided a better reproducibility compared to its initial guess, which indicates a lower sensitivity regarding accommodation. More specifically, moderate levels of accommodation were successfully compensated as depicted for the first three cases in \fref{fig:07_robusntessAccommodationExample} (bottom row). In this experiment, severe levels of eye accommodation that are related to a substantial amount of time variant blur as shown for the last two cases could not be compensated.

\begin{figure}[!t]
	\centering
	\small
	\subfloat{\includegraphics[width=0.185\textwidth]{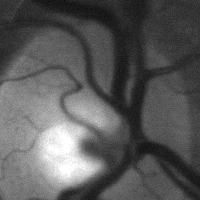}}~
	\subfloat{\includegraphics[width=0.185\textwidth]{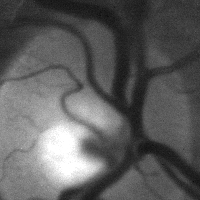}}~
	\subfloat{\includegraphics[width=0.185\textwidth]{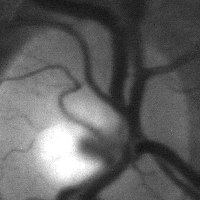}}~
	\subfloat{\includegraphics[width=0.185\textwidth]{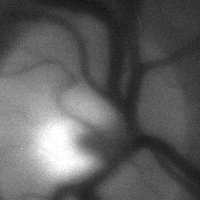}}~
	\subfloat{\includegraphics[width=0.185\textwidth]{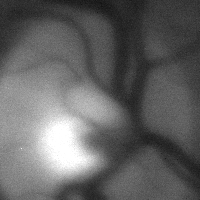}}\\
	\textbf{(a)} Low-resolution video frames with eye accommodation over time\\
	\subfloat{\includegraphics[width=0.185\textwidth]{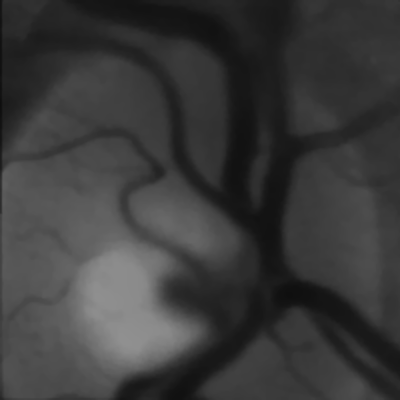}}~
	\subfloat{\includegraphics[width=0.185\textwidth]{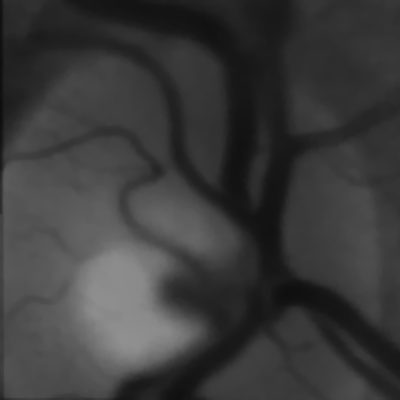}}~
	\subfloat{\includegraphics[width=0.185\textwidth]{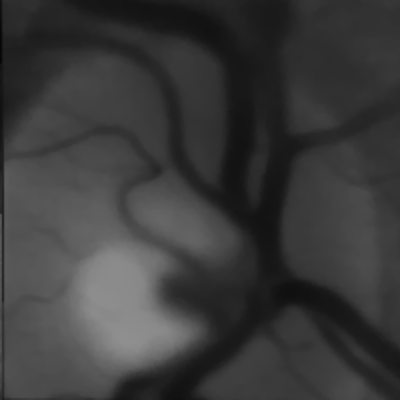}}~
	\subfloat{\includegraphics[width=0.185\textwidth]{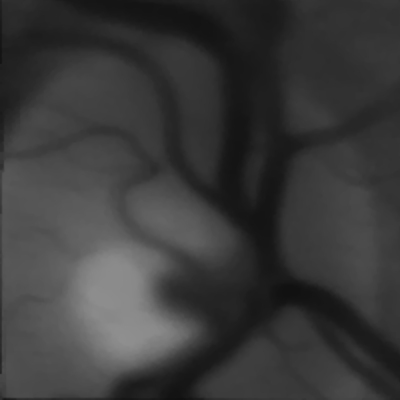}}~
	\subfloat{\includegraphics[width=0.185\textwidth]{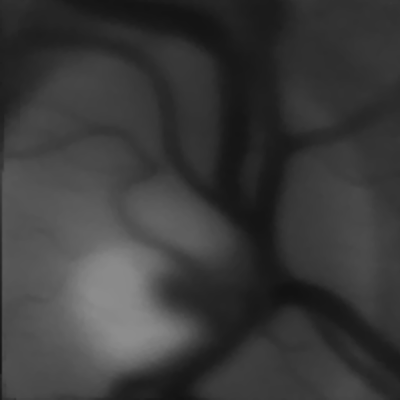}}\\
	\textbf{(b)} Super-resolution under increasing eye accommodation \\
	\caption[Super-resolution under eye accommodation over time]{Super-resolution under out-of-focus blur due to eye accommodation over time. (a) Low-resolution frames taken from a sequence with accommodation and increasing out-of-focus blur from the first to the last frame. (b) Super-resolution for five different subsequences taken from the input video with $\NumFrames = 8$ frames using the same reference frame and increasing out-of-focus blur from the first to the last frame.}
	\label{fig:07_robusntessAccommodationExample}
\end{figure}
\begin{figure}[!t]
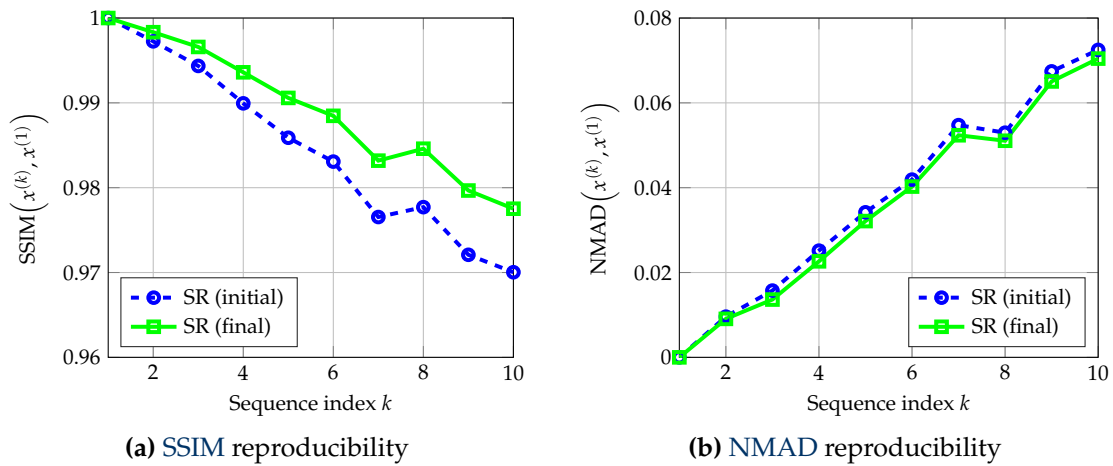

	\centering
	\scriptsize 
	\setlength \figurewidth{0.37\textwidth}
	\setlength \figureheight{0.80\figurewidth} 
	\subfloat[\gls{ssim} reproducibility]{\input{images/chapter7/accommodation_ssim.tikz}}\qquad
	\subfloat[\gls{nmad} reproducibility]{\input{images/chapter7/accommodation_nmad.tikz}}
	\caption[Sensitivity of super-resolution against eye accommodation]{Sensitivity of super-resolution against out-of-focus blur due to eye accommodation. The sensitivity was assessed by measuring reproducibility of super-resolved images associated with ten subsequences that show increasing out-of-focus blur relative to the first subsequence. Reproducibility was measured by the \gls{ssim} and the \gls{nmad}.}
	\label{fig:07_robusntessAccommodationErrorMeasures}
\end{figure}

\subsection{Application to Super-Resolved Mosaicing}

Besides the spatial resolution, another quality criterion of ophthalmic imaging systems is their \gls{fov}. In order to get a comprehensive view of the human retina for diagnostic or interventional purposes, there is a strong need to capture retinal images with a wide \gls{fov}. This applies to technologies like the slit lamp, scanning laser ophthalmoscopy or digital fundus cameras. In practice, however, this is challenging due the finite size of the human pupil and the fact that the pupil needs to be dilated to increase the \gls{fov}. For this reason, software-based image registration and mosaicing \cite{Can2002,Cattin2006,Adal2014,Zheng2014a} have been proposed, which aims at stitching multiple views of the retina. 

This section demonstrates a novel combination of multi-frame super-resolution with mosaicing techniques to enable super-resolved mosaicing. Unlike related methods, this joint approach recovers a single mosaic view from low-resolution video while simultaneously enhancing the spatial resolution. For the study of super-resolved mosaicing, we employ the method proposed in \cite{Kohler2016} that reconstructs a single retinal mosaic from multiple super-resolved images. This approach exploits a set of low-resolution frames $\mathcal{Y}$ that consists of $n$ disjoint subsets $\mathcal{Y}_i$, $i = 1, \ldots, n$. Each subset $\mathcal{Y}_i$ is represented by $K_i$ consecutive frames taken from $\mathcal{Y}$ and is referred to as a \textit{view}. These views capture complementary regions on the human retina due to eye motion during the examination. 

In summary, super-resolved mosaicing is described by the following three-stage procedure:
\begin{enumerate}
	\item Eye tracking is used for a fully automatic selection of $n$ views. This is done in real-time using the optic disk as a robust feature for tracking \cite{Kurten2014}.
	\item For each view $\mathcal{Y}_i$ that is selected according to the tracking procedure, $K_i$ frames are utilized to reconstruct the corresponding super-resolved view $\HR_i$.
	\item The super-resolved views $\HR_1, \ldots, \HR_n$ are first geometrically and photometrically registered and then stitched to a mosaic by adaptive averaging.
\end{enumerate}

\Fref{fig:07_superresolvedMosaicingResult} demonstrates super-resolved mosaicing on video data acquired from one healthy subject. In this experiment, we examined the left eye without dilating the pupil and asked the subject to fixate nine different positions on a fixation target. This resulted in eye movements across the frames in the acquired video sequence, and hence a scan of different regions of the retina as depicted in \fref{fig:07_superresolvedMosaicingResult}. We employed super-resolution with magnification $\MagFac = 2$ for these views with $K_i = 8$ frames as embedded in the proposed mosaicing framework. The final mosaic was assembled from nine different views. On the one hand, the super-resolution stage enhanced the spatial resolution of the original video data. On the other hand, stitching of super-resolved views enlarged the \gls{fov} from $\approx 15^\circ$ in the video data to $\approx 30^\circ$ in the mosaic image.
\begin{figure}[!t]
	\centering
	~~\begin{minipage}[b]{0.502\textwidth}
		\subfloat{\includegraphics[width=0.32\textwidth]{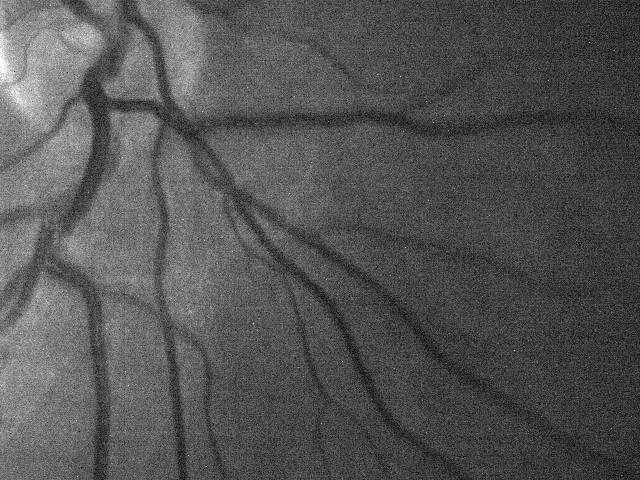}}~
		\subfloat{\includegraphics[width=0.32\textwidth]{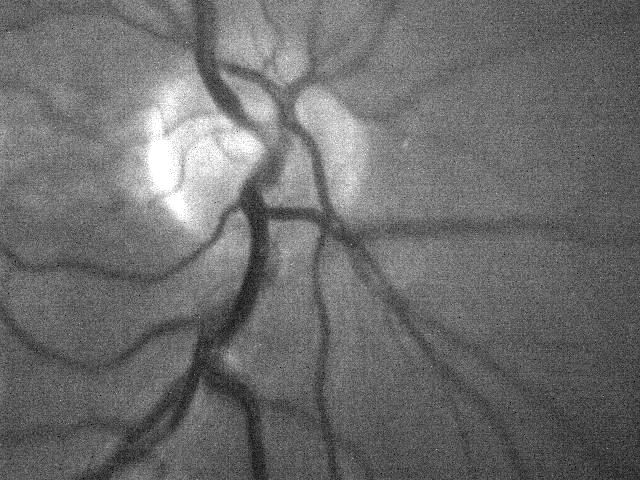}}~
		\subfloat{\includegraphics[width=0.32\textwidth]{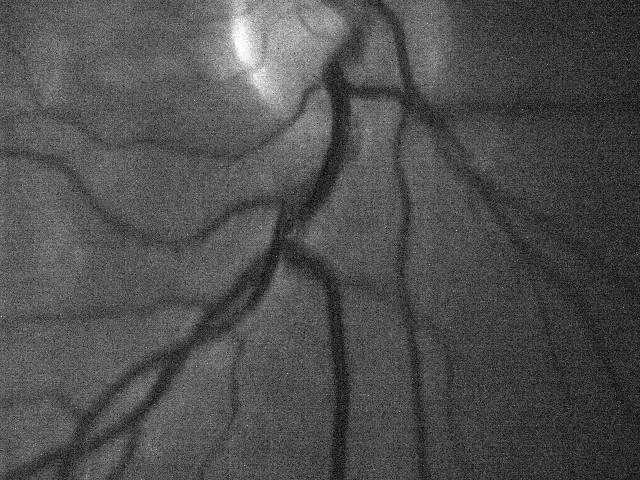}}\\[-0.5em]
		\subfloat{\includegraphics[width=0.32\textwidth]{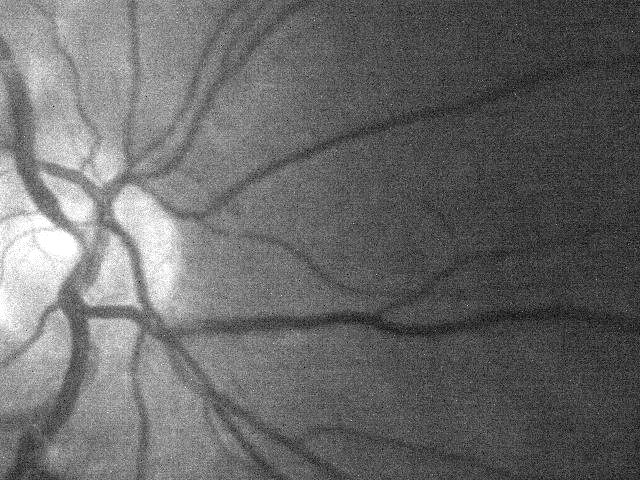}}~
		\subfloat{\includegraphics[width=0.32\textwidth]{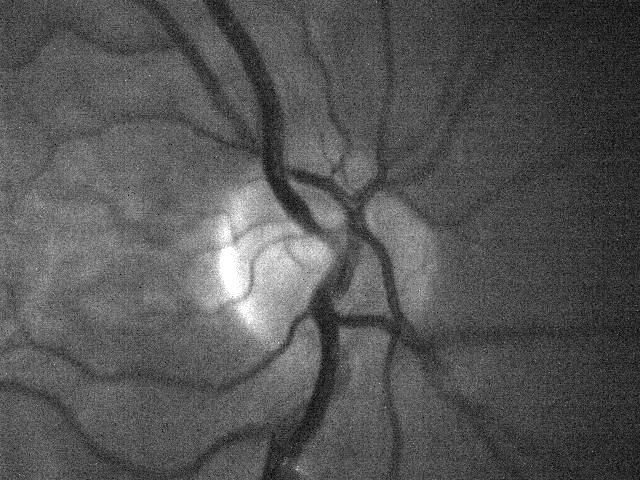}}~
		\subfloat{\includegraphics[width=0.32\textwidth]{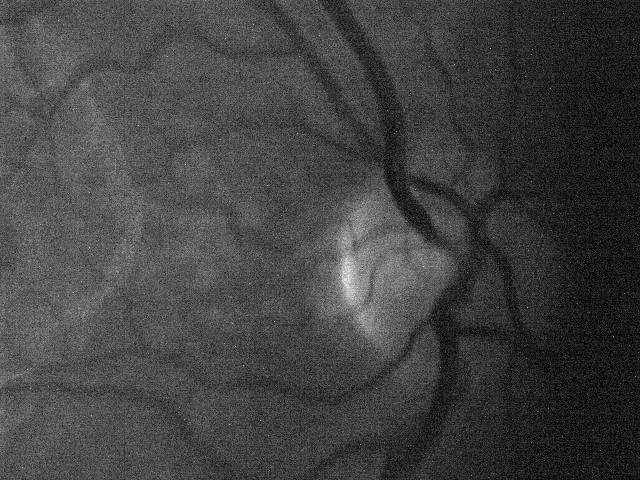}}\\[-0.5em]
		\subfloat{\includegraphics[width=0.32\textwidth]{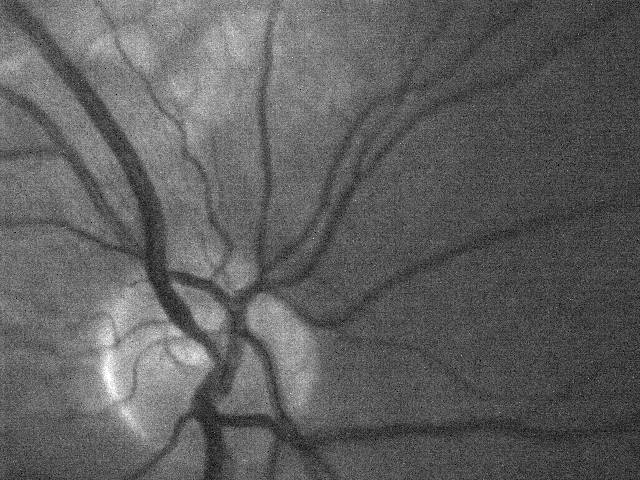}}~
		\subfloat{\includegraphics[width=0.32\textwidth]{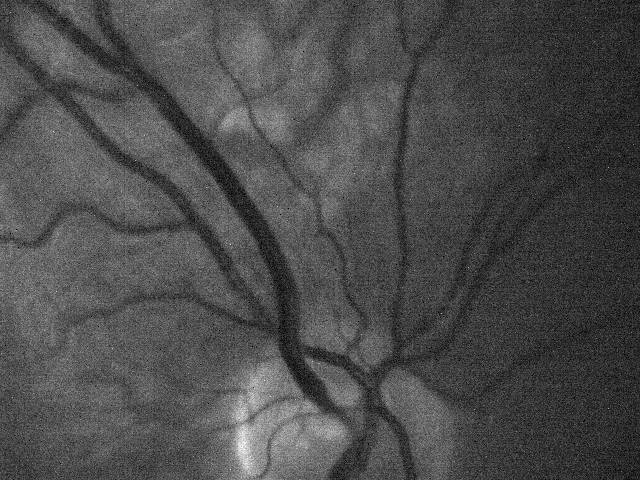}}~
		\subfloat{\includegraphics[width=0.32\textwidth]{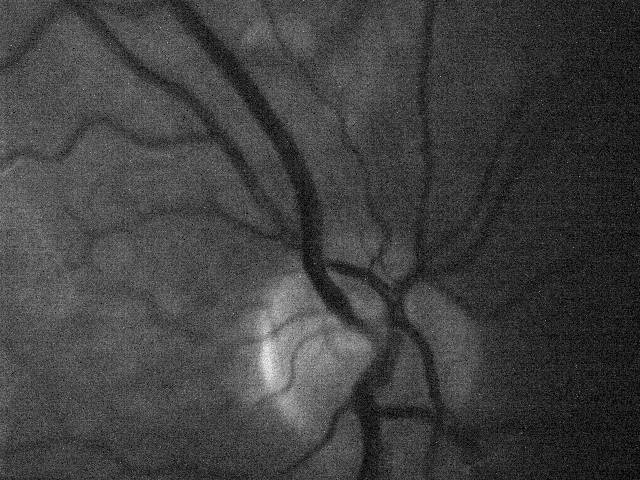}}\\
		\small
		\centering
		\textbf{(a)} Low-resolution views
	\end{minipage}~
	\begin{minipage}[b]{0.472\textwidth}
		\includegraphics[width=0.96\textwidth]{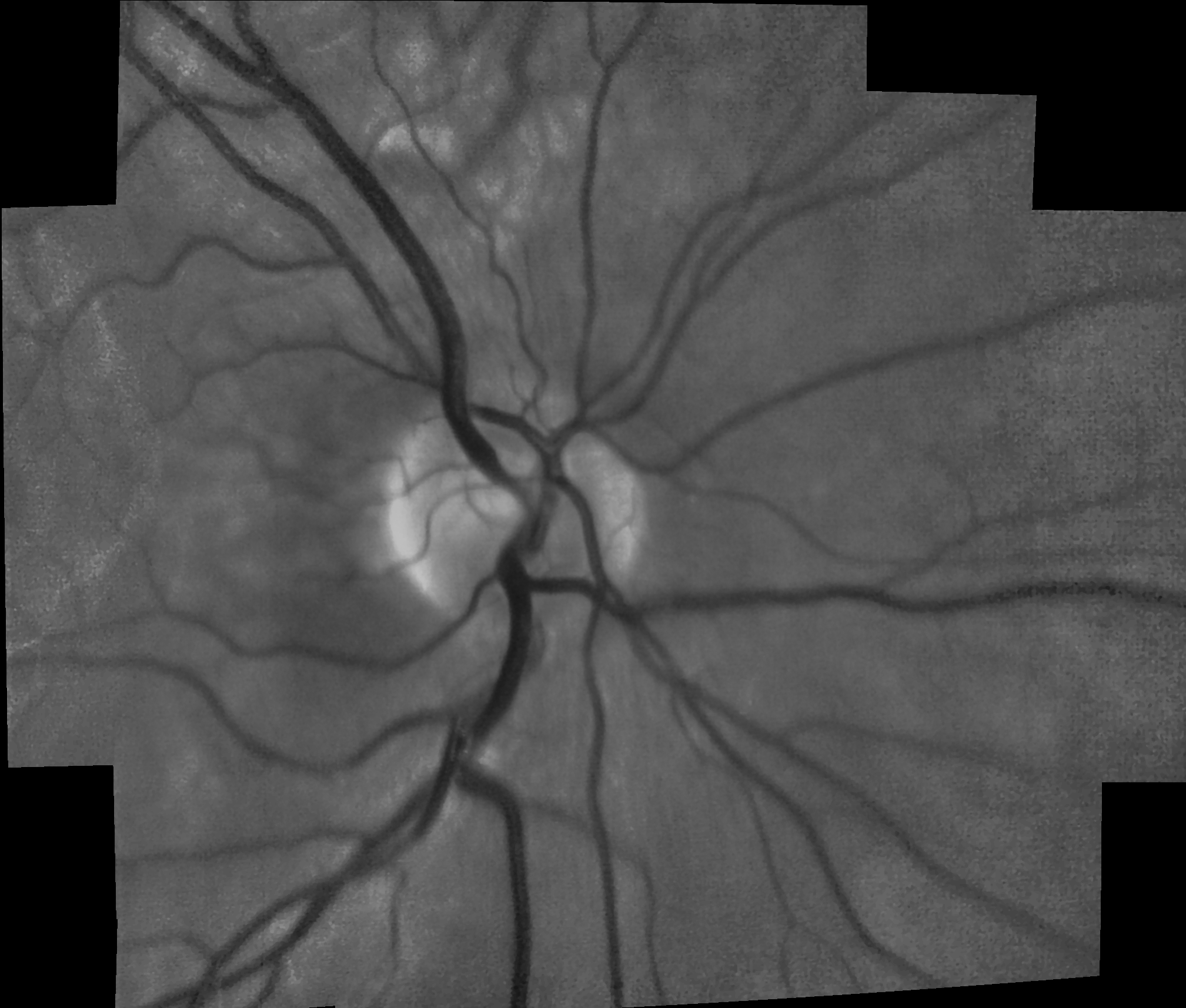}
		\small
		\centering
		\textbf{(b)} Super-resolved mosaic
	\end{minipage}
	\caption[Image mosaicing by scanning different regions on the human retina]{Application of multi-frame super-resolution for image mosaicing. (a) Single low-resolution images for nine different regions of the human retina extracted from video data of a healthy subject. The different regions were scanned by asking the subject to fixate nine different positions on a fixation target. (b) Super-resolved mosaic reconstructed from the original video data of the nine regions. Figure reused from \cite{Kohler2016} with the publisher's permission \copyright2016 IEEE.}
	\label{fig:07_superresolvedMosaicingResult}
\end{figure}

\section{Conclusion}
\label{sec:07_Conclusion}

This chapter introduced a new super-resolution framework for retinal fundus video imaging as a novel diagnostic technique in ophthalmology. In order to reconstruct high-resolution fundus images from a low-resolution video, natural human eye movements during an examination are exploited. An image formation model was introduced that models eye movements by affine image-to-image transformations and considers spatial and temporal photometric variations across multiple frames. Based on this model, an iterative super-resolution algorithm was introduced that is steered by image quality self-assessment for the automatic selection of regularization hyperparameters. Quality self-assessment is based on a continuous quality measure that characterizes the level of sharpness and noise. The proposed measure has a mean Spearman rank correlation of 0.70 \wrt the \gls{psnr}, which shows that it can act as surrogate for full-reference quality assessment in the absence of ground truth data.

In a quantitative study, super-resolution enhanced the \gls{psnr} by $3.5$\,\gls{db} and the \gls{ssim} by $0.07$ compared to low-resolution data. Moreover, the sensitivity of automatic blood vessel segmentation was improved by $10$\,\%. Super-resolution on video data acquired with a mobile low-cost fundus camera provided images of comparable quality to those of commercially available, but expensive and stationary cameras. This encourages the use of the proposed method within clinical workflows, where cost-efficiency and mobility are essential, \eg computer-aided screening. In addition, super-resolved mosaicing was presented to reconstruct high-resolution fundus images with enlarged \gls{fov}.

Future work needs to study the impact of the proposed framework to the diagnostic usability of fundus video imaging. One promising direction for future research is the adoption of super-resolution within machine learning techniques for computer-aided diagnosis regarding prevalent eye diseases \cite{Bock2010,Abramoff2015}.


\chapter{Applications in Image-Guided Surgery}
\label{sec:HybridRangeImagingForImageGuidedSurgery}

\myminitoc

\noindent
This chapter investigates new applications of super-resolution to facilitate interventional medical imaging. In this context, one emerging field of research is the development of image guidance systems by means of hybrid range imaging to assist surgeons during minimally invasive or open surgical procedures. These systems can be implemented based on active range sensor technologies that enable the joint acquisition of surface data besides photometric information to provide a comprehensive view of the underlying scene. However, one common issue of these technologies is the low spatial resolution of today’s range sensors, which limits their applicability in medical workflows. In order to enhance the reliability of image guidance, this chapter adopts the multi-sensor super-resolution framework presented in \cref{sec:MultiSensorSuperResolutionUsingGuidanceImages}. The following complementary imaging setups are examined: 1) 3-D endoscopy to enhance minimally invasive surgical procedures as well as 2) 3-D image guidance for open surgery. This chapter presents system calibration approaches for both setups as a prerequisite for multi-sensor super-resolution. In addition, a comprehensive evaluation for super-resolution based on synthetic and ex-vivo datasets in both applications is reported.

An early study of super-resolution in 3-D endoscopy has been published by K\"ohler \etal \cite{Kohler2013a} and Haase \cite{Haase2016}. These concepts have been later extended in \cite{Kohler2014a} and \cite{Kohler2015a} including their application in open surgery.

\section{Introduction and Medical Background}
\label{sec:08_IntroductionAndMedicalBackground}

In the area of interventional medical imaging, one recent trend is the usage of range imaging technologies to gain 3-D surface information of patient anatomy in addition to 2-D photometric data \cite{Bauer2013}. If both approaches are aggregated, the combined setup enables intra-operative hybrid imaging of the anatomy. Compared to pure 2-D imaging, the existence of additional range information features various advantages for medical interventions. One of the most obvious benefits is the expectation that range data holds the potential to offer the surgeon a more comprehensive view of patient anatomy in order to enhance the safety and efficiency of surgical procedures. In addition, it initiated the development of novel techniques for computer-assisted interventions to aid the surgeon. Some prominent examples for such applications in the area of minimally invasive surgery include automatic localization and collision avoidance for surgical instruments \cite{Haase2013b,CongcongWang2014a}. More recently, 3-D abdomen reconstruction using range satellite cameras has been proposed to improve orientation and navigation during minimally invasive procedures \cite{Haase2013a}. Another use case widely investigated for open surgery is the multi-modal registration of pre-operative, tomographic planning data with intra-operative range information \cite{Mersmann2011a}. This has widespread applications for augmented reality to aid surgeries or forensic medicine \cite{Kilgus2015}.
\begin{figure}[!t]
	\centering
		\raisebox{0.90cm}{\rotatebox{90}{\footnotesize Endoscopy}}\quad
		\subfloat{\includegraphics[width=0.305\textwidth, height=0.2425\linewidth]{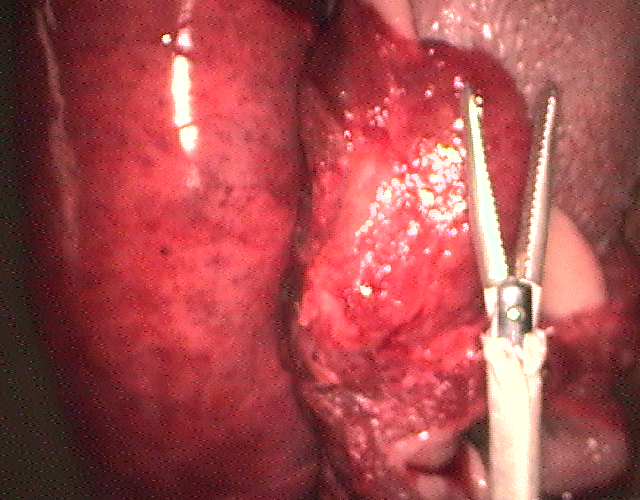}}~
		\subfloat{\includegraphics[width=0.305\textwidth, height=0.2425\linewidth]{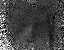}}~
		\subfloat{\includegraphics[width=0.305\textwidth, height=0.2425\linewidth]{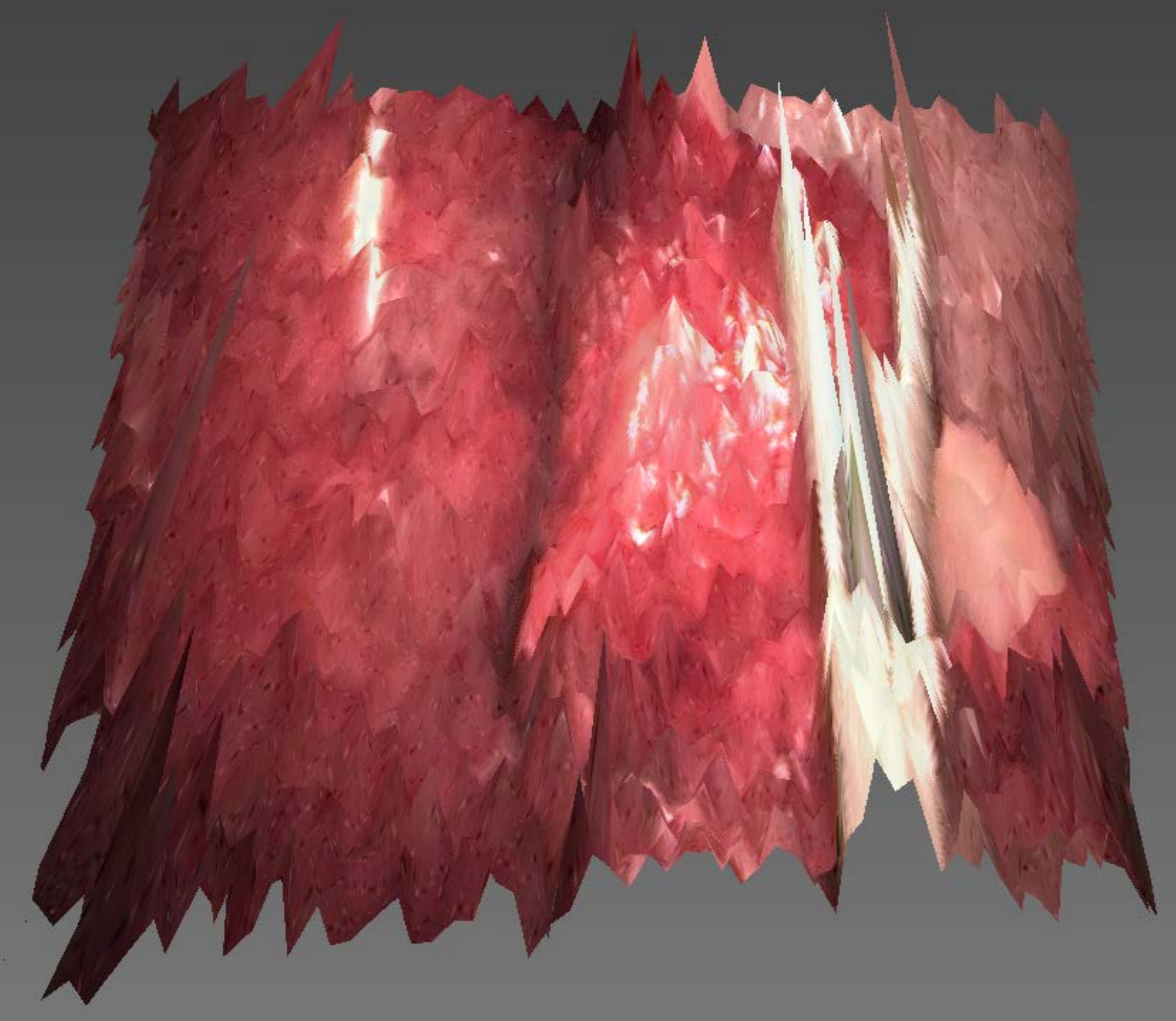}}\\
		\setcounter{subfigure}{0}
		\raisebox{1.0cm}{\rotatebox{90}{\footnotesize Open surgery}}\quad
		\subfloat[Color image]{\includegraphics[width=0.305\textwidth, height=0.285\linewidth]{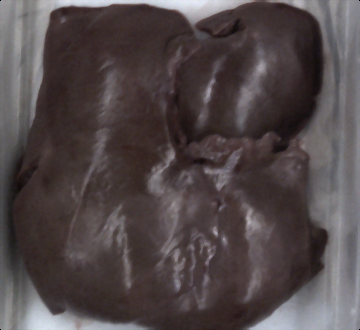}}~
		\subfloat[Range image]{\includegraphics[width=0.305\textwidth, height=0.285\linewidth]{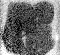}}~
		\subfloat[Textured 3-D mesh]{\includegraphics[width=0.305\linewidth, height=0.285\linewidth]{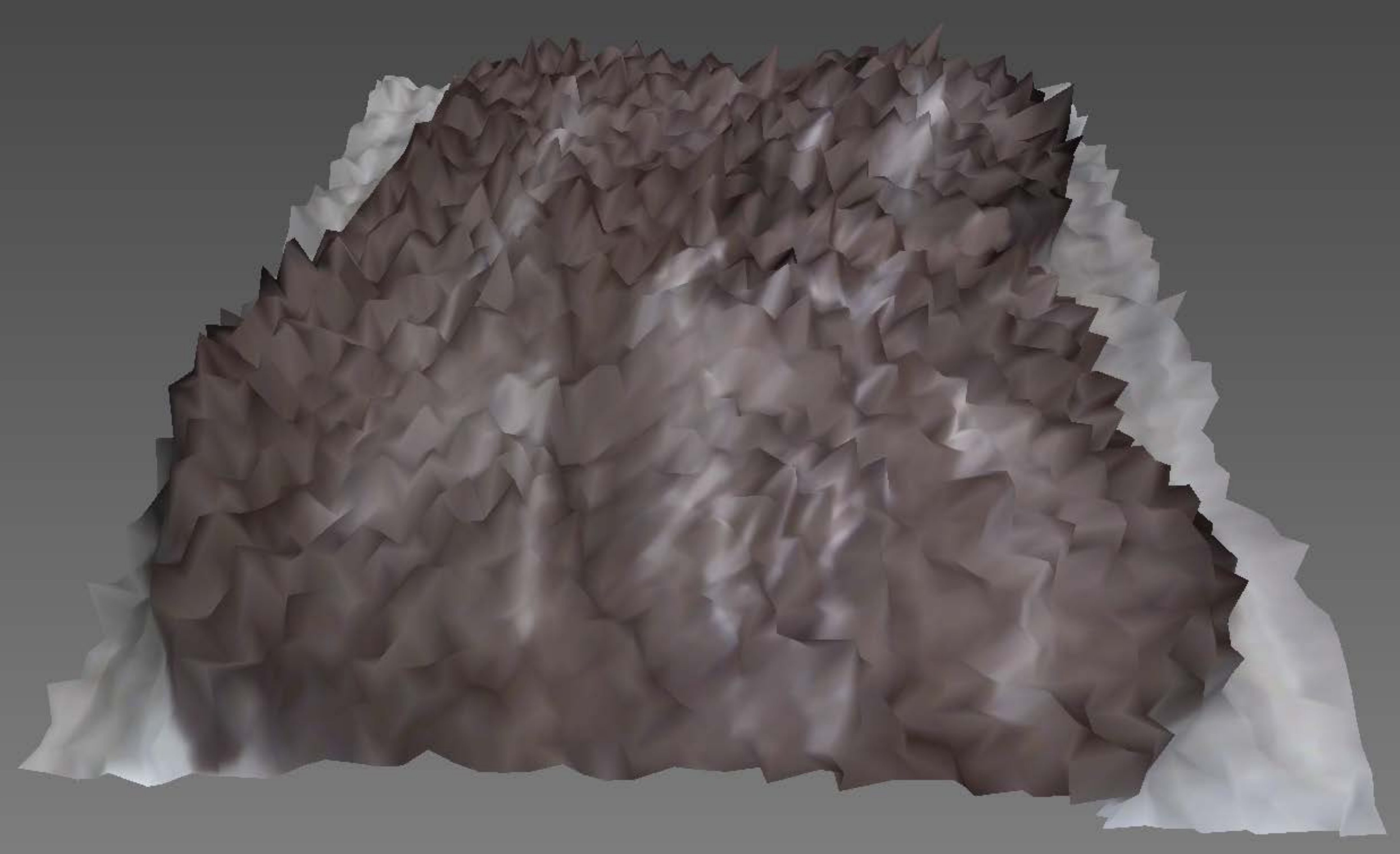}}
	\caption[Color and \gls{tof} sensor data fusion for image-guided surgery]{Color and \gls{tof} range measurements of porcine organs along with textured mesh representations to visualize sensor data fusion. Top row: ex-vivo data captured with a hybrid 3-D endoscope in minimally invasive surgery (see \sref{sec:08_ApplicationToHybrid3DEndoscopy}). Bottom row: ex-vivo data acquired with an imaging setup applicable for open surgery (see \sref{sec:08_ApplicationToOpenSurgery}).}
	\label{fig:exVivoDataExample}
\end{figure}

In terms of the technical implementation of hybrid range imaging, there exist various approaches with individual pros and cons in image-guided surgery \cite{Maier-Hein2013, Maier-Hein2014}. One of the historically first approaches to gain range data is stereo vision. Stereoscopy features a \textit{passive} approach that utilizes geometric correspondences across two views of the same scene, \eg pairs of corresponding points, to triangulate range data. The advantage of stereoscopy is that it can capture highly accurate measurements under ideal situations and has been engineered in stereo-based endoscopes \cite{Field2009}. However, under realistic conditions in image-guided surgery it is error prone due to repetitive image structures or texture-less surfaces. In image-guided surgery, \textit{active} senor technologies such as \gls{tof} \cite{Penne2009} or structured light \cite{Schmalz2012} provide a promising alternative and hold the potential to capture dense range images in real-time. Unfortunately, one of their major shortcomings are their spatial resolutions that are rather low compared to modern color cameras. This means a major barrier to employ such sensors in clinical workflows. \Fref{fig:exVivoDataExample} depicts ex-vivo \gls{tof} measurements alongside with high-resolution color images to visualize this issue for endoscopy and open surgery. Here, the overlay of range and color data demonstrates the complementary natures of both technologies. While color sensors capture photometric information of high spatial resolution, range sensors provide 3-D information that is acquired at a lower resolution. Range sensors might also suffer from a low \gls{snr} due to random or systematic errors, which is common in case of current \gls{tof} sensors \cite{Kolb2010,Fursattel2016}. In order to overcome the low spatial resolution of range sensors in these systems, this chapter adopts the multi-sensor super-resolution framework introduced in \cref{sec:MultiSensorSuperResolutionUsingGuidanceImages}. Accordingly, we employ high-resolution color images as guidance to super-resolve low-resolution range data.

The remainder of this chapter is structured as follows. \sref{sec:08_SystemCalibrationAndSensorDataFusion} introduces system calibration techniques to facilitate sensor data fusion for hybrid range imaging in image-guided surgery as prerequisite for multi-sensor super-resolution. \sref{sec:08_ExperimentsAndResults} presents a quantitative simulation study along with ex-vivo experiments for super-resolution in minimally invasive and open surgery workflows. \sref{sec:08_Conclusion} draws a conclusion for these studies.

\section{System Calibration and Sensor Data Fusion}
\label{sec:08_SystemCalibrationAndSensorDataFusion}

The super-resolution method proposed in \cref{sec:MultiSensorSuperResolutionUsingGuidanceImages} has the goal to reconstruct high-resolution range data from a set of low-resolution range images \smash{$\LRSequence{1}{K}$} to facilitate accurate 3-D measurements for image-guided surgery. This framework is driven by high-resolution color images \smash{$\GuideSequence{1}{K}$} that encode photometric information of the same scene. According to \cref{sec:MultiSensorSuperResolutionUsingGuidanceImages}, color images are exploited for motion estimation, spatially adaptive regularization as well as outlier detection in the underlying reconstruction algorithm, see \fref{fig:multisensorFlowchart}. This requires a pixel-wise mapping across both modalities, which is unknown a priori. Accordingly, multi-sensor super-resolution necessitates a system calibration to establish the mapping.

In this section, two calibration schemes are presented that are applicable to hybrid range imaging systems in image-guided surgery. This includes a homography approach that is applicable to \textit{beam splitter} setups as well as a \textit{stereo vision} approach that involves intrinsic and extrinsic camera calibrations. For more technical details on these approaches and their comparative evaluation, we refer to \cite{Haase2016}.
\begin{figure}[!t]
	\centering
		\includegraphics[width=1.00\textwidth]{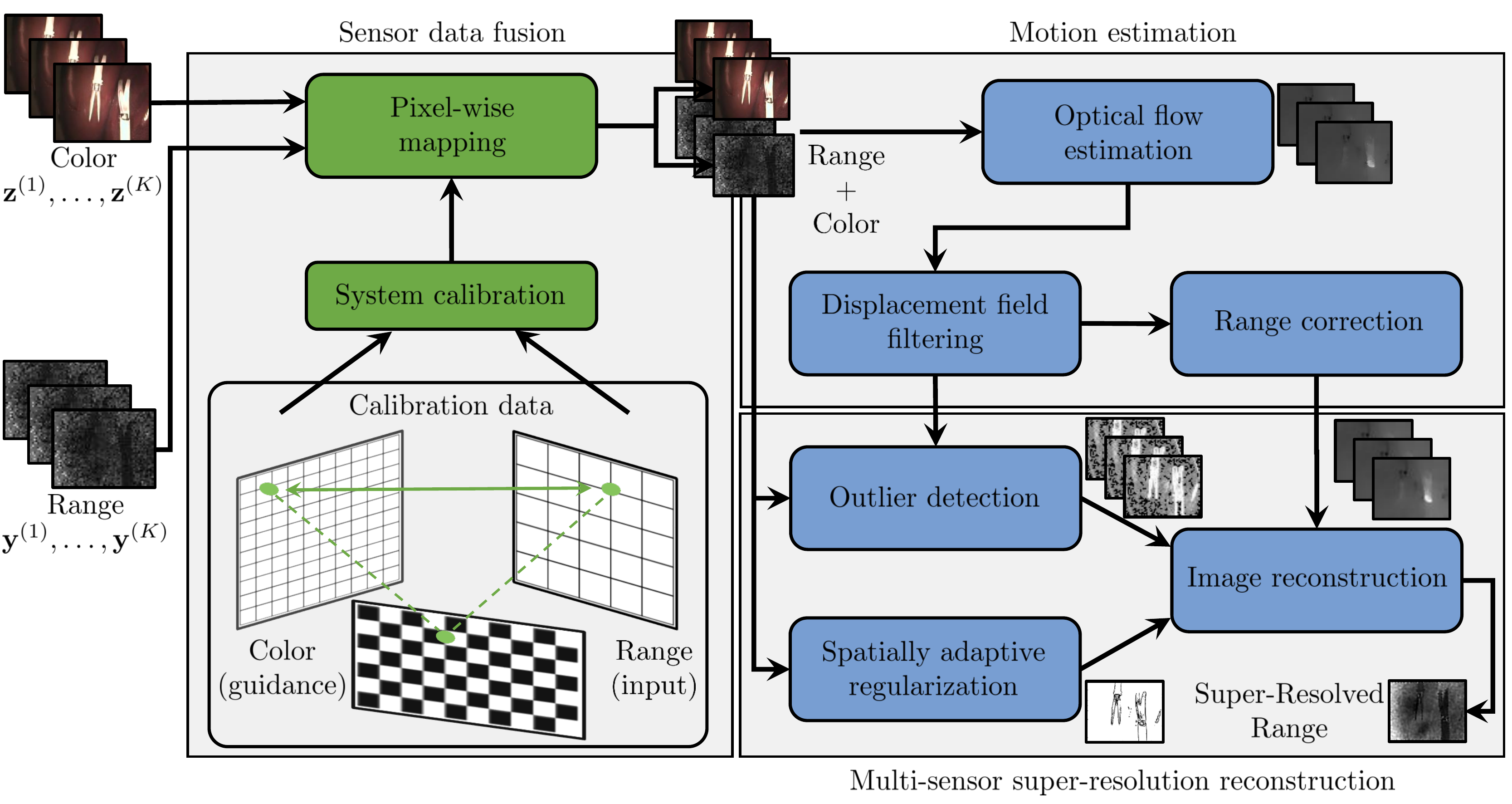}
	\caption[Multi-sensor super-resolution for hybrid range imaging in image-guided surgery]{Flowchart of multi-sensor super-resolution for hybrid range imaging in image-guided surgery. The sensor fusion between range and color images is gained by system calibration using geometric correspondences among both modalities. Then, the fused images are used for motion estimation and the reconstruction of super-resolved range data.}
	\label{fig:multisensorFlowchart}
\end{figure}

\subsection{Sensor Data Fusion using a Homography}
\label{sec:08_SensorDataFusionUsingHomographicMapping}

The first approach assumes that a 3-D surface is measured by a single optical system that acquires range and photometric information simultaneously. This can be implemented by means of a beam splitter that decomposes incoming light into two parts according to the wavelength, see \fref{fig:beamSplitterSetup}. Range and photometric data is captured by two separate sensors that have the same view to the underlying scene. 

In order to perform sensor data fusion, we employ the calibration approach introduced by Haase \etal \cite{Haase2013} that has been later adopted for multi-sensor super-resolution in \cite{Kohler2013a}. For the task of system calibration, let $\Point_{\LRSym}$ and $\Point_{\GuideSym}$ be a pair of corresponding points on a checkerboard calibration pattern in a range and a color image, respectively\footnote{In order to detect feature points in range data, the amplitude images provided by a \gls{tof} sensor after contrast enhancement and binarization as shown in \cite{Haase2013} can be used.}. The relationship between these points is modeled according to:
\begin{equation}
	\tilde{\Point}_{\GuideSym} \cong \vec{H}_{\LRSym \GuideSym} \tilde{\Point}_{\LRSym},
\end{equation}
where \smash{$\tilde{\Point}_{\GuideSym} \in \RealN{3}$} and \smash{$\tilde{\Point}_{\LRSym} \in \RealN{3}$} denote the points $\Point_{\GuideSym}$ and $\Point_{\LRSym}$ in homogeneous coordinates \cite{Hartley2004}. The homography $\vec{H}_{\LRSym \GuideSym} \in \RealMN{3}{3}$\label{notation:sensorFusionHomography} describes this mapping up to scale denoted by $\cong$. For system calibration, a set of point correspondences is identified by a self-encoded marker \cite{Forman2011} and the homography $\vec{H}_{\LRSym \GuideSym}$ is found by least-squares estimation \cite{Bradski2000}. Then, the homography is used to fuse range and color images in a common coordinate system. In the proposed framework, each color image is warped towards the corresponding range image, up to a scale factor to preserve the spatial resolution.
\begin{figure}[!t]
	\centering
		\includegraphics[width=0.51\textwidth]{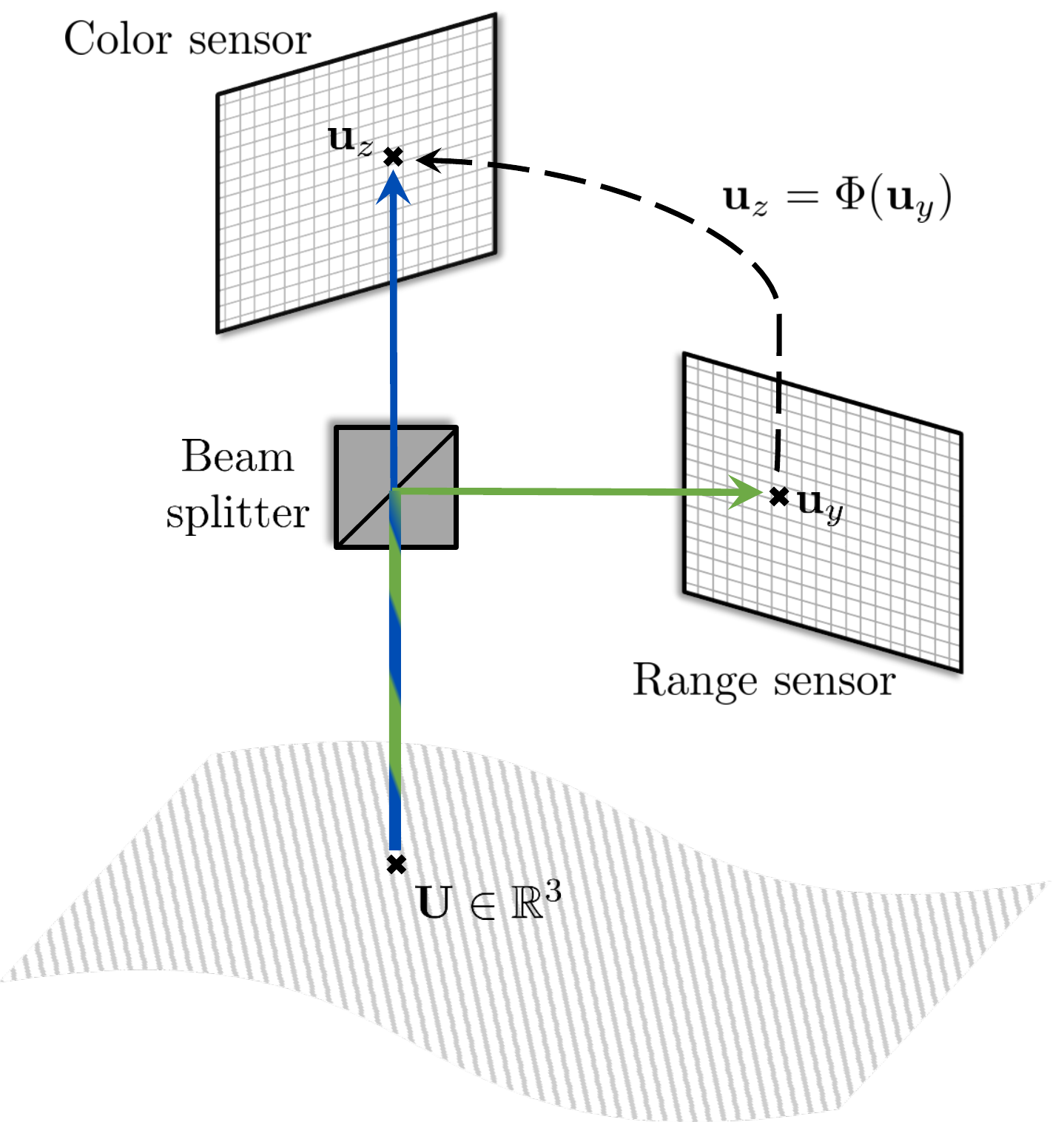}
	\caption[Simultaneous acquisition of range and photometric data with one common optical system]{Geometry of the system setup for the simultaneous acquisition of range and photometric data with one common optical system. The beam splitter decomposes incoming light into a range and a photometric signal. In this approach, the mapping of photometric data towards a range image (and vice versa) is modeled by a homography.}
	\label{fig:beamSplitterSetup}
\end{figure}

The use of a homography offers a couple of useful properties. One essential property is the possible inversion of the mapping between both modalities. In addition, the homography enables sensor data fusion solely with corresponding point pairs without involving intrinsic or extrinsic camera calibration. In practice, the estimated homography leads to re-projection errors of subpixel accuracy \cite{Haase2013}, which enables a highly accurate sensor data fusion. In this chapter, this approach is utilized for hybrid 3-D endoscopy. Here, it provides the acquisition of color and range data through one single endoscope equipped with a beam splitter.  

\subsection{Sensor Data Fusion using Stereo Vision}
\label{sec:08_SensorDataFusionUsingStereoVision}

The second approach considers the case that a 3-D surface is captured with two separate cameras with respective optical systems. These cameras acquire color and range information of the same scene from different viewpoints. Using a temporal synchronization, we can combine both modalities. The advantage of this setup is that it is not necessary to combine range and color sensors by a common optical system, see \fref{fig:stereoVisionSetup}. This is beneficial as it increases the flexibility in terms of the camera hardware. In the applications presented below, this setup is examined for image-guided open surgery.
\begin{figure}[!t]
	\centering
		\includegraphics[width=0.51\textwidth]{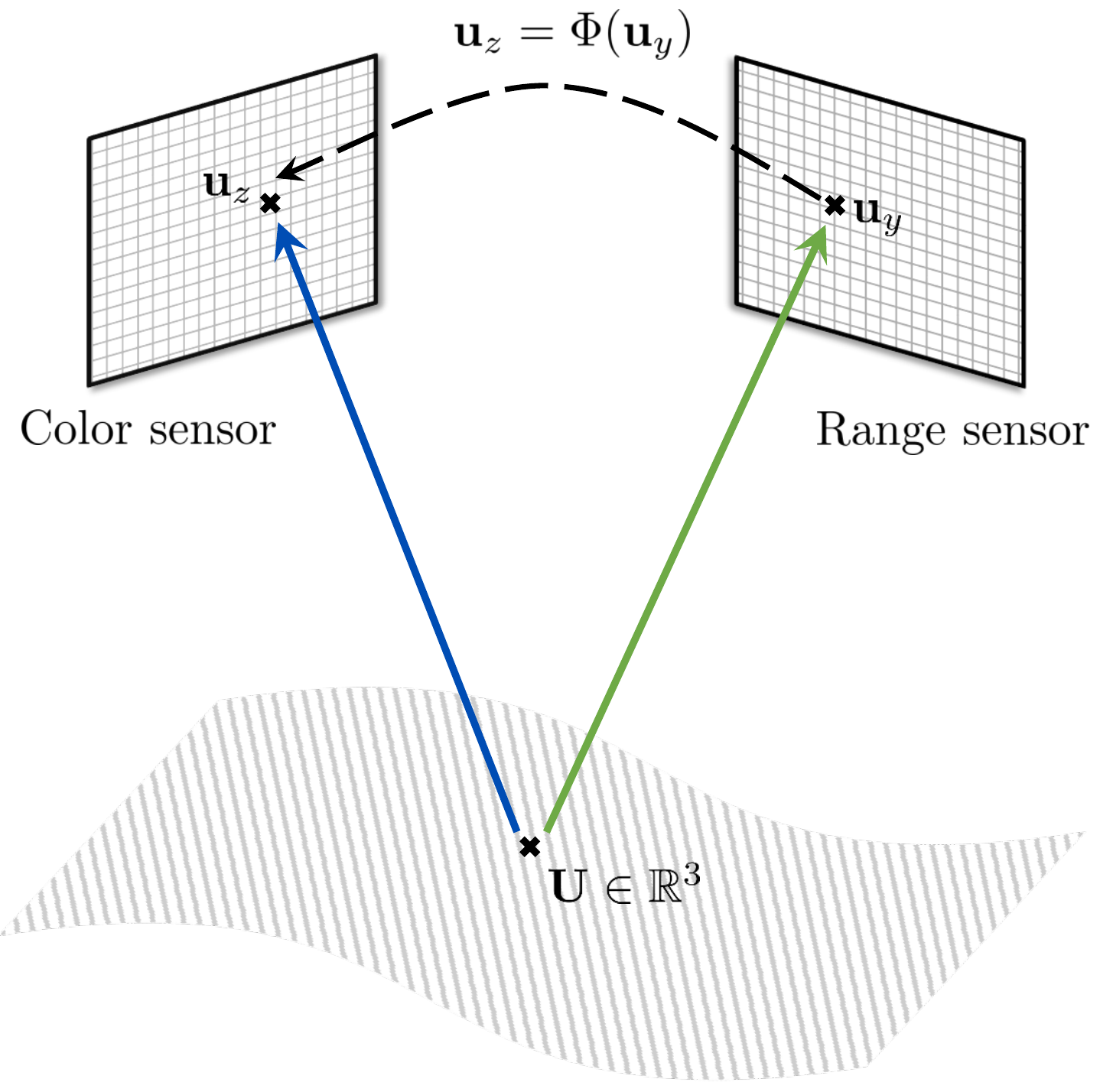}
	\caption[Simultaneous acquisition of range and photometric data with two separate sensors and optics]{Geometry of the system setup for the simultaneous acquisition of range and photometric data with two separate sensors and optics. The mapping of photometric data towards a range image is determined by stereo calibration for both cameras.}
	\label{fig:stereoVisionSetup}
\end{figure}

Unlike in the beam splitter setup, the system calibration cannot be described by a homography. As a consequence, one needs to perform stereo camera calibration \cite{Hartley2004} to fuse color and range images. In this work, sensor fusion is accomplished according to the calibration method introduced in \cite{Haase2012} that has been later used for multi-sensor super-resolution in \cite{Kohler2015a}. First, a point $\Point = (\CoordU_\LRSym, \CoordV_\LRSym)$ on the range camera image plane with the measured range value $\LRSym$ is re-projected to the 3-D space according to:
\begin{equation}
	\label{eqn:rangePointReprojection}
	\tilde{\vec{U}} \cong \vec{P}_{\LRSym}^{-1} \begin{pmatrix} \CoordU_\LRSym & \CoordV_\LRSym & \LRSym & 1 \end{pmatrix}^\top,
\end{equation}
where $\vec{P}_{\LRSym} \in \RealMN{4}{4}$\label{notation:rangeCameraProjMatrix} is the range camera projection matrix as shown in the calibration approach of Park \etal \cite{Park2011}. Subsequently, the re-projected point given in homogeneous coordinates $\tilde{\vec{U}} \in \RealN{4}$ can be projected onto the image plane of the color camera. Using the re-projected point in \eref{eqn:rangePointReprojection}, the corresponding pixel position in homogenous coordinates is given by:
\begin{equation}
	\label{eqn:colorPointProjection}
	\begin{split}
		\tilde{\Point}_\GuideSym 
			&\cong \vec{P}_{\GuideSym} \tilde{\vec{U}} \\
			&= \vec{P}_{\GuideSym} \vec{P}_{\LRSym}^{-1} \begin{pmatrix} \CoordU_\LRSym & \CoordV_\LRSym & \LRSym & 1 \end{pmatrix}^\top,
	\end{split}
\end{equation}
where $\vec{P}_{\GuideSym} \in \RealMN{3}{4}$\label{notation:colorCameraProjMatrix} denotes the projection matrix of the color camera. The projection matrices $\vec{P}_{\GuideSym}$ and $\vec{P}_{\LRSym}$ in \eref{eqn:colorPointProjection} are determined by intrinsic and extrinsic camera calibration. This is done using checkerboard calibration patterns to establish point correspondences for the calibration and least-squares optimization. The calibration procedure yields the intrinsic calibration matrices $\vec{K}_{\LRSym} \in \RealMN{3}{3}$\label{notation:rangeCameraCalibMatrix} for the range camera and $\vec{K}_{\GuideSym} \in \RealMN{3}{3}$\label{notation:colorCameraCalibMatrix} for the color camera as well as the extrinsic parameters given by the rotation matrix $\RotMat \in \RealMN{4}{4}$\label{notation:extrinsicRotation} and the translation vector $\TransVec \in \RealN{4}$\label{notation:extrinsicTranslation}. Similar to the homography approach, this method is used to warp color images to the domain of the range data up to a scale factor. However, notice that the calibrated mapping is not invertible and is affected by occlusions.
 
Compared to the homography approach, a stereo calibration suffers from shortcomings in terms of its accuracy under practical conditions. In particular, the calibration accuracy is highly dependent on the reliability of the measured range data as these measurements are used for the re-projection in \eref{eqn:rangePointReprojection}. In order to deal with random measurement noise, the calibration is performed on preprocessed range data using the filter pipeline proposed in \cite{Wasza2011}. Moreover, for the compensation of systematic errors in the range data \cite{Kolb2010}, the extrinsic parameters are further refined after stereo calibration. For this purpose, the translation vector $\TransVec$ is refined by optimizing the normalized mutual information \cite{Pluim2003} between range and color data to alleviate a potential bias in the calibration.
   
\section{Experiments and Results}
\label{sec:08_ExperimentsAndResults}

This section present an experimental evaluation of multi-sensor super-resolution for hybrid range imaging systems in image-guided surgery. For a quantitative evaluation, a comprehensive simulation study is presented to validate accuracy and robustness of super-resolution in the desired applications. Subsequently, the applicability of the proposed framework is demonstrated by ex-vivo experiments for hybrid 3-D endoscopy as well as image guidance in open surgery.

\subsection{Simulated Data Experiments}
\label{sec:08_SimulatedDataExperiments}

\begin{figure}[!t]
	\centering
	\raisebox{0.65cm}{\rotatebox{90}{\footnotesize Ground truth}}\quad
	\subfloat{\includegraphics[width=0.33\textwidth]{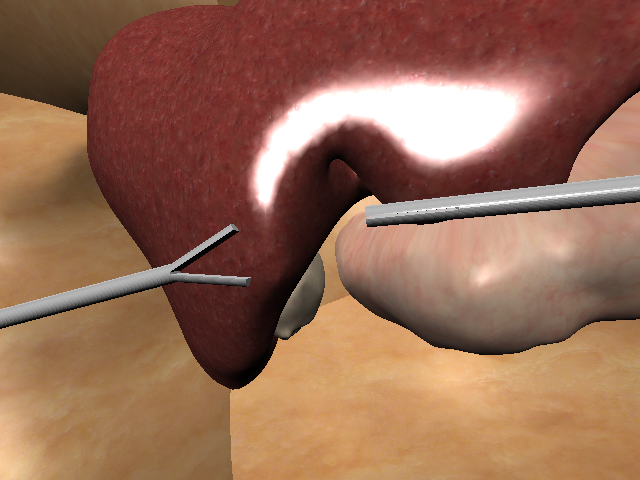}}~
	\subfloat{\includegraphics[width=0.33\textwidth]{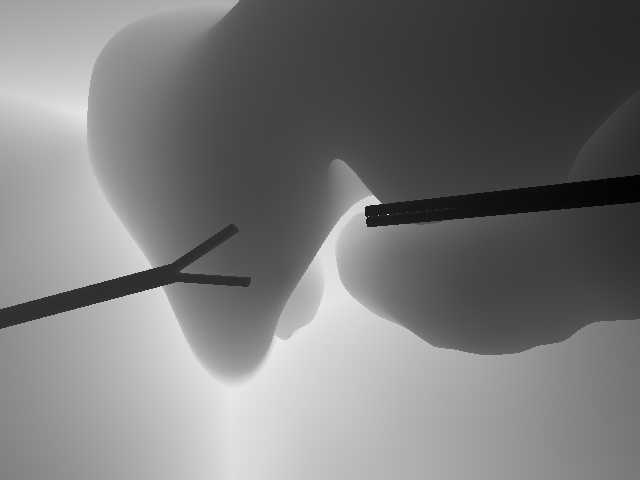}}\\[-0.7ex]
	\setcounter{subfigure}{0}
	\raisebox{0.65cm}{\rotatebox{90}{\footnotesize Simulated data}}\quad
	\subfloat[Color data]{\includegraphics[width=0.33\textwidth]{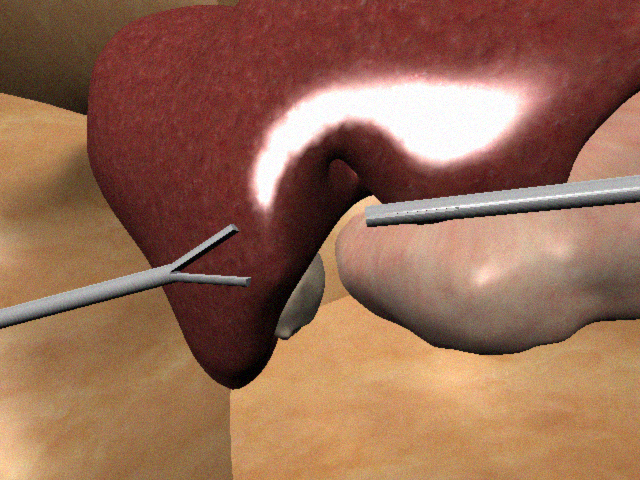}}~
	\subfloat[Range data]{\includegraphics[width=0.33\textwidth]{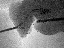}}
	\caption[Color and range images from an artificial laparoscopic scene]{Color and range image obtained from an artificial laparoscopic scene. Simulated data (bottom row) is generated from the ground truth (top row) using subsampling, blurring as well as conditions of \gls{tof} endoscopy like specular highlights.}
	\label{fig:simulatedDatasetExample}
\end{figure}

In the following study, we used artificial hybrid range data from the publicly available Multi-Sensor Super-Resolution Datasets\footnote{\url{https://www5.cs.fau.de/research/data/multi-sensor-super-resolution-datasets/}} for a quantitative evaluation. These range and color images were obtained from an artificial laparoscopic scene under the conditions of minimally invasive surgery using the \gls{ritk} \cite{Wasza2011b}. Ground truth range and color images were gained from the artificial 3-D model and both modalities were perfectly aligned to exclude the influence of  calibration errors in this baseline experiment. Color images were encoded with a pixel resolution of 640$\times$480\,px and disturbed by a Gaussian \gls{psf} ($\PSFWidth = 0.5$) as well as additive Gaussian noise ($\NoiseStd = 0.001$). The corresponding range images were simulated with a pixel resolution of 64$\times$48\,px. To analyze the influence of systematic errors, the simulation considered the following effects of \gls{tof} imaging and surgical interventions. First, as opposed to space invariant noise, Gaussian noise in range data was simulated to be distance-dependent with a maximum standard deviation of $\NoiseStd = 10$\,mm in the scene space. Second, specular highlights in color images were simulated as a common issue in endoscopy \cite{Haase2014}. These specular highlights resulted in range measurements disturbed by Perlin noise in the affected image regions. Finally, range data was corrupted by \textit{flying pixels} \cite{Kolb2010} that were generated by randomly flipping 20\,\% of all pixels located on depth discontinuities.

The artificial model was used to generate image sequences of the scene from different perspectives along with surgical instruments, see \fref{fig:simulatedDatasetExample}. Movements of an hand-held endoscope were simulated by a randomly generated rigid motion of the virtual camera in the scene space. In addition, movements of surgical instruments and soft tissue were simulated to consider realistic conditions in minimally invasive surgery. We use the superposition of these 3-D movements that appear as 2-D subpixel motion in range and color images as a cue for super-resolution. 

We examine the reconstruction methods that were previously introduced in \sref{sec:05_ApplicationToHybridRangeImaging} using a Huber prior with $\HuberThresh = 5 \cdot 10^{-4}$ and $\RegWeight = 0.8$. See \tref{tab:08_msrReconstructionAlgos} for an overview of the configurations of the different multi-sensor techniques. Throughout the following experiments, the single-sensor reconstruction algorithm (\gls{ssr}) that works solely on the range data is considered as the baseline and compared to the different multi-sensor methods (\gls{msr}, \gls{amsr}, and \gls{amsrod}).
\begin{table}[t]
  \centering
  \caption[Range super-resolution algorithms along with their parameter settings]{Overview of range super-resolution algorithms along with their parameter settings. Multi-sensor super-resolution is employed with different algorithm profiles (\gls{msr}, \gls{amsr}, \gls{amsrod}). Single-sensor super-resolution (\gls{ssr}) is considered as the baseline.}
	\small
	\begin{tabular}{l ccc}
		\toprule
		\textbf{Reconstruction algorithm} & \multicolumn{3}{c}{\textbf{Algorithm properties}} \\
		\cmidrule{2-4}
		& \textbf{Motion} & \textbf{Adaptive} & \textbf{Outlier} \\
		& \textbf{estimation} & \textbf{regularization} & \textbf{detection} \\
		\midrule
		Single-sensor super-resolution & direct & \xmark & \xmark \\
		(SSR) & & & \\
		\midrule
		Multi-sensor super-resolution & filter-based & \xmark & \xmark \\
		(MSR) & & & \\
		\midrule
		Adaptive multi-sensor super-resolution & filter-based & \cmark & \xmark \\
		(AMSR) & & $\AMSRContrastFactor = 0.025$, $\AMSRPatchSize = 7$ & \\
		\midrule
		Adaptive multi-sensor super-resolution & filter-based & \cmark & \cmark \\
		with outlier detection (AMSR-OD) & & $\AMSRContrastFactor = 0.025$, $\AMSRPatchSize = 7$ & $\rho_{0} = 0.5$ \\
		\bottomrule
	\end{tabular}
	\label{tab:08_msrReconstructionAlgos}
\end{table}

\paragraph{Accuracy Analysis.}
In order to conduct a baseline experiment, four artificial datasets (S1 - S4) were generated from the given laparoscopic scene. Throughout all experiments, super-resolution was performed with magnification $s = 4$ and $K = 31$ frames, where the central one was used as reference for variational optical flow computation \cite{Liu2009}. The reconstructions were conducted in a sliding window scheme using $K$ successive frames to obtain single super-resolved images.

The statistics of the \gls{psnr} and \gls{ssim} of super-resolved range data reconstructed by the different algorithms on ten randomly generated image sequences per dataset are reported in \fref{fig:msrAccuracyAnalysis}. Notice that multi-sensor super-resolution consistently outperformed the single-sensor approach on all datasets. The most substantial improvements were obtained under challenging situations for optical flow estimation due to motion of soft tissue and instruments along with movements of the virtual endoscope. The multi-sensor formulation considerably increased the accuracy of the motion estimate using the underlying filter-based technique, which resulted in accurate range super-resolution. In addition, spatially adaptive regularization (\gls{amsr}) leveraged the reconstruction of depth discontinuities compared to non-adaptive regularization (\gls{msr}). This affects the measurement of anatomical structures or surgical instruments. Moreover, outlier detection (\gls{amsrod}) enhanced the robustness against individual misregistered frames as well as outliers in the low-resolution range data, \eg space variant random noise or systematic errors. This is notably for situations where the filter-based motion estimation cannot establish a reliable displacement field since optical  flow computation on color images entirely failed. See \fref{fig:msrAccuracyAnalysisExample} for a visual comparison on two example datasets (S2 and S3) in these situations. Here, the full combination of the proposed multi-sensor techniques (\gls{amsrod}) provides reliable range data including accurate reconstructions of soft tissue and surgical instruments.
\begin{figure}[!t]
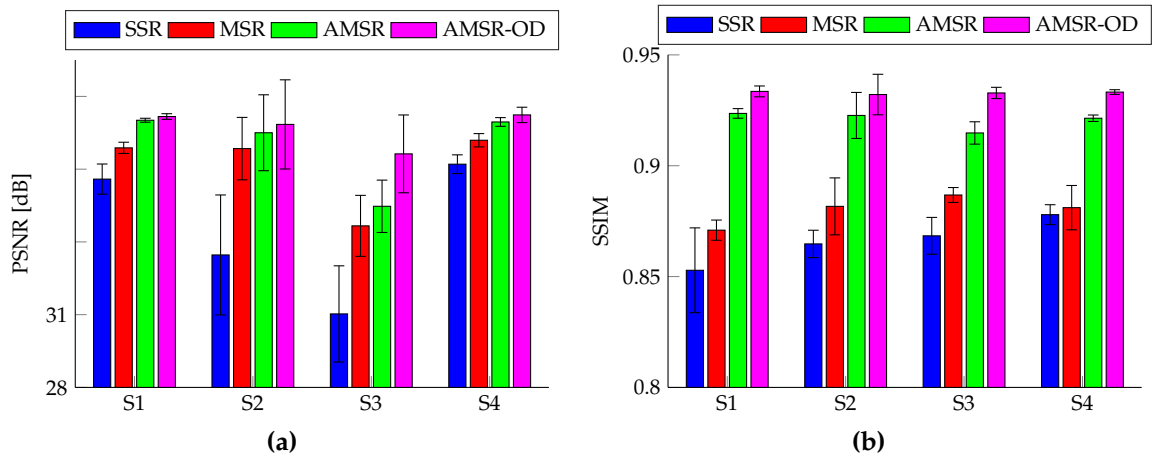

	\centering
	\scriptsize
	\setlength \figurewidth{0.41\textwidth}
	\setlength \figureheight{0.78\figurewidth} 
	\tikzset{external/export next=false}	
	\subfloat[]{\input{images/chapter8/simulatedData_psnr.tikz}}\quad
	\tikzset{external/export next=false}	
	\subfloat[]{\input{images/chapter8/simulatedData_ssim.tikz}}
	\caption[\gls{psnr} and \gls{ssim} of single-sensor and multi-sensor algorithms on simulated laparoscopic data]{Single-sensor super-resolution (\gls{ssr}) versus the different multi-sensor algorithms (\gls{msr}, \gls{amsr}, and \gls{amsrod}) on an artificial laparoscopic scene. The statistics (mean $\pm$ standard deviation) of the \gls{psnr} and \gls{ssim} measures were determined on four datasets (S1 - S4) with known ground truth range data. For each dataset, super-resolution was applied for ten consecutive image sequences using sliding window processing.}
	\label{fig:msrAccuracyAnalysis}
\end{figure}

\begin{figure}[!p]
	\centering
	\subfloat{\includegraphics[width=0.32\textwidth]{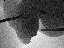}}~
	\subfloat{\includegraphics[width=0.32\textwidth]{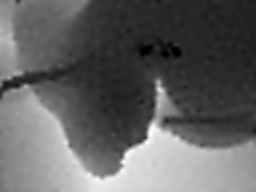}}~
	\subfloat{\includegraphics[width=0.32\textwidth]{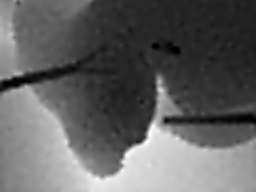}}\\[-0.7em]
	\setcounter{subfigure}{0}
	\subfloat[Input range]{\includegraphics[width=0.32\textwidth]{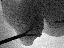}}~
	\subfloat[\gls{ssr}]{\includegraphics[width=0.32\textwidth]{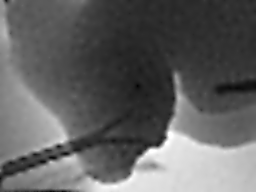}}~
	\subfloat[\gls{msr}]{\includegraphics[width=0.32\textwidth]{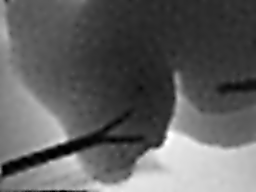}}\\
	\subfloat{\includegraphics[width=0.32\textwidth]{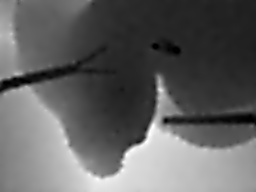}}~
	\subfloat{\includegraphics[width=0.32\textwidth]{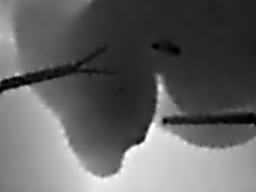}}~
	\subfloat{\includegraphics[width=0.32\textwidth]{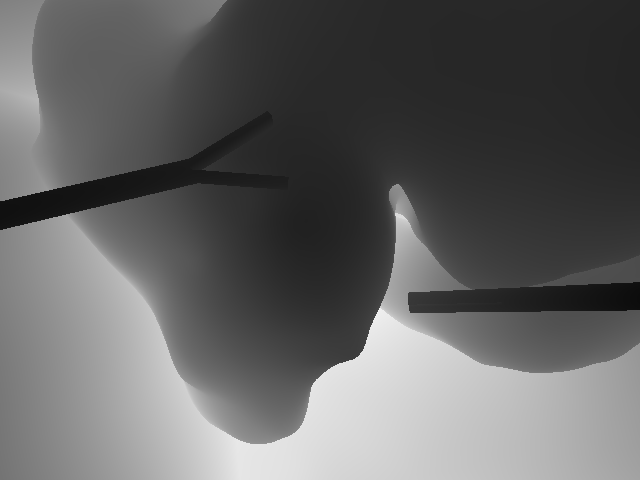}}\\[-0.7em]
	\setcounter{subfigure}{3}
	\subfloat[\gls{amsr}]{\includegraphics[width=0.32\textwidth]{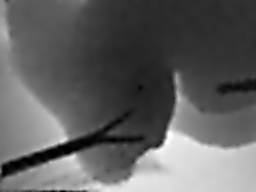}}~
	\subfloat[\gls{amsrod}]{\includegraphics[width=0.32\textwidth]{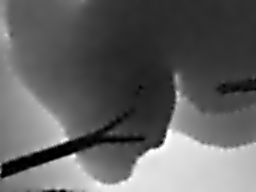}}~
	\subfloat[Ground truth]{\includegraphics[width=0.32\textwidth]{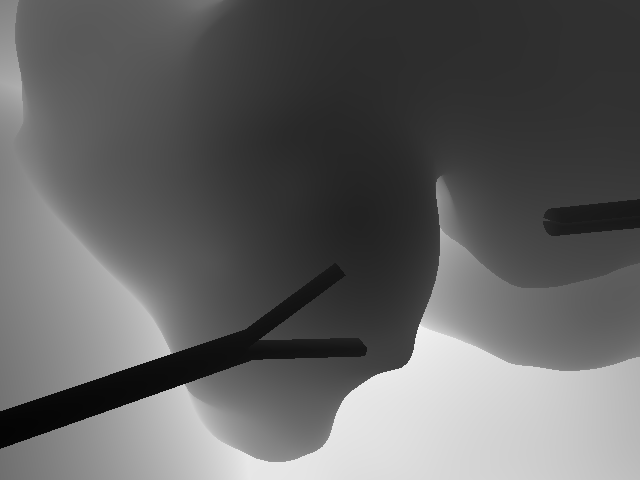}}
	\caption[Super-resolution on an artificial laparoscopic scene]{Super-resolution reconstruction ($K = 31$ frames, $4 \times$ magnification) on two datasets of an artificial laparoscopic scene. First and third row: comparison of low-resolution range data to single-sensor super-resolution (\gls{ssr}) and the different multi-sensor approaches (\gls{msr}, \gls{amsr}, and \gls{amsrod}) for the dataset S2. Notice that direct motion estimation on range data as implemented by the \gls{ssr} approach failed, which resulted in unreliable super-resolved range information. Second and fourth row: comparison for the dataset S3. Note that outlier detection as implemented by \gls{amsrod} compensated for outliers in optical flow that are related to difficult motion types, \eg endoscope movements superimposed with independently moving surgical instruments.}
	\label{fig:msrAccuracyAnalysisExample}
\end{figure}

\paragraph{Robustness Analysis.}
In terms of the robustness, multi-sensor super-resolution is affected by the conditions in image-guided surgery. Let us study two important issues that are related to motion estimation and system calibration.

One important problem case is the uncertainty of optical flow estimation under realistic conditions. This issue was investigated by intentionally disturbing the optical flow determined on color images by zero-mean, normal distributed noise with standard deviation $\sigma_{\text{OFL}}$ in each displacement component. In practice, this situation might appear in texture-less regions on organ surfaces resulting in unreliable displacement fields. \Fref{fig:msrRobustnessAnalysis:opticalFlowNoise} shows the impact of noisy optical flow for the dataset S1 at different noise levels $\sigma_{\text{OFL}}$ measured in terms of pixels of the color images. Notice that even for large noise levels, the different multi-sensor approaches were nearly insensitive to noisy optical flow and considerably outperformed the single-sensor reconstruction. This behavior is related to the filter-based motion estimation as an integral part of multi-sensor super-resolution, which gets rid of the noise present in the optical flow of the color images.

So far, the fusion among color and range data was assumed to be exact. In practice, however, the actual accuracy is highly affected by unavoidable calibration errors. This influences the reliability of the multi-sensor reconstruction as it relies on accurate sensor data fusion. For a robustness analysis, small misalignments between both modalities were intentionally induced in the simulation process to consider calibration errors. These misalignments were obtained by randomly generated translations of the color images relative to the range data. The behavior of super-resolution at different misalignments measured by the translation length $\epsilon_t$ in terms of pixels of color data  is shown in \fref{fig:msrRobustnessAnalysis:mappingError} for the dataset S1. Notice that in contrast to single-sensor super-resolution, the accuracy of the different multi-sensor approaches dropped under increasing misalignments. In particular, spatially adaptive regularization as well as outlier detection were sensitive regarding this effect, while filter-based motion estimation was less affected. However, the different multi-sensor approaches still outperformed the competing single-sensor approach confirming a reasonable robustness against calibration errors.
\begin{figure}[!t]
	\centering
	\scriptsize 
	\setlength \figurewidth{0.40\textwidth}
	\setlength \figureheight{0.77\figurewidth} 
	\tikzset{external/export next=false}
	\subfloat[\gls{psnr} vs. flow noise level $\sigma_{\text{OFL}}$]{\input{images/chapter8/oflNoise_psnr.tikz}
	\label{fig:msrRobustnessAnalysis:opticalFlowNoise}}\quad
	\tikzset{external/export next=false}
	\subfloat[\gls{psnr} vs. calibration error $\epsilon_t$]{\input{images/chapter8/mappingError_psnr.tikz}
	\label{fig:msrRobustnessAnalysis:mappingError}}
	\caption[Robustness analysis of multi-sensor super-resolution]{Robustness analysis of different multi-sensor approaches (\gls{msr}, \gls{amsr}, and \gls{amsrod}) using the single-sensor approach (\gls{ssr}) as baseline. 
	\protect\subref{fig:msrRobustnessAnalysis:opticalFlowNoise} Sensitivity regarding optical flow corrupted by Gaussian noise with standard deviation $\sigma_{\text{OFL}}$ measured in pixels of the color images. 
	\protect\subref{fig:msrRobustnessAnalysis:mappingError} Sensitivity regarding calibration errors simulated by translational misalignments of length $\epsilon_t$ measured in pixels of the color images.}
	\label{fig:msrRobustnessAnalysis}
\end{figure}

\subsection{Application to Hybrid 3-D Endoscopy}
\label{sec:08_ApplicationToHybrid3DEndoscopy}

This section demonstrates the application of multi-sensor super-resolution in the area of minimally invasive surgery. For this study, ex-vivo experiments were conducted by measuring porcine organs with a hybrid 3-D endoscope\footnote{All experiments for this study were conducted with the hybrid 3-D endoscope prototype manufactured by the Richard Wolf GmbH, Knittlingen, Germany.}. Range data was captured with a \gls{tof} sensor that features a pixel resolution of 64$\times$48\,px at a frame rate of 30\,Hz. The corresponding color sensor provides a resolution of 640$\times$480\,px. Both sensors are combined in a single optical system that is equipped with a beam splitter to synchronize the acquisition of range and color images \cite{Haase2013}. Therefore, the homography approach presented in \sref{sec:08_SensorDataFusionUsingHomographicMapping} was used for calibration and sensor data fusion. 

To induce motion, the endoscope was shifted over time relative to the organ surface. In addition, surgical instruments were moved to consider the conditions of minimally invasive procedures. Super-resolution was performed with magnification $\MagFac = 4$ and $K = 31$ consecutive frames, where the central frame was chosen as reference for optical flow \cite{Liu2009}.

\paragraph{Reconstruction Results.}
\Fref{fig:endoExampleResults} depicts a comparison among the different reconstruction algorithms to low-resolution range data on one example dataset. In this application, we are particularly interested in reliable reconstructions of soft tissue surfaces and artificial objects, \eg the instrument tips. It is worth noting that these structures are difficult to detect in the measured range data. 

In comparison to the single-sensor reconstruction (\gls{ssr}), the proposed filter-based motion estimation substantially improved the reliability of the computed displacement fields. This resulted in a superior accuracy of the multi-sensor algorithm (\gls{msr}) in terms of the reconstruction of anatomical structures and surgical instruments. In contrast to the filter-based approach, direct optical flow estimation on range data as implemented for the single-sensor reconstruction was error prone and did not capture endoscope or instrument movements appropriately. Notice that spatially adaptive regularization (\gls{amsr}) enhanced the multi-sensor reconstruction even further. This translated into a superior recovery of depth discontinuities. Moreover, outlier detection (\gls{amsrod}) got rid of non-Gaussian noise related to systematic distance- and intensity-dependent errors in \gls{tof} imaging. This considerably improved the reconstruction of soft tissue surfaces.

\tikzset{external/export next=false}
\begin{figure}[!p]
	\centering
	\tikzset{external/export next=false}
	\subfloat[Input RGB]{	
		\begin{tikzpicture}[spy using outlines={rectangle, blue, magnification=2.0, height=3cm, width=4.5cm, connect spies, every spy on node/.append style={thick}}] 
			\node {\pgfimage[width=0.315\linewidth]{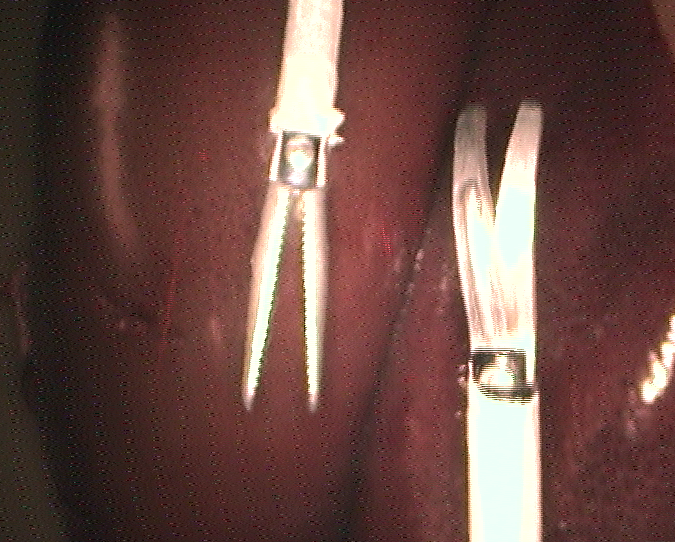}}; 
			\spy on (0.5, 0.5) in node [left] at (2.15, -3.5);
		\end{tikzpicture}
	}\hspace{-1.25em}
	\tikzset{external/export next=false}
	\subfloat[Input range]{	
		\begin{tikzpicture}[spy using outlines={rectangle, blue, magnification=2.0, height=3cm, width=4.5cm, connect spies, every spy on node/.append style={thick}}] 
			\node {\pgfimage[width=0.315\linewidth]{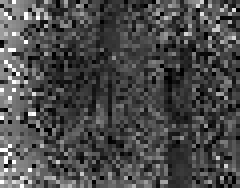}}; 
			\spy on (0.5, 0.5) in node [left] at (2.15, -3.5);
		\end{tikzpicture}
	}\hspace{-1.25em}
	\tikzset{external/export next=false}
	\subfloat[\gls{ssr}]{	
		\begin{tikzpicture}[spy using outlines={rectangle, blue, magnification=2.0, height=3cm, width=4.5cm, connect spies, every spy on node/.append style={thick}}] 
			\node {\pgfimage[width=0.315\linewidth]{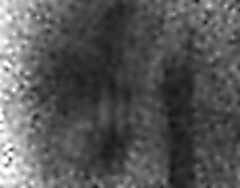}}; 
			\spy on (0.5, 0.5) in node [left] at (2.15, -3.5);
		\end{tikzpicture}
	}\\
	\tikzset{external/export next=false}
	\subfloat[\gls{msr}]{	
		\begin{tikzpicture}[spy using outlines={rectangle, blue, magnification=2.0, height=3cm, width=4.5cm, connect spies, every spy on node/.append style={thick}}] 
			\node {\pgfimage[width=0.315\linewidth]{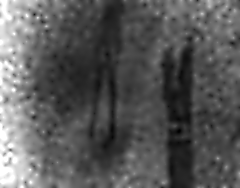}}; 
			\spy on (0.5, 0.5) in node [left] at (2.15, -3.5);
		\end{tikzpicture}
	}\hspace{-1.25em}
	\tikzset{external/export next=false}
	\subfloat[\gls{amsr}]{	
		\begin{tikzpicture}[spy using outlines={rectangle, blue, magnification=2.0, height=3cm, width=4.5cm, connect spies, every spy on node/.append style={thick}}] 
			\node {\pgfimage[width=0.315\linewidth]{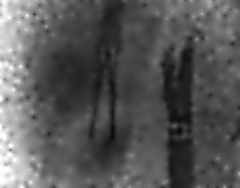}}; 
			\spy on (0.5, 0.5) in node [left] at (2.15, -3.5);
		\end{tikzpicture}
	}\hspace{-1.25em}
	\tikzset{external/export next=false}
	\subfloat[\gls{amsrod}]{	
		\begin{tikzpicture}[spy using outlines={rectangle, blue, magnification=2.0, height=3cm, width=4.5cm, connect spies, every spy on node/.append style={thick}}] 
			\node {\pgfimage[width=0.315\linewidth]{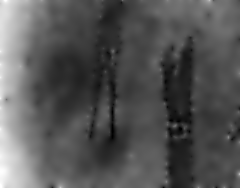}}; 
			\spy on (0.5, 0.5) in node [left] at (2.15, -3.5);
		\end{tikzpicture}
	}
	\caption[Ex-vivo experiments for hybrid \gls{tof}/RGB endoscopy on a porcine liver]{Super-resolution reconstruction ($K = 31$ frames, $4 \times$ magnification) for hybrid \gls{tof}/RGB endoscopy on a porcine liver. First and third row: high-resolution color image and low-resolution range data in comparison to super-resolved range images obtained by single-sensor super-resolution (\gls{ssr}) as well as the different multi-sensor algorithms (\gls{msr}, \gls{amsr}, and \gls{amsrod}). Second and fourth row: zoom-in for an image region that contains surgical instruments. Notice that single-sensor super-resolution failed to provide reliable range information, while the different multi-sensor algorithms considerably enhanced the accuracy of range information for soft tissue and the surgical instruments.}
	\label{fig:endoExampleResults}
\end{figure}
\tikzset{external/export next=true}

\paragraph{Range Data Quality Assessment.}
Two no-reference quality measures are used to quantitatively assess the reliability of range data. One criterion is noise reduction on reconstructed surfaces. To this end, a blind \gls{snr} estimation is performed. The measure that is used in this work is computed for flat surfaces according to:\label{notation:blindSNR}
\begin{equation}
	\label{eqn:blindSNR}
	Q_{\text{snr}} = 10 \log_{10} \left( \frac{\mu_{\text{flat}}}{\sigma_{\text{flat}}} \right),
\end{equation}
where $\mu_{\text{flat}}$ and $\sigma_{\text{flat}}$ denote the mean and standard deviation of the range measurements in a rectangular region of interest, respectively. This measure is defined in \gls{db} and the higher $Q_{\text{snr}}$, the more accurate the reconstruction of the surface.

Besides reliable surface reconstruction, another goal is the accurate reconstruction of object transitions. For this purpose, regions of interests that contain an edge between two structures in the range data are analyzed. The range values are described by a \gls{gmm} consisting of two components that represent foreground and background range values, respectively. Then, the quality measure to assess the reconstruction of depth discontinuities is defined as:\label{notation:edgeRecoMeasure}
\begin{equation}
	\label{eqn:edgeRecoMeasure}
	Q_{\text{edge}} = \frac{w_b (\mu_b - \mu)^2 + w_f (\mu_f - \mu)^2}{w_b \sigma_b^2 - w_f\sigma_f^2},
\end{equation}
where $\mu$ denotes the mean range value in the selected region, and $\mu_b$ and $\mu_f$ are the mean values of the background and the foreground, respectively. Similarly, $\sigma_b$ and $\sigma_f$ are the standard deviations, and $w_b$ and $w_f$ are the weights associated with the \gls{gmm} components. This model is fitted to the range values using $k$-means clustering ($k = 2$). Notice that lower estimates of $\sigma_b$ and $\sigma_f$ along with larger differences between $\mu_b$ and $\mu_f$ indicate a better discrimination between foreground and background. Accordingly, the higher $Q_{\text{edge}}$ the better the underlying reconstruction.

Six image regions per dataset containing flat surfaces and depth discontinuities were manually selected. The respective statistics of $Q_{\text{snr}}$ and $Q_{\text{edge}}$ for nine datasets are summarized in \tref{tab:blindMeasuresEndoscopy}. In comparison to raw range data, the combination of all multi-sensor techniques (\gls{amsrod}) leads to an increase of $Q_{\text{snr}}$ and $Q_{\text{edge}}$ by $64$\,\% and $68$\,\%, respectively. The different multi-sensor algorithms also improved the reconstruction of flat surfaces and depth discontinuities in comparison to a single-sensor approach (\gls{ssr}). These properties express a higher reliability of range data to facilitate segmentation or object detection \cite{Haase2013c}.
\begin{table}[t]
  \centering
	\small
  \caption[No-reference quality measures in the ex-vivo study for hybrid 3-D endoscopy]{Mean $\pm$ standard deviation of the no-reference quality measures $Q_{\text{snr}}$ and $Q_{\text{edge}}$ in the ex-vivo study for hybrid 3-D endoscopy. Both measures were determined using manually selected regions of interest in low-resolution and super-resolved range data gained by the different reconstruction algorithms. In total, nine datasets with six manually selected regions per dataset were used.}
	\begin{tabular}{p{4.0em} ccccc}
		\toprule
		\textbf{Measure} & \textbf{Low-res. data} & \multicolumn{4}{c}{\textbf{Super-resolved data}} \\[0.6ex]
		\cline{3-6} \\
		 &  & \gls{ssr} & \gls{msr} & \gls{amsr} & \gls{amsrod} \\
		\midrule
		$Q_{\text{snr}}$	& 7.2 $\pm$ 2.8 & 10.6 $\pm$ 2.1 & 10.6 $\pm$ 2.1 	& 11.4 $\pm$ 2.3 	& 11.8 $\pm$ 2.4  \\
		$Q_{\text{edge}}$	& 1.9 $\pm$ 0.4 & 2.4 $\pm$ 0.7 & 2.7 $\pm$ 1.1 	& 3.1 $\pm$ 1.7 	& 3.2 $\pm$ 2.0 	\\
		\bottomrule
	\end{tabular}
  \label{tab:blindMeasuresEndoscopy}
\end{table}

\subsection{Application to Image Guidance in Open Surgery}
\label{sec:08_ApplicationToOpenSurgery}

Let us now demonstrate the application of multi-sensor super-resolution for image guidance in open surgery. In contrast to hybrid 3-D endoscopy, a stereo camera setup was developed for ex-vivo measurements on a porcine liver. In order to measure the liver surface, range data was captured with a PMD\,CamCube\,3 \gls{tof} camera that provides a pixel resolution of 200$\times$200\,px at a frame rate of 30\,Hz. A Grasshopper2 camera with a resolution of 1200$\times$1200\,px was used to acquire color images and was temporally synchronized to the range sensor. Both cameras were coupled on a tripod with a baseline that was chosen as small as possible to minimize occlusions. This stereo setup was calibrated according to \sref{sec:08_SensorDataFusionUsingStereoVision} and color images were fused with the range data based on the intrinsic and extrinsic camera parameters. 

The motion required for super-resolution was induced by small vibrations of the tripod. Super-resolution was performed with magnification factor $\MagFac = 4$ using $K = 31$ consecutive range images, where the central one was chosen as reference for optical flow \cite{Liu2009}.

\paragraph{Reconstruction Results.}
\Fref{fig:camCubeExampleResults} shows a qualitative comparison among low-resolution range data acquired under this setup and super-resolution using the different reconstruction algorithms. In this use case, super-resolution aims at reconstructing reliable surface information of the measured porcine liver.

Similar to the ex-vivo experiments for minimally invasive surgery, one can observe that direct optical flow estimation on low-resolution range data was error prone. This resulted in a poor surface reconstruction provided by the single-sensor approach (\gls{ssr}), which is particularly visible by blurred boundaries of the porcine liver. The different multi-sensor approaches (\gls{msr}, \gls{amsr}, and \gls{amsrod}) were less sensitive to this issue due to the higher reliability of the filter-based motion estimation driven by the color images. Note that in this application, the quality gain achieved by spatially adaptive regularization was limited since surfaces were typically more smooth and thus edges could not be exploited by the underlying regularization technique. Moreover, outlier detection did not achieved substantial quality gains due to the higher reliability of the motion estimate and the measured range data.

\tikzset{external/export next=false}
\begin{figure}[!p]
	\centering
	\tikzset{external/export next=false}
	\subfloat[Input RGB]{	
		\begin{tikzpicture}[spy using outlines={rectangle, blue, magnification=2.0, height=3cm, width=2.3cm, connect spies, every spy on node/.append style={thick}}] 
			\node {\pgfimage[width=0.315\linewidth]{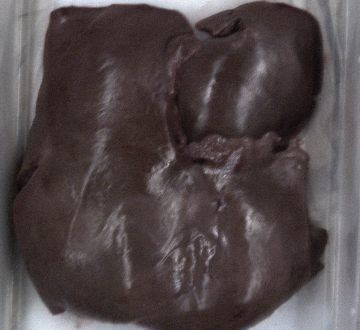}}; 
			\spy on (-1.4, -1.4) in node [left] at (-0.1, -3.8);
			\spy[green] on (-1.6, 1.45) in node [left] at (2.4, -3.8);
		\end{tikzpicture}
	}\hspace{-1.25em}
	\tikzset{external/export next=false}
	\subfloat[Input range]{	
		\begin{tikzpicture}[spy using outlines={rectangle, blue, magnification=2.0, height=3cm, width=2.3cm, connect spies, every spy on node/.append style={thick}}] 
			\node {\pgfimage[width=0.315\linewidth]{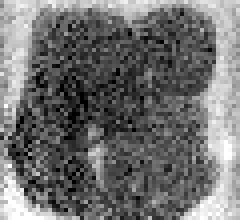}}; 
			\spy on (-1.4, -1.4) in node [left] at (-0.1, -3.8);
			\spy[green] on (-1.6, 1.45) in node [left] at (2.4, -3.8);
		\end{tikzpicture}
	}\hspace{-1.25em}
	\tikzset{external/export next=false}
	\subfloat[\gls{ssr}]{	
		\begin{tikzpicture}[spy using outlines={rectangle, blue, magnification=2.0, height=3cm, width=2.3cm, connect spies, every spy on node/.append style={thick}}] 
			\node {\pgfimage[width=0.315\linewidth]{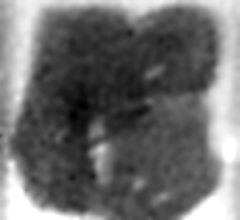}}; 
			\spy on (-1.4, -1.4) in node [left] at (-0.1, -3.8);
			\spy[green] on (-1.6, 1.45) in node [left] at (2.4, -3.8);
		\end{tikzpicture}
	}\\
	\tikzset{external/export next=false}
	\subfloat[\gls{msr}]{	
		\begin{tikzpicture}[spy using outlines={rectangle, blue, magnification=2.0, height=3cm, width=2.3cm, connect spies, every spy on node/.append style={thick}}] 
			\node {\pgfimage[width=0.315\linewidth]{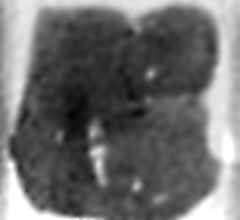}}; 
			\spy on (-1.4, -1.4) in node [left] at (-0.1, -3.8);
			\spy[green] on (-1.6, 1.45) in node [left] at (2.4, -3.8);
		\end{tikzpicture}
	}\hspace{-1.25em}
	\tikzset{external/export next=false}
	\subfloat[\gls{amsr}]{	
		\begin{tikzpicture}[spy using outlines={rectangle, blue, magnification=2.0, height=3cm, width=2.3cm, connect spies, every spy on node/.append style={thick}}] 
			\node {\pgfimage[width=0.315\linewidth]{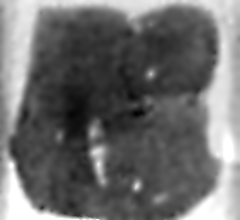}}; 
			\spy on (-1.4, -1.4) in node [left] at (-0.1, -3.8);
			\spy[green] on (-1.6, 1.45) in node [left] at (2.4, -3.8);
		\end{tikzpicture}
	}\hspace{-1.25em}
	\tikzset{external/export next=false}
	\subfloat[\gls{amsrod}]{	
		\begin{tikzpicture}[spy using outlines={rectangle, blue, magnification=2.0, height=3cm, width=2.3cm, connect spies, every spy on node/.append style={thick}}] 
			\node {\pgfimage[width=0.315\linewidth]{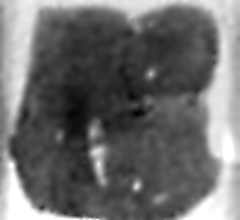}}; 
			\spy on (-1.4, -1.4) in node [left] at (-0.1, -3.8);
			\spy[green] on (-1.6, 1.45) in node [left] at (2.4, -3.8);
		\end{tikzpicture}
	}
	\caption[Ex-vivo experiments for image-guided open surgery on a porcine liver]{Super-resolution reconstruction ($K = 31$ frames, $4 \times$ magnification) for ex-vivo experiments in image-guided open surgery on a porcine liver. First and third row: high-resolution color image and low-resolution range data in comparison to super-resolved range images obtained by single-sensor super-resolution (\gls{ssr}) as well as the different multi-sensor algorithms (\gls{msr}, \gls{amsr}, and \gls{amsrod}). Second and fourth row: zoom-in for different areas on the boundary of the porcine liver. Notice that multi-sensor super-resolution considerably enhanced the accuracy of the reconstructed liver surface.}
	\label{fig:camCubeExampleResults}
\end{figure}
\tikzset{external/export next=false}

\begin{table}[t]
  \centering
	\small
  \caption[No-reference quality measures in the ex-vivo study of image-guided open surgery]{Mean $\pm$ standard deviation of the no-reference quality measures $Q_{\text{snr}}$ and $Q_{\text{edge}}$ in the ex-vivo study of image-guided open surgery. Both measures were determined using manually selected regions of interest in low-resolution and super-resolved range data gained by the different reconstruction algorithms. In total, four datasets with six manually selected regions per dataset were used.}
	\begin{tabular}{p{4.0em} ccccc}
		\toprule
		\textbf{Measure} & \textbf{Low-res. data} & \multicolumn{4}{c}{\textbf{Super-resolved data}} \\[0.6ex]
		\cline{3-6} \\
		 &  & \gls{ssr} & \gls{msr} & \gls{amsr} & \gls{amsrod} \\
		\midrule
		$Q_{\text{snr}}$	& 17.8 $\pm$ 0.9 & 21.7 $\pm$ 1.5 & 22.0 $\pm$ 1.4 	& 22.1 $\pm$ 1.4 	& 22.1 $\pm$ 1.4  \\
		$Q_{\text{edge}}$	& 3.9 $\pm$ 0.9 & 5.0 $\pm$ 1.0 & 6.6 $\pm$ 1.3 	& 6.7 $\pm$ 1.4 	& 6.7 $\pm$ 1.4 	\\
		\bottomrule
	\end{tabular}
  \label{tab:blindMeasuresOpenSurgery}
\end{table}

\paragraph{Range Data Quality Assessment.}
Super-resolved and low-resolution range data was quantitatively assessed in six image regions per dataset. The statistics of $Q_{\text{snr}}$ and $Q_{\text{edge}}$ evaluated on four datasets are summarized in \tref{tab:blindMeasuresOpenSurgery}. In comparison to raw range data, combining the multi-sensor techniques (\gls{amsrod}) leads to an increase of $Q_{\text{snr}}$ and $Q_{\text{edge}}$ by $24$\,\% and $72$\,\%, respectively. The multi-sensor reconstruction algorithms also outperformed the single-sensor approach (\gls{ssr}) regarding the reconstruction of depth discontinuities. It is worth noting that this is essential for the further usage of range information for image guidance, \eg in augmented reality \cite{Mersmann2011a, Kilgus2015}. 

\section{Conclusion}
\label{sec:08_Conclusion}

\begin{figure}[!t]
	\centering
	\subfloat{\includegraphics[width=0.454\linewidth, height=0.314\linewidth]{images/chapter8/endo_exampleMesh}}~
	\subfloat{\includegraphics[width=0.454\linewidth, height=0.314\linewidth]{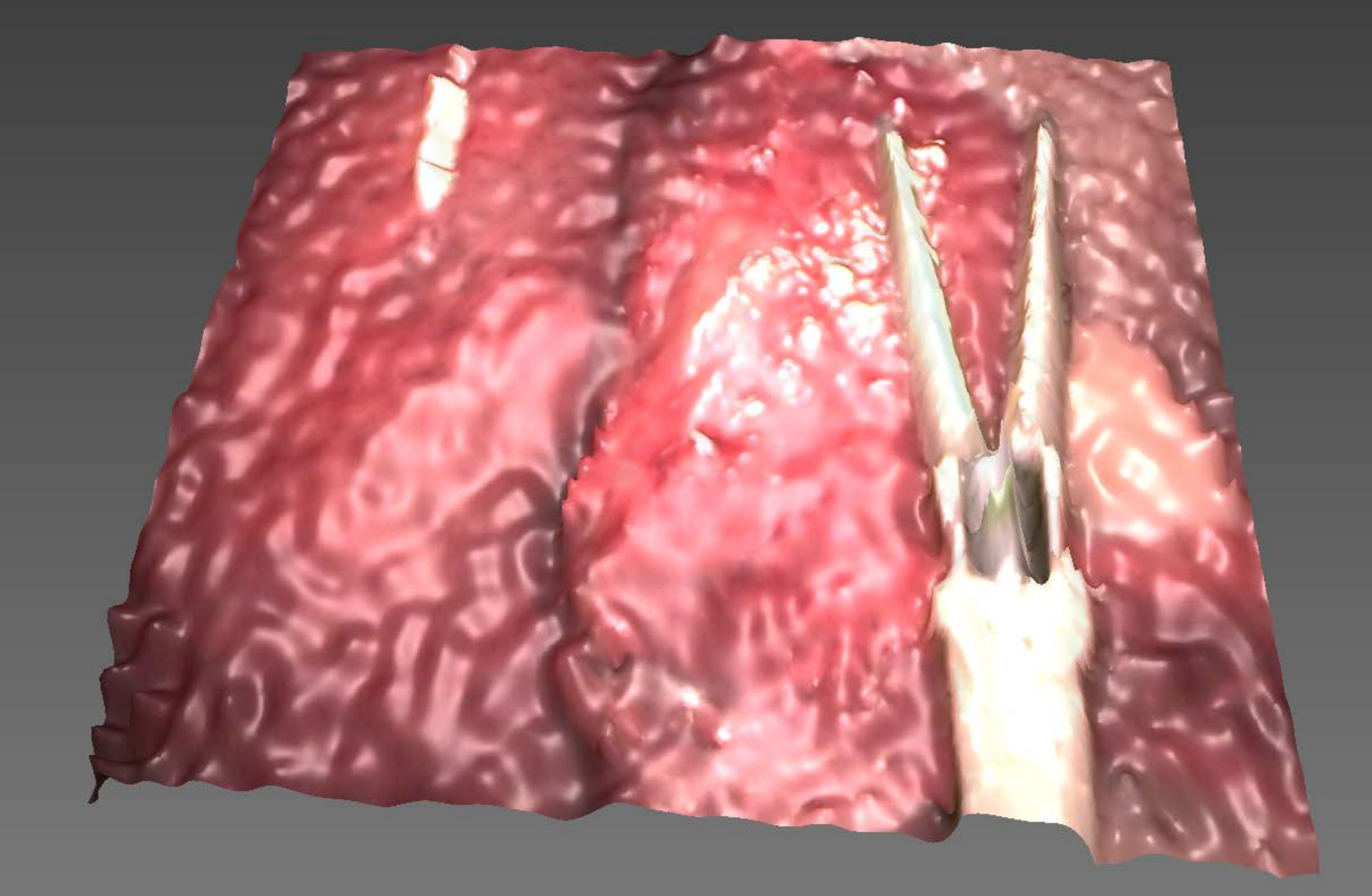}}\\
	\setcounter{subfigure}{0}
	\subfloat[Original]{\includegraphics[width=0.454\linewidth, height=0.314\linewidth]{images/chapter8/camCube_exampleMesh}\label{fig:meshResults:lr}}~
	\subfloat[Super-resolution ($4 \times$ magnification)]{\includegraphics[width=0.454\linewidth, height=0.314\linewidth]{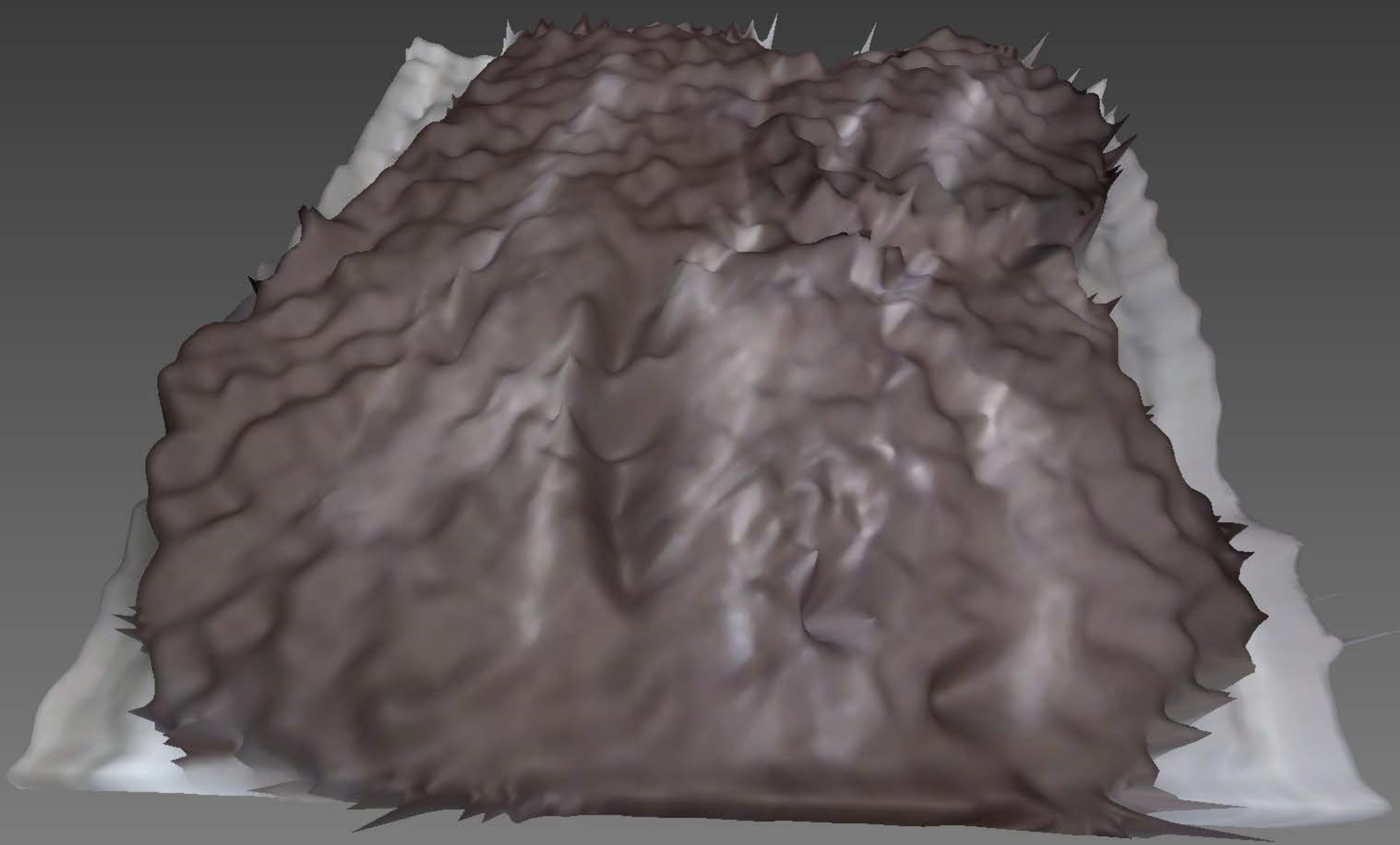}\label{fig:meshResults:sr}}
	\caption[Fusion of color and super-resolved range information]{Fusion of color and range information depicted as a textured 3-D mesh visualization in hybrid 3-D endoscopy (top row) and image guidance for open surgery (bottom row). \protect\subref{fig:meshResults:lr} 3-D mesh based on low-resolution range data acquired with \gls{tof} sensors. \protect\subref{fig:meshResults:sr} 3-D mesh based on super-resolved range data obtained by multi-sensor super-resolution ($4 \times$ magnification). Figure reused from \cite{Kohler2015a} with the publisher's permission.}
	\label{fig:meshResults}
\end{figure}

This chapter investigated super-resolution in image-guided surgery based on hybrid range imaging. One of the fundamental limitations of this technology towards clinical applications is the low spatial resolution of current range sensors. In order to alleviate this issue, multi-sensor super-resolution was adopted to gain high-resolution range images from low-resolution ones using color images as guidance. To this end, domain-specific system calibration schemes to enable sensor data fusion among range and color images were introduced. 

Two application areas were investigated in a simulation study as well as ex-vivo experiments on porcine organs: 1) hybrid 3-D endoscopy for minimally invasive surgery, and 2) image guidance in open surgery. In both areas, super-resolution enhanced the reliability of surface information and enabled the reconstruction of anatomical structures or artificial objects like surgical instruments that were barley detectable in low-resolution measurements. In ex-vivo experiments using \gls{tof} sensors, the proposed techniques improved the reliability of surface and depth discontinuity measurements compared to raw range data by more than $24$\,\% and $68$\,\%, respectively. Super-resolved range data can be augmented with high-resolution color images for a comprehensive representation of the measured scene, see \fref{fig:meshResults}. This can be considered as a key requirement regarding applications for computer-assisted interventions, \eg tracking \cite{Haase2013b} and segmentation \cite{Haase2013c} of surgical instruments as well as augmented reality \cite{Mersmann2011a} to name a few. 

Different to related filter-based preprocessing techniques proposed for image-guided surgery \cite{Wasza2011a}, super-resolution appropriately models physical effects of image formation like motion among successive frames or the underlying camera \gls{psf}. The general-purpose model used in this work can be extended even further by domain-specific effects like specular highlights or intensity related uncertainty of range measurements \cite{Reynolds2011}. However, super-resolution comes to a higher computational effort, which means a practical limitation for specific workflows with real-time constraints. Therefore, future work needs to consider an efficient implementation to enable real-time processing. This might be achieved by the use of efficient motion estimation methods \cite{Plyer2014} or by a massively parallel implementation of the reconstruction algorithm \cite{Wetzl2013}.


\part{Summary and Outlook}

\glsresetall

\chapter{Summary}
\label{sec:09_Summary}

\noindent
Multi-frame super-resolution is a software-based approach to overcome physical limitations regarding the spatial resolution of digital sensor technologies. This pursues the objective of enabling high-resolution imagery based on cost-effective systems retrospectively. To this end, we investigated novel computational methods and their applications with emphasis on medical imaging workflows. This thesis covers both, classical super-resolution applied to image data of a single modality as well as multi-sensor super-resolution in hybrid imaging.

\cref{sec:MultiFrameSuperResolutionAndTheSamplingTheorem} of this work concerned a theoretical study of multi-frame super-resolution using the Fourier transform. More specifically, we described super-resolution from a signal processing point of view as a linear inverse problem in multi-channel sampling. This concept was utilized to show the relationship to the Nyquist-Shannon sampling theorem and to discuss the meaning of the magnification factor. Eventually, we proved and discussed conditions to achieve uniqueness of super-resolution reconstruction. 

In the sequel, the main research findings of this thesis were divided into three parts. 

\paragraph{Numerical Methods for Multi-Frame Super-Resolution.}
\pref{sec:NumericalMethodsForMultiFrameSuperResolution} concerned the development of super-resolution methods for a single imaging modality. We focused on multi-frame algorithms that aim at reconstructing single high-resolution images from sequences of low-resolution frames by exploiting subpixel motion across the input images.

\cref{sec:ComputationalFrameworkForMultiFrameSuperResolution} introduced the computational framework utilized throughout this work along with a review on current super-resolution paradigms. We introduced a mathematical model to describe the physics of digital imaging. This image formation model was discretized to make it accessible for the development of numerical algorithms from a Bayesian estimation perspective. More specifically, as two of the most fundamental approaches, we presented \acrfull{ml} and \acrfull{map} estimation to gain point estimates of a latent high-resolution image from noisy, low-resolution observations. Numerous algorithms, building upon the Bayesian paradigm, that have been developed over the past years are prone to fail under practical conditions. Most of these methods are particularly sensitive regarding model parameter uncertainties like inaccurate subpixel motion estimation. 

\cref{sec:RobustMultiFrameSuperResolutionWithSparseRegularization} proposed a novel algorithm to meet the requirements regarding robustness in real-world applications. To this end, we introduced a weighted Bayesian observation model to consider outliers in the reconstruction algorithm. Furthermore, we introduced a weighted prior distribution that encourages sparsity to model the statistical appearance of natural images. Super-resolution was implemented as iteratively re-weighted energy minimization to simultaneously estimate high-resolution images and latent model confidence weights. We showed the relationship of this iteration scheme to \acrfull{mm} algorithms and rigorously proved its convergence. In comparative experimental evaluations with focus on challenging real-world conditions like space variant noise, inaccurate motion estimation, or photometric variations, the proposed method outperformed the state-of-the-art. For instance, in a benchmark with inaccurate motion estimation, iteratively re-weighted minimization improved the \acrfull{psnr} by $0.7$\,\gls{db} and the \acrfull{ssim} by $0.04$ compared to related robust algorithms. The optimization algorithm also relies on a minimal amount of manual parameter tuning making it attractive for real applications. It was also further customized in the super-resolution algorithms developed in the remainder of this thesis.

\paragraph{Multi-Sensor Super-Resolution for Hybrid Imaging.}
\pref{sec:MultiSensorSuperResolutionForHybridImaging} concerned super-resolution for multiple modalities. In this area, referred to as hybrid imaging, we studied two complementary problem statements.

\cref{sec:MultiSensorSuperResolutionUsingGuidanceImages} introduced multi-sensor super-resolution for a single modality under the guidance of a second one. We studied the case that both modalities are co-registered but complementary regarding their spatial resolutions. Accordingly, we proposed a guidance image driven framework comprising three key components: First, filter-based motion estimation is used to obtain displacement fields from optical flow on high-resolution guidance data to avoid error-prone motion estimation on low-resolution frames. Second, feature-based adaptive regularization is used to exploit correlations in terms of discontinuities between low-resolution and guidance data. Third, outlier detection using iteratively re-weighted minimization driven by image similarity assessment on the guidance data is employed. These techniques were validated for hybrid 3-D range imaging, where high-quality color images steer super-resolution of range data. Overall, the multi-sensor methodology led to gains of $0.9$\,\gls{db} in terms of \gls{psnr} and $0.02$ in terms of \gls{ssim} over a straightforward application of super-resolution solely on the range data. 

\cref{sec:SuperResolutionForMultiChannelImages} generalized multi-sensor super-resolution to jointly super-resolve a set of modalities. To this end, we dropped the usage of guidance data to facilitate a wider range of hybrid imaging setups. This methodology builds on multi-channel images as the underlying mathematical concept. Its key notion is the consideration of mutual dependencies between image channels in a Bayesian model. Different to feature-based regularization, mutual dependencies are captured by a novel \acrfull{llr} prior. This model was used to develop an alternating minimization scheme building upon the robust algorithm presented in \cref{sec:RobustMultiFrameSuperResolutionWithSparseRegularization}. It is applicable in color-, multispectral-, and range imaging as well as further applications beyond classical multi-frame super-resolution like joint segmentation and resolution enhancement. As the primary insight, multi-channel reconstructions outperformed sequential channel-wise reconstructions that essentially ignore inter-channel dependencies. In color imaging as a classical use case, the proposed method led to a gain of $1.5$\,\gls{db} in terms of \gls{psnr} and $0.04$ in terms of \gls{ssim} compared to channel-wise super-resolution of color images.   

\paragraph{Super-Resolution in Medical Imaging.}
\pref{sec:SuperResolutionInMedicalImaging} addressed applications in medical imaging with focus on diagnostic and interventional use cases. The methods investigated in this part pursue the common goal of overcoming the resolution limitations of recently developed imaging technologies as important step towards their clinical use.

\cref{sec:RetinalFundusVideoImaging} presented a new framework to approach super-resolution in the area of retinal imaging. This framework targets at the reconstruction of a high-resolution retinal image from a low-resolution video acquired from the human eye background. For this purpose, we introduced an image formation model tailored to the conditions of retinal video imaging and exploited natural eye movements. Moreover, we presented a quality self-assessment scheme to estimate a high-resolution image driven by an objective no-reference measure of image noise and sharpness. This method was evaluated for low-cost retinal imaging on real video data of healthy subjects and glaucoma patients. In this study, it led to an image quality comparable to those of commercially available, but expensive and stationary cameras. Furthermore, super-resolution was able to enhance common image analysis tasks like automatic blood vessel segmentation, where it increased the sensitivity by $10$\,\% compared to a direct segmentation on low-resolution images. The proposed method can serve as a valuable tool for high-resolution imagery in clinical workflows with high demands on cost-efficiency and mobility, \eg screening applications.

\cref{sec:HybridRangeImagingForImageGuidedSurgery} examined super-resolution to aid hybrid range imaging for image-guided surgery. This concerns an adoption of the multi-sensor framework introduced in \cref{sec:MultiSensorSuperResolutionUsingGuidanceImages}, where high-resolution color images steer super-resolution on low-resolution range data. To make this method usable for this particular domain, we introduced two system calibration schemes for sensor data fusion among both modalities: a beam splitter setup to measure geometric and photometric information through a single optical system as well as a stereo vision setup that combines distinct cameras. We conducted comprehensive experiments in two fields of today's surgery using these setups, namely 3-D endoscopy for minimally invasive procedures and image guidance for open surgery. In ex-vivo experiments using \acrfull{tof} sensors, multi-sensor super-resolution improved the reliability of surface and depth discontinuity measurements compared to raw range data by more than $24$\,\% and $68$\,\%, respectively. This is an essential step towards reliable geometric measurements of anatomical structures or artificial objects like surgical instruments. In combination with high-resolution photometric information, this can provide surgeons a comprehensive view of the underlying scene.

\chapter{Outlook}
\label{sec:10_Outlook}

\noindent
Apart from the theory and applications investigated in this thesis, there is a great number of opportunities for future work. Below, we summarize several promising directions for further research that are related to this work.

\paragraph{Extension of the Image Formation Model.}
Throughout this work, we limited ourselves to the algorithm design for linear image formation models that build upon several idealizing assumptions. One of the pitfalls is the assumption of isotropic and space invariant blur related to the camera \gls{psf} that is known a priori. This assumption is reasonable in case of optical blur as mainly considered in the investigated applications but might be violated under different conditions. Some typical examples include atmospheric or motion blur, where the modeling by simple isotropic kernels is inappropriate. Consequently, the underlying image formation model needs to be revised to tackle these effects. However, recent attempts to handle motion blur \cite{Ma2015} or more general space variant models \cite{Sorel2010} might provide a basis towards tackling these challenging situations.

Another crucial limitation is the assumption that raw data untouched by camera internal preprocessing is accessible by super-resolution algorithms via the camera interface. This is convincing for scientific or medical applications but might be violated by low-cost consumer cameras that employ compression codecs, which limits the efficiency of super-resolution. Thus, modeling data compression is important to break into new application areas. Related work considered this aspect by new image priors tailored to compressed video reconstruction \cite{Belekos2009}.

Despite the simplicity of the underlying model and the aforementioned limitations, the modular design of the proposed algorithms developed from a Bayesian perceptive makes them flexible regarding revisions or extensions. This enables the tailoring of these algorithms to new domains either by adapting the image formation model or by considering new effects in the design of image priors. 

\paragraph{Extension to Video Super-Resolution.}
This thesis considered the use case of reconstructing single images of enhanced spatial resolution from a set of low-resolution frames. Consequently, super-resolution buys an improved spatial resolution at the price of a decreased temporal one that is lost in the reconstruction. One interesting extension comprises \textit{video super-resolution} that targets at the simultaneous estimation of an entire high-resolution video from a low-resolution one. Although this could be approached by a successive use of image super-resolution in a temporal sliding window mode, special algorithms have already been introduced to solve this highly ill-posed problem by exploiting temporal consistencies in natural videos \cite{Zibetti2007,Dirks2016}. This can also be achieved by fast incremental algorithms \cite{Su2011} to accelerate video super-resolution. 

As most of the presented algorithms are extendable by these concepts, one promising direction for future work is their transfer from classical image to video super-resolution. In particular, the robust estimation techniques proposed in \cref{sec:RobustMultiFrameSuperResolutionWithSparseRegularization} provide a sensible basis to approach video super-resolution that involves additional issues regarding robustness. Such techniques might also contribute to new applications that would benefit from additional temporal information, \eg diagnostic medical imaging investigated in \cref{sec:RetinalFundusVideoImaging}. 

\paragraph{Extension to Learning-Based Methods.}
In contrast to this work that approaches super-resolution in an unsupervised way, \textit{learning-based} methods gained enormous interest over the past years. This class of algorithms aims at learning the mapping among low-resolution and high-resolution images from training data. This can be done via sparse signal representation and dictionary learning \cite{Yang2010}. More recently, current deep learning architectures made their entrance into single-image super-resolution including deep convolutional neural networks \cite{Dong2014,Kim2016,Kim2016a} or generative adversarial learning \cite{Ledig2016}. 

Such architectures are also extendable to the multi-frame case \cite{Liao2015,Kappeler2016}. In spite of their success as demonstrated in recent works, these methods heavily rely on the existence of large training datasets to learn the mappings among the low-resolution and high-resolution domains. Nevertheless, given a reliable training, they enable efficient resolution enhancement in contrary to algorithms that involve time-consuming numerical optimizations based on generative modeling. Hence, future research needs to consider \textit{hybrid} super-resolution schemes by combining the individual strengths of these complementary paradigms. 

\paragraph{Theoretical Considerations.}
An essential question in the research of image super-resolution is the question whether there exist fundamental limits of the investigated algorithms. More specifically, it is worthwhile to derive upper bounds regarding the spatial resolution reachable by super-resolution and thus an effective magnification factor. A basic study of this question under ideal conditions in the Fourier domain comprising noise-free sampling was presented in \cref{sec:MultiFrameSuperResolutionAndTheSamplingTheorem}. This demonstrated that the effective magnification is bounded by the band limitation of the signal that needs to be reconstructed. 

Several attempts have been made to derive fundamental limits in more general situations but based on simplifying assumptions. In \cite{Baker2002}, Baker and Kanade reported that classical reconstruction-based algorithms as studied in this thesis tend to be less profitable under an increasing magnification factor. Later, Lin and Shum \cite{Lin2004} presented quantitative statements for this fact using the perturbation theory of linear systems under translational motion and a box shaped \gls{psf}. Tanaka and Okutomi \cite{Tanaka2005} extended these studies for an arbitrary space invariant \gls{psf} by formulating the condition number theorem. However, in addition to a pure translation, the underlying assumption of an infinite number of low-resolution observations means an oversimplification. Thus, one unsolved aspect in the research community is the derivation of tighter bounds regarding the performance of super-resolution. This needs to consider real-world conditions including more general motion models and different sources of error like image noise.

\paragraph{Practical Considerations.}
The practical aspects for future work concern the usability of the presented algorithms. This mainly includes two considerations.

One aspect is the amount of parameter calibration that comes along with the usage of super-resolution. This is due to the fact that most of the presented techniques involve several tuning parameters in the underlying image formation model or the employed numerical optimization algorithms. Some examples are the \gls{psf} kernel, scale parameters, or regularization weights. In \cref{sec:RobustMultiFrameSuperResolutionWithSparseRegularization}, we introduced an optimization scheme that provides a scale and regularization parameter estimation to minimize the amount of manual parameter tuning. With a similar motivation, we proposed quality self-assessment in \cref{sec:RetinalFundusVideoImaging} to objectify the choice of a regularization weight in a particular application domain. Future work needs to extend these concepts to rigorously reduce the number of user-defined parameters following the same lines of thought as used in related field, \eg image auto-denoising \cite{Kong2013a}. This also includes research in the area of objective quality assessment \cite{Yeganeh2012} as the main building block of these techniques. Such a fully automatic parameter tuning could lead to a further improved robustness, flexibility, and user-friendliness of super-resolution algorithms.

Another aspect is the consideration of the computational complexity. While the main scope of the proposed algorithms is accurate resolution enhancement in a retrospective way, their use is computationally demanding. Non-parallelized implementations of these methods yield run times in the range of seconds up to several minutes growing linearly with the most relevant parameters, \ie the number of observed low-resolution pixels and the desired magnification. This might reduce the acceptability in time-critical environments and does not meet real-time constraints. Example applications concerned by this issue include computer-assisted interventions as addressed in \cref{sec:HybridRangeImagingForImageGuidedSurgery}. For these reasons, future work needs to study efficient implementations of the proposed algorithms. Promising opportunities towards an interactive use of super-resolution could exploit parallelizations of the reconstruction algorithm \cite{Wetzl2013} or the motion estimation \cite{Plyer2014} using modern graphics processing units. Other possibilities are hardware-based implementations, \eg using field programmable gate arrays \cite{Bowen2008}.

\appendix
\chapter{Appendix}
\label{sec:appendixA}

\section{Multi-Frame Super-Resolution and the Sampling Theorem}
\label{sec:A_MultiFrameSuperResolutionAndTheSamplingTheorem}

\subsection{Uniqueness for Ideal Sampling}
\label{sec:A_UniquenessForIdealSampling}

In this appendix, we investigate the conditions to provide unique super-resolution for ideal and real sampling in the Fourier domain based on \cite{Tsai1984,Kim1990,Tekalp1992a}. Let us first consider the case of ideal sampling. The conditions to guarantee a unique solution are summarized by \theoref{theo:02_uniqunessIdealSampling}.
\uniqunessIdealSampling*
\begin{proof}
For a unique solution of the linear problem in \eref{eqn:02_sampling_aliasingEqSystemTotal}, the matrix $\vec{W}$ needs to be non-singular. Due to the block structure of $\vec{W}$ according to \eref{eqn:02_sampling_aliasingEqSystem}, we consider the reconstruction of frequencies $X_{n,-L}, \ldots, X_{n,L-1}$ for $n = 1, \ldots, N$. This can be written as the solution of the linear equation system:
\begin{equation}
	\begin{pmatrix}
		\FrameIdx{\mathcal{Y}}{1}[n] \\ \FrameIdx{\mathcal{Y}}{2}[n] \\ \vdots \\ \FrameIdx{\mathcal{Y}}{K}[n]
	\end{pmatrix}
	=
	\underbrace{
	\begin{pmatrix}
		\FrameIdx{W_{n,-L}}{1}	& \FrameIdx{W_{n,-L+1}}{1}	& \ldots	&	\FrameIdx{W_{n,L-1}}{1} \\
		\FrameIdx{W_{n,-L}}{2} 	& \FrameIdx{W_{n,-L+1}}{2} 	& \ldots 	&	\FrameIdx{W_{n,L-1}}{2} \\
		\vdots									& \vdots										&  				&	\vdots \\
		\FrameIdx{W_{n,-L}}{K} 	& \FrameIdx{W_{n,-L+1}}{K} 	& \ldots 	&	\FrameIdx{W_{n,L-1}}{K}
	\end{pmatrix}
	}_{\vec{W}_n}
	\begin{pmatrix}
		X_{n,-L} \\ X_{n,-L+1} \\ \vdots \\ X_{n, L-1}
	\end{pmatrix},
\end{equation}
where the elements $\FrameIdx{W_{n,m}}{k}$ are calculated according to \eref{eqn:02_sampling_systemMatrix} for $k = 1, \ldots, K$ and $m = -L, \ldots, L-1$. Notice that the overall system matrix $\vec{W}$ in \eref{eqn:02_sampling_aliasingEqSystemTotal} is non-singular iff the matrices $\vec{W}_n$ are non-singular for all $n = 1, \ldots, N$. 

Based on \eref{eqn:02_sampling_systemMatrix}, we can decompose $\vec{W}_n$ for an arbitrary sample index $n$ with $n = 1, \ldots, N$ according to \cite{Kim1990}:
\begin{equation}
	\vec{W}_n = \vec{U}_n \vec{V}_n,
\end{equation}
where $\vec{U}_n$ is a complex-valued diagonal matrix that contains the non-zero elements $U_{n,kk} = \exp \left( -j 2\pi f_s t_k \left( \frac{n}{N} + L \right) \right)$ for $k = 1 \ldots, K$. $\vec{V}_n$ is given by the Vandermonde matrix:
\begin{equation}
	\vec{V}_n =
	\begin{pmatrix}
		1 & \exp\left( j 2\pi f_s t_1 \right) & \exp\left( j 2\pi f_s t_1 \right)^2 & \ldots & \exp\left( j 2\pi f_s t_1 \right)^{2L-1}\\
		1 & \exp\left( j 2\pi f_s t_2 \right) & \exp\left( j 2\pi f_s t_2 \right)^2 & \ldots & \exp\left( j 2\pi f_s t_2 \right)^{2L-1}\\
		\vdots	& \vdots	& \vdots & & \vdots \\
		1 & \exp\left( j 2\pi f_s t_K \right) & \exp\left( j 2\pi f_s t_K \right)^2 & \ldots & \exp\left( j 2\pi f_s t_K \right)^{2L-1}\\
	\end{pmatrix},
\end{equation}
where the nodes $\exp\left( j 2\pi f_s t_k \right)$ are located on the complex unit circle.

Since $\vec{U}_n$ is always non-singular, $\vec{W}_n$ is non-singular iff $\vec{V}_n$ is non-singular. That is, the Vandermonde determinant $\mathrm{det}(\vec{V}_n)$ of $\vec{V}_n$ needs to be non-zero, which is the case for distinct nodes \cite{Horn2012}. Thus, we have:
\begin{equation}
	\begin{split}
		\mathrm{det}(\vec{V}_n) 
		&= \prod_{i = 1}^K \prod_{j = i + 1}^K \Big(\exp\left( j 2\pi f_s t_i \right) - \exp\left(j 2\pi f_s t_j \right)  \Big) \neq 0 \\
		&\Leftrightarrow \exp\left( j 2\pi f_s t_i \right) - \exp\left( j 2\pi f_s t_j \right) \neq 0 \\
		&\Leftrightarrow \exp\left( j 2\pi f_s t_i \right) \neq \exp\left( j 2\pi f_s t_j \right),
	\end{split}
\end{equation}
for all $1 \leq i < j \leq K$. In order to guarantee a non-zero determinant $\mathrm{det}(\vec{V}_n)$ and thus a unique solution of \eref{eqn:02_sampling_aliasingEqSystemTotal}, it follows:
\begin{equation}
	\begin{split}
		\mathrm{det}(\vec{V}_n)  \neq 0
		\Leftrightarrow ~&~ \exp\left( j 2\pi f_s t_i \right) \neq \exp\left( j 2\pi f_s t_j \right) \\
		\Leftrightarrow ~&~ t_j \neq c_1 t_i + c_2 \frac{1}{f_s} \\
		\Leftrightarrow	~&~ t_j \neq c_1 t_i + c_2 T,
	\end{split}
\end{equation}
for all $1 \leq i < j \leq K ~\text{and}~ c_1, c_2 \in \mathbb{Z}$. This shows the desired condition and completes the proof.
\end{proof}

\subsection{Uniqueness for Real Sampling}
\label{sec:A_UniquenessForRealSampling}

Let us now prove the conditions regarding uniqueness in case of real sampling. These conditions are summarized by the following theorem.
\uniqunessRealSampling*
\begin{proof}
If we consider the reconstruction of frequencies $X_{n,-L}, \ldots, X_{n,L-1}$ for an arbitrary $n$ with $n = 1, \ldots, N$ and exploit the block structure of the matrix $\vec{H}$ in \eref{eqn:02_sampling_aliasingEqSystemTotalBlurKernel}, the corresponding linear system is given by:
\begin{equation}
	\begin{pmatrix}
		\FrameIdx{\mathcal{Y}}{1}[n] \\ \FrameIdx{\mathcal{Y}}{2}[n] \\ \vdots \\ \FrameIdx{\mathcal{Y}}{K}[n]
	\end{pmatrix}
	=
	\underbrace{
	\begin{pmatrix}
		\FrameIdx{H_{n,-L}}{1}	& \FrameIdx{H_{n,-L+1}}{1}	& \ldots	&	\FrameIdx{H_{n,L-1}}{1} \\
		\FrameIdx{H_{n,-L}}{2} 	& \FrameIdx{H_{n,-L+1}}{2} 	& \ldots 	&	\FrameIdx{H_{n,L-1}}{2} \\
		\vdots									& \vdots										&  				&	\vdots \\
		\FrameIdx{H_{n,-L}}{K} 	& \FrameIdx{H_{n,-L+1}}{K} 	& \ldots 	&	\FrameIdx{H_{n,L-1}}{K}
	\end{pmatrix}
	}_{\vec{H}_n}
	\begin{pmatrix}
		X_{n,-L} \\ X_{n,-L+1} \\ \vdots \\ X_{n, L-1}
	\end{pmatrix},
\end{equation}
where the elements $\FrameIdx{H_{n,m}}{k}$ in the quadratic matrix $\vec{H}_n$ are calculated according to the corresponding blur kernels for $k = 1, \ldots, K$ and $m = -L, \ldots, L-1$. Let us assume that the channel offsets in \eref{eqn:02_sampling_aliasingEqSystemTotalBlurKernel} are given by $t_i = 0$ for all $i = 1, \ldots, K$.  Then, we can assemble the matrix $\vec{H}_n$ according to:
\begin{equation}
	\begin{split}
	&\vec{H}_n = \\
	&\begin{pmatrix}
		\FrameIdx{H}{1}\left( \left(\frac{n}{N} + L \right) f_s \right) 
		& \FrameIdx{H}{1}\left( \left( \frac{n}{N} + (L - 1) \right) f_s \right) 
		& \ldots 
		& \FrameIdx{H}{1}\left( \left( \frac{n}{N} - (L - 1) \right) f_s \right) \\
		\FrameIdx{H}{2}\left( \left( \frac{n}{N} + L \right) f_s \right) 
		& \FrameIdx{H}{2}\left( \left( \frac{n}{N} + (L - 1) \right) f_s \right) 
		& \ldots 
		& \FrameIdx{H}{2} \left( \left( \frac{n}{N} - (L - 1) \right) f_s \right)\\
		\vdots & \vdots & & \vdots \\
		\FrameIdx{H}{K}\left( \left( \frac{n}{N} + L \right) f_s \right) 
		& \FrameIdx{H}{K}\left( \left( \frac{n}{N} + (L - 1) \right) f_s \right) 
		& \ldots 
		& \FrameIdx{H}{K}\left( \left( \frac{n}{N} - (L - 1) \right) f_s \right) \\
	\end{pmatrix}.
	\end{split}
\end{equation}
Note that $\vec{H}$ in \eref{eqn:02_sampling_aliasingEqSystemTotalBlurKernel} is non-singular iff $\vec{H}_n$ has full rank for all $n = 1, \ldots, N$. A full rank of the quadratic matrix $\vec{H}_n$ is equivalent to linearly independent rows. That is:
\begin{equation}
	\sum_{i = 1}^K c_{n,i} 
	\begin{pmatrix} 
		\FrameIdx{H}{i}\left( \left( \frac{n}{N} + L \right) f_s \right) 
		& \ldots 
		& \FrameIdx{H}{i}\left( \left( \frac{n}{N} - (L - 1) \right) f_s \right)
	\end{pmatrix}^\top 
	\neq \vec{0},
\end{equation}
for all $c_{n,i} \neq 0$. This translates into:
\begin{equation}
	\sum_{i = 1}^K c_i \FrameIdx{\vec{H}}{i} \neq \vec{0},
\end{equation}
for all $c_i \neq 0$, which is equivalent to independent blur kernels and shows the first condition in \theoref{theo:02_uniqunessRealSampling}. Moreover, we need to guarantee that all columns of $\vec{H}_n$ are non-zero. That is:
\begin{equation}
	\sum_{i = 1}^K \left| \FrameIdx{H}{i} \left(\frac{n}{N} f_s + m f_s \right) \right| \neq 0
\end{equation}
for all $m = -L, \ldots, L-1$. This shows the second condition in \theoref{theo:02_uniqunessRealSampling} and completes the proof. 

\end{proof}

\section{Robust Multi-Frame Super-Resolution with Sparse Regularization}
\label{sec:A_RobustMultiFrameSuperResolutionWithSparseRegularization}

\subsection{Relationship to Majorization-Minimization Algorithms}
\label{sec:A_RelationshipToMajorizationMinimizationAlgorithms}

In this appendix, we prove \theoref{theo:04_mmTheorem} to establish the connection between iteratively re-weighted minimization and \gls{mm} algorithms. First, let us present several important properties of the Huber loss and the mixed \LOne/\Lp{\SparsityParam} norm. The following lemma states that the Huber loss function can be written as the solution of a weighted minimization problem.
\begin{restatable}{lemma}{A_lemmaHuber}
	The Huber loss function $\HuberLoss{z}$ in \eref{eqn:04_nonConvexEnergy_huber} can be written as a weighted minimization problem:
	\begin{align}
		\label{eqn:A_lemmaHuberDecomp}
		\HuberLoss{z} &= \min_{\WeightsBFun \in \RealNonNeg} \left\{ \WeightsBFun z^2 + \NoiseStd^2 \rho(\WeightsBFun) \right\} \\
		\rho(\WeightsBFun) &=
		\begin{cases}
			\frac{1}{\WeightsBFun} - 1 
				& \mathrm{if}~ 0 \leq \WeightsBFun < 1\\
			0 
				& \mathrm{if}~ \WeightsBFun \geq 1.
		\end{cases}.
	\end{align}
	\label{lemma:A_lemmaHuber}
\end{restatable}
\begin{proof}
Obviously, $\HuberLoss{z}$ is a convex function and when $\WeightsBFun \geq 1$, it is monotonically increasing. Thus, the optimal weight is $\WeightsBFun^* =  1$ in case of $z^2 \leq \NoiseStd^2$ or $\WeightsBFun^* = \NoiseStd/|z|$ in case of $z^2 > \NoiseStd^2$, where the later comes from the first order optimality condition. Comparing the objective values we get the optimal weight:
\begin{equation}
	\label{eqn:A_huberWeights}
	\WeightsBFun^* =
	\begin{cases}
		1        
			& \mathrm{if}~~ |z| \leq \NoiseStd\\
		\frac{\NoiseStd}{|z|} 
			& \mathrm{otherwise}
	\end{cases}.
\end{equation}
Therefore, the solution of the weighted minimization in \eref{eqn:A_lemmaHuberDecomp} yields:
\begin{equation}
	\min_{\WeightsBFun \in \RealNonNeg} \left\{ \WeightsBFun z^2 + \NoiseStd^2 \rho(\WeightsBFun) \right\} =
	\begin{cases}
		z^2        
			& \mathrm{if}~~ |z| \leq \NoiseStd\\
		2 \NoiseStd |z| - \NoiseStd^2 
			& \mathrm{otherwise}
	\end{cases},
\end{equation}
which coincides with the Huber loss $\HuberLoss{z}$ in \eref{eqn:04_nonConvexEnergy_huber}.
\end{proof}

Next, let us employ the weighted minimization problem in \eqref{eqn:A_lemmaHuberDecomp} to define a majorizing function for the Huber loss. This function is provided by the following lemma.
\begin{restatable}{lemma}{A_majorizationHuber}
	The Huber loss $\HuberLoss{z}$ in \eref{eqn:04_nonConvexEnergy_huber} is majorized at $\IterationIdx{z}{t-1}$ by:
	\begin{equation}
		\HuberLossMajorize{z, \IterationIdx{z}{t-1}} =
		\begin{cases}
			z^2
				& \text{if}~|\IterationIdx{z}{t-1}| \leq \NoiseStd \\
			\frac{\NoiseStd}{|\IterationIdx{z}{t-1}|} z^2 + \NoiseStd^2 \left( \frac{|\IterationIdx{z}{t-1}|}{\NoiseStd} - 1 \right)
				& \text{otherwise}
		\end{cases}.
	\end{equation}
	\label{lemma:A_majorizationHuber}
\end{restatable}
\begin{proof}
	For this proof, let us first assume that $|\IterationIdx{z}{t-1}| \leq \NoiseStd$. Then, $\HuberLossMajorize{z, \IterationIdx{z}{t-1}}$ coincides with $\HuberLoss{z}$ according to the definitions of both functions.
	
	If $|\IterationIdx{z}{t-1}| > \NoiseStd$ and $\beta = \NoiseStd / |\IterationIdx{z}{t-1}| \in [0, 1]$, the function $\HuberLossMajorize{z, \IterationIdx{z}{t-1}}$ can be reformulated according to \lemmaref{lemma:A_lemmaHuber}:
	\begin{equation}
		\begin{split}
			\HuberLossMajorize{z, \IterationIdx{z}{t-1}}
			&= \WeightsBFun z^2 + \NoiseStd^2 \rho(\WeightsBFun) \\ 
			&= \frac{\NoiseStd}{|\IterationIdx{z}{t-1}|} z^2 + \NoiseStd \left| \IterationIdx{z}{t-1} \right| - \NoiseStd^2 \\
			&\geq \HuberLoss{z},
		\end{split}
	\end{equation}
	where the equality holds true for $z = \IterationIdx{z}{t-1}$. Thus, $\HuberLossMajorize{z, \IterationIdx{z}{t-1}}$ is a majorizing function for $\HuberLoss{z}$ at $\IterationIdx{z}{t-1}$.
\end{proof}

Next, we define a majorizing function for the mixed \LOne/\Lp{\SparsityParam} norm. This function is established by the following lemma.
\begin{restatable}{lemma}{A_majorizationL1Lp}
	The mixed \LOne/\Lp{\SparsityParam} norm $\LossFun[\SparsityParam]{z}$ in \eref{eqn:04_nonConvexEnergy_l1lp} for $\SparsityParam \in [0, 1]$ is majorized at $\IterationIdx{z}{t-1}$ by:
	\begin{equation}
		\label{eqn:A_majorizationL1Lp}
		\LossFunMajorize[\SparsityParam]{z, \IterationIdx{z}{t-1}} =
		\begin{cases}
			|z|
				& \text{if}~ |\IterationIdx{z}{t-1}| \leq \PriorStd \\
			\SparsityParam \left( \frac{\PriorStd}{|\IterationIdx{z}{t-1}|} \right)^{1-\SparsityParam} |z| + (1-\SparsityParam) \PriorStd^{1-\SparsityParam} |\IterationIdx{z}{t-1}|^\SparsityParam
				& \text{otherwise}
		\end{cases}.
	\end{equation}
	\label{lemma:A_majorizationL1Lp}
\end{restatable}
\begin{proof}
Let us first consider the case $|\IterationIdx{z}{t-1}| \leq \PriorStd$. Then, \smash{$\LossFunMajorize[\SparsityParam]{\IterationIdx{z}{t-1}, \IterationIdx{z}{t-1}}$} coincides with \smash{$\LossFun[\SparsityParam]{\IterationIdx{z}{t-1}}$} according to the definitions of both functions.

Let us now consider the case $|\IterationIdx{z}{t-1}| > \PriorStd$. Since $\LossFun[\SparsityParam]{z}$ is monotone and concave downwards, the Taylor series expansion for $|z| > \PriorStd$ yields the inequality:
\begin{equation}
	\begin{split}
		\LossFun[\SparsityParam]{z} 
		&\leq \LossFun[\SparsityParam]{\IterationIdx{z}{t-1}} + \left( z - \IterationIdx{z}{t-1} \right) 
		\cdot \frac{d}{dz} \LossFun[\SparsityParam]{z} \Big|_{z = \IterationIdx{z}{t-1}} \\
		&= \PriorStd^{1-\SparsityParam} \left|\IterationIdx{z}{t-1}\right|^\SparsityParam 
		+ \left( z - \IterationIdx{z}{t-1} \right) 
		\cdot \Sign{\IterationIdx{z}{t-1}} \PriorStd^{1-\SparsityParam} \SparsityParam \left| \IterationIdx{z}{t-1} \right|^{\SparsityParam-1} \\
		&= \SparsityParam \left( \frac{\PriorStd}{|\IterationIdx{z}{t-1}|} \right)^{1-\SparsityParam} |z| + (1 - \SparsityParam) \PriorStd^{1-\SparsityParam} \left| \IterationIdx{z}{t-1} \right|^\SparsityParam \\
		&\leq \LossFunMajorize[\SparsityParam]{z, \IterationIdx{z}{t-1}},
	\end{split}
\end{equation}
and $\LossFun[\SparsityParam]{\IterationIdx{z}{t-1}} = \LossFunMajorize[\SparsityParam]{\IterationIdx{z}{t-1}, \IterationIdx{z}{t-1}}$ if $z = \IterationIdx{z}{t-1}$. Hence, $\LossFunMajorize[\SparsityParam]{z, \IterationIdx{z}{t-1}}$ is a majorizing function for $\LossFun[\SparsityParam]{z}$ at $\IterationIdx{z}{t-1}$. 
\end{proof}

Notice that the majorizing functions \smash{$\HuberLossMajorize{z, \IterationIdx{z}{t-1}}$} and \smash{$\LossFunMajorize[\SparsityParam]{z, \IterationIdx{z}{t-1}}$} are non-negative and provide upper bounds for \smash{$\HuberLoss{z}$} and \smash{$\LossFunMajorize[\SparsityParam]{z}$}, respectively. Hence, minimization can be performed by an \gls{mm} algorithm \cite{Hunter2004}. Based on these properties, we can establish \theoref{theo:04_mmTheorem}.
\mmTheorem*
\begin{proof}
According to \lemmaref{lemma:A_majorizationHuber} and since the majorization relation is closed under the formation of a sum of non-negative terms \cite{Hunter2004}, it follows: 
\begin{equation}
	\begin{split}
		\DataTermMajorize{\HR, \IterationIdx{\HR}{t-1}} 
			&= \left(\LR - \SystemMat \HR \right)^\top \IterationIdx{\WeightsBMat}{t} \left( \LR - \SystemMat \HR \right)
				+ \sum_{i = 1}^{\NumFrames\LRSize} \rho \left( \VecEl{\LR - \SystemMat \IterationIdx{\HR}{t-1}}{i} \right) \\
			&\geq \DataTerm{\HR},
	\end{split}
\end{equation}
with equality for $\HR = \IterationIdx{\HR}{t-1}$, where $\rho(\cdot)$ is given by \eqref{eqn:04_majorizingFunctionRho} and:
\begin{equation}
	\DataTerm{\HR} = \sum_{i=1}^{\NumFrames \LRSize} \HuberLoss{ \VecEl{\LR - \SystemMat \HR}{i} }.
\end{equation}
Hence, the confidence-aware data fidelity term $\DataTermMajorize{\HR, \IterationIdx{\HR}{t-1}}$ is a majorizing function for $\DataTerm{\HR}$. Similarly, \lemmaref{lemma:A_majorizationL1Lp} yields:
\begin{equation}
	\begin{split}
		\RegTermMajorize{\HR, \IterationIdx{\HR}{t-1}}
			&= \big| \big| \IterationIdx{\WeightsAMat}{t} \SparseTransMat \HR \big| \big|_1 
				+ \sum_{i = 0}^{\SparseTransDim} \tau \left( \VecEl{\SparseTransMat \IterationIdx{\HR}{t-1}}{i} \right)\\
			&\geq \RegTerm{\HR},
	\end{split}
\end{equation}
with equality for $\HR = \IterationIdx{\HR}{t-1}$, where $\tau(\cdot)$ is given by \eqref{eqn:04_majorizingFunctionTau} and:
\begin{equation}
	\RegTerm{\HR} = \sum_{i = 1}^{\SparseTransDim} \LossFun[\SparsityParam]{ \VecEl{\SparseTransMat \HR}{i} }.
\end{equation}
Thus, the regularization term $\RegTermMajorize{\HR, \IterationIdx{\HR}{t-1}}$ is a majorizing function for $\RegTerm{\HR}$. Then, $\EnergyFunMajorize{\HR, \IterationIdx{\HR}{t-1}} = \DataTermMajorize{\HR, \IterationIdx{\HR}{t-1}} + \RegWeight \RegTermMajorize{\HR, \IterationIdx{\HR}{t-1}}$ majorizes $\EnergyFun{\HR}$ in \eref{eqn:04_nonConvexEnergy} at $\HR = \IterationIdx{\HR}{t-1}$ for $\RegWeight \geq 0$, which completes the proof.
\end{proof}

\subsection{Convergence Analysis}
\label{sec:A_ConvergenceAnalysis}

This appendix provides the proof of \theoref{theorem:04_convergenceTheorem} according to \cite{Kohler2015c} to analyze the convergence of iteratively re-weighted minimization. Let us first present one important property of the \LOne/\Lp{\SparsityParam} norm regularization term. For the sake of notational brevity, this regularization is reformulated according to:
\begin{equation}
	\RegTerm{\vec{z}} = \sum_{i \not\in \cI(\vec{z})} |z_i| + \sum_{i \in \cI(\vec{z})} \PriorStd^{1-\SparsityParam} |z_i|^\SparsityParam, 
\end{equation}
where the index set $\cI(\vec{z})$ is defined as $\cI(\vec{z}) = \{ i: z_i > \PriorStd \}$ based on the scale parameter $\PriorStd$. Without loss of generality, we consider in the following analysis the case $\PriorStd = 1$. Note that if $\PriorStd \neq 1$, one can use a normalization of $\vec{z}$ to satisfy this condition. In this situation, for the \LOne/\Lp{\SparsityParam} norm regularization term with the index set $\cI(\vec{z})$, the following inequality is fulfilled:
\begin{restatable}{lemma}{A_indexSetInequality}
	For all index sets $\cI(\vec{z}) = \{i: z_i > 1\}$ and $\cI'$ with the sparsity parameter $\SparsityParam$, where $\SparsityParam \in [0, 1]$, there is:
	\begin{equation}
		\label{eqn:A_indexSetInequality}
		\sum_{i \not \in \cI(\vec{z})} |z_i| + \sum_{i \in \cI(\vec{z})} |z_i|^\SparsityParam \leq \sum_{i \not \in \cI'} |z_i| + \sum_{i \in \cI'} |z_i|^\SparsityParam.
	\end{equation}
	\label{lemma:A_indexSetInequality}
\end{restatable}
\begin{proof}
Subtracting the left hand side in \eref{eqn:A_indexSetInequality} by the right hand side, we have:
\begin{equation}
	\begin{split}
		& \sum_{i \not \in \cI(\vec{z})} |z_i| - \sum_{i \not \in \cI'} |z_i| 
		+ \sum_{i \in \cI(\vec{z})} |z_i|^\SparsityParam  - \sum_{i \in \cI'} |z_i|^\SparsityParam \\
		&\qquad = \sum_{i \in \cI'\backslash \cI(\vec{z})} |z_i| - \sum_{i \in \cI(\vec{z}) \backslash \cI'} |z_i| 
		+ \sum_{i \in \cI(\vec{z}) \backslash \cI' } |z_i|^\SparsityParam - \sum_{i \in \cI'\backslash \cI(\vec{z})} |z_i|^\SparsityParam \\
		&\qquad 
		= \sum_{i \in \cI' \backslash \cI(\vec{z})} \left( |z_i| - |z_i|^\SparsityParam \right) 
		- \sum_{i \in \cI(\vec{z}) \backslash \cI' }\left( |z_i| - |z_i|^\SparsityParam \right).
	\end{split}
\end{equation}
Notice that for $\SparsityParam \in [0, 1]$ it follows that $|z_i| \geq |z_i|^\SparsityParam$ if and only if $z_i \geq 1$, \ie, $i \in \cI(\vec{z})$. Thus, $|z_i| - |z_i|^\SparsityParam < 0, \forall i \in \cI'\backslash \cI(\vec{z})$ and $|z_i| - |z_i|^\SparsityParam \geq 0, \forall i \in \cI(\vec{z})\backslash \cI'$. From these inequalities, it follows that:
\[
	\sum_{i \in \cI'\backslash \cI(\vec{z})} \left( |z_i| - |z_i|^\SparsityParam \right) - \sum_{i \in \cI(\vec{z})\backslash \cI' }\left( |z_i| - |z_i|^\SparsityParam \right) \leq 0.
\]
Hence, the inequality in \eref{eqn:A_indexSetInequality} is true for all index sets $\cI(\vec{z})$ and $\cI'$.
\end{proof}

Now, we can establish \theoref{theorem:04_convergenceTheorem}, which shows the convergence of iteratively re-weighted minimization in terms of the energy function value $\EnergyFun{\HR}$.
\convergenceTheorem*
\begin{proof}
For the sake of notational brevity, we present this proof by assuming the identity for the sparsifying transform, \ie $\SparseTransMat = \Id$. Note that in the general case where $\SparseTransMat \neq \Id$, we can include the transform $\SparseTransMat$ to the optimization problem by re-formulation as a constrained problem. 

According to \lemmaref{lemma:A_lemmaHuber}, we reformulate the energy function $F(\IterationIdx{\HR}{t})$ by writing the Huber loss as a weighted minimization problem:
\begin{equation}
	\begin{split}
		F(\IterationIdx{\HR}{t})
		& = \RegWeight R(\IterationIdx{\HR}{t})
		+ \sum_{i=1}^{KM} \min_{\WeightsBFun \in \RealNonNeg} 
		\left\{ \WeightsBFun \VecEl{\SystemMat \IterationIdx{\HR}{t} - \LR}{i} ^2 + \NoiseStd^2 \rho(\WeightsBFun) \right\} \\
		& \leq \RegWeight R(\IterationIdx{\HR}{t}) 
		+ \sum_{i=1}^{KM} 
		\left\{ \IterationIdx{\WeightsBFun_i}{t} \VecEl{\SystemMat \IterationIdx{\HR}{t} - \LR}{i} ^2 
		+ \NoiseStd^2 \rho(\IterationIdx{\WeightsBFun_i}{t}) \right\},
	\end{split}
\end{equation}
where the weights $\IterationIdx{\WeightsBFun_i}{t}$ for $i = 1, \ldots, KM$ are computed from $\IterationIdx{\HR}{t-1}$ according to \eref{eqn:04_localDataFidelityWeights} and $\NoiseStd$ is the scale parameter of the observation model that is assumed to be constant over the iterations. Comparing the weights $\IterationIdx{\WeightsBFun_i}{t}$ given by \eref{eqn:04_localDataFidelityWeights} and \eref{eqn:A_huberWeights}, one can verify that:
\begin{equation}
	F(\IterationIdx{\HR}{t-1}) = 
	\RegWeight R(\IterationIdx{\HR}{t-1}) 
	+ \sum_{i=1}^{KM} 
	\left\{ \IterationIdx{\WeightsBFun_i}{t} \VecEl{\SystemMat \IterationIdx{\HR}{t-1} - \LR}{i} ^2 
	+ \NoiseStd^2 \rho(\IterationIdx{\WeightsBFun_i}{t}) \right\}.
\end{equation}
Hence, we can derive the inequality condition for the energy function value among successive iterations:
\begin{equation}
	\begin{split}
		&F(\IterationIdx{\HR}{t-1}) - F(\IterationIdx{\HR}{t}) \\
		&\quad \geq \big( \SystemMat \IterationIdx{\HR}{t-1} - \LR \big)^\top \IterationIdx{\WeightsBMat}{t} \big( \SystemMat \IterationIdx{\HR}{t-1} - \LR \big) 
		- \big (\SystemMat \IterationIdx{\HR}{t} - \LR \big)^\top \IterationIdx{\WeightsBMat}{t} \big( \SystemMat \IterationIdx{\HR}{t} - \LR \big) \\
		&\quad\quad + \RegWeight 
		\bigg( 
			\sum_{i \not\in \cI(\IterationIdx{\HR}{t-1})} \big| \IterationIdx{\HRSym_i}{t-1} \big| 
			+ \sum_{i \in \cI(\IterationIdx{\HR}{t-1})} \big| \IterationIdx{\HRSym_i}{t-1} \big|^\SparsityParam
			- \sum_{i \not\in \cI(\IterationIdx{\HR}{t})} \big| \IterationIdx{\HRSym_i}{t} \big|
			- \sum_{i \in \cI(\IterationIdx{\HR}{t})} \big| \IterationIdx{\HRSym_i}{t} \big|^\SparsityParam
		\bigg),
	\end{split}
\end{equation}
where $\IterationIdx{\WeightsBMat}{t}$ is constructed as $\IterationIdx{\WeightsBMat}{t} = \Diag{\IterationIdx{\WeightsBFun_1}{t}, \ldots, \IterationIdx{\WeightsBFun_{KM}}{t}}$. This inequality condition can be rearranged according to:
\begin{equation}
	\begin{split}
		& F(\IterationIdx{\HR}{t-1}) - F(\IterationIdx{\HR}{t}) \\
		& \quad \geq \big( \IterationIdx{\HR}{t-1} - \IterationIdx{\HR}{t} \big)^\top \SystemMat^\top \IterationIdx{\WeightsBMat}{t} \SystemMat \big( \IterationIdx{\HR}{t-1} - \IterationIdx{\HR}{t} \big)
		+ 2 \big( \SystemMat \IterationIdx{\HR}{t-1} - \SystemMat \IterationIdx{\HR}{t} \big)^\top \IterationIdx{\WeightsBMat}{t} \big( \SystemMat \IterationIdx{\HR}{t} - \LR \big) \\
		&\quad\quad + \RegWeight 
		\bigg( 
			\sum_{i \not\in \cI(\IterationIdx{\HR}{t-1})} \big| \IterationIdx{\HRSym_i}{t-1} \big| 
			+ \sum_{i \in \cI(\IterationIdx{\HR}{t-1})} \big| \IterationIdx{\HRSym_i}{t-1} \big|^\SparsityParam
			- \sum_{i \not\in \cI(\IterationIdx{\HR}{t})} \big| \IterationIdx{\HRSym_i}{t} \big|
			- \sum_{i \in \cI(\IterationIdx{\HR}{t})} \big| \IterationIdx{\HRSym_i}{t} \big|^\SparsityParam
		\bigg) \\
		&\quad\geq \IterationIdx{\underline{\WeightsBFun}}{t} 
		\left \|\SystemMat \IterationIdx{\HR}{t-1} - \SystemMat \IterationIdx{\HR}{t} \right \|_2 ^2 
		+ 2 \big( \SystemMat \IterationIdx{\HR}{t-1} - \SystemMat \IterationIdx{\HR}{t} \big)^\top \IterationIdx{\WeightsBMat}{t} \big(\SystemMat \IterationIdx{\HR}{t} - \LR \big) \\
		&\quad\quad + \RegWeight 
		\bigg( 
			\sum_{i \not\in \cI(\IterationIdx{\HR}{t-1})} \big| \IterationIdx{\HRSym_i}{t-1} \big| 
			+ \sum_{i \in \cI(\IterationIdx{\HR}{t-1})} \big| \IterationIdx{\HRSym_i}{t-1} \big|^\SparsityParam
			- \sum_{i \not\in \cI(\IterationIdx{\HR}{t-1})} \big| \IterationIdx{\HRSym_i}{t} \big|
			- \sum_{i \in \cI(\IterationIdx{\HR}{t-1})} \big| \IterationIdx{\HRSym_i}{t} \big|^\SparsityParam
		\bigg),
	\end{split}
\end{equation}
where the last inequality is based on \lemmaref{lemma:A_indexSetInequality} and $\IterationIdx{\underline{\WeightsBFun}}{t} = \min_i \IterationIdx{\underline{\WeightsBFun_i}}{t}$. Then, the weight \smash{$\underline{\WeightsBFun} =  \min_t \IterationIdx{\underline{\WeightsBFun}}{t}$} is strictly positive and:
\begin{equation}
	\label{ineqaulity-2}
	\begin{split}
		& F(\IterationIdx{\HR}{t-1}) - F(\IterationIdx{\HR}{t}) \\
		& \quad \geq \underline{\WeightsBFun} \left \|\SystemMat \IterationIdx{\HR}{t-1} - \SystemMat \IterationIdx{\HR}{t} \right \|_2^2 
		+ 2 \big( \SystemMat \IterationIdx{\HR}{t-1} - \SystemMat \IterationIdx{\HR}{t} \big)^\top \IterationIdx{\WeightsBMat}{t} \big(\SystemMat \IterationIdx{\HR}{t} - \LR \big) \\
		& \quad\quad + \RegWeight 
		\bigg( 
			\sum_{i \not\in \cI(\IterationIdx{\HR}{t-1})} \big| \IterationIdx{\HRSym_i}{t-1} \big| 
			+ \sum_{i \in \cI(\IterationIdx{\HR}{t-1})} \big| \IterationIdx{\HRSym_i}{t-1} \big|^\SparsityParam
			- \sum_{i \not\in \cI(\IterationIdx{\HR}{t-1})} \big| \IterationIdx{\HRSym_i}{t} \big|
			- \sum_{i \in \cI(\IterationIdx{\HR}{t-1})} \big| \IterationIdx{\HRSym_i}{t} \big|^\SparsityParam
		\bigg).
	\end{split}
\end{equation}
Since $\IterationIdx{\HR}{t}$ is the solution of iteratively re-weighted minimization at iteration $t$, it follows:
\begin{equation}
	\Zeros \in \frac{\partial}{\partial \HR} 
	\left\{ 
		\left( \SystemMat \HR - \LR \right)^\top \IterationIdx{\WeightsBMat}{t} \left( \SystemMat \HR - \LR \right) 
		+ \RegWeight \sum_{i=1}^{\SparseTransDim} \IterationIdx{\alpha_i}{t} |\HRSym_i| 
	\right\}
	\Bigg|_{\HR = \IterationIdx{\HR}{t}},
\end{equation}
where $\IterationIdx{\alpha_i}{t}$ is computed from the weighting function in \eref{eqn:04_regularizationWeights}. Thus, it follows for the subgradient:
\begin{equation}
	\label{subgradiant}
	\begin{split}
		2 \SystemMat^\top \IterationIdx{\WeightsBMat}{t} 
		\left (\SystemMat \IterationIdx{\HR}{t} - \LR \right) + \RegWeight \IterationIdx{c_i}{t} \IterationIdx{\alpha_i}{t} &= 0, \forall i \\
		\IterationIdx{c_i}{t} &\in
		\begin{cases}
			\{1\}, 
				& \text{if\,\,} \IterationIdx{\HRSym_i}{t} > 0 \\
			\left[-1,1 \right] 
				& \text{if\,\,} \IterationIdx{\HRSym_i}{t} = 0 \\
			\{-1\}, 
				&  \text{if\,\,} \IterationIdx{\HRSym_i}{t} < 0
		\end{cases}.
	\end{split}
\end{equation}
Substituting the condition in \eref{subgradiant} into \eref{ineqaulity-2} and using the fact that $\IterationIdx{c_i}{t} \IterationIdx{\HRSym_i}{t} = |\IterationIdx{\HRSym_i}{t}|$ and $|\IterationIdx{c_i}{t}| \leq 1$, leads to:
\begin{equation}
	\begin{split}
		F(\IterationIdx{\HR}{t-1}) - F(\IterationIdx{\HR}{t})
		&\geq \underline{\WeightsBFun} \left \| \SystemMat \IterationIdx{\HR}{t-1} - \SystemMat \IterationIdx{\HR}{t} \right \|_2^2 \\
		&\quad + \RegWeight \sum_{i \not \in \cI(\IterationIdx{\HR}{t-1})} \left( \big| \IterationIdx{\HRSym_i}{t-1} \big| - \big|\IterationIdx{\HRSym_i}{t} \big| 
		+ \IterationIdx{c_i}{t} \big( \IterationIdx{\HRSym_i}{t} - \IterationIdx{\HRSym_i}{t-1} \big)  \right) \\
		&\quad + \RegWeight \sum_{i \in \cI(\IterationIdx{\HR}{t-1})} \Big( \big|\IterationIdx{\HRSym_i}{t-1} \big|^\SparsityParam - \big |\IterationIdx{\HRSym_i}{t} \big|^\SparsityParam 
		+ \SparsityParam \big| \IterationIdx{\HRSym_i}{t-1} \big|^{\SparsityParam-1} \IterationIdx{c_i}{t} \big( \IterationIdx{\HRSym_i}{t} - \IterationIdx{\HRSym_i}{t-1} \big)  \Big)  \\
		&\geq \underline{\WeightsBFun} \left \|\SystemMat \IterationIdx{\HR}{t-1} - \SystemMat \IterationIdx{\HR}{t} \right \|_2^2 \\
		&\quad  + \RegWeight \sum_{i \not \in \cI(\IterationIdx{\HR}{t-1})} 
				\Big( \big |\IterationIdx{\HRSym_i}{t-1} \big| - \big| \IterationIdx{\HRSym_i}{t} \big| + \big( \big| \IterationIdx{\HRSym_i}{t} \big| - \big| \IterationIdx{\HRSym_i}{t-1} \big| \big) \Big) \\
		&\quad  + \RegWeight \sum_{i \in \cI(\IterationIdx{\HR}{t-1})} \Big( \big| \IterationIdx{\HRSym_i}{t-1} \big|^\SparsityParam - \big| \IterationIdx{\HRSym_i}{t} \big|^\SparsityParam 
		+ \SparsityParam \big| \IterationIdx{\HRSym_i}{t-1} \big|^{\SparsityParam-1} \big( \big| \IterationIdx{\HRSym_i}{t} \big| - \big| \IterationIdx{\HRSym_i}{t-1}\big| \big) \Big) \\
		&\geq \underline{\WeightsBFun} \left \|\SystemMat \IterationIdx{\HR}{t-1} - \SystemMat \IterationIdx{\HR}{t} \right \|_2^2  \\
		&\quad + \RegWeight \sum_{i \in \cI(\IterationIdx{\HR}{t-1})} \Big( \big| \IterationIdx{\HRSym_i}{t-1} \big|^{\SparsityParam-1} \Big( (1-\SparsityParam) \big| \IterationIdx{\HRSym_i}{t-1} \big| 
		+ \SparsityParam \big| \IterationIdx{\HRSym_i}{t} \big| - \big| \IterationIdx{\HRSym_i}{t-1} \big|^{1-\SparsityParam} \big| \IterationIdx{\HRSym_i}{t} \big|^\SparsityParam \Big)  \Big) \\
		&\geq \underline{\WeightsBFun}  \left \|\SystemMat \IterationIdx{\HR}{t-1} - \SystemMat \IterationIdx{\HR}{t} \right \|_2^2, 
	\end{split}
\end{equation}
where the last inequality is according to Lemma 1 in \cite{Chen2014} as corollary of Young's inequality, which completes the proof.
\end{proof}

Since $\EnergyFun{\HR}$ is a lower-bounded function, $\EnergyFun{\IterationIdx{\HR}{t}}$ converges to an extreme value. If $\IterationIdx{\HR}{t}$ also converges to an extreme value denoted by $\HR^*$, this estimate satisfies:
\begin{equation}
	\Zeros \in \frac{\partial}{\partial \HR} 
	\left\{
		\left( \SystemMat \HR - \LR \right)^\top \IterationIdx{\WeightsBMat}{t} \left( \SystemMat \HR - \LR \right)  
		+ \RegWeight \sum_{i=1}^{\SparseTransDim} \IterationIdx{\alpha_i}{t} |\HRSym_i| 
	\right\}
	\Bigg|_{\HR = \HR^*}.
\end{equation}
As a consequence, $\HR^*$ is a stationary point of the non-convex problem in \eref{eqn:04_nonConvexEnergy}.

Notice that there can be the situation $\EnergyFun{\IterationIdx{\HR}{t}} = \EnergyFun{\IterationIdx{\HR}{t-1}}$ but $\HR$ does not converge to a stationary point, \ie $\IterationIdx{\HR}{t} \neq \IterationIdx{\HR}{t-1}$. This is the case if the following conditions are fulfilled:
\begin{enumerate}
	\item $\IterationIdx{\HR}{t} - \IterationIdx{\HR}{t-1}$ is in the null space of the system matrix, \ie $\SystemMat(\IterationIdx{\HR}{t-1} - \IterationIdx{\HR}{t}) = \Zeros$.
	\item The index sets among successive iterations are identical, \ie $\cI(\IterationIdx{\HR}{t}) = \cI(\IterationIdx{\HR}{t-1})$.
	\item The estimates in the index set $\cI(\IterationIdx{\HR}{t})$ among successive iterations are identical, \ie $|\IterationIdx{\HRSym_i}{t}| = |\IterationIdx{\HRSym_i}{t-1}|$ for all $i \in \cI(\IterationIdx{\HR}{t})$.
	\item The objective value of the regularization term among successive iterations is identical, \ie $\RegTerm{\IterationIdx{\HR}{t}} = \RegTerm{\IterationIdx{\HR}{t-1}}$ 
\end{enumerate}
However, in practice this situation can be avoided. For instance, if the system matrix $\SystemMat$ has full rank, the null space is $\{ \Zeros \}$, \ie $\IterationIdx{\HR}{t} = \IterationIdx{\HR}{t-1}$ if $\EnergyFun{\IterationIdx{\HR}{t}} = \EnergyFun{\IterationIdx{\HR}{t-1}}$.

\section{Multi-Sensor Super-Resolution using Locally Linear Regression}
\label{sec:A_MultiSensorSuperResolutionUsingLocallyLinearRegression}

\subsection{Majorization-Minimization for Tukey's Biweight Loss}
\label{sec:A_MajorizationMinimizationForTheRegressionCoefficients}

Let us consider the 1-D minimization problem:
\begin{equation}
	\hat{z} = \argmin_z \LossFun[\HRChannel{ij}]{z}.
\end{equation}
We can reformulate the minimization of the non-convex loss function $\LossFun[\HRChannel{ij}]{z}$ as an \gls{mm} algorithm according to \cite{Ochs2015}. This leads to the \gls{irls} scheme:
\begin{align}
	\IterationIdx{z}{t} 
		&= \argmin_{z} \kappa(\IterationIdx{z}{t-1}) z^2, \\
	\kappa(z)
		&= \frac{\frac{d}{dz} \LossFun[\HRChannel{ij}]{z}}{z},
\end{align}
where $\kappa(z)$ is the underlying weighting function. In case of Tukey's biweight loss defined in \eref{eqn:06_interChannelLossFun}, we can compute the gradient according to:
\begin{equation}
	\frac{d}{dz} \LossFun[\HRChannel{ij}]{z} =
	\begin{cases}
		z \left(1 - \frac{z^2}{\LLRStd{i}{j}^2} \right)^2 
			& \text{if}~ |z| \leq \LLRStd{i}{j}^2 \\
		0
			& \text{otherwise}
	\end{cases}.
\end{equation}
This leads to corresponding weighting function for \gls{irls}:
\begin{equation}
	\kappa(z) =
	\begin{cases}
		\left(1 - \frac{z^2}{\LLRStd{i}{j}^2} \right)^2 
			& \text{if}~ |z| \leq \LLRStd{i}{j}^2 \\
		0
			& \text{otherwise}
	\end{cases}.
\end{equation}

\subsection{Estimation of the Regression Coefficients}
\label{sec:A_EstimationOfTheRegressionCoefficients}

If we omit the iteration index for the sake of notational clarity, the regression coefficients associated with the $k$-th pixel in the image channels $\HRChannel{i}$ and $\HRChannel{j}$ are estimated according to:
\begin{equation}
	(\ChannelIdx{\tilde{C}}{ij, k}, \ChannelIdx{\tilde{b}}{ij, k}) 
	= \argmin_{\ChannelIdx{C}{ij, k}, \ChannelIdx{b}{ij, k}} \EnergyFun{ \ChannelIdx{C}{ij, k}, \ChannelIdx{b}{ij, k} },
\end{equation}
where:
\begin{equation}
	\EnergyFun{ \ChannelIdx{C}{ij, k}, \ChannelIdx{b}{ij, k} }
	= \sum_{l \in \LLRPatch(k)} \ChannelIdx{\kappa}{ij,l} \left( \ChannelIdx{C}{ij, k} \ChannelIdx{\HRSym}{i,l} + \ChannelIdx{b}{ij, k} - \ChannelIdx{\HRSym}{j,l} \right)^2 
	+ \ChannelIdx{\epsilon}{ij} \ChannelIdx{C}{ij, k}^2.
\end{equation}

Computing the zero-crossings of the derivative of this energy function \wrt the unknown regression coefficient $\ChannelIdx{b}{ij, k}$ yields:
\begin{align}
	\frac{\partial}{\partial \ChannelIdx{b}{ij, k}} \EnergyFun{ \ChannelIdx{C}{ij, k}, \ChannelIdx{b}{ij, k} }
	&= 2 \sum_{l \in \LLRPatch(k)} \ChannelIdx{\kappa}{ij,l}  \left( \ChannelIdx{C}{ij, k} \ChannelIdx{\HRSym}{i,l} + \ChannelIdx{b}{ij, k} - \ChannelIdx{\HRSym}{j,l} \right) \nonumber \\
	&= 0.
\end{align}
If we rearrange this condition, we can compute the regression coefficient $\ChannelIdx{b}{ij, k}$ in closed-form:
\begin{align}
	\ChannelIdx{b}{ij, k}
		&= 
		\frac{1}{Z_{\LLRPatch(k)} (\ChannelIdx{\vec{K}}{ij})}
			\left( 
				\sum_{l \in \LLRPatch(k)} \ChannelIdx{\kappa}{ij,l} \ChannelIdx{\HRSym}{j,l} 
				- \ChannelIdx{C}{ij, k} \sum_{l \in \LLRPatch(k)} \ChannelIdx{\kappa}{ij,l} \ChannelIdx{\HRSym}{i,l} 
			\right), \nonumber \\
		&=  \text{E}_{\LLRPatch(k)} \left( \HRChannel{j}, \ChannelIdx{\vec{K}}{ij} \right) - \ChannelIdx{C}{ij,k} \cdot \text{E}_{\LLRPatch(k)} \left( \HRChannel{i}, \ChannelIdx{\vec{K}}{ij} \right),
		\label{eqn:A_regressionCoefficientB}
\end{align}
where:
\begin{align}
	\text{E}_{\LLRPatch(k)}(\vec{z}, \vec{K}) 
		&= \frac{1}{Z_{\LLRPatch(k)} (\vec{K})} \sum_{l \in \LLRPatch(k)} \kappa_l  z_l,  \\
	Z_{\LLRPatch(k)} (\vec{K})
		&= \sum_{l \in \LLRPatch(k)} \kappa_l.
\end{align}
Notice that the estimation of this regression coefficient involves box filtering operations denoted by $\text{E}_{\LLRPatch(k)}(\cdot, \cdot)$ for the channels $\HRChannel{i}$ and $\HRChannel{j}$ with a normalization of the filter kernel according to $Z_{\LLRPatch(k)}(\vec{K})$. These box filters can be implemented efficiently using integral images, see \eg \cite{He2013,Horentrup2014}.

Computing the zero-crossings of the derivative of the energy function \wrt the regression coefficient $\ChannelIdx{C}{ij, k}$ yields:
\begin{align}
	\frac{\partial}{\partial \ChannelIdx{C}{ij, k}} \EnergyFun{ \ChannelIdx{C}{ij, k}, \ChannelIdx{b}{ij, k} }
	&= 2 \sum_{l \in \LLRPatch(k)} \ChannelIdx{\kappa}{ij,l}  \left( \ChannelIdx{C}{ij, k} \ChannelIdx{\HRSym}{i,l} + \ChannelIdx{b}{ij, k} - \ChannelIdx{\HRSym}{j,l} \right) \ChannelIdx{\HRSym}{i,l}
		+ 2 \ChannelIdx{\epsilon}{ij} \ChannelIdx{C}{ij, k} \nonumber \\
	&= 0.
\end{align}
If we substitute the expression for the regression coefficient $\ChannelIdx{b}{ij, k}$ given by \eref{eqn:A_regressionCoefficientB}, this condition can be rearranged according to:
\begin{align}
	& \ChannelIdx{C}{ij, k} \sum_{l \in \LLRPatch(k)}  \ChannelIdx{\kappa}{ij,l} \ChannelIdx{\HRSym}{i,l}^2 + \ChannelIdx{b}{ij, k} \sum_{l \in \LLRPatch(k)}  \ChannelIdx{\kappa}{ij,l} \ChannelIdx{\HRSym}{i,l}
	- \sum_{l \in \LLRPatch(k)}  \ChannelIdx{\kappa}{ij,l} \ChannelIdx{\HRSym}{i,l} \ChannelIdx{\HRSym}{j,l} + \ChannelIdx{\epsilon}{ij} \ChannelIdx{C}{ij, k} \nonumber \\
	&\quad= \ChannelIdx{C}{ij, k} \sum_{l \in \LLRPatch(k)}  \ChannelIdx{\kappa}{ij,l} \ChannelIdx{\HRSym}{i,l}^2 
	- \sum_{l \in \LLRPatch(k)}  \ChannelIdx{\kappa}{ij,l} \ChannelIdx{\HRSym}{i,l} \ChannelIdx{\HRSym}{j,l} + \ChannelIdx{\epsilon}{ij} \ChannelIdx{C}{ij, k} \nonumber \\
	&\quad\quad + \frac{1}{ Z_{\LLRPatch(k)} (\ChannelIdx{\vec{K}}{ij}) } 
			\left(
				\sum_{l \in \LLRPatch(k)} \ChannelIdx{\kappa}{ij,l} \ChannelIdx{\HRSym}{j,l}  - \ChannelIdx{C}{ij, k} \sum_{l \in \LLRPatch(k)} \ChannelIdx{\kappa}{ij,l} \ChannelIdx{\HRSym}{i,l} 
			\right) \sum_{l \in \LLRPatch(k)}  \ChannelIdx{\kappa}{ij,l} \ChannelIdx{\HRSym}{i,l} \nonumber \\
	&\quad= \ChannelIdx{C}{ij, k} 
				\left( 
					\sum_{l \in \LLRPatch(k)}  \ChannelIdx{\kappa}{ij,l} \ChannelIdx{\HRSym}{i,l}^2 
					- \frac{1}{ Z_{\LLRPatch(k)} (\ChannelIdx{\vec{K}}{ij}) }  \left( \sum_{l \in \LLRPatch(k)}  \ChannelIdx{\kappa}{ij,l} \ChannelIdx{\HRSym}{i,l} \right)^2
					+ \ChannelIdx{\epsilon}{ij}
				\right) \nonumber \\
	&\quad\quad + \frac{1}{ Z_{\LLRPatch(k)} (\ChannelIdx{\vec{K}}{ij}) }  \sum_{l \in \LLRPatch(k)}  \ChannelIdx{\kappa}{ij,l} \ChannelIdx{\HRSym}{i,l}  \sum_{l \in \LLRPatch(k)}  \ChannelIdx{\kappa}{ij,l} \ChannelIdx{\HRSym}{j,l} -  \sum_{l \in \LLRPatch(k)}  \ChannelIdx{\kappa}{ij,l} \ChannelIdx{\HRSym}{i,l} \ChannelIdx{\HRSym}{j,l} \nonumber \\
	&\quad= 0.
\end{align}
Thus, we can compute the regression coefficient $\ChannelIdx{C}{ij, k}$ in closed-form:
\begin{equation}
	\begin{split}
	\displaystyle
	\ChannelIdx{C}{ij, k} &= 
		\frac{
			\text{E}_{\LLRPatch(k)} \left( \HRChannel{i} \odot \HRChannel{j}, \ChannelIdx{\vec{K}}{ij} \right) 
			- \text{E}_{\LLRPatch(k)} \left( \HRChannel{i}, \ChannelIdx{\vec{K}}{ij} \right) 
			\cdot \text{E}_{\LLRPatch(k)} \left(\HRChannel{j}, \ChannelIdx{\vec{K}}{ij} \right)
		}
		{
			\text{E}_{\LLRPatch(k)} \left( \HRChannel{i} \odot \HRChannel{i}, \ChannelIdx{\vec{K}}{ij} \right) + \frac{1}{ Z_{\LLRPatch(k)} (\ChannelIdx{\vec{K}}{ij}) } \ChannelIdx{\epsilon}{ij}
		},
	\end{split}
\end{equation}
where we can again use box filtering of the image channels $\HRChannel{i}$ and $\HRChannel{j}$.
	\else
		\cleardoublepage  
	\fi
  
  \backmatter
	

\phantomsection

\chapter{List of Symbols}

\subsubsection{Chapter \ref{sec:MultiFrameSuperResolutionAndTheSamplingTheorem}}
\noindent
\addnotation{$x(t)$}{1-D continuous signal}{realValuedSignal}
\addnotation{$y(t)$}{Sampling of $x(t)$ as a continuous signal}{sampledSignal}
\addnotation{$y[n]$}{Discretization of the sampled signal $y(t)$}{discretizedSignal}
\addnotation{$T$}{Sampling pitch}{samplingPitch}
\addnotation{$\mathcal{D}_T\{\cdot\}$}{Sampling operator with sampling pitch $T$}{samplingOperator}
\addnotation{$\delta(t)$}{Dirac delta impulse}{diracDelta}
\addnotation{$j$}{Imaginary unit of complex number}{imagUnit}
\addnotation{$\conv$}{Convolution operator}{conv}
\addnotation{$\mathcal{F}\{\cdot\}$}{Continuous Fourier transform (CFT)}{cftOp}
\addnotation{$X(f)$}{CFT of a continuous signal $x(t)$}{cft}
\addnotation{$f_s$}{Sampling frequency (sampling rate)}{samplingRate}
\addnotation{$f_0$}{Band limitation frequency}{bandLimit}
\addnotation{$\mathcal{F}^{-1}\{\cdot\}$}{Inverse of the CFT}{cftOpInv}
\addnotation{$h(t)$}{Linear and shift invariant blur kernel}{blurKernel1D}
\addnotation{$\FrameIdx{x}{k}(t)$}{1-D continuous signal of the $k$-th channel}{continousChannel}
\addnotation{$\FrameIdx{y}{k}(t)$}{Sampling of $\FrameIdx{x}{k}(t)$ as a continuous signal}{sampledChannel}
\addnotation{$\FrameIdx{y}{k}[n]$}{Discretization the sampled signal $\FrameIdx{y}{k}(t)$}{discreteChannel}
\addnotation{$t_k$}{Offset of the $k$-th channel}{channelOffset}
\addnotation{$\mathcal{Y}[n]$}{Discrete Fourier transform (DFT) of $y[n]$}{dft}
\addnotation{$\vec{\mathcal{Y}}$}{Samples of the DFT $\mathcal{Y}[n]$ in vector notation}{dftCoeff}
\addnotation{$\vec{X}$}{Samples of the CFT $X(f)$ in vector notation}{cftCoeff}
\addnotation{$\vec{W}$}{System matrix in the Fourier domain}{systemMatrixFourier}

\subsubsection{Chapter \ref{sec:ComputationalFrameworkForMultiFrameSuperResolution}}
\noindent
\addnotation{$\Point = (\CoordU, \CoordV)^\top$}{Position in 2-D space}{point}
\addnotation{$\HRFun{\Point}$}{High-resolution image as continuous function}{hrFun}
\addnotation{$\LRFun{\Point}$}{Low-resolution image as continuous function}{hrFun}
\addnotation{$\NumFrames$}{Number of frames in a sequence}{numFrames}
\addnotation{$\FrameIdx{(\cdot)}{k}$}{Frame index}{frameIdx}
\addnotation{$\MotionOpFrame{\cdot}{k}$}{Motion model}{motionOpFrame}
\addnotation{$\FrameIdx{m}{k}(\Point)$}{Displacement vector field}{dispField}
\addnotation{$\BlurKernel{\Point}$}{Linear and shift invariant blur kernel}{blurKernel}
\addnotation{$\SamplingOp{\cdot}$}{Sampling model}{samplingOp}
\addnotation{$\Noise{\Point}$}{Stochastic noise signal}{noiseSignal}
\addnotation{$\DomainLR \subset \RealN{2}$}{Domain of low-resolution data}{domainLR}
\addnotation{$\LRDimU \times \LRDimV$}{Pixel resolution of low-res. image (width $\times$ height)}{lrDim}
\addnotation{$\LRFrame{k} \in \RealN{M}$}{Low-resolution image in vector notation}{lrFrame}
\addnotation{$\LR \in \RealN{KM}$}{Low-resolution observations (sequence of frames)}{lr}
\addnotation{$\DomainHR \subset \RealN{2}$}{Domain of high-resolution data}{domainHR}
\addnotation{$\HRDimU \times \HRDimV$}{Pixel resolution of high-res. image (width $\times$ height)}{hrDim}
\addnotation{$\HR \in \RealN{N}$}{High-resolution image in vector notation}{hrImage}
\addnotation{$\LRSize$}{Size of a low-resolution image (number of pixels)}{numLRPixels}
\addnotation{$\HRSize$}{Size of a high-resolution image (number of pixels) }{numHRPixels}
\addnotation{$\MagFac$}{Magnification factor}{magFac}
\addnotation{$\Homography$}{Homography in projective space}{homography}
\addnotation{$\TransVec$}{Translation vector}{transVec}
\addnotation{$\RotMat(\RotAngle)$}{Rotation matrix with rotation angle $\RotAngle$}{rotAngle}
\addnotation{$\SystemMatFrame{k} \in \RealMN{M}{N}$}{System matrix for the $k$-th frame}{systemMatFrame}
\addnotation{$\SystemMat \in \RealMN{KM}{N}$}{Joint system matrix}{systemMat}
\addnotation{$\NoiseVec$}{Observation noise vector}{noiseVec}
\addnotation{$\Neighbor_{\text{PSF}}(\Point)$}{Support (set of pixels) of the PSF centered at $\Point$}{neighborPSF}
\addnotation{$\PSFKernelSize$}{Size (radius) of the PSF support $\Neighbor_{\text{PSF}}(\Point)$}{psfKernelSize}
\addnotation{$\PSFWidth$}{Width of isotropic Gaussian PSF}{psfWidth}
\addnotation{$\Pdf{\HR}$}{Probability density function}{pdf}
\addnotation{$\PdfCond{\HR}{\LR}$}{Conditional probability density function}{pdfCond}
\addnotation{$\NormalDist{\cdot}{\cdot}{\cdot}$}{Multivariate normal distribution}{normalDist}
\addnotation{$\NoiseStd$}{Standard deviation of additive Gaussian noise}{noiseStd}
\addnotation{$\DataTerm{\HR}$}{Data fidelity term}{dataTerm}
\addnotation{$\ResidualFun{\HR}{\LR}$}{Residual error of $\HR$ with respect to $\LR$}{residualError}
\addnotation{$\RegTerm{\HR}$}{Regularization term}{regTerm}
\addnotation{$\RegWeight$}{Regularization weight}{regWeight}
\addnotation{$\PartFun(\cdot)$}{Partition function}{partFun}
\addnotation{$\RegTerm[Gauss]{\HR}$}{Gaussian prior regularization term}{regTermGauss}
\addnotation{$\HPKernel$}{High-pass filter kernel}{hpKernel}
\addnotation{$\HPMat$}{High-pass filter $\HPKernel$ as circulant matrix}{hpMat}
\addnotation{$\RegTerm[Huber]{\HR}$}{Huber prior regularization term}{regTermHuber}
\addnotation{$\VecEl{\vec{z}}{i}$}{$i$-th element of vector $\vec{z}$}{vecEl}
\addnotation{$\RegTerm[TV]{\HR}$}{Total variation regularization term}{regTermTV}
\addnotation{$\nabla_i \HR$}{Discrete image gradient in in direction $i \in \{u, v\}$}{imageGrad}
\addnotation{$\RegTerm[BTV]{\HR}$}{Bilateral total variation regularization term}{regTermBTV}
\addnotation{$\vec{S}_i^m$}{Shift operation by $m$ pixels in direction $i \in \{u, v\}$}{shiftMat}
\addnotation{$\BTVSize$}{Bilateral total variation window size}{btvSize}
\addnotation{$\BTVWeight$}{Bilateral total variation weighting factor}{btvWeight}

\subsubsection{Chapter \ref{sec:RobustMultiFrameSuperResolutionWithSparseRegularization}}
\noindent
\addnotation{$\WeightsB$}{Observation confidence weights}{weightsB}
\addnotation{$\PdfCond{ \LR }{ \HR , \WeightsB }$}{Weighted observation model}{weightedObservationModel}
\addnotation{$\NormalDistWeighted{\cdot}{\cdot}{\cdot}{\cdot}$}{Weighted normal distribution}{normalDistWeighted}
\addnotation{$\WeightsBMat = \Diag{\WeightsB}$}{Weights $\WeightsB$ as diagonal matrix}{weightsBMat}
\addnotation{$\SparseTransFun{\cdot}$}{Sparsifying transform}{sparseTransFun}
\addnotation{$\SparseTransMat$}{Linear sparsifying transformation matrix}{sparseTransMat}
\addnotation{$\SparseTransDomain \subset \RealN{\SparseTransDim}$}{Sparsifying transform domain}{sparseTransDomain}
\addnotation{$\SparseTransDim$}{Size of the sparsifying transform domain}{sparseTransDim}
\addnotation{$\HyperLapDist{\cdot}{\cdot}{\cdot}{\cdot}$}{Hyper-Laplacian distribution}{hyperLapDist}
\addnotation{$\WeightsA$}{Spatially adaptive prior weights}{weightsA}
\addnotation{$\PdfCond{\HR}{\WeightsA}$}{Spatially adaptive prior distribution}{adaptPrior}
\addnotation{$\LapDistWeighted{\cdot}{\cdot}{\cdot}{\cdot}$}{Weighted Laplacian distribution}{lapDistWeighted}
\addnotation{$\WeightsAMat = \Diag{\WeightsA}$}{Weights $\WeightsA$ as diagonal matrix}{weightsAMat}
\addnotation{$\WeightsBFun(\HR, \LR)$}{Observation weighting function}{weightsBFun}
\addnotation{$\WeightsAFun(\HR)$}{Prior weighting function}{weightsAFun}
\addnotation{$\SparsityParam \in [0, 1]$}{Sparsity parameter}{sparsityParam}
\addnotation{$\WMedian{\Residual}{\WeightsB}$}{Weighted median of $\Residual$ under the weights $\WeightsB$}{wMedian}
\addnotation{$\WMad{\Residual}{\WeightsB}$}{Weighted median absolute deviation}{wMad}
\addnotation{$\Id_{\delta}$}{Random binary diagonal matrix (training set)}{trainingObservations}
\addnotation{$\overline{\Id_{\delta}}$}{Element-wise flipping of $\Id_{\delta}$ (validation set)}{validationObservations}
\addnotation{$\DataTerm[\text{cv}]{\RegWeight, \overline{\Id_{\delta}} }$}{Cross validation error for regularization weight $\RegWeight$}{cvError}
\addnotation{$[\log \RegWeight_l, \log \RegWeight_u]$}{Cross validation search range (log-transformed)}{cvSearchRange}
\addnotation{$\NumIter[cv]$}{Number of cross validation iterations}{cvNumIter}
\addnotation{$\IterationIdx{\EnergyFunSym}{t}(\HR)$}{Energy function at iteration $t$}{energyFunIter}
\addnotation{$\LossFun[\text{Char}]{\vec{z}}$}{Charbonnier loss function}{charbonnierLoss}
\addnotation{$\NumIter[irwsr]$}{Number of re-weighted minimization iterations}{irwsrNumIter}
\addnotation{$\NumIter[scg]$}{Number of SCG iterations}{scgNumIter}
\addnotation{$\TermTolerance$}{Termination tolerance}{irwsrTerminationTol}
\addnotation{$\EnergyFunMajorize{\HR, \IterationIdx{\HR}{t-1}}$}{Majorizing function for $\EnergyFun{\HR}$ at $\IterationIdx{\HR}{t-1}$}{majorizingFun}
\addnotation{$\LossFun[\SparsityParam]{z}$}{Mixed \LOne/\Lp{\SparsityParam} norm}{mixedL1Lp}

\subsubsection{Chapter \ref{sec:MultiSensorSuperResolutionUsingGuidanceImages}}
\noindent
\addnotation{$\GuideFrame{k}$}{Guidance image}{guideFrame}
\addnotation{$\GuideDimU \times \GuideDimV$}{Pixel resolution of guidance image (width $\times$ height)}{guideDim}
\addnotation{$\GuideSize$}{Size of a guidance image (number of pixels)}{guideSize}
\addnotation{$\SenFusMapping{\Point_{\GuideSym}}$}{Mapping from guidance image to input image}{senFusMapping}
\addnotation{$\Point_{\GuideSym}$}{Pixel position in guidance image}{pointGuide}
\addnotation{$\Point_{\LRSym}$}{Pixel position in input image}{pointLr}
\addnotation{$\DomainGuide$}{Domain of guidance images}{domainGuide}
\addnotation{$\DomainLR$}{Domain of input images}{domainInput}
\addnotation{$\SenFusNeighbor(\Point_{\GuideSym})$}{Support (set of pixels) for sensor data fusion}{senFusNeighbor}
\addnotation{$\DataTerm[MSR]{\HR, \Guide}$}{Data fidelity term for multi-sensor super-resolution}{msrDataFidelity}
\addnotation{$\RegTerm[MSR]{\HR, \Guide}$}{Regularization term for multi-sensor super-resolution}{msrRegTerm}
\addnotation{$\DispVecField[\LRSym]{\Point_{\LRSym}}$}{Displacement vector field on input image}{dispVecFieldInput}
\addnotation{$\DispVecField[\GuideSym]{\Point_{\GuideSym}}$}{Displacement vector field on guidance image}{dispVecFieldGuide}
\addnotation{$\MotionResamplingOp{\cdot}$}{Displacement vector field filter (resampling) operator}{motionResamplingOp}
\addnotation{$\LossFun[\text{MSR}]{\cdot}$}{Loss function for spatially adaptive regularization}{msrRegLossFun}
\addnotation{$\WeightsAFun(\HR, \Guide)$}{Spatially adaptive regularization weights (in $\DomainLR$)}{amsrWeights}
\addnotation{$\tilde{\WeightsAFun}( \tilde{\HR}, \Guide)$}{Spatially adaptive regularization weights (in $\DomainGuide$)}{amsrWeightsGuide}
\addnotation{$\tau(\Point) \in \{0, 1\}$}{Binary edge map}{edgeMap}
\addnotation{$\AMSRNeighbor(\Point)$}{Image patch for spatially adaptive regularization}{amsrNeighbor}
\addnotation{$\AMSRPatchSize$}{Size of image patch $\AMSRNeighbor(\Point)$}{amsrPatchSize}
\addnotation{$\AMSRContrastFactor$}{Contrast factor for spatially adaptive regularization}{amsrContrastFactor}
\addnotation{$\SimMeasureLocal{\cdot}{\cdot}{\cdot}$}{Local image similarity measure}{simMeasureLocal}
\addnotation{$\MiMeasureLocal{\cdot}{\cdot}{\cdot}$}{Local mutual information}{miMeasureLocal}
\addnotation{$\NccMeasureLocal{\cdot}{\cdot}{\cdot}$}{Local normalized cross correlation}{nccMeasureLocal}
\addnotation{$\SimMeasureSym_{0} \in [-1, +1]$}{Normalized cross correlation outlier threshold}{nccThreshold}
\addnotation{$\WeightsBFun_{\GuideSym, i} ( \GuideFrame{k} )$}{Confidence weight for $i$-th pixel in guidance image $\GuideFrame{k}$}{confWeightGuidePixel}
\addnotation{$\WeightsBFun_{\GuideSym} (\GuideFrame{k} )$}{Confidence map for guidance image $\GuideFrame{k}$}{confWeightGuideFrame}
\addnotation{$\WeightsBFun_{\GuideSym} (\Guide )$}{Confidence map for guidance data $\Guide$}{confWeightGuide}
\addnotation{$\WeightsBFun_{\LRSym,i} (\HR, \LR)$}{Confidence weight for $i$-th pixel in $\LR$}{confWeightInputPixel}
\addnotation{$\WeightsBFun_{\LRSym} (\HR, \LR)$}{Confidence map for input data $\LR$}{confWeightInput}
\addnotation{$\IterationIdx{\WeightsBMat}{t}$}{Joint confidence map for input and guidance data}{jointConfidence}

\subsubsection{Chapter \ref{sec:SuperResolutionForMultiChannelImages}}
\noindent
\addnotation{$\HR$}{High-resolution multi-channel image}{hrChannelComposite}
\addnotation{$\HRChannel{i}$}{High-resolution channel ($i$-th channel)}{hrChannel}
\addnotation{$\ChannelIdx{\HRSize}{i}$}{Size of the $i$-th high-resolution channel}{hrChannelSize}
\addnotation{$\NumChannels$}{Number of channels}{numChannels}
\addnotation{$\LRChannel{i}$}{Low-resolution channel ($i$-th channel)}{lrChannel}
\addnotation{$\ChannelIdx{\LRSize}{i}$}{Size of the $i$-th low-resolution channel}{lrChannelSize}
\addnotation{$\LR$}{Set of all low-resolution channels}{lrChannelComposite}
\addnotation{$\SystemMatChannel{i}$}{System matrix of the $i$-th channel}{systemMatChannel}
\addnotation{$\RegTerm[intra]{\HRChannel{i}}$}{Intra-channel regularization term for $\HRChannel{i}$}{intraChannelRegTerm}
\addnotation{$\ChannelIdx{\RegWeight}{i}$}{Intra-channel regularization weight for $\HRChannel{i}$}{intraChannelRegWeight}
\addnotation{$\RegTerm[inter]{\HRChannel{i}, \HRChannel{j}, \ChannelIdx{\LLRParams}{ij}}$}{Inter-channel regularization term for for $\HRChannel{i}$ and $\HRChannel{j}$}{interChannelRegTerm}
\addnotation{$\ChannelIdx{\mu}{ij}$}{Inter-channel regularization weight for $\HRChannel{i}$ and $\HRChannel{j}$}{interChannelRegWeight}
\addnotation{$\ChannelIdx{\LLRParams}{ij}$}{Inter-channel prior hyperparameters for $\HRChannel{i}$ and $\HRChannel{j}$}{llrParams}
\addnotation{$\LLRPatch(n)$}{Quadratic image patch for locally linear regression}{llrPatch}
\addnotation{$\LLRPatchSize$}{Size (edge length) of $\LLRPatch(n)$}{llrPatchSize}
\addnotation{$\ChannelIdx{C}{ij,n}$}{Pixel-wise multiplicative regression coefficients}{llrCoeffsMultLocal}
\addnotation{$\ChannelIdx{b}{ij,n}$}{Pixel-wise additive regression coefficients}{llrCoeffsAddLocal}
\addnotation{$\ChannelIdx{\vec{C}}{ij}$}{Channel-wise multiplicative regression coefficients}{llrCoeffsMultGlobal}
\addnotation{$\ChannelIdx{\vec{b}}{ij}$}{Channel-wise additive regression coefficients}{llrCoeffsAddGlobal}

\subsubsection{Chapter \ref{sec:RetinalFundusVideoImaging}}
\noindent
\addnotation{$\MotionParamsSingle{i}$}{Eye motion parameter}{motionParamsSingle}
\addnotation{$\MotionParams = \{\MotionParamsSingle{1}, \ldots, \MotionParamsSingle{n} \}$}{Set of eye motion parameters ($n$ degrees of freedom)}{motionParams}
\addnotation{$\BiasFieldMult \in \RealN{M}$}{Multiplicative photometric parameters (bias field)}{biasFieldMult}
\addnotation{$\BiasFieldAdd \in \Real$}{Additive photometric parameters (brightness offset)}{biasFieldAdd}
\addnotation{$\PhotometricParams = \{\BiasFieldMult, \BiasFieldAdd \}$}{Set of photometric parameters}{photometricParams}
\addnotation{$\vec{p}$}{$N_p \times N_p$ image patch}{imagePatch}
\addnotation{$\vec{G}(\vec{p})$}{Gradient matrix for image patch $\vec{p}$}{gradientMatrix}
\addnotation{$s_1(\vec{p})$, $s_2(\vec{p})$}{Singular values of the gradient matrix $G(\vec{p})$}{gradientMatrixSingularValues}
\addnotation{$\sigma_j$}{Laplacian of Gaussian kernel size}{logKernelStd}
\addnotation{$\vec{H}_i (\sigma_j)$}{Hessian for the $i$-th pixel with kernel size $\sigma_j$}{hessian}
\addnotation{$\lambda_{1,i} (\sigma_j)$, $\lambda_{2,i} (\sigma_j)$}{Eigenvalues of the Hessian $\vec{H}_i (\sigma_j)$}{hessianEigenvalues}
\addnotation{$V_i (\sigma_j)$}{Vesselness for the $i$-th pixel with kernel size $\sigma_j$}{vesselnessSingleScale}
\addnotation{$V_i^*$}{Vesselness for the $i$-th pixel}{vesselness}
\addnotation{$V(\vec{p})$}{Vesselness variance for image patch $\vec{p}$}{vesselnessVariance}
\addnotation{$c(\vec{p})$}{Coherence for image patch $\vec{p}$}{coherence}
\addnotation{$q(\vec{p})$}{Local quality measure for image patch $\vec{p}$}{localQualityMeasure}
\addnotation{$Q(\HR)$}{Global quality measure for image $\HR$}{globalQualityMeasure}
\addnotation{$\mathcal{A}(\HR)$}{Set of anisotropic patches for image $\HR$}{anisoSet}
\addnotation{$\alpha_c$}{Anisotropic patch significance level}{anisoSetLevel}

\subsubsection{Chapter \ref{sec:HybridRangeImagingForImageGuidedSurgery}}
\noindent
\addnotation{$\vec{H}_{\LRSym \GuideSym}$}{Homography from range image to color image domain}{sensorFusionHomography}
\addnotation{$\vec{P}_{\LRSym}$}{Range camera projection matrix}{rangeCameraProjMatrix}
\addnotation{$\vec{P}_{\GuideSym}$}{Color camera projection matrix}{colorCameraProjMatrix}
\addnotation{$\vec{K}_{\LRSym}$}{Intrinsic range camera calibration matrix}{rangeCameraCalibMatrix}
\addnotation{$\vec{K}_{\GuideSym}$}{Intrinsic color camera calibration matrix}{colorCameraCalibMatrix}
\addnotation{$\RotMat$}{Extrinsic rotation matrix of the color camera}{extrinsicRotation}
\addnotation{$\TransVec$}{Extrinsic translation vector of the color camera}{extrinsicTranslation}
\addnotation{$Q_{\text{snr}}$}{Blind signal-to-noise ratio measure}{blindSNR}
\addnotation{$Q_{\text{edge}}$}{Blind edge reconstruction measure}{edgeRecoMeasure}

\cleardoublepage

	\printglossary[type=\acronymtype,style=list,title=List of Abbreviations,toctitle=List of Abbreviations]
	\glsresetall

\phantomsection
\addcontentsline{toc}{chapter}{\listfigurename}
\listoffigures
\cleardoublepage


\phantomsection
\addcontentsline{toc}{chapter}{\listtablename}
\listoftables
\cleardoublepage
	

%

\bibliographystyle{wmaainf}
\phantomsection
\addcontentsline{toc}{chapter}{\bibname}
\renewcommand\refname{Bibliography}
\baselineskip = 12pt
\bibliography{library}
\baselineskip = 18pt
\cleardoublepage


\end{document}